\def\sphinxdocclass{jupyterBook}
\documentclass[letterpaper,10pt,english]{jupyterBook}
\ifdefined\pdfpxdimen
   \let\sphinxpxdimen\pdfpxdimen\else\newdimen\sphinxpxdimen
\fi \sphinxpxdimen=.75bp\relax
\PassOptionsToPackage{hyperindex=false}{hyperref}

\PassOptionsToPackage{warn}{textcomp}

\catcode`^^^^00a0\active\protected\def^^^^00a0{\leavevmode\nobreak\ }
\usepackage{cmap}
\usepackage{fontspec}
\defaultfontfeatures[\rmfamily,\sffamily,\ttfamily]{}
\usepackage{amsmath,amssymb,amstext}
\usepackage{polyglossia}
\setmainlanguage{english}

\usepackage[Bjarne]{fncychap}
\usepackage[,numfigreset=0,mathnumfig]{sphinx}

\fvset{fontsize=\small}
\usepackage{geometry}

\usepackage{hyperref}
\usepackage{hypcap}% it must be loaded after hyperref.
\addto\captionsenglish{}

\usepackage{sphinxmessages}

         \usepackage[Latin,Greek]{ucharclasses}
        \usepackage{unicode-math}
        \addto\captionsenglish{}

\title{Physics-based Deep Learning}
\date{Sep 11, 2021}
\release{}
\author{N.\@{} Thuerey, P.\@{} Holl, M.\@{} Mueller, P.\@{} Schnell, F.\@{} Trost, K.\@{} Um}
\newcommand{\sphinxlogo}{\vbox{}}
\renewcommand{\releasename}{}
\makeindex
\begin{document}

\pagestyle{empty}
\sphinxmaketitle
\pagestyle{plain}
\sphinxtableofcontents
\pagestyle{normal}
\phantomsection\label{\detokenize{intro::doc}}

\begin{figure}[htbp]
\centering

\noindent\sphinxincludegraphics{{logo-xl}.jpg}
\end{figure}

Welcome to the \sphinxstyleemphasis{Physics\sphinxhyphen{}based Deep Learning Book} (v0.1) 👋

\sphinxstylestrong{TL;DR}:
This document contains a practical and comprehensive introduction of everything
related to deep learning in the context of physical simulations.
As much as possible, all topics come with hands\sphinxhyphen{}on code examples in the
form of Jupyter notebooks to quickly get started.
Beyond standard \sphinxstyleemphasis{supervised} learning from data, we’ll look at \sphinxstyleemphasis{physical loss} constraints,
more tightly coupled learning algorithms with \sphinxstyleemphasis{differentiable simulations}, as well as
reinforcement learning and uncertainty modeling.
We live in exciting times: these methods have a huge potential to fundamentally
change what computer simulations can achieve.

\bigskip\hrule\bigskip

\begin{DUlineblock}{0em}
\item[] \sphinxstylestrong{\Large Coming up}
\end{DUlineblock}

As a \sphinxstyleemphasis{sneak preview}, the next chapters will show:
\begin{itemize}
\item {} 
How to train networks to infer a fluid flow around shapes like airfoils, and estimate the uncertainty of the prediction. This gives a \sphinxstyleemphasis{surrogate model} that replaces a traditional numerical simulation.

\item {} 
How to use model equations as residuals to train networks that represent solutions, and how to improve upon these residual constraints by using \sphinxstyleemphasis{differentiable simulations}.

\item {} 
How to more tightly interact with a full simulator for \sphinxstyleemphasis{inverse problems}. E.g., we’ll demonstrate how to circumvent the convergence problems of standard reinforcement learning techniques by leveraging simulators in the training loop.

\end{itemize}

Throughout this text,
we will introduce different approaches for introducing physical models
into deep learning, i.e., \sphinxstyleemphasis{physics\sphinxhyphen{}based deep learning} (PBDL) approaches.
These algorithmic variants will be introduced in order of increasing
tightness of the integration, and the pros and cons of the different approaches
will be discussed. It’s important to know in which scenarios each of the
different techniques is particularly useful.

\begin{sphinxadmonition}{note}{Executable code, right here, right now}

We focus on Jupyter notebooks, a key advantage of which is that all code examples
can be executed \sphinxstyleemphasis{on the spot}, from your browser. You can modify things and
immediately see what happens – give it a try…

Plus, Jupyter notebooks are great because they’re a form of \sphinxhref{https://en.wikipedia.org/wiki/Literate\_programming}{literate programming}.
\end{sphinxadmonition}

\begin{DUlineblock}{0em}
\item[] \sphinxstylestrong{\large Comments and suggestions}
\end{DUlineblock}

This \sphinxstyleemphasis{book}, where “book” stands for a collection of digital texts and code examples,
is maintained by the
\sphinxhref{https://ge.in.tum.de}{TUM Physics\sphinxhyphen{}based Simulation Group}. Feel free to contact us
if you have any comments, e.g., via \sphinxhref{mailto:i15ge@cs.tum.edu}{old fashioned email}.
If you find mistakes, please also let us know! We’re aware that this document is far from perfect,
and we’re eager to improve it. Thanks in advance 😀! Btw., we also maintain a \sphinxhref{https://github.com/thunil/Physics-Based-Deep-Learning}{link collection} with recent research papers.

\begin{figure}[htbp]
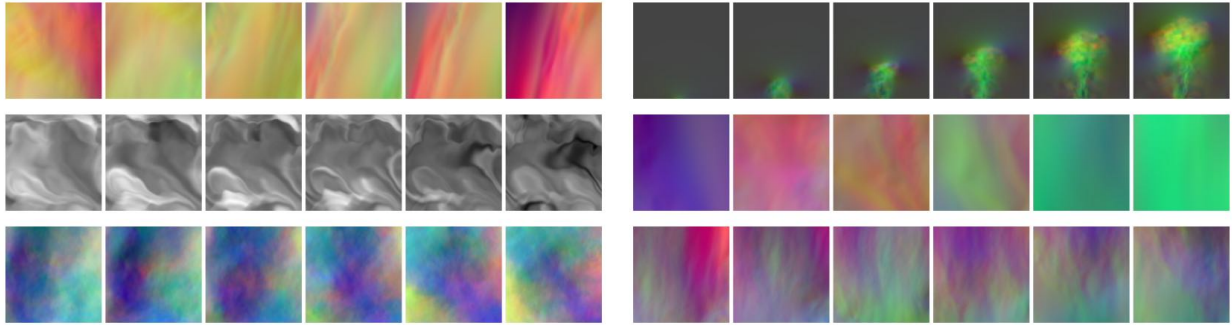

\centering
\capstart

\noindent\sphinxincludegraphics[height=220\sphinxpxdimen]{{divider-mult}.jpg}
\caption{Some visual examples of numerically simulated time sequences. In this book, we explain how to realize algorithms that use neural networks alongside numerical solvers.}\label{\detokenize{intro:divider-mult}}\end{figure}

\begin{DUlineblock}{0em}
\item[] \sphinxstylestrong{\large Thanks!}
\end{DUlineblock}

This project would not have been possible without the help of many people who contributed. Thanks to everyone 🙏 Here’s an alphabetical list:
\begin{itemize}
\item {} 
\sphinxhref{https://ge.in.tum.de/about/philipp-holl/}{Philipp Holl}

\item {} 
\sphinxhref{https://ge.in.tum.de/}{Maximilian Mueller}

\item {} 
\sphinxhref{https://ge.in.tum.de/about/patrick-schnell/}{Patrick Schnell}

\item {} 
\sphinxhref{https://ge.in.tum.de/}{Felix Trost}

\item {} 
\sphinxhref{https://ge.in.tum.de/about/n-thuerey/}{Nils Thuerey}

\item {} 
\sphinxhref{https://ge.in.tum.de/about/kiwon/}{Kiwon Um}

\end{itemize}

Additional thanks go to
Georg Kohl for the nice divider images (cf. {[}\hyperlink{cite.references:id7}{KUT20}{]}),
Li\sphinxhyphen{}Wei Chen for the airfoil data image,
and to
Chloe Paillard for proofreading parts of the document.

\begin{DUlineblock}{0em}
\item[] \sphinxstylestrong{\large Citation}
\end{DUlineblock}

If you find this book useful, please cite it via:

\begin{sphinxVerbatim}[commandchars=\\\{\}]
\PYG{n+nd}{@book}\PYG{p}{\PYGZob{}}\PYG{n}{thuerey2021pbdl}\PYG{p}{,}
  \PYG{n}{title}\PYG{o}{=}\PYG{p}{\PYGZob{}}\PYG{n}{Physics}\PYG{o}{\PYGZhy{}}\PYG{n}{based} \PYG{n}{Deep} \PYG{n}{Learning}\PYG{p}{\PYGZcb{}}\PYG{p}{,}
  \PYG{n}{author}\PYG{o}{=}\PYG{p}{\PYGZob{}}\PYG{n}{Nils} \PYG{n}{Thuerey} \PYG{o+ow}{and} \PYG{n}{Philipp} \PYG{n}{Holl} \PYG{o+ow}{and} \PYG{n}{Maximilian} \PYG{n}{Mueller} \PYG{o+ow}{and} \PYG{n}{Patrick} \PYG{n}{Schnell} \PYG{o+ow}{and} \PYG{n}{Felix} \PYG{n}{Trost} \PYG{o+ow}{and} \PYG{n}{Kiwon} \PYG{n}{Um}\PYG{p}{\PYGZcb{}}\PYG{p}{,}
  \PYG{n}{url}\PYG{o}{=}\PYG{p}{\PYGZob{}}\PYG{n}{https}\PYG{p}{:}\PYG{o}{/}\PYG{o}{/}\PYG{n}{physicsbaseddeeplearning}\PYG{o}{.}\PYG{n}{org}\PYG{p}{\PYGZcb{}}\PYG{p}{,}
  \PYG{n}{year}\PYG{o}{=}\PYG{p}{\PYGZob{}}\PYG{l+m+mi}{2021}\PYG{p}{\PYGZcb{}}\PYG{p}{,}
  \PYG{n}{publisher}\PYG{o}{=}\PYG{p}{\PYGZob{}}\PYG{n}{WWW}\PYG{p}{\PYGZcb{}}
\PYG{p}{\PYGZcb{}}
\end{sphinxVerbatim}

\part{Introduction}

\chapter{A Teaser Example}
\label{\detokenize{intro-teaser:a-teaser-example}}\label{\detokenize{intro-teaser::doc}}
Let’s start with a very reduced example that highlights some of the key capabilities of physics\sphinxhyphen{}based learning approaches. Let’s assume our physical model is a very simple equation: a parabola along the positive x\sphinxhyphen{}axis.

Despite being very simple, for every point along there are two solutions, i.e. we have two modes, one above the other one below the x\sphinxhyphen{}axis, as shown on the left below. If we don’t take care a conventional learning approach will give us an approximation like the red one shown in the middle, which is completely off. With an improved learning setup, ideally, by using a discretized numerical solver, we can at least accurately represent one of the modes of the solution (shown in green on the right).

\begin{figure}[htbp]
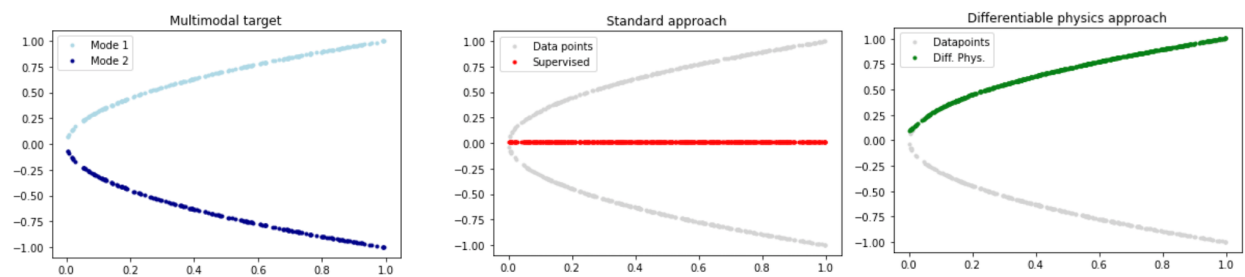

\centering
\capstart

\noindent\sphinxincludegraphics[height=180\sphinxpxdimen]{{intro-teaser-side-by-side}.png}
\caption{Side by side \sphinxhyphen{} supervised versus differentiable physics training.}\label{\detokenize{intro-teaser:intro-teaser-side-by-side}}\end{figure}

\section{Differentiable physics}
\label{\detokenize{intro-teaser:differentiable-physics}}
One of the key concepts of the following chapters is what we’ll call \sphinxstyleemphasis{differentiable physics} (DP). This means that we use domain knowledge in the form of model equations, and then integrate discretized versions of these models into the training process. As implied by the name, having differentiable formulations is crucial for this process to support the training of neural networks.

Let’s illustrate the properties of deep learning via DP with the following example: We’d like to find an unknown function \(f^*\) that generates solutions from a space \(Y\), taking inputs from \(X\), i.e. \(f^*: X \to Y\). In the following, we’ll often denote \sphinxstyleemphasis{idealized}, and unknown functions with a \(*\) superscript, in contrast to their discretized, realizable counterparts without this superscript.

Let’s additionally assume we have a generic differential equation \(\mathcal P^*: Y \to Z\) (our \sphinxstyleemphasis{model} equation), that encodes a property of the solutions, e.g. some real world behavior we’d like to match. Later on, \(P^*\) will often represent time evolutions, but it could also be a constraint for conservation of mass (then \(\mathcal P^*\) would measure divergence). But to keep things as simple as possible here, the model we’ll look at in the following is a mapping back to the input space \(X\), i.e. \(\mathcal P^*: Y \to X\).

Using a neural network \(f\) to learn the unknown and ideal function \(f^*\), we could turn to classic \sphinxstyleemphasis{supervised} training to obtain \(f\) by collecting data. This classical setup requires a dataset by sampling \(x\) from \(X\) and adding the corresponding solutions \(y\) from \(Y\). We could obtain these, e.g., by classical numerical techniques. Then we train the NN \(f\) in the usual way using this dataset.

In contrast to this supervised approach, employing a differentiable physics approach takes advantage of the fact that we can often use a discretized version of the physical model \(\mathcal P\) and employ it to guide the training of \(f\). I.e., we want \(f\) to be aware of our \sphinxstyleemphasis{simulator} \(\mathcal P\), and to \sphinxstyleemphasis{interact} with it. This can vastly improve the learning, as we’ll illustrate below with a very simple example (more complex ones will follow later on).

Note that in order for the DP approach to work, \(\mathcal P\) has to be differentiable, as implied by the name. These differentials, in the form of a gradient, are what’s driving the learning process.

\section{Finding the inverse function of a parabola}
\label{\detokenize{intro-teaser:finding-the-inverse-function-of-a-parabola}}
To illustrate the difference of supervised and DP approaches, we consider the following simplified setting: Given the function \(\mathcal P: y\to y^2\) for \(y\) in the interval \([0,1]\), find the unknown function \(f\) such that \(\mathcal P(f(x)) = x\) for all \(x\) in \([0,1]\). Note: to make things a bit more interesting, we’re using \(y^2\) here for \(\mathcal P\) instead of the more common \(x^2\) parabola, and the \sphinxstyleemphasis{discretization} is simply given by representing the \(x\) and \(y\) via floating point numbers in the computer for this simple case.

We know that possible solutions for \(f\) are the positive or negative square root function (for completeness: piecewise combinations would also be possible).
Knowing that this is not overly difficult, a solution that suggests itself is to train a neural network to approximate this inverse mapping \(f\).
Doing this in the “classical” supervised manner, i.e. purely based on data, is an obvious starting point. After all, this approach was shown to be a powerful tool for a variety of other applications, e.g., in computer vision.

\begin{sphinxVerbatim}[commandchars=\\\{\}]
\PYG{k+kn}{import} \PYG{n+nn}{numpy} \PYG{k}{as} \PYG{n+nn}{np}
\PYG{k+kn}{import} \PYG{n+nn}{tensorflow} \PYG{k}{as} \PYG{n+nn}{tf}
\PYG{k+kn}{import} \PYG{n+nn}{matplotlib}\PYG{n+nn}{.}\PYG{n+nn}{pyplot} \PYG{k}{as} \PYG{n+nn}{plt}
\end{sphinxVerbatim}

For supervised training, we can employ our solver \(\mathcal P\) for the problem to pre\sphinxhyphen{}compute the solutions we need for training: We randomly choose between the positive and the negative square root. This resembles the  general case, where we would gather all data available to us (e.g., using optimization techniques to compute the solutions). Such data collection typically does not favor one particular mode from multimodal solutions.

\begin{sphinxVerbatim}[commandchars=\\\{\}]
\PYG{c+c1}{\PYGZsh{} X\PYGZhy{}Data}
\PYG{n}{N} \PYG{o}{=} \PYG{l+m+mi}{200}
\PYG{n}{X} \PYG{o}{=} \PYG{n}{np}\PYG{o}{.}\PYG{n}{random}\PYG{o}{.}\PYG{n}{random}\PYG{p}{(}\PYG{n}{N}\PYG{p}{)}
\end{sphinxVerbatim}

\begin{sphinxVerbatim}[commandchars=\\\{\}]
\PYG{c+c1}{\PYGZsh{} Generation Y\PYGZhy{}Data}
\PYG{n}{sign} \PYG{o}{=} \PYG{p}{(}\PYG{o}{\PYGZhy{}} \PYG{n}{np}\PYG{o}{.}\PYG{n}{ones}\PYG{p}{(}\PYG{p}{(}\PYG{n}{N}\PYG{p}{,}\PYG{p}{)}\PYG{p}{)}\PYG{p}{)}\PYG{o}{*}\PYG{o}{*}\PYG{n}{np}\PYG{o}{.}\PYG{n}{random}\PYG{o}{.}\PYG{n}{randint}\PYG{p}{(}\PYG{l+m+mi}{2}\PYG{p}{,}\PYG{n}{size}\PYG{o}{=}\PYG{n}{N}\PYG{p}{)}
\PYG{n}{Y} \PYG{o}{=} \PYG{n}{np}\PYG{o}{.}\PYG{n}{sqrt}\PYG{p}{(}\PYG{n}{X}\PYG{p}{)} \PYG{o}{*} \PYG{n}{sign}
\end{sphinxVerbatim}

Now we can define a network, the loss, and the training configuration. We’ll use a simple \sphinxcode{\sphinxupquote{keras}} architecture with three hidden layers, ReLU activations.

\begin{sphinxVerbatim}[commandchars=\\\{\}]
\PYG{c+c1}{\PYGZsh{} Neural network}
\PYG{n}{act} \PYG{o}{=} \PYG{n}{tf}\PYG{o}{.}\PYG{n}{keras}\PYG{o}{.}\PYG{n}{layers}\PYG{o}{.}\PYG{n}{ReLU}\PYG{p}{(}\PYG{p}{)}
\PYG{n}{nn\PYGZus{}sv} \PYG{o}{=} \PYG{n}{tf}\PYG{o}{.}\PYG{n}{keras}\PYG{o}{.}\PYG{n}{models}\PYG{o}{.}\PYG{n}{Sequential}\PYG{p}{(}\PYG{p}{[}
  \PYG{n}{tf}\PYG{o}{.}\PYG{n}{keras}\PYG{o}{.}\PYG{n}{layers}\PYG{o}{.}\PYG{n}{Dense}\PYG{p}{(}\PYG{l+m+mi}{10}\PYG{p}{,} \PYG{n}{activation}\PYG{o}{=}\PYG{n}{act}\PYG{p}{)}\PYG{p}{,}
  \PYG{n}{tf}\PYG{o}{.}\PYG{n}{keras}\PYG{o}{.}\PYG{n}{layers}\PYG{o}{.}\PYG{n}{Dense}\PYG{p}{(}\PYG{l+m+mi}{10}\PYG{p}{,} \PYG{n}{activation}\PYG{o}{=}\PYG{n}{act}\PYG{p}{)}\PYG{p}{,}
  \PYG{n}{tf}\PYG{o}{.}\PYG{n}{keras}\PYG{o}{.}\PYG{n}{layers}\PYG{o}{.}\PYG{n}{Dense}\PYG{p}{(}\PYG{l+m+mi}{1}\PYG{p}{,}\PYG{n}{activation}\PYG{o}{=}\PYG{l+s+s1}{\PYGZsq{}}\PYG{l+s+s1}{linear}\PYG{l+s+s1}{\PYGZsq{}}\PYG{p}{)}\PYG{p}{]}\PYG{p}{)}
\end{sphinxVerbatim}

And we can start training via a simple mean squared error loss, using \sphinxcode{\sphinxupquote{fit}} function from keras:

\begin{sphinxVerbatim}[commandchars=\\\{\}]
\PYG{c+c1}{\PYGZsh{} Loss function}
\PYG{n}{loss\PYGZus{}sv} \PYG{o}{=} \PYG{n}{tf}\PYG{o}{.}\PYG{n}{keras}\PYG{o}{.}\PYG{n}{losses}\PYG{o}{.}\PYG{n}{MeanSquaredError}\PYG{p}{(}\PYG{p}{)}
\PYG{n}{optimizer\PYGZus{}sv} \PYG{o}{=} \PYG{n}{tf}\PYG{o}{.}\PYG{n}{keras}\PYG{o}{.}\PYG{n}{optimizers}\PYG{o}{.}\PYG{n}{Adam}\PYG{p}{(}\PYG{n}{lr}\PYG{o}{=}\PYG{l+m+mf}{0.001}\PYG{p}{)}
\PYG{n}{nn\PYGZus{}sv}\PYG{o}{.}\PYG{n}{compile}\PYG{p}{(}\PYG{n}{optimizer}\PYG{o}{=}\PYG{n}{optimizer\PYGZus{}sv}\PYG{p}{,} \PYG{n}{loss}\PYG{o}{=}\PYG{n}{loss\PYGZus{}sv}\PYG{p}{)}

\PYG{c+c1}{\PYGZsh{} Training}
\PYG{n}{results\PYGZus{}sv} \PYG{o}{=} \PYG{n}{nn\PYGZus{}sv}\PYG{o}{.}\PYG{n}{fit}\PYG{p}{(}\PYG{n}{X}\PYG{p}{,} \PYG{n}{Y}\PYG{p}{,} \PYG{n}{epochs}\PYG{o}{=}\PYG{l+m+mi}{5}\PYG{p}{,} \PYG{n}{batch\PYGZus{}size}\PYG{o}{=} \PYG{l+m+mi}{5}\PYG{p}{,} \PYG{n}{verbose}\PYG{o}{=}\PYG{l+m+mi}{1}\PYG{p}{)}
\end{sphinxVerbatim}

\begin{sphinxVerbatim}[commandchars=\\\{\}]
Epoch 1/5
40/40 [==============================] \PYGZhy{} 0s 1ms/step \PYGZhy{} loss: 0.5084
Epoch 2/5
40/40 [==============================] \PYGZhy{} 0s 1ms/step \PYGZhy{} loss: 0.5022
Epoch 3/5
40/40 [==============================] \PYGZhy{} 0s 1ms/step \PYGZhy{} loss: 0.5011
Epoch 4/5
40/40 [==============================] \PYGZhy{} 0s 1ms/step \PYGZhy{} loss: 0.5002
Epoch 5/5
40/40 [==============================] \PYGZhy{} 0s 1ms/step \PYGZhy{} loss: 0.5007
\end{sphinxVerbatim}

As both NN and the data set are very small, the training converges very quickly. However, if we inspect the predictions of the network, we can see that it is nowhere near the solution we were hoping to find: it averages between the data points on both sides of the x\sphinxhyphen{}axis and therefore fails to find satisfying solutions to the problem above.

The following plot nicely highlights this: it shows the data in light gray, and the supervised solution in red.

\begin{sphinxVerbatim}[commandchars=\\\{\}]
\PYG{c+c1}{\PYGZsh{} Results}
\PYG{n}{plt}\PYG{o}{.}\PYG{n}{plot}\PYG{p}{(}\PYG{n}{X}\PYG{p}{,}\PYG{n}{Y}\PYG{p}{,}\PYG{l+s+s1}{\PYGZsq{}}\PYG{l+s+s1}{.}\PYG{l+s+s1}{\PYGZsq{}}\PYG{p}{,}\PYG{n}{label}\PYG{o}{=}\PYG{l+s+s1}{\PYGZsq{}}\PYG{l+s+s1}{Data points}\PYG{l+s+s1}{\PYGZsq{}}\PYG{p}{,} \PYG{n}{color}\PYG{o}{=}\PYG{l+s+s2}{\PYGZdq{}}\PYG{l+s+s2}{lightgray}\PYG{l+s+s2}{\PYGZdq{}}\PYG{p}{)}
\PYG{n}{plt}\PYG{o}{.}\PYG{n}{plot}\PYG{p}{(}\PYG{n}{X}\PYG{p}{,}\PYG{n}{nn\PYGZus{}sv}\PYG{o}{.}\PYG{n}{predict}\PYG{p}{(}\PYG{n}{X}\PYG{p}{)}\PYG{p}{,}\PYG{l+s+s1}{\PYGZsq{}}\PYG{l+s+s1}{.}\PYG{l+s+s1}{\PYGZsq{}}\PYG{p}{,}\PYG{n}{label}\PYG{o}{=}\PYG{l+s+s1}{\PYGZsq{}}\PYG{l+s+s1}{Supervised}\PYG{l+s+s1}{\PYGZsq{}}\PYG{p}{,} \PYG{n}{color}\PYG{o}{=}\PYG{l+s+s2}{\PYGZdq{}}\PYG{l+s+s2}{red}\PYG{l+s+s2}{\PYGZdq{}}\PYG{p}{)}
\PYG{n}{plt}\PYG{o}{.}\PYG{n}{xlabel}\PYG{p}{(}\PYG{l+s+s1}{\PYGZsq{}}\PYG{l+s+s1}{y}\PYG{l+s+s1}{\PYGZsq{}}\PYG{p}{)}
\PYG{n}{plt}\PYG{o}{.}\PYG{n}{ylabel}\PYG{p}{(}\PYG{l+s+s1}{\PYGZsq{}}\PYG{l+s+s1}{x}\PYG{l+s+s1}{\PYGZsq{}}\PYG{p}{)}
\PYG{n}{plt}\PYG{o}{.}\PYG{n}{title}\PYG{p}{(}\PYG{l+s+s1}{\PYGZsq{}}\PYG{l+s+s1}{Standard approach}\PYG{l+s+s1}{\PYGZsq{}}\PYG{p}{)}
\PYG{n}{plt}\PYG{o}{.}\PYG{n}{legend}\PYG{p}{(}\PYG{p}{)}
\PYG{n}{plt}\PYG{o}{.}\PYG{n}{show}\PYG{p}{(}\PYG{p}{)}
\end{sphinxVerbatim}

\noindent\sphinxincludegraphics{{intro-teaser_15_0}.png}

😱 This is obviously completely wrong! The red solution is nowhere near one of the two modes of our solution shown in gray.

Note that the red line is often not perfectly at zero, which is where the two modes of the solution should average out in the continuous setting. This is caused by the relatively coarse sampling with only 200 points in this example.

\bigskip\hrule\bigskip

\section{A differentiable physics approach}
\label{\detokenize{intro-teaser:a-differentiable-physics-approach}}
Now let’s apply a differentiable physics approach to find \(f\): we’ll directly include our discretized model \(\mathcal P\) in the training.

There is no real data generation step; we only need to sample from the \([0,1]\) interval. We’ll simply keep the same \(x\) locations used in the previous case, and a new instance of a NN with the same architecture as before \sphinxcode{\sphinxupquote{nn\_dp}}:

\begin{sphinxVerbatim}[commandchars=\\\{\}]
\PYG{c+c1}{\PYGZsh{} X\PYGZhy{}Data}
\PYG{c+c1}{\PYGZsh{} X = X , we can directly re\PYGZhy{}use the X from above, nothing has changed...}
\PYG{c+c1}{\PYGZsh{} Y is evaluated on the fly}

\PYG{c+c1}{\PYGZsh{} Model}
\PYG{n}{nn\PYGZus{}dp} \PYG{o}{=} \PYG{n}{tf}\PYG{o}{.}\PYG{n}{keras}\PYG{o}{.}\PYG{n}{models}\PYG{o}{.}\PYG{n}{Sequential}\PYG{p}{(}\PYG{p}{[}
  \PYG{n}{tf}\PYG{o}{.}\PYG{n}{keras}\PYG{o}{.}\PYG{n}{layers}\PYG{o}{.}\PYG{n}{Dense}\PYG{p}{(}\PYG{l+m+mi}{10}\PYG{p}{,} \PYG{n}{activation}\PYG{o}{=}\PYG{n}{act}\PYG{p}{)}\PYG{p}{,}
  \PYG{n}{tf}\PYG{o}{.}\PYG{n}{keras}\PYG{o}{.}\PYG{n}{layers}\PYG{o}{.}\PYG{n}{Dense}\PYG{p}{(}\PYG{l+m+mi}{10}\PYG{p}{,} \PYG{n}{activation}\PYG{o}{=}\PYG{n}{act}\PYG{p}{)}\PYG{p}{,}
  \PYG{n}{tf}\PYG{o}{.}\PYG{n}{keras}\PYG{o}{.}\PYG{n}{layers}\PYG{o}{.}\PYG{n}{Dense}\PYG{p}{(}\PYG{l+m+mi}{1}\PYG{p}{,} \PYG{n}{activation}\PYG{o}{=}\PYG{l+s+s1}{\PYGZsq{}}\PYG{l+s+s1}{linear}\PYG{l+s+s1}{\PYGZsq{}}\PYG{p}{)}\PYG{p}{]}\PYG{p}{)}
\end{sphinxVerbatim}

The loss function is the crucial point for training: we directly incorporate the function f into the loss. In this simple case, the \sphinxcode{\sphinxupquote{loss\_dp}} function simply computes the square of the prediction \sphinxcode{\sphinxupquote{y\_pred}}.

Later on, a lot more could happen here: we could evaluate finite\sphinxhyphen{}difference stencils on the predicted solution, or compute a whole implicit time\sphinxhyphen{}integration step of a solver. Here we have a simple \sphinxstyleemphasis{mean\sphinxhyphen{}squared error} term of the form \(|y_{\text{pred}}^2 - y_{\text{true}}|^2\), which we are minimizing during training. It’s not necessary to make it so simple: the more knowledge and numerical methods we can incorporate, the better we can guide the training process.

\begin{sphinxVerbatim}[commandchars=\\\{\}]
\PYG{c+c1}{\PYGZsh{}Loss}
\PYG{n}{mse} \PYG{o}{=} \PYG{n}{tf}\PYG{o}{.}\PYG{n}{keras}\PYG{o}{.}\PYG{n}{losses}\PYG{o}{.}\PYG{n}{MeanSquaredError}\PYG{p}{(}\PYG{p}{)}
\PYG{k}{def} \PYG{n+nf}{loss\PYGZus{}dp}\PYG{p}{(}\PYG{n}{y\PYGZus{}true}\PYG{p}{,} \PYG{n}{y\PYGZus{}pred}\PYG{p}{)}\PYG{p}{:}
    \PYG{k}{return} \PYG{n}{mse}\PYG{p}{(}\PYG{n}{y\PYGZus{}true}\PYG{p}{,}\PYG{n}{y\PYGZus{}pred}\PYG{o}{*}\PYG{o}{*}\PYG{l+m+mi}{2}\PYG{p}{)}

\PYG{n}{optimizer\PYGZus{}dp} \PYG{o}{=} \PYG{n}{tf}\PYG{o}{.}\PYG{n}{keras}\PYG{o}{.}\PYG{n}{optimizers}\PYG{o}{.}\PYG{n}{Adam}\PYG{p}{(}\PYG{n}{lr}\PYG{o}{=}\PYG{l+m+mf}{0.001}\PYG{p}{)}
\PYG{n}{nn\PYGZus{}dp}\PYG{o}{.}\PYG{n}{compile}\PYG{p}{(}\PYG{n}{optimizer}\PYG{o}{=}\PYG{n}{optimizer\PYGZus{}dp}\PYG{p}{,} \PYG{n}{loss}\PYG{o}{=}\PYG{n}{loss\PYGZus{}dp}\PYG{p}{)}
\end{sphinxVerbatim}

\begin{sphinxVerbatim}[commandchars=\\\{\}]
\PYG{c+c1}{\PYGZsh{}Training}
\PYG{n}{results\PYGZus{}dp} \PYG{o}{=} \PYG{n}{nn\PYGZus{}dp}\PYG{o}{.}\PYG{n}{fit}\PYG{p}{(}\PYG{n}{X}\PYG{p}{,} \PYG{n}{X}\PYG{p}{,} \PYG{n}{epochs}\PYG{o}{=}\PYG{l+m+mi}{5}\PYG{p}{,} \PYG{n}{batch\PYGZus{}size}\PYG{o}{=}\PYG{l+m+mi}{5}\PYG{p}{,} \PYG{n}{verbose}\PYG{o}{=}\PYG{l+m+mi}{1}\PYG{p}{)}
\end{sphinxVerbatim}

\begin{sphinxVerbatim}[commandchars=\\\{\}]
Epoch 1/5
40/40 [==============================] \PYGZhy{} 0s 656us/step \PYGZhy{} loss: 0.2814
Epoch 2/5
40/40 [==============================] \PYGZhy{} 0s 1ms/step \PYGZhy{} loss: 0.1259
Epoch 3/5
40/40 [==============================] \PYGZhy{} 0s 962us/step \PYGZhy{} loss: 0.0038
Epoch 4/5
40/40 [==============================] \PYGZhy{} 0s 949us/step \PYGZhy{} loss: 0.0014
Epoch 5/5
40/40 [==============================] \PYGZhy{} 0s 645us/step \PYGZhy{} loss: 0.0012
\end{sphinxVerbatim}

Now the network actually has learned a good inverse of the parabola function! The following plot shows the solution in green.

\begin{sphinxVerbatim}[commandchars=\\\{\}]
\PYG{c+c1}{\PYGZsh{} Results}
\PYG{n}{plt}\PYG{o}{.}\PYG{n}{plot}\PYG{p}{(}\PYG{n}{X}\PYG{p}{,}\PYG{n}{Y}\PYG{p}{,}\PYG{l+s+s1}{\PYGZsq{}}\PYG{l+s+s1}{.}\PYG{l+s+s1}{\PYGZsq{}}\PYG{p}{,}\PYG{n}{label}\PYG{o}{=}\PYG{l+s+s1}{\PYGZsq{}}\PYG{l+s+s1}{Datapoints}\PYG{l+s+s1}{\PYGZsq{}}\PYG{p}{,} \PYG{n}{color}\PYG{o}{=}\PYG{l+s+s2}{\PYGZdq{}}\PYG{l+s+s2}{lightgray}\PYG{l+s+s2}{\PYGZdq{}}\PYG{p}{)}
\PYG{c+c1}{\PYGZsh{}plt.plot(X,nn\PYGZus{}sv.predict(X),\PYGZsq{}.\PYGZsq{},label=\PYGZsq{}Supervised\PYGZsq{}, color=\PYGZdq{}red\PYGZdq{}) \PYGZsh{} optional for comparison}
\PYG{n}{plt}\PYG{o}{.}\PYG{n}{plot}\PYG{p}{(}\PYG{n}{X}\PYG{p}{,}\PYG{n}{nn\PYGZus{}dp}\PYG{o}{.}\PYG{n}{predict}\PYG{p}{(}\PYG{n}{X}\PYG{p}{)}\PYG{p}{,}\PYG{l+s+s1}{\PYGZsq{}}\PYG{l+s+s1}{.}\PYG{l+s+s1}{\PYGZsq{}}\PYG{p}{,}\PYG{n}{label}\PYG{o}{=}\PYG{l+s+s1}{\PYGZsq{}}\PYG{l+s+s1}{Diff. Phys.}\PYG{l+s+s1}{\PYGZsq{}}\PYG{p}{,} \PYG{n}{color}\PYG{o}{=}\PYG{l+s+s2}{\PYGZdq{}}\PYG{l+s+s2}{green}\PYG{l+s+s2}{\PYGZdq{}}\PYG{p}{)} 
\PYG{n}{plt}\PYG{o}{.}\PYG{n}{xlabel}\PYG{p}{(}\PYG{l+s+s1}{\PYGZsq{}}\PYG{l+s+s1}{x}\PYG{l+s+s1}{\PYGZsq{}}\PYG{p}{)}
\PYG{n}{plt}\PYG{o}{.}\PYG{n}{ylabel}\PYG{p}{(}\PYG{l+s+s1}{\PYGZsq{}}\PYG{l+s+s1}{y}\PYG{l+s+s1}{\PYGZsq{}}\PYG{p}{)}
\PYG{n}{plt}\PYG{o}{.}\PYG{n}{title}\PYG{p}{(}\PYG{l+s+s1}{\PYGZsq{}}\PYG{l+s+s1}{Differentiable physics approach}\PYG{l+s+s1}{\PYGZsq{}}\PYG{p}{)}
\PYG{n}{plt}\PYG{o}{.}\PYG{n}{legend}\PYG{p}{(}\PYG{p}{)}
\PYG{n}{plt}\PYG{o}{.}\PYG{n}{show}\PYG{p}{(}\PYG{p}{)}
\end{sphinxVerbatim}

\noindent\sphinxincludegraphics{{intro-teaser_24_0}.png}

This looks much better 😎, at least in the range of 0.1 to 1.

What has happened here?
\begin{itemize}
\item {} 
We’ve prevented an undesired averaging of multiple modes in the solution by evaluating our discrete model w.r.t. current prediction of the network, rather than using a pre\sphinxhyphen{}computed solution. This lets us find the best mode near the network prediction, and prevents an averaging of the modes that exist in the solution manifold.

\item {} 
We’re still only getting one side of the curve! This is to be expected because we’re representing the solutions with a deterministic function. Hence, we can only represent a single mode. Interestingly, whether it’s the top or bottom mode is determined by the random initialization of the weights in \(f\) \sphinxhyphen{} run the example a couple of times to see this effect in action. To capture multiple modes we’d need to extend the NN to capture the full distribution of the outputs and parametrize it with additional dimensions.

\item {} 
The region with \(x\) near zero is typically still off in this example. The network essentially learns a linear approximation of one half of the parabola here. This is partially caused by the weak neural network: it is very small and shallow. In addition, the evenly spread of sample points along the x\sphinxhyphen{}axis bias the NN towards the larger y values. These contribute more to the loss, and hence the network invests most of its resources to reduce the error in this region.

\end{itemize}

\section{Discussion}
\label{\detokenize{intro-teaser:discussion}}
It’s a very simple example, but it very clearly shows a failure case for supervised learning. While it might seem very artificial at first sight, many practical PDEs exhibit a variety of these modes, and it’s often not clear where (and how many) exist in the solution manifold we’re interested in. Using supervised learning is very dangerous in such cases. We might unknowingly get an average of these different modes.

Good and obvious examples are bifurcations in fluid flow. Smoke rising above a candle will start out straight, and then, due to tiny perturbations in its motion, start oscillating in a random direction. The images below illustrate this case via \sphinxstyleemphasis{numerical perturbations}: the perfectly symmetric setup will start turning left or right, depending on how the approximation errors build up. Averaging the two modes would give an unphysical, straight flow similar to the parabola example above.

Similarly, we have different modes in many numerical solutions, and typically it’s important to recover them, rather than averaging them out. Hence, we’ll show how to leverage training via \sphinxstyleemphasis{differentiable physics} in the following chapters for more practical and complex cases.

\begin{figure}[htbp]
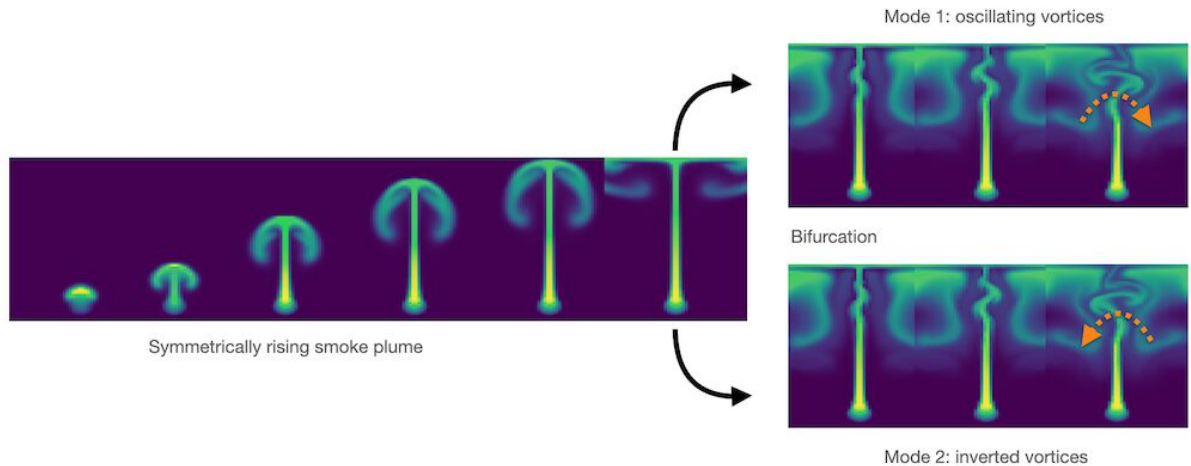

\centering
\capstart

\noindent\sphinxincludegraphics[height=240\sphinxpxdimen]{{intro-fluid-bifurcation}.jpg}
\caption{A bifurcation in a buoyancy\sphinxhyphen{}driven fluid flow: the “smoke” shown in green color starts rising in a perfectly straight manner, but tiny numerical inaccuracies grow over time to lead to an instability with vortices alternating to one side (top\sphinxhyphen{}right), or in the opposite direction (bottom\sphinxhyphen{}right).}\label{\detokenize{intro-teaser:intro-fluid-bifurcation}}\end{figure}

\section{Next steps}
\label{\detokenize{intro-teaser:next-steps}}
For each of the following notebooks, there’s a “next steps” section like the one below which contains recommendations about where to start modifying the code. After all, the whole point of these notebooks is to have readily executable programs as a basis for own experiments. The data set and NN sizes of the examples are often quite small to reduce the runtime of the notebooks, but they’re nonetheless good starting points for potentially complex and large projects.

For the simple DP example above:
\begin{itemize}
\item {} 
This notebook is intentionally using a very simple setup. Change the training setup and NN above to obtain a higher\sphinxhyphen{}quality solution such as the green one shown in the very first image at the top.

\item {} 
Or try extending the setup to a 2D case, i.e. a paraboloid. Given the function \(\mathcal P:(y_1,y_2)\to y_1^2+y_2^2\), find an inverse function \(f\) such that \(\mathcal P(f(x)) = x\) for all \(x\) in \([0,1]\).

\item {} 
If you want to experiment without installing anything, you can also \sphinxhref{https://colab.research.google.com/github/tum-pbs/pbdl-book/blob/main/intro-teaser.ipynb}{{[}run this notebook in colab{]}}.

\end{itemize}

\chapter{Overview}
\label{\detokenize{overview:overview}}\label{\detokenize{overview::doc}}
The name of this book, \sphinxstyleemphasis{Physics\sphinxhyphen{}Based Deep Learning},
denotes combinations of physical modeling and numerical simulations with
methods based on artificial neural networks.
The general direction of Physics\sphinxhyphen{}Based Deep Learning represents a very
active, quickly growing and exciting field of research. The following chapter will
give a more thorough introduction to the topic and establish the basics
for following chapters.

\begin{figure}[htbp]
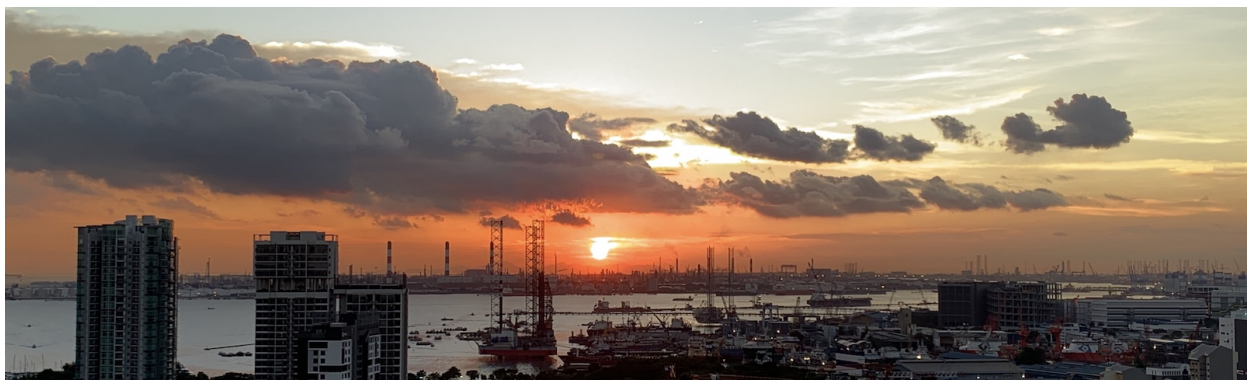

\centering
\capstart

\noindent\sphinxincludegraphics[height=240\sphinxpxdimen]{{overview-pano}.jpg}
\caption{Understanding our environment, and predicting how it will evolve is one of the key challenges of humankind.
A key tool for achieving these goals are simulations, and next\sphinxhyphen{}gen simulations
could strongly profit from integrating deep learning components to make even
more accurate predictions about our world.}\label{\detokenize{overview:overview-pano}}\end{figure}

\section{Motivation}
\label{\detokenize{overview:motivation}}
From weather and climate forecasts {[}\hyperlink{cite.references:id82}{Sto14}{]} (see the picture above),
over quantum physics {[}\hyperlink{cite.references:id84}{OMalleyBK+16}{]},
to the control of plasma fusion {[}\hyperlink{cite.references:id83}{MLA+19}{]},
using numerical analysis to obtain solutions for physical models has
become an integral part of science.

In recent years, machine learning technologies and \sphinxstyleemphasis{deep neural networks} in particular,
have led to impressive achievements in a variety of fields:
from image classification {[}\hyperlink{cite.references:id85}{KSH12}{]} over
natural language processing {[}\hyperlink{cite.references:id86}{RWC+19}{]},
and more recently also for protein folding {[}\hyperlink{cite.references:id87}{Qur19}{]}.
The field is very vibrant and quickly developing, with the promise of vast possibilities.

\subsection{Replacing traditional simulations?}
\label{\detokenize{overview:replacing-traditional-simulations}}
These success stories of deep learning (DL) approaches
have given rise to concerns that this technology has
the potential to replace the traditional, simulation\sphinxhyphen{}driven approach to science.
E.g., recent works show that NN\sphinxhyphen{}based surrogate models achieve accuracies required
for real\sphinxhyphen{}world, industrial applications such as airfoil flows {[}\hyperlink{cite.references:id2}{CT21}{]}, while at the
same time outperforming traditional solvers by orders of magnitude in terms of runtime.

Instead of relying on models that are carefully crafted
from first principles, can data collections of sufficient size
be processed to provide the correct answers?
As we’ll show in the next chapters, this concern is unfounded.
Rather, it is crucial for the next generation of simulation systems
to bridge both worlds: to
combine \sphinxstyleemphasis{classical numerical} techniques with \sphinxstyleemphasis{deep learning} methods.

One central reason for the importance of this combination is
that DL approaches are powerful, but at the same time strongly profit
from domain knowledge in the form of physical models.
DL techniques and NNs are novel, sometimes difficult to apply, and
it is admittedly often non\sphinxhyphen{}trivial to properly integrate our understanding
of physical processes into the learning algorithms.

Over the last decades,
highly specialized and accurate discretization schemes have
been developed to solve fundamental model equations such
as the Navier\sphinxhyphen{}Stokes, Maxwell’s, or Schroedinger’s equations.
Seemingly trivial changes to the discretization can determine
whether key phenomena are visible in the solutions or not.
Rather than discarding the powerful methods that have been
developed in the field of numerical mathematics, this book will
show that it is highly beneficial to use them as much as possible
when applying DL.

\subsection{Black boxes and magic?}
\label{\detokenize{overview:black-boxes-and-magic}}
People who are unfamiliar with DL methods often associate neural networks
with \sphinxstyleemphasis{black boxes}, and see the training processes as something that is beyond the grasp
of human understanding. However, these viewpoints typically stem from
relying on hearsay and not dealing with the topic enough.

Rather, the situation is a very common one in science: we are facing a new class of methods,
and “all the gritty details” are not yet fully worked out. However, this is pretty common
for scientific advances.
Numerical methods themselves are a good example. Around 1950, numerical approximations
and solvers had a tough standing. E.g., to cite H. Goldstine,
numerical instabilities were considered to be a “constant source of
anxiety in the future” {[}\hyperlink{cite.references:id79}{Gol90}{]}.
By now we have a pretty good grasp of these instabilities, and numerical methods
are ubiquitous and well established.

Thus, it is important to be aware of the fact that – in a way – there is nothing
magical or otherworldly to deep learning methods. They’re simply another set of
numerical tools. That being said, they’re clearly fairly new, and right now
definitely the most powerful set of tools we have for non\sphinxhyphen{}linear problems.
Just because all the details aren’t fully worked out and nicely written up,
that shouldn’t stop us from including these powerful methods in our numerical toolbox.

\subsection{Reconciling DL and simulations}
\label{\detokenize{overview:reconciling-dl-and-simulations}}
Taking a step back, the aim of this book is to build on all the powerful techniques that we have
at our disposal for numerical simulations, and use them wherever we can in conjunction
with deep learning.
As such, a central goal is to \sphinxstyleemphasis{reconcile} the data\sphinxhyphen{}centered viewpoint with physical simulations.

\begin{sphinxadmonition}{note}{Goals of this document}

The key aspects that we will address in the following are:
\begin{itemize}
\item {} 
explain how to use deep learning techniques to solve PDE problems,

\item {} 
how to combine them with \sphinxstylestrong{existing knowledge} of physics,

\item {} 
without \sphinxstylestrong{discarding} our knowledge about numerical methods.

\end{itemize}
\end{sphinxadmonition}

The resulting methods have a huge potential to improve
what can be done with numerical methods: in scenarios
where a solver targets cases from a certain well\sphinxhyphen{}defined problem
domain repeatedly, it can for instance make a lot of sense to once invest
significant resources to train
a neural network that supports the repeated solves. Based on the
domain\sphinxhyphen{}specific specialization of this network, such a hybrid
could vastly outperform traditional, generic solvers. And despite
the many open questions, first publications have demonstrated
that this goal is not overly far away {[}\hyperlink{cite.references:id94}{KSA+21}, \hyperlink{cite.references:id6}{UBH+20}{]}.

Another way to look at it is that all mathematical models of our nature
are idealized approximations and contain errors. A lot of effort has been
made to obtain very good model equations, but to make the next
big step forward, DL methods offer a very powerful tool to close the
remaining gap towards reality {[}\hyperlink{cite.references:id95}{AAC+19}{]}.

\section{Categorization}
\label{\detokenize{overview:categorization}}
Within the area of \sphinxstyleemphasis{physics\sphinxhyphen{}based deep learning},
we can distinguish a variety of different
approaches, from targeting constraints, combined methods, and
optimizations to applications. More specifically, all approaches either target
\sphinxstyleemphasis{forward} simulations (predicting state or temporal evolution) or \sphinxstyleemphasis{inverse}
problems (e.g., obtaining a parametrization for a physical system from
observations).

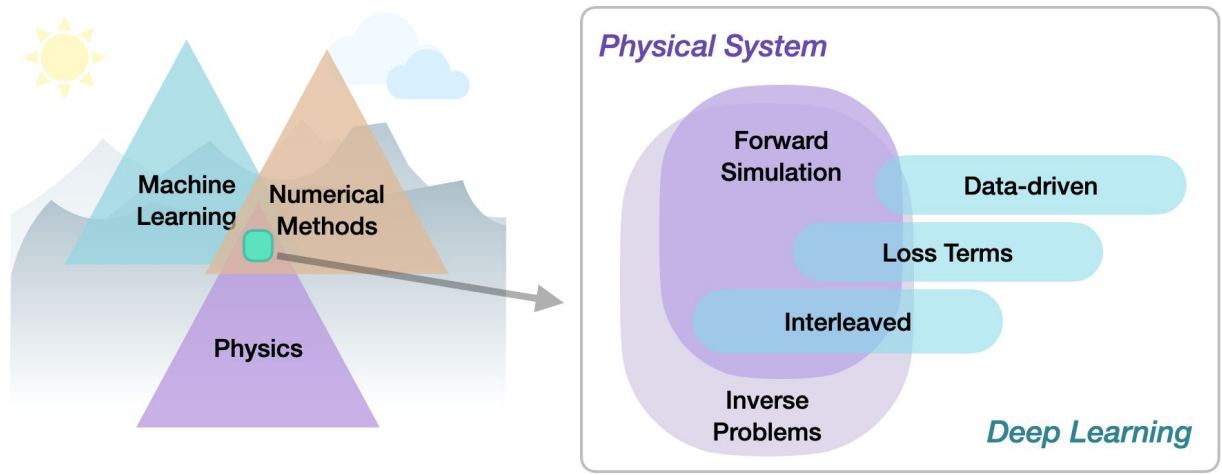
\sphinxincludegraphics{{physics-based-deep-learning-overview}.jpg}

No matter whether we’re considering forward or inverse problems,
the most crucial differentiation for the following topics lies in the
nature of the integration  between DL techniques
and the domain knowledge, typically in the form of model equations
via partial differential equations (PDEs).
The following three categories can be
identified to roughly categorize \sphinxstyleemphasis{physics\sphinxhyphen{}based deep learning} (PBDL)
techniques:
\begin{itemize}
\item {} 
\sphinxstyleemphasis{Supervised}: the data is produced by a physical system (real or simulated),
but no further interaction exists. This is the classic machine learning approach.

\item {} 
\sphinxstyleemphasis{Loss\sphinxhyphen{}terms}: the physical dynamics (or parts thereof) are encoded in the
loss function, typically in the form of differentiable operations. The
learning process can repeatedly evaluate the loss, and usually receives
gradients from a PDE\sphinxhyphen{}based formulation. These soft\sphinxhyphen{}constraints sometimes also go
under the name “physics\sphinxhyphen{}informed” training.

\item {} 
\sphinxstyleemphasis{Interleaved}: the full physical simulation is interleaved and combined with
an output from a deep neural network; this requires a fully differentiable
simulator and represents the tightest coupling between the physical system and
the learning process. Interleaved differentiable physics approaches are especially important for
temporal evolutions, where they can yield an estimate of the future behavior of the
dynamics.

\end{itemize}

Thus, methods can be categorized in terms of forward versus inverse
solve, and how tightly the physical model is integrated into the
optimization loop that trains the deep neural network. Here, especially
interleaved approaches that leverage \sphinxstyleemphasis{differentiable physics} allow for
very tight integration of deep learning and numerical simulation methods.

\section{Looking ahead}
\label{\detokenize{overview:looking-ahead}}
\sphinxstyleemphasis{Physical simulations} are a huge field, and we won’t be able to cover all possible types of physical models and simulations.

\begin{sphinxadmonition}{note}{Note:}
Rather, the focus of this book lies on:
\begin{itemize}
\item {} 
\sphinxstyleemphasis{Field\sphinxhyphen{}based simulations} (no Lagrangian methods)

\item {} 
Combinations with \sphinxstyleemphasis{deep learning} (plenty of other interesting ML techniques exist, but won’t be discussed here)

\item {} 
Experiments are left as an \sphinxstyleemphasis{outlook} (i.e., replacing synthetic data with real\sphinxhyphen{}world observations)

\end{itemize}
\end{sphinxadmonition}

It’s also worth noting that we’re starting to build the methods from some very
fundamental building blocks. Here are some considerations for skipping ahead to the later chapters.

\begin{sphinxadmonition}{note}{Hint: You can skip ahead if…}
\begin{itemize}
\item {} 
you’re very familiar with numerical methods and PDE solvers, and want to get started with DL topics right away. The {\hyperref[\detokenize{supervised::doc}]{\sphinxcrossref{\DUrole{doc}{Supervised Training}}}} chapter is a good starting point then.

\item {} 
On the other hand, if you’re already deep into NNs\&Co, and you’d like to skip ahead to the research related topics, we recommend starting in the {\hyperref[\detokenize{physicalloss::doc}]{\sphinxcrossref{\DUrole{doc}{Physical Loss Terms}}}} chapter, which lays the foundations for the next chapters.

\end{itemize}

A brief look at our \sphinxstyleemphasis{notation} in the {\hyperref[\detokenize{notation::doc}]{\sphinxcrossref{\DUrole{doc}{Notation and Abbreviations}}}} chapter won’t hurt in both cases, though!
\end{sphinxadmonition}

\section{Implementations}
\label{\detokenize{overview:implementations}}
This text also represents an introduction to a wide range of deep learning and simulation APIs.
We’ll use popular deep learning APIs such as \sphinxstyleemphasis{pytorch} \sphinxurl{https://pytorch.org} and \sphinxstyleemphasis{tensorflow} \sphinxurl{https://www.tensorflow.org}, and additionally
give introductions into the differentiable simulation framework \sphinxstyleemphasis{ΦFlow (phiflow)} \sphinxurl{https://github.com/tum-pbs/PhiFlow}. Some examples also use \sphinxstyleemphasis{JAX} \sphinxurl{https://github.com/google/jax}. Thus after going through
these examples, you should have a good overview of what’s available in current APIs, such that
the best one can be selected for new tasks.

As we’re (in most Jupyter notebook examples) dealing with stochastic optimizations, many of the following code examples will produce slightly different results each time they’re run. This is fairly common with NN training, but it’s important to keep in mind when executing the code. It also means that the numbers discussed in the text might not exactly match the numbers you’ll see after re\sphinxhyphen{}running the examples.

\bigskip\hrule\bigskip

\section{Models and Equations}
\label{\detokenize{overview-equations:models-and-equations}}\label{\detokenize{overview-equations::doc}}
Below we’ll give a brief (really \sphinxstyleemphasis{very} brief!) intro to deep learning, primarily to introduce the notation.
In addition we’ll discuss some \sphinxstyleemphasis{model equations} below. Note that we’ll avoid using \sphinxstyleemphasis{model} to denote trained neural networks, in contrast to some other texts and APIs. These will be called “NNs” or “networks”. A “model” will typically denote a set of model equations for a physical effect, usually PDEs.

\subsection{Deep learning and neural networks}
\label{\detokenize{overview-equations:deep-learning-and-neural-networks}}
In this book we focus on the connection with physical
models, and there are lots of great introductions to deep learning.
Hence, we’ll keep it short:
the goal in deep learning is to approximate an unknown function
\begin{equation}\label{equation:overview-equations:learn-base}
\begin{split}
f^*(x) = y^* , 
\end{split}
\end{equation}
where \(y^*\) denotes reference or “ground truth” solutions.
\(f^*(x)\) should be approximated with an NN representation \(f(x;\theta)\). We typically determine \(f\)
with the help of some variant of an error function \(e(y,y^*)\), where \(y=f(x;\theta)\) is the output
of the NN.
This gives a minimization problem to find \(f(x;\theta)\) such that \(e\) is minimized.
In the simplest case, we can use an \(L^2\) error, giving
\begin{equation}\label{equation:overview-equations:learn-l2}
\begin{split}
\text{arg min}_{\theta} | f(x;\theta) - y^* |_2^2
\end{split}
\end{equation}
We typically optimize, i.e. \sphinxstyleemphasis{train},
with a stochastic gradient descent (SGD) optimizer of choice, e.g. Adam {[}\hyperlink{cite.references:id89}{KB14}{]}.
We’ll rely on auto\sphinxhyphen{}diff to compute the gradient w.r.t. weights, \(\partial f / \partial \theta\),
We will also assume that \(e\) denotes a \sphinxstyleemphasis{scalar} error function (also
called cost, or objective function).
It is crucial for the efficient calculation of gradients that this function is scalar.

For training we distinguish: the \sphinxstylestrong{training} data set drawn from some distribution,
the \sphinxstylestrong{validation} set (from the same distribution, but different data),
and \sphinxstylestrong{test} data sets with \sphinxstyleemphasis{some} different distribution than the training one.
The latter distinction is important. For the test set we want
\sphinxstyleemphasis{out of distribution} (OOD) data to check how well our trained model generalizes.
Note that this gives a huge range of possibilities for the test data set:
from tiny changes that will certainly work,
up to completely different inputs that are essentially guaranteed to fail.
There’s no gold standard, but test data should be generated with care.

Enough for now \sphinxhyphen{} if all the above wasn’t totally obvious for you, we very strongly recommend to
read chapters 6 to 9 of the \sphinxhref{https://www.deeplearningbook.org}{Deep Learning book},
especially the sections about \sphinxhref{https://www.deeplearningbook.org/contents/mlp.html}{MLPs}
and “Conv\sphinxhyphen{}Nets”, i.e. \sphinxhref{https://www.deeplearningbook.org/contents/convnets.html}{CNNs}.

\begin{sphinxadmonition}{note}{Note:}
Classification vs Regression

The classic ML distinction between \sphinxstyleemphasis{classification} and \sphinxstyleemphasis{regression} problems is not so important here:
we only deal with \sphinxstyleemphasis{regression} problems in the following.
\end{sphinxadmonition}

\subsection{Partial differential equations as physical models}
\label{\detokenize{overview-equations:partial-differential-equations-as-physical-models}}
The following section will give a brief outlook for the model equations
we’ll be using later on in the DL examples.
We typically target continuous PDEs denoted by \(\mathcal P^*\)
whose solution is of interest in a spatial domain \(\Omega \subset \mathbb{R}^d\) in \(d \in {1,2,3} \) dimensions.
In addition, wo often consider a time evolution for a finite time interval \(t \in \mathbb{R}^{+}\).
The corresponding fields are either d\sphinxhyphen{}dimensional vector fields, for instance \(\mathbf{u}: \mathbb{R}^d \times \mathbb{R}^{+} \rightarrow \mathbb{R}^d\),
or scalar \(\mathbf{p}: \mathbb{R}^d \times \mathbb{R}^{+} \rightarrow \mathbb{R}\).
The components of a vector are typically denoted by \(x,y,z\) subscripts, i.e.,
\(\mathbf{v} = (v_x, v_y, v_z)^T\) for \(d=3\), while
positions are denoted by \(\mathbf{x} \in \Omega\).

To obtain unique solutions for \(\mathcal P^*\) we need to specify suitable
initial conditions, typically for all quantities of interest at \(t=0\),
and boundary conditions for the boundary of \(\Omega\), denoted by \(\Gamma\) in
the following.

\(\mathcal P^*\) denotes
a continuous formulation, where we make mild assumptions about
its continuity, we will typically assume that first and second derivatives exist.

We can then use numerical methods to obtain approximations
of a smooth function such as \(\mathcal P^*\) via discretization.
These invariably introduce discretization errors, which we’d like to keep as small as possible.
These errors can be measured in terms of the deviation from the exact analytical solution,
and for discrete simulations of PDEs, they are typically expressed as a function of the truncation error
\(O( \Delta x^k )\), where \(\Delta x\) denotes the spatial step size of the discretization.
Likewise, we typically have a temporal discretization via a time step \(\Delta t\).

\begin{sphinxadmonition}{note}{Notation and abbreviations}

If unsure, please check the summary of our mathematical notation
and the abbreviations used in: {\hyperref[\detokenize{notation::doc}]{\sphinxcrossref{\DUrole{doc}{Notation and Abbreviations}}}}.
\end{sphinxadmonition}

We solve a discretized PDE \(\mathcal{P}\) by performing steps of size \(\Delta t\).
The solution can be expressed as a function of \(\mathbf{u}\) and its derivatives:
\(\mathbf{u}(\mathbf{x},t+\Delta t) = 
\mathcal{P}( \mathbf{u}_{x}, \mathbf{u}_{xx}, ... \mathbf{u}_{xx...x} )\), where
\(\mathbf{u}_{x}\) denotes the spatial derivatives \(\partial \mathbf{u}(\mathbf{x},t) / \partial \mathbf{x}\).

For all PDEs, we will assume non\sphinxhyphen{}dimensional parametrizations as outlined below,
which could be re\sphinxhyphen{}scaled to real world quantities with suitable scaling factors.
Next, we’ll give an overview of the model equations, before getting started
with actual simulations and implementation examples on the next page.

\bigskip\hrule\bigskip

\subsection{Some example PDEs}
\label{\detokenize{overview-equations:some-example-pdes}}
The following PDEs are good examples, and we’ll use them later on in different settings to show how to incorporate them into DL approaches.

\subsubsection{Burgers}
\label{\detokenize{overview-equations:burgers}}
We’ll often consider Burgers’ equation
in 1D or 2D as a starting point.
It represents a well\sphinxhyphen{}studied PDE, which (unlike Navier\sphinxhyphen{}Stokes)
does not include any additional constraints such as conservation of mass.
Hence, it leads to interesting shock formations.
It contains an advection term (motion / transport) and a diffusion term (dissipation due to the second law of thermodynamics).
In 2D, it is given by:
\begin{equation}\label{equation:overview-equations:model-burgers2d}
\begin{split}\begin{aligned}
  \frac{\partial u_x}{\partial{t}} + \mathbf{u} \cdot \nabla u_x &=
  \nu \nabla\cdot \nabla u_x + g_x, 
  \\
  \frac{\partial u_y}{\partial{t}} + \mathbf{u} \cdot \nabla u_y &=
  \nu \nabla\cdot \nabla u_y + g_y \ ,
\end{aligned}\end{split}
\end{equation}
where \(\nu\) and \(\mathbf{g}\) denote diffusion constant and external forces, respectively.

A simpler variant of Burgers’ equation in 1D without forces,
denoting the single 1D velocity component as \(u = u_x\),
is given by:
\begin{equation}\label{equation:overview-equations:model-burgers1d}
\begin{split}
\frac{\partial u}{\partial{t}} + u \nabla u = \nu \nabla \cdot \nabla u \ . 
\end{split}
\end{equation}

\subsubsection{Navier\sphinxhyphen{}Stokes}
\label{\detokenize{overview-equations:navier-stokes}}
A good next step in terms of complexity is given by the
Navier\sphinxhyphen{}Stokes equations, which are a well\sphinxhyphen{}established model for fluids.
In addition to an equation for the conservation of momentum (similar to Burgers),
they include an equation for the conservation of mass. This prevents the
formation of shock waves, but introduces a new challenge for numerical methods
in the form of a hard\sphinxhyphen{}constraint for divergence free motions.

In 2D, the Navier\sphinxhyphen{}Stokes equations without any external forces can be written as:
\begin{equation}\label{equation:overview-equations:model-ns2d}
\begin{split}\begin{aligned}
    \frac{\partial u_x}{\partial{t}} + \mathbf{u} \cdot \nabla u_x &=
    - \frac{\Delta t}{\rho}\nabla{p} + \nu \nabla\cdot \nabla u_x  
    \\
    \frac{\partial u_y}{\partial{t}} + \mathbf{u} \cdot \nabla u_y &=
    - \frac{\Delta t}{\rho}\nabla{p} + \nu \nabla\cdot \nabla u_y  
    \\
    \text{subject to} \quad \nabla \cdot \mathbf{u} &= 0
\end{aligned}\end{split}
\end{equation}
where, like before, \(\nu\) denotes a diffusion constant for viscosity.
In practice, the \(\Delta t\) factor for the pressure term can be often simplified to
\(1/\rho\) as it simply yields a scaling of the pressure gradient used to make
the velocity divergence free. We’ll typically use this simplification later on
in implementations, effectively computing an instantaneous pressure.

An interesting variant is obtained by including the
\sphinxhref{https://en.wikipedia.org/wiki/Boussinesq\_approximation\_(buoyancy)}{Boussinesq approximation}
for varying densities, e.g., for simple temperature changes of the fluid.
With a marker field \(v\) that indicates regions of high temperature,
it yields the following set of equations:
\begin{equation}\label{equation:overview-equations:model-boussinesq2d}
\begin{split}\begin{aligned}
  \frac{\partial u_x}{\partial{t}} + \mathbf{u} \cdot \nabla u_x &= - \frac{\Delta t}{\rho} \nabla p 
  \\
  \frac{\partial u_y}{\partial{t}} + \mathbf{u} \cdot \nabla u_y &= - \frac{\Delta t}{\rho} \nabla p + \xi v
  \\
  \text{subject to} \quad \nabla \cdot \mathbf{u} &= 0,
  \\
  \frac{\partial v}{\partial{t}} + \mathbf{u} \cdot \nabla v &= 0 
\end{aligned}\end{split}
\end{equation}
where \(\xi\) denotes the strength of the buoyancy force.

And finally, the Navier\sphinxhyphen{}Stokes model in 3D give the following set of equations:
\begin{equation}\label{equation:overview-equations:model-ns3d}
\begin{split}
\begin{aligned}
  \frac{\partial u_x}{\partial{t}} + \mathbf{u} \cdot \nabla u_x &= - \frac{\Delta t}{\rho} \nabla p + \nu \nabla\cdot \nabla u_x 
  \\
  \frac{\partial u_y}{\partial{t}} + \mathbf{u} \cdot \nabla u_y &= - \frac{\Delta t}{\rho} \nabla p + \nu \nabla\cdot \nabla u_y 
  \\
  \frac{\partial u_z}{\partial{t}} + \mathbf{u} \cdot \nabla u_z &= - \frac{\Delta t}{\rho} \nabla p + \nu \nabla\cdot \nabla u_z 
  \\
  \text{subject to} \quad \nabla \cdot \mathbf{u} &= 0.
\end{aligned}
\end{split}
\end{equation}

\subsection{Forward Simulations}
\label{\detokenize{overview-equations:forward-simulations}}
Before we really start with learning methods, it’s important to cover the most basic variant of using the above model equations: a regular “forward” simulation, that starts from a set of initial conditions, and evolves the state of the system over time with a discretized version of the model equation. We’ll show how to run such forward simulations for Burgers’ equation in 1D and for a 2D Navier\sphinxhyphen{}Stokes simulation.

\section{Simple Forward Simulation of Burgers Equation with phiflow}
\label{\detokenize{overview-burgers-forw:simple-forward-simulation-of-burgers-equation-with-phiflow}}\label{\detokenize{overview-burgers-forw::doc}}
This chapter will give an introduction for how to run \sphinxstyleemphasis{forward}, i.e., regular simulations starting with a given initial state and approximating a later state numerically, and introduce the ΦFlow framework. ΦFlow provides a set of differentiable building blocks that directly interface with deep learning frameworks, and hence is a very good basis for the topics of this book. Before going for deeper and more complicated integrations, this notebook (and the next one), will show how regular simulations can be done with ΦFlow. Later on, we’ll show that these simulations can be easily coupled with neural networks.

The main repository for ΦFlow (in the following “phiflow”) is \sphinxurl{https://github.com/tum-pbs/PhiFlow}, and additional API documentation and examples can be found at \sphinxurl{https://tum-pbs.github.io/PhiFlow/}.

For this jupyter notebook (and all following ones), you can find a \sphinxstyleemphasis{“{[}run in colab{]}”} link at the end of the first paragraph (alternatively you can use the launch button at the top of the page). This will load the latest version from the PBDL github repo in a colab notebook that you can execute on the spot:
\sphinxhref{https://colab.research.google.com/github/tum-pbs/pbdl-book/blob/main/overview-burgers-forw.ipynb}{{[}run in colab{]}}

\subsection{Model}
\label{\detokenize{overview-burgers-forw:model}}
As physical model we’ll use Burgers equation.
This equation is a very simple, yet non\sphinxhyphen{}linear and non\sphinxhyphen{}trivial, model equation that can lead to interesting shock formations. Hence, it’s a very good starting point for experiments, and it’s 1D version (from equation \eqref{equation:overview-equations:model-burgers1d}) is given by:
\begin{equation*}
\begin{split}
    \frac{\partial u}{\partial{t}} + u \nabla u =
    \nu \nabla\cdot \nabla u
\end{split}
\end{equation*}

\subsection{Importing and loading phiflow}
\label{\detokenize{overview-burgers-forw:importing-and-loading-phiflow}}
Let’s get some preliminaries out of the way: first we’ll import the phiflow library, more specifically the \sphinxcode{\sphinxupquote{numpy}} operators for fluid flow simulations: \sphinxcode{\sphinxupquote{phi.flow}} (differentiable versions for a DL framework \sphinxstyleemphasis{X} are loaded via \sphinxcode{\sphinxupquote{phi.X.flow}} instead).

\sphinxstylestrong{Note:} Below, the first command with a “!” prefix will install the \sphinxhref{https://github.com/tum-pbs/PhiFlow}{phiflow python package from GitHub} via \sphinxcode{\sphinxupquote{pip}} in your python environment once you uncomment it. We’ve assumed that phiflow isn’t installed, but if you have already done so, just comment out the first line (the same will hold for all following notebooks).

\begin{sphinxVerbatim}[commandchars=\\\{\}]
\PYG{c+ch}{\PYGZsh{}!pip install \PYGZhy{}\PYGZhy{}upgrade \PYGZhy{}\PYGZhy{}quiet phiflow}
\PYG{o}{!}pip install \PYGZhy{}\PYGZhy{}upgrade \PYGZhy{}\PYGZhy{}quiet git+https://github.com/tum\PYGZhy{}pbs/PhiFlow@develop

\PYG{k+kn}{from} \PYG{n+nn}{phi}\PYG{n+nn}{.}\PYG{n+nn}{flow} \PYG{k+kn}{import} \PYG{o}{*}

\PYG{k+kn}{from} \PYG{n+nn}{phi} \PYG{k+kn}{import} \PYG{n}{\PYGZus{}\PYGZus{}version\PYGZus{}\PYGZus{}}
\PYG{n+nb}{print}\PYG{p}{(}\PYG{l+s+s2}{\PYGZdq{}}\PYG{l+s+s2}{Using phiflow version: }\PYG{l+s+si}{\PYGZob{}\PYGZcb{}}\PYG{l+s+s2}{\PYGZdq{}}\PYG{o}{.}\PYG{n}{format}\PYG{p}{(}\PYG{n}{phi}\PYG{o}{.}\PYG{n}{\PYGZus{}\PYGZus{}version\PYGZus{}\PYGZus{}}\PYG{p}{)}\PYG{p}{)}
\end{sphinxVerbatim}

\begin{sphinxVerbatim}[commandchars=\\\{\}]
Using phiflow version: 2.0.0rc2
\end{sphinxVerbatim}

Next we can define and initialize the necessary constants (denoted by upper\sphinxhyphen{}case names):
our simulation domain will have \sphinxcode{\sphinxupquote{N=128}} cells as discretization points for the 1D velocity \(u\) in a periodic domain \(\Omega\) for the interval \([-1,1]\). We’ll use 32 time \sphinxcode{\sphinxupquote{STEPS}} for a time interval of 1, giving us \sphinxcode{\sphinxupquote{DT=1/32}}. Additionally, we’ll use a viscosity \sphinxcode{\sphinxupquote{NU}} of \(\nu=0.01/\pi\).

We’ll also define an initial state given by \(-\text{sin}(\pi x)\) in the numpy array \sphinxcode{\sphinxupquote{INITIAL\_NUMPY}}, which we’ll use to initialize the velocity \(u\) in the simulation in the next cell. This initialization will produce a nice shock in the center of our domain.

Phiflow is object\sphinxhyphen{}oriented and centered around field data in the form of grids (internally represented by a tensor object). I.e. you assemble your simulation by constructing a number of grids, and updating them over the course of time steps.

Phiflow internally works with tensors that have named dimensions. This will be especially handy later on for 2D simulations with additional batch and channel dimensions, but for now we’ll simply convert the 1D array into a phiflow tensor that has a single spatial dimension \sphinxcode{\sphinxupquote{'x'}}.

\begin{sphinxVerbatim}[commandchars=\\\{\}]
\PYG{n}{N} \PYG{o}{=} \PYG{l+m+mi}{128}
\PYG{n}{STEPS} \PYG{o}{=} \PYG{l+m+mi}{32}
\PYG{n}{DT} \PYG{o}{=} \PYG{l+m+mf}{1.}\PYG{o}{/}\PYG{n}{STEPS}
\PYG{n}{NU} \PYG{o}{=} \PYG{l+m+mf}{0.01}\PYG{o}{/}\PYG{n}{np}\PYG{o}{.}\PYG{n}{pi}

\PYG{c+c1}{\PYGZsh{} initialization of velocities}
\PYG{n}{INITIAL\PYGZus{}NUMPY} \PYG{o}{=} \PYG{n}{np}\PYG{o}{.}\PYG{n}{asarray}\PYG{p}{(} \PYG{p}{[}\PYG{o}{\PYGZhy{}}\PYG{n}{np}\PYG{o}{.}\PYG{n}{sin}\PYG{p}{(}\PYG{n}{np}\PYG{o}{.}\PYG{n}{pi} \PYG{o}{*} \PYG{n}{x}\PYG{p}{)} \PYG{o}{*} \PYG{l+m+mf}{1.} \PYG{k}{for} \PYG{n}{x} \PYG{o+ow}{in} \PYG{n}{np}\PYG{o}{.}\PYG{n}{linspace}\PYG{p}{(}\PYG{o}{\PYGZhy{}}\PYG{l+m+mi}{1}\PYG{p}{,}\PYG{l+m+mi}{1}\PYG{p}{,}\PYG{n}{N}\PYG{p}{)}\PYG{p}{]} \PYG{p}{)} \PYG{c+c1}{\PYGZsh{} 1D numpy array}

\PYG{n}{INITIAL} \PYG{o}{=} \PYG{n}{math}\PYG{o}{.}\PYG{n}{tensor}\PYG{p}{(}\PYG{n}{INITIAL\PYGZus{}NUMPY}\PYG{p}{,} \PYG{n}{spatial}\PYG{p}{(}\PYG{l+s+s1}{\PYGZsq{}}\PYG{l+s+s1}{x}\PYG{l+s+s1}{\PYGZsq{}}\PYG{p}{)} \PYG{p}{)} \PYG{c+c1}{\PYGZsh{} convert to phiflow tensor}
\end{sphinxVerbatim}

Next, we initialize a 1D \sphinxcode{\sphinxupquote{velocity}} grid from the \sphinxcode{\sphinxupquote{INITIAL}} numpy array that was converted into a tensor.
The extent of our domain \(\Omega\) is specifiied via the \sphinxcode{\sphinxupquote{bounds}} parameter \([-1,1]\), and the grid uses periodic boundary conditions (\sphinxcode{\sphinxupquote{extrapolation.PERIODIC}}). These two properties are the main difference between a tensor and a grid: the latter has boundary conditions and a physical extent.

Just to illustrate, we’ll also print some info about the velocity object: it’s a \sphinxcode{\sphinxupquote{phi.math}} tensor with a size of 128. Note that the actual grid content is contained in the \sphinxcode{\sphinxupquote{values}} of the grid. Below we’re printing five entries by using the \sphinxcode{\sphinxupquote{numpy()}} function to convert the content of the phiflow tensor into a numpy array. For tensors with more dimensions, we’d need to specify the order here, e.g., \sphinxcode{\sphinxupquote{'y,x,vector'}} for a 2D velocity field. (If we’d call \sphinxcode{\sphinxupquote{numpy('x,vector')}} below, this would convert the 1D array into one with an additional dimension for each 1D velocity component.)

\begin{sphinxVerbatim}[commandchars=\\\{\}]
\PYG{n}{velocity} \PYG{o}{=} \PYG{n}{CenteredGrid}\PYG{p}{(}\PYG{n}{INITIAL}\PYG{p}{,} \PYG{n}{extrapolation}\PYG{o}{.}\PYG{n}{PERIODIC}\PYG{p}{,} \PYG{n}{x}\PYG{o}{=}\PYG{n}{N}\PYG{p}{,} \PYG{n}{bounds}\PYG{o}{=}\PYG{n}{Box}\PYG{p}{[}\PYG{o}{\PYGZhy{}}\PYG{l+m+mi}{1}\PYG{p}{:}\PYG{l+m+mi}{1}\PYG{p}{]}\PYG{p}{)}
\PYG{c+c1}{\PYGZsh{}velocity = CenteredGrid(Noise(), extrapolation.PERIODIC, x=N, bounds=Box[\PYGZhy{}1:1]) \PYGZsh{} random init}

\PYG{n+nb}{print}\PYG{p}{(}\PYG{l+s+s2}{\PYGZdq{}}\PYG{l+s+s2}{Velocity tensor shape: }\PYG{l+s+s2}{\PYGZdq{}}   \PYG{o}{+} \PYG{n+nb}{format}\PYG{p}{(} \PYG{n}{velocity}\PYG{o}{.}\PYG{n}{shape} \PYG{p}{)}\PYG{p}{)} \PYG{c+c1}{\PYGZsh{} == velocity.values.shape}
\PYG{n+nb}{print}\PYG{p}{(}\PYG{l+s+s2}{\PYGZdq{}}\PYG{l+s+s2}{Velocity tensor type: }\PYG{l+s+s2}{\PYGZdq{}}    \PYG{o}{+} \PYG{n+nb}{format}\PYG{p}{(} \PYG{n+nb}{type}\PYG{p}{(}\PYG{n}{velocity}\PYG{o}{.}\PYG{n}{values}\PYG{p}{)} \PYG{p}{)}\PYG{p}{)}
\PYG{n+nb}{print}\PYG{p}{(}\PYG{l+s+s2}{\PYGZdq{}}\PYG{l+s+s2}{Velocity tensor entries 10 to 14: }\PYG{l+s+s2}{\PYGZdq{}} \PYG{o}{+} \PYG{n+nb}{format}\PYG{p}{(} \PYG{n}{velocity}\PYG{o}{.}\PYG{n}{values}\PYG{o}{.}\PYG{n}{numpy}\PYG{p}{(}\PYG{p}{)}\PYG{p}{[}\PYG{l+m+mi}{10}\PYG{p}{:}\PYG{l+m+mi}{15}\PYG{p}{]} \PYG{p}{)}\PYG{p}{)}
\end{sphinxVerbatim}

\begin{sphinxVerbatim}[commandchars=\\\{\}]
Velocity tensor shape: (xˢ=128)
Velocity tensor type: \PYGZlt{}class \PYGZsq{}phi.math.\PYGZus{}tensors.CollapsedTensor\PYGZsq{}\PYGZgt{}
Velocity tensor entries 10 to 14: [0.47480196 0.51774486 0.55942075 0.59972764 0.6385669 ]
\end{sphinxVerbatim}

\subsection{Running the simulation}
\label{\detokenize{overview-burgers-forw:running-the-simulation}}
Now we’re ready to run the simulation itself. To ccompute the diffusion and advection components of our model equation we can simply call the existing \sphinxcode{\sphinxupquote{diffusion}} and \sphinxcode{\sphinxupquote{semi\_lagrangian}} operators in phiflow: \sphinxcode{\sphinxupquote{diffuse.explicit(u,...)}} computes an explicit diffusion step via central differences for the term \(\nu \nabla\cdot \nabla u\) of our model. Next, \sphinxcode{\sphinxupquote{advect.semi\_lagrangian(f,u)}} is used for a stable first\sphinxhyphen{}order approximation of the transport of an arbitrary field \sphinxcode{\sphinxupquote{f}} by a velocity \sphinxcode{\sphinxupquote{u}}. In our model we have \(\partial u / \partial{t} + u \nabla f\), hence we use the \sphinxcode{\sphinxupquote{semi\_lagrangian}} function to transport the velocity with itself in the implementation:

\begin{sphinxVerbatim}[commandchars=\\\{\}]
\PYG{n}{velocities} \PYG{o}{=} \PYG{p}{[}\PYG{n}{velocity}\PYG{p}{]}
\PYG{n}{age} \PYG{o}{=} \PYG{l+m+mf}{0.}
\PYG{k}{for} \PYG{n}{i} \PYG{o+ow}{in} \PYG{n+nb}{range}\PYG{p}{(}\PYG{n}{STEPS}\PYG{p}{)}\PYG{p}{:}
    \PYG{n}{v1} \PYG{o}{=} \PYG{n}{diffuse}\PYG{o}{.}\PYG{n}{explicit}\PYG{p}{(}\PYG{n}{velocities}\PYG{p}{[}\PYG{o}{\PYGZhy{}}\PYG{l+m+mi}{1}\PYG{p}{]}\PYG{p}{,} \PYG{n}{NU}\PYG{p}{,} \PYG{n}{DT}\PYG{p}{)}
    \PYG{n}{v2} \PYG{o}{=} \PYG{n}{advect}\PYG{o}{.}\PYG{n}{semi\PYGZus{}lagrangian}\PYG{p}{(}\PYG{n}{v1}\PYG{p}{,} \PYG{n}{v1}\PYG{p}{,} \PYG{n}{DT}\PYG{p}{)}
    \PYG{n}{age} \PYG{o}{+}\PYG{o}{=} \PYG{n}{DT}
    \PYG{n}{velocities}\PYG{o}{.}\PYG{n}{append}\PYG{p}{(}\PYG{n}{v2}\PYG{p}{)}

\PYG{n+nb}{print}\PYG{p}{(}\PYG{l+s+s2}{\PYGZdq{}}\PYG{l+s+s2}{New velocity content at t=}\PYG{l+s+si}{\PYGZob{}\PYGZcb{}}\PYG{l+s+s2}{: }\PYG{l+s+si}{\PYGZob{}\PYGZcb{}}\PYG{l+s+s2}{\PYGZdq{}}\PYG{o}{.}\PYG{n}{format}\PYG{p}{(} \PYG{n}{age}\PYG{p}{,} \PYG{n}{velocities}\PYG{p}{[}\PYG{o}{\PYGZhy{}}\PYG{l+m+mi}{1}\PYG{p}{]}\PYG{o}{.}\PYG{n}{values}\PYG{o}{.}\PYG{n}{numpy}\PYG{p}{(}\PYG{p}{)}\PYG{p}{[}\PYG{l+m+mi}{0}\PYG{p}{:}\PYG{l+m+mi}{5}\PYG{p}{]} \PYG{p}{)}\PYG{p}{)}
\end{sphinxVerbatim}

\begin{sphinxVerbatim}[commandchars=\\\{\}]
New velocity content at t=1.0: [0.00274862 0.01272991 0.02360343 0.03478042 0.0460869 ]
\end{sphinxVerbatim}

Here we’re actually collecting all time steps in the list \sphinxcode{\sphinxupquote{velocities}}. This is not necessary in general (and could consume lots of memory for long\sphinxhyphen{}running sims), but useful here to plot the evolution of the velocity states later on.

The print statements print a few of the velocity entries, and already show that something is happening in our simulation, but it’s difficult to get an intuition for the behavior of the PDE just from these numbers. Hence, let’s visualize the states over time to show what is happening.

\subsection{Visualization}
\label{\detokenize{overview-burgers-forw:visualization}}
We can visualize this 1D case easily in a graph: the following code shows the initial state in blue, and then times \(10/32, 20/32, 1\) in green, cyan and purple.

\begin{sphinxVerbatim}[commandchars=\\\{\}]
\PYG{c+c1}{\PYGZsh{} get \PYGZdq{}velocity.values\PYGZdq{} from each phiflow state with a channel dimensions, i.e. \PYGZdq{}vector\PYGZdq{}}
\PYG{n}{vels} \PYG{o}{=} \PYG{p}{[}\PYG{n}{v}\PYG{o}{.}\PYG{n}{values}\PYG{o}{.}\PYG{n}{numpy}\PYG{p}{(}\PYG{l+s+s1}{\PYGZsq{}}\PYG{l+s+s1}{x,vector}\PYG{l+s+s1}{\PYGZsq{}}\PYG{p}{)} \PYG{k}{for} \PYG{n}{v} \PYG{o+ow}{in} \PYG{n}{velocities}\PYG{p}{]} \PYG{c+c1}{\PYGZsh{} gives a list of 2D arrays }

\PYG{k+kn}{import} \PYG{n+nn}{pylab}

\PYG{n}{fig} \PYG{o}{=} \PYG{n}{pylab}\PYG{o}{.}\PYG{n}{figure}\PYG{p}{(}\PYG{p}{)}\PYG{o}{.}\PYG{n}{gca}\PYG{p}{(}\PYG{p}{)}
\PYG{n}{fig}\PYG{o}{.}\PYG{n}{plot}\PYG{p}{(}\PYG{n}{np}\PYG{o}{.}\PYG{n}{linspace}\PYG{p}{(}\PYG{o}{\PYGZhy{}}\PYG{l+m+mi}{1}\PYG{p}{,}\PYG{l+m+mi}{1}\PYG{p}{,}\PYG{n+nb}{len}\PYG{p}{(}\PYG{n}{vels}\PYG{p}{[} \PYG{l+m+mi}{0}\PYG{p}{]}\PYG{o}{.}\PYG{n}{flatten}\PYG{p}{(}\PYG{p}{)}\PYG{p}{)}\PYG{p}{)}\PYG{p}{,} \PYG{n}{vels}\PYG{p}{[} \PYG{l+m+mi}{0}\PYG{p}{]}\PYG{o}{.}\PYG{n}{flatten}\PYG{p}{(}\PYG{p}{)}\PYG{p}{,} \PYG{n}{lw}\PYG{o}{=}\PYG{l+m+mi}{2}\PYG{p}{,} \PYG{n}{color}\PYG{o}{=}\PYG{l+s+s1}{\PYGZsq{}}\PYG{l+s+s1}{blue}\PYG{l+s+s1}{\PYGZsq{}}\PYG{p}{,}  \PYG{n}{label}\PYG{o}{=}\PYG{l+s+s2}{\PYGZdq{}}\PYG{l+s+s2}{t=0}\PYG{l+s+s2}{\PYGZdq{}}\PYG{p}{)}
\PYG{n}{fig}\PYG{o}{.}\PYG{n}{plot}\PYG{p}{(}\PYG{n}{np}\PYG{o}{.}\PYG{n}{linspace}\PYG{p}{(}\PYG{o}{\PYGZhy{}}\PYG{l+m+mi}{1}\PYG{p}{,}\PYG{l+m+mi}{1}\PYG{p}{,}\PYG{n+nb}{len}\PYG{p}{(}\PYG{n}{vels}\PYG{p}{[}\PYG{l+m+mi}{10}\PYG{p}{]}\PYG{o}{.}\PYG{n}{flatten}\PYG{p}{(}\PYG{p}{)}\PYG{p}{)}\PYG{p}{)}\PYG{p}{,} \PYG{n}{vels}\PYG{p}{[}\PYG{l+m+mi}{10}\PYG{p}{]}\PYG{o}{.}\PYG{n}{flatten}\PYG{p}{(}\PYG{p}{)}\PYG{p}{,} \PYG{n}{lw}\PYG{o}{=}\PYG{l+m+mi}{2}\PYG{p}{,} \PYG{n}{color}\PYG{o}{=}\PYG{l+s+s1}{\PYGZsq{}}\PYG{l+s+s1}{green}\PYG{l+s+s1}{\PYGZsq{}}\PYG{p}{,} \PYG{n}{label}\PYG{o}{=}\PYG{l+s+s2}{\PYGZdq{}}\PYG{l+s+s2}{t=0.3125}\PYG{l+s+s2}{\PYGZdq{}}\PYG{p}{)}
\PYG{n}{fig}\PYG{o}{.}\PYG{n}{plot}\PYG{p}{(}\PYG{n}{np}\PYG{o}{.}\PYG{n}{linspace}\PYG{p}{(}\PYG{o}{\PYGZhy{}}\PYG{l+m+mi}{1}\PYG{p}{,}\PYG{l+m+mi}{1}\PYG{p}{,}\PYG{n+nb}{len}\PYG{p}{(}\PYG{n}{vels}\PYG{p}{[}\PYG{l+m+mi}{20}\PYG{p}{]}\PYG{o}{.}\PYG{n}{flatten}\PYG{p}{(}\PYG{p}{)}\PYG{p}{)}\PYG{p}{)}\PYG{p}{,} \PYG{n}{vels}\PYG{p}{[}\PYG{l+m+mi}{20}\PYG{p}{]}\PYG{o}{.}\PYG{n}{flatten}\PYG{p}{(}\PYG{p}{)}\PYG{p}{,} \PYG{n}{lw}\PYG{o}{=}\PYG{l+m+mi}{2}\PYG{p}{,} \PYG{n}{color}\PYG{o}{=}\PYG{l+s+s1}{\PYGZsq{}}\PYG{l+s+s1}{cyan}\PYG{l+s+s1}{\PYGZsq{}}\PYG{p}{,}  \PYG{n}{label}\PYG{o}{=}\PYG{l+s+s2}{\PYGZdq{}}\PYG{l+s+s2}{t=0.625}\PYG{l+s+s2}{\PYGZdq{}}\PYG{p}{)}
\PYG{n}{fig}\PYG{o}{.}\PYG{n}{plot}\PYG{p}{(}\PYG{n}{np}\PYG{o}{.}\PYG{n}{linspace}\PYG{p}{(}\PYG{o}{\PYGZhy{}}\PYG{l+m+mi}{1}\PYG{p}{,}\PYG{l+m+mi}{1}\PYG{p}{,}\PYG{n+nb}{len}\PYG{p}{(}\PYG{n}{vels}\PYG{p}{[}\PYG{l+m+mi}{32}\PYG{p}{]}\PYG{o}{.}\PYG{n}{flatten}\PYG{p}{(}\PYG{p}{)}\PYG{p}{)}\PYG{p}{)}\PYG{p}{,} \PYG{n}{vels}\PYG{p}{[}\PYG{l+m+mi}{32}\PYG{p}{]}\PYG{o}{.}\PYG{n}{flatten}\PYG{p}{(}\PYG{p}{)}\PYG{p}{,} \PYG{n}{lw}\PYG{o}{=}\PYG{l+m+mi}{2}\PYG{p}{,} \PYG{n}{color}\PYG{o}{=}\PYG{l+s+s1}{\PYGZsq{}}\PYG{l+s+s1}{purple}\PYG{l+s+s1}{\PYGZsq{}}\PYG{p}{,}\PYG{n}{label}\PYG{o}{=}\PYG{l+s+s2}{\PYGZdq{}}\PYG{l+s+s2}{t=1}\PYG{l+s+s2}{\PYGZdq{}}\PYG{p}{)}
\PYG{n}{pylab}\PYG{o}{.}\PYG{n}{xlabel}\PYG{p}{(}\PYG{l+s+s1}{\PYGZsq{}}\PYG{l+s+s1}{x}\PYG{l+s+s1}{\PYGZsq{}}\PYG{p}{)}\PYG{p}{;} \PYG{n}{pylab}\PYG{o}{.}\PYG{n}{ylabel}\PYG{p}{(}\PYG{l+s+s1}{\PYGZsq{}}\PYG{l+s+s1}{u}\PYG{l+s+s1}{\PYGZsq{}}\PYG{p}{)}\PYG{p}{;} \PYG{n}{pylab}\PYG{o}{.}\PYG{n}{legend}\PYG{p}{(}\PYG{p}{)}
\end{sphinxVerbatim}

\begin{sphinxVerbatim}[commandchars=\\\{\}]
\PYGZlt{}matplotlib.legend.Legend at 0x7f9ca2d4ef70\PYGZgt{}
\end{sphinxVerbatim}

\noindent\sphinxincludegraphics{{overview-burgers-forw_10_1}.png}

This nicely shows the shock developing in the center of our domain, which forms from the collision of the two initial velocity “bumps”, the positive one on left (moving right) and the negative one right of the center (moving left).

As these lines can overlap quite a bit we’ll also use a different visualization in the following chapters that shows the evolution over the course of all time steps in a 2D image. Our 1D domain will be shown along the Y\sphinxhyphen{}axis, and each point along X will represent one time step.

The code below converts our collection of velocity states into a 2D array, repeating individual time steps 8 times to make the image a bit wider. This is purely optional, of course, but makes it easier to see what’s happening in our Burgers simulation.

\begin{sphinxVerbatim}[commandchars=\\\{\}]
\PYG{k}{def} \PYG{n+nf}{show\PYGZus{}state}\PYG{p}{(}\PYG{n}{a}\PYG{p}{,} \PYG{n}{title}\PYG{p}{)}\PYG{p}{:}
    \PYG{c+c1}{\PYGZsh{} we only have 33 time steps, blow up by a factor of 2\PYGZca{}4 to make it easier to see}
    \PYG{c+c1}{\PYGZsh{} (could also be done with more evaluations of network)}
    \PYG{n}{a}\PYG{o}{=}\PYG{n}{np}\PYG{o}{.}\PYG{n}{expand\PYGZus{}dims}\PYG{p}{(}\PYG{n}{a}\PYG{p}{,} \PYG{n}{axis}\PYG{o}{=}\PYG{l+m+mi}{2}\PYG{p}{)}
    \PYG{k}{for} \PYG{n}{i} \PYG{o+ow}{in} \PYG{n+nb}{range}\PYG{p}{(}\PYG{l+m+mi}{4}\PYG{p}{)}\PYG{p}{:}
        \PYG{n}{a} \PYG{o}{=} \PYG{n}{np}\PYG{o}{.}\PYG{n}{concatenate}\PYG{p}{(} \PYG{p}{[}\PYG{n}{a}\PYG{p}{,}\PYG{n}{a}\PYG{p}{]} \PYG{p}{,} \PYG{n}{axis}\PYG{o}{=}\PYG{l+m+mi}{2}\PYG{p}{)}

    \PYG{n}{a} \PYG{o}{=} \PYG{n}{np}\PYG{o}{.}\PYG{n}{reshape}\PYG{p}{(} \PYG{n}{a}\PYG{p}{,} \PYG{p}{[}\PYG{n}{a}\PYG{o}{.}\PYG{n}{shape}\PYG{p}{[}\PYG{l+m+mi}{0}\PYG{p}{]}\PYG{p}{,}\PYG{n}{a}\PYG{o}{.}\PYG{n}{shape}\PYG{p}{[}\PYG{l+m+mi}{1}\PYG{p}{]}\PYG{o}{*}\PYG{n}{a}\PYG{o}{.}\PYG{n}{shape}\PYG{p}{[}\PYG{l+m+mi}{2}\PYG{p}{]}\PYG{p}{]} \PYG{p}{)}
    \PYG{c+c1}{\PYGZsh{}print(\PYGZdq{}Resulting image size\PYGZdq{} +format(a.shape))}

    \PYG{n}{fig}\PYG{p}{,} \PYG{n}{axes} \PYG{o}{=} \PYG{n}{pylab}\PYG{o}{.}\PYG{n}{subplots}\PYG{p}{(}\PYG{l+m+mi}{1}\PYG{p}{,} \PYG{l+m+mi}{1}\PYG{p}{,} \PYG{n}{figsize}\PYG{o}{=}\PYG{p}{(}\PYG{l+m+mi}{16}\PYG{p}{,} \PYG{l+m+mi}{5}\PYG{p}{)}\PYG{p}{)}
    \PYG{n}{im} \PYG{o}{=} \PYG{n}{axes}\PYG{o}{.}\PYG{n}{imshow}\PYG{p}{(}\PYG{n}{a}\PYG{p}{,} \PYG{n}{origin}\PYG{o}{=}\PYG{l+s+s1}{\PYGZsq{}}\PYG{l+s+s1}{upper}\PYG{l+s+s1}{\PYGZsq{}}\PYG{p}{,} \PYG{n}{cmap}\PYG{o}{=}\PYG{l+s+s1}{\PYGZsq{}}\PYG{l+s+s1}{inferno}\PYG{l+s+s1}{\PYGZsq{}}\PYG{p}{)}
    \PYG{n}{pylab}\PYG{o}{.}\PYG{n}{colorbar}\PYG{p}{(}\PYG{n}{im}\PYG{p}{)} \PYG{p}{;} \PYG{n}{pylab}\PYG{o}{.}\PYG{n}{xlabel}\PYG{p}{(}\PYG{l+s+s1}{\PYGZsq{}}\PYG{l+s+s1}{time}\PYG{l+s+s1}{\PYGZsq{}}\PYG{p}{)}\PYG{p}{;} \PYG{n}{pylab}\PYG{o}{.}\PYG{n}{ylabel}\PYG{p}{(}\PYG{l+s+s1}{\PYGZsq{}}\PYG{l+s+s1}{x}\PYG{l+s+s1}{\PYGZsq{}}\PYG{p}{)}\PYG{p}{;} \PYG{n}{pylab}\PYG{o}{.}\PYG{n}{title}\PYG{p}{(}\PYG{n}{title}\PYG{p}{)}
        
\PYG{n}{vels\PYGZus{}img} \PYG{o}{=} \PYG{n}{np}\PYG{o}{.}\PYG{n}{asarray}\PYG{p}{(} \PYG{n}{np}\PYG{o}{.}\PYG{n}{concatenate}\PYG{p}{(}\PYG{n}{vels}\PYG{p}{,} \PYG{n}{axis}\PYG{o}{=}\PYG{o}{\PYGZhy{}}\PYG{l+m+mi}{1}\PYG{p}{)}\PYG{p}{,} \PYG{n}{dtype}\PYG{o}{=}\PYG{n}{np}\PYG{o}{.}\PYG{n}{float32} \PYG{p}{)} 

\PYG{c+c1}{\PYGZsh{} save for comparison with reconstructions later on}
\PYG{k+kn}{import} \PYG{n+nn}{os}\PYG{p}{;} \PYG{n}{os}\PYG{o}{.}\PYG{n}{makedirs}\PYG{p}{(}\PYG{l+s+s2}{\PYGZdq{}}\PYG{l+s+s2}{./temp}\PYG{l+s+s2}{\PYGZdq{}}\PYG{p}{,}\PYG{n}{exist\PYGZus{}ok}\PYG{o}{=}\PYG{k+kc}{True}\PYG{p}{)}
\PYG{n}{np}\PYG{o}{.}\PYG{n}{savez\PYGZus{}compressed}\PYG{p}{(}\PYG{l+s+s2}{\PYGZdq{}}\PYG{l+s+s2}{./temp/burgers\PYGZhy{}groundtruth\PYGZhy{}solution.npz}\PYG{l+s+s2}{\PYGZdq{}}\PYG{p}{,} \PYG{n}{np}\PYG{o}{.}\PYG{n}{reshape}\PYG{p}{(}\PYG{n}{vels\PYGZus{}img}\PYG{p}{,}\PYG{p}{[}\PYG{n}{N}\PYG{p}{,}\PYG{n}{STEPS}\PYG{o}{+}\PYG{l+m+mi}{1}\PYG{p}{]}\PYG{p}{)}\PYG{p}{)} \PYG{c+c1}{\PYGZsh{} remove batch \PYGZam{} channel dimension}

\PYG{n}{show\PYGZus{}state}\PYG{p}{(}\PYG{n}{vels\PYGZus{}img}\PYG{p}{,} \PYG{l+s+s2}{\PYGZdq{}}\PYG{l+s+s2}{Velocity}\PYG{l+s+s2}{\PYGZdq{}}\PYG{p}{)}
\end{sphinxVerbatim}

\noindent\sphinxincludegraphics{{overview-burgers-forw_12_0}.png}

This concludes a first simulation in phiflow. It’s not overly complex, but because of that it’s a good starting point for evaluating and comparing different physics\sphinxhyphen{}based deep learning approaches in the next chapter. But before that, we’ll target a more complex simulation type in the next section.

\subsection{Next steps}
\label{\detokenize{overview-burgers-forw:next-steps}}
Some things to try based on this simulation setup:
\begin{itemize}
\item {} 
Feel free to experiment \sphinxhyphen{} the setup above is very simple, you can change the simulation parameters, or the initialization. E.g., you can use a noise field via \sphinxcode{\sphinxupquote{Noise()}} to get more chaotic results (cf. the comment in the \sphinxcode{\sphinxupquote{velocity}} cell above).

\item {} 
A bit more complicated: extend the simulation to 2D (or higher). This will require changes throughout, but all operators above support higher dimensions. Before trying this, you probably will want to check out the next example, which covers a 2D Navier\sphinxhyphen{}Stokes case.

\end{itemize}

\section{Navier\sphinxhyphen{}Stokes Forward Simulation}
\label{\detokenize{overview-ns-forw:navier-stokes-forward-simulation}}\label{\detokenize{overview-ns-forw::doc}}
Now let’s target a somewhat more complex example: a fluid simulation based on the Navier\sphinxhyphen{}Stokes equations. This is still very simple with ΦFlow (phiflow), as differentiable operators for all steps exist there. The Navier\sphinxhyphen{}Stokes equations (in their incompressible form) introduce an additional pressure field \(p\), and a constraint for conservation of mass, as introduced in equation \eqref{equation:overview-equations:model-boussinesq2d}. We’re also moving a marker field, denoted by \(d\) here, with the flow. It indicates regions of higher temperature, and exerts a force via a buouyancy factor \(\xi\):
\begin{equation*}
\begin{split}\begin{aligned}
    \frac{\partial \mathbf{u}}{\partial{t}} + \mathbf{u} \cdot \nabla \mathbf{u} &= - \frac{1}{\rho} \nabla p + \nu \nabla\cdot \nabla \mathbf{u} + (0,1)^T \xi d
  \quad \text{s.t.} \quad \nabla \cdot \mathbf{u} = 0,
  \\
  \frac{\partial d}{\partial{t}} + \mathbf{u} \cdot \nabla d &= 0 
\end{aligned}\end{split}
\end{equation*}
Here, we’re aiming for an incompressible flow (i.e., \(\rho = \text{const}\)), and use a simple buoyancy model (the Boussinesq approximation) via the term \((0,1)^T \xi d\). This models changes in density without explicitly calculating \(\rho\), and we assume a gravity force that acts along the y direction via the vector \((0,1)^T\).
We’ll solve this PDE on a closed domain with Dirichlet boundary conditions \(\mathbf{u}=0\) for the velocity, and Neumann boundaries \(\frac{\partial p}{\partial x}=0\) for pressure, on a domain \(\Omega\) with a physical size of \(100 \times 80\) units.
\sphinxhref{https://colab.research.google.com/github/tum-pbs/pbdl-book/blob/main/overview-ns-forw.ipynb}{{[}run in colab{]}}

\subsection{Implementation}
\label{\detokenize{overview-ns-forw:implementation}}
As in the previous section, the first command with a “!” prefix installs the \sphinxhref{https://github.com/tum-pbs/PhiFlow}{phiflow python package from GitHub} via \sphinxcode{\sphinxupquote{pip}} in your python environment. (Skip or modify this command if necessary.)

\begin{sphinxVerbatim}[commandchars=\\\{\}]
\PYG{c+ch}{\PYGZsh{}!pip install \PYGZhy{}\PYGZhy{}upgrade \PYGZhy{}\PYGZhy{}quiet phiflow }
\PYG{o}{!}pip install \PYGZhy{}\PYGZhy{}upgrade \PYGZhy{}\PYGZhy{}quiet git+https://github.com/tum\PYGZhy{}pbs/PhiFlow@develop

\PYG{k+kn}{from} \PYG{n+nn}{phi}\PYG{n+nn}{.}\PYG{n+nn}{flow} \PYG{k+kn}{import} \PYG{o}{*}  \PYG{c+c1}{\PYGZsh{} The Dash GUI is not supported on Google colab, ignore the warning}
\PYG{k+kn}{import} \PYG{n+nn}{pylab}
\end{sphinxVerbatim}

\subsection{Setting up the simulation}
\label{\detokenize{overview-ns-forw:setting-up-the-simulation}}
The following code sets up a few constants, which are denoted by upper case names. We’ll use \(40\times32\) cells to discretize our domain, introduce a slight viscosity via \(\nu\), and define the time step to be \(\Delta t=1.5\).

We’re creating a first \sphinxcode{\sphinxupquote{CenteredGrid}} here, which is initialized by a \sphinxcode{\sphinxupquote{Sphere}} geometry object. This will represent the inflow region \sphinxcode{\sphinxupquote{INFLOW}} where hot smoke is generated.

\begin{sphinxVerbatim}[commandchars=\\\{\}]
\PYG{n}{DT} \PYG{o}{=} \PYG{l+m+mf}{1.5}
\PYG{n}{NU} \PYG{o}{=} \PYG{l+m+mf}{0.01}

\PYG{n}{INFLOW} \PYG{o}{=} \PYG{n}{CenteredGrid}\PYG{p}{(}\PYG{n}{Sphere}\PYG{p}{(}\PYG{n}{center}\PYG{o}{=}\PYG{p}{(}\PYG{l+m+mi}{30}\PYG{p}{,}\PYG{l+m+mi}{15}\PYG{p}{)}\PYG{p}{,} \PYG{n}{radius}\PYG{o}{=}\PYG{l+m+mi}{10}\PYG{p}{)}\PYG{p}{,} \PYG{n}{extrapolation}\PYG{o}{.}\PYG{n}{BOUNDARY}\PYG{p}{,} \PYG{n}{x}\PYG{o}{=}\PYG{l+m+mi}{32}\PYG{p}{,} \PYG{n}{y}\PYG{o}{=}\PYG{l+m+mi}{40}\PYG{p}{,} \PYG{n}{bounds}\PYG{o}{=}\PYG{n}{Box}\PYG{p}{[}\PYG{l+m+mi}{0}\PYG{p}{:}\PYG{l+m+mi}{80}\PYG{p}{,} \PYG{l+m+mi}{0}\PYG{p}{:}\PYG{l+m+mi}{100}\PYG{p}{]}\PYG{p}{)} \PYG{o}{*} \PYG{l+m+mf}{0.2}
\end{sphinxVerbatim}

The inflow will be used to inject smoke into a second centered grid \sphinxcode{\sphinxupquote{smoke}} that represents the marker field \(d\) from above. Note that we’ve defined a \sphinxcode{\sphinxupquote{Box}} of size \(100x80\) above. This is the physical scale in terms of spatial units in our simulation, i.e., a velocity of magnitude \(1\) will move the smoke density by 1 unit per 1 time unit, which may be larger or smaller than a cell in the discretized grid, depending on the settings for \sphinxcode{\sphinxupquote{x,y}}. You could parametrize your simulation grid to directly resemble real\sphinxhyphen{}world units, or keep appropriate conversion factors in mind.

The inflow sphere above is already using the “world” coordinates: it is located at \(x=30\) along the first axis, and \(y=15\) (within the \(100x80\) domain box).

Next, we create grids for the quantities we want to simulate. For this example, we require a velocity field and a smoke density field.

\begin{sphinxVerbatim}[commandchars=\\\{\}]
\PYG{n}{smoke} \PYG{o}{=} \PYG{n}{CenteredGrid}\PYG{p}{(}\PYG{l+m+mi}{0}\PYG{p}{,} \PYG{n}{extrapolation}\PYG{o}{.}\PYG{n}{BOUNDARY}\PYG{p}{,} \PYG{n}{x}\PYG{o}{=}\PYG{l+m+mi}{32}\PYG{p}{,} \PYG{n}{y}\PYG{o}{=}\PYG{l+m+mi}{40}\PYG{p}{,} \PYG{n}{bounds}\PYG{o}{=}\PYG{n}{Box}\PYG{p}{[}\PYG{l+m+mi}{0}\PYG{p}{:}\PYG{l+m+mi}{80}\PYG{p}{,} \PYG{l+m+mi}{0}\PYG{p}{:}\PYG{l+m+mi}{100}\PYG{p}{]}\PYG{p}{)}  \PYG{c+c1}{\PYGZsh{} sampled at cell centers}
\PYG{n}{velocity} \PYG{o}{=} \PYG{n}{StaggeredGrid}\PYG{p}{(}\PYG{l+m+mi}{0}\PYG{p}{,} \PYG{n}{extrapolation}\PYG{o}{.}\PYG{n}{ZERO}\PYG{p}{,} \PYG{n}{x}\PYG{o}{=}\PYG{l+m+mi}{32}\PYG{p}{,} \PYG{n}{y}\PYG{o}{=}\PYG{l+m+mi}{40}\PYG{p}{,} \PYG{n}{bounds}\PYG{o}{=}\PYG{n}{Box}\PYG{p}{[}\PYG{l+m+mi}{0}\PYG{p}{:}\PYG{l+m+mi}{80}\PYG{p}{,} \PYG{l+m+mi}{0}\PYG{p}{:}\PYG{l+m+mi}{100}\PYG{p}{]}\PYG{p}{)}  \PYG{c+c1}{\PYGZsh{} sampled in staggered form at face centers }
\end{sphinxVerbatim}

We sample the smoke field at the cell centers and the velocity in \sphinxhref{https://tum-pbs.github.io/PhiFlow/Staggered\_Grids.html}{staggered form}. The staggered grid internally contains 2 centered grids with different dimensions, and can be converted into centered grids (or simply numpy arrays) via the \sphinxcode{\sphinxupquote{unstack}} function, as explained in the link above.

Next we define the update step of the simulation, which calls the necessary functions to advance the state of our fluid system by \sphinxcode{\sphinxupquote{dt}}. The next cell computes one such step, and plots the marker density after one simulation frame.

\begin{sphinxVerbatim}[commandchars=\\\{\}]
\PYG{k}{def} \PYG{n+nf}{step}\PYG{p}{(}\PYG{n}{velocity}\PYG{p}{,} \PYG{n}{smoke}\PYG{p}{,} \PYG{n}{pressure}\PYG{p}{,} \PYG{n}{dt}\PYG{o}{=}\PYG{l+m+mf}{1.0}\PYG{p}{,} \PYG{n}{buoyancy\PYGZus{}factor}\PYG{o}{=}\PYG{l+m+mf}{1.0}\PYG{p}{)}\PYG{p}{:}
    \PYG{n}{smoke} \PYG{o}{=} \PYG{n}{advect}\PYG{o}{.}\PYG{n}{semi\PYGZus{}lagrangian}\PYG{p}{(}\PYG{n}{smoke}\PYG{p}{,} \PYG{n}{velocity}\PYG{p}{,} \PYG{n}{dt}\PYG{p}{)} \PYG{o}{+} \PYG{n}{INFLOW}
    \PYG{n}{buoyancy\PYGZus{}force} \PYG{o}{=} \PYG{n}{smoke} \PYG{o}{*} \PYG{p}{(}\PYG{l+m+mi}{0}\PYG{p}{,} \PYG{n}{buoyancy\PYGZus{}factor}\PYG{p}{)} \PYG{o}{\PYGZgt{}\PYGZgt{}} \PYG{n}{velocity}  \PYG{c+c1}{\PYGZsh{} resamples smoke to velocity sample points}
    \PYG{n}{velocity} \PYG{o}{=} \PYG{n}{advect}\PYG{o}{.}\PYG{n}{semi\PYGZus{}lagrangian}\PYG{p}{(}\PYG{n}{velocity}\PYG{p}{,} \PYG{n}{velocity}\PYG{p}{,} \PYG{n}{dt}\PYG{p}{)} \PYG{o}{+} \PYG{n}{dt} \PYG{o}{*} \PYG{n}{buoyancy\PYGZus{}force}
    \PYG{n}{velocity} \PYG{o}{=} \PYG{n}{diffuse}\PYG{o}{.}\PYG{n}{explicit}\PYG{p}{(}\PYG{n}{velocity}\PYG{p}{,} \PYG{n}{NU}\PYG{p}{,} \PYG{n}{dt}\PYG{p}{)}
    \PYG{n}{velocity}\PYG{p}{,} \PYG{n}{pressure} \PYG{o}{=} \PYG{n}{fluid}\PYG{o}{.}\PYG{n}{make\PYGZus{}incompressible}\PYG{p}{(}\PYG{n}{velocity}\PYG{p}{)}
    \PYG{k}{return} \PYG{n}{velocity}\PYG{p}{,} \PYG{n}{smoke}\PYG{p}{,} \PYG{n}{pressure}

\PYG{n}{velocity}\PYG{p}{,} \PYG{n}{smoke}\PYG{p}{,} \PYG{n}{pressure} \PYG{o}{=} \PYG{n}{step}\PYG{p}{(}\PYG{n}{velocity}\PYG{p}{,} \PYG{n}{smoke}\PYG{p}{,} \PYG{k+kc}{None}\PYG{p}{,} \PYG{n}{dt}\PYG{o}{=}\PYG{n}{DT}\PYG{p}{)}

\PYG{n+nb}{print}\PYG{p}{(}\PYG{l+s+s2}{\PYGZdq{}}\PYG{l+s+s2}{Max. velocity and mean marker density: }\PYG{l+s+s2}{\PYGZdq{}} \PYG{o}{+} \PYG{n+nb}{format}\PYG{p}{(} \PYG{p}{[} \PYG{n}{math}\PYG{o}{.}\PYG{n}{max}\PYG{p}{(}\PYG{n}{velocity}\PYG{o}{.}\PYG{n}{values}\PYG{p}{)} \PYG{p}{,} \PYG{n}{math}\PYG{o}{.}\PYG{n}{mean}\PYG{p}{(}\PYG{n}{smoke}\PYG{o}{.}\PYG{n}{values}\PYG{p}{)} \PYG{p}{]} \PYG{p}{)}\PYG{p}{)}

\PYG{n}{pylab}\PYG{o}{.}\PYG{n}{imshow}\PYG{p}{(}\PYG{n}{np}\PYG{o}{.}\PYG{n}{asarray}\PYG{p}{(}\PYG{n}{smoke}\PYG{o}{.}\PYG{n}{values}\PYG{o}{.}\PYG{n}{numpy}\PYG{p}{(}\PYG{l+s+s1}{\PYGZsq{}}\PYG{l+s+s1}{y,x}\PYG{l+s+s1}{\PYGZsq{}}\PYG{p}{)}\PYG{p}{)}\PYG{p}{,} \PYG{n}{origin}\PYG{o}{=}\PYG{l+s+s1}{\PYGZsq{}}\PYG{l+s+s1}{lower}\PYG{l+s+s1}{\PYGZsq{}}\PYG{p}{,} \PYG{n}{cmap}\PYG{o}{=}\PYG{l+s+s1}{\PYGZsq{}}\PYG{l+s+s1}{magma}\PYG{l+s+s1}{\PYGZsq{}}\PYG{p}{)}
\end{sphinxVerbatim}

\begin{sphinxVerbatim}[commandchars=\\\{\}]
Max. velocity and mean marker density: [0.15530995, 0.008125]
\end{sphinxVerbatim}

\begin{sphinxVerbatim}[commandchars=\\\{\}]
\PYGZlt{}matplotlib.image.AxesImage at 0x7fa2f8d16eb0\PYGZgt{}
\end{sphinxVerbatim}

\noindent\sphinxincludegraphics{{overview-ns-forw_8_2}.png}

A lot has happened in this \sphinxcode{\sphinxupquote{step()}} call: we’ve advected the smoke field, added an upwards force via a Boussinesq model, advected the velocity field, and finally made it divergence free via a pressure solve.

The Boussinesq model uses a multiplication by a tuple \sphinxcode{\sphinxupquote{(0, buoyancy\_factor)}} to turn the smoke field into a staggered, 2 component force field, sampled at the locations of the velocity components via the \sphinxcode{\sphinxupquote{>>}} operator.
This operator makes sure the individual force components are correctly interpolated for the velocity components of the staggered velocity. \sphinxcode{\sphinxupquote{>>}} could be rewritten via the sampling function \sphinxcode{\sphinxupquote{at()}} as \sphinxcode{\sphinxupquote{(smoke*(0,buoyancy\_factor)).at(velocity)}}. However, \sphinxcode{\sphinxupquote{>>}} also directly ensure the boundary conditions of the original grid are kept. Hence, it internally also does \sphinxcode{\sphinxupquote{StaggeredGrid(..., extrapolation.ZERO,...)}} for the force grid. Long story short: the \sphinxcode{\sphinxupquote{>>}} operator does the same thing, and typically produces shorter and more readable code.

The pressure projection step in \sphinxcode{\sphinxupquote{make\_incompressible}} is typically the computationally most expensive step in the sequence above. It solves a Poisson equation for the boundary conditions of the domain, and updates the velocity field with the gradient of the computed pressure.

Just for testing, we’ve also printed the mean value of the velocities, and the max density after the update. As you can see in the resulting image, we have a first round region of smoke, with a slight upwards motion (which does not show here yet).

\subsection{Datatypes and dimensions}
\label{\detokenize{overview-ns-forw:datatypes-and-dimensions}}
The variables we created for the fields of the simulation here are instances of the class \sphinxcode{\sphinxupquote{Grid}}.
Like tensors, grids also have the \sphinxcode{\sphinxupquote{shape}} attribute which lists all batch, spatial and channel dimensions.
\sphinxhref{https://tum-pbs.github.io/PhiFlow/Math.html\#shapes}{Shapes in phiflow} store not only the sizes of the dimensions but also their names and types.

\begin{sphinxVerbatim}[commandchars=\\\{\}]
\PYG{n+nb}{print}\PYG{p}{(}\PYG{l+s+sa}{f}\PYG{l+s+s2}{\PYGZdq{}}\PYG{l+s+s2}{Smoke: }\PYG{l+s+si}{\PYGZob{}}\PYG{n}{smoke}\PYG{o}{.}\PYG{n}{shape}\PYG{l+s+si}{\PYGZcb{}}\PYG{l+s+s2}{\PYGZdq{}}\PYG{p}{)}
\PYG{n+nb}{print}\PYG{p}{(}\PYG{l+s+sa}{f}\PYG{l+s+s2}{\PYGZdq{}}\PYG{l+s+s2}{Velocity: }\PYG{l+s+si}{\PYGZob{}}\PYG{n}{velocity}\PYG{o}{.}\PYG{n}{shape}\PYG{l+s+si}{\PYGZcb{}}\PYG{l+s+s2}{\PYGZdq{}}\PYG{p}{)}
\PYG{n+nb}{print}\PYG{p}{(}\PYG{l+s+sa}{f}\PYG{l+s+s2}{\PYGZdq{}}\PYG{l+s+s2}{Inflow: }\PYG{l+s+si}{\PYGZob{}}\PYG{n}{INFLOW}\PYG{o}{.}\PYG{n}{shape}\PYG{l+s+si}{\PYGZcb{}}\PYG{l+s+s2}{, spatial only: }\PYG{l+s+si}{\PYGZob{}}\PYG{n}{INFLOW}\PYG{o}{.}\PYG{n}{shape}\PYG{o}{.}\PYG{n}{spatial}\PYG{l+s+si}{\PYGZcb{}}\PYG{l+s+s2}{\PYGZdq{}}\PYG{p}{)}
\end{sphinxVerbatim}

\begin{sphinxVerbatim}[commandchars=\\\{\}]
Smoke: (xˢ=32, yˢ=40)
Velocity: (xˢ=32, yˢ=40, vectorᵛ=2)
Inflow: (xˢ=32, yˢ=40), spatial only: (xˢ=32, yˢ=40)
\end{sphinxVerbatim}

Note that the phiflow output here indicates the type of a dimension, e.g., \(^S\) for a spatial, and \(^V\) for a vector dimension. Later on for learning, we’ll also introduce batch dimensions.

The actual content of a shape object can be obtained via \sphinxcode{\sphinxupquote{.sizes}}, or alternatively we can query the size of a specific dimension \sphinxcode{\sphinxupquote{dim}} via \sphinxcode{\sphinxupquote{.get\_size('dim')}}. Here are two examples:

\begin{sphinxVerbatim}[commandchars=\\\{\}]
\PYG{n+nb}{print}\PYG{p}{(}\PYG{l+s+sa}{f}\PYG{l+s+s2}{\PYGZdq{}}\PYG{l+s+s2}{Shape content: }\PYG{l+s+si}{\PYGZob{}}\PYG{n}{velocity}\PYG{o}{.}\PYG{n}{shape}\PYG{o}{.}\PYG{n}{sizes}\PYG{l+s+si}{\PYGZcb{}}\PYG{l+s+s2}{\PYGZdq{}}\PYG{p}{)}
\PYG{n+nb}{print}\PYG{p}{(}\PYG{l+s+sa}{f}\PYG{l+s+s2}{\PYGZdq{}}\PYG{l+s+s2}{Vector dimension: }\PYG{l+s+si}{\PYGZob{}}\PYG{n}{velocity}\PYG{o}{.}\PYG{n}{shape}\PYG{o}{.}\PYG{n}{get\PYGZus{}size}\PYG{p}{(}\PYG{l+s+s1}{\PYGZsq{}}\PYG{l+s+s1}{vector}\PYG{l+s+s1}{\PYGZsq{}}\PYG{p}{)}\PYG{l+s+si}{\PYGZcb{}}\PYG{l+s+s2}{\PYGZdq{}}\PYG{p}{)}
\end{sphinxVerbatim}

\begin{sphinxVerbatim}[commandchars=\\\{\}]
Shape content: (32, 40, 2)
Vector dimension: 2
\end{sphinxVerbatim}

The grid values can be accessed using the \sphinxcode{\sphinxupquote{values}} property. This is an important difference to a phiflow tensor object, which does not have \sphinxcode{\sphinxupquote{values}}, as illustrated in the code example below.

\begin{sphinxVerbatim}[commandchars=\\\{\}]
\PYG{n+nb}{print}\PYG{p}{(}\PYG{l+s+s2}{\PYGZdq{}}\PYG{l+s+s2}{Statistics of the different simulation grids:}\PYG{l+s+s2}{\PYGZdq{}}\PYG{p}{)}
\PYG{n+nb}{print}\PYG{p}{(}\PYG{n}{smoke}\PYG{o}{.}\PYG{n}{values}\PYG{p}{)}
\PYG{n+nb}{print}\PYG{p}{(}\PYG{n}{velocity}\PYG{o}{.}\PYG{n}{values}\PYG{p}{)}

\PYG{c+c1}{\PYGZsh{} in contrast to a simple tensor:}
\PYG{n}{test\PYGZus{}tensor} \PYG{o}{=} \PYG{n}{math}\PYG{o}{.}\PYG{n}{tensor}\PYG{p}{(}\PYG{n}{numpy}\PYG{o}{.}\PYG{n}{zeros}\PYG{p}{(}\PYG{p}{[}\PYG{l+m+mi}{2}\PYG{p}{,} \PYG{l+m+mi}{5}\PYG{p}{,} \PYG{l+m+mi}{3}\PYG{p}{]}\PYG{p}{)}\PYG{p}{,} \PYG{n}{spatial}\PYG{p}{(}\PYG{l+s+s1}{\PYGZsq{}}\PYG{l+s+s1}{x,y}\PYG{l+s+s1}{\PYGZsq{}}\PYG{p}{)}\PYG{p}{,} \PYG{n}{channel}\PYG{p}{(}\PYG{l+s+s1}{\PYGZsq{}}\PYG{l+s+s1}{vector}\PYG{l+s+s1}{\PYGZsq{}}\PYG{p}{)}\PYG{p}{)}
\PYG{n+nb}{print}\PYG{p}{(}\PYG{l+s+s2}{\PYGZdq{}}\PYG{l+s+s2}{Reordered test tensor shape: }\PYG{l+s+s2}{\PYGZdq{}} \PYG{o}{+} \PYG{n+nb}{format}\PYG{p}{(}\PYG{n}{test\PYGZus{}tensor}\PYG{o}{.}\PYG{n}{numpy}\PYG{p}{(}\PYG{l+s+s1}{\PYGZsq{}}\PYG{l+s+s1}{vector,y,x}\PYG{l+s+s1}{\PYGZsq{}}\PYG{p}{)}\PYG{o}{.}\PYG{n}{shape}\PYG{p}{)} \PYG{p}{)}
\PYG{c+c1}{\PYGZsh{}print(test\PYGZus{}tensor.values.numpy(\PYGZsq{}y,x\PYGZsq{})) \PYGZsh{} error! tensors don\PYGZsq{}t return their content via \PYGZdq{}.values\PYGZdq{}}
\end{sphinxVerbatim}

\begin{sphinxVerbatim}[commandchars=\\\{\}]
Statistics of the different simulation grids:
(xˢ=32, yˢ=40) float32  0.0 \PYGZlt{} ... \PYGZlt{} 0.20000000298023224
(xˢ=(31, 32), yˢ=(40, 39), vectorᵛ=2) float32  \PYGZhy{}0.12352858483791351 \PYGZlt{} ... \PYGZlt{} 0.15530994534492493
Reordered test tensor shape: (3, 5, 2)
\end{sphinxVerbatim}

Grids have many more properties which are documented \sphinxhref{https://tum-pbs.github.io/PhiFlow/phi/field/\#phi.field.Grid}{here}.
Also note that the staggered grid has a \sphinxhref{https://tum-pbs.github.io/PhiFlow/Math.html\#non-uniform-tensors}{non\sphinxhyphen{}uniform shape} because the number of faces is not equal to the number of cells (in this example the x component has \(31 \times 40\) cells, while y has \(32 \times 39\)). The \sphinxcode{\sphinxupquote{INFLOW}} grid naturally has the same dimensions as the \sphinxcode{\sphinxupquote{smoke}} grid.

\subsection{Time evolution}
\label{\detokenize{overview-ns-forw:time-evolution}}
With this setup, we can easily advance the simulation forward in time a bit more by repeatedly calling the \sphinxcode{\sphinxupquote{step}} function.

\begin{sphinxVerbatim}[commandchars=\\\{\}]
\PYG{k}{for} \PYG{n}{time\PYGZus{}step} \PYG{o+ow}{in} \PYG{n+nb}{range}\PYG{p}{(}\PYG{l+m+mi}{10}\PYG{p}{)}\PYG{p}{:}
    \PYG{n}{velocity}\PYG{p}{,} \PYG{n}{smoke}\PYG{p}{,} \PYG{n}{pressure} \PYG{o}{=} \PYG{n}{step}\PYG{p}{(}\PYG{n}{velocity}\PYG{p}{,} \PYG{n}{smoke}\PYG{p}{,} \PYG{n}{pressure}\PYG{p}{,} \PYG{n}{dt}\PYG{o}{=}\PYG{n}{DT}\PYG{p}{)}
    \PYG{n+nb}{print}\PYG{p}{(}\PYG{l+s+s1}{\PYGZsq{}}\PYG{l+s+s1}{Computed frame }\PYG{l+s+si}{\PYGZob{}\PYGZcb{}}\PYG{l+s+s1}{, max velocity }\PYG{l+s+si}{\PYGZob{}\PYGZcb{}}\PYG{l+s+s1}{\PYGZsq{}}\PYG{o}{.}\PYG{n}{format}\PYG{p}{(}\PYG{n}{time\PYGZus{}step} \PYG{p}{,} \PYG{n}{np}\PYG{o}{.}\PYG{n}{asarray}\PYG{p}{(}\PYG{n}{math}\PYG{o}{.}\PYG{n}{max}\PYG{p}{(}\PYG{n}{velocity}\PYG{o}{.}\PYG{n}{values}\PYG{p}{)}\PYG{p}{)} \PYG{p}{)}\PYG{p}{)}
\end{sphinxVerbatim}

\begin{sphinxVerbatim}[commandchars=\\\{\}]
Computed frame 0, max velocity 0.4609803557395935
Computed frame 1, max velocity 0.8926814794540405
Computed frame 2, max velocity 1.4052708148956299
Computed frame 3, max velocity 2.036500930786133
Computed frame 4, max velocity 2.921010971069336
Computed frame 5, max velocity 3.828195810317993
Computed frame 6, max velocity 4.516851425170898
Computed frame 7, max velocity 4.861286640167236
Computed frame 8, max velocity 5.125314235687256
Computed frame 9, max velocity 5.476282119750977
\end{sphinxVerbatim}

Now the hot plume is starting to rise:

\begin{sphinxVerbatim}[commandchars=\\\{\}]
\PYG{n}{pylab}\PYG{o}{.}\PYG{n}{imshow}\PYG{p}{(}\PYG{n}{smoke}\PYG{o}{.}\PYG{n}{values}\PYG{o}{.}\PYG{n}{numpy}\PYG{p}{(}\PYG{l+s+s1}{\PYGZsq{}}\PYG{l+s+s1}{y,x}\PYG{l+s+s1}{\PYGZsq{}}\PYG{p}{)}\PYG{p}{,} \PYG{n}{origin}\PYG{o}{=}\PYG{l+s+s1}{\PYGZsq{}}\PYG{l+s+s1}{lower}\PYG{l+s+s1}{\PYGZsq{}}\PYG{p}{,} \PYG{n}{cmap}\PYG{o}{=}\PYG{l+s+s1}{\PYGZsq{}}\PYG{l+s+s1}{magma}\PYG{l+s+s1}{\PYGZsq{}}\PYG{p}{)}
\end{sphinxVerbatim}

\begin{sphinxVerbatim}[commandchars=\\\{\}]
\PYGZlt{}matplotlib.image.AxesImage at 0x7fa2f902cf40\PYGZgt{}
\end{sphinxVerbatim}

\noindent\sphinxincludegraphics{{overview-ns-forw_19_1}.png}

Let’s compute and show a few more steps of the simulation. Because of the inflow being located off\sphinxhyphen{}center to the left (with x position 30), the plume will curve towards the right when it hits the top wall of the domain.

\begin{sphinxVerbatim}[commandchars=\\\{\}]
\PYG{n}{steps} \PYG{o}{=} \PYG{p}{[}\PYG{p}{[} \PYG{n}{smoke}\PYG{o}{.}\PYG{n}{values}\PYG{p}{,} \PYG{n}{velocity}\PYG{o}{.}\PYG{n}{values}\PYG{o}{.}\PYG{n}{vector}\PYG{p}{[}\PYG{l+m+mi}{0}\PYG{p}{]}\PYG{p}{,} \PYG{n}{velocity}\PYG{o}{.}\PYG{n}{values}\PYG{o}{.}\PYG{n}{vector}\PYG{p}{[}\PYG{l+m+mi}{1}\PYG{p}{]} \PYG{p}{]}\PYG{p}{]}
\PYG{k}{for} \PYG{n}{time\PYGZus{}step} \PYG{o+ow}{in} \PYG{n+nb}{range}\PYG{p}{(}\PYG{l+m+mi}{20}\PYG{p}{)}\PYG{p}{:}
  \PYG{k}{if} \PYG{n}{time\PYGZus{}step}\PYG{o}{\PYGZlt{}}\PYG{l+m+mi}{3} \PYG{o+ow}{or} \PYG{n}{time\PYGZus{}step}\PYG{o}{\PYGZpc{}}\PYG{k}{10}==0: 
    \PYG{n+nb}{print}\PYG{p}{(}\PYG{l+s+s1}{\PYGZsq{}}\PYG{l+s+s1}{Computing time step }\PYG{l+s+si}{\PYGZpc{}d}\PYG{l+s+s1}{\PYGZsq{}} \PYG{o}{\PYGZpc{}} \PYG{n}{time\PYGZus{}step}\PYG{p}{)}
  \PYG{n}{velocity}\PYG{p}{,} \PYG{n}{smoke}\PYG{p}{,} \PYG{n}{pressure} \PYG{o}{=} \PYG{n}{step}\PYG{p}{(}\PYG{n}{velocity}\PYG{p}{,} \PYG{n}{smoke}\PYG{p}{,} \PYG{n}{pressure}\PYG{p}{,} \PYG{n}{dt}\PYG{o}{=}\PYG{n}{DT}\PYG{p}{)}
  \PYG{k}{if} \PYG{n}{time\PYGZus{}step}\PYG{o}{\PYGZpc{}}\PYG{k}{5}==0:
    \PYG{n}{steps}\PYG{o}{.}\PYG{n}{append}\PYG{p}{(} \PYG{p}{[}\PYG{n}{smoke}\PYG{o}{.}\PYG{n}{values}\PYG{p}{,} \PYG{n}{velocity}\PYG{o}{.}\PYG{n}{values}\PYG{o}{.}\PYG{n}{vector}\PYG{p}{[}\PYG{l+m+mi}{0}\PYG{p}{]}\PYG{p}{,} \PYG{n}{velocity}\PYG{o}{.}\PYG{n}{values}\PYG{o}{.}\PYG{n}{vector}\PYG{p}{[}\PYG{l+m+mi}{1}\PYG{p}{]}\PYG{p}{]} \PYG{p}{)}

\PYG{n}{fig}\PYG{p}{,} \PYG{n}{axes} \PYG{o}{=} \PYG{n}{pylab}\PYG{o}{.}\PYG{n}{subplots}\PYG{p}{(}\PYG{l+m+mi}{1}\PYG{p}{,} \PYG{n+nb}{len}\PYG{p}{(}\PYG{n}{steps}\PYG{p}{)}\PYG{p}{,} \PYG{n}{figsize}\PYG{o}{=}\PYG{p}{(}\PYG{l+m+mi}{16}\PYG{p}{,} \PYG{l+m+mi}{5}\PYG{p}{)}\PYG{p}{)}
\PYG{k}{for} \PYG{n}{i} \PYG{o+ow}{in} \PYG{n+nb}{range}\PYG{p}{(}\PYG{n+nb}{len}\PYG{p}{(}\PYG{n}{steps}\PYG{p}{)}\PYG{p}{)}\PYG{p}{:}
    \PYG{n}{axes}\PYG{p}{[}\PYG{n}{i}\PYG{p}{]}\PYG{o}{.}\PYG{n}{imshow}\PYG{p}{(}\PYG{n}{steps}\PYG{p}{[}\PYG{n}{i}\PYG{p}{]}\PYG{p}{[}\PYG{l+m+mi}{0}\PYG{p}{]}\PYG{o}{.}\PYG{n}{numpy}\PYG{p}{(}\PYG{l+s+s1}{\PYGZsq{}}\PYG{l+s+s1}{y,x}\PYG{l+s+s1}{\PYGZsq{}}\PYG{p}{)}\PYG{p}{,} \PYG{n}{origin}\PYG{o}{=}\PYG{l+s+s1}{\PYGZsq{}}\PYG{l+s+s1}{lower}\PYG{l+s+s1}{\PYGZsq{}}\PYG{p}{,} \PYG{n}{cmap}\PYG{o}{=}\PYG{l+s+s1}{\PYGZsq{}}\PYG{l+s+s1}{magma}\PYG{l+s+s1}{\PYGZsq{}}\PYG{p}{)}
    \PYG{n}{axes}\PYG{p}{[}\PYG{n}{i}\PYG{p}{]}\PYG{o}{.}\PYG{n}{set\PYGZus{}title}\PYG{p}{(}\PYG{l+s+sa}{f}\PYG{l+s+s2}{\PYGZdq{}}\PYG{l+s+s2}{d at t=}\PYG{l+s+si}{\PYGZob{}}\PYG{n}{i}\PYG{o}{*}\PYG{l+m+mi}{5}\PYG{l+s+si}{\PYGZcb{}}\PYG{l+s+s2}{\PYGZdq{}}\PYG{p}{)}
\end{sphinxVerbatim}

\begin{sphinxVerbatim}[commandchars=\\\{\}]
Computing time step 0
Computing time step 1
Computing time step 2
Computing time step 10
\end{sphinxVerbatim}

\noindent\sphinxincludegraphics{{overview-ns-forw_21_1}.png}

We can also take a look at the velocities. The \sphinxcode{\sphinxupquote{steps}} list above already stores \sphinxcode{\sphinxupquote{vector{[}0{]}}} and \sphinxcode{\sphinxupquote{vector{[}1{]}}} components of the velocities as numpy arrays, which we can show next.

\begin{sphinxVerbatim}[commandchars=\\\{\}]
\PYG{n}{fig}\PYG{p}{,} \PYG{n}{axes} \PYG{o}{=} \PYG{n}{pylab}\PYG{o}{.}\PYG{n}{subplots}\PYG{p}{(}\PYG{l+m+mi}{1}\PYG{p}{,} \PYG{n+nb}{len}\PYG{p}{(}\PYG{n}{steps}\PYG{p}{)}\PYG{p}{,} \PYG{n}{figsize}\PYG{o}{=}\PYG{p}{(}\PYG{l+m+mi}{16}\PYG{p}{,} \PYG{l+m+mi}{5}\PYG{p}{)}\PYG{p}{)}
\PYG{k}{for} \PYG{n}{i} \PYG{o+ow}{in} \PYG{n+nb}{range}\PYG{p}{(}\PYG{n+nb}{len}\PYG{p}{(}\PYG{n}{steps}\PYG{p}{)}\PYG{p}{)}\PYG{p}{:}
    \PYG{n}{axes}\PYG{p}{[}\PYG{n}{i}\PYG{p}{]}\PYG{o}{.}\PYG{n}{imshow}\PYG{p}{(}\PYG{n}{steps}\PYG{p}{[}\PYG{n}{i}\PYG{p}{]}\PYG{p}{[}\PYG{l+m+mi}{1}\PYG{p}{]}\PYG{o}{.}\PYG{n}{numpy}\PYG{p}{(}\PYG{l+s+s1}{\PYGZsq{}}\PYG{l+s+s1}{y,x}\PYG{l+s+s1}{\PYGZsq{}}\PYG{p}{)}\PYG{p}{,} \PYG{n}{origin}\PYG{o}{=}\PYG{l+s+s1}{\PYGZsq{}}\PYG{l+s+s1}{lower}\PYG{l+s+s1}{\PYGZsq{}}\PYG{p}{,} \PYG{n}{cmap}\PYG{o}{=}\PYG{l+s+s1}{\PYGZsq{}}\PYG{l+s+s1}{magma}\PYG{l+s+s1}{\PYGZsq{}}\PYG{p}{)}
    \PYG{n}{axes}\PYG{p}{[}\PYG{n}{i}\PYG{p}{]}\PYG{o}{.}\PYG{n}{set\PYGZus{}title}\PYG{p}{(}\PYG{l+s+sa}{f}\PYG{l+s+s2}{\PYGZdq{}}\PYG{l+s+s2}{u\PYGZus{}x at t=}\PYG{l+s+si}{\PYGZob{}}\PYG{n}{i}\PYG{o}{*}\PYG{l+m+mi}{5}\PYG{l+s+si}{\PYGZcb{}}\PYG{l+s+s2}{\PYGZdq{}}\PYG{p}{)}
    
\PYG{n}{fig}\PYG{p}{,} \PYG{n}{axes} \PYG{o}{=} \PYG{n}{pylab}\PYG{o}{.}\PYG{n}{subplots}\PYG{p}{(}\PYG{l+m+mi}{1}\PYG{p}{,} \PYG{n+nb}{len}\PYG{p}{(}\PYG{n}{steps}\PYG{p}{)}\PYG{p}{,} \PYG{n}{figsize}\PYG{o}{=}\PYG{p}{(}\PYG{l+m+mi}{16}\PYG{p}{,} \PYG{l+m+mi}{5}\PYG{p}{)}\PYG{p}{)}
\PYG{k}{for} \PYG{n}{i} \PYG{o+ow}{in} \PYG{n+nb}{range}\PYG{p}{(}\PYG{n+nb}{len}\PYG{p}{(}\PYG{n}{steps}\PYG{p}{)}\PYG{p}{)}\PYG{p}{:}
    \PYG{n}{axes}\PYG{p}{[}\PYG{n}{i}\PYG{p}{]}\PYG{o}{.}\PYG{n}{imshow}\PYG{p}{(}\PYG{n}{steps}\PYG{p}{[}\PYG{n}{i}\PYG{p}{]}\PYG{p}{[}\PYG{l+m+mi}{2}\PYG{p}{]}\PYG{o}{.}\PYG{n}{numpy}\PYG{p}{(}\PYG{l+s+s1}{\PYGZsq{}}\PYG{l+s+s1}{y,x}\PYG{l+s+s1}{\PYGZsq{}}\PYG{p}{)}\PYG{p}{,} \PYG{n}{origin}\PYG{o}{=}\PYG{l+s+s1}{\PYGZsq{}}\PYG{l+s+s1}{lower}\PYG{l+s+s1}{\PYGZsq{}}\PYG{p}{,} \PYG{n}{cmap}\PYG{o}{=}\PYG{l+s+s1}{\PYGZsq{}}\PYG{l+s+s1}{magma}\PYG{l+s+s1}{\PYGZsq{}}\PYG{p}{)}
    \PYG{n}{axes}\PYG{p}{[}\PYG{n}{i}\PYG{p}{]}\PYG{o}{.}\PYG{n}{set\PYGZus{}title}\PYG{p}{(}\PYG{l+s+sa}{f}\PYG{l+s+s2}{\PYGZdq{}}\PYG{l+s+s2}{u\PYGZus{}y at t=}\PYG{l+s+si}{\PYGZob{}}\PYG{n}{i}\PYG{o}{*}\PYG{l+m+mi}{5}\PYG{l+s+si}{\PYGZcb{}}\PYG{l+s+s2}{\PYGZdq{}}\PYG{p}{)}
    
\end{sphinxVerbatim}

\noindent\sphinxincludegraphics{{overview-ns-forw_23_0}.png}

\noindent\sphinxincludegraphics{{overview-ns-forw_23_1}.png}

It looks simple here, but this simulation setup is a powerful tool. The simulation could easily be extended to more complex cases or 3D, and it is already fully compatible with backpropagation pipelines of deep learning frameworks.

In the next chapters we’ll show how to use these simulations for training NNs, and how to steer and modify them via trained NNs. This will illustrate how much we can improve the training process by having a solver in the loop, especially when the solver is \sphinxstyleemphasis{differentiable}. Before moving to these more complex training processes, we will cover a simpler supervised approach in the next chapter. This is very fundamental: even when aiming for advanced physics\sphinxhyphen{}based learning setups, a working supervised training is always the first step.

\subsection{Next steps}
\label{\detokenize{overview-ns-forw:next-steps}}
You could create a variety of nice fluid simulations based on this setup. E.g., try changing the spatial resolution, the buoyancy factors, and the overall length of the simulation run.

\chapter{Supervised Training}
\label{\detokenize{supervised:supervised-training}}\label{\detokenize{supervised::doc}}
\sphinxstyleemphasis{Supervised} here essentially means: “doing things the old fashioned way”. Old fashioned in the context of
deep learning (DL), of course, so it’s still fairly new.
Also, “old fashioned” doesn’t always mean bad \sphinxhyphen{} it’s just that later on we’ll discuss ways to train networks that clearly outperform approaches using supervised training.

Nonetheless, “supervised training” is a starting point for all projects one would encounter in the context of DL, and
hence it is worth studying. Also, while it typically yields inferior results to approaches that more tightly
couple with physics, it can be the only choice in certain application scenarios where no good
model equations exist.

\section{Problem setting}
\label{\detokenize{supervised:problem-setting}}
For supervised training, we’re faced with an
unknown function \(f^*(x)=y^*\), collect lots of pairs of data \([x_0,y^*_0], ...[x_n,y^*_n]\) (the training data set)
and directly train an NN to represent an approximation of \(f^*\) denoted as \(f\).

The \(f\) we can obtain in this way is typically not exact,
but instead we obtain it via a minimization problem:
by adjusting the weights \(\theta\) of our NN representation of \(f\) such that
\begin{equation}\label{equation:supervised:supervised-training}
\begin{split}
\text{arg min}_{\theta} \sum_i (f(x_i ; \theta)-y^*_i)^2 .
\end{split}
\end{equation}
This will give us \(\theta\) such that \(f(x;\theta) =  y \approx y^*\) as accurately as possible given
our choice of \(f\) and the hyperparameters for training. Note that above we’ve assumed
the simplest case of an \(L^2\) loss. A more general version would use an error metric \(e(x,y)\)
to be minimized via \(\text{arg min}_{\theta} \sum_i e( f(x_i ; \theta) , y^*_i) )\). The choice
of a suitable metric is a topic we will get back to later on.

Irrespective of our choice of metric, this formulation
gives the actual “learning” process for a supervised approach.

The training data typically needs to be of substantial size, and hence it is attractive
to use numerical simulations solving a physical model \(\mathcal{P}\)
to produce a large number of reliable input\sphinxhyphen{}output pairs for training.
This means that the training process uses a set of model equations, and approximates
them numerically, in order to train the NN representation \(f\). This
has quite a few advantages, e.g., we don’t have measurement noise of real\sphinxhyphen{}world devices
and we don’t need manual labour to annotate a large number of samples to get training data.

On the other hand, this approach inherits the common challenges of replacing experiments
with simulations: first, we need to ensure the chosen model has enough power to predict the
behavior of real\sphinxhyphen{}world phenomena that we’re interested in.
In addition, the numerical approximations have numerical errors
which need to be kept small enough for a chosen application. As these topics are studied in depth
for classical simulations, and the existing knowledge can likewise be leveraged to
set up DL training tasks.

\begin{figure}[htbp]
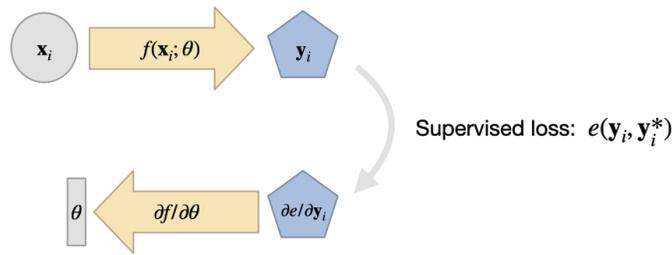

\centering
\capstart

\noindent\sphinxincludegraphics[height=220\sphinxpxdimen]{{supervised-training}.jpg}
\caption{A visual overview of supervised training. Quite simple overall, but it’s good to keep this
in mind in comparison to the more complex variants we’ll encounter later on.}\label{\detokenize{supervised:id1}}\end{figure}

\section{Surrogate models}
\label{\detokenize{supervised:surrogate-models}}
One of the central advantages of the supervised approach above is that
we obtain a \sphinxstyleemphasis{surrogate model}, i.e., a new function that mimics the behavior of the original \(\mathcal{P}\).
The numerical approximations
of PDE models for real world phenomena are often very expensive to compute. A trained
NN on the other hand incurs a constant cost per evaluation, and is typically trivial
to evaluate on specialized hardware such as GPUs or NN units.

Despite this, it’s important to be careful:
NNs can quickly generate huge numbers of in between results. Consider a CNN layer with
\(128\) features. If we apply it to an input of \(128^2\), i.e. ca. 16k cells, we get \(128^3\) intermediate values.
That’s more than 2 million.
All these values at least need to be momentarily stored in memory, and processed by the next layer.

Nonetheless, replacing complex and expensive solvers with fast, learned approximations
is a very attractive and interesting direction.

\section{Show me some code!}
\label{\detokenize{supervised:show-me-some-code}}
Let’s finally look at a code example that trains a neural network:
we’ll replace a full solver for \sphinxstyleemphasis{turbulent flows around airfoils} with a surrogate model from {[}\hyperlink{cite.references:id16}{TWPH20}{]}.

\section{Supervised training for RANS flows around airfoils}
\label{\detokenize{supervised-airfoils:supervised-training-for-rans-flows-around-airfoils}}\label{\detokenize{supervised-airfoils::doc}}

\subsection{Overview}
\label{\detokenize{supervised-airfoils:overview}}
For this example of supervised training
we have a turbulent airflow around wing profiles, and we’d like to know the average motion
and pressure distribution around this airfoil for different Reynolds numbers and angles of attack.
Thus, given an airfoil shape, Reynolds numbers, and angle of attack, we’d like to obtain
a velocity field and a pressure field around the airfoil.

This is classically approximated with \sphinxstyleemphasis{Reynolds\sphinxhyphen{}Averaged Navier Stokes} (RANS) models, and this
setting is still one of the most widely used applications of Navier\sphinxhyphen{}Stokes solver in industry.
However, instead of relying on traditional numerical methods to solve the RANS equations,
we now aim for training a surrogate model via a neural network that completely bypasses the numerical solver,
and produces the solution in terms of velocity and pressure.
\sphinxhref{https://colab.research.google.com/github/tum-pbs/pbdl-book/blob/main/supervised-airfoils.ipynb}{{[}run in colab{]}}

\subsection{Formulation}
\label{\detokenize{supervised-airfoils:formulation}}
With the supervised formulation from {\hyperref[\detokenize{supervised::doc}]{\sphinxcrossref{\DUrole{doc}{Supervised Training}}}}, our learning task is pretty straight\sphinxhyphen{}forward, and can be written as
\begin{equation*}
\begin{split}\begin{aligned}
\text{arg min}_{\theta} \sum_i ( f(x_i ; \theta)-y^*_i )^2 ,
\end{aligned}\end{split}
\end{equation*}
where \(x\) and \(y^*\) each consist of a set of physical fields,
and the index \(i\) evaluates the difference across all discretization points in our data sets.

The goal is to infer velocity \(\mathbf{u} = u_x,u_y\) and a pressure field \(p\) in a computational domain \(\Omega\)
around the airfoil in the center of \(\Omega\).
\(u_x,u_y\) and \(p\) each have a dimension of \(128^2\).
As inputs we have the Reynolds number \(\text{Re} \in \mathbb{R}\), the angle of attack
\(\alpha \in \mathbb{R}\), and the airfoil shape \(\mathbf{s}\) encoded as a rasterized grid with \(128^2\).
Both constant, scalar inputs \(\text{Re}\) and \(\alpha\) are likewise extended to a size of \(128^2\).
Thus, put together, both input and output have the same dimensions: \(x,y^* \in \mathbb{R}^{3\times128\times128}\).
This is exactly what we’ll specify as input and output dimensions for the NN below.

A point to keep in mind here is that our quantities of interest in \(y^*\) contain three different physical fields. While the two velocity components are quite similar in spirit, the pressure field typically has a different behavior with an approximately squared scaling with respect to the velocity (cf. \sphinxhref{https://en.wikipedia.org/wiki/Bernoulli\%27s\_principle}{Bernoulli}). This implies that we need to be careful with simple summations (as in the minimization problem above), and that we should take care to normalize the data.

\subsection{Code coming up…}
\label{\detokenize{supervised-airfoils:code-coming-up}}
Let’s get started with the implementation. Note that we’ll skip the data generation process here. The code below is adapted from {[}\hyperlink{cite.references:id16}{TWPH20}{]} and \sphinxhref{https://github.com/thunil/Deep-Flow-Prediction}{this codebase}, which you can check out for details. Here, we’ll simply download a small set of training data generated with a Spalart\sphinxhyphen{}Almaras RANS simulation in \sphinxhref{https://openfoam.org/}{OpenFOAM}.

\begin{sphinxVerbatim}[commandchars=\\\{\}]
\PYG{k+kn}{import} \PYG{n+nn}{numpy} \PYG{k}{as} \PYG{n+nn}{np}
\PYG{k+kn}{import} \PYG{n+nn}{os}\PYG{n+nn}{.}\PYG{n+nn}{path}\PYG{o}{,} \PYG{n+nn}{random}
\PYG{k+kn}{import} \PYG{n+nn}{torch}
\PYG{k+kn}{from} \PYG{n+nn}{torch}\PYG{n+nn}{.}\PYG{n+nn}{utils}\PYG{n+nn}{.}\PYG{n+nn}{data} \PYG{k+kn}{import} \PYG{n}{Dataset}
\PYG{n+nb}{print}\PYG{p}{(}\PYG{l+s+s2}{\PYGZdq{}}\PYG{l+s+s2}{Torch version }\PYG{l+s+si}{\PYGZob{}\PYGZcb{}}\PYG{l+s+s2}{\PYGZdq{}}\PYG{o}{.}\PYG{n}{format}\PYG{p}{(}\PYG{n}{torch}\PYG{o}{.}\PYG{n}{\PYGZus{}\PYGZus{}version\PYGZus{}\PYGZus{}}\PYG{p}{)}\PYG{p}{)}

\PYG{c+c1}{\PYGZsh{} get training data}
\PYG{n+nb}{dir} \PYG{o}{=} \PYG{l+s+s2}{\PYGZdq{}}\PYG{l+s+s2}{./}\PYG{l+s+s2}{\PYGZdq{}}
\PYG{k}{if} \PYG{k+kc}{True}\PYG{p}{:}
    \PYG{c+c1}{\PYGZsh{} download}
    \PYG{k}{if} \PYG{o+ow}{not} \PYG{n}{os}\PYG{o}{.}\PYG{n}{path}\PYG{o}{.}\PYG{n}{isfile}\PYG{p}{(}\PYG{l+s+s1}{\PYGZsq{}}\PYG{l+s+s1}{data\PYGZhy{}airfoils.npz}\PYG{l+s+s1}{\PYGZsq{}}\PYG{p}{)}\PYG{p}{:}
        \PYG{k+kn}{import} \PYG{n+nn}{requests}
        \PYG{n+nb}{print}\PYG{p}{(}\PYG{l+s+s2}{\PYGZdq{}}\PYG{l+s+s2}{Downloading training data (300MB), this can take a few minutes the first time...}\PYG{l+s+s2}{\PYGZdq{}}\PYG{p}{)}
        \PYG{k}{with} \PYG{n+nb}{open}\PYG{p}{(}\PYG{l+s+s2}{\PYGZdq{}}\PYG{l+s+s2}{data\PYGZhy{}airfoils.npz}\PYG{l+s+s2}{\PYGZdq{}}\PYG{p}{,} \PYG{l+s+s1}{\PYGZsq{}}\PYG{l+s+s1}{wb}\PYG{l+s+s1}{\PYGZsq{}}\PYG{p}{)} \PYG{k}{as} \PYG{n}{datafile}\PYG{p}{:}
            \PYG{n}{resp} \PYG{o}{=} \PYG{n}{requests}\PYG{o}{.}\PYG{n}{get}\PYG{p}{(}\PYG{l+s+s1}{\PYGZsq{}}\PYG{l+s+s1}{https://dataserv.ub.tum.de/s/m1615239/download?path=}\PYG{l+s+si}{\PYGZpc{}2F}\PYG{l+s+s1}{\PYGZam{}files=dfp\PYGZhy{}data\PYGZhy{}400.npz}\PYG{l+s+s1}{\PYGZsq{}}\PYG{p}{,} \PYG{n}{verify}\PYG{o}{=}\PYG{k+kc}{False}\PYG{p}{)}
            \PYG{n}{datafile}\PYG{o}{.}\PYG{n}{write}\PYG{p}{(}\PYG{n}{resp}\PYG{o}{.}\PYG{n}{content}\PYG{p}{)}
\PYG{k}{else}\PYG{p}{:} 
    \PYG{c+c1}{\PYGZsh{} alternative: load from google drive (upload there beforehand):}
    \PYG{k+kn}{from} \PYG{n+nn}{google}\PYG{n+nn}{.}\PYG{n+nn}{colab} \PYG{k+kn}{import} \PYG{n}{drive}
    \PYG{n}{drive}\PYG{o}{.}\PYG{n}{mount}\PYG{p}{(}\PYG{l+s+s1}{\PYGZsq{}}\PYG{l+s+s1}{/content/gdrive}\PYG{l+s+s1}{\PYGZsq{}}\PYG{p}{)}
    \PYG{n+nb}{dir} \PYG{o}{=} \PYG{l+s+s2}{\PYGZdq{}}\PYG{l+s+s2}{./gdrive/My Drive/}\PYG{l+s+s2}{\PYGZdq{}}

\PYG{n}{npfile}\PYG{o}{=}\PYG{n}{np}\PYG{o}{.}\PYG{n}{load}\PYG{p}{(}\PYG{n+nb}{dir}\PYG{o}{+}\PYG{l+s+s1}{\PYGZsq{}}\PYG{l+s+s1}{data\PYGZhy{}airfoils.npz}\PYG{l+s+s1}{\PYGZsq{}}\PYG{p}{)}
    
\PYG{n+nb}{print}\PYG{p}{(}\PYG{l+s+s2}{\PYGZdq{}}\PYG{l+s+s2}{Loaded data, }\PYG{l+s+si}{\PYGZob{}\PYGZcb{}}\PYG{l+s+s2}{ training, }\PYG{l+s+si}{\PYGZob{}\PYGZcb{}}\PYG{l+s+s2}{ validation samples}\PYG{l+s+s2}{\PYGZdq{}}\PYG{o}{.}\PYG{n}{format}\PYG{p}{(}\PYG{n+nb}{len}\PYG{p}{(}\PYG{n}{npfile}\PYG{p}{[}\PYG{l+s+s2}{\PYGZdq{}}\PYG{l+s+s2}{inputs}\PYG{l+s+s2}{\PYGZdq{}}\PYG{p}{]}\PYG{p}{)}\PYG{p}{,}\PYG{n+nb}{len}\PYG{p}{(}\PYG{n}{npfile}\PYG{p}{[}\PYG{l+s+s2}{\PYGZdq{}}\PYG{l+s+s2}{vinputs}\PYG{l+s+s2}{\PYGZdq{}}\PYG{p}{]}\PYG{p}{)}\PYG{p}{)}\PYG{p}{)}

\PYG{n+nb}{print}\PYG{p}{(}\PYG{l+s+s2}{\PYGZdq{}}\PYG{l+s+s2}{Size of the inputs array: }\PYG{l+s+s2}{\PYGZdq{}}\PYG{o}{+}\PYG{n+nb}{format}\PYG{p}{(}\PYG{n}{npfile}\PYG{p}{[}\PYG{l+s+s2}{\PYGZdq{}}\PYG{l+s+s2}{inputs}\PYG{l+s+s2}{\PYGZdq{}}\PYG{p}{]}\PYG{o}{.}\PYG{n}{shape}\PYG{p}{)}\PYG{p}{)}
\end{sphinxVerbatim}

\begin{sphinxVerbatim}[commandchars=\\\{\}]
Torch version 1.8.1+cu101
Downloading training data (300MB), this can take a few minutes the first time...
Loaded data, 320 training, 80 validation samples
Size of the inputs array: (320, 3, 128, 128)
\end{sphinxVerbatim}

If you run this notebook in colab, the \sphinxcode{\sphinxupquote{else}} statement above (which is deactivated by default) might be interesting for you: instead of downloading the training data anew every time, you can manually download it once and store it in your google drive. We assume it’s stored in the root directory as \sphinxcode{\sphinxupquote{data\sphinxhyphen{}airfoils.npz}}. Afterwards, you can use the code above to load the file from your google drive, which is typically much faster. This is highly recommended if you want to experiment more extensively via colab.

\subsection{RANS training data}
\label{\detokenize{supervised-airfoils:rans-training-data}}
Now we have some training data. In general it’s very important to understand the data we’re working with as much as possible (for any ML task the \sphinxstyleemphasis{garbage\sphinxhyphen{}in\sphinxhyphen{}gargabe\sphinxhyphen{}out} principle definitely holds). We should at least understand the data in terms of dimensions and rough statistics, but ideally also in terms of content. Otherwise we’ll have a very hard time interpreting the results of a training run. And despite all the DL magic: if you can’t make out any patterns in your data, NNs surely won’t find any useful ones.

Hence, let’s look at one of the training samples… The following is just some helper code to show images side by side.

\begin{sphinxVerbatim}[commandchars=\\\{\}]
\PYG{k+kn}{import} \PYG{n+nn}{pylab}

\PYG{c+c1}{\PYGZsh{} helper to show three target channels: normalized, with colormap, side by side}
\PYG{k}{def} \PYG{n+nf}{showSbs}\PYG{p}{(}\PYG{n}{a1}\PYG{p}{,}\PYG{n}{a2}\PYG{p}{,} \PYG{n}{stats}\PYG{o}{=}\PYG{k+kc}{False}\PYG{p}{,} \PYG{n}{bottom}\PYG{o}{=}\PYG{l+s+s2}{\PYGZdq{}}\PYG{l+s+s2}{NN Output}\PYG{l+s+s2}{\PYGZdq{}}\PYG{p}{,} \PYG{n}{top}\PYG{o}{=}\PYG{l+s+s2}{\PYGZdq{}}\PYG{l+s+s2}{Reference}\PYG{l+s+s2}{\PYGZdq{}}\PYG{p}{,} \PYG{n}{title}\PYG{o}{=}\PYG{k+kc}{None}\PYG{p}{)}\PYG{p}{:} 
  \PYG{n}{c}\PYG{o}{=}\PYG{p}{[}\PYG{p}{]}
  \PYG{k}{for} \PYG{n}{i} \PYG{o+ow}{in} \PYG{n+nb}{range}\PYG{p}{(}\PYG{l+m+mi}{3}\PYG{p}{)}\PYG{p}{:}
    \PYG{n}{b} \PYG{o}{=} \PYG{n}{np}\PYG{o}{.}\PYG{n}{flipud}\PYG{p}{(} \PYG{n}{np}\PYG{o}{.}\PYG{n}{concatenate}\PYG{p}{(}\PYG{p}{(}\PYG{n}{a2}\PYG{p}{[}\PYG{n}{i}\PYG{p}{]}\PYG{p}{,}\PYG{n}{a1}\PYG{p}{[}\PYG{n}{i}\PYG{p}{]}\PYG{p}{)}\PYG{p}{,}\PYG{n}{axis}\PYG{o}{=}\PYG{l+m+mi}{1}\PYG{p}{)}\PYG{o}{.}\PYG{n}{transpose}\PYG{p}{(}\PYG{p}{)}\PYG{p}{)}
    \PYG{n+nb}{min}\PYG{p}{,} \PYG{n}{mean}\PYG{p}{,} \PYG{n+nb}{max} \PYG{o}{=} \PYG{n}{np}\PYG{o}{.}\PYG{n}{min}\PYG{p}{(}\PYG{n}{b}\PYG{p}{)}\PYG{p}{,} \PYG{n}{np}\PYG{o}{.}\PYG{n}{mean}\PYG{p}{(}\PYG{n}{b}\PYG{p}{)}\PYG{p}{,} \PYG{n}{np}\PYG{o}{.}\PYG{n}{max}\PYG{p}{(}\PYG{n}{b}\PYG{p}{)}\PYG{p}{;} 
    \PYG{k}{if} \PYG{n}{stats}\PYG{p}{:} \PYG{n+nb}{print}\PYG{p}{(}\PYG{l+s+s2}{\PYGZdq{}}\PYG{l+s+s2}{Stats }\PYG{l+s+si}{\PYGZpc{}d}\PYG{l+s+s2}{: }\PYG{l+s+s2}{\PYGZdq{}}\PYG{o}{\PYGZpc{}}\PYG{k}{i} + format([min,mean,max]))
    \PYG{n}{b} \PYG{o}{\PYGZhy{}}\PYG{o}{=} \PYG{n+nb}{min}\PYG{p}{;} \PYG{n}{b} \PYG{o}{/}\PYG{o}{=} \PYG{p}{(}\PYG{n+nb}{max}\PYG{o}{\PYGZhy{}}\PYG{n+nb}{min}\PYG{p}{)}
    \PYG{n}{c}\PYG{o}{.}\PYG{n}{append}\PYG{p}{(}\PYG{n}{b}\PYG{p}{)}
  \PYG{n}{fig}\PYG{p}{,} \PYG{n}{axes} \PYG{o}{=} \PYG{n}{pylab}\PYG{o}{.}\PYG{n}{subplots}\PYG{p}{(}\PYG{l+m+mi}{1}\PYG{p}{,} \PYG{l+m+mi}{1}\PYG{p}{,} \PYG{n}{figsize}\PYG{o}{=}\PYG{p}{(}\PYG{l+m+mi}{16}\PYG{p}{,} \PYG{l+m+mi}{5}\PYG{p}{)}\PYG{p}{)}
  \PYG{n}{axes}\PYG{o}{.}\PYG{n}{set\PYGZus{}xticks}\PYG{p}{(}\PYG{p}{[}\PYG{p}{]}\PYG{p}{)}\PYG{p}{;} \PYG{n}{axes}\PYG{o}{.}\PYG{n}{set\PYGZus{}yticks}\PYG{p}{(}\PYG{p}{[}\PYG{p}{]}\PYG{p}{)}\PYG{p}{;} 
  \PYG{n}{im} \PYG{o}{=} \PYG{n}{axes}\PYG{o}{.}\PYG{n}{imshow}\PYG{p}{(}\PYG{n}{np}\PYG{o}{.}\PYG{n}{concatenate}\PYG{p}{(}\PYG{n}{c}\PYG{p}{,}\PYG{n}{axis}\PYG{o}{=}\PYG{l+m+mi}{1}\PYG{p}{)}\PYG{p}{,} \PYG{n}{origin}\PYG{o}{=}\PYG{l+s+s1}{\PYGZsq{}}\PYG{l+s+s1}{upper}\PYG{l+s+s1}{\PYGZsq{}}\PYG{p}{,} \PYG{n}{cmap}\PYG{o}{=}\PYG{l+s+s1}{\PYGZsq{}}\PYG{l+s+s1}{magma}\PYG{l+s+s1}{\PYGZsq{}}\PYG{p}{)}

  \PYG{n}{pylab}\PYG{o}{.}\PYG{n}{colorbar}\PYG{p}{(}\PYG{n}{im}\PYG{p}{)}\PYG{p}{;} \PYG{n}{pylab}\PYG{o}{.}\PYG{n}{xlabel}\PYG{p}{(}\PYG{l+s+s1}{\PYGZsq{}}\PYG{l+s+s1}{p, ux, uy}\PYG{l+s+s1}{\PYGZsq{}}\PYG{p}{)}\PYG{p}{;} \PYG{n}{pylab}\PYG{o}{.}\PYG{n}{ylabel}\PYG{p}{(}\PYG{l+s+s1}{\PYGZsq{}}\PYG{l+s+si}{\PYGZpc{}s}\PYG{l+s+s1}{           }\PYG{l+s+si}{\PYGZpc{}s}\PYG{l+s+s1}{\PYGZsq{}}\PYG{o}{\PYGZpc{}}\PYG{p}{(}\PYG{n}{bottom}\PYG{p}{,}\PYG{n}{top}\PYG{p}{)}\PYG{p}{)}
  \PYG{k}{if} \PYG{n}{title} \PYG{o+ow}{is} \PYG{o+ow}{not} \PYG{k+kc}{None}\PYG{p}{:} \PYG{n}{pylab}\PYG{o}{.}\PYG{n}{title}\PYG{p}{(}\PYG{n}{title}\PYG{p}{)}

\PYG{n}{NUM}\PYG{o}{=}\PYG{l+m+mi}{72}
\PYG{n}{showSbs}\PYG{p}{(}\PYG{n}{npfile}\PYG{p}{[}\PYG{l+s+s2}{\PYGZdq{}}\PYG{l+s+s2}{inputs}\PYG{l+s+s2}{\PYGZdq{}}\PYG{p}{]}\PYG{p}{[}\PYG{n}{NUM}\PYG{p}{]}\PYG{p}{,}\PYG{n}{npfile}\PYG{p}{[}\PYG{l+s+s2}{\PYGZdq{}}\PYG{l+s+s2}{targets}\PYG{l+s+s2}{\PYGZdq{}}\PYG{p}{]}\PYG{p}{[}\PYG{n}{NUM}\PYG{p}{]}\PYG{p}{,} \PYG{n}{stats}\PYG{o}{=}\PYG{k+kc}{False}\PYG{p}{,} \PYG{n}{bottom}\PYG{o}{=}\PYG{l+s+s2}{\PYGZdq{}}\PYG{l+s+s2}{Target Output}\PYG{l+s+s2}{\PYGZdq{}}\PYG{p}{,} \PYG{n}{top}\PYG{o}{=}\PYG{l+s+s2}{\PYGZdq{}}\PYG{l+s+s2}{Inputs}\PYG{l+s+s2}{\PYGZdq{}}\PYG{p}{,} \PYG{n}{title}\PYG{o}{=}\PYG{l+s+s2}{\PYGZdq{}}\PYG{l+s+s2}{3 inputs are shown at the top (mask, in\PYGZhy{}ux, in\PYGZhy{}uy), with the 3 output channels (p,ux,uy) at the bottom}\PYG{l+s+s2}{\PYGZdq{}}\PYG{p}{)}
\end{sphinxVerbatim}

\noindent\sphinxincludegraphics{{supervised-airfoils_6_0}.png}

Next, let’s define a small helper class \sphinxcode{\sphinxupquote{DfpDataset}} to organize inputs and targets. We’ll transfer the corresponding data to the pytorch \sphinxcode{\sphinxupquote{DataLoader}} class.

We also set up some globals to control training parameters, maybe most importantly: the learning rate \sphinxcode{\sphinxupquote{LR}}, i.e. \(\eta\) from the previous setions. When your training run doesn’t converge this is the first parameter to experiment with.

Here, we’ll keep it relatively small throughout. (Using \sphinxstyleemphasis{learning rate decay} would be better, i.e. potentially give an improed convergence, but is omitted here for clarity.)

\begin{sphinxVerbatim}[commandchars=\\\{\}]
\PYG{c+c1}{\PYGZsh{} some global training constants}

\PYG{c+c1}{\PYGZsh{} number of training epochs}
\PYG{n}{EPOCHS} \PYG{o}{=} \PYG{l+m+mi}{100}
\PYG{c+c1}{\PYGZsh{} batch size}
\PYG{n}{BATCH\PYGZus{}SIZE} \PYG{o}{=} \PYG{l+m+mi}{10}
\PYG{c+c1}{\PYGZsh{} learning rate}
\PYG{n}{LR} \PYG{o}{=} \PYG{l+m+mf}{0.00002}

\PYG{k}{class} \PYG{n+nc}{DfpDataset}\PYG{p}{(}\PYG{p}{)}\PYG{p}{:}
    \PYG{k}{def} \PYG{n+nf+fm}{\PYGZus{}\PYGZus{}init\PYGZus{}\PYGZus{}}\PYG{p}{(}\PYG{n+nb+bp}{self}\PYG{p}{,} \PYG{n}{inputs}\PYG{p}{,}\PYG{n}{targets}\PYG{p}{)}\PYG{p}{:} 
        \PYG{n+nb+bp}{self}\PYG{o}{.}\PYG{n}{inputs}  \PYG{o}{=} \PYG{n}{inputs}
        \PYG{n+nb+bp}{self}\PYG{o}{.}\PYG{n}{targets} \PYG{o}{=} \PYG{n}{targets}

    \PYG{k}{def} \PYG{n+nf+fm}{\PYGZus{}\PYGZus{}len\PYGZus{}\PYGZus{}}\PYG{p}{(}\PYG{n+nb+bp}{self}\PYG{p}{)}\PYG{p}{:}
        \PYG{k}{return} \PYG{n+nb}{len}\PYG{p}{(}\PYG{n+nb+bp}{self}\PYG{o}{.}\PYG{n}{inputs}\PYG{p}{)}

    \PYG{k}{def} \PYG{n+nf+fm}{\PYGZus{}\PYGZus{}getitem\PYGZus{}\PYGZus{}}\PYG{p}{(}\PYG{n+nb+bp}{self}\PYG{p}{,} \PYG{n}{idx}\PYG{p}{)}\PYG{p}{:}
        \PYG{k}{return} \PYG{n+nb+bp}{self}\PYG{o}{.}\PYG{n}{inputs}\PYG{p}{[}\PYG{n}{idx}\PYG{p}{]}\PYG{p}{,} \PYG{n+nb+bp}{self}\PYG{o}{.}\PYG{n}{targets}\PYG{p}{[}\PYG{n}{idx}\PYG{p}{]}

\PYG{n}{tdata} \PYG{o}{=} \PYG{n}{DfpDataset}\PYG{p}{(}\PYG{n}{npfile}\PYG{p}{[}\PYG{l+s+s2}{\PYGZdq{}}\PYG{l+s+s2}{inputs}\PYG{l+s+s2}{\PYGZdq{}}\PYG{p}{]}\PYG{p}{,}\PYG{n}{npfile}\PYG{p}{[}\PYG{l+s+s2}{\PYGZdq{}}\PYG{l+s+s2}{targets}\PYG{l+s+s2}{\PYGZdq{}}\PYG{p}{]}\PYG{p}{)}
\PYG{n}{vdata} \PYG{o}{=} \PYG{n}{DfpDataset}\PYG{p}{(}\PYG{n}{npfile}\PYG{p}{[}\PYG{l+s+s2}{\PYGZdq{}}\PYG{l+s+s2}{vinputs}\PYG{l+s+s2}{\PYGZdq{}}\PYG{p}{]}\PYG{p}{,}\PYG{n}{npfile}\PYG{p}{[}\PYG{l+s+s2}{\PYGZdq{}}\PYG{l+s+s2}{vtargets}\PYG{l+s+s2}{\PYGZdq{}}\PYG{p}{]}\PYG{p}{)}

\PYG{n}{trainLoader} \PYG{o}{=} \PYG{n}{torch}\PYG{o}{.}\PYG{n}{utils}\PYG{o}{.}\PYG{n}{data}\PYG{o}{.}\PYG{n}{DataLoader}\PYG{p}{(}\PYG{n}{tdata}\PYG{p}{,} \PYG{n}{batch\PYGZus{}size}\PYG{o}{=}\PYG{n}{BATCH\PYGZus{}SIZE}\PYG{p}{,} \PYG{n}{shuffle}\PYG{o}{=}\PYG{k+kc}{True} \PYG{p}{,} \PYG{n}{drop\PYGZus{}last}\PYG{o}{=}\PYG{k+kc}{True}\PYG{p}{)} 
\PYG{n}{valiLoader}  \PYG{o}{=} \PYG{n}{torch}\PYG{o}{.}\PYG{n}{utils}\PYG{o}{.}\PYG{n}{data}\PYG{o}{.}\PYG{n}{DataLoader}\PYG{p}{(}\PYG{n}{vdata}\PYG{p}{,} \PYG{n}{batch\PYGZus{}size}\PYG{o}{=}\PYG{n}{BATCH\PYGZus{}SIZE}\PYG{p}{,} \PYG{n}{shuffle}\PYG{o}{=}\PYG{k+kc}{False}\PYG{p}{,} \PYG{n}{drop\PYGZus{}last}\PYG{o}{=}\PYG{k+kc}{True}\PYG{p}{)} 

\PYG{n+nb}{print}\PYG{p}{(}\PYG{l+s+s2}{\PYGZdq{}}\PYG{l+s+s2}{Training \PYGZam{} validation batches: }\PYG{l+s+si}{\PYGZob{}\PYGZcb{}}\PYG{l+s+s2}{ , }\PYG{l+s+si}{\PYGZob{}\PYGZcb{}}\PYG{l+s+s2}{\PYGZdq{}}\PYG{o}{.}\PYG{n}{format}\PYG{p}{(}\PYG{n+nb}{len}\PYG{p}{(}\PYG{n}{trainLoader}\PYG{p}{)}\PYG{p}{,}\PYG{n+nb}{len}\PYG{p}{(}\PYG{n}{valiLoader}\PYG{p}{)} \PYG{p}{)}\PYG{p}{)}
\end{sphinxVerbatim}

\begin{sphinxVerbatim}[commandchars=\\\{\}]
Training \PYGZam{} validation batches: 32 , 8
\end{sphinxVerbatim}

\subsection{Network setup}
\label{\detokenize{supervised-airfoils:network-setup}}
Now we can set up the architecture of our neural network, we’ll use a fully convolutional U\sphinxhyphen{}net. This is a widely used architecture that uses a stack of convolutions across different spatial resolutions. The main deviation from a regular conv\sphinxhyphen{}net is to introduce \sphinxstyleemphasis{skip connection} from the encoder to the decoder part. This ensures that no information is lost during feature extraction. (Note that this only works if the network is to be used as a whole. It doesn’t work in situations where we’d, e.g., want to use the decoder as a standalone component.)

Here’s a overview of the architecure:

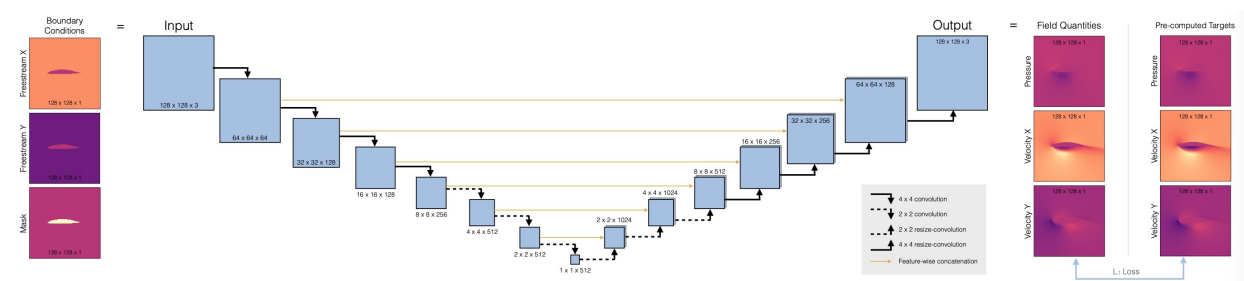
\sphinxincludegraphics{{supervised-airfoils-unet}.jpg}

First, we’ll define a helper to set up a convolutional block in the network, \sphinxcode{\sphinxupquote{blockUNet}}. Note, we don’t use any pooling! Instead we use strides and transpose convolutions (these need to be symmetric for the decoder part, i.e. have an uneven kernel size), following \sphinxhref{https://distill.pub/2016/deconv-checkerboard/}{best practices}. The full pytroch neural network is managed via the \sphinxcode{\sphinxupquote{DfpNet}} class.

\begin{sphinxVerbatim}[commandchars=\\\{\}]
\PYG{k+kn}{import} \PYG{n+nn}{os}\PYG{o}{,} \PYG{n+nn}{sys}\PYG{o}{,} \PYG{n+nn}{random}
\PYG{k+kn}{import} \PYG{n+nn}{numpy} \PYG{k}{as} \PYG{n+nn}{np}

\PYG{k+kn}{import} \PYG{n+nn}{torch}
\PYG{k+kn}{import} \PYG{n+nn}{torch}\PYG{n+nn}{.}\PYG{n+nn}{nn} \PYG{k}{as} \PYG{n+nn}{nn}
\PYG{k+kn}{import} \PYG{n+nn}{torch}\PYG{n+nn}{.}\PYG{n+nn}{optim} \PYG{k}{as} \PYG{n+nn}{optim}
\PYG{k+kn}{import} \PYG{n+nn}{torch}\PYG{n+nn}{.}\PYG{n+nn}{autograd} 
\PYG{k+kn}{import} \PYG{n+nn}{torch}\PYG{n+nn}{.}\PYG{n+nn}{utils}\PYG{n+nn}{.}\PYG{n+nn}{data} 

\PYG{k}{def} \PYG{n+nf}{blockUNet}\PYG{p}{(}\PYG{n}{in\PYGZus{}c}\PYG{p}{,} \PYG{n}{out\PYGZus{}c}\PYG{p}{,} \PYG{n}{name}\PYG{p}{,} \PYG{n}{transposed}\PYG{o}{=}\PYG{k+kc}{False}\PYG{p}{,} \PYG{n}{bn}\PYG{o}{=}\PYG{k+kc}{True}\PYG{p}{,} \PYG{n}{relu}\PYG{o}{=}\PYG{k+kc}{True}\PYG{p}{,} \PYG{n}{size}\PYG{o}{=}\PYG{l+m+mi}{4}\PYG{p}{,} \PYG{n}{pad}\PYG{o}{=}\PYG{l+m+mi}{1}\PYG{p}{,} \PYG{n}{dropout}\PYG{o}{=}\PYG{l+m+mf}{0.}\PYG{p}{)}\PYG{p}{:}
    \PYG{n}{block} \PYG{o}{=} \PYG{n}{nn}\PYG{o}{.}\PYG{n}{Sequential}\PYG{p}{(}\PYG{p}{)}

    \PYG{k}{if} \PYG{n}{relu}\PYG{p}{:}
        \PYG{n}{block}\PYG{o}{.}\PYG{n}{add\PYGZus{}module}\PYG{p}{(}\PYG{l+s+s1}{\PYGZsq{}}\PYG{l+s+si}{\PYGZpc{}s}\PYG{l+s+s1}{\PYGZus{}relu}\PYG{l+s+s1}{\PYGZsq{}} \PYG{o}{\PYGZpc{}} \PYG{n}{name}\PYG{p}{,} \PYG{n}{nn}\PYG{o}{.}\PYG{n}{ReLU}\PYG{p}{(}\PYG{n}{inplace}\PYG{o}{=}\PYG{k+kc}{True}\PYG{p}{)}\PYG{p}{)}
    \PYG{k}{else}\PYG{p}{:}
        \PYG{n}{block}\PYG{o}{.}\PYG{n}{add\PYGZus{}module}\PYG{p}{(}\PYG{l+s+s1}{\PYGZsq{}}\PYG{l+s+si}{\PYGZpc{}s}\PYG{l+s+s1}{\PYGZus{}leakyrelu}\PYG{l+s+s1}{\PYGZsq{}} \PYG{o}{\PYGZpc{}} \PYG{n}{name}\PYG{p}{,} \PYG{n}{nn}\PYG{o}{.}\PYG{n}{LeakyReLU}\PYG{p}{(}\PYG{l+m+mf}{0.2}\PYG{p}{,} \PYG{n}{inplace}\PYG{o}{=}\PYG{k+kc}{True}\PYG{p}{)}\PYG{p}{)}

    \PYG{k}{if} \PYG{o+ow}{not} \PYG{n}{transposed}\PYG{p}{:}
        \PYG{n}{block}\PYG{o}{.}\PYG{n}{add\PYGZus{}module}\PYG{p}{(}\PYG{l+s+s1}{\PYGZsq{}}\PYG{l+s+si}{\PYGZpc{}s}\PYG{l+s+s1}{\PYGZus{}conv}\PYG{l+s+s1}{\PYGZsq{}} \PYG{o}{\PYGZpc{}} \PYG{n}{name}\PYG{p}{,} \PYG{n}{nn}\PYG{o}{.}\PYG{n}{Conv2d}\PYG{p}{(}\PYG{n}{in\PYGZus{}c}\PYG{p}{,} \PYG{n}{out\PYGZus{}c}\PYG{p}{,} \PYG{n}{kernel\PYGZus{}size}\PYG{o}{=}\PYG{n}{size}\PYG{p}{,} \PYG{n}{stride}\PYG{o}{=}\PYG{l+m+mi}{2}\PYG{p}{,} \PYG{n}{padding}\PYG{o}{=}\PYG{n}{pad}\PYG{p}{,} \PYG{n}{bias}\PYG{o}{=}\PYG{k+kc}{True}\PYG{p}{)}\PYG{p}{)}
    \PYG{k}{else}\PYG{p}{:}
        \PYG{n}{block}\PYG{o}{.}\PYG{n}{add\PYGZus{}module}\PYG{p}{(}\PYG{l+s+s1}{\PYGZsq{}}\PYG{l+s+si}{\PYGZpc{}s}\PYG{l+s+s1}{\PYGZus{}upsam}\PYG{l+s+s1}{\PYGZsq{}} \PYG{o}{\PYGZpc{}} \PYG{n}{name}\PYG{p}{,} \PYG{n}{nn}\PYG{o}{.}\PYG{n}{Upsample}\PYG{p}{(}\PYG{n}{scale\PYGZus{}factor}\PYG{o}{=}\PYG{l+m+mi}{2}\PYG{p}{,} \PYG{n}{mode}\PYG{o}{=}\PYG{l+s+s1}{\PYGZsq{}}\PYG{l+s+s1}{bilinear}\PYG{l+s+s1}{\PYGZsq{}}\PYG{p}{)}\PYG{p}{)}
        \PYG{c+c1}{\PYGZsh{} reduce kernel size by one for the upsampling (ie decoder part)}
        \PYG{n}{block}\PYG{o}{.}\PYG{n}{add\PYGZus{}module}\PYG{p}{(}\PYG{l+s+s1}{\PYGZsq{}}\PYG{l+s+si}{\PYGZpc{}s}\PYG{l+s+s1}{\PYGZus{}tconv}\PYG{l+s+s1}{\PYGZsq{}} \PYG{o}{\PYGZpc{}} \PYG{n}{name}\PYG{p}{,} \PYG{n}{nn}\PYG{o}{.}\PYG{n}{Conv2d}\PYG{p}{(}\PYG{n}{in\PYGZus{}c}\PYG{p}{,} \PYG{n}{out\PYGZus{}c}\PYG{p}{,} \PYG{n}{kernel\PYGZus{}size}\PYG{o}{=}\PYG{p}{(}\PYG{n}{size}\PYG{o}{\PYGZhy{}}\PYG{l+m+mi}{1}\PYG{p}{)}\PYG{p}{,} \PYG{n}{stride}\PYG{o}{=}\PYG{l+m+mi}{1}\PYG{p}{,} \PYG{n}{padding}\PYG{o}{=}\PYG{n}{pad}\PYG{p}{,} \PYG{n}{bias}\PYG{o}{=}\PYG{k+kc}{True}\PYG{p}{)}\PYG{p}{)}

    \PYG{k}{if} \PYG{n}{bn}\PYG{p}{:}
        \PYG{n}{block}\PYG{o}{.}\PYG{n}{add\PYGZus{}module}\PYG{p}{(}\PYG{l+s+s1}{\PYGZsq{}}\PYG{l+s+si}{\PYGZpc{}s}\PYG{l+s+s1}{\PYGZus{}bn}\PYG{l+s+s1}{\PYGZsq{}} \PYG{o}{\PYGZpc{}} \PYG{n}{name}\PYG{p}{,} \PYG{n}{nn}\PYG{o}{.}\PYG{n}{BatchNorm2d}\PYG{p}{(}\PYG{n}{out\PYGZus{}c}\PYG{p}{)}\PYG{p}{)}
    \PYG{k}{if} \PYG{n}{dropout}\PYG{o}{\PYGZgt{}}\PYG{l+m+mf}{0.}\PYG{p}{:}
        \PYG{n}{block}\PYG{o}{.}\PYG{n}{add\PYGZus{}module}\PYG{p}{(}\PYG{l+s+s1}{\PYGZsq{}}\PYG{l+s+si}{\PYGZpc{}s}\PYG{l+s+s1}{\PYGZus{}dropout}\PYG{l+s+s1}{\PYGZsq{}} \PYG{o}{\PYGZpc{}} \PYG{n}{name}\PYG{p}{,} \PYG{n}{nn}\PYG{o}{.}\PYG{n}{Dropout2d}\PYG{p}{(} \PYG{n}{dropout}\PYG{p}{,} \PYG{n}{inplace}\PYG{o}{=}\PYG{k+kc}{True}\PYG{p}{)}\PYG{p}{)}

    \PYG{k}{return} \PYG{n}{block}
    
\PYG{k}{class} \PYG{n+nc}{DfpNet}\PYG{p}{(}\PYG{n}{nn}\PYG{o}{.}\PYG{n}{Module}\PYG{p}{)}\PYG{p}{:}
    \PYG{k}{def} \PYG{n+nf+fm}{\PYGZus{}\PYGZus{}init\PYGZus{}\PYGZus{}}\PYG{p}{(}\PYG{n+nb+bp}{self}\PYG{p}{,} \PYG{n}{channelExponent}\PYG{o}{=}\PYG{l+m+mi}{6}\PYG{p}{,} \PYG{n}{dropout}\PYG{o}{=}\PYG{l+m+mf}{0.}\PYG{p}{)}\PYG{p}{:}
        \PYG{n+nb}{super}\PYG{p}{(}\PYG{n}{DfpNet}\PYG{p}{,} \PYG{n+nb+bp}{self}\PYG{p}{)}\PYG{o}{.}\PYG{n+nf+fm}{\PYGZus{}\PYGZus{}init\PYGZus{}\PYGZus{}}\PYG{p}{(}\PYG{p}{)}
        \PYG{n}{channels} \PYG{o}{=} \PYG{n+nb}{int}\PYG{p}{(}\PYG{l+m+mi}{2} \PYG{o}{*}\PYG{o}{*} \PYG{n}{channelExponent} \PYG{o}{+} \PYG{l+m+mf}{0.5}\PYG{p}{)}

        \PYG{n+nb+bp}{self}\PYG{o}{.}\PYG{n}{layer1} \PYG{o}{=} \PYG{n}{nn}\PYG{o}{.}\PYG{n}{Sequential}\PYG{p}{(}\PYG{p}{)}
        \PYG{n+nb+bp}{self}\PYG{o}{.}\PYG{n}{layer1}\PYG{o}{.}\PYG{n}{add\PYGZus{}module}\PYG{p}{(}\PYG{l+s+s1}{\PYGZsq{}}\PYG{l+s+s1}{layer1}\PYG{l+s+s1}{\PYGZsq{}}\PYG{p}{,} \PYG{n}{nn}\PYG{o}{.}\PYG{n}{Conv2d}\PYG{p}{(}\PYG{l+m+mi}{3}\PYG{p}{,} \PYG{n}{channels}\PYG{p}{,} \PYG{l+m+mi}{4}\PYG{p}{,} \PYG{l+m+mi}{2}\PYG{p}{,} \PYG{l+m+mi}{1}\PYG{p}{,} \PYG{n}{bias}\PYG{o}{=}\PYG{k+kc}{True}\PYG{p}{)}\PYG{p}{)}

        \PYG{n+nb+bp}{self}\PYG{o}{.}\PYG{n}{layer2} \PYG{o}{=} \PYG{n}{blockUNet}\PYG{p}{(}\PYG{n}{channels}  \PYG{p}{,} \PYG{n}{channels}\PYG{o}{*}\PYG{l+m+mi}{2}\PYG{p}{,} \PYG{l+s+s1}{\PYGZsq{}}\PYG{l+s+s1}{enc\PYGZus{}layer2}\PYG{l+s+s1}{\PYGZsq{}}\PYG{p}{,} \PYG{n}{transposed}\PYG{o}{=}\PYG{k+kc}{False}\PYG{p}{,} \PYG{n}{bn}\PYG{o}{=}\PYG{k+kc}{True}\PYG{p}{,} \PYG{n}{relu}\PYG{o}{=}\PYG{k+kc}{False}\PYG{p}{,} \PYG{n}{dropout}\PYG{o}{=}\PYG{n}{dropout} \PYG{p}{)}
        \PYG{n+nb+bp}{self}\PYG{o}{.}\PYG{n}{layer3} \PYG{o}{=} \PYG{n}{blockUNet}\PYG{p}{(}\PYG{n}{channels}\PYG{o}{*}\PYG{l+m+mi}{2}\PYG{p}{,} \PYG{n}{channels}\PYG{o}{*}\PYG{l+m+mi}{2}\PYG{p}{,} \PYG{l+s+s1}{\PYGZsq{}}\PYG{l+s+s1}{enc\PYGZus{}layer3}\PYG{l+s+s1}{\PYGZsq{}}\PYG{p}{,} \PYG{n}{transposed}\PYG{o}{=}\PYG{k+kc}{False}\PYG{p}{,} \PYG{n}{bn}\PYG{o}{=}\PYG{k+kc}{True}\PYG{p}{,} \PYG{n}{relu}\PYG{o}{=}\PYG{k+kc}{False}\PYG{p}{,} \PYG{n}{dropout}\PYG{o}{=}\PYG{n}{dropout} \PYG{p}{)}
        \PYG{n+nb+bp}{self}\PYG{o}{.}\PYG{n}{layer4} \PYG{o}{=} \PYG{n}{blockUNet}\PYG{p}{(}\PYG{n}{channels}\PYG{o}{*}\PYG{l+m+mi}{2}\PYG{p}{,} \PYG{n}{channels}\PYG{o}{*}\PYG{l+m+mi}{4}\PYG{p}{,} \PYG{l+s+s1}{\PYGZsq{}}\PYG{l+s+s1}{enc\PYGZus{}layer4}\PYG{l+s+s1}{\PYGZsq{}}\PYG{p}{,} \PYG{n}{transposed}\PYG{o}{=}\PYG{k+kc}{False}\PYG{p}{,} \PYG{n}{bn}\PYG{o}{=}\PYG{k+kc}{True}\PYG{p}{,} \PYG{n}{relu}\PYG{o}{=}\PYG{k+kc}{False}\PYG{p}{,} \PYG{n}{dropout}\PYG{o}{=}\PYG{n}{dropout} \PYG{p}{)}
        \PYG{n+nb+bp}{self}\PYG{o}{.}\PYG{n}{layer5} \PYG{o}{=} \PYG{n}{blockUNet}\PYG{p}{(}\PYG{n}{channels}\PYG{o}{*}\PYG{l+m+mi}{4}\PYG{p}{,} \PYG{n}{channels}\PYG{o}{*}\PYG{l+m+mi}{8}\PYG{p}{,} \PYG{l+s+s1}{\PYGZsq{}}\PYG{l+s+s1}{enc\PYGZus{}layer5}\PYG{l+s+s1}{\PYGZsq{}}\PYG{p}{,} \PYG{n}{transposed}\PYG{o}{=}\PYG{k+kc}{False}\PYG{p}{,} \PYG{n}{bn}\PYG{o}{=}\PYG{k+kc}{True}\PYG{p}{,} \PYG{n}{relu}\PYG{o}{=}\PYG{k+kc}{False}\PYG{p}{,} \PYG{n}{dropout}\PYG{o}{=}\PYG{n}{dropout} \PYG{p}{)} 
        \PYG{n+nb+bp}{self}\PYG{o}{.}\PYG{n}{layer6} \PYG{o}{=} \PYG{n}{blockUNet}\PYG{p}{(}\PYG{n}{channels}\PYG{o}{*}\PYG{l+m+mi}{8}\PYG{p}{,} \PYG{n}{channels}\PYG{o}{*}\PYG{l+m+mi}{8}\PYG{p}{,} \PYG{l+s+s1}{\PYGZsq{}}\PYG{l+s+s1}{enc\PYGZus{}layer6}\PYG{l+s+s1}{\PYGZsq{}}\PYG{p}{,} \PYG{n}{transposed}\PYG{o}{=}\PYG{k+kc}{False}\PYG{p}{,} \PYG{n}{bn}\PYG{o}{=}\PYG{k+kc}{True}\PYG{p}{,} \PYG{n}{relu}\PYG{o}{=}\PYG{k+kc}{False}\PYG{p}{,} \PYG{n}{dropout}\PYG{o}{=}\PYG{n}{dropout} \PYG{p}{,} \PYG{n}{size}\PYG{o}{=}\PYG{l+m+mi}{2}\PYG{p}{,}\PYG{n}{pad}\PYG{o}{=}\PYG{l+m+mi}{0}\PYG{p}{)}
        \PYG{n+nb+bp}{self}\PYG{o}{.}\PYG{n}{layer7} \PYG{o}{=} \PYG{n}{blockUNet}\PYG{p}{(}\PYG{n}{channels}\PYG{o}{*}\PYG{l+m+mi}{8}\PYG{p}{,} \PYG{n}{channels}\PYG{o}{*}\PYG{l+m+mi}{8}\PYG{p}{,} \PYG{l+s+s1}{\PYGZsq{}}\PYG{l+s+s1}{enc\PYGZus{}layer7}\PYG{l+s+s1}{\PYGZsq{}}\PYG{p}{,} \PYG{n}{transposed}\PYG{o}{=}\PYG{k+kc}{False}\PYG{p}{,} \PYG{n}{bn}\PYG{o}{=}\PYG{k+kc}{True}\PYG{p}{,} \PYG{n}{relu}\PYG{o}{=}\PYG{k+kc}{False}\PYG{p}{,} \PYG{n}{dropout}\PYG{o}{=}\PYG{n}{dropout} \PYG{p}{,} \PYG{n}{size}\PYG{o}{=}\PYG{l+m+mi}{2}\PYG{p}{,}\PYG{n}{pad}\PYG{o}{=}\PYG{l+m+mi}{0}\PYG{p}{)}
     
        \PYG{c+c1}{\PYGZsh{} note, kernel size is internally reduced by one for the decoder part}
        \PYG{n+nb+bp}{self}\PYG{o}{.}\PYG{n}{dlayer7} \PYG{o}{=} \PYG{n}{blockUNet}\PYG{p}{(}\PYG{n}{channels}\PYG{o}{*}\PYG{l+m+mi}{8}\PYG{p}{,} \PYG{n}{channels}\PYG{o}{*}\PYG{l+m+mi}{8}\PYG{p}{,} \PYG{l+s+s1}{\PYGZsq{}}\PYG{l+s+s1}{dec\PYGZus{}layer7}\PYG{l+s+s1}{\PYGZsq{}}\PYG{p}{,} \PYG{n}{transposed}\PYG{o}{=}\PYG{k+kc}{True}\PYG{p}{,} \PYG{n}{bn}\PYG{o}{=}\PYG{k+kc}{True}\PYG{p}{,} \PYG{n}{relu}\PYG{o}{=}\PYG{k+kc}{True}\PYG{p}{,} \PYG{n}{dropout}\PYG{o}{=}\PYG{n}{dropout} \PYG{p}{,} \PYG{n}{size}\PYG{o}{=}\PYG{l+m+mi}{2}\PYG{p}{,}\PYG{n}{pad}\PYG{o}{=}\PYG{l+m+mi}{0}\PYG{p}{)}
        \PYG{n+nb+bp}{self}\PYG{o}{.}\PYG{n}{dlayer6} \PYG{o}{=} \PYG{n}{blockUNet}\PYG{p}{(}\PYG{n}{channels}\PYG{o}{*}\PYG{l+m+mi}{16}\PYG{p}{,}\PYG{n}{channels}\PYG{o}{*}\PYG{l+m+mi}{8}\PYG{p}{,} \PYG{l+s+s1}{\PYGZsq{}}\PYG{l+s+s1}{dec\PYGZus{}layer6}\PYG{l+s+s1}{\PYGZsq{}}\PYG{p}{,} \PYG{n}{transposed}\PYG{o}{=}\PYG{k+kc}{True}\PYG{p}{,} \PYG{n}{bn}\PYG{o}{=}\PYG{k+kc}{True}\PYG{p}{,} \PYG{n}{relu}\PYG{o}{=}\PYG{k+kc}{True}\PYG{p}{,} \PYG{n}{dropout}\PYG{o}{=}\PYG{n}{dropout} \PYG{p}{,} \PYG{n}{size}\PYG{o}{=}\PYG{l+m+mi}{2}\PYG{p}{,}\PYG{n}{pad}\PYG{o}{=}\PYG{l+m+mi}{0}\PYG{p}{)}
        \PYG{n+nb+bp}{self}\PYG{o}{.}\PYG{n}{dlayer5} \PYG{o}{=} \PYG{n}{blockUNet}\PYG{p}{(}\PYG{n}{channels}\PYG{o}{*}\PYG{l+m+mi}{16}\PYG{p}{,}\PYG{n}{channels}\PYG{o}{*}\PYG{l+m+mi}{4}\PYG{p}{,} \PYG{l+s+s1}{\PYGZsq{}}\PYG{l+s+s1}{dec\PYGZus{}layer5}\PYG{l+s+s1}{\PYGZsq{}}\PYG{p}{,} \PYG{n}{transposed}\PYG{o}{=}\PYG{k+kc}{True}\PYG{p}{,} \PYG{n}{bn}\PYG{o}{=}\PYG{k+kc}{True}\PYG{p}{,} \PYG{n}{relu}\PYG{o}{=}\PYG{k+kc}{True}\PYG{p}{,} \PYG{n}{dropout}\PYG{o}{=}\PYG{n}{dropout} \PYG{p}{)} 
        \PYG{n+nb+bp}{self}\PYG{o}{.}\PYG{n}{dlayer4} \PYG{o}{=} \PYG{n}{blockUNet}\PYG{p}{(}\PYG{n}{channels}\PYG{o}{*}\PYG{l+m+mi}{8}\PYG{p}{,} \PYG{n}{channels}\PYG{o}{*}\PYG{l+m+mi}{2}\PYG{p}{,} \PYG{l+s+s1}{\PYGZsq{}}\PYG{l+s+s1}{dec\PYGZus{}layer4}\PYG{l+s+s1}{\PYGZsq{}}\PYG{p}{,} \PYG{n}{transposed}\PYG{o}{=}\PYG{k+kc}{True}\PYG{p}{,} \PYG{n}{bn}\PYG{o}{=}\PYG{k+kc}{True}\PYG{p}{,} \PYG{n}{relu}\PYG{o}{=}\PYG{k+kc}{True}\PYG{p}{,} \PYG{n}{dropout}\PYG{o}{=}\PYG{n}{dropout} \PYG{p}{)}
        \PYG{n+nb+bp}{self}\PYG{o}{.}\PYG{n}{dlayer3} \PYG{o}{=} \PYG{n}{blockUNet}\PYG{p}{(}\PYG{n}{channels}\PYG{o}{*}\PYG{l+m+mi}{4}\PYG{p}{,} \PYG{n}{channels}\PYG{o}{*}\PYG{l+m+mi}{2}\PYG{p}{,} \PYG{l+s+s1}{\PYGZsq{}}\PYG{l+s+s1}{dec\PYGZus{}layer3}\PYG{l+s+s1}{\PYGZsq{}}\PYG{p}{,} \PYG{n}{transposed}\PYG{o}{=}\PYG{k+kc}{True}\PYG{p}{,} \PYG{n}{bn}\PYG{o}{=}\PYG{k+kc}{True}\PYG{p}{,} \PYG{n}{relu}\PYG{o}{=}\PYG{k+kc}{True}\PYG{p}{,} \PYG{n}{dropout}\PYG{o}{=}\PYG{n}{dropout} \PYG{p}{)}
        \PYG{n+nb+bp}{self}\PYG{o}{.}\PYG{n}{dlayer2} \PYG{o}{=} \PYG{n}{blockUNet}\PYG{p}{(}\PYG{n}{channels}\PYG{o}{*}\PYG{l+m+mi}{4}\PYG{p}{,} \PYG{n}{channels}  \PYG{p}{,} \PYG{l+s+s1}{\PYGZsq{}}\PYG{l+s+s1}{dec\PYGZus{}layer2}\PYG{l+s+s1}{\PYGZsq{}}\PYG{p}{,} \PYG{n}{transposed}\PYG{o}{=}\PYG{k+kc}{True}\PYG{p}{,} \PYG{n}{bn}\PYG{o}{=}\PYG{k+kc}{True}\PYG{p}{,} \PYG{n}{relu}\PYG{o}{=}\PYG{k+kc}{True}\PYG{p}{,} \PYG{n}{dropout}\PYG{o}{=}\PYG{n}{dropout} \PYG{p}{)}

        \PYG{n+nb+bp}{self}\PYG{o}{.}\PYG{n}{dlayer1} \PYG{o}{=} \PYG{n}{nn}\PYG{o}{.}\PYG{n}{Sequential}\PYG{p}{(}\PYG{p}{)}
        \PYG{n+nb+bp}{self}\PYG{o}{.}\PYG{n}{dlayer1}\PYG{o}{.}\PYG{n}{add\PYGZus{}module}\PYG{p}{(}\PYG{l+s+s1}{\PYGZsq{}}\PYG{l+s+s1}{dec\PYGZus{}layer1\PYGZus{}relu}\PYG{l+s+s1}{\PYGZsq{}}\PYG{p}{,} \PYG{n}{nn}\PYG{o}{.}\PYG{n}{ReLU}\PYG{p}{(}\PYG{n}{inplace}\PYG{o}{=}\PYG{k+kc}{True}\PYG{p}{)}\PYG{p}{)}
        \PYG{n+nb+bp}{self}\PYG{o}{.}\PYG{n}{dlayer1}\PYG{o}{.}\PYG{n}{add\PYGZus{}module}\PYG{p}{(}\PYG{l+s+s1}{\PYGZsq{}}\PYG{l+s+s1}{dec\PYGZus{}layer1\PYGZus{}tconv}\PYG{l+s+s1}{\PYGZsq{}}\PYG{p}{,} \PYG{n}{nn}\PYG{o}{.}\PYG{n}{ConvTranspose2d}\PYG{p}{(}\PYG{n}{channels}\PYG{o}{*}\PYG{l+m+mi}{2}\PYG{p}{,} \PYG{l+m+mi}{3}\PYG{p}{,} \PYG{l+m+mi}{4}\PYG{p}{,} \PYG{l+m+mi}{2}\PYG{p}{,} \PYG{l+m+mi}{1}\PYG{p}{,} \PYG{n}{bias}\PYG{o}{=}\PYG{k+kc}{True}\PYG{p}{)}\PYG{p}{)}

    \PYG{k}{def} \PYG{n+nf}{forward}\PYG{p}{(}\PYG{n+nb+bp}{self}\PYG{p}{,} \PYG{n}{x}\PYG{p}{)}\PYG{p}{:}
        \PYG{c+c1}{\PYGZsh{} note, this Unet stack could be allocated with a loop, of course... }
        \PYG{n}{out1} \PYG{o}{=} \PYG{n+nb+bp}{self}\PYG{o}{.}\PYG{n}{layer1}\PYG{p}{(}\PYG{n}{x}\PYG{p}{)}
        \PYG{n}{out2} \PYG{o}{=} \PYG{n+nb+bp}{self}\PYG{o}{.}\PYG{n}{layer2}\PYG{p}{(}\PYG{n}{out1}\PYG{p}{)}
        \PYG{n}{out3} \PYG{o}{=} \PYG{n+nb+bp}{self}\PYG{o}{.}\PYG{n}{layer3}\PYG{p}{(}\PYG{n}{out2}\PYG{p}{)}
        \PYG{n}{out4} \PYG{o}{=} \PYG{n+nb+bp}{self}\PYG{o}{.}\PYG{n}{layer4}\PYG{p}{(}\PYG{n}{out3}\PYG{p}{)}
        \PYG{n}{out5} \PYG{o}{=} \PYG{n+nb+bp}{self}\PYG{o}{.}\PYG{n}{layer5}\PYG{p}{(}\PYG{n}{out4}\PYG{p}{)}
        \PYG{n}{out6} \PYG{o}{=} \PYG{n+nb+bp}{self}\PYG{o}{.}\PYG{n}{layer6}\PYG{p}{(}\PYG{n}{out5}\PYG{p}{)}
        \PYG{n}{out7} \PYG{o}{=} \PYG{n+nb+bp}{self}\PYG{o}{.}\PYG{n}{layer7}\PYG{p}{(}\PYG{n}{out6}\PYG{p}{)}
        \PYG{c+c1}{\PYGZsh{} ... bottleneck ...}
        \PYG{n}{dout6} \PYG{o}{=} \PYG{n+nb+bp}{self}\PYG{o}{.}\PYG{n}{dlayer7}\PYG{p}{(}\PYG{n}{out7}\PYG{p}{)}
        \PYG{n}{dout6\PYGZus{}out6} \PYG{o}{=} \PYG{n}{torch}\PYG{o}{.}\PYG{n}{cat}\PYG{p}{(}\PYG{p}{[}\PYG{n}{dout6}\PYG{p}{,} \PYG{n}{out6}\PYG{p}{]}\PYG{p}{,} \PYG{l+m+mi}{1}\PYG{p}{)}
        \PYG{n}{dout6} \PYG{o}{=} \PYG{n+nb+bp}{self}\PYG{o}{.}\PYG{n}{dlayer6}\PYG{p}{(}\PYG{n}{dout6\PYGZus{}out6}\PYG{p}{)}
        \PYG{n}{dout6\PYGZus{}out5} \PYG{o}{=} \PYG{n}{torch}\PYG{o}{.}\PYG{n}{cat}\PYG{p}{(}\PYG{p}{[}\PYG{n}{dout6}\PYG{p}{,} \PYG{n}{out5}\PYG{p}{]}\PYG{p}{,} \PYG{l+m+mi}{1}\PYG{p}{)}
        \PYG{n}{dout5} \PYG{o}{=} \PYG{n+nb+bp}{self}\PYG{o}{.}\PYG{n}{dlayer5}\PYG{p}{(}\PYG{n}{dout6\PYGZus{}out5}\PYG{p}{)}
        \PYG{n}{dout5\PYGZus{}out4} \PYG{o}{=} \PYG{n}{torch}\PYG{o}{.}\PYG{n}{cat}\PYG{p}{(}\PYG{p}{[}\PYG{n}{dout5}\PYG{p}{,} \PYG{n}{out4}\PYG{p}{]}\PYG{p}{,} \PYG{l+m+mi}{1}\PYG{p}{)}
        \PYG{n}{dout4} \PYG{o}{=} \PYG{n+nb+bp}{self}\PYG{o}{.}\PYG{n}{dlayer4}\PYG{p}{(}\PYG{n}{dout5\PYGZus{}out4}\PYG{p}{)}
        \PYG{n}{dout4\PYGZus{}out3} \PYG{o}{=} \PYG{n}{torch}\PYG{o}{.}\PYG{n}{cat}\PYG{p}{(}\PYG{p}{[}\PYG{n}{dout4}\PYG{p}{,} \PYG{n}{out3}\PYG{p}{]}\PYG{p}{,} \PYG{l+m+mi}{1}\PYG{p}{)}
        \PYG{n}{dout3} \PYG{o}{=} \PYG{n+nb+bp}{self}\PYG{o}{.}\PYG{n}{dlayer3}\PYG{p}{(}\PYG{n}{dout4\PYGZus{}out3}\PYG{p}{)}
        \PYG{n}{dout3\PYGZus{}out2} \PYG{o}{=} \PYG{n}{torch}\PYG{o}{.}\PYG{n}{cat}\PYG{p}{(}\PYG{p}{[}\PYG{n}{dout3}\PYG{p}{,} \PYG{n}{out2}\PYG{p}{]}\PYG{p}{,} \PYG{l+m+mi}{1}\PYG{p}{)}
        \PYG{n}{dout2} \PYG{o}{=} \PYG{n+nb+bp}{self}\PYG{o}{.}\PYG{n}{dlayer2}\PYG{p}{(}\PYG{n}{dout3\PYGZus{}out2}\PYG{p}{)}
        \PYG{n}{dout2\PYGZus{}out1} \PYG{o}{=} \PYG{n}{torch}\PYG{o}{.}\PYG{n}{cat}\PYG{p}{(}\PYG{p}{[}\PYG{n}{dout2}\PYG{p}{,} \PYG{n}{out1}\PYG{p}{]}\PYG{p}{,} \PYG{l+m+mi}{1}\PYG{p}{)}
        \PYG{n}{dout1} \PYG{o}{=} \PYG{n+nb+bp}{self}\PYG{o}{.}\PYG{n}{dlayer1}\PYG{p}{(}\PYG{n}{dout2\PYGZus{}out1}\PYG{p}{)}
        \PYG{k}{return} \PYG{n}{dout1}

\PYG{k}{def} \PYG{n+nf}{weights\PYGZus{}init}\PYG{p}{(}\PYG{n}{m}\PYG{p}{)}\PYG{p}{:}
    \PYG{n}{classname} \PYG{o}{=} \PYG{n}{m}\PYG{o}{.}\PYG{n+nv+vm}{\PYGZus{}\PYGZus{}class\PYGZus{}\PYGZus{}}\PYG{o}{.}\PYG{n+nv+vm}{\PYGZus{}\PYGZus{}name\PYGZus{}\PYGZus{}}
    \PYG{k}{if} \PYG{n}{classname}\PYG{o}{.}\PYG{n}{find}\PYG{p}{(}\PYG{l+s+s1}{\PYGZsq{}}\PYG{l+s+s1}{Conv}\PYG{l+s+s1}{\PYGZsq{}}\PYG{p}{)} \PYG{o}{!=} \PYG{o}{\PYGZhy{}}\PYG{l+m+mi}{1}\PYG{p}{:}
        \PYG{n}{m}\PYG{o}{.}\PYG{n}{weight}\PYG{o}{.}\PYG{n}{data}\PYG{o}{.}\PYG{n}{normal\PYGZus{}}\PYG{p}{(}\PYG{l+m+mf}{0.0}\PYG{p}{,} \PYG{l+m+mf}{0.02}\PYG{p}{)}
    \PYG{k}{elif} \PYG{n}{classname}\PYG{o}{.}\PYG{n}{find}\PYG{p}{(}\PYG{l+s+s1}{\PYGZsq{}}\PYG{l+s+s1}{BatchNorm}\PYG{l+s+s1}{\PYGZsq{}}\PYG{p}{)} \PYG{o}{!=} \PYG{o}{\PYGZhy{}}\PYG{l+m+mi}{1}\PYG{p}{:}
        \PYG{n}{m}\PYG{o}{.}\PYG{n}{weight}\PYG{o}{.}\PYG{n}{data}\PYG{o}{.}\PYG{n}{normal\PYGZus{}}\PYG{p}{(}\PYG{l+m+mf}{1.0}\PYG{p}{,} \PYG{l+m+mf}{0.02}\PYG{p}{)}
        \PYG{n}{m}\PYG{o}{.}\PYG{n}{bias}\PYG{o}{.}\PYG{n}{data}\PYG{o}{.}\PYG{n}{fill\PYGZus{}}\PYG{p}{(}\PYG{l+m+mi}{0}\PYG{p}{)}
\end{sphinxVerbatim}

Next, we can initialize an instance of the \sphinxcode{\sphinxupquote{DfpNet}}.

Below, the \sphinxcode{\sphinxupquote{EXPO}} parameter here controls the exponent for the feature maps of our Unet: this directly scales the network size (3 gives a network with ca. 150k parameters). This is relatively small for a generative NN with \(3 \times 128^2 = \text{ca. }49k\) outputs, but yields fast training times and prevents overfitting given the relatively small data set we’re using here. Hence it’s a good starting point.

\begin{sphinxVerbatim}[commandchars=\\\{\}]
\PYG{c+c1}{\PYGZsh{} channel exponent to control network size}
\PYG{n}{EXPO} \PYG{o}{=} \PYG{l+m+mi}{3}

\PYG{c+c1}{\PYGZsh{} setup network}
\PYG{n}{net} \PYG{o}{=} \PYG{n}{DfpNet}\PYG{p}{(}\PYG{n}{channelExponent}\PYG{o}{=}\PYG{n}{EXPO}\PYG{p}{)}
\PYG{c+c1}{\PYGZsh{}print(net) \PYGZsh{} to double check the details...}

\PYG{n}{nn\PYGZus{}parameters} \PYG{o}{=} \PYG{n+nb}{filter}\PYG{p}{(}\PYG{k}{lambda} \PYG{n}{p}\PYG{p}{:} \PYG{n}{p}\PYG{o}{.}\PYG{n}{requires\PYGZus{}grad}\PYG{p}{,} \PYG{n}{net}\PYG{o}{.}\PYG{n}{parameters}\PYG{p}{(}\PYG{p}{)}\PYG{p}{)}
\PYG{n}{params} \PYG{o}{=} \PYG{n+nb}{sum}\PYG{p}{(}\PYG{p}{[}\PYG{n}{np}\PYG{o}{.}\PYG{n}{prod}\PYG{p}{(}\PYG{n}{p}\PYG{o}{.}\PYG{n}{size}\PYG{p}{(}\PYG{p}{)}\PYG{p}{)} \PYG{k}{for} \PYG{n}{p} \PYG{o+ow}{in} \PYG{n}{nn\PYGZus{}parameters}\PYG{p}{]}\PYG{p}{)}

\PYG{c+c1}{\PYGZsh{} crucial parameter to keep in view: how many parameters do we have?}
\PYG{n+nb}{print}\PYG{p}{(}\PYG{l+s+s2}{\PYGZdq{}}\PYG{l+s+s2}{Trainable params: }\PYG{l+s+si}{\PYGZob{}\PYGZcb{}}\PYG{l+s+s2}{   \PYGZhy{}\PYGZgt{} crucial! always keep in view... }\PYG{l+s+s2}{\PYGZdq{}}\PYG{o}{.}\PYG{n}{format}\PYG{p}{(}\PYG{n}{params}\PYG{p}{)}\PYG{p}{)} 

\PYG{n}{net}\PYG{o}{.}\PYG{n}{apply}\PYG{p}{(}\PYG{n}{weights\PYGZus{}init}\PYG{p}{)}

\PYG{n}{criterionL1} \PYG{o}{=} \PYG{n}{nn}\PYG{o}{.}\PYG{n}{L1Loss}\PYG{p}{(}\PYG{p}{)}
\PYG{n}{optimizerG} \PYG{o}{=} \PYG{n}{optim}\PYG{o}{.}\PYG{n}{Adam}\PYG{p}{(}\PYG{n}{net}\PYG{o}{.}\PYG{n}{parameters}\PYG{p}{(}\PYG{p}{)}\PYG{p}{,} \PYG{n}{lr}\PYG{o}{=}\PYG{n}{LR}\PYG{p}{,} \PYG{n}{betas}\PYG{o}{=}\PYG{p}{(}\PYG{l+m+mf}{0.5}\PYG{p}{,} \PYG{l+m+mf}{0.999}\PYG{p}{)}\PYG{p}{,} \PYG{n}{weight\PYGZus{}decay}\PYG{o}{=}\PYG{l+m+mf}{0.0}\PYG{p}{)}

\PYG{n}{targets} \PYG{o}{=} \PYG{n}{torch}\PYG{o}{.}\PYG{n}{autograd}\PYG{o}{.}\PYG{n}{Variable}\PYG{p}{(}\PYG{n}{torch}\PYG{o}{.}\PYG{n}{FloatTensor}\PYG{p}{(}\PYG{n}{BATCH\PYGZus{}SIZE}\PYG{p}{,} \PYG{l+m+mi}{3}\PYG{p}{,} \PYG{l+m+mi}{128}\PYG{p}{,} \PYG{l+m+mi}{128}\PYG{p}{)}\PYG{p}{)}
\PYG{n}{inputs}  \PYG{o}{=} \PYG{n}{torch}\PYG{o}{.}\PYG{n}{autograd}\PYG{o}{.}\PYG{n}{Variable}\PYG{p}{(}\PYG{n}{torch}\PYG{o}{.}\PYG{n}{FloatTensor}\PYG{p}{(}\PYG{n}{BATCH\PYGZus{}SIZE}\PYG{p}{,} \PYG{l+m+mi}{3}\PYG{p}{,} \PYG{l+m+mi}{128}\PYG{p}{,} \PYG{l+m+mi}{128}\PYG{p}{)}\PYG{p}{)}
\end{sphinxVerbatim}

\begin{sphinxVerbatim}[commandchars=\\\{\}]
Trainable params: 147555   \PYGZhy{}\PYGZgt{} crucial! always keep in view... 
\end{sphinxVerbatim}

With an exponent of 3, this network has 147555 trainable parameters. As the subtle hint in the print statement indicates, this is a crucial number to always have in view when training NNs. It’s easy to change settings, and get a network that has millions of parameters, and as a result probably all kinds of convergence and overfitting problems. The number of parameters definitely has to be matched with the amount of training data, and should also scale with the depth of the network. How these three relate to each other exactly is problem dependent, though.

\subsection{Training}
\label{\detokenize{supervised-airfoils:training}}
Finally, we can train the NN. This step can take a while, as the training runs over all 320 samples 100 times, and continually evaluates the validation samples to keep track of how well the current state of the NN is doing.

\begin{sphinxVerbatim}[commandchars=\\\{\}]
\PYG{n}{history\PYGZus{}L1} \PYG{o}{=} \PYG{p}{[}\PYG{p}{]}
\PYG{n}{history\PYGZus{}L1val} \PYG{o}{=} \PYG{p}{[}\PYG{p}{]}

\PYG{k}{if} \PYG{n}{os}\PYG{o}{.}\PYG{n}{path}\PYG{o}{.}\PYG{n}{isfile}\PYG{p}{(}\PYG{l+s+s2}{\PYGZdq{}}\PYG{l+s+s2}{network}\PYG{l+s+s2}{\PYGZdq{}}\PYG{p}{)}\PYG{p}{:}
  \PYG{n+nb}{print}\PYG{p}{(}\PYG{l+s+s2}{\PYGZdq{}}\PYG{l+s+s2}{Found existing network, loading \PYGZam{} skipping training}\PYG{l+s+s2}{\PYGZdq{}}\PYG{p}{)}
  \PYG{n}{net}\PYG{o}{.}\PYG{n}{load\PYGZus{}state\PYGZus{}dict}\PYG{p}{(}\PYG{n}{torch}\PYG{o}{.}\PYG{n}{load}\PYG{p}{(}\PYG{l+s+s2}{\PYGZdq{}}\PYG{l+s+s2}{network}\PYG{l+s+s2}{\PYGZdq{}}\PYG{p}{)}\PYG{p}{)} \PYG{c+c1}{\PYGZsh{} optionally, load existing network}

\PYG{k}{else}\PYG{p}{:}
  \PYG{n+nb}{print}\PYG{p}{(}\PYG{l+s+s2}{\PYGZdq{}}\PYG{l+s+s2}{Training from scratch}\PYG{l+s+s2}{\PYGZdq{}}\PYG{p}{)}
  \PYG{k}{for} \PYG{n}{epoch} \PYG{o+ow}{in} \PYG{n+nb}{range}\PYG{p}{(}\PYG{n}{EPOCHS}\PYG{p}{)}\PYG{p}{:}
      \PYG{n}{net}\PYG{o}{.}\PYG{n}{train}\PYG{p}{(}\PYG{p}{)}
      \PYG{n}{L1\PYGZus{}accum} \PYG{o}{=} \PYG{l+m+mf}{0.0}
      \PYG{k}{for} \PYG{n}{i}\PYG{p}{,} \PYG{n}{traindata} \PYG{o+ow}{in} \PYG{n+nb}{enumerate}\PYG{p}{(}\PYG{n}{trainLoader}\PYG{p}{,} \PYG{l+m+mi}{0}\PYG{p}{)}\PYG{p}{:}
          \PYG{n}{inputs\PYGZus{}curr}\PYG{p}{,} \PYG{n}{targets\PYGZus{}curr} \PYG{o}{=} \PYG{n}{traindata}
          \PYG{n}{inputs}\PYG{o}{.}\PYG{n}{data}\PYG{o}{.}\PYG{n}{copy\PYGZus{}}\PYG{p}{(}\PYG{n}{inputs\PYGZus{}curr}\PYG{o}{.}\PYG{n}{float}\PYG{p}{(}\PYG{p}{)}\PYG{p}{)}
          \PYG{n}{targets}\PYG{o}{.}\PYG{n}{data}\PYG{o}{.}\PYG{n}{copy\PYGZus{}}\PYG{p}{(}\PYG{n}{targets\PYGZus{}curr}\PYG{o}{.}\PYG{n}{float}\PYG{p}{(}\PYG{p}{)}\PYG{p}{)}

          \PYG{n}{net}\PYG{o}{.}\PYG{n}{zero\PYGZus{}grad}\PYG{p}{(}\PYG{p}{)}
          \PYG{n}{gen\PYGZus{}out} \PYG{o}{=} \PYG{n}{net}\PYG{p}{(}\PYG{n}{inputs}\PYG{p}{)}

          \PYG{n}{lossL1} \PYG{o}{=} \PYG{n}{criterionL1}\PYG{p}{(}\PYG{n}{gen\PYGZus{}out}\PYG{p}{,} \PYG{n}{targets}\PYG{p}{)}
          \PYG{n}{lossL1}\PYG{o}{.}\PYG{n}{backward}\PYG{p}{(}\PYG{p}{)}
          \PYG{n}{optimizerG}\PYG{o}{.}\PYG{n}{step}\PYG{p}{(}\PYG{p}{)}
          \PYG{n}{L1\PYGZus{}accum} \PYG{o}{+}\PYG{o}{=} \PYG{n}{lossL1}\PYG{o}{.}\PYG{n}{item}\PYG{p}{(}\PYG{p}{)}

      \PYG{c+c1}{\PYGZsh{} validation}
      \PYG{n}{net}\PYG{o}{.}\PYG{n}{eval}\PYG{p}{(}\PYG{p}{)}
      \PYG{n}{L1val\PYGZus{}accum} \PYG{o}{=} \PYG{l+m+mf}{0.0}
      \PYG{k}{for} \PYG{n}{i}\PYG{p}{,} \PYG{n}{validata} \PYG{o+ow}{in} \PYG{n+nb}{enumerate}\PYG{p}{(}\PYG{n}{valiLoader}\PYG{p}{,} \PYG{l+m+mi}{0}\PYG{p}{)}\PYG{p}{:}
          \PYG{n}{inputs\PYGZus{}curr}\PYG{p}{,} \PYG{n}{targets\PYGZus{}curr} \PYG{o}{=} \PYG{n}{validata}
          \PYG{n}{inputs}\PYG{o}{.}\PYG{n}{data}\PYG{o}{.}\PYG{n}{copy\PYGZus{}}\PYG{p}{(}\PYG{n}{inputs\PYGZus{}curr}\PYG{o}{.}\PYG{n}{float}\PYG{p}{(}\PYG{p}{)}\PYG{p}{)}
          \PYG{n}{targets}\PYG{o}{.}\PYG{n}{data}\PYG{o}{.}\PYG{n}{copy\PYGZus{}}\PYG{p}{(}\PYG{n}{targets\PYGZus{}curr}\PYG{o}{.}\PYG{n}{float}\PYG{p}{(}\PYG{p}{)}\PYG{p}{)}

          \PYG{n}{outputs} \PYG{o}{=} \PYG{n}{net}\PYG{p}{(}\PYG{n}{inputs}\PYG{p}{)}
          \PYG{n}{outputs\PYGZus{}curr} \PYG{o}{=} \PYG{n}{outputs}\PYG{o}{.}\PYG{n}{data}\PYG{o}{.}\PYG{n}{cpu}\PYG{p}{(}\PYG{p}{)}\PYG{o}{.}\PYG{n}{numpy}\PYG{p}{(}\PYG{p}{)}

          \PYG{n}{lossL1val} \PYG{o}{=} \PYG{n}{criterionL1}\PYG{p}{(}\PYG{n}{outputs}\PYG{p}{,} \PYG{n}{targets}\PYG{p}{)}
          \PYG{n}{L1val\PYGZus{}accum} \PYG{o}{+}\PYG{o}{=} \PYG{n}{lossL1val}\PYG{o}{.}\PYG{n}{item}\PYG{p}{(}\PYG{p}{)}

      \PYG{c+c1}{\PYGZsh{} data for graph plotting}
      \PYG{n}{history\PYGZus{}L1}\PYG{o}{.}\PYG{n}{append}\PYG{p}{(} \PYG{n}{L1\PYGZus{}accum} \PYG{o}{/} \PYG{n+nb}{len}\PYG{p}{(}\PYG{n}{trainLoader}\PYG{p}{)} \PYG{p}{)}
      \PYG{n}{history\PYGZus{}L1val}\PYG{o}{.}\PYG{n}{append}\PYG{p}{(} \PYG{n}{L1val\PYGZus{}accum} \PYG{o}{/} \PYG{n+nb}{len}\PYG{p}{(}\PYG{n}{valiLoader}\PYG{p}{)} \PYG{p}{)}

      \PYG{k}{if} \PYG{n}{epoch}\PYG{o}{\PYGZlt{}}\PYG{l+m+mi}{3} \PYG{o+ow}{or} \PYG{n}{epoch}\PYG{o}{\PYGZpc{}}\PYG{k}{20}==0:
          \PYG{n+nb}{print}\PYG{p}{(} \PYG{l+s+s2}{\PYGZdq{}}\PYG{l+s+s2}{Epoch: }\PYG{l+s+si}{\PYGZob{}\PYGZcb{}}\PYG{l+s+s2}{, L1 train: }\PYG{l+s+si}{\PYGZob{}:7.5f\PYGZcb{}}\PYG{l+s+s2}{, L1 vali: }\PYG{l+s+si}{\PYGZob{}:7.5f\PYGZcb{}}\PYG{l+s+s2}{\PYGZdq{}}\PYG{o}{.}\PYG{n}{format}\PYG{p}{(}\PYG{n}{epoch}\PYG{p}{,} \PYG{n}{history\PYGZus{}L1}\PYG{p}{[}\PYG{o}{\PYGZhy{}}\PYG{l+m+mi}{1}\PYG{p}{]}\PYG{p}{,} \PYG{n}{history\PYGZus{}L1val}\PYG{p}{[}\PYG{o}{\PYGZhy{}}\PYG{l+m+mi}{1}\PYG{p}{]}\PYG{p}{)} \PYG{p}{)}

  \PYG{n}{torch}\PYG{o}{.}\PYG{n}{save}\PYG{p}{(}\PYG{n}{net}\PYG{o}{.}\PYG{n}{state\PYGZus{}dict}\PYG{p}{(}\PYG{p}{)}\PYG{p}{,} \PYG{l+s+s2}{\PYGZdq{}}\PYG{l+s+s2}{network}\PYG{l+s+s2}{\PYGZdq{}} \PYG{p}{)}
  \PYG{n+nb}{print}\PYG{p}{(}\PYG{l+s+s2}{\PYGZdq{}}\PYG{l+s+s2}{Training done, saved network}\PYG{l+s+s2}{\PYGZdq{}}\PYG{p}{)}
\end{sphinxVerbatim}

\begin{sphinxVerbatim}[commandchars=\\\{\}]
Training from scratch
\end{sphinxVerbatim}

\begin{sphinxVerbatim}[commandchars=\\\{\}]
  \PYGZdq{}See the documentation of nn.Upsample for details.\PYGZdq{}.format(mode)
\end{sphinxVerbatim}

\begin{sphinxVerbatim}[commandchars=\\\{\}]
Epoch: 0, L1 train: 0.20192, L1 vali: 0.17916
Epoch: 1, L1 train: 0.18912, L1 vali: 0.18013
Epoch: 2, L1 train: 0.17917, L1 vali: 0.17145
Epoch: 20, L1 train: 0.08637, L1 vali: 0.08432
Epoch: 40, L1 train: 0.03682, L1 vali: 0.03222
Epoch: 60, L1 train: 0.03176, L1 vali: 0.02710
Epoch: 80, L1 train: 0.02772, L1 vali: 0.02522
Training done, saved trained network
\end{sphinxVerbatim}

The NN is finally trained! The losses should have nicely gone down in terms of absolute values: With the standard settings from an initial value of around 0.2 for the validation  loss, to ca. 0.02 after 100 epochs.

Let’s look at the graphs to get some intuition for how the training progressed over time. This is typically important to identify longer\sphinxhyphen{}term trends in the training. In practice it’s tricky to spot whether the overall trend of 100 or so noisy numbers in a command line log is going slightly up or down \sphinxhyphen{} this is much easier to spot in a visualization.

\begin{sphinxVerbatim}[commandchars=\\\{\}]
\PYG{k+kn}{import} \PYG{n+nn}{matplotlib}\PYG{n+nn}{.}\PYG{n+nn}{pyplot} \PYG{k}{as} \PYG{n+nn}{plt}

\PYG{n}{l1train} \PYG{o}{=} \PYG{n}{np}\PYG{o}{.}\PYG{n}{asarray}\PYG{p}{(}\PYG{n}{history\PYGZus{}L1}\PYG{p}{)}
\PYG{n}{l1vali}  \PYG{o}{=} \PYG{n}{np}\PYG{o}{.}\PYG{n}{asarray}\PYG{p}{(}\PYG{n}{history\PYGZus{}L1val}\PYG{p}{)}

\PYG{n}{plt}\PYG{o}{.}\PYG{n}{plot}\PYG{p}{(}\PYG{n}{np}\PYG{o}{.}\PYG{n}{arange}\PYG{p}{(}\PYG{n}{l1train}\PYG{o}{.}\PYG{n}{shape}\PYG{p}{[}\PYG{l+m+mi}{0}\PYG{p}{]}\PYG{p}{)}\PYG{p}{,}\PYG{n}{l1train}\PYG{p}{,}\PYG{l+s+s1}{\PYGZsq{}}\PYG{l+s+s1}{b}\PYG{l+s+s1}{\PYGZsq{}}\PYG{p}{,}\PYG{n}{label}\PYG{o}{=}\PYG{l+s+s1}{\PYGZsq{}}\PYG{l+s+s1}{Training loss}\PYG{l+s+s1}{\PYGZsq{}}\PYG{p}{)}
\PYG{n}{plt}\PYG{o}{.}\PYG{n}{plot}\PYG{p}{(}\PYG{n}{np}\PYG{o}{.}\PYG{n}{arange}\PYG{p}{(}\PYG{n}{l1vali}\PYG{o}{.}\PYG{n}{shape}\PYG{p}{[}\PYG{l+m+mi}{0}\PYG{p}{]} \PYG{p}{)}\PYG{p}{,}\PYG{n}{l1vali} \PYG{p}{,}\PYG{l+s+s1}{\PYGZsq{}}\PYG{l+s+s1}{g}\PYG{l+s+s1}{\PYGZsq{}}\PYG{p}{,}\PYG{n}{label}\PYG{o}{=}\PYG{l+s+s1}{\PYGZsq{}}\PYG{l+s+s1}{Validation loss}\PYG{l+s+s1}{\PYGZsq{}}\PYG{p}{)}
\PYG{n}{plt}\PYG{o}{.}\PYG{n}{legend}\PYG{p}{(}\PYG{p}{)}
\PYG{n}{plt}\PYG{o}{.}\PYG{n}{show}\PYG{p}{(}\PYG{p}{)}
\end{sphinxVerbatim}

\noindent\sphinxincludegraphics{{supervised-airfoils_17_0}.png}

You should see a curve that goes down for ca. 40 epochs, and then starts to flatten out. In the last part, it’s still slowly decreasing, and most importantly, the validation loss is not increasing. This would be a certain sign of overfitting, and something that we should avoid. (Try decreasing the amount of training data artificially, then you should be able to intentionally cause overfitting.)

\subsection{Training progress and validation}
\label{\detokenize{supervised-airfoils:training-progress-and-validation}}
If you look closely at this graph, you should spot something peculiar:
\sphinxstyleemphasis{Why is the validation loss lower than the training loss}?
The data is similar to the training data of course, but in a way it’s slightly “tougher”, because the network certainly never received any validation samples during training. It is natural that the validation loss slightly deviates from the training loss, but how can the L1 loss be \sphinxstyleemphasis{lower} for these inputs?

This is a subtlety of the training loop above: it runs a training step first, and the loss for each point in the graph is measured with the evolving state of the network in an epoch. The network is updated, and afterwards runs through the validation samples. Thus all validation samples are using a state that is definitely different (and hopefully a bit better) than the initial states of the epoch. Hence, the validation loss can be slightly lower.

A general word of caution here: never evaluate your network with training data! That won’t tell you much because overfitting is a very common problem. At least use data the network hasn’t seen before, i.e. validation data, and if that looks good, try some more different (at least slightly out\sphinxhyphen{}of\sphinxhyphen{}distribution) inputs, i.e., \sphinxstyleemphasis{test data}. The next cell runs the trained network over the validation data, and displays one of them with the \sphinxcode{\sphinxupquote{showSbs}} function.

\begin{sphinxVerbatim}[commandchars=\\\{\}]
\PYG{n}{net}\PYG{o}{.}\PYG{n}{eval}\PYG{p}{(}\PYG{p}{)}
\PYG{k}{for} \PYG{n}{i}\PYG{p}{,} \PYG{n}{validata} \PYG{o+ow}{in} \PYG{n+nb}{enumerate}\PYG{p}{(}\PYG{n}{valiLoader}\PYG{p}{,} \PYG{l+m+mi}{0}\PYG{p}{)}\PYG{p}{:}
    \PYG{n}{inputs\PYGZus{}curr}\PYG{p}{,} \PYG{n}{targets\PYGZus{}curr} \PYG{o}{=} \PYG{n}{validata}
    \PYG{n}{inputs}\PYG{o}{.}\PYG{n}{data}\PYG{o}{.}\PYG{n}{copy\PYGZus{}}\PYG{p}{(}\PYG{n}{inputs\PYGZus{}curr}\PYG{o}{.}\PYG{n}{float}\PYG{p}{(}\PYG{p}{)}\PYG{p}{)}
    \PYG{n}{targets}\PYG{o}{.}\PYG{n}{data}\PYG{o}{.}\PYG{n}{copy\PYGZus{}}\PYG{p}{(}\PYG{n}{targets\PYGZus{}curr}\PYG{o}{.}\PYG{n}{float}\PYG{p}{(}\PYG{p}{)}\PYG{p}{)}
    
    \PYG{n}{outputs} \PYG{o}{=} \PYG{n}{net}\PYG{p}{(}\PYG{n}{inputs}\PYG{p}{)}
    \PYG{n}{outputs\PYGZus{}curr} \PYG{o}{=} \PYG{n}{outputs}\PYG{o}{.}\PYG{n}{data}\PYG{o}{.}\PYG{n}{cpu}\PYG{p}{(}\PYG{p}{)}\PYG{o}{.}\PYG{n}{numpy}\PYG{p}{(}\PYG{p}{)}
    \PYG{k}{if} \PYG{n}{i}\PYG{o}{\PYGZlt{}}\PYG{l+m+mi}{1}\PYG{p}{:} \PYG{n}{showSbs}\PYG{p}{(}\PYG{n}{targets\PYGZus{}curr}\PYG{p}{[}\PYG{l+m+mi}{0}\PYG{p}{]} \PYG{p}{,} \PYG{n}{outputs\PYGZus{}curr}\PYG{p}{[}\PYG{l+m+mi}{0}\PYG{p}{]}\PYG{p}{,} \PYG{n}{title}\PYG{o}{=}\PYG{l+s+s2}{\PYGZdq{}}\PYG{l+s+s2}{Validation sample }\PYG{l+s+si}{\PYGZpc{}d}\PYG{l+s+s2}{\PYGZdq{}}\PYG{o}{\PYGZpc{}}\PYG{p}{(}\PYG{n}{i}\PYG{o}{*}\PYG{n}{BATCH\PYGZus{}SIZE}\PYG{p}{)}\PYG{p}{)}
\end{sphinxVerbatim}

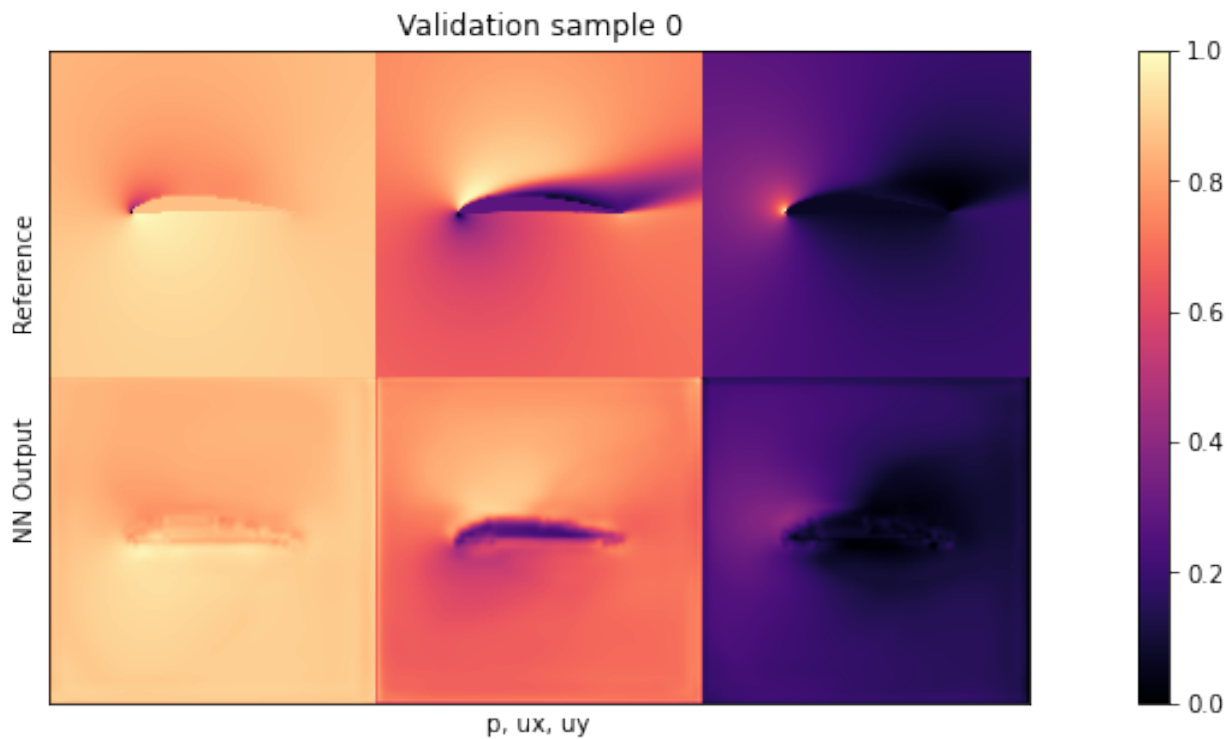
\begin{sphinxVerbatim}[commandchars=\\\{\}]
  \PYGZdq{}See the documentation of nn.Upsample for details.\PYGZdq{}.format(mode)
\end{sphinxVerbatim}

\noindent\sphinxincludegraphics{{supervised-airfoils_19_1}.png}

Visually, there should at least be a rough resemblance here between input out network output. We’ll save the more detailed evaluation for the test data, though.

\subsection{Test evaluation}
\label{\detokenize{supervised-airfoils:test-evaluation}}
Now let’s look at actual test samples: In this case we’ll use new airfoil shapes as out\sphinxhyphen{}of\sphinxhyphen{}distribution (OOD) data. These are shapes that the network never saw in any training samples, and hence it tells us a bit about how well the NN generalizes to unseen inputs (the validation data wouldn’t suffice to draw conclusions about generalization).

We’ll use the same visualization as before, and as indicated by the Bernoulli equation, especially the \sphinxstyleemphasis{pressure} in the first column is a challenging quantity for the network. Due to it’s cubic scaling w.r.t. the input freestream velocity and localized peaks, it is the toughest quantity to infer for the network.

The cell below first downloads a smaller archive with these test data samples, and then runs them through the network. The evaluation loop also computes the accumulated L1 error such that we can quantify how well the network does on the test samples.

\begin{sphinxVerbatim}[commandchars=\\\{\}]
\PYG{k}{if} \PYG{o+ow}{not} \PYG{n}{os}\PYG{o}{.}\PYG{n}{path}\PYG{o}{.}\PYG{n}{isfile}\PYG{p}{(}\PYG{l+s+s1}{\PYGZsq{}}\PYG{l+s+s1}{data\PYGZhy{}airfoils\PYGZhy{}test.npz}\PYG{l+s+s1}{\PYGZsq{}}\PYG{p}{)}\PYG{p}{:}
  \PYG{k+kn}{import} \PYG{n+nn}{urllib}\PYG{n+nn}{.}\PYG{n+nn}{request}
  \PYG{n}{url}\PYG{o}{=}\PYG{l+s+s2}{\PYGZdq{}}\PYG{l+s+s2}{https://physicsbaseddeeplearning.org/data/data\PYGZus{}test.npz}\PYG{l+s+s2}{\PYGZdq{}}
  \PYG{n+nb}{print}\PYG{p}{(}\PYG{l+s+s2}{\PYGZdq{}}\PYG{l+s+s2}{Downloading test data, this should be fast...}\PYG{l+s+s2}{\PYGZdq{}}\PYG{p}{)}
  \PYG{n}{urllib}\PYG{o}{.}\PYG{n}{request}\PYG{o}{.}\PYG{n}{urlretrieve}\PYG{p}{(}\PYG{n}{url}\PYG{p}{,} \PYG{l+s+s1}{\PYGZsq{}}\PYG{l+s+s1}{data\PYGZhy{}airfoils\PYGZhy{}test.npz}\PYG{l+s+s1}{\PYGZsq{}}\PYG{p}{)}

\PYG{n}{nptfile}\PYG{o}{=}\PYG{n}{np}\PYG{o}{.}\PYG{n}{load}\PYG{p}{(}\PYG{l+s+s1}{\PYGZsq{}}\PYG{l+s+s1}{data\PYGZhy{}airfoils\PYGZhy{}test.npz}\PYG{l+s+s1}{\PYGZsq{}}\PYG{p}{)}
\PYG{n+nb}{print}\PYG{p}{(}\PYG{l+s+s2}{\PYGZdq{}}\PYG{l+s+s2}{Loaded }\PYG{l+s+si}{\PYGZob{}\PYGZcb{}}\PYG{l+s+s2}{/}\PYG{l+s+si}{\PYGZob{}\PYGZcb{}}\PYG{l+s+s2}{ test samples}\PYG{l+s+se}{\PYGZbs{}n}\PYG{l+s+s2}{\PYGZdq{}}\PYG{o}{.}\PYG{n}{format}\PYG{p}{(}\PYG{n+nb}{len}\PYG{p}{(}\PYG{n}{nptfile}\PYG{p}{[}\PYG{l+s+s2}{\PYGZdq{}}\PYG{l+s+s2}{test\PYGZus{}inputs}\PYG{l+s+s2}{\PYGZdq{}}\PYG{p}{]}\PYG{p}{)}\PYG{p}{,}\PYG{n+nb}{len}\PYG{p}{(}\PYG{n}{nptfile}\PYG{p}{[}\PYG{l+s+s2}{\PYGZdq{}}\PYG{l+s+s2}{test\PYGZus{}targets}\PYG{l+s+s2}{\PYGZdq{}}\PYG{p}{]}\PYG{p}{)}\PYG{p}{)}\PYG{p}{)}

\PYG{n}{testdata} \PYG{o}{=} \PYG{n}{DfpDataset}\PYG{p}{(}\PYG{n}{nptfile}\PYG{p}{[}\PYG{l+s+s2}{\PYGZdq{}}\PYG{l+s+s2}{test\PYGZus{}inputs}\PYG{l+s+s2}{\PYGZdq{}}\PYG{p}{]}\PYG{p}{,}\PYG{n}{nptfile}\PYG{p}{[}\PYG{l+s+s2}{\PYGZdq{}}\PYG{l+s+s2}{test\PYGZus{}targets}\PYG{l+s+s2}{\PYGZdq{}}\PYG{p}{]}\PYG{p}{)}
\PYG{n}{testLoader}  \PYG{o}{=} \PYG{n}{torch}\PYG{o}{.}\PYG{n}{utils}\PYG{o}{.}\PYG{n}{data}\PYG{o}{.}\PYG{n}{DataLoader}\PYG{p}{(}\PYG{n}{testdata}\PYG{p}{,} \PYG{n}{batch\PYGZus{}size}\PYG{o}{=}\PYG{l+m+mi}{1}\PYG{p}{,} \PYG{n}{shuffle}\PYG{o}{=}\PYG{k+kc}{False}\PYG{p}{,} \PYG{n}{drop\PYGZus{}last}\PYG{o}{=}\PYG{k+kc}{True}\PYG{p}{)} 

\PYG{n}{net}\PYG{o}{.}\PYG{n}{eval}\PYG{p}{(}\PYG{p}{)}
\PYG{n}{L1t\PYGZus{}accum} \PYG{o}{=} \PYG{l+m+mf}{0.}
\PYG{k}{for} \PYG{n}{i}\PYG{p}{,} \PYG{n}{validata} \PYG{o+ow}{in} \PYG{n+nb}{enumerate}\PYG{p}{(}\PYG{n}{testLoader}\PYG{p}{,} \PYG{l+m+mi}{0}\PYG{p}{)}\PYG{p}{:}
    \PYG{n}{inputs\PYGZus{}curr}\PYG{p}{,} \PYG{n}{targets\PYGZus{}curr} \PYG{o}{=} \PYG{n}{validata}
    \PYG{n}{inputs}\PYG{o}{.}\PYG{n}{data}\PYG{o}{.}\PYG{n}{copy\PYGZus{}}\PYG{p}{(}\PYG{n}{inputs\PYGZus{}curr}\PYG{o}{.}\PYG{n}{float}\PYG{p}{(}\PYG{p}{)}\PYG{p}{)}
    \PYG{n}{targets}\PYG{o}{.}\PYG{n}{data}\PYG{o}{.}\PYG{n}{copy\PYGZus{}}\PYG{p}{(}\PYG{n}{targets\PYGZus{}curr}\PYG{o}{.}\PYG{n}{float}\PYG{p}{(}\PYG{p}{)}\PYG{p}{)}

    \PYG{n}{outputs} \PYG{o}{=} \PYG{n}{net}\PYG{p}{(}\PYG{n}{inputs}\PYG{p}{)}
    \PYG{n}{outputs\PYGZus{}curr} \PYG{o}{=} \PYG{n}{outputs}\PYG{o}{.}\PYG{n}{data}\PYG{o}{.}\PYG{n}{cpu}\PYG{p}{(}\PYG{p}{)}\PYG{o}{.}\PYG{n}{numpy}\PYG{p}{(}\PYG{p}{)}

    \PYG{n}{lossL1t} \PYG{o}{=} \PYG{n}{criterionL1}\PYG{p}{(}\PYG{n}{outputs}\PYG{p}{,} \PYG{n}{targets}\PYG{p}{)}
    \PYG{n}{L1t\PYGZus{}accum} \PYG{o}{+}\PYG{o}{=} \PYG{n}{lossL1t}\PYG{o}{.}\PYG{n}{item}\PYG{p}{(}\PYG{p}{)}
    \PYG{k}{if} \PYG{n}{i}\PYG{o}{\PYGZlt{}}\PYG{l+m+mi}{3}\PYG{p}{:} \PYG{n}{showSbs}\PYG{p}{(}\PYG{n}{targets\PYGZus{}curr}\PYG{p}{[}\PYG{l+m+mi}{0}\PYG{p}{]} \PYG{p}{,} \PYG{n}{outputs\PYGZus{}curr}\PYG{p}{[}\PYG{l+m+mi}{0}\PYG{p}{]}\PYG{p}{,}  \PYG{n}{title}\PYG{o}{=}\PYG{l+s+s2}{\PYGZdq{}}\PYG{l+s+s2}{Test sample }\PYG{l+s+si}{\PYGZpc{}d}\PYG{l+s+s2}{\PYGZdq{}}\PYG{o}{\PYGZpc{}}\PYG{p}{(}\PYG{n}{i}\PYG{p}{)}\PYG{p}{)}

\PYG{n+nb}{print}\PYG{p}{(}\PYG{l+s+s2}{\PYGZdq{}}\PYG{l+s+se}{\PYGZbs{}n}\PYG{l+s+s2}{Average test error: }\PYG{l+s+si}{\PYGZob{}\PYGZcb{}}\PYG{l+s+s2}{\PYGZdq{}}\PYG{o}{.}\PYG{n}{format}\PYG{p}{(} \PYG{n}{L1t\PYGZus{}accum}\PYG{o}{/}\PYG{n+nb}{len}\PYG{p}{(}\PYG{n}{testLoader}\PYG{p}{)} \PYG{p}{)}\PYG{p}{)}
\end{sphinxVerbatim}

\begin{sphinxVerbatim}[commandchars=\\\{\}]
Downloading test data, this should be fast...
Loaded 10/10 test samples
\end{sphinxVerbatim}

\begin{sphinxVerbatim}[commandchars=\\\{\}]
  \PYGZdq{}See the documentation of nn.Upsample for details.\PYGZdq{}.format(mode)
\end{sphinxVerbatim}

\begin{sphinxVerbatim}[commandchars=\\\{\}]
Average test error: 0.029000501427799464
\end{sphinxVerbatim}

\noindent\sphinxincludegraphics{{supervised-airfoils_22_3}.png}

\noindent\sphinxincludegraphics{{supervised-airfoils_22_4}.png}

\noindent\sphinxincludegraphics{{supervised-airfoils_22_5}.png}

The average test error with the default settings should be ca. 0.03. As the inputs are normalized, this means the average error across all three fields is 3\% w.r.t. the maxima of each quantity. This is not too bad for new shapes, but clearly leaves room for improvement.

Looking at the visualizations, you’ll notice that especially high\sphinxhyphen{}pressure peaks and pockets of larger y\sphinxhyphen{}velocities are missing in the outputs. This is primarily caused by the small network, which does not have enough resources to reconstruct details.

Nonetheless, we have successfully replaced a fairly sophisticated RANS solver with a very small and fast neural network architecture. It has GPU support “out\sphinxhyphen{}of\sphinxhyphen{}the\sphinxhyphen{}box” (via pytorch), is differentiable, and introduces an error of only a few per\sphinxhyphen{}cent.

\bigskip\hrule\bigskip

\subsection{Next steps}
\label{\detokenize{supervised-airfoils:next-steps}}
There are many obvious things to try here (see the suggestions below), e.g. longer training, larger data sets, larger networks etc.
\begin{itemize}
\item {} 
Experiment with learning rate, dropout, and network size to reduce the error on the test set. How small can you make it with the given training data?

\item {} 
The setup above uses normalized data. Instead you can recover \sphinxhref{https://github.com/thunil/Deep-Flow-Prediction}{the original fields by undoing the normalization} to check how well the network does w.r.t. the original quantities.

\item {} 
As you’ll see, it’s a bit limited here what you can get out of this dataset, head over to \sphinxhref{https://github.com/thunil/Deep-Flow-Prediction}{the main github repo of this project} to download larger data sets, or generate own data.

\end{itemize}

\section{Discussion of Supervised Approaches}
\label{\detokenize{supervised-discuss:discussion-of-supervised-approaches}}\label{\detokenize{supervised-discuss::doc}}
The previous example illustrates that we can quite easily use
supervised training to solve complex tasks. The main workload is
collecting a large enough dataset of examples. Once that exists, we can
train a network to approximate the solution manifold sampled
by these solutions, and the trained network can give us predictions
very quickly. There are a few important points to keep in mind when
using supervised training.

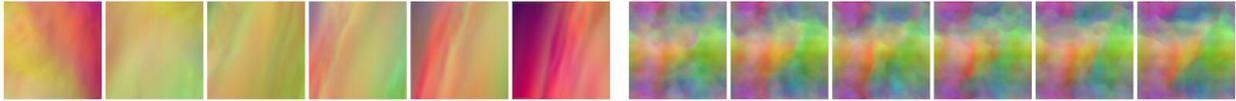
\sphinxincludegraphics{{divider1}.jpg}

\subsection{Some things to keep in mind…}
\label{\detokenize{supervised-discuss:some-things-to-keep-in-mind}}

\subsubsection{Natural starting point}
\label{\detokenize{supervised-discuss:natural-starting-point}}
\sphinxstyleemphasis{Supervised training} is the natural starting point for \sphinxstylestrong{any} DL project. It always,
and we really mean \sphinxstylestrong{always} here, makes sense to start with a fully supervised
test using as little data as possible. This will be a pure overfitting test,
but if your network can’t quickly converge and give a very good performance
on a single example, then there’s something fundamentally wrong
with your code or data. Thus, there’s no reason to move on to more complex
setups that will make finding these fundamental problems more difficult.

\begin{sphinxadmonition}{note}{Best practices 👑}

To summarize the scattered comments of the previous sections, here’s a set of “golden rules”  for setting up a DL project.
\begin{itemize}
\item {} 
Always start with a 1\sphinxhyphen{}sample overfitting test.

\item {} 
Check how many trainable parameters your network has.

\item {} 
Slowly increase the amount of training data (and potentially network parameters and depth).

\item {} 
Adjust hyperparameters (especially the learning rate).

\item {} 
Then introduce other components such as differentiable solvers or adversarial training.

\end{itemize}
\end{sphinxadmonition}

\subsubsection{Stability}
\label{\detokenize{supervised-discuss:stability}}
A nice property of the supervised training is also that it’s very stable.
Things won’t get any better when we include more complex physical
models, or look at more complicated NN architectures.

Thus, again, make sure you can see a nice exponential falloff in your training
loss when starting with the simple overfitting tests. This is a good
setup to figure out an upper bound and reasonable range for the learning rate
as the most central hyperparameter.
You’ll probably need to reduce it later on, but you should at least get a
rough estimate of suitable values for \(\eta\).

\subsubsection{Know your data}
\label{\detokenize{supervised-discuss:know-your-data}}
All data\sphinxhyphen{}driven methods obey the \sphinxstyleemphasis{garbage\sphinxhyphen{}in\sphinxhyphen{}garbage\sphinxhyphen{}out} principle. Because of this it’s important
to work on getting to know the data you are dealing with. While there’s no one\sphinxhyphen{}size\sphinxhyphen{}fits\sphinxhyphen{}all
approach for how to best achieve this, we can strongly recommend to track
a broad range of statistics of your data set. A good starting point are
per quantity mean, standard deviation, min and max values.
If some of these contain unusual values, this is a first indicator of bad
samples in the dataset.

These values can
also be easily visualized in terms of histograms, to track down
unwanted outliers. A small number of such outliers
can easily skew a data set in undesirable ways.

Finally, checking the relationships between different quantities
is often a good idea to get some intuition for what’s contained in the
data set. The next figure gives an example for this step.

\begin{figure}[htbp]
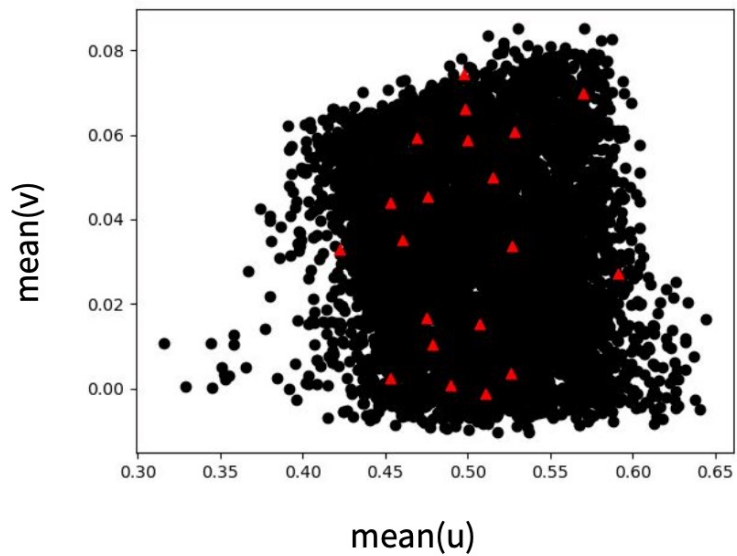

\centering
\capstart

\noindent\sphinxincludegraphics[height=300\sphinxpxdimen]{{supervised-example-plot}.jpg}
\caption{An example from the airfoil case of the previous section: a visualization of a training data
set in terms of mean u and v velocity of 2D flow fields. It nicely shows that there are no extreme outliers,
but there are a few entries with relatively low mean u velocity on the left side.
A second, smaller data set is shown on top in red, showing that its samples cover the range of mean motions quite well.}\label{\detokenize{supervised-discuss:supervised-example-plot}}\end{figure}

\subsubsection{Where’s the magic? 🦄}
\label{\detokenize{supervised-discuss:where-s-the-magic}}
A comment that you’ll often hear when talking about DL approaches, and especially
when using relatively simple training methodologies is: “Isn’t it just interpolating the data?”

Well, \sphinxstylestrong{yes} it is! And that’s exactly what the NN should do. In a way \sphinxhyphen{} there isn’t
anything else to do. This is what \sphinxstyleemphasis{all} DL approaches are about. They give us smooth
representations of the data seen at training time. Even if we’ll use fancy physical
models at training time later on, the NNs just adjust their weights to represent the signals
they receive, and reproduce it.

Due to the hype and numerous success stories, people not familiar with DL often have
the impression that DL works like a human mind, and is able to extract fundamental
and general principles from data sets (\sphinxhref{https://dilbert.com/strip/2000-01-03}{“messages from god”} anyone?).
That’s not what happens with the current state of the art. Nonetheless, it’s
the most powerful tool we have to approximate complex, non\sphinxhyphen{}linear functions.
It is a great tool, but it’s important to keep in mind, that once we set up the training
correctly, all we’ll get out of it is an approximation of the function the NN
was trained for \sphinxhyphen{} no magic involved.

An implication of this is that you shouldn’t expect the network
to work on data it has never seen. In a way, the NNs are so good exactly
because they can accurately adapt to the signals they receive at training time,
but in contrast to other learned representations, they’re actually not very good
at extrapolation. So we can’t expect an NN to magically work with new inputs.
Rather, we need to make sure that we can properly shape the input space,
e.g., by normalization and by focusing on invariants.

To give a more specific example: if you always train
your networks for inputs in the range \([0\dots1]\), don’t expect it to work
with inputs of \([27\dots39]\). In certain cases it’s valid to normalize
inputs and outputs by subtracting the mean, and normalize via the standard
deviation or a suitable quantile (make sure this doesn’t destroy important
correlations in your data).

As a rule of thumb: make sure you actually train the NN on the
inputs that are as similar as possible to those you want to use at inference time.

This is important to keep in mind during the next chapters: e.g., if we
want an NN to work in conjunction with a certain simulation environment,
it’s important to actually include the simulator in the training process. Otherwise,
the network might specialize on pre\sphinxhyphen{}computed data that differs from what is produced
when combining the NN with the solver, i.e it will suffer from \sphinxstyleemphasis{distribution shift}.

\subsubsection{Meshes and grids}
\label{\detokenize{supervised-discuss:meshes-and-grids}}
The previous airfoil example used Cartesian grids with standard
convolutions. These typically give the most \sphinxstyleemphasis{bang\sphinxhyphen{}for\sphinxhyphen{}the\sphinxhyphen{}buck}, in terms
of performance and stability. Nonetheless, the whole discussion here of course
also holds for other types of convolutions, e.g., a less regular mesh
in conjunction with graph\sphinxhyphen{}convolutions, or particle\sphinxhyphen{}based data
with continuous convolutions (cf {\hyperref[\detokenize{others-lagrangian::doc}]{\sphinxcrossref{\DUrole{doc}{Unstructured Meshes and Meshless Methods}}}}). You will typically see reduced learning
performance in exchange for improved sampling flexibility when switching to these.

Finally, a word on fully\sphinxhyphen{}connected layers or \sphinxstyleemphasis{MLPs} in general: we’d recommend
to avoid these as much as possible. For any structured data, like spatial functions,
or \sphinxstyleemphasis{field data} in general, convolutions are preferable, and less likely to overfit.
E.g., you’ll notice that CNNs typically don’t need dropout, as they’re nicely
regularized by construction. For MLPs, you typically need quite a bit to
avoid overfitting.

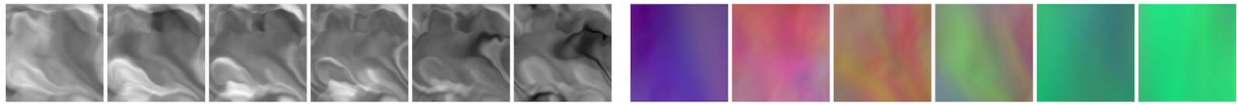
\sphinxincludegraphics{{divider2}.jpg}

\subsection{Supervised training in a nutshell}
\label{\detokenize{supervised-discuss:supervised-training-in-a-nutshell}}
To summarize, supervised training has the following properties.

✅ Pros:
\begin{itemize}
\item {} 
Very fast training.

\item {} 
Stable and simple.

\item {} 
Great starting point.

\end{itemize}

❌ Con:
\begin{itemize}
\item {} 
Lots of data needed.

\item {} 
Sub\sphinxhyphen{}optimal performance, accuracy and generalization.

\item {} 
Interactions with external “processes” (such as embedding into a solver) are difficult.

\end{itemize}

The next chapters will explain how to alleviate these shortcomings of supervised training.
First, we’ll look at bringing model equations into the picture via soft\sphinxhyphen{}constraints, and afterwards
we’ll revisit the challenges of bringing together numerical simulations and learned approaches.

\part{Physical Losses}

\chapter{Physical Loss Terms}
\label{\detokenize{physicalloss:physical-loss-terms}}\label{\detokenize{physicalloss::doc}}
The supervised setting of the previous sections can quickly
yield approximate solutions with a fairly simple training process. However, what’s
quite sad to see here is that we only use physical models and numerical methods
as an “external” tool to produce a big pile of data 😢.

We as humans have a lot of knowledge about how to describe physical processes
mathematically. As the following chapters will show, we can improve the
training process by guiding it with our human knowledge of physics.

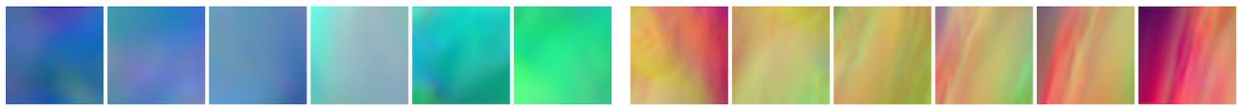
\begin{figure}[htbp]
\centering
\capstart

\noindent\sphinxincludegraphics[height=220\sphinxpxdimen]{{physloss-overview}.jpg}
\caption{Physical losses typically combine a supervised loss with a combination of derivatives from the neural network.}\label{\detokenize{physicalloss:physloss-overview}}\end{figure}

\section{Using physical models}
\label{\detokenize{physicalloss:using-physical-models}}
Given a PDE for \(\mathbf{u}(\mathbf{x},t)\) with a time evolution,
we can typically express it in terms of a function \(\mathcal F\) of the derivatives
of \(\mathbf{u}\) via
\begin{equation*}
\begin{split}
  \mathbf{u}_t = \mathcal F ( \mathbf{u}_{x}, \mathbf{u}_{xx}, ... \mathbf{u}_{xx...x} ) ,
\end{split}
\end{equation*}
where the \(_{\mathbf{x}}\) subscripts denote spatial derivatives with respect to one of the spatial dimensions
of higher and higher order (this can of course also include mixed derivatives with respect to different axes). \textbackslash{}mathbf\{u\}\_t denotes the changes over time.

In this context, we can approximate the unknown \(\mathbf{u}\) itself with a neural network. If the approximation, which we call \(\tilde{\mathbf{u}}\), is accurate, the PDE should be satisfied naturally. In other words, the residual R should be equal to zero:
\begin{equation*}
\begin{split}
  R = \mathbf{u}_t - \mathcal F ( \mathbf{u}_{x}, \mathbf{u}_{xx}, ... \mathbf{u}_{xx...x} ) = 0 .
\end{split}
\end{equation*}
This nicely integrates with the objective for training a neural network: we can train for
minimizing this residual in combination with direct loss terms.
Similar to before, we can make use of sample solutions
\([x_0,y_0], ...[x_n,y_n]\) for \(\mathbf{u}\) with \(\mathbf{u}(\mathbf{x})=y\).
This is typically important, as most practical PDEs we encounter do not have unique solutions
unless initial and boundary conditions are specified. Hence, if we only consider \(R\) we might
get solutions with random offset or other undesirable components. The supervised sample points
therefore help to \sphinxstyleemphasis{pin down} the solution in certain places.
Now our training objective becomes
\begin{equation}\label{equation:physicalloss:physloss-training}
\begin{split}
\text{arg min}_{\theta} \ \alpha_0 \sum_i \big( f(x_i ; \theta)-y_i \big)^2 + \alpha_1 R(x_i) ,
\end{split}
\end{equation}
where \(\alpha_{0,1}\) denote hyperparameters that scale the contribution of the supervised term and
the residual term, respectively. We could of course add additional residual terms with suitable scaling factors here.

It is instructive to note what the two different terms in equation \eqref{equation:physicalloss:physloss-training} mean: The first term is a conventional, supervised L2\sphinxhyphen{}loss. If we were to optimize only this loss, our network would learn to approximate the training samples well, but might average multiple modes in the solutions, and do poorly in regions in between the sample points.
If we, instead, were to optimize only the second term (the physical residual), our neural network might be able to locally satisfy the PDE, but still could produce solutions that are still far away from our training data. This can happen due to “null spaces” in the solutions, i.e., different solutions that all satisfy the residuals.
Therefore, we optimize both objectives simultaneously such that, in the best case, the network learns to approximate the specific solutions of the training data while still capturing knowledge about the underlying PDE.

Note that, similar to the data samples used for supervised training, we have no guarantees that the
residual terms \(R\) will actually reach zero during training. The non\sphinxhyphen{}linear optimization of the training process
will minimize the supervised and residual terms as much as possible, but there is no guarantee. Large, non\sphinxhyphen{}zero residual
contributions can remain. We’ll look at this in more detail in the upcoming code example, for now it’s important
to remember that physical constraints in this way only represent \sphinxstyleemphasis{soft\sphinxhyphen{}constraints}, without guarantees
of minimizing these constraints.

\section{Neural network derivatives}
\label{\detokenize{physicalloss:neural-network-derivatives}}
In order to compute the residuals at training time, it would be possible to store
the unknowns of \(\mathbf{u}\) on a computational mesh, e.g., a grid, and discretize the equations of
\(R\) there. This has a fairly long “tradition” in DL, and was proposed by Tompson et al. {[}\hyperlink{cite.references:id80}{TSSP17}{]} early on.

A popular variant of employing physical soft\sphinxhyphen{}constraints {[}\hyperlink{cite.references:id81}{RK18}{]}
instead uses fully connected NNs to represent \(\mathbf{u}\). This has some interesting pros and cons that we’ll outline in the following, and we will also focus on it in the following code examples and comparisons.

The central idea here is that the aforementioned general function \(f\) that we’re after in our learning problems
can also be used to obtain a representation of a physical field, e.g., a field \(\mathbf{u}\) that satisfies \(R=0\). This means \(\mathbf{u}(\mathbf{x})\) will
be turned into \(\mathbf{u}(\mathbf{x}, \theta)\) where we choose the NN parameters \(\theta\) such that a desired \(\mathbf{u}\) is
represented as precisely as possible.

One nice side effect of this viewpoint is that NN representations inherently support the calculation of derivatives.
The derivative \(\partial f / \partial \theta\) was a key building block for learning via gradient descent, as explained
in {\hyperref[\detokenize{overview::doc}]{\sphinxcrossref{\DUrole{doc}{Overview}}}}. Now, we can use the same tools to compute spatial derivatives such as \(\partial \mathbf{u} / \partial x\),
Note that above for \(R\) we’ve written this derivative in the shortened notation as \(\mathbf{u}_{x}\).
For functions over time this of course also works for \(\partial \mathbf{u} / \partial t\), i.e. \(\mathbf{u}_{t}\) in the notation above.

Thus, for some generic \(R\), made up of \(\mathbf{u}_t\) and \(\mathbf{u}_{x}\) terms, we can rely on the backpropagation algorithm
of DL frameworks to compute these derivatives once we have a NN that represents \(\mathbf{u}\). Essentially, this gives us a
function (the NN) that receives space and time coordinates to produce a solution for \(\mathbf{u}\). Hence, the input is typically
quite low\sphinxhyphen{}dimensional, e.g., 3+1 values for a 3D case over time, and often produces a scalar value or a spatial vector.
Due to the lack of explicit spatial sampling points, an MLP, i.e., fully\sphinxhyphen{}connected NN is the architecture of choice here.

To pick a simple example, Burgers equation in 1D,
\(\frac{\partial u}{\partial{t}} + u \nabla u = \nu \nabla \cdot \nabla u \) , we can directly
formulate a loss term \(R = \frac{\partial u}{\partial t} + u \frac{\partial u}{\partial x} - \nu \frac{\partial^2 u}{\partial x^2} u\) that should be minimized as much as possible at training time. For each of the terms, e.g. \(\frac{\partial u}{\partial x}\),
we can simply query the DL framework that realizes \(u\) to obtain the corresponding derivative.
For higher order derivatives, such as \(\frac{\partial^2 u}{\partial x^2}\), we can simply query the derivative function of the framework multiple times. In the following section, we’ll give a specific example of how that works in tensorflow.

\section{Summary so far}
\label{\detokenize{physicalloss:summary-so-far}}
The approach above gives us a method to include physical equations into DL learning as a soft\sphinxhyphen{}constraint: the residual loss.
Typically, this setup is suitable for \sphinxstyleemphasis{inverse problems}, where we have certain measurements or observations
for which we want to find a PDE solution. Because of the high cost of the reconstruction (to be
demonstrated in the following), the solution manifold shouldn’t be overly complex. E.g., it is not possible
to capture a wide range of solutions, such as with the previous supervised airfoil example, with such a physical residual loss.

\chapter{Burgers Optimization with a Physics\sphinxhyphen{}Informed NN}
\label{\detokenize{physicalloss-code:burgers-optimization-with-a-physics-informed-nn}}\label{\detokenize{physicalloss-code::doc}}
To illustrate how the physics\sphinxhyphen{}informed losses work, let’s consider a reconstruction task
as an inverse problem example.
We’ll use Burgers equation \(\frac{\partial u}{\partial{t}} + u \nabla u = \nu \nabla \cdot \nabla u\) as a simple yet non\sphinxhyphen{}linear equation in 1D, for which we have a series of \sphinxstyleemphasis{observations} at time \(t=0.5\).
The solution should fulfill the residual formulation for Burgers equation and match the observations.
In addition, let’s impose Dirichlet boundary conditions \(u=0\)
at the sides of our computational domain, and define the solution in
the time interval \(t \in [0,1]\).

Note that similar to the previous forward simulation example,
we will still be sampling the solution with 128 points (\(n=128\)), but now we have a discretization via the NN. So we could also sample points in between without having to explicitly choose a basis function for interpolation. The discretization via the NN now internally determines how to use its degrees of freedom to arrange the activation functions as basis functions. So we have no direct control over the reconstruction.
\sphinxhref{https://colab.research.google.com/github/tum-pbs/pbdl-book/blob/main/physicalloss-code.ipynb}{{[}run in colab{]}}

\section{Formulation}
\label{\detokenize{physicalloss-code:formulation}}
In terms of the \(x,y^*\) notation from {\hyperref[\detokenize{overview-equations::doc}]{\sphinxcrossref{\DUrole{doc}{Models and Equations}}}} and the previous section, this reconstruction problem means we are solving
\begin{equation*}
\begin{split}
\text{arg min}_{\theta} \sum_i ( f(x_i ; \theta)-y^*_i )^2 + R(x_i) ,
\end{split}
\end{equation*}
where \(x\) and \(y^*\) are solutions of \(u\) at different locations in space and time. As we’re dealing with a 1D velocity, \(x,y^* \in \mathbb{R}\).
They both represent two\sphinxhyphen{}dimensional solutions
\(x(p_i,t_i)\) and \(y^*(p_i,t_i)\) for a spatial coordinate \(p_i\) and a time \(t_i\), where the index \(i\) sums over a set of chosen \(p_i,t_i\) locations at which we evaluate the PDE and the approximated solutions. Thus \(y^*\) denotes a reference \(u\) for \(\mathcal{P}\) being Burgers equation, which \(x\) should approximate as closely as possible. Thus our neural network representation of \(x\) will receive \(p,t\) as input to produce a velocity solution at the specified position.

The residual function \(R\) above collects additional evaluations of \(f(;\theta)\) and its derivatives to formulate the residual for \(\mathcal{P}\). This approach – using derivatives of a neural network to compute a PDE residual – is typically called a \sphinxstyleemphasis{physics\sphinxhyphen{}informed} approach, yielding a \sphinxstyleemphasis{physics\sphinxhyphen{}informed neural network} (PINN) to represent a solution for the inverse reconstruction problem.

Thus, in the formulation above, \(R\) should simply converge to zero above. We’ve omitted scaling factors in the objective function for simplicity. Note that, effectively, we’re only dealing with individual point samples of a single solution \(u\) for \(\mathcal{P}\) here.

\section{Preliminaries}
\label{\detokenize{physicalloss-code:preliminaries}}
Let’s just load phiflow with the tensorflow backend for now, and initialize the random sampling. (\sphinxstyleemphasis{Note: this example uses an older version 1.5.1 of phiflow.})

\begin{sphinxVerbatim}[commandchars=\\\{\}]
\PYG{o}{!}pip install \PYGZhy{}\PYGZhy{}upgrade \PYGZhy{}\PYGZhy{}quiet git+https://github.com/tum\PYGZhy{}pbs/PhiFlow@1.5.1

\PYG{k+kn}{from} \PYG{n+nn}{phi}\PYG{n+nn}{.}\PYG{n+nn}{tf}\PYG{n+nn}{.}\PYG{n+nn}{flow} \PYG{k+kn}{import} \PYG{o}{*}
\PYG{k+kn}{import} \PYG{n+nn}{numpy} \PYG{k}{as} \PYG{n+nn}{np}

\PYG{c+c1}{\PYGZsh{}rnd = TF\PYGZus{}BACKEND  \PYGZsh{} for phiflow: sample different points in the domain each iteration}
\PYG{n}{rnd} \PYG{o}{=} \PYG{n}{math}\PYG{o}{.}\PYG{n}{choose\PYGZus{}backend}\PYG{p}{(}\PYG{l+m+mi}{1}\PYG{p}{)}  \PYG{c+c1}{\PYGZsh{} use same random points for all iterations}
\end{sphinxVerbatim}

We’re importing phiflow here, but we won’t use it to compute a solution to the PDE as in {\hyperref[\detokenize{overview-burgers-forw::doc}]{\sphinxcrossref{\DUrole{doc}{Simple Forward Simulation of Burgers Equation with phiflow}}}}. Instead, we’ll use the  derivatives of an NN (as explained in the previous section) to set up a loss formulation for training.

Next, we set up a simple NN with 8 fully connected layers and \sphinxcode{\sphinxupquote{tanh}} activations with 20 units each.

We’ll also define the \sphinxcode{\sphinxupquote{boundary\_tx}} function which gives an array of constraints for the solution (all for \(=0.5\) in this example), and the \sphinxcode{\sphinxupquote{open\_boundary}} function which stores constraints for \(x= \pm1\) being 0.

\begin{sphinxVerbatim}[commandchars=\\\{\}]
\PYG{k}{def} \PYG{n+nf}{network}\PYG{p}{(}\PYG{n}{x}\PYG{p}{,} \PYG{n}{t}\PYG{p}{)}\PYG{p}{:}
    \PYG{l+s+sd}{\PYGZdq{}\PYGZdq{}\PYGZdq{} Dense neural network with 8 hidden layers and 3021 parameters in total.}
\PYG{l+s+sd}{        Parameters will only be allocated once (auto reuse).}
\PYG{l+s+sd}{    \PYGZdq{}\PYGZdq{}\PYGZdq{}}
    \PYG{n}{y} \PYG{o}{=} \PYG{n}{math}\PYG{o}{.}\PYG{n}{stack}\PYG{p}{(}\PYG{p}{[}\PYG{n}{x}\PYG{p}{,} \PYG{n}{t}\PYG{p}{]}\PYG{p}{,} \PYG{n}{axis}\PYG{o}{=}\PYG{o}{\PYGZhy{}}\PYG{l+m+mi}{1}\PYG{p}{)}
    \PYG{k}{for} \PYG{n}{i} \PYG{o+ow}{in} \PYG{n+nb}{range}\PYG{p}{(}\PYG{l+m+mi}{8}\PYG{p}{)}\PYG{p}{:}
        \PYG{n}{y} \PYG{o}{=} \PYG{n}{tf}\PYG{o}{.}\PYG{n}{layers}\PYG{o}{.}\PYG{n}{dense}\PYG{p}{(}\PYG{n}{y}\PYG{p}{,} \PYG{l+m+mi}{20}\PYG{p}{,} \PYG{n}{activation}\PYG{o}{=}\PYG{n}{tf}\PYG{o}{.}\PYG{n}{math}\PYG{o}{.}\PYG{n}{tanh}\PYG{p}{,} \PYG{n}{name}\PYG{o}{=}\PYG{l+s+s1}{\PYGZsq{}}\PYG{l+s+s1}{layer}\PYG{l+s+si}{\PYGZpc{}d}\PYG{l+s+s1}{\PYGZsq{}} \PYG{o}{\PYGZpc{}} \PYG{n}{i}\PYG{p}{,} \PYG{n}{reuse}\PYG{o}{=}\PYG{n}{tf}\PYG{o}{.}\PYG{n}{AUTO\PYGZus{}REUSE}\PYG{p}{)}
    \PYG{k}{return} \PYG{n}{tf}\PYG{o}{.}\PYG{n}{layers}\PYG{o}{.}\PYG{n}{dense}\PYG{p}{(}\PYG{n}{y}\PYG{p}{,} \PYG{l+m+mi}{1}\PYG{p}{,} \PYG{n}{activation}\PYG{o}{=}\PYG{k+kc}{None}\PYG{p}{,} \PYG{n}{name}\PYG{o}{=}\PYG{l+s+s1}{\PYGZsq{}}\PYG{l+s+s1}{layer\PYGZus{}out}\PYG{l+s+s1}{\PYGZsq{}}\PYG{p}{,} \PYG{n}{reuse}\PYG{o}{=}\PYG{n}{tf}\PYG{o}{.}\PYG{n}{AUTO\PYGZus{}REUSE}\PYG{p}{)}

\PYG{k}{def} \PYG{n+nf}{boundary\PYGZus{}tx}\PYG{p}{(}\PYG{n}{N}\PYG{p}{)}\PYG{p}{:}
    \PYG{n}{x} \PYG{o}{=} \PYG{n}{np}\PYG{o}{.}\PYG{n}{linspace}\PYG{p}{(}\PYG{o}{\PYGZhy{}}\PYG{l+m+mi}{1}\PYG{p}{,}\PYG{l+m+mi}{1}\PYG{p}{,}\PYG{l+m+mi}{128}\PYG{p}{)}
    \PYG{c+c1}{\PYGZsh{} precomputed solution from forward simulation:}
    \PYG{n}{u} \PYG{o}{=} \PYG{n}{np}\PYG{o}{.}\PYG{n}{asarray}\PYG{p}{(} \PYG{p}{[}\PYG{l+m+mf}{0.008612174447657694}\PYG{p}{,} \PYG{l+m+mf}{0.02584669669548606}\PYG{p}{,} \PYG{l+m+mf}{0.043136357266407785} \PYG{o}{.}\PYG{o}{.}\PYG{o}{.} \PYG{p}{]} \PYG{p}{)}\PYG{p}{;}
    \PYG{n}{t} \PYG{o}{=} \PYG{n}{np}\PYG{o}{.}\PYG{n}{asarray}\PYG{p}{(}\PYG{n}{rnd}\PYG{o}{.}\PYG{n}{ones\PYGZus{}like}\PYG{p}{(}\PYG{n}{x}\PYG{p}{)}\PYG{p}{)} \PYG{o}{*} \PYG{l+m+mf}{0.5}
    \PYG{n}{perm} \PYG{o}{=} \PYG{n}{np}\PYG{o}{.}\PYG{n}{random}\PYG{o}{.}\PYG{n}{permutation}\PYG{p}{(}\PYG{l+m+mi}{128}\PYG{p}{)} 
    \PYG{k}{return} \PYG{p}{(}\PYG{n}{x}\PYG{p}{[}\PYG{n}{perm}\PYG{p}{]}\PYG{p}{)}\PYG{p}{[}\PYG{l+m+mi}{0}\PYG{p}{:}\PYG{n}{N}\PYG{p}{]}\PYG{p}{,} \PYG{p}{(}\PYG{n}{t}\PYG{p}{[}\PYG{n}{perm}\PYG{p}{]}\PYG{p}{)}\PYG{p}{[}\PYG{l+m+mi}{0}\PYG{p}{:}\PYG{n}{N}\PYG{p}{]}\PYG{p}{,} \PYG{p}{(}\PYG{n}{u}\PYG{p}{[}\PYG{n}{perm}\PYG{p}{]}\PYG{p}{)}\PYG{p}{[}\PYG{l+m+mi}{0}\PYG{p}{:}\PYG{n}{N}\PYG{p}{]}

\PYG{k}{def} \PYG{n+nf}{\PYGZus{}ALT\PYGZus{}t0}\PYG{p}{(}\PYG{n}{N}\PYG{p}{)}\PYG{p}{:} \PYG{c+c1}{\PYGZsh{} alternative, impose original initial state at t=0}
    \PYG{n}{x} \PYG{o}{=} \PYG{n}{rnd}\PYG{o}{.}\PYG{n}{random\PYGZus{}uniform}\PYG{p}{(}\PYG{p}{[}\PYG{n}{N}\PYG{p}{]}\PYG{p}{,} \PYG{o}{\PYGZhy{}}\PYG{l+m+mi}{1}\PYG{p}{,} \PYG{l+m+mi}{1}\PYG{p}{)}
    \PYG{n}{t} \PYG{o}{=} \PYG{n}{rnd}\PYG{o}{.}\PYG{n}{zeros\PYGZus{}like}\PYG{p}{(}\PYG{n}{x}\PYG{p}{)}
    \PYG{n}{u} \PYG{o}{=} \PYG{o}{\PYGZhy{}} \PYG{n}{math}\PYG{o}{.}\PYG{n}{sin}\PYG{p}{(}\PYG{n}{np}\PYG{o}{.}\PYG{n}{pi} \PYG{o}{*} \PYG{n}{x}\PYG{p}{)}
    \PYG{k}{return} \PYG{n}{x}\PYG{p}{,} \PYG{n}{t}\PYG{p}{,} \PYG{n}{u}

\PYG{k}{def} \PYG{n+nf}{open\PYGZus{}boundary}\PYG{p}{(}\PYG{n}{N}\PYG{p}{)}\PYG{p}{:}
    \PYG{n}{t} \PYG{o}{=} \PYG{n}{rnd}\PYG{o}{.}\PYG{n}{random\PYGZus{}uniform}\PYG{p}{(}\PYG{p}{[}\PYG{n}{N}\PYG{p}{]}\PYG{p}{,} \PYG{l+m+mi}{0}\PYG{p}{,} \PYG{l+m+mi}{1}\PYG{p}{)}
    \PYG{n}{x} \PYG{o}{=} \PYG{n}{math}\PYG{o}{.}\PYG{n}{concat}\PYG{p}{(}\PYG{p}{[}\PYG{n}{math}\PYG{o}{.}\PYG{n}{zeros}\PYG{p}{(}\PYG{p}{[}\PYG{n}{N}\PYG{o}{/}\PYG{o}{/}\PYG{l+m+mi}{2}\PYG{p}{]}\PYG{p}{)} \PYG{o}{+} \PYG{l+m+mi}{1}\PYG{p}{,} \PYG{n}{math}\PYG{o}{.}\PYG{n}{zeros}\PYG{p}{(}\PYG{p}{[}\PYG{n}{N}\PYG{o}{/}\PYG{o}{/}\PYG{l+m+mi}{2}\PYG{p}{]}\PYG{p}{)} \PYG{o}{\PYGZhy{}} \PYG{l+m+mi}{1}\PYG{p}{]}\PYG{p}{,} \PYG{n}{axis}\PYG{o}{=}\PYG{l+m+mi}{0}\PYG{p}{)}
    \PYG{n}{u} \PYG{o}{=} \PYG{n}{math}\PYG{o}{.}\PYG{n}{zeros}\PYG{p}{(}\PYG{p}{[}\PYG{n}{N}\PYG{p}{]}\PYG{p}{)}
    \PYG{k}{return} \PYG{n}{x}\PYG{p}{,} \PYG{n}{t}\PYG{p}{,} \PYG{n}{u}
\end{sphinxVerbatim}

Most importantly, we can now also construct the residual loss function \sphinxcode{\sphinxupquote{f}} that we’d like to minimize in order to guide the NN to retrieve a solution for our model equation. As can be seen in the equation at the top, we need derivatives w.r.t. \(t\), \(x\) and a second derivative for \(x\). The first three lines of \sphinxcode{\sphinxupquote{f}} below do just that.

Afterwards, we simply combine the derivates to form Burgers equation. Here we make use of phiflow’s \sphinxcode{\sphinxupquote{gradient}} function:

\begin{sphinxVerbatim}[commandchars=\\\{\}]
\PYG{k}{def} \PYG{n+nf}{f}\PYG{p}{(}\PYG{n}{u}\PYG{p}{,} \PYG{n}{x}\PYG{p}{,} \PYG{n}{t}\PYG{p}{)}\PYG{p}{:}
    \PYG{l+s+sd}{\PYGZdq{}\PYGZdq{}\PYGZdq{} Physics\PYGZhy{}based loss function with Burgers equation \PYGZdq{}\PYGZdq{}\PYGZdq{}}
    \PYG{n}{u\PYGZus{}t} \PYG{o}{=} \PYG{n}{gradients}\PYG{p}{(}\PYG{n}{u}\PYG{p}{,} \PYG{n}{t}\PYG{p}{)}
    \PYG{n}{u\PYGZus{}x} \PYG{o}{=} \PYG{n}{gradients}\PYG{p}{(}\PYG{n}{u}\PYG{p}{,} \PYG{n}{x}\PYG{p}{)}
    \PYG{n}{u\PYGZus{}xx} \PYG{o}{=} \PYG{n}{gradients}\PYG{p}{(}\PYG{n}{u\PYGZus{}x}\PYG{p}{,} \PYG{n}{x}\PYG{p}{)}
    \PYG{k}{return} \PYG{n}{u\PYGZus{}t} \PYG{o}{+} \PYG{n}{u}\PYG{o}{*}\PYG{n}{u\PYGZus{}x} \PYG{o}{\PYGZhy{}} \PYG{p}{(}\PYG{l+m+mf}{0.01} \PYG{o}{/} \PYG{n}{np}\PYG{o}{.}\PYG{n}{pi}\PYG{p}{)} \PYG{o}{*} \PYG{n}{u\PYGZus{}xx}
\end{sphinxVerbatim}

Next, let’s set up the sampling points in the inner domain, such that we can compare the solution with the previous forward simulation in phiflow.

The next cell allocates two tensors: \sphinxcode{\sphinxupquote{grid\_x}} will cover the size of our domain, i.e., the \sphinxhyphen{}1 to 1 range, with 128 cells, while \sphinxcode{\sphinxupquote{grid\_t}} will sample the time interval \([0,1]\) with 33 time stamps.

The last \sphinxcode{\sphinxupquote{math.expand\_dims()}} call simply adds another \sphinxcode{\sphinxupquote{batch}} dimension, so that the resulting tensor is compatible with the following examples.

\begin{sphinxVerbatim}[commandchars=\\\{\}]
\PYG{n}{N}\PYG{o}{=}\PYG{l+m+mi}{128}
\PYG{n}{grids\PYGZus{}xt} \PYG{o}{=} \PYG{n}{np}\PYG{o}{.}\PYG{n}{meshgrid}\PYG{p}{(}\PYG{n}{np}\PYG{o}{.}\PYG{n}{linspace}\PYG{p}{(}\PYG{o}{\PYGZhy{}}\PYG{l+m+mi}{1}\PYG{p}{,} \PYG{l+m+mi}{1}\PYG{p}{,} \PYG{n}{N}\PYG{p}{)}\PYG{p}{,} \PYG{n}{np}\PYG{o}{.}\PYG{n}{linspace}\PYG{p}{(}\PYG{l+m+mi}{0}\PYG{p}{,} \PYG{l+m+mi}{1}\PYG{p}{,} \PYG{l+m+mi}{33}\PYG{p}{)}\PYG{p}{,} \PYG{n}{indexing}\PYG{o}{=}\PYG{l+s+s1}{\PYGZsq{}}\PYG{l+s+s1}{ij}\PYG{l+s+s1}{\PYGZsq{}}\PYG{p}{)}
\PYG{n}{grid\PYGZus{}x}\PYG{p}{,} \PYG{n}{grid\PYGZus{}t} \PYG{o}{=} \PYG{p}{[}\PYG{n}{tf}\PYG{o}{.}\PYG{n}{convert\PYGZus{}to\PYGZus{}tensor}\PYG{p}{(}\PYG{n}{t}\PYG{p}{,} \PYG{n}{tf}\PYG{o}{.}\PYG{n}{float32}\PYG{p}{)} \PYG{k}{for} \PYG{n}{t} \PYG{o+ow}{in} \PYG{n}{grids\PYGZus{}xt}\PYG{p}{]}

\PYG{c+c1}{\PYGZsh{} create 4D tensor with batch and channel dimensions in addition to space and time}
\PYG{c+c1}{\PYGZsh{} in this case gives shape=(1, N, 33, 1)}
\PYG{n}{grid\PYGZus{}u} \PYG{o}{=} \PYG{n}{math}\PYG{o}{.}\PYG{n}{expand\PYGZus{}dims}\PYG{p}{(}\PYG{n}{network}\PYG{p}{(}\PYG{n}{grid\PYGZus{}x}\PYG{p}{,} \PYG{n}{grid\PYGZus{}t}\PYG{p}{)}\PYG{p}{)}
\end{sphinxVerbatim}

\begin{sphinxVerbatim}[commandchars=\\\{\}]
Instructions for updating:
Use keras.layers.Dense instead.
Instructions for updating:
Please use `layer.\PYGZus{}\PYGZus{}call\PYGZus{}\PYGZus{}` method instead.
\end{sphinxVerbatim}

Now, \sphinxcode{\sphinxupquote{grid\_u}} contains a full graph to evaluate our NN at \(128 \times 33\) positions, and returns the results in a \([1,128,33,1]\) array once we run it through \sphinxcode{\sphinxupquote{session.run}}. Let’s give this a try: we can initialize a TF session, evaluate \sphinxcode{\sphinxupquote{grid\_u}} and show it in an image, just like the phiflow solution we computed previously.

(Note, we’ll use the \sphinxcode{\sphinxupquote{show\_state}} as in {\hyperref[\detokenize{overview-burgers-forw::doc}]{\sphinxcrossref{\DUrole{doc}{Simple Forward Simulation of Burgers Equation with phiflow}}}}. Hence, the x axis does not show actual simulation time, but is showing 32 steps “blown” up by a factor of 16 to make the changes over time easier to see in the image.)

\begin{sphinxVerbatim}[commandchars=\\\{\}]
\PYG{k+kn}{import} \PYG{n+nn}{pylab} \PYG{k}{as} \PYG{n+nn}{plt}
\PYG{n+nb}{print}\PYG{p}{(}\PYG{l+s+s2}{\PYGZdq{}}\PYG{l+s+s2}{Size of grid\PYGZus{}u: }\PYG{l+s+s2}{\PYGZdq{}}\PYG{o}{+}\PYG{n+nb}{format}\PYG{p}{(}\PYG{n}{grid\PYGZus{}u}\PYG{o}{.}\PYG{n}{shape}\PYG{p}{)}\PYG{p}{)}

\PYG{n}{session} \PYG{o}{=} \PYG{n}{Session}\PYG{p}{(}\PYG{k+kc}{None}\PYG{p}{)}
\PYG{n}{session}\PYG{o}{.}\PYG{n}{initialize\PYGZus{}variables}\PYG{p}{(}\PYG{p}{)}

\PYG{k}{def} \PYG{n+nf}{show\PYGZus{}state}\PYG{p}{(}\PYG{n}{a}\PYG{p}{,} \PYG{n}{title}\PYG{p}{)}\PYG{p}{:}
    \PYG{k}{for} \PYG{n}{i} \PYG{o+ow}{in} \PYG{n+nb}{range}\PYG{p}{(}\PYG{l+m+mi}{4}\PYG{p}{)}\PYG{p}{:} \PYG{n}{a} \PYG{o}{=} \PYG{n}{np}\PYG{o}{.}\PYG{n}{concatenate}\PYG{p}{(} \PYG{p}{[}\PYG{n}{a}\PYG{p}{,}\PYG{n}{a}\PYG{p}{]} \PYG{p}{,} \PYG{n}{axis}\PYG{o}{=}\PYG{l+m+mi}{3}\PYG{p}{)}
    \PYG{n}{a} \PYG{o}{=} \PYG{n}{np}\PYG{o}{.}\PYG{n}{reshape}\PYG{p}{(} \PYG{n}{a}\PYG{p}{,} \PYG{p}{[}\PYG{n}{a}\PYG{o}{.}\PYG{n}{shape}\PYG{p}{[}\PYG{l+m+mi}{1}\PYG{p}{]}\PYG{p}{,}\PYG{n}{a}\PYG{o}{.}\PYG{n}{shape}\PYG{p}{[}\PYG{l+m+mi}{2}\PYG{p}{]}\PYG{o}{*}\PYG{n}{a}\PYG{o}{.}\PYG{n}{shape}\PYG{p}{[}\PYG{l+m+mi}{3}\PYG{p}{]}\PYG{p}{]} \PYG{p}{)}
    \PYG{n}{fig}\PYG{p}{,} \PYG{n}{axes} \PYG{o}{=} \PYG{n}{plt}\PYG{o}{.}\PYG{n}{subplots}\PYG{p}{(}\PYG{l+m+mi}{1}\PYG{p}{,} \PYG{l+m+mi}{1}\PYG{p}{,} \PYG{n}{figsize}\PYG{o}{=}\PYG{p}{(}\PYG{l+m+mi}{16}\PYG{p}{,} \PYG{l+m+mi}{5}\PYG{p}{)}\PYG{p}{)}
    \PYG{n}{im} \PYG{o}{=} \PYG{n}{axes}\PYG{o}{.}\PYG{n}{imshow}\PYG{p}{(}\PYG{n}{a}\PYG{p}{,} \PYG{n}{origin}\PYG{o}{=}\PYG{l+s+s1}{\PYGZsq{}}\PYG{l+s+s1}{upper}\PYG{l+s+s1}{\PYGZsq{}}\PYG{p}{,} \PYG{n}{cmap}\PYG{o}{=}\PYG{l+s+s1}{\PYGZsq{}}\PYG{l+s+s1}{inferno}\PYG{l+s+s1}{\PYGZsq{}}\PYG{p}{)}
    \PYG{n}{plt}\PYG{o}{.}\PYG{n}{colorbar}\PYG{p}{(}\PYG{n}{im}\PYG{p}{)} \PYG{p}{;} \PYG{n}{plt}\PYG{o}{.}\PYG{n}{xlabel}\PYG{p}{(}\PYG{l+s+s1}{\PYGZsq{}}\PYG{l+s+s1}{time}\PYG{l+s+s1}{\PYGZsq{}}\PYG{p}{)}\PYG{p}{;} \PYG{n}{plt}\PYG{o}{.}\PYG{n}{ylabel}\PYG{p}{(}\PYG{l+s+s1}{\PYGZsq{}}\PYG{l+s+s1}{x}\PYG{l+s+s1}{\PYGZsq{}}\PYG{p}{)}\PYG{p}{;} \PYG{n}{plt}\PYG{o}{.}\PYG{n}{title}\PYG{p}{(}\PYG{n}{title}\PYG{p}{)}
    
\PYG{n+nb}{print}\PYG{p}{(}\PYG{l+s+s2}{\PYGZdq{}}\PYG{l+s+s2}{Randomly initialized network state:}\PYG{l+s+s2}{\PYGZdq{}}\PYG{p}{)}
\PYG{n}{show\PYGZus{}state}\PYG{p}{(}\PYG{n}{session}\PYG{o}{.}\PYG{n}{run}\PYG{p}{(}\PYG{n}{grid\PYGZus{}u}\PYG{p}{)}\PYG{p}{,}\PYG{l+s+s2}{\PYGZdq{}}\PYG{l+s+s2}{Uninitialized NN}\PYG{l+s+s2}{\PYGZdq{}}\PYG{p}{)}
\end{sphinxVerbatim}

\begin{sphinxVerbatim}[commandchars=\\\{\}]
Size of grid\PYGZus{}u: (1, 128, 33, 1)
Randomly initialized network state:
\end{sphinxVerbatim}

\noindent\sphinxincludegraphics{{physicalloss-code_11_1}.png}

This visualization already shows a smooth transition over space and time. So far, this is purely the random initialization of the NN that we’re sampling here. So it has nothing to do with a solution of our PDE\sphinxhyphen{}based model up to now.

The next steps will actually evaluate the constraints in terms of data (from the \sphinxcode{\sphinxupquote{boundary}} functions), and the model constraints from \sphinxcode{\sphinxupquote{f}} to retrieve an actual solution to the PDE.

\section{Loss function and training}
\label{\detokenize{physicalloss-code:loss-function-and-training}}
As objective for the learning process we can now combine the \sphinxstyleemphasis{direct} constraints, i.e., the solution at \(t=0.5\) and the Dirichlet \(u=0\) boundary conditions with the loss from the PDE residuals. For both boundary constraints we’ll use 100 points below, and then sample the solution in the inner region with an additional 1000 points.

The direct constraints are evaluated via \sphinxcode{\sphinxupquote{network(x, t){[}:, 0{]} \sphinxhyphen{} u}}, where \sphinxcode{\sphinxupquote{x}} and \sphinxcode{\sphinxupquote{t}} are the space\sphinxhyphen{}time location where we’d like to sample the solution, and \sphinxcode{\sphinxupquote{u}} provides the corresponding ground truth value.

For the physical loss points, we have no ground truth solutions, but we’ll only evaluate the PDE residual via the NN derivatives, to see whether the solution satisfies the PDE model. If not, this directly gives us an error to be reduced via an update step in the optimization. The corresponding expression is of the form  \sphinxcode{\sphinxupquote{f(network(x, t){[}:, 0{]}, x, t)}} below. Note that for both data and physics terms the \sphinxcode{\sphinxupquote{network(){[}:, 0{]}}} expressions don’t remove any data from the \(L^2\) evaluation, they simply discard the last size\sphinxhyphen{}1 dimension of the \((n,1)\) tensor returned by the network.

\begin{sphinxVerbatim}[commandchars=\\\{\}]
\PYG{c+c1}{\PYGZsh{} Boundary loss}
\PYG{n}{N\PYGZus{}SAMPLE\PYGZus{}POINTS\PYGZus{}BND} \PYG{o}{=} \PYG{l+m+mi}{100}
\PYG{n}{x\PYGZus{}bc}\PYG{p}{,} \PYG{n}{t\PYGZus{}bc}\PYG{p}{,} \PYG{n}{u\PYGZus{}bc} \PYG{o}{=} \PYG{p}{[}\PYG{n}{math}\PYG{o}{.}\PYG{n}{concat}\PYG{p}{(}\PYG{p}{[}\PYG{n}{v\PYGZus{}t0}\PYG{p}{,} \PYG{n}{v\PYGZus{}x}\PYG{p}{]}\PYG{p}{,} \PYG{n}{axis}\PYG{o}{=}\PYG{l+m+mi}{0}\PYG{p}{)} \PYG{k}{for} \PYG{n}{v\PYGZus{}t0}\PYG{p}{,} \PYG{n}{v\PYGZus{}x} \PYG{o+ow}{in} \PYG{n+nb}{zip}\PYG{p}{(}\PYG{n}{boundary\PYGZus{}tx}\PYG{p}{(}\PYG{n}{N\PYGZus{}SAMPLE\PYGZus{}POINTS\PYGZus{}BND}\PYG{p}{)}\PYG{p}{,} \PYG{n}{open\PYGZus{}boundary}\PYG{p}{(}\PYG{n}{N\PYGZus{}SAMPLE\PYGZus{}POINTS\PYGZus{}BND}\PYG{p}{)}\PYG{p}{)}\PYG{p}{]}
\PYG{n}{x\PYGZus{}bc}\PYG{p}{,} \PYG{n}{t\PYGZus{}bc}\PYG{p}{,} \PYG{n}{u\PYGZus{}bc} \PYG{o}{=} \PYG{n}{np}\PYG{o}{.}\PYG{n}{asarray}\PYG{p}{(}\PYG{n}{x\PYGZus{}bc}\PYG{p}{,}\PYG{n}{dtype}\PYG{o}{=}\PYG{n}{np}\PYG{o}{.}\PYG{n}{float32}\PYG{p}{)}\PYG{p}{,} \PYG{n}{np}\PYG{o}{.}\PYG{n}{asarray}\PYG{p}{(}\PYG{n}{t\PYGZus{}bc}\PYG{p}{,}\PYG{n}{dtype}\PYG{o}{=}\PYG{n}{np}\PYG{o}{.}\PYG{n}{float32}\PYG{p}{)} \PYG{p}{,}\PYG{n}{np}\PYG{o}{.}\PYG{n}{asarray}\PYG{p}{(}\PYG{n}{u\PYGZus{}bc}\PYG{p}{,}\PYG{n}{dtype}\PYG{o}{=}\PYG{n}{np}\PYG{o}{.}\PYG{n}{float32}\PYG{p}{)}
\PYG{c+c1}{\PYGZsh{}with app.model\PYGZus{}scope():}
\PYG{n}{loss\PYGZus{}u} \PYG{o}{=} \PYG{n}{math}\PYG{o}{.}\PYG{n}{l2\PYGZus{}loss}\PYG{p}{(}\PYG{n}{network}\PYG{p}{(}\PYG{n}{x\PYGZus{}bc}\PYG{p}{,} \PYG{n}{t\PYGZus{}bc}\PYG{p}{)}\PYG{p}{[}\PYG{p}{:}\PYG{p}{,} \PYG{l+m+mi}{0}\PYG{p}{]} \PYG{o}{\PYGZhy{}} \PYG{n}{u\PYGZus{}bc}\PYG{p}{)}  \PYG{c+c1}{\PYGZsh{} normalizes by first dimension, N\PYGZus{}bc}

\PYG{c+c1}{\PYGZsh{} Physics loss inside of domain}
\PYG{n}{N\PYGZus{}SAMPLE\PYGZus{}POINTS\PYGZus{}INNER} \PYG{o}{=} \PYG{l+m+mi}{1000}
\PYG{n}{x\PYGZus{}ph}\PYG{p}{,} \PYG{n}{t\PYGZus{}ph} \PYG{o}{=} \PYG{n}{tf}\PYG{o}{.}\PYG{n}{convert\PYGZus{}to\PYGZus{}tensor}\PYG{p}{(}\PYG{n}{rnd}\PYG{o}{.}\PYG{n}{random\PYGZus{}uniform}\PYG{p}{(}\PYG{p}{[}\PYG{n}{N\PYGZus{}SAMPLE\PYGZus{}POINTS\PYGZus{}INNER}\PYG{p}{]}\PYG{p}{,} \PYG{o}{\PYGZhy{}}\PYG{l+m+mi}{1}\PYG{p}{,} \PYG{l+m+mi}{1}\PYG{p}{)}\PYG{p}{)}\PYG{p}{,} \PYG{n}{tf}\PYG{o}{.}\PYG{n}{convert\PYGZus{}to\PYGZus{}tensor}\PYG{p}{(}\PYG{n}{rnd}\PYG{o}{.}\PYG{n}{random\PYGZus{}uniform}\PYG{p}{(}\PYG{p}{[}\PYG{n}{N\PYGZus{}SAMPLE\PYGZus{}POINTS\PYGZus{}INNER}\PYG{p}{]}\PYG{p}{,} \PYG{l+m+mi}{0}\PYG{p}{,} \PYG{l+m+mi}{1}\PYG{p}{)}\PYG{p}{)}
\PYG{n}{loss\PYGZus{}ph} \PYG{o}{=} \PYG{n}{math}\PYG{o}{.}\PYG{n}{l2\PYGZus{}loss}\PYG{p}{(}\PYG{n}{f}\PYG{p}{(}\PYG{n}{network}\PYG{p}{(}\PYG{n}{x\PYGZus{}ph}\PYG{p}{,} \PYG{n}{t\PYGZus{}ph}\PYG{p}{)}\PYG{p}{[}\PYG{p}{:}\PYG{p}{,} \PYG{l+m+mi}{0}\PYG{p}{]}\PYG{p}{,} \PYG{n}{x\PYGZus{}ph}\PYG{p}{,} \PYG{n}{t\PYGZus{}ph}\PYG{p}{)}\PYG{p}{)}  \PYG{c+c1}{\PYGZsh{} normalizes by first dimension, N\PYGZus{}ph}

\PYG{c+c1}{\PYGZsh{} Combine}
\PYG{n}{ph\PYGZus{}factor} \PYG{o}{=} \PYG{l+m+mf}{1.}
\PYG{n}{loss} \PYG{o}{=} \PYG{n}{loss\PYGZus{}u} \PYG{o}{+} \PYG{n}{ph\PYGZus{}factor} \PYG{o}{*} \PYG{n}{loss\PYGZus{}ph} \PYG{c+c1}{\PYGZsh{} allows us to control the relative influence of loss\PYGZus{}ph }

\PYG{n}{optim} \PYG{o}{=} \PYG{n}{tf}\PYG{o}{.}\PYG{n}{train}\PYG{o}{.}\PYG{n}{GradientDescentOptimizer}\PYG{p}{(}\PYG{n}{learning\PYGZus{}rate}\PYG{o}{=}\PYG{l+m+mf}{0.02}\PYG{p}{)}\PYG{o}{.}\PYG{n}{minimize}\PYG{p}{(}\PYG{n}{loss}\PYG{p}{)}
\PYG{c+c1}{\PYGZsh{}optim = tf.train.AdamOptimizer(learning\PYGZus{}rate=0.001).minimize(loss) \PYGZsh{} alternative, but not much benefit here}
\end{sphinxVerbatim}

The code above just initializes the evaluation of the loss, we still didn’t do any optimization steps, but we’re finally in a good position to get started with this.

Despite the simple equation, the convergence is typically very slow. The iterations themselves are fast to compute, but this setup needs a \sphinxstyleemphasis{lot} of iterations. To keep the runtime in a reasonable range, we only do 10k iterations by default below (\sphinxcode{\sphinxupquote{ITERS}}). You can increase this value to get better results.

\begin{sphinxVerbatim}[commandchars=\\\{\}]
\PYG{n}{session}\PYG{o}{.}\PYG{n}{initialize\PYGZus{}variables}\PYG{p}{(}\PYG{p}{)}

\PYG{k+kn}{import} \PYG{n+nn}{time}
\PYG{n}{start} \PYG{o}{=} \PYG{n}{time}\PYG{o}{.}\PYG{n}{time}\PYG{p}{(}\PYG{p}{)}

\PYG{n}{ITERS} \PYG{o}{=} \PYG{l+m+mi}{10000}
\PYG{k}{for} \PYG{n}{optim\PYGZus{}step} \PYG{o+ow}{in} \PYG{n+nb}{range}\PYG{p}{(}\PYG{n}{ITERS}\PYG{o}{+}\PYG{l+m+mi}{1}\PYG{p}{)}\PYG{p}{:}
  \PYG{n}{\PYGZus{}}\PYG{p}{,} \PYG{n}{loss\PYGZus{}value} \PYG{o}{=} \PYG{n}{session}\PYG{o}{.}\PYG{n}{run}\PYG{p}{(}\PYG{p}{[}\PYG{n}{optim}\PYG{p}{,} \PYG{n}{loss}\PYG{p}{]}\PYG{p}{)}
  \PYG{k}{if} \PYG{n}{optim\PYGZus{}step}\PYG{o}{\PYGZlt{}}\PYG{l+m+mi}{3} \PYG{o+ow}{or} \PYG{n}{optim\PYGZus{}step}\PYG{o}{\PYGZpc{}}\PYG{k}{1000}==0: 
        \PYG{n+nb}{print}\PYG{p}{(}\PYG{l+s+s1}{\PYGZsq{}}\PYG{l+s+s1}{Step }\PYG{l+s+si}{\PYGZpc{}d}\PYG{l+s+s1}{, loss: }\PYG{l+s+si}{\PYGZpc{}f}\PYG{l+s+s1}{\PYGZsq{}} \PYG{o}{\PYGZpc{}} \PYG{p}{(}\PYG{n}{optim\PYGZus{}step}\PYG{p}{,}\PYG{n}{loss\PYGZus{}value}\PYG{p}{)}\PYG{p}{)}
        \PYG{c+c1}{\PYGZsh{}show\PYGZus{}state(grid\PYGZus{}u)}
        
\PYG{n}{end} \PYG{o}{=} \PYG{n}{time}\PYG{o}{.}\PYG{n}{time}\PYG{p}{(}\PYG{p}{)}
\PYG{n+nb}{print}\PYG{p}{(}\PYG{l+s+s2}{\PYGZdq{}}\PYG{l+s+s2}{Runtime }\PYG{l+s+si}{\PYGZob{}:.2f\PYGZcb{}}\PYG{l+s+s2}{s}\PYG{l+s+s2}{\PYGZdq{}}\PYG{o}{.}\PYG{n}{format}\PYG{p}{(}\PYG{n}{end}\PYG{o}{\PYGZhy{}}\PYG{n}{start}\PYG{p}{)}\PYG{p}{)}
\end{sphinxVerbatim}

\begin{sphinxVerbatim}[commandchars=\\\{\}]
Step 0, loss: 0.108430
Step 1, loss: 0.103907
Step 2, loss: 0.100539
Step 1000, loss: 0.057177
Step 2000, loss: 0.053230
Step 3000, loss: 0.049326
Step 4000, loss: 0.046316
Step 5000, loss: 0.044153
Step 6000, loss: 0.041827
Step 7000, loss: 0.039517
Step 8000, loss: 0.037237
Step 9000, loss: 0.035256
Step 10000, loss: 0.033931
Runtime 110.15s
\end{sphinxVerbatim}

The training can take a significant amount of time, around 2 minutes on a typical notebook, but at least the error goes down significantly (roughly from around 0.2 to ca. 0.03), and the network seems to successfully converge to a solution.

Let’s show the reconstruction of the network, by evaluating the network at the centers of a regular grid, so that we can show the solution as an image. Note that this is actually fairly expensive, we have to run through the whole network with a few thousand weights for all of the \(128 \times 32\) sampling points in the grid.

It looks pretty good on first sight, though. There’s been a very noticeable change compared to the random initialization shown above:

\begin{sphinxVerbatim}[commandchars=\\\{\}]
\PYG{n}{show\PYGZus{}state}\PYG{p}{(}\PYG{n}{session}\PYG{o}{.}\PYG{n}{run}\PYG{p}{(}\PYG{n}{grid\PYGZus{}u}\PYG{p}{)}\PYG{p}{,}\PYG{l+s+s2}{\PYGZdq{}}\PYG{l+s+s2}{After Training}\PYG{l+s+s2}{\PYGZdq{}}\PYG{p}{)} 
\end{sphinxVerbatim}

\noindent\sphinxincludegraphics{{physicalloss-code_18_0}.png}

\bigskip\hrule\bigskip

\section{Evaluation}
\label{\detokenize{physicalloss-code:evaluation}}
Let’s compare solution in a bit more detail. Here are the actual sample points used for constraining the solution (at time step 16, \(t=1/2\)) shown in gray, versus the reconstructed solution in blue:

\begin{sphinxVerbatim}[commandchars=\\\{\}]
\PYG{n}{u} \PYG{o}{=} \PYG{n}{session}\PYG{o}{.}\PYG{n}{run}\PYG{p}{(}\PYG{n}{grid\PYGZus{}u}\PYG{p}{)}

\PYG{c+c1}{\PYGZsh{} solution is imposed at t=1/2 , which is 16 in the array}
\PYG{n}{BC\PYGZus{}TX} \PYG{o}{=} \PYG{l+m+mi}{16} 
\PYG{n}{uT} \PYG{o}{=} \PYG{n}{u}\PYG{p}{[}\PYG{l+m+mi}{0}\PYG{p}{,}\PYG{p}{:}\PYG{p}{,}\PYG{n}{BC\PYGZus{}TX}\PYG{p}{,}\PYG{l+m+mi}{0}\PYG{p}{]}

\PYG{n}{fig} \PYG{o}{=} \PYG{n}{plt}\PYG{o}{.}\PYG{n}{figure}\PYG{p}{(}\PYG{p}{)}\PYG{o}{.}\PYG{n}{gca}\PYG{p}{(}\PYG{p}{)}
\PYG{n}{fig}\PYG{o}{.}\PYG{n}{plot}\PYG{p}{(}\PYG{n}{np}\PYG{o}{.}\PYG{n}{linspace}\PYG{p}{(}\PYG{o}{\PYGZhy{}}\PYG{l+m+mi}{1}\PYG{p}{,}\PYG{l+m+mi}{1}\PYG{p}{,}\PYG{n+nb}{len}\PYG{p}{(}\PYG{n}{uT}\PYG{p}{)}\PYG{p}{)}\PYG{p}{,} \PYG{n}{uT}\PYG{p}{,} \PYG{n}{lw}\PYG{o}{=}\PYG{l+m+mi}{2}\PYG{p}{,} \PYG{n}{color}\PYG{o}{=}\PYG{l+s+s1}{\PYGZsq{}}\PYG{l+s+s1}{blue}\PYG{l+s+s1}{\PYGZsq{}}\PYG{p}{,} \PYG{n}{label}\PYG{o}{=}\PYG{l+s+s2}{\PYGZdq{}}\PYG{l+s+s2}{NN}\PYG{l+s+s2}{\PYGZdq{}}\PYG{p}{)}
\PYG{n}{fig}\PYG{o}{.}\PYG{n}{scatter}\PYG{p}{(}\PYG{n}{x\PYGZus{}bc}\PYG{p}{[}\PYG{l+m+mi}{0}\PYG{p}{:}\PYG{l+m+mi}{100}\PYG{p}{]}\PYG{p}{,} \PYG{n}{u\PYGZus{}bc}\PYG{p}{[}\PYG{l+m+mi}{0}\PYG{p}{:}\PYG{l+m+mi}{100}\PYG{p}{]}\PYG{p}{,} \PYG{n}{color}\PYG{o}{=}\PYG{l+s+s1}{\PYGZsq{}}\PYG{l+s+s1}{gray}\PYG{l+s+s1}{\PYGZsq{}}\PYG{p}{,} \PYG{n}{label}\PYG{o}{=}\PYG{l+s+s2}{\PYGZdq{}}\PYG{l+s+s2}{Reference}\PYG{l+s+s2}{\PYGZdq{}}\PYG{p}{)}
\PYG{n}{plt}\PYG{o}{.}\PYG{n}{title}\PYG{p}{(}\PYG{l+s+s2}{\PYGZdq{}}\PYG{l+s+s2}{Comparison at t=1/2}\PYG{l+s+s2}{\PYGZdq{}}\PYG{p}{)}
\PYG{n}{plt}\PYG{o}{.}\PYG{n}{xlabel}\PYG{p}{(}\PYG{l+s+s1}{\PYGZsq{}}\PYG{l+s+s1}{x}\PYG{l+s+s1}{\PYGZsq{}}\PYG{p}{)}\PYG{p}{;} \PYG{n}{plt}\PYG{o}{.}\PYG{n}{ylabel}\PYG{p}{(}\PYG{l+s+s1}{\PYGZsq{}}\PYG{l+s+s1}{u}\PYG{l+s+s1}{\PYGZsq{}}\PYG{p}{)}\PYG{p}{;} \PYG{n}{plt}\PYG{o}{.}\PYG{n}{legend}\PYG{p}{(}\PYG{p}{)}
\end{sphinxVerbatim}

\begin{sphinxVerbatim}[commandchars=\\\{\}]
\PYGZlt{}matplotlib.legend.Legend at 0x7f84721c3fd0\PYGZgt{}
\end{sphinxVerbatim}

\noindent\sphinxincludegraphics{{physicalloss-code_20_1}.png}

Not too bad at the sides of the domain (the Dirichlet boundary conditions \(u=0\) are fulfilled), but the shock in the center (at \(x=0\)) is not well represented.

Let’s check how well the initial state at \(t=0\) was reconstructed. That’s the most interesting, and toughest part of the problem (the rest basically follows from the model equation and boundary conditions, given the first state).

It turns out that the accuracy of the initial state is actually not that good: the blue curve from the PINN is quite far away from the constraints via the reference data (shown in gray)… The solution will get better with larger number of iterations, but it requires a surprisingly large number of iterations for this fairly simple case.

\begin{sphinxVerbatim}[commandchars=\\\{\}]
\PYG{c+c1}{\PYGZsh{} ground truth solution at t0}
\PYG{n}{t0gt} \PYG{o}{=} \PYG{n}{np}\PYG{o}{.}\PYG{n}{asarray}\PYG{p}{(} \PYG{p}{[} \PYG{p}{[}\PYG{o}{\PYGZhy{}}\PYG{n}{math}\PYG{o}{.}\PYG{n}{sin}\PYG{p}{(}\PYG{n}{np}\PYG{o}{.}\PYG{n}{pi} \PYG{o}{*} \PYG{n}{x}\PYG{p}{)} \PYG{o}{*} \PYG{l+m+mf}{1.}\PYG{p}{]} \PYG{k}{for} \PYG{n}{x} \PYG{o+ow}{in} \PYG{n}{np}\PYG{o}{.}\PYG{n}{linspace}\PYG{p}{(}\PYG{o}{\PYGZhy{}}\PYG{l+m+mi}{1}\PYG{p}{,}\PYG{l+m+mi}{1}\PYG{p}{,}\PYG{n}{N}\PYG{p}{)}\PYG{p}{]} \PYG{p}{)}
\PYG{n}{velP0} \PYG{o}{=} \PYG{n}{u}\PYG{p}{[}\PYG{l+m+mi}{0}\PYG{p}{,}\PYG{p}{:}\PYG{p}{,}\PYG{l+m+mi}{0}\PYG{p}{,}\PYG{l+m+mi}{0}\PYG{p}{]}

\PYG{n}{fig} \PYG{o}{=} \PYG{n}{plt}\PYG{o}{.}\PYG{n}{figure}\PYG{p}{(}\PYG{p}{)}\PYG{o}{.}\PYG{n}{gca}\PYG{p}{(}\PYG{p}{)}
\PYG{n}{fig}\PYG{o}{.}\PYG{n}{plot}\PYG{p}{(}\PYG{n}{np}\PYG{o}{.}\PYG{n}{linspace}\PYG{p}{(}\PYG{o}{\PYGZhy{}}\PYG{l+m+mi}{1}\PYG{p}{,}\PYG{l+m+mi}{1}\PYG{p}{,}\PYG{n+nb}{len}\PYG{p}{(}\PYG{n}{velP0}\PYG{p}{)}\PYG{p}{)}\PYG{p}{,} \PYG{n}{velP0}\PYG{p}{,} \PYG{n}{lw}\PYG{o}{=}\PYG{l+m+mi}{2}\PYG{p}{,} \PYG{n}{color}\PYG{o}{=}\PYG{l+s+s1}{\PYGZsq{}}\PYG{l+s+s1}{blue}\PYG{l+s+s1}{\PYGZsq{}}\PYG{p}{,} \PYG{n}{label}\PYG{o}{=}\PYG{l+s+s2}{\PYGZdq{}}\PYG{l+s+s2}{NN}\PYG{l+s+s2}{\PYGZdq{}}\PYG{p}{)}
\PYG{n}{fig}\PYG{o}{.}\PYG{n}{plot}\PYG{p}{(}\PYG{n}{np}\PYG{o}{.}\PYG{n}{linspace}\PYG{p}{(}\PYG{o}{\PYGZhy{}}\PYG{l+m+mi}{1}\PYG{p}{,}\PYG{l+m+mi}{1}\PYG{p}{,}\PYG{n+nb}{len}\PYG{p}{(}\PYG{n}{t0gt}\PYG{p}{)}\PYG{p}{)}\PYG{p}{,} \PYG{n}{t0gt}\PYG{p}{,} \PYG{n}{lw}\PYG{o}{=}\PYG{l+m+mi}{2}\PYG{p}{,} \PYG{n}{color}\PYG{o}{=}\PYG{l+s+s1}{\PYGZsq{}}\PYG{l+s+s1}{gray}\PYG{l+s+s1}{\PYGZsq{}}\PYG{p}{,} \PYG{n}{label}\PYG{o}{=}\PYG{l+s+s2}{\PYGZdq{}}\PYG{l+s+s2}{Reference}\PYG{l+s+s2}{\PYGZdq{}}\PYG{p}{)} 
\PYG{n}{plt}\PYG{o}{.}\PYG{n}{title}\PYG{p}{(}\PYG{l+s+s2}{\PYGZdq{}}\PYG{l+s+s2}{Comparison at t=0}\PYG{l+s+s2}{\PYGZdq{}}\PYG{p}{)}
\PYG{n}{plt}\PYG{o}{.}\PYG{n}{xlabel}\PYG{p}{(}\PYG{l+s+s1}{\PYGZsq{}}\PYG{l+s+s1}{x}\PYG{l+s+s1}{\PYGZsq{}}\PYG{p}{)}\PYG{p}{;} \PYG{n}{plt}\PYG{o}{.}\PYG{n}{ylabel}\PYG{p}{(}\PYG{l+s+s1}{\PYGZsq{}}\PYG{l+s+s1}{u}\PYG{l+s+s1}{\PYGZsq{}}\PYG{p}{)}\PYG{p}{;} \PYG{n}{plt}\PYG{o}{.}\PYG{n}{legend}\PYG{p}{(}\PYG{p}{)}
\end{sphinxVerbatim}

\begin{sphinxVerbatim}[commandchars=\\\{\}]
\PYGZlt{}matplotlib.legend.Legend at 0x7f847223da60\PYGZgt{}
\end{sphinxVerbatim}

\noindent\sphinxincludegraphics{{physicalloss-code_22_1}.png}

Especially the maximum / minimum at \(x=\pm 1/2\) are far off, and the boundaries at \(x=\pm 1\) are not fulfilled: the solution is not at zero.

We have the forward simulator for this simulation, so we can use the \(t=0\) solution of the network to
evaluate how well the temporal evaluation was reconstructed. This measures how well the temporal evolution of the model equation was captured via the soft constraints of the PINN loss.

The graph below shows the initial state in blue, and two evolved states at \(t=8/32\) and \(t=15/32\). Note that this is all from the simulated version, we’ll show the learned version next.

(Note: The code segments below also have some optional code to show the states at \sphinxcode{\sphinxupquote{{[}STEPS//4{]}}}. It’s commented out by default, you can uncomment or add additional ones to visualize more of the time evolution if you like.)

\begin{sphinxVerbatim}[commandchars=\\\{\}]
\PYG{c+c1}{\PYGZsh{} re\PYGZhy{}simulate with phiflow from solution at t=0}
\PYG{n}{DT} \PYG{o}{=} \PYG{l+m+mf}{1.}\PYG{o}{/}\PYG{l+m+mf}{32.}
\PYG{n}{STEPS} \PYG{o}{=} \PYG{l+m+mi}{32}\PYG{o}{\PYGZhy{}}\PYG{n}{BC\PYGZus{}TX} \PYG{c+c1}{\PYGZsh{} depends on where BCs were imposed}
\PYG{n}{INITIAL} \PYG{o}{=} \PYG{n}{u}\PYG{p}{[}\PYG{o}{.}\PYG{o}{.}\PYG{o}{.}\PYG{p}{,}\PYG{n}{BC\PYGZus{}TX}\PYG{p}{:}\PYG{p}{(}\PYG{n}{BC\PYGZus{}TX}\PYG{o}{+}\PYG{l+m+mi}{1}\PYG{p}{)}\PYG{p}{,}\PYG{l+m+mi}{0}\PYG{p}{]} \PYG{c+c1}{\PYGZsh{} np.reshape(u0, [1,len(u0),1]) }
\PYG{n+nb}{print}\PYG{p}{(}\PYG{n}{INITIAL}\PYG{o}{.}\PYG{n}{shape}\PYG{p}{)}

\PYG{n}{DOMAIN} \PYG{o}{=} \PYG{n}{Domain}\PYG{p}{(}\PYG{p}{[}\PYG{n}{N}\PYG{p}{]}\PYG{p}{,} \PYG{n}{boundaries}\PYG{o}{=}\PYG{n}{PERIODIC}\PYG{p}{,} \PYG{n}{box}\PYG{o}{=}\PYG{n}{box}\PYG{p}{[}\PYG{o}{\PYGZhy{}}\PYG{l+m+mi}{1}\PYG{p}{:}\PYG{l+m+mi}{1}\PYG{p}{]}\PYG{p}{)}
\PYG{n}{state} \PYG{o}{=} \PYG{p}{[}\PYG{n}{BurgersVelocity}\PYG{p}{(}\PYG{n}{DOMAIN}\PYG{p}{,} \PYG{n}{velocity}\PYG{o}{=}\PYG{n}{INITIAL}\PYG{p}{,} \PYG{n}{viscosity}\PYG{o}{=}\PYG{l+m+mf}{0.01}\PYG{o}{/}\PYG{n}{np}\PYG{o}{.}\PYG{n}{pi}\PYG{p}{)}\PYG{p}{]}
\PYG{n}{physics} \PYG{o}{=} \PYG{n}{Burgers}\PYG{p}{(}\PYG{p}{)}

\PYG{k}{for} \PYG{n}{i} \PYG{o+ow}{in} \PYG{n+nb}{range}\PYG{p}{(}\PYG{n}{STEPS}\PYG{p}{)}\PYG{p}{:}
    \PYG{n}{state}\PYG{o}{.}\PYG{n}{append}\PYG{p}{(} \PYG{n}{physics}\PYG{o}{.}\PYG{n}{step}\PYG{p}{(}\PYG{n}{state}\PYG{p}{[}\PYG{o}{\PYGZhy{}}\PYG{l+m+mi}{1}\PYG{p}{]}\PYG{p}{,}\PYG{n}{dt}\PYG{o}{=}\PYG{n}{DT}\PYG{p}{)} \PYG{p}{)}

\PYG{c+c1}{\PYGZsh{} we only need \PYGZdq{}velocity.data\PYGZdq{} from each phiflow state}
\PYG{n}{vel\PYGZus{}resim} \PYG{o}{=} \PYG{p}{[}\PYG{n}{x}\PYG{o}{.}\PYG{n}{velocity}\PYG{o}{.}\PYG{n}{data} \PYG{k}{for} \PYG{n}{x} \PYG{o+ow}{in} \PYG{n}{state}\PYG{p}{]}

\PYG{n}{fig} \PYG{o}{=} \PYG{n}{plt}\PYG{o}{.}\PYG{n}{figure}\PYG{p}{(}\PYG{p}{)}\PYG{o}{.}\PYG{n}{gca}\PYG{p}{(}\PYG{p}{)}
\PYG{n}{pltx} \PYG{o}{=} \PYG{n}{np}\PYG{o}{.}\PYG{n}{linspace}\PYG{p}{(}\PYG{o}{\PYGZhy{}}\PYG{l+m+mi}{1}\PYG{p}{,}\PYG{l+m+mi}{1}\PYG{p}{,}\PYG{n+nb}{len}\PYG{p}{(}\PYG{n}{vel\PYGZus{}resim}\PYG{p}{[}\PYG{l+m+mi}{0}\PYG{p}{]}\PYG{o}{.}\PYG{n}{flatten}\PYG{p}{(}\PYG{p}{)}\PYG{p}{)}\PYG{p}{)}
\PYG{n}{fig}\PYG{o}{.}\PYG{n}{plot}\PYG{p}{(}\PYG{n}{pltx}\PYG{p}{,} \PYG{n}{vel\PYGZus{}resim}\PYG{p}{[} \PYG{l+m+mi}{0}\PYG{p}{]}\PYG{o}{.}\PYG{n}{flatten}\PYG{p}{(}\PYG{p}{)}\PYG{p}{,}       \PYG{n}{lw}\PYG{o}{=}\PYG{l+m+mi}{2}\PYG{p}{,} \PYG{n}{color}\PYG{o}{=}\PYG{l+s+s1}{\PYGZsq{}}\PYG{l+s+s1}{blue}\PYG{l+s+s1}{\PYGZsq{}}\PYG{p}{,}  \PYG{n}{label}\PYG{o}{=}\PYG{l+s+s2}{\PYGZdq{}}\PYG{l+s+s2}{t=0}\PYG{l+s+s2}{\PYGZdq{}}\PYG{p}{)}
\PYG{c+c1}{\PYGZsh{}fig.plot(pltx, vel\PYGZus{}resim[STEPS//4].flatten(), lw=2, color=\PYGZsq{}green\PYGZsq{}, label=\PYGZdq{}t=0.125\PYGZdq{})}
\PYG{n}{fig}\PYG{o}{.}\PYG{n}{plot}\PYG{p}{(}\PYG{n}{pltx}\PYG{p}{,} \PYG{n}{vel\PYGZus{}resim}\PYG{p}{[}\PYG{n}{STEPS}\PYG{o}{/}\PYG{o}{/}\PYG{l+m+mi}{2}\PYG{p}{]}\PYG{o}{.}\PYG{n}{flatten}\PYG{p}{(}\PYG{p}{)}\PYG{p}{,} \PYG{n}{lw}\PYG{o}{=}\PYG{l+m+mi}{2}\PYG{p}{,} \PYG{n}{color}\PYG{o}{=}\PYG{l+s+s1}{\PYGZsq{}}\PYG{l+s+s1}{cyan}\PYG{l+s+s1}{\PYGZsq{}}\PYG{p}{,}  \PYG{n}{label}\PYG{o}{=}\PYG{l+s+s2}{\PYGZdq{}}\PYG{l+s+s2}{t=0.25}\PYG{l+s+s2}{\PYGZdq{}}\PYG{p}{)}
\PYG{n}{fig}\PYG{o}{.}\PYG{n}{plot}\PYG{p}{(}\PYG{n}{pltx}\PYG{p}{,} \PYG{n}{vel\PYGZus{}resim}\PYG{p}{[}\PYG{n}{STEPS}\PYG{o}{\PYGZhy{}}\PYG{l+m+mi}{1}\PYG{p}{]}\PYG{o}{.}\PYG{n}{flatten}\PYG{p}{(}\PYG{p}{)}\PYG{p}{,}  \PYG{n}{lw}\PYG{o}{=}\PYG{l+m+mi}{2}\PYG{p}{,} \PYG{n}{color}\PYG{o}{=}\PYG{l+s+s1}{\PYGZsq{}}\PYG{l+s+s1}{purple}\PYG{l+s+s1}{\PYGZsq{}}\PYG{p}{,}\PYG{n}{label}\PYG{o}{=}\PYG{l+s+s2}{\PYGZdq{}}\PYG{l+s+s2}{t=0.5}\PYG{l+s+s2}{\PYGZdq{}}\PYG{p}{)}
\PYG{c+c1}{\PYGZsh{}fig.plot(pltx, t0gt, lw=2, color=\PYGZsq{}gray\PYGZsq{}, label=\PYGZdq{}t=0 Reference\PYGZdq{}) \PYGZsh{} optionally show GT, compare to blue}
\PYG{n}{plt}\PYG{o}{.}\PYG{n}{title}\PYG{p}{(}\PYG{l+s+s2}{\PYGZdq{}}\PYG{l+s+s2}{Resimulated u from solution at t=0}\PYG{l+s+s2}{\PYGZdq{}}\PYG{p}{)}
\PYG{n}{plt}\PYG{o}{.}\PYG{n}{xlabel}\PYG{p}{(}\PYG{l+s+s1}{\PYGZsq{}}\PYG{l+s+s1}{x}\PYG{l+s+s1}{\PYGZsq{}}\PYG{p}{)}\PYG{p}{;} \PYG{n}{plt}\PYG{o}{.}\PYG{n}{ylabel}\PYG{p}{(}\PYG{l+s+s1}{\PYGZsq{}}\PYG{l+s+s1}{u}\PYG{l+s+s1}{\PYGZsq{}}\PYG{p}{)}\PYG{p}{;} \PYG{n}{plt}\PYG{o}{.}\PYG{n}{legend}\PYG{p}{(}\PYG{p}{)}
\end{sphinxVerbatim}

\begin{sphinxVerbatim}[commandchars=\\\{\}]
(1, 128, 1)
\end{sphinxVerbatim}

\begin{sphinxVerbatim}[commandchars=\\\{\}]
\PYGZlt{}matplotlib.legend.Legend at 0x7f84722de760\PYGZgt{}
\end{sphinxVerbatim}

\noindent\sphinxincludegraphics{{physicalloss-code_24_2}.png}

And here is the PINN output from \sphinxcode{\sphinxupquote{u}} at the same time steps:

\begin{sphinxVerbatim}[commandchars=\\\{\}]
\PYG{n}{velP} \PYG{o}{=} \PYG{p}{[}\PYG{n}{u}\PYG{p}{[}\PYG{l+m+mi}{0}\PYG{p}{,}\PYG{p}{:}\PYG{p}{,}\PYG{n}{x}\PYG{p}{,}\PYG{l+m+mi}{0}\PYG{p}{]} \PYG{k}{for} \PYG{n}{x} \PYG{o+ow}{in} \PYG{n+nb}{range}\PYG{p}{(}\PYG{l+m+mi}{33}\PYG{p}{)}\PYG{p}{]}
\PYG{n+nb}{print}\PYG{p}{(}\PYG{n}{velP}\PYG{p}{[}\PYG{l+m+mi}{0}\PYG{p}{]}\PYG{o}{.}\PYG{n}{shape}\PYG{p}{)}

\PYG{n}{fig} \PYG{o}{=} \PYG{n}{plt}\PYG{o}{.}\PYG{n}{figure}\PYG{p}{(}\PYG{p}{)}\PYG{o}{.}\PYG{n}{gca}\PYG{p}{(}\PYG{p}{)}
\PYG{n}{fig}\PYG{o}{.}\PYG{n}{plot}\PYG{p}{(}\PYG{n}{pltx}\PYG{p}{,} \PYG{n}{velP}\PYG{p}{[}\PYG{n}{BC\PYGZus{}TX}\PYG{o}{+} \PYG{l+m+mi}{0}\PYG{p}{]}\PYG{o}{.}\PYG{n}{flatten}\PYG{p}{(}\PYG{p}{)}\PYG{p}{,}       \PYG{n}{lw}\PYG{o}{=}\PYG{l+m+mi}{2}\PYG{p}{,} \PYG{n}{color}\PYG{o}{=}\PYG{l+s+s1}{\PYGZsq{}}\PYG{l+s+s1}{blue}\PYG{l+s+s1}{\PYGZsq{}}\PYG{p}{,}  \PYG{n}{label}\PYG{o}{=}\PYG{l+s+s2}{\PYGZdq{}}\PYG{l+s+s2}{t=0}\PYG{l+s+s2}{\PYGZdq{}}\PYG{p}{)}
\PYG{c+c1}{\PYGZsh{}fig.plot(pltx, velP[BC\PYGZus{}TX+STEPS//4].flatten(), lw=2, color=\PYGZsq{}green\PYGZsq{}, label=\PYGZdq{}t=0.125\PYGZdq{})}
\PYG{n}{fig}\PYG{o}{.}\PYG{n}{plot}\PYG{p}{(}\PYG{n}{pltx}\PYG{p}{,} \PYG{n}{velP}\PYG{p}{[}\PYG{n}{BC\PYGZus{}TX}\PYG{o}{+}\PYG{n}{STEPS}\PYG{o}{/}\PYG{o}{/}\PYG{l+m+mi}{2}\PYG{p}{]}\PYG{o}{.}\PYG{n}{flatten}\PYG{p}{(}\PYG{p}{)}\PYG{p}{,} \PYG{n}{lw}\PYG{o}{=}\PYG{l+m+mi}{2}\PYG{p}{,} \PYG{n}{color}\PYG{o}{=}\PYG{l+s+s1}{\PYGZsq{}}\PYG{l+s+s1}{cyan}\PYG{l+s+s1}{\PYGZsq{}}\PYG{p}{,}  \PYG{n}{label}\PYG{o}{=}\PYG{l+s+s2}{\PYGZdq{}}\PYG{l+s+s2}{t=0.25}\PYG{l+s+s2}{\PYGZdq{}}\PYG{p}{)}
\PYG{n}{fig}\PYG{o}{.}\PYG{n}{plot}\PYG{p}{(}\PYG{n}{pltx}\PYG{p}{,} \PYG{n}{velP}\PYG{p}{[}\PYG{n}{BC\PYGZus{}TX}\PYG{o}{+}\PYG{n}{STEPS}\PYG{o}{\PYGZhy{}}\PYG{l+m+mi}{1}\PYG{p}{]}\PYG{o}{.}\PYG{n}{flatten}\PYG{p}{(}\PYG{p}{)}\PYG{p}{,}  \PYG{n}{lw}\PYG{o}{=}\PYG{l+m+mi}{2}\PYG{p}{,} \PYG{n}{color}\PYG{o}{=}\PYG{l+s+s1}{\PYGZsq{}}\PYG{l+s+s1}{purple}\PYG{l+s+s1}{\PYGZsq{}}\PYG{p}{,}\PYG{n}{label}\PYG{o}{=}\PYG{l+s+s2}{\PYGZdq{}}\PYG{l+s+s2}{t=0.5}\PYG{l+s+s2}{\PYGZdq{}}\PYG{p}{)}
\PYG{n}{plt}\PYG{o}{.}\PYG{n}{title}\PYG{p}{(}\PYG{l+s+s2}{\PYGZdq{}}\PYG{l+s+s2}{NN Output}\PYG{l+s+s2}{\PYGZdq{}}\PYG{p}{)}
\PYG{n}{plt}\PYG{o}{.}\PYG{n}{xlabel}\PYG{p}{(}\PYG{l+s+s1}{\PYGZsq{}}\PYG{l+s+s1}{x}\PYG{l+s+s1}{\PYGZsq{}}\PYG{p}{)}\PYG{p}{;} \PYG{n}{plt}\PYG{o}{.}\PYG{n}{ylabel}\PYG{p}{(}\PYG{l+s+s1}{\PYGZsq{}}\PYG{l+s+s1}{u}\PYG{l+s+s1}{\PYGZsq{}}\PYG{p}{)}\PYG{p}{;} \PYG{n}{plt}\PYG{o}{.}\PYG{n}{legend}\PYG{p}{(}\PYG{p}{)}
\end{sphinxVerbatim}

\begin{sphinxVerbatim}[commandchars=\\\{\}]
(128,)
\end{sphinxVerbatim}

\begin{sphinxVerbatim}[commandchars=\\\{\}]
\PYGZlt{}matplotlib.legend.Legend at 0x7f8470cfb1f0\PYGZgt{}
\end{sphinxVerbatim}

\noindent\sphinxincludegraphics{{physicalloss-code_26_2}.png}

Judging via eyeball norm, these two versions of \(u\) look quite similar, but not surprisingly the errors grow over time and there are significant differences. Especially the steepening of the solution near the shock at \(x=0\) is not “captured” well. It’s a bit difficult to see in these two graphs, though, let’s quantify the error and show the actual difference:

\begin{sphinxVerbatim}[commandchars=\\\{\}]
\PYG{n}{error} \PYG{o}{=} \PYG{n}{np}\PYG{o}{.}\PYG{n}{sum}\PYG{p}{(} \PYG{n}{np}\PYG{o}{.}\PYG{n}{abs}\PYG{p}{(} \PYG{n}{np}\PYG{o}{.}\PYG{n}{asarray}\PYG{p}{(}\PYG{n}{vel\PYGZus{}resim}\PYG{p}{[}\PYG{l+m+mi}{0}\PYG{p}{:}\PYG{l+m+mi}{16}\PYG{p}{]}\PYG{p}{)}\PYG{o}{.}\PYG{n}{flatten}\PYG{p}{(}\PYG{p}{)} \PYG{o}{\PYGZhy{}} \PYG{n}{np}\PYG{o}{.}\PYG{n}{asarray}\PYG{p}{(}\PYG{n}{velP}\PYG{p}{[}\PYG{n}{BC\PYGZus{}TX}\PYG{p}{:}\PYG{n}{BC\PYGZus{}TX}\PYG{o}{+}\PYG{n}{STEPS}\PYG{p}{]}\PYG{p}{)}\PYG{o}{.}\PYG{n}{flatten}\PYG{p}{(}\PYG{p}{)} \PYG{p}{)}\PYG{p}{)} \PYG{o}{/} \PYG{p}{(}\PYG{n}{STEPS}\PYG{o}{*}\PYG{n}{N}\PYG{p}{)}
\PYG{n+nb}{print}\PYG{p}{(}\PYG{l+s+s2}{\PYGZdq{}}\PYG{l+s+s2}{Mean absolute error for re\PYGZhy{}simulation across }\PYG{l+s+si}{\PYGZob{}\PYGZcb{}}\PYG{l+s+s2}{ steps: }\PYG{l+s+si}{\PYGZob{}:7.5f\PYGZcb{}}\PYG{l+s+s2}{\PYGZdq{}}\PYG{o}{.}\PYG{n}{format}\PYG{p}{(}\PYG{n}{STEPS}\PYG{p}{,}\PYG{n}{error}\PYG{p}{)}\PYG{p}{)}

\PYG{n}{fig} \PYG{o}{=} \PYG{n}{plt}\PYG{o}{.}\PYG{n}{figure}\PYG{p}{(}\PYG{p}{)}\PYG{o}{.}\PYG{n}{gca}\PYG{p}{(}\PYG{p}{)}
\PYG{n}{fig}\PYG{o}{.}\PYG{n}{plot}\PYG{p}{(}\PYG{n}{pltx}\PYG{p}{,} \PYG{p}{(}\PYG{n}{vel\PYGZus{}resim}\PYG{p}{[}\PYG{l+m+mi}{0}       \PYG{p}{]}\PYG{o}{.}\PYG{n}{flatten}\PYG{p}{(}\PYG{p}{)}\PYG{o}{\PYGZhy{}}\PYG{n}{velP}\PYG{p}{[}\PYG{n}{BC\PYGZus{}TX}         \PYG{p}{]}\PYG{o}{.}\PYG{n}{flatten}\PYG{p}{(}\PYG{p}{)}\PYG{p}{)}\PYG{p}{,} \PYG{n}{lw}\PYG{o}{=}\PYG{l+m+mi}{2}\PYG{p}{,} \PYG{n}{color}\PYG{o}{=}\PYG{l+s+s1}{\PYGZsq{}}\PYG{l+s+s1}{blue}\PYG{l+s+s1}{\PYGZsq{}}\PYG{p}{,}  \PYG{n}{label}\PYG{o}{=}\PYG{l+s+s2}{\PYGZdq{}}\PYG{l+s+s2}{t=5}\PYG{l+s+s2}{\PYGZdq{}}\PYG{p}{)}
\PYG{n}{fig}\PYG{o}{.}\PYG{n}{plot}\PYG{p}{(}\PYG{n}{pltx}\PYG{p}{,} \PYG{p}{(}\PYG{n}{vel\PYGZus{}resim}\PYG{p}{[}\PYG{n}{STEPS}\PYG{o}{/}\PYG{o}{/}\PYG{l+m+mi}{4}\PYG{p}{]}\PYG{o}{.}\PYG{n}{flatten}\PYG{p}{(}\PYG{p}{)}\PYG{o}{\PYGZhy{}}\PYG{n}{velP}\PYG{p}{[}\PYG{n}{BC\PYGZus{}TX}\PYG{o}{+}\PYG{n}{STEPS}\PYG{o}{/}\PYG{o}{/}\PYG{l+m+mi}{4}\PYG{p}{]}\PYG{o}{.}\PYG{n}{flatten}\PYG{p}{(}\PYG{p}{)}\PYG{p}{)}\PYG{p}{,} \PYG{n}{lw}\PYG{o}{=}\PYG{l+m+mi}{2}\PYG{p}{,} \PYG{n}{color}\PYG{o}{=}\PYG{l+s+s1}{\PYGZsq{}}\PYG{l+s+s1}{green}\PYG{l+s+s1}{\PYGZsq{}}\PYG{p}{,} \PYG{n}{label}\PYG{o}{=}\PYG{l+s+s2}{\PYGZdq{}}\PYG{l+s+s2}{t=0.625}\PYG{l+s+s2}{\PYGZdq{}}\PYG{p}{)}
\PYG{n}{fig}\PYG{o}{.}\PYG{n}{plot}\PYG{p}{(}\PYG{n}{pltx}\PYG{p}{,} \PYG{p}{(}\PYG{n}{vel\PYGZus{}resim}\PYG{p}{[}\PYG{n}{STEPS}\PYG{o}{/}\PYG{o}{/}\PYG{l+m+mi}{2}\PYG{p}{]}\PYG{o}{.}\PYG{n}{flatten}\PYG{p}{(}\PYG{p}{)}\PYG{o}{\PYGZhy{}}\PYG{n}{velP}\PYG{p}{[}\PYG{n}{BC\PYGZus{}TX}\PYG{o}{+}\PYG{n}{STEPS}\PYG{o}{/}\PYG{o}{/}\PYG{l+m+mi}{2}\PYG{p}{]}\PYG{o}{.}\PYG{n}{flatten}\PYG{p}{(}\PYG{p}{)}\PYG{p}{)}\PYG{p}{,} \PYG{n}{lw}\PYG{o}{=}\PYG{l+m+mi}{2}\PYG{p}{,} \PYG{n}{color}\PYG{o}{=}\PYG{l+s+s1}{\PYGZsq{}}\PYG{l+s+s1}{cyan}\PYG{l+s+s1}{\PYGZsq{}}\PYG{p}{,}  \PYG{n}{label}\PYG{o}{=}\PYG{l+s+s2}{\PYGZdq{}}\PYG{l+s+s2}{t=0.75}\PYG{l+s+s2}{\PYGZdq{}}\PYG{p}{)}
\PYG{n}{fig}\PYG{o}{.}\PYG{n}{plot}\PYG{p}{(}\PYG{n}{pltx}\PYG{p}{,} \PYG{p}{(}\PYG{n}{vel\PYGZus{}resim}\PYG{p}{[}\PYG{n}{STEPS}\PYG{o}{\PYGZhy{}}\PYG{l+m+mi}{1} \PYG{p}{]}\PYG{o}{.}\PYG{n}{flatten}\PYG{p}{(}\PYG{p}{)}\PYG{o}{\PYGZhy{}}\PYG{n}{velP}\PYG{p}{[}\PYG{n}{BC\PYGZus{}TX}\PYG{o}{+}\PYG{n}{STEPS}\PYG{o}{\PYGZhy{}}\PYG{l+m+mi}{1} \PYG{p}{]}\PYG{o}{.}\PYG{n}{flatten}\PYG{p}{(}\PYG{p}{)}\PYG{p}{)}\PYG{p}{,} \PYG{n}{lw}\PYG{o}{=}\PYG{l+m+mi}{2}\PYG{p}{,} \PYG{n}{color}\PYG{o}{=}\PYG{l+s+s1}{\PYGZsq{}}\PYG{l+s+s1}{purple}\PYG{l+s+s1}{\PYGZsq{}}\PYG{p}{,}\PYG{n}{label}\PYG{o}{=}\PYG{l+s+s2}{\PYGZdq{}}\PYG{l+s+s2}{t=1}\PYG{l+s+s2}{\PYGZdq{}}\PYG{p}{)}
\PYG{n}{plt}\PYG{o}{.}\PYG{n}{title}\PYG{p}{(}\PYG{l+s+s2}{\PYGZdq{}}\PYG{l+s+s2}{u Error}\PYG{l+s+s2}{\PYGZdq{}}\PYG{p}{)}
\PYG{n}{plt}\PYG{o}{.}\PYG{n}{xlabel}\PYG{p}{(}\PYG{l+s+s1}{\PYGZsq{}}\PYG{l+s+s1}{x}\PYG{l+s+s1}{\PYGZsq{}}\PYG{p}{)}\PYG{p}{;} \PYG{n}{plt}\PYG{o}{.}\PYG{n}{ylabel}\PYG{p}{(}\PYG{l+s+s1}{\PYGZsq{}}\PYG{l+s+s1}{MAE}\PYG{l+s+s1}{\PYGZsq{}}\PYG{p}{)}\PYG{p}{;} \PYG{n}{plt}\PYG{o}{.}\PYG{n}{legend}\PYG{p}{(}\PYG{p}{)}
\end{sphinxVerbatim}

\begin{sphinxVerbatim}[commandchars=\\\{\}]
Mean absolute error for re\PYGZhy{}simulation across 16 steps: 0.01067
\end{sphinxVerbatim}

\begin{sphinxVerbatim}[commandchars=\\\{\}]
\PYGZlt{}matplotlib.legend.Legend at 0x7f8472550e50\PYGZgt{}
\end{sphinxVerbatim}

\noindent\sphinxincludegraphics{{physicalloss-code_28_2}.png}

The code above will compute a mean absolute error of ca. \(1.5 \cdot 10^{-2}\) between ground truth re\sphinxhyphen{}simulation and the PINN evolution, which is significant for the value range of the simulation.

And for comparison with the forward simulation and following cases, here are also all steps over time with a color map.

\begin{sphinxVerbatim}[commandchars=\\\{\}]
\PYG{c+c1}{\PYGZsh{} show re\PYGZhy{}simulated solution again as full image over time}
\PYG{n}{sn} \PYG{o}{=} \PYG{n}{np}\PYG{o}{.}\PYG{n}{concatenate}\PYG{p}{(}\PYG{n}{vel\PYGZus{}resim}\PYG{p}{,} \PYG{n}{axis}\PYG{o}{=}\PYG{o}{\PYGZhy{}}\PYG{l+m+mi}{1}\PYG{p}{)}
\PYG{n}{sn} \PYG{o}{=} \PYG{n}{np}\PYG{o}{.}\PYG{n}{reshape}\PYG{p}{(}\PYG{n}{sn}\PYG{p}{,} \PYG{n+nb}{list}\PYG{p}{(}\PYG{n}{sn}\PYG{o}{.}\PYG{n}{shape}\PYG{p}{)}\PYG{o}{+}\PYG{p}{[}\PYG{l+m+mi}{1}\PYG{p}{]} \PYG{p}{)} \PYG{c+c1}{\PYGZsh{} print(sn.shape)}
\PYG{n}{show\PYGZus{}state}\PYG{p}{(}\PYG{n}{sn}\PYG{p}{,}\PYG{l+s+s2}{\PYGZdq{}}\PYG{l+s+s2}{Re\PYGZhy{}simulated u}\PYG{l+s+s2}{\PYGZdq{}}\PYG{p}{)}
\end{sphinxVerbatim}

\noindent\sphinxincludegraphics{{physicalloss-code_30_0}.png}

Next, we’ll store the full solution over the course of the \(t=0 \dots 1\) time interval, so that we can compare it later on to the full solution from a regular forward solve and compare it to the differential physics solution.

Thus, stay tuned for the full evaluation and the comparison. This will follow in {\hyperref[\detokenize{diffphys-code-burgers::doc}]{\sphinxcrossref{\DUrole{doc}{Burgers Optimization with a Differentiable Physics Gradient}}}}, after we’ve discussed the details of how to run the differential physics optimization.

\begin{sphinxVerbatim}[commandchars=\\\{\}]
\PYG{n}{vels} \PYG{o}{=} \PYG{n}{session}\PYG{o}{.}\PYG{n}{run}\PYG{p}{(}\PYG{n}{grid\PYGZus{}u}\PYG{p}{)} \PYG{c+c1}{\PYGZsh{} special for showing NN results, run through TF }
\PYG{n}{vels} \PYG{o}{=} \PYG{n}{np}\PYG{o}{.}\PYG{n}{reshape}\PYG{p}{(} \PYG{n}{vels}\PYG{p}{,} \PYG{p}{[}\PYG{n}{vels}\PYG{o}{.}\PYG{n}{shape}\PYG{p}{[}\PYG{l+m+mi}{1}\PYG{p}{]}\PYG{p}{,}\PYG{n}{vels}\PYG{o}{.}\PYG{n}{shape}\PYG{p}{[}\PYG{l+m+mi}{2}\PYG{p}{]}\PYG{p}{]} \PYG{p}{)}

\PYG{c+c1}{\PYGZsh{} save for comparison with other methods}
\PYG{k+kn}{import} \PYG{n+nn}{os}\PYG{p}{;} \PYG{n}{os}\PYG{o}{.}\PYG{n}{makedirs}\PYG{p}{(}\PYG{l+s+s2}{\PYGZdq{}}\PYG{l+s+s2}{./temp}\PYG{l+s+s2}{\PYGZdq{}}\PYG{p}{,}\PYG{n}{exist\PYGZus{}ok}\PYG{o}{=}\PYG{k+kc}{True}\PYG{p}{)}
\PYG{n}{np}\PYG{o}{.}\PYG{n}{savez\PYGZus{}compressed}\PYG{p}{(}\PYG{l+s+s2}{\PYGZdq{}}\PYG{l+s+s2}{./temp/burgers\PYGZhy{}pinn\PYGZhy{}solution.npz}\PYG{l+s+s2}{\PYGZdq{}}\PYG{p}{,}\PYG{n}{vels}\PYG{p}{)} \PYG{p}{;} \PYG{n+nb}{print}\PYG{p}{(}\PYG{l+s+s2}{\PYGZdq{}}\PYG{l+s+s2}{Vels array shape: }\PYG{l+s+s2}{\PYGZdq{}}\PYG{o}{+}\PYG{n+nb}{format}\PYG{p}{(}\PYG{n}{vels}\PYG{o}{.}\PYG{n}{shape}\PYG{p}{)}\PYG{p}{)}
\end{sphinxVerbatim}

\begin{sphinxVerbatim}[commandchars=\\\{\}]
Vels array shape: (128, 33)
\end{sphinxVerbatim}

\bigskip\hrule\bigskip

\section{Next steps}
\label{\detokenize{physicalloss-code:next-steps}}
This setup is just a starting point for PINNs and physical soft\sphinxhyphen{}constraints, of course. The parameters of the setup were chosen to run relatively quickly. As we’ll show in the next sections, the behavior of such an inverse solve can be improved substantially by a tighter integration of solver and learning.

The solution of the PINN setup above can also directly be improved, however. E.g., try to:
\begin{itemize}
\item {} 
Adjust parameters of the training to further decrease the error without making the solution diverge.

\item {} 
Adapt the NN architecture for further improvements (keep track of the weight count, though).

\item {} 
Activate a different optimizer, and observe the change in behavior (this typically requires adjusting the learning rate). Note that the more complex optimizers don’t necessarily do better in this relatively simple example.

\item {} 
Or modify the setup to make the test case more interesting: e.g., move the boundary conditions to a later point in simulation time, to give the reconstruction a larger time interval to reconstruct.

\end{itemize}

\chapter{Discussion of Physical Soft\sphinxhyphen{}Constraints}
\label{\detokenize{physicalloss-discuss:discussion-of-physical-soft-constraints}}\label{\detokenize{physicalloss-discuss::doc}}
The good news so far is \sphinxhyphen{} we have a DL method that can include
physical laws in the form of soft constraints by minimizing residuals.
However, as the very simple previous example illustrates, this is just a conceptual
starting point.

On the positive side, we can leverage DL frameworks with backpropagation to compute
the derivatives of the model. At the same time, this puts us at the mercy of the learned
representation regarding the reliability of these derivatives. Also, each derivative
requires backpropagation through the full network, which can be very expensive. Especially so
for higher\sphinxhyphen{}order derivatives.

And while the setup is relatively simple, it is generally difficult to control. The NN
has flexibility to refine the solution by itself, but at the same time, tricks are necessary
when it doesn’t focus on the right regions of the solution.

\section{Is it “Machine Learning”?}
\label{\detokenize{physicalloss-discuss:is-it-machine-learning}}
One question that might also come to mind at this point is: \sphinxstyleemphasis{can we really call it machine learning}?
Of course, such denomination questions are superficial \sphinxhyphen{} if an algorithm is useful, it doesn’t matter
what name it has. However, here the question helps to highlight some important properties
that are typically associated with algorithms from fields like machine learning or optimization.

One main reason \sphinxstyleemphasis{not} to call these physical constraints machine learning (ML), is that the
positions where we test and constrain the solution are the final positions we are interested in.
As such, there is no real distinction between training, validation and (out of distribution) test sets.
Computing the solution for a known and given set of samples is much more akin to classical optimization,
where inverse problems like the previous Burgers example stem from.

For machine learning, we typically work under the assumption that the final performance of our
model will be evaluated on a different, potentially unknown set of inputs. The \sphinxstyleemphasis{test data}
should usually capture such out of distribution (OOD) behavior, so that we can make estimates
about how well our model will generalize to “real\sphinxhyphen{}world” cases that we will encounter when
we deploy it into an application.

In contrast, for the PINN training as described here, we reconstruct a single solution in a known
and given space\sphinxhyphen{}time region. As such, any samples from this domain follow the same distribution
and hence don’t really represent test or OOD sampes. As the NN directly encodes the solution,
there is also little hope that it will yield different solutions, or perform well outside
of the training distribution. If we’re interested in a different solution, we most likely
have to start training the NN from scratch.

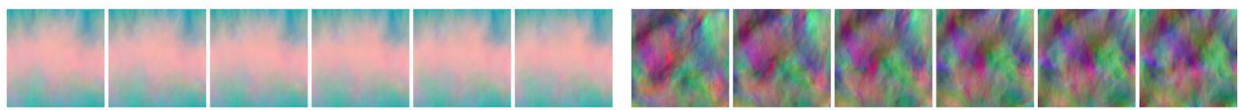
\sphinxincludegraphics{{divider5}.jpg}

\section{Summary}
\label{\detokenize{physicalloss-discuss:summary}}
Thus, the physical soft constraints allow us to encode solutions to
PDEs with the tools of NNs.
An inherent drawback of this approach is that it yields single solutions,
and that it does not combine with traditional numerical techniques well.
E.g., the learned representation is not suitable to be refined with
a classical iterative solver such as the conjugate gradient method.

This means many
powerful techniques that were developed in the past decades cannot be used in this context.
Bringing these numerical methods back into the picture will be one of the central
goals of the next sections.

✅ Pro:
\begin{itemize}
\item {} 
Uses physical model.

\item {} 
Derivatives can be conveniently computed via backpropagation.

\end{itemize}

❌ Con:
\begin{itemize}
\item {} 
Quite slow …

\item {} 
Physical constraints are enforced only as soft constraints.

\item {} 
Largely incompatible with \sphinxstyleemphasis{classical} numerical methods.

\item {} 
Accuracy of derivatives relies on learned representation.

\end{itemize}

Next, let’s look at how we can leverage numerical methods to improve the DL accuracy and efficiency
by making use of differentiable solvers.

\part{Differentiable Physics}

\chapter{Introduction to Differentiable Physics}
\label{\detokenize{diffphys:introduction-to-differentiable-physics}}\label{\detokenize{diffphys::doc}}
As a next step towards a tighter and more generic combination of deep learning
methods and physical simulations we will target incorporating \sphinxstyleemphasis{differentiable
simulations} into the learning process. In the following, we’ll shorten
that to “differentiable physics” (DP).

The central goal of these methods is to use existing numerical solvers, and equip
them with functionality to compute gradients with respect to their inputs.
Once this is realized for all operators of a simulation, we can leverage
the autodiff functionality of DL frameworks with backpropagation to let gradient
information flow from from a simulator into an NN and vice versa. This has numerous
advantages such as improved learning feedback and generalization, as we’ll outline below.

In contrast to physics\sphinxhyphen{}informed loss functions, it also enables handling more complex
solution manifolds instead of single inverse problems.
E.g., instead of using deep learning
to solve single inverse problems as in the previous chapter,
differentiable physics can be used to train NNs that learn to solve
larger classes of inverse problems very efficiently.

\begin{figure}[htbp]
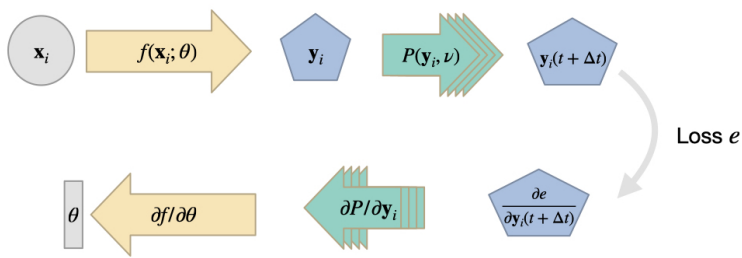

\centering
\capstart

\noindent\sphinxincludegraphics[height=220\sphinxpxdimen]{{diffphys-shortened}.jpg}
\caption{Training with differentiable physics means that a chain of differentiable operators
provide directions in the form of gradients to steer the learning process.}\label{\detokenize{diffphys:diffphys-short-overview}}\end{figure}

\section{Differentiable operators}
\label{\detokenize{diffphys:differentiable-operators}}
With the DP direction we build on existing numerical solvers. I.e.,
the approach is strongly relying on the algorithms developed in the larger field
of computational methods for a vast range of physical effects in our world.
To start with, we need a continuous formulation as model for the physical effect that we’d like
to simulate – if this is missing we’re in trouble. But luckily, we can
tap into existing collections of model equations and established methods
for discretizing continuous models.

Let’s assume we have a continuous formulation \(\mathcal P^*(\mathbf{x}, \nu)\) of the physical quantity of
interest \(\mathbf{u}(\mathbf{x}, t): \mathbb R^d \times \mathbb R^+ \rightarrow \mathbb R^d\),
with model parameters \(\nu\) (e.g., diffusion, viscosity, or conductivity constants).
The component of \(\mathbf{u}\) will be denoted by a numbered subscript, i.e.,
\(\mathbf{u} = (u_1,u_2,\dots,u_d)^T\).

Typically, we are interested in the temporal evolution of such a system.
Discretization yields a formulation \(\mathcal P(\mathbf{x}, \nu)\)
that we can re\sphinxhyphen{}arrange to compute a future state after a time step \(\Delta t\).
The state at \(t+\Delta t\) is computed via sequence of
operations \(\mathcal P_1, \mathcal P_2 \dots \mathcal P_m\) such that
\(\mathbf{u}(t+\Delta t) = \mathcal P_1 \circ \mathcal P_2 \circ \dots \mathcal P_m ( \mathbf{u}(t),\nu )\),
where \(\circ\) denotes function decomposition, i.e. \(f(g(x)) = f \circ g(x)\).

\begin{sphinxadmonition}{note}{Note:}
In order to integrate this solver into a DL process, we need to ensure that every operator
\(\mathcal P_i\) provides a gradient w.r.t. its inputs, i.e. in the example above
\(\partial \mathcal P_i / \partial \mathbf{u}\).
\end{sphinxadmonition}

Note that we typically don’t need derivatives
for all parameters of \(\mathcal P(\mathbf{x}, \nu)\), e.g.,
we omit \(\nu\) in the following, assuming that this is a
given model parameter with which the NN should not interact.
Naturally, it can vary within the solution manifold that we’re interested in,
but \(\nu\) will not be the output of an NN representation. If this is the case, we can omit
providing \(\partial \mathcal P_i / \partial \nu\) in our solver. However, the following learning process
naturally transfers to including \(\nu\) as a degree of freedom.

\section{Jacobians}
\label{\detokenize{diffphys:jacobians}}
As \(\mathbf{u}\) is typically a vector\sphinxhyphen{}valued function, \(\partial \mathcal P_i / \partial \mathbf{u}\) denotes
a Jacobian matrix \(J\) rather than a single value:
\begin{equation*}
\begin{split} \begin{aligned}
    \frac{ \partial \mathcal P_i }{ \partial \mathbf{u} } = 
    \begin{bmatrix} 
    \partial \mathcal P_{i,1} / \partial u_{1} 
    & \  \cdots \ &
    \partial \mathcal P_{i,1} / \partial u_{d} 
    \\
    \vdots & \ & \ 
    \\
    \partial \mathcal P_{i,d} / \partial u_{1} 
    & \  \cdots \ &
    \partial \mathcal P_{i,d} / \partial u_{d} 
    \end{bmatrix} 
\end{aligned} \end{split}
\end{equation*}
where, as above, \(d\) denotes the number of components in \(\mathbf{u}\). As \(\mathcal P\) maps one value of
\(\mathbf{u}\) to another, the Jacobian is square and symmetric here. Of course this isn’t necessarily the case
for general model equations, but non\sphinxhyphen{}square Jacobian matrices would not cause any problems for differentiable
simulations.

In practice, we can rely on the \sphinxstyleemphasis{reverse mode} differentiation that all modern DL
frameworks provide, and focus on computing a matrix vector product of the Jacobian transpose
with a vector \(\mathbf{a}\), i.e. the expression:
\(
    ( \frac{\partial \mathcal P_i }{ \partial \mathbf{u} } )^T \mathbf{a}
\).
If we’d need to construct and store all full Jacobian matrices that we encounter during training,
this would cause huge memory overheads and unnecessarily slow down training.
Instead, for backpropagation, we can provide faster operations that compute products
with the Jacobian transpose because we always have a scalar loss function at the end of the chain.

Given the formulation above, we need to resolve the derivatives
of the chain of function compositions of the \(\mathcal P_i\) at some current state \(\mathbf{u}^n\) via the chain rule.
E.g., for two of them
\begin{equation*}
\begin{split}
    \frac{ \partial (\mathcal P_1 \circ \mathcal P_2) }{ \partial \mathbf{u} }|_{\mathbf{u}^n}
    = 
    \frac{ \partial \mathcal P_1 }{ \partial \mathbf{u} }|_{\mathcal P_2(\mathbf{u}^n)}
    \ 
    \frac{ \partial \mathcal P_2 }{ \partial \mathbf{u} }|_{\mathbf{u}^n} \ , 
\end{split}
\end{equation*}
which is just the vector valued version of the “classic” chain rule
\(f(g(x))' = f'(g(x)) g'(x)\), and directly extends for larger numbers of composited functions, i.e. \(i>2\).

Here, the derivatives for \(\mathcal P_1\) and \(\mathcal P_2\) are still Jacobian matrices, but knowing that
at the “end” of the chain we have our scalar loss (cf. {\hyperref[\detokenize{overview::doc}]{\sphinxcrossref{\DUrole{doc}{Overview}}}}), the right\sphinxhyphen{}most Jacobian will invariably
be a matrix with 1 column, i.e. a vector. During reverse mode, we start with this vector, and compute
the multiplications with the left Jacobians, \(\frac{ \partial \mathcal P_1 }{ \partial \mathbf{u} }\) above,
one by one.

For the details of forward and reverse mode differentiation, please check out external materials such
as this \sphinxhref{https://arxiv.org/pdf/1502.05767.pdf}{nice survey by Baydin et al.}.

\section{Learning via DP operators}
\label{\detokenize{diffphys:learning-via-dp-operators}}
Thus, once the operators of our simulator support computations of the Jacobian\sphinxhyphen{}vector
products, we can integrate them into DL pipelines just like you would include a regular fully\sphinxhyphen{}connected layer
or a ReLU activation.

At this point, the following (very valid) question arises: “\sphinxstyleemphasis{Most physics solvers can be broken down into a
sequence of vector and matrix operations. All state\sphinxhyphen{}of\sphinxhyphen{}the\sphinxhyphen{}art DL frameworks support these, so why don’t we just
use these operators to realize our physics solver?}”

It’s true that this would theoretically be possible. The problem here is that each of the vector and matrix
operations in tensorflow and pytorch is computed individually, and internally needs to store the current
state of the forward evaluation for backpropagation (the “\(g(x)\)” above). For a typical
simulation, however, we’re not overly interested in every single intermediate result our solver produces.
Typically, we’re more concerned with significant updates such as the step from \(\mathbf{u}(t)\) to  \(\mathbf{u}(t+\Delta t)\).

Thus, in practice it is a very good idea to break down the solving process into a sequence
of meaningful but \sphinxstyleemphasis{monolithic} operators. This not only saves a lot of work by preventing the calculation
of unnecessary intermediate results, it also allows us to choose the best possible numerical methods
to compute the updates (and derivatives) for these operators.
E.g., as this process is very similar to adjoint method optimizations, we can re\sphinxhyphen{}use many of the techniques
that were developed in this field, or leverage established numerical methods. E.g.,
we could leverage the \(O(n)\) runtime of multigrid solvers for matrix inversion.

The flip\sphinxhyphen{}side of this approach is that it requires some understanding of the problem at hand,
and of the numerical methods. Also, a given solver might not provide gradient calculations out of the box.
Thus, if we want to employ DL for model equations that we don’t have a proper grasp of, it might not be a good
idea to directly go for learning via a DP approach. However, if we don’t really understand our model, we probably
should go back to studying it a bit more anyway…

Also, in practice we can be \sphinxstyleemphasis{greedy} with the derivative operators, and only
provide those which are relevant for the learning task. E.g., if our network
never produces the parameter \(\nu\) in the example above, and it doesn’t appear in our
loss formulation, we will never encounter a \(\partial/\partial \nu\) derivative
in our backpropagation step.

The following figure summarizes the DP\sphinxhyphen{}based learning approach, and illustrates the sequence of operations that are typically processed within a single PDE solve. As many of the operations are non\sphinxhyphen{}linear in practice, this often leads to a challenging learning task for the NN:

\begin{figure}[htbp]
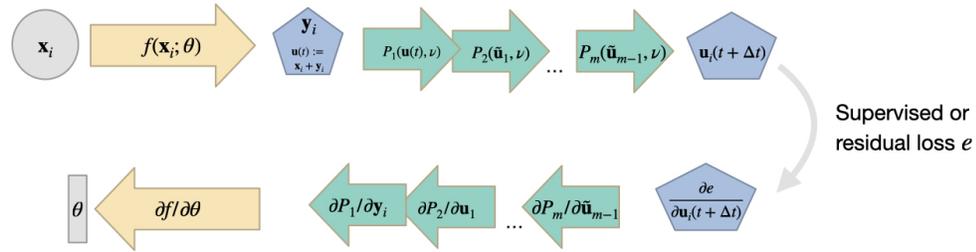

\centering
\capstart

\noindent\sphinxincludegraphics[height=220\sphinxpxdimen]{{diffphys-overview}.jpg}
\caption{DP learning with a PDE solver that consists of \(m\) individual operators \(\mathcal P_i\). The gradient travels backward through all \(m\) operators before influencing the network weights \(\theta\).}\label{\detokenize{diffphys:diffphys-full-overview}}\end{figure}

\bigskip\hrule\bigskip

\section{A practical example}
\label{\detokenize{diffphys:a-practical-example}}
As a simple example let’s consider the advection of a passive scalar density \(d(\mathbf{x},t)\) in
a velocity field \(\mathbf{u}\) as physical model \(\mathcal P^*\):
\begin{equation*}
\begin{split}
  \frac{\partial d}{\partial{t}} + \mathbf{u} \cdot \nabla d = 0 
\end{split}
\end{equation*}
Instead of using this formulation as a residual equation right away (as in {\hyperref[\detokenize{physicalloss::doc}]{\sphinxcrossref{\DUrole{doc}{Physical Loss Terms}}}}),
we can discretize it with our favorite mesh and discretization scheme,
to obtain a formulation that updates the state of our system over time. This is a standard
procedure for a \sphinxstyleemphasis{forward} solve.
To simplify things, we assume here that \(\mathbf{u}\) is only a function in space,
i.e. constant over time. We’ll bring back the time evolution of \(\mathbf{u}\) later on.

Let’s denote this re\sphinxhyphen{}formulation as \(\mathcal P\). It maps a state of \(d(t)\) into a
new state at an evolved time, i.e.:
\begin{equation*}
\begin{split}
    d(t+\Delta t) = \mathcal P ( ~ d(t), \mathbf{u}, t+\Delta t) 
\end{split}
\end{equation*}
As a simple example of an inverse problem and learning task, let’s consider the problem of
finding an unknown motion \(\mathbf{u}\):
this motion should transform a given initial scalar density state \(d^{~0}\) at time \(t^0\)
into a state that’s evolved by \(\mathcal P\) to a later “end” time \(t^e\)
with a certain shape or configuration \(d^{\text{target}}\).
Informally, we’d like to find a motion that deforms \(d^{~0}\) through the PDE model into a target state.
The simplest way to express this goal is via an \(L^2\) loss between the two states. So we want
to minimize the loss function \(L=|d(t^e) - d^{\text{target}}|^2\).

Note that as described here this inverse problem is a pure optimization task: there’s no NN involved,
and our goal is to obtain \(\mathbf{u}\). We do not want to apply this motion to other, unseen \sphinxstyleemphasis{test data},
as would be custom in a real learning task.

The final state of our marker density \(d(t^e)\) is fully determined by the evolution
of \(\mathcal P\) via \(\mathbf{u}\), which gives the following minimization problem:
\begin{equation*}
\begin{split}
    \text{arg min}_{~\mathbf{u}} | \mathcal P ( d^{~0}, \mathbf{u}, t^e) - d^{\text{target}}|^2
\end{split}
\end{equation*}
We’d now like to find the minimizer for this objective by
\sphinxstyleemphasis{gradient descent} (GD), where the
gradient is determined by the differentiable physics approach described earlier in this chapter.
Once things are working with GD, we can relatively easily switch to better optimizers or bring
an NN into the picture, hence it’s always a good starting point.

As the discretized velocity field \(\mathbf{u}\) contains all our degrees of freedom,
all we need to do is to update the velocity by an amount
\(\Delta \mathbf{u} = \partial L / \partial \mathbf{u}\),
which can be decomposed into
\(\Delta \mathbf{u} = 
\frac{ \partial d }{ \partial \mathbf{u}}
\frac{ \partial L }{ \partial d} \).

The \(\frac{ \partial L }{ \partial d}\) component is typically simple enough: we’ll get
\begin{equation*}
\begin{split} 
\frac{ \partial L }{ \partial d} 
    = \partial | \mathcal P ( d^{~0}, \mathbf{u}, t^e) - d^{\text{target}}|^2 / \partial d 
    = 2 (d(t^e)-d^{\text{target}}).
\end{split}
\end{equation*}
If \(d\) is represented as a vector, e.g., for one entry per cell of a mesh,
\(\frac{ \partial L }{ \partial d}\) will likewise be a column vector of equivalent size.
This is thanks to the fact that \(L\) is always a scalar loss function, and hence the Jacobian
matrix will have a dimension of 1 along the \(L\) dimension.
Intuitively, this vector will simply contain the differences between \(d\) at the end time
in comparison to the target densities \(d^{\text{target}}\).

The evolution of \(d\) itself is given by our discretized physical model \(\mathcal P\),
and we use \(\mathcal P\) and \(d\) interchangeably.
Hence, the more interesting component is the Jacobian
\(\partial d / \partial \mathbf{u} = \partial \mathcal P / \partial \mathbf{u}\) to
compute the full \(\Delta \mathbf{u} = 
 \frac{ \partial d }{ \partial \mathbf{u}}
 \frac{ \partial L }{ \partial d}\).
We luckily don’t need \(\partial d / \partial \mathbf{u}\) as a full
matrix, but instead only multiplied by \(\frac{ \partial L }{ \partial d}\).

So what is the actual Jacobian for \(d\)? To compute it we first need
to finalize our PDE model \(\mathcal P\), such that we get an expression which we can derive.
In the next section we’ll choose a specific advection scheme and a discretization
so that we can be more specific.

\subsection{Introducing a specific advection scheme}
\label{\detokenize{diffphys:introducing-a-specific-advection-scheme}}
In the following we’ll make use of a simple \sphinxhref{https://en.wikipedia.org/wiki/Upwind\_scheme}{first order upwinding scheme}
on a Cartesian grid in 1D, with marker density \(d_i\) and velocity \(u_i\) for cell \(i\).
We omit the \((t)\) for quantities at time \(t\) for brevity, i.e., \(d_i(t)\) is written as \(d_i\) below.
From above, we’ll use our \sphinxstyleemphasis{physical model} that updates the marker density
\(d_i(t+\Delta t) = \mathcal P ( d_i(t), \mathbf{u}(t), t + \Delta t)\), which
gives the following:
\begin{equation*}
\begin{split} \begin{aligned}
    & d_i(t+\Delta t) = d_i - u_i^+ (d_{i+1} - d_{i}) +  u_i^- (d_{i} - d_{i-1}) \text{ with }  \\
    & u_i^+ = \text{max}(u_i \Delta t / \Delta x,0) \\
    & u_i^- = \text{min}(u_i \Delta t / \Delta x,0)
\end{aligned} \end{split}
\end{equation*}
\begin{figure}[htbp]
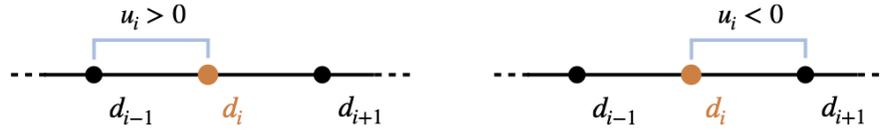

\centering
\capstart

\noindent\sphinxincludegraphics[height=150\sphinxpxdimen]{{diffphys-advect1d}.jpg}
\caption{1st\sphinxhyphen{}order upwinding uses a simple one\sphinxhyphen{}sided finite\sphinxhyphen{}difference stencil that takes into account the direction of the motion}\label{\detokenize{diffphys:advection-upwind}}\end{figure}

Thus, for a positive \(u_i\) we have
\begin{equation}\label{equation:diffphys:eq:advection}
\begin{split}
    \mathcal P ( d_i(t), \mathbf{u}(t), t + \Delta t) = (1 + \frac{u_i \Delta t }{ \Delta x}) d_i - \frac{u_i \Delta t }{ \Delta x} d_{i+1}
\end{split}
\end{equation}
and hence \(\partial \mathcal P / \partial u_i\) gives
\(\frac{\Delta t }{ \Delta x} d_i - \frac{\Delta t }{ \Delta x} d_{i+1}\). Intuitively,
the change of the velocity \(u_i\) depends on the spatial derivatives of the densities.
Due to the first order upwinding, we only include two neighbors (higher order methods would depend on
additional entries of \(d\))

In practice this step is equivalent to evaluating a transposed matrix multiplication.
If we rewrite the calculation above as
\( \mathcal P ( d_i(t), \mathbf{u}(t), t + \Delta t) = A \mathbf{u}\),
then \(\partial \mathcal P / \partial \mathbf{u} = A^T\).
However, in many practical cases, a matrix free implementation of this multiplication might
be preferable to actually constructing \(A\).

Another derivative that we can consider for the advection scheme is that w.r.t. the previous
density state, i.e. \(d_i(t)\), which is \(d_i\) in the shortened notation.
\(\partial \mathcal P / \partial d_i\) for cell \(i\) from above gives \(1 + \frac{u_i \Delta t }{ \Delta x}\). However, for the full gradient we’d need to add the potential contributions from cells \(i+1\) and \(i-1\), depending on the sign of their velocities. This derivative will come into play in the next section.

\subsection{Time evolution}
\label{\detokenize{diffphys:time-evolution}}
So far we’ve only dealt with a single update step of
\(d\) from time \(t\) to \(t+\Delta t\), but we could of course have an arbitrary number of such
steps. After all, above we stated the goal to advance the initial marker state \(d(t^0)\) to
the target state at time \(t^e\), which could encompass a long interval of time.

In the expression above for \(d_i(t+\Delta t)\), each of the \(d_i(t)\) in turn depends
on the velocity and density states at time \(t-\Delta t\), i.e., \(d_i(t-\Delta t)\). Thus we have to trace back
the influence of our loss \(L\) all the way back to how \(\mathbf{u}\) influences the initial marker
state. This can involve a large number of evaluations of our advection scheme via \(\mathcal P\).

This sounds challenging at first:
e.g., one could try to insert equation \eqref{equation:diffphys:eq:advection} at time \(t-\Delta t\)
into equation \eqref{equation:diffphys:eq:advection} at \(t\) and repeat this process recursively until
we have a single expression relating \(d^{~0}\) to the targets. However, thanks
to the linear nature of the Jacobians, we can treat each advection step, i.e.,
each invocation of our PDE \(\mathcal P\) as a seperate, modular
operation. And each of these invocations follows the procedure described
in the previous section.

Hence, given the machinery above, the backtrace is fairly simple to realize:
for each of the advection steps
in \(\mathcal P\) we can compute a Jacobian product with the \sphinxstyleemphasis{incoming} vector of derivatives
from the loss \(L\) or a previous advection step. We repeat this until we have traced the chain from the
loss with \(d^{\text{target}}\) all the way back to \(d^{~0}\).
Theoretically, the velocity \(\mathbf{u}\) could be a function of time, in which
case we’d get a gradient \(\Delta \mathbf{u}(t)\) for every time step \(t\). To simplify things
below, let’s we assume we have field that is constant in time, i.e., we’re
reusing the same velocities \(\mathbf{u}\) for every advection via \(\mathcal P\). Now, each time step
will give us a contribution to \(\Delta \mathbf{u}\) which we can accumulate for all steps.
\begin{equation*}
\begin{split} \begin{aligned}
    \Delta \mathbf{u} =& 
        \frac{ \partial d(t^e) }{ \partial \mathbf{u} }
        \frac{ \partial L }{ \partial d(t^e) }
        +
        \frac{ \partial d(t^e - \Delta t) }{ \partial \mathbf{u}}
        \frac{ \partial d(t^e) }{ \partial d(t^e - \Delta t) }
        \frac{ \partial L }{ \partial d(t^e)}
        \\
    & 
        + \ \cdots \ + \\
    & 
       \Big( \frac{ \partial d(t^0) }{ \partial \mathbf{u}} \cdots 
        \frac{ \partial d(t^e - \Delta t) }{ \partial d(t^e - 2 \Delta t) }
        \frac{ \partial d(t^e) }{ \partial d(t^e - \Delta t) }
        \frac{ \partial L }{ \partial d(t^e)} \Big)
\end{aligned} \end{split}
\end{equation*}
Here the last term above contains the full backtrace of the marker density to time \(t^0\).
The terms of this sum look unwieldy
at first, but they contain a lot of similar Jacobians, and in practice can be computed efficiently
by backtracing through the sequence of computational steps in the forward evaluation of our PDE.

This structure also makes clear that the process is very similar to the regular training
process of an NN: the evaluations of these Jacobian vector products is exactly what
a deep learning framework does for training an NN (we just have weights \(\theta\) instead
of a velocity field there). And hence all we need to do in practice is to provide a custom
function the Jacobian vector product for \(\mathcal P\).

\bigskip\hrule\bigskip

\section{Implicit gradient calculations}
\label{\detokenize{diffphys:implicit-gradient-calculations}}
As a slightly more complex example let’s consider Poisson’s equation \(\nabla^2 a = b\), where
\(a\) is the quantity of interest, and \(b\) is given.
This is a very fundamental elliptic PDE that is important for
a variety of physical problems, from electrostatics to gravitational fields. It also arises
in the context of fluids, where \(a\) takes the role of a scalar pressure field in the fluid, and
the right hand side \(b\) is given by the divergence of the fluid velocity \(\mathbf{u}\).

For fluids, we typically have
\(\mathbf{u}^{n} = \mathbf{u} - \nabla p\), with
\(\nabla^2 p = \nabla \cdot \mathbf{u}\). Here, \(\mathbf{u}^{n}\) denotes the \sphinxstyleemphasis{new}, divergence\sphinxhyphen{}free
velocity field. This step is typically crucial to enforce the hard\sphinxhyphen{}constraint \(\nabla \cdot \mathbf{u}=0\),
and is often called \sphinxstyleemphasis{Chorin Projection}, or \sphinxstyleemphasis{Helmholtz decomposition}. It is also closely related to the fundamental theorem of vector calculus.

If we now introduce an NN that modifies \(\mathbf{u}\) in a solver, we inevitably have to
backpropagate through the Poisson solve. I.e., we need a gradient for \(\mathbf{u}^{n}\), which in this
notation takes the form \(\partial \mathbf{u}^{n} / \partial \mathbf{u}\).

In combination, \(\mathbf{u}^{n} = \mathbf{u} - \nabla \left(  (\nabla^2)^{-1} \nabla \cdot \mathbf{u} \right)\). The outer gradient (from \(\nabla p\)) and the inner divergence (\(\nabla \cdot \mathbf{u}\)) are both linear operators, and their gradients are simple to compute. The main difficulty lies in obtaining the
matrix inverse \((\nabla^2)^{-1}\) from Poisson’s equation for pressure (we’ll keep it a bit simpler here, but it’s often time\sphinxhyphen{}dependent, and non\sphinxhyphen{}linear).

In practice, the matrix vector product for \((\nabla^2)^{-1} b\) with \(b=\nabla \cdot \mathbf{u}\) is not explicitly computed via matrix operations, but approximated with a (potentially matrix\sphinxhyphen{}free) iterative solver. E.g., conjugate gradient (CG) methods are a very popular choice here. Thus, we could treat this iterative solver as a function \(S\),
with \(p = S(\nabla \cdot \mathbf{u})\). Note that matrix inversion is a non\sphinxhyphen{}linear process, despite the matrix itself being linear. As solvers like CG are also based on matrix and vector operations, we could decompose \(S\) into a sequence of simpler operations \(S(x) = S_n( S_{n-1}(...S_{1}(x)))\), and backpropagate through each of them. This is certainly possible, but not a good idea: it can introduce numerical problems, and can be very slow.
As mentioned above, by default DL frameworks store the internal states for every differentiable operator like the \(S_i()\) in this example, and hence we’d organize and keep \(n\) intermediate states in memory. These states are completely uninteresting for our original PDE, though. They’re just intermediate states of the CG solver.

If we take a step back and look at \(p = (\nabla^2)^{-1} b\), it’s gradient \(\partial p / \partial b\)
is just \(((\nabla^2)^{-1})^T\). And in this case, \((\nabla^2)\) is a symmetric matrix, and so \(((\nabla^2)^{-1})^T=(\nabla^2)^{-1}\). This is the identical inverse matrix that we encountered in the original equation above, and hence we can re\sphinxhyphen{}use our unmodified iterative solver to compute the gradient. We don’t need to take it apart and slow it down by storing intermediate states. However, the iterative solver computes the matrix\sphinxhyphen{}vector\sphinxhyphen{}products for \((\nabla^2)^{-1} b\). So what is \(b\) during backpropagation? In an optimization setting we’ll always have our loss function \(L\) at the end of the forward chain. The backpropagation step will then give a gradient for the output, let’s assume it is \(\partial L/\partial p\) here, which needs to be propagated to the earlier operations of the forward pass. Thus, we can simply invoke our iterative solve during the backward pass to compute \(\partial p / \partial b = S(\partial L/\partial p)\). And assuming that we’ve chosen a good solver as \(S\) for the forward pass, we get exactly the same performance and accuracy in the backwards pass.

If you’re interested in a code example, the \sphinxhref{https://github.com/tum-pbs/PhiFlow/blob/master/demos/differentiate\_pressure.py}{differentiate\sphinxhyphen{}pressure example} of phiflow uses exactly this process for an optimization through a pressure projection step: a flow field that is constrained on the right side, is optimized for the content on the left, such that it matches the target on the right after a pressure projection step.

The main take\sphinxhyphen{}away here is: it is important \sphinxstyleemphasis{not to blindly backpropagate} through the forward computation, but to think about which steps of the analytic equations for the forward pass to compute gradients for. In cases like the above, we can often find improved analytic expressions for the gradients, which we can then compute numerically.

\begin{sphinxadmonition}{note}{Implicit Function Theorem \& Time}

\sphinxstylestrong{IFT}:
The process above essentially yields an \sphinxstyleemphasis{implicit derivative}. Instead of explicitly deriving all forward steps, we’ve relied on the \sphinxhref{https://en.wikipedia.org/wiki/Implicit\_function\_theorem}{implicit function theorem} to compute the derivative.

\sphinxstylestrong{Time}: we \sphinxstyleemphasis{can} actually consider the steps of an iterative solver as a virtual “time”,
and backpropagate through these steps. In line with other DP approaches, this enables an NN to \sphinxstyleemphasis{interact} with an iterative solver. An example is to learn initial guesses of CG solvers from {[}\hyperlink{cite.references:id6}{UBH+20}{]}.
\sphinxhref{https://github.com/tum-pbs/CG-Solver-in-the-Loop}{Details can be found here}
\end{sphinxadmonition}

\bigskip\hrule\bigskip

\section{Summary of differentiable physics so far}
\label{\detokenize{diffphys:summary-of-differentiable-physics-so-far}}
To summarize, using differentiable physical simulations
gives us a tool to include physical equations with a chosen discretization into DL.
In contrast to the residual constraints of the previous chapter,
this makes it possible to let NNs seamlessly interact with physical solvers.

We’d previously fully discard our physical model and solver
once the NN is trained: in the example from {\hyperref[\detokenize{physicalloss-code::doc}]{\sphinxcrossref{\DUrole{doc}{Burgers Optimization with a Physics\sphinxhyphen{}Informed NN}}}}
the NN gives us the solution directly, bypassing
any solver or model equation. With the DP approach we can train an NN alongside
a numerical solver, and thus we can make use of the physical model (as represented by
the solver) later on at inference time. This allows us to move beyond solving single
inverse problems, and can yield NNs that quite robustly generalize to new inputs.
Let’s revisit this sample problem in the context of DPs.

\chapter{Burgers Optimization with a Differentiable Physics Gradient}
\label{\detokenize{diffphys-code-burgers:burgers-optimization-with-a-differentiable-physics-gradient}}\label{\detokenize{diffphys-code-burgers::doc}}
To illustrate the process of computing gradients in a \sphinxstyleemphasis{differentiable physics} (DP) setting, we target the same inverse problem (the reconstruction task) used for the PINN example  in {\hyperref[\detokenize{physicalloss-code::doc}]{\sphinxcrossref{\DUrole{doc}{Burgers Optimization with a Physics\sphinxhyphen{}Informed NN}}}}. The choice of DP as a method has some immediate implications: we start with a discretized PDE, and the evolution of the  system is now fully determined by the resulting numerical solver. Hence, the only real unknown is the initial state. We will still need to re\sphinxhyphen{}compute all the states between the initial and target state many times, just now we won’t need an NN for this step. Instead, we can rely on our discretized model.

Also, as we choose an initial discretization for the DP approach, the unknown initial state consists of the sampling points of the involved physical fields, and we can simply represent these unknowns as floating point variables. Hence, even for the initial state we do not need to set up an NN. Thus, our Burgers reconstruction problem reduces to a gradient\sphinxhyphen{}based optimization without any NN when solving it with DP. Nonetheless, it’s a very good starting point to illustrate the process.

First, we’ll set up our discretized simulation. Here we can employ phiflow, as shown in the overview section on \sphinxstyleemphasis{Burgers forward simulations}.
\sphinxhref{https://colab.research.google.com/github/tum-pbs/pbdl-book/blob/main/diffphys-code-burgers.ipynb}{{[}run in colab{]}}

\section{Initialization}
\label{\detokenize{diffphys-code-burgers:initialization}}
phiflow directly gives us a sequence of differentiable operations, provided that we don’t use the \sphinxstyleemphasis{numpy} backend.
The important step here is to include \sphinxcode{\sphinxupquote{phi.tf.flow}} instad of \sphinxcode{\sphinxupquote{phi.flow}} (for \sphinxstyleemphasis{pytorch} you could use \sphinxcode{\sphinxupquote{phi.torch.flow}}).

So, as a first step, let’s set up some constants, and initialize a \sphinxcode{\sphinxupquote{velocity}} field with zeros, and our constraint at \(t=0.5\) (step 16), now as a \sphinxcode{\sphinxupquote{CenteredGrid}} in phiflow. Both are using periodic boundary conditions (via \sphinxcode{\sphinxupquote{extrapolation.PERIODIC}}) and a spatial discretization of \(\Delta x = 1/128\).

\begin{sphinxVerbatim}[commandchars=\\\{\}]
\PYG{c+ch}{\PYGZsh{}!pip install \PYGZhy{}\PYGZhy{}upgrade \PYGZhy{}\PYGZhy{}quiet phiflow}
\PYG{o}{!}pip install \PYGZhy{}\PYGZhy{}upgrade \PYGZhy{}\PYGZhy{}quiet git+https://github.com/tum\PYGZhy{}pbs/PhiFlow@develop 
\PYG{k+kn}{from} \PYG{n+nn}{phi}\PYG{n+nn}{.}\PYG{n+nn}{tf}\PYG{n+nn}{.}\PYG{n+nn}{flow} \PYG{k+kn}{import} \PYG{o}{*}

\PYG{n}{N} \PYG{o}{=} \PYG{l+m+mi}{128}
\PYG{n}{STEPS} \PYG{o}{=} \PYG{l+m+mi}{32}
\PYG{n}{DT} \PYG{o}{=} \PYG{l+m+mf}{1.}\PYG{o}{/}\PYG{n}{STEPS}
\PYG{n}{NU} \PYG{o}{=} \PYG{l+m+mf}{0.2}\PYG{o}{/}\PYG{n}{np}\PYG{o}{.}\PYG{n}{pi}

\PYG{c+c1}{\PYGZsh{} allocate velocity grid}
\PYG{n}{velocity} \PYG{o}{=} \PYG{n}{CenteredGrid}\PYG{p}{(}\PYG{l+m+mi}{0}\PYG{p}{,} \PYG{n}{extrapolation}\PYG{o}{.}\PYG{n}{PERIODIC}\PYG{p}{,} \PYG{n}{x}\PYG{o}{=}\PYG{n}{N}\PYG{p}{,} \PYG{n}{bounds}\PYG{o}{=}\PYG{n}{Box}\PYG{p}{[}\PYG{o}{\PYGZhy{}}\PYG{l+m+mi}{1}\PYG{p}{:}\PYG{l+m+mi}{1}\PYG{p}{]}\PYG{p}{)}

\PYG{c+c1}{\PYGZsh{} and a grid with the reference solution }
\PYG{n}{REFERENCE\PYGZus{}DATA} \PYG{o}{=} \PYG{n}{math}\PYG{o}{.}\PYG{n}{tensor}\PYG{p}{(}\PYG{p}{[}\PYG{l+m+mf}{0.008612174447657694}\PYG{p}{,} \PYG{l+m+mf}{0.02584669669548606}\PYG{p}{,} \PYG{l+m+mf}{0.043136357266407785} \PYG{o}{.}\PYG{o}{.}\PYG{o}{.} \PYG{p}{]} \PYG{p}{,} \PYG{n}{math}\PYG{o}{.}\PYG{n}{spatial}\PYG{p}{(}\PYG{l+s+s1}{\PYGZsq{}}\PYG{l+s+s1}{x}\PYG{l+s+s1}{\PYGZsq{}}\PYG{p}{)}\PYG{p}{)}
\PYG{n}{SOLUTION\PYGZus{}T16} \PYG{o}{=} \PYG{n}{CenteredGrid}\PYG{p}{(} \PYG{n}{REFERENCE\PYGZus{}DATA}\PYG{p}{,} \PYG{n}{extrapolation}\PYG{o}{.}\PYG{n}{PERIODIC}\PYG{p}{,} \PYG{n}{x}\PYG{o}{=}\PYG{n}{N}\PYG{p}{,} \PYG{n}{bounds}\PYG{o}{=}\PYG{n}{Box}\PYG{p}{[}\PYG{o}{\PYGZhy{}}\PYG{l+m+mi}{1}\PYG{p}{:}\PYG{l+m+mi}{1}\PYG{p}{]}\PYG{p}{)}
\end{sphinxVerbatim}

We can verify that the fields of our simulation are now backed by TensorFlow.

\begin{sphinxVerbatim}[commandchars=\\\{\}]
\PYG{n+nb}{type}\PYG{p}{(}\PYG{n}{velocity}\PYG{o}{.}\PYG{n}{values}\PYG{o}{.}\PYG{n}{native}\PYG{p}{(}\PYG{p}{)}\PYG{p}{)}
\end{sphinxVerbatim}

\begin{sphinxVerbatim}[commandchars=\\\{\}]
tensorflow.python.framework.ops.EagerTensor
\end{sphinxVerbatim}

\section{Gradients}
\label{\detokenize{diffphys-code-burgers:gradients}}
The \sphinxcode{\sphinxupquote{record\_gradients}} function of phiflow triggers the generation of a gradient tape to compute gradients of a simulation via \sphinxcode{\sphinxupquote{math.gradients(loss, values)}}.

To use it for the Burgers case we need to specify a loss function: we want the solution at \(t=0.5\) to match the reference data. Thus we simply compute an \(L^2\) difference between step number 16 and our constraint array as \sphinxcode{\sphinxupquote{loss}}. Afterwards, we evaluate the gradient of the initial velocity state \sphinxcode{\sphinxupquote{velocity}} with respect to this loss.

\begin{sphinxVerbatim}[commandchars=\\\{\}]
\PYG{n}{velocities} \PYG{o}{=} \PYG{p}{[}\PYG{n}{velocity}\PYG{p}{]}

\PYG{k}{with} \PYG{n}{math}\PYG{o}{.}\PYG{n}{record\PYGZus{}gradients}\PYG{p}{(}\PYG{n}{velocity}\PYG{o}{.}\PYG{n}{values}\PYG{p}{)}\PYG{p}{:}

    \PYG{k}{for} \PYG{n}{time\PYGZus{}step} \PYG{o+ow}{in} \PYG{n+nb}{range}\PYG{p}{(}\PYG{n}{STEPS}\PYG{p}{)}\PYG{p}{:}
        \PYG{n}{v1} \PYG{o}{=} \PYG{n}{diffuse}\PYG{o}{.}\PYG{n}{explicit}\PYG{p}{(}\PYG{l+m+mf}{1.0}\PYG{o}{*}\PYG{n}{velocities}\PYG{p}{[}\PYG{o}{\PYGZhy{}}\PYG{l+m+mi}{1}\PYG{p}{]}\PYG{p}{,} \PYG{n}{NU}\PYG{p}{,} \PYG{n}{DT}\PYG{p}{,} \PYG{n}{substeps}\PYG{o}{=}\PYG{l+m+mi}{1}\PYG{p}{)}
        \PYG{n}{v2} \PYG{o}{=} \PYG{n}{advect}\PYG{o}{.}\PYG{n}{semi\PYGZus{}lagrangian}\PYG{p}{(}\PYG{n}{v1}\PYG{p}{,} \PYG{n}{v1}\PYG{p}{,} \PYG{n}{DT}\PYG{p}{)}
        \PYG{n}{velocities}\PYG{o}{.}\PYG{n}{append}\PYG{p}{(}\PYG{n}{v2}\PYG{p}{)}

    \PYG{n}{loss} \PYG{o}{=} \PYG{n}{field}\PYG{o}{.}\PYG{n}{l2\PYGZus{}loss}\PYG{p}{(}\PYG{n}{velocities}\PYG{p}{[}\PYG{l+m+mi}{16}\PYG{p}{]} \PYG{o}{\PYGZhy{}} \PYG{n}{SOLUTION\PYGZus{}T16}\PYG{p}{)}\PYG{o}{*}\PYG{l+m+mf}{2.}\PYG{o}{/}\PYG{n}{N} \PYG{c+c1}{\PYGZsh{} MSE}

    \PYG{n}{grad} \PYG{o}{=} \PYG{n}{math}\PYG{o}{.}\PYG{n}{gradients}\PYG{p}{(}\PYG{n}{loss}\PYG{p}{,} \PYG{n}{velocity}\PYG{o}{.}\PYG{n}{values}\PYG{p}{)}

\PYG{n+nb}{print}\PYG{p}{(}\PYG{l+s+s1}{\PYGZsq{}}\PYG{l+s+s1}{Loss: }\PYG{l+s+si}{\PYGZpc{}f}\PYG{l+s+s1}{\PYGZsq{}} \PYG{o}{\PYGZpc{}} \PYG{p}{(}\PYG{n}{loss}\PYG{p}{)}\PYG{p}{)}
\end{sphinxVerbatim}

\begin{sphinxVerbatim}[commandchars=\\\{\}]
Instructions for updating:
This op will be removed after the deprecation date. Please switch to tf.sets.difference().
Loss: 0.382915
\end{sphinxVerbatim}

Because we’re only constraining time step 16, we could actually omit steps 17 to 31 in this setup. They don’t have any degrees of freedom and are not constrained in any way. However, for fairness regarding a comparison with the previous PINN case, we include them.

Note that we’ve done a lot of calculations here: first the 32 steps of our simulation, and then another 16 steps backwards from the loss. They were recorded by the gradient tape, and used to backpropagate the loss to the initial state of the simulation.

Not surprisingly, because we’re starting from zero, there’s also a significant initial error of ca. 0.38 for the 16th simulation step.

So what do we get as a gradient here? It has the same dimensions as the velocity, and we can easily visualize it:
Starting from the zero state for \sphinxcode{\sphinxupquote{velocity}} (shown in blue), the first gradient is shown as a green line below. If you compare it with the solution it points in the opposite direction, as expected. The solution is much larger in magnitude, so we omit it here (see the next graph).

\begin{sphinxVerbatim}[commandchars=\\\{\}]
\PYG{k+kn}{import} \PYG{n+nn}{pylab} \PYG{k}{as} \PYG{n+nn}{plt}

\PYG{n}{fig} \PYG{o}{=} \PYG{n}{plt}\PYG{o}{.}\PYG{n}{figure}\PYG{p}{(}\PYG{p}{)}\PYG{o}{.}\PYG{n}{gca}\PYG{p}{(}\PYG{p}{)}
\PYG{n}{pltx} \PYG{o}{=} \PYG{n}{np}\PYG{o}{.}\PYG{n}{linspace}\PYG{p}{(}\PYG{o}{\PYGZhy{}}\PYG{l+m+mi}{1}\PYG{p}{,}\PYG{l+m+mi}{1}\PYG{p}{,}\PYG{n}{N}\PYG{p}{)}

\PYG{c+c1}{\PYGZsh{} first gradient}
\PYG{n}{fig}\PYG{o}{.}\PYG{n}{plot}\PYG{p}{(}\PYG{n}{pltx}\PYG{p}{,} \PYG{n}{grad}\PYG{o}{.}\PYG{n}{numpy}\PYG{p}{(}\PYG{l+s+s1}{\PYGZsq{}}\PYG{l+s+s1}{x}\PYG{l+s+s1}{\PYGZsq{}}\PYG{p}{)}           \PYG{p}{,} \PYG{n}{lw}\PYG{o}{=}\PYG{l+m+mi}{2}\PYG{p}{,} \PYG{n}{color}\PYG{o}{=}\PYG{l+s+s1}{\PYGZsq{}}\PYG{l+s+s1}{green}\PYG{l+s+s1}{\PYGZsq{}}\PYG{p}{,}      \PYG{n}{label}\PYG{o}{=}\PYG{l+s+s2}{\PYGZdq{}}\PYG{l+s+s2}{Gradient}\PYG{l+s+s2}{\PYGZdq{}}\PYG{p}{)}  
\PYG{n}{fig}\PYG{o}{.}\PYG{n}{plot}\PYG{p}{(}\PYG{n}{pltx}\PYG{p}{,} \PYG{n}{velocity}\PYG{o}{.}\PYG{n}{values}\PYG{o}{.}\PYG{n}{numpy}\PYG{p}{(}\PYG{l+s+s1}{\PYGZsq{}}\PYG{l+s+s1}{x}\PYG{l+s+s1}{\PYGZsq{}}\PYG{p}{)}\PYG{p}{,} \PYG{n}{lw}\PYG{o}{=}\PYG{l+m+mi}{2}\PYG{p}{,} \PYG{n}{color}\PYG{o}{=}\PYG{l+s+s1}{\PYGZsq{}}\PYG{l+s+s1}{mediumblue}\PYG{l+s+s1}{\PYGZsq{}}\PYG{p}{,} \PYG{n}{label}\PYG{o}{=}\PYG{l+s+s2}{\PYGZdq{}}\PYG{l+s+s2}{u at t=0}\PYG{l+s+s2}{\PYGZdq{}}\PYG{p}{)}
\PYG{n}{plt}\PYG{o}{.}\PYG{n}{xlabel}\PYG{p}{(}\PYG{l+s+s1}{\PYGZsq{}}\PYG{l+s+s1}{x}\PYG{l+s+s1}{\PYGZsq{}}\PYG{p}{)}\PYG{p}{;} \PYG{n}{plt}\PYG{o}{.}\PYG{n}{ylabel}\PYG{p}{(}\PYG{l+s+s1}{\PYGZsq{}}\PYG{l+s+s1}{u}\PYG{l+s+s1}{\PYGZsq{}}\PYG{p}{)}\PYG{p}{;} \PYG{n}{plt}\PYG{o}{.}\PYG{n}{legend}\PYG{p}{(}\PYG{p}{)}\PYG{p}{;}

\PYG{c+c1}{\PYGZsh{} some (optional) other fields to plot:}
\PYG{c+c1}{\PYGZsh{}fig.plot(pltx,  (velocities[16]).values.numpy(\PYGZsq{}x\PYGZsq{})   , lw=2, color=\PYGZsq{}cyan\PYGZsq{}, label=\PYGZdq{}u at t=0.5\PYGZdq{})  }
\PYG{c+c1}{\PYGZsh{}fig.plot(pltx,  (SOLUTION\PYGZus{}T16).values.numpy(\PYGZsq{}x\PYGZsq{})   , lw=2, color=\PYGZsq{}red\PYGZsq{}, label=\PYGZdq{}solution at t=0.5\PYGZdq{})  }
\PYG{c+c1}{\PYGZsh{}fig.plot(pltx,  (velocities[16] \PYGZhy{} SOLUTION\PYGZus{}T16).values.numpy(\PYGZsq{}x\PYGZsq{})   , lw=2, color=\PYGZsq{}blue\PYGZsq{}, label=\PYGZdq{}difference at t=0.5\PYGZdq{})  }
\end{sphinxVerbatim}

\noindent\sphinxincludegraphics{{diffphys-code-burgers_7_0}.png}

This gives us a “search direction” for each velocity variable. Based on a linear approximation, the gradient tells us how to change each of them to increase the loss function (gradients \sphinxstyleemphasis{always} point “upwards”). Thus, we can use the gradient to run an optimization and find an initial state \sphinxcode{\sphinxupquote{velocity}} that minimizes our loss.

\section{Optimization}
\label{\detokenize{diffphys-code-burgers:optimization}}
Equipped with the gradient we can run a gradient descent optimization. Below, we’re using a learning rate of \sphinxcode{\sphinxupquote{LR=5}}, and we’re re\sphinxhyphen{}evaluating the loss for the updated state to track convergence.

In the following code block, we’re additionally saving the gradients in a list called \sphinxcode{\sphinxupquote{grads}}, such that we can visualize them later on. For a regular optimization we could of course discard the gradient after performing an update of the velocity.

\begin{sphinxVerbatim}[commandchars=\\\{\}]
\PYG{n}{LR} \PYG{o}{=} \PYG{l+m+mf}{5.}

\PYG{n}{grads}\PYG{o}{=}\PYG{p}{[}\PYG{p}{]}
\PYG{k}{for} \PYG{n}{optim\PYGZus{}step} \PYG{o+ow}{in} \PYG{n+nb}{range}\PYG{p}{(}\PYG{l+m+mi}{5}\PYG{p}{)}\PYG{p}{:}
    \PYG{n}{velocities} \PYG{o}{=} \PYG{p}{[}\PYG{n}{velocity}\PYG{p}{]}
    \PYG{k}{with} \PYG{n}{math}\PYG{o}{.}\PYG{n}{record\PYGZus{}gradients}\PYG{p}{(}\PYG{n}{velocity}\PYG{o}{.}\PYG{n}{values}\PYG{p}{)}\PYG{p}{:}
        \PYG{k}{for} \PYG{n}{time\PYGZus{}step} \PYG{o+ow}{in} \PYG{n+nb}{range}\PYG{p}{(}\PYG{n}{STEPS}\PYG{p}{)}\PYG{p}{:}
            \PYG{n}{v1} \PYG{o}{=} \PYG{n}{diffuse}\PYG{o}{.}\PYG{n}{explicit}\PYG{p}{(}\PYG{l+m+mf}{1.0}\PYG{o}{*}\PYG{n}{velocities}\PYG{p}{[}\PYG{o}{\PYGZhy{}}\PYG{l+m+mi}{1}\PYG{p}{]}\PYG{p}{,} \PYG{n}{NU}\PYG{p}{,} \PYG{n}{DT}\PYG{p}{)}
            \PYG{n}{v2} \PYG{o}{=} \PYG{n}{advect}\PYG{o}{.}\PYG{n}{semi\PYGZus{}lagrangian}\PYG{p}{(}\PYG{n}{v1}\PYG{p}{,} \PYG{n}{v1}\PYG{p}{,} \PYG{n}{DT}\PYG{p}{)}
            \PYG{n}{velocities}\PYG{o}{.}\PYG{n}{append}\PYG{p}{(}\PYG{n}{v2}\PYG{p}{)}

        \PYG{n}{loss} \PYG{o}{=} \PYG{n}{field}\PYG{o}{.}\PYG{n}{l2\PYGZus{}loss}\PYG{p}{(}\PYG{n}{velocities}\PYG{p}{[}\PYG{l+m+mi}{16}\PYG{p}{]} \PYG{o}{\PYGZhy{}} \PYG{n}{SOLUTION\PYGZus{}T16}\PYG{p}{)}\PYG{o}{*}\PYG{l+m+mf}{2.}\PYG{o}{/}\PYG{n}{N} \PYG{c+c1}{\PYGZsh{} MSE}
        \PYG{n+nb}{print}\PYG{p}{(}\PYG{l+s+s1}{\PYGZsq{}}\PYG{l+s+s1}{Optimization step }\PYG{l+s+si}{\PYGZpc{}d}\PYG{l+s+s1}{, loss: }\PYG{l+s+si}{\PYGZpc{}f}\PYG{l+s+s1}{\PYGZsq{}} \PYG{o}{\PYGZpc{}} \PYG{p}{(}\PYG{n}{optim\PYGZus{}step}\PYG{p}{,}\PYG{n}{loss}\PYG{p}{)}\PYG{p}{)}

        \PYG{n}{grads}\PYG{o}{.}\PYG{n}{append}\PYG{p}{(} \PYG{n}{math}\PYG{o}{.}\PYG{n}{gradients}\PYG{p}{(}\PYG{n}{loss}\PYG{p}{,} \PYG{n}{velocity}\PYG{o}{.}\PYG{n}{values}\PYG{p}{)} \PYG{p}{)}

    \PYG{n}{velocity} \PYG{o}{=} \PYG{n}{velocity} \PYG{o}{\PYGZhy{}} \PYG{n}{LR} \PYG{o}{*} \PYG{n}{grads}\PYG{p}{[}\PYG{o}{\PYGZhy{}}\PYG{l+m+mi}{1}\PYG{p}{]}
\end{sphinxVerbatim}

\begin{sphinxVerbatim}[commandchars=\\\{\}]
Optimization step 0, loss: 0.382915
Optimization step 1, loss: 0.329677
Optimization step 2, loss: 0.284684
Optimization step 3, loss: 0.246604
Optimization step 4, loss: 0.214305
\end{sphinxVerbatim}

Now we can check well the 16th state of the simulation actually matches the target after the 5 update steps. This is what the loss measures, after all. The next graph shows the constraints (i.e. the solution we’d like to obtain) in green, and the reconstructed state after the initial state \sphinxcode{\sphinxupquote{velocity}} (which we updated five times via the gradient by now) was updated 16 times by the solver.

\begin{sphinxVerbatim}[commandchars=\\\{\}]
\PYG{n}{fig} \PYG{o}{=} \PYG{n}{plt}\PYG{o}{.}\PYG{n}{figure}\PYG{p}{(}\PYG{p}{)}\PYG{o}{.}\PYG{n}{gca}\PYG{p}{(}\PYG{p}{)}

\PYG{c+c1}{\PYGZsh{} target constraint at t=0.5}
\PYG{n}{fig}\PYG{o}{.}\PYG{n}{plot}\PYG{p}{(}\PYG{n}{pltx}\PYG{p}{,} \PYG{n}{SOLUTION\PYGZus{}T16}\PYG{o}{.}\PYG{n}{values}\PYG{o}{.}\PYG{n}{numpy}\PYG{p}{(}\PYG{l+s+s1}{\PYGZsq{}}\PYG{l+s+s1}{x}\PYG{l+s+s1}{\PYGZsq{}}\PYG{p}{)}\PYG{p}{,} \PYG{n}{lw}\PYG{o}{=}\PYG{l+m+mi}{2}\PYG{p}{,} \PYG{n}{color}\PYG{o}{=}\PYG{l+s+s1}{\PYGZsq{}}\PYG{l+s+s1}{forestgreen}\PYG{l+s+s1}{\PYGZsq{}}\PYG{p}{,}  \PYG{n}{label}\PYG{o}{=}\PYG{l+s+s2}{\PYGZdq{}}\PYG{l+s+s2}{Reference}\PYG{l+s+s2}{\PYGZdq{}}\PYG{p}{)} 

\PYG{c+c1}{\PYGZsh{} optimized state of our simulation after 16 steps}
\PYG{n}{fig}\PYG{o}{.}\PYG{n}{plot}\PYG{p}{(}\PYG{n}{pltx}\PYG{p}{,} \PYG{n}{velocities}\PYG{p}{[}\PYG{l+m+mi}{16}\PYG{p}{]}\PYG{o}{.}\PYG{n}{values}\PYG{o}{.}\PYG{n}{numpy}\PYG{p}{(}\PYG{l+s+s1}{\PYGZsq{}}\PYG{l+s+s1}{x}\PYG{l+s+s1}{\PYGZsq{}}\PYG{p}{)}\PYG{p}{,} \PYG{n}{lw}\PYG{o}{=}\PYG{l+m+mi}{2}\PYG{p}{,} \PYG{n}{color}\PYG{o}{=}\PYG{l+s+s1}{\PYGZsq{}}\PYG{l+s+s1}{mediumblue}\PYG{l+s+s1}{\PYGZsq{}}\PYG{p}{,} \PYG{n}{label}\PYG{o}{=}\PYG{l+s+s2}{\PYGZdq{}}\PYG{l+s+s2}{Simulated velocity}\PYG{l+s+s2}{\PYGZdq{}}\PYG{p}{)}

\PYG{n}{plt}\PYG{o}{.}\PYG{n}{xlabel}\PYG{p}{(}\PYG{l+s+s1}{\PYGZsq{}}\PYG{l+s+s1}{x}\PYG{l+s+s1}{\PYGZsq{}}\PYG{p}{)}\PYG{p}{;} \PYG{n}{plt}\PYG{o}{.}\PYG{n}{ylabel}\PYG{p}{(}\PYG{l+s+s1}{\PYGZsq{}}\PYG{l+s+s1}{u}\PYG{l+s+s1}{\PYGZsq{}}\PYG{p}{)}\PYG{p}{;} \PYG{n}{plt}\PYG{o}{.}\PYG{n}{legend}\PYG{p}{(}\PYG{p}{)}\PYG{p}{;} \PYG{n}{plt}\PYG{o}{.}\PYG{n}{title}\PYG{p}{(}\PYG{l+s+s2}{\PYGZdq{}}\PYG{l+s+s2}{After 5 Optimization Steps at t=0.5}\PYG{l+s+s2}{\PYGZdq{}}\PYG{p}{)}\PYG{p}{;}
\end{sphinxVerbatim}

\noindent\sphinxincludegraphics{{diffphys-code-burgers_12_0}.png}

This seems to be going in the right direction! It’s definitely not perfect, but we’ve only computed 5 GD update steps so far. The two peaks with a positive velocity on the left side of the shock and the negative peak on the right side are starting to show.

This is a good indicator that the backpropagation of gradients through all of our 16 simulated steps is behaving correctly, and that it’s driving the solution in the right direction. The graph above only hints at how powerful the setup is: the gradient that we obtain from each of the simulation steps (and each operation within them) can easily be chained together into more complex sequences. In the example above, we’re backpropagating through all 16 steps of the simulation, and we could easily enlarge this “look\sphinxhyphen{}ahead” of the optimization with minor changes to the code.

\section{More optimization steps}
\label{\detokenize{diffphys-code-burgers:more-optimization-steps}}
Before moving on to more complex physics simulations, or involving NNs, let’s finish the optimization task at hand, and run more steps to get a better solution.

\begin{sphinxVerbatim}[commandchars=\\\{\}]
\PYG{k+kn}{import} \PYG{n+nn}{time}
\PYG{n}{start} \PYG{o}{=} \PYG{n}{time}\PYG{o}{.}\PYG{n}{time}\PYG{p}{(}\PYG{p}{)}

\PYG{k}{for} \PYG{n}{optim\PYGZus{}step} \PYG{o+ow}{in} \PYG{n+nb}{range}\PYG{p}{(}\PYG{l+m+mi}{45}\PYG{p}{)}\PYG{p}{:}
    \PYG{n}{velocities} \PYG{o}{=} \PYG{p}{[}\PYG{n}{velocity}\PYG{p}{]}
    \PYG{k}{with} \PYG{n}{math}\PYG{o}{.}\PYG{n}{record\PYGZus{}gradients}\PYG{p}{(}\PYG{n}{velocity}\PYG{o}{.}\PYG{n}{values}\PYG{p}{)}\PYG{p}{:}
        \PYG{k}{for} \PYG{n}{time\PYGZus{}step} \PYG{o+ow}{in} \PYG{n+nb}{range}\PYG{p}{(}\PYG{n}{STEPS}\PYG{p}{)}\PYG{p}{:}
            \PYG{n}{v1} \PYG{o}{=} \PYG{n}{diffuse}\PYG{o}{.}\PYG{n}{explicit}\PYG{p}{(}\PYG{l+m+mf}{1.0}\PYG{o}{*}\PYG{n}{velocities}\PYG{p}{[}\PYG{o}{\PYGZhy{}}\PYG{l+m+mi}{1}\PYG{p}{]}\PYG{p}{,} \PYG{n}{NU}\PYG{p}{,} \PYG{n}{DT}\PYG{p}{)}
            \PYG{n}{v2} \PYG{o}{=} \PYG{n}{advect}\PYG{o}{.}\PYG{n}{semi\PYGZus{}lagrangian}\PYG{p}{(}\PYG{n}{v1}\PYG{p}{,} \PYG{n}{v1}\PYG{p}{,} \PYG{n}{DT}\PYG{p}{)}
            \PYG{n}{velocities}\PYG{o}{.}\PYG{n}{append}\PYG{p}{(}\PYG{n}{v2}\PYG{p}{)}

        \PYG{n}{loss} \PYG{o}{=} \PYG{n}{field}\PYG{o}{.}\PYG{n}{l2\PYGZus{}loss}\PYG{p}{(}\PYG{n}{velocities}\PYG{p}{[}\PYG{l+m+mi}{16}\PYG{p}{]} \PYG{o}{\PYGZhy{}} \PYG{n}{SOLUTION\PYGZus{}T16}\PYG{p}{)}\PYG{o}{*}\PYG{l+m+mf}{2.}\PYG{o}{/}\PYG{n}{N} \PYG{c+c1}{\PYGZsh{} MSE}
        \PYG{k}{if} \PYG{n}{optim\PYGZus{}step}\PYG{o}{\PYGZpc{}}\PYG{k}{5}==0: 
            \PYG{n+nb}{print}\PYG{p}{(}\PYG{l+s+s1}{\PYGZsq{}}\PYG{l+s+s1}{Optimization step }\PYG{l+s+si}{\PYGZpc{}d}\PYG{l+s+s1}{, loss: }\PYG{l+s+si}{\PYGZpc{}f}\PYG{l+s+s1}{\PYGZsq{}} \PYG{o}{\PYGZpc{}} \PYG{p}{(}\PYG{n}{optim\PYGZus{}step}\PYG{p}{,}\PYG{n}{loss}\PYG{p}{)}\PYG{p}{)}

        \PYG{n}{grad} \PYG{o}{=} \PYG{n}{math}\PYG{o}{.}\PYG{n}{gradients}\PYG{p}{(}\PYG{n}{loss}\PYG{p}{,} \PYG{n}{velocity}\PYG{o}{.}\PYG{n}{values}\PYG{p}{)}

    \PYG{n}{velocity} \PYG{o}{=} \PYG{n}{velocity} \PYG{o}{\PYGZhy{}} \PYG{n}{LR} \PYG{o}{*} \PYG{n}{grad}

\PYG{n}{end} \PYG{o}{=} \PYG{n}{time}\PYG{o}{.}\PYG{n}{time}\PYG{p}{(}\PYG{p}{)}
\PYG{n+nb}{print}\PYG{p}{(}\PYG{l+s+s2}{\PYGZdq{}}\PYG{l+s+s2}{Runtime }\PYG{l+s+si}{\PYGZob{}:.2f\PYGZcb{}}\PYG{l+s+s2}{s}\PYG{l+s+s2}{\PYGZdq{}}\PYG{o}{.}\PYG{n}{format}\PYG{p}{(}\PYG{n}{end}\PYG{o}{\PYGZhy{}}\PYG{n}{start}\PYG{p}{)}\PYG{p}{)}
\end{sphinxVerbatim}

\begin{sphinxVerbatim}[commandchars=\\\{\}]
Optimization step 0, loss: 0.186823
Optimization step 5, loss: 0.098101
Optimization step 10, loss: 0.055802
Optimization step 15, loss: 0.033452
Optimization step 20, loss: 0.020761
Optimization step 25, loss: 0.013248
Optimization step 30, loss: 0.008643
Optimization step 35, loss: 0.005733
Optimization step 40, loss: 0.003848
Runtime 204.01s
\end{sphinxVerbatim}

Thinking back to the PINN version from {\hyperref[\detokenize{diffphys-code-burgers::doc}]{\sphinxcrossref{\DUrole{doc}{Burgers Optimization with a Differentiable Physics Gradient}}}}, we have a much lower error here after only 50 steps (by ca. an order of magnitude), and the runtime is also lower (roughly by a factor of 2).

Let’s plot again how well our solution at \(t=0.5\) (blue) matches the constraints (green) now:

\begin{sphinxVerbatim}[commandchars=\\\{\}]
\PYG{n}{fig} \PYG{o}{=} \PYG{n}{plt}\PYG{o}{.}\PYG{n}{figure}\PYG{p}{(}\PYG{p}{)}\PYG{o}{.}\PYG{n}{gca}\PYG{p}{(}\PYG{p}{)}
\PYG{n}{fig}\PYG{o}{.}\PYG{n}{plot}\PYG{p}{(}\PYG{n}{pltx}\PYG{p}{,} \PYG{n}{SOLUTION\PYGZus{}T16}\PYG{o}{.}\PYG{n}{values}\PYG{o}{.}\PYG{n}{numpy}\PYG{p}{(}\PYG{l+s+s1}{\PYGZsq{}}\PYG{l+s+s1}{x}\PYG{l+s+s1}{\PYGZsq{}}\PYG{p}{)}\PYG{p}{,} \PYG{n}{lw}\PYG{o}{=}\PYG{l+m+mi}{2}\PYG{p}{,} \PYG{n}{color}\PYG{o}{=}\PYG{l+s+s1}{\PYGZsq{}}\PYG{l+s+s1}{forestgreen}\PYG{l+s+s1}{\PYGZsq{}}\PYG{p}{,}  \PYG{n}{label}\PYG{o}{=}\PYG{l+s+s2}{\PYGZdq{}}\PYG{l+s+s2}{Reference}\PYG{l+s+s2}{\PYGZdq{}}\PYG{p}{)} 
\PYG{n}{fig}\PYG{o}{.}\PYG{n}{plot}\PYG{p}{(}\PYG{n}{pltx}\PYG{p}{,} \PYG{n}{velocities}\PYG{p}{[}\PYG{l+m+mi}{16}\PYG{p}{]}\PYG{o}{.}\PYG{n}{values}\PYG{o}{.}\PYG{n}{numpy}\PYG{p}{(}\PYG{l+s+s1}{\PYGZsq{}}\PYG{l+s+s1}{x}\PYG{l+s+s1}{\PYGZsq{}}\PYG{p}{)}\PYG{p}{,} \PYG{n}{lw}\PYG{o}{=}\PYG{l+m+mi}{2}\PYG{p}{,} \PYG{n}{color}\PYG{o}{=}\PYG{l+s+s1}{\PYGZsq{}}\PYG{l+s+s1}{mediumblue}\PYG{l+s+s1}{\PYGZsq{}}\PYG{p}{,} \PYG{n}{label}\PYG{o}{=}\PYG{l+s+s2}{\PYGZdq{}}\PYG{l+s+s2}{Simulated velocity}\PYG{l+s+s2}{\PYGZdq{}}\PYG{p}{)}
\PYG{n}{plt}\PYG{o}{.}\PYG{n}{xlabel}\PYG{p}{(}\PYG{l+s+s1}{\PYGZsq{}}\PYG{l+s+s1}{x}\PYG{l+s+s1}{\PYGZsq{}}\PYG{p}{)}\PYG{p}{;} \PYG{n}{plt}\PYG{o}{.}\PYG{n}{ylabel}\PYG{p}{(}\PYG{l+s+s1}{\PYGZsq{}}\PYG{l+s+s1}{u}\PYG{l+s+s1}{\PYGZsq{}}\PYG{p}{)}\PYG{p}{;} \PYG{n}{plt}\PYG{o}{.}\PYG{n}{legend}\PYG{p}{(}\PYG{p}{)}\PYG{p}{;} \PYG{n}{plt}\PYG{o}{.}\PYG{n}{title}\PYG{p}{(}\PYG{l+s+s2}{\PYGZdq{}}\PYG{l+s+s2}{After 50 Optimization Steps at t=0.5}\PYG{l+s+s2}{\PYGZdq{}}\PYG{p}{)}\PYG{p}{;}
\end{sphinxVerbatim}

\noindent\sphinxincludegraphics{{diffphys-code-burgers_17_0}.png}

Not bad. But how well is the initial state recovered via backpropagation through the 16 simulation steps? This is what we’re changing, and because it’s only indirectly constrained via the observation later in time there is more room to deviate from a desired or expected solution.

This is shown in the next plot:

\begin{sphinxVerbatim}[commandchars=\\\{\}]
\PYG{n}{fig} \PYG{o}{=} \PYG{n}{plt}\PYG{o}{.}\PYG{n}{figure}\PYG{p}{(}\PYG{p}{)}\PYG{o}{.}\PYG{n}{gca}\PYG{p}{(}\PYG{p}{)}
\PYG{n}{pltx} \PYG{o}{=} \PYG{n}{np}\PYG{o}{.}\PYG{n}{linspace}\PYG{p}{(}\PYG{o}{\PYGZhy{}}\PYG{l+m+mi}{1}\PYG{p}{,}\PYG{l+m+mi}{1}\PYG{p}{,}\PYG{n}{N}\PYG{p}{)}

\PYG{c+c1}{\PYGZsh{} ground truth state at time=0 , move down}
\PYG{n}{INITIAL\PYGZus{}GT} \PYG{o}{=} \PYG{n}{np}\PYG{o}{.}\PYG{n}{asarray}\PYG{p}{(} \PYG{p}{[}\PYG{o}{\PYGZhy{}}\PYG{n}{np}\PYG{o}{.}\PYG{n}{sin}\PYG{p}{(}\PYG{n}{np}\PYG{o}{.}\PYG{n}{pi} \PYG{o}{*} \PYG{n}{x}\PYG{p}{)} \PYG{o}{*} \PYG{l+m+mf}{1.} \PYG{k}{for} \PYG{n}{x} \PYG{o+ow}{in} \PYG{n}{np}\PYG{o}{.}\PYG{n}{linspace}\PYG{p}{(}\PYG{o}{\PYGZhy{}}\PYG{l+m+mi}{1}\PYG{p}{,}\PYG{l+m+mi}{1}\PYG{p}{,}\PYG{n}{N}\PYG{p}{)}\PYG{p}{]} \PYG{p}{)} \PYG{c+c1}{\PYGZsh{} 1D array}
\PYG{n}{fig}\PYG{o}{.}\PYG{n}{plot}\PYG{p}{(}\PYG{n}{pltx}\PYG{p}{,} \PYG{n}{INITIAL\PYGZus{}GT}\PYG{o}{.}\PYG{n}{flatten}\PYG{p}{(}\PYG{p}{)}      \PYG{p}{,} \PYG{n}{lw}\PYG{o}{=}\PYG{l+m+mi}{2}\PYG{p}{,} \PYG{n}{color}\PYG{o}{=}\PYG{l+s+s1}{\PYGZsq{}}\PYG{l+s+s1}{forestgreen}\PYG{l+s+s1}{\PYGZsq{}}\PYG{p}{,} \PYG{n}{label}\PYG{o}{=}\PYG{l+s+s2}{\PYGZdq{}}\PYG{l+s+s2}{Ground truth initial state}\PYG{l+s+s2}{\PYGZdq{}}\PYG{p}{)}  \PYG{c+c1}{\PYGZsh{} ground truth initial state of sim}
\PYG{n}{fig}\PYG{o}{.}\PYG{n}{plot}\PYG{p}{(}\PYG{n}{pltx}\PYG{p}{,} \PYG{n}{velocity}\PYG{o}{.}\PYG{n}{values}\PYG{o}{.}\PYG{n}{numpy}\PYG{p}{(}\PYG{l+s+s1}{\PYGZsq{}}\PYG{l+s+s1}{x}\PYG{l+s+s1}{\PYGZsq{}}\PYG{p}{)}\PYG{p}{,} \PYG{n}{lw}\PYG{o}{=}\PYG{l+m+mi}{2}\PYG{p}{,} \PYG{n}{color}\PYG{o}{=}\PYG{l+s+s1}{\PYGZsq{}}\PYG{l+s+s1}{mediumblue}\PYG{l+s+s1}{\PYGZsq{}}\PYG{p}{,}  \PYG{n}{label}\PYG{o}{=}\PYG{l+s+s2}{\PYGZdq{}}\PYG{l+s+s2}{Optimized initial state}\PYG{l+s+s2}{\PYGZdq{}}\PYG{p}{)} \PYG{c+c1}{\PYGZsh{} manual}
\PYG{n}{plt}\PYG{o}{.}\PYG{n}{xlabel}\PYG{p}{(}\PYG{l+s+s1}{\PYGZsq{}}\PYG{l+s+s1}{x}\PYG{l+s+s1}{\PYGZsq{}}\PYG{p}{)}\PYG{p}{;} \PYG{n}{plt}\PYG{o}{.}\PYG{n}{ylabel}\PYG{p}{(}\PYG{l+s+s1}{\PYGZsq{}}\PYG{l+s+s1}{u}\PYG{l+s+s1}{\PYGZsq{}}\PYG{p}{)}\PYG{p}{;} \PYG{n}{plt}\PYG{o}{.}\PYG{n}{legend}\PYG{p}{(}\PYG{p}{)}\PYG{p}{;} \PYG{n}{plt}\PYG{o}{.}\PYG{n}{title}\PYG{p}{(}\PYG{l+s+s2}{\PYGZdq{}}\PYG{l+s+s2}{Initial State After 50 Optimization Steps}\PYG{l+s+s2}{\PYGZdq{}}\PYG{p}{)}\PYG{p}{;}
\end{sphinxVerbatim}

\noindent\sphinxincludegraphics{{diffphys-code-burgers_19_0}.png}

Naturally, this is a tougher task: the optimization receives direct feedback what the state at \(t=0.5\) should look like, but due to the non\sphinxhyphen{}linear model equation, we typically have a large number of solutions that exactly or numerically very closely satisfy the constraints. Hence, our minimizer does not necessarily find the exact state we started from. However, it’s still quite close in this Burgers scenario.

Before measuring the overall error of the reconstruction, let’s visualize the full evolution of our system over time as this also yields the solution in the form of a numpy array that we can compare to the other versions:

\begin{sphinxVerbatim}[commandchars=\\\{\}]
\PYG{k+kn}{import} \PYG{n+nn}{pylab}

\PYG{k}{def} \PYG{n+nf}{show\PYGZus{}state}\PYG{p}{(}\PYG{n}{a}\PYG{p}{)}\PYG{p}{:}
    \PYG{n}{a}\PYG{o}{=}\PYG{n}{np}\PYG{o}{.}\PYG{n}{expand\PYGZus{}dims}\PYG{p}{(}\PYG{n}{a}\PYG{p}{,} \PYG{n}{axis}\PYG{o}{=}\PYG{l+m+mi}{2}\PYG{p}{)}
    \PYG{k}{for} \PYG{n}{i} \PYG{o+ow}{in} \PYG{n+nb}{range}\PYG{p}{(}\PYG{l+m+mi}{4}\PYG{p}{)}\PYG{p}{:}
        \PYG{n}{a} \PYG{o}{=} \PYG{n}{np}\PYG{o}{.}\PYG{n}{concatenate}\PYG{p}{(} \PYG{p}{[}\PYG{n}{a}\PYG{p}{,}\PYG{n}{a}\PYG{p}{]} \PYG{p}{,} \PYG{n}{axis}\PYG{o}{=}\PYG{l+m+mi}{2}\PYG{p}{)}
    \PYG{n}{a} \PYG{o}{=} \PYG{n}{np}\PYG{o}{.}\PYG{n}{reshape}\PYG{p}{(} \PYG{n}{a}\PYG{p}{,} \PYG{p}{[}\PYG{n}{a}\PYG{o}{.}\PYG{n}{shape}\PYG{p}{[}\PYG{l+m+mi}{0}\PYG{p}{]}\PYG{p}{,}\PYG{n}{a}\PYG{o}{.}\PYG{n}{shape}\PYG{p}{[}\PYG{l+m+mi}{1}\PYG{p}{]}\PYG{o}{*}\PYG{n}{a}\PYG{o}{.}\PYG{n}{shape}\PYG{p}{[}\PYG{l+m+mi}{2}\PYG{p}{]}\PYG{p}{]} \PYG{p}{)}
    \PYG{n}{fig}\PYG{p}{,} \PYG{n}{axes} \PYG{o}{=} \PYG{n}{pylab}\PYG{o}{.}\PYG{n}{subplots}\PYG{p}{(}\PYG{l+m+mi}{1}\PYG{p}{,} \PYG{l+m+mi}{1}\PYG{p}{,} \PYG{n}{figsize}\PYG{o}{=}\PYG{p}{(}\PYG{l+m+mi}{16}\PYG{p}{,} \PYG{l+m+mi}{5}\PYG{p}{)}\PYG{p}{)}
    \PYG{n}{im} \PYG{o}{=} \PYG{n}{axes}\PYG{o}{.}\PYG{n}{imshow}\PYG{p}{(}\PYG{n}{a}\PYG{p}{,} \PYG{n}{origin}\PYG{o}{=}\PYG{l+s+s1}{\PYGZsq{}}\PYG{l+s+s1}{upper}\PYG{l+s+s1}{\PYGZsq{}}\PYG{p}{,} \PYG{n}{cmap}\PYG{o}{=}\PYG{l+s+s1}{\PYGZsq{}}\PYG{l+s+s1}{inferno}\PYG{l+s+s1}{\PYGZsq{}}\PYG{p}{)}
    \PYG{n}{pylab}\PYG{o}{.}\PYG{n}{colorbar}\PYG{p}{(}\PYG{n}{im}\PYG{p}{)} 
        
\PYG{c+c1}{\PYGZsh{} get numpy versions of all states }
\PYG{n}{vels} \PYG{o}{=} \PYG{p}{[} \PYG{n}{x}\PYG{o}{.}\PYG{n}{values}\PYG{o}{.}\PYG{n}{numpy}\PYG{p}{(}\PYG{l+s+s1}{\PYGZsq{}}\PYG{l+s+s1}{x,vector}\PYG{l+s+s1}{\PYGZsq{}}\PYG{p}{)} \PYG{k}{for} \PYG{n}{x} \PYG{o+ow}{in} \PYG{n}{velocities}\PYG{p}{]} 
\PYG{c+c1}{\PYGZsh{} concatenate along vector/features dimension}
\PYG{n}{vels} \PYG{o}{=} \PYG{n}{np}\PYG{o}{.}\PYG{n}{concatenate}\PYG{p}{(}\PYG{n}{vels}\PYG{p}{,} \PYG{n}{axis}\PYG{o}{=}\PYG{o}{\PYGZhy{}}\PYG{l+m+mi}{1}\PYG{p}{)} 

\PYG{c+c1}{\PYGZsh{} save for comparison with other methods}
\PYG{k+kn}{import} \PYG{n+nn}{os}\PYG{p}{;} \PYG{n}{os}\PYG{o}{.}\PYG{n}{makedirs}\PYG{p}{(}\PYG{l+s+s2}{\PYGZdq{}}\PYG{l+s+s2}{./temp}\PYG{l+s+s2}{\PYGZdq{}}\PYG{p}{,}\PYG{n}{exist\PYGZus{}ok}\PYG{o}{=}\PYG{k+kc}{True}\PYG{p}{)}
\PYG{n}{np}\PYG{o}{.}\PYG{n}{savez\PYGZus{}compressed}\PYG{p}{(}\PYG{l+s+s2}{\PYGZdq{}}\PYG{l+s+s2}{./temp/burgers\PYGZhy{}diffphys\PYGZhy{}solution.npz}\PYG{l+s+s2}{\PYGZdq{}}\PYG{p}{,} \PYG{n}{np}\PYG{o}{.}\PYG{n}{reshape}\PYG{p}{(}\PYG{n}{vels}\PYG{p}{,}\PYG{p}{[}\PYG{n}{N}\PYG{p}{,}\PYG{n}{STEPS}\PYG{o}{+}\PYG{l+m+mi}{1}\PYG{p}{]}\PYG{p}{)}\PYG{p}{)} \PYG{c+c1}{\PYGZsh{} remove batch \PYGZam{} channel dimension}

\PYG{n}{show\PYGZus{}state}\PYG{p}{(}\PYG{n}{vels}\PYG{p}{)}
\end{sphinxVerbatim}

\noindent\sphinxincludegraphics{{diffphys-code-burgers_21_0}.png}

\section{Physics\sphinxhyphen{}informed vs. differentiable physics reconstruction}
\label{\detokenize{diffphys-code-burgers:physics-informed-vs-differentiable-physics-reconstruction}}
Now we have both versions, the one with the PINN, and the DP version, so let’s compare both reconstructions in more detail. (Note: The following cells expect that the Burgers\sphinxhyphen{}forward and PINN notebooks were executed in the same environment beforehand.)

Let’s first look at the solutions side by side. The code below generates an image with 3 versions, from top to bottom: the “ground truth” (GT) solution as given by the regular forward simulation, in the middle the PINN reconstruction, and at the bottom the differentiable physics version.

\begin{sphinxVerbatim}[commandchars=\\\{\}]
\PYG{c+c1}{\PYGZsh{} note, this requires previous runs of the forward\PYGZhy{}sim \PYGZam{} PINN notebooks in the same environment}
\PYG{n}{sol\PYGZus{}gt}\PYG{o}{=}\PYG{n}{npfile}\PYG{o}{=}\PYG{n}{np}\PYG{o}{.}\PYG{n}{load}\PYG{p}{(}\PYG{l+s+s2}{\PYGZdq{}}\PYG{l+s+s2}{./temp/burgers\PYGZhy{}groundtruth\PYGZhy{}solution.npz}\PYG{l+s+s2}{\PYGZdq{}}\PYG{p}{)}\PYG{p}{[}\PYG{l+s+s2}{\PYGZdq{}}\PYG{l+s+s2}{arr\PYGZus{}0}\PYG{l+s+s2}{\PYGZdq{}}\PYG{p}{]} 
\PYG{n}{sol\PYGZus{}pi}\PYG{o}{=}\PYG{n}{npfile}\PYG{o}{=}\PYG{n}{np}\PYG{o}{.}\PYG{n}{load}\PYG{p}{(}\PYG{l+s+s2}{\PYGZdq{}}\PYG{l+s+s2}{./temp/burgers\PYGZhy{}pinn\PYGZhy{}solution.npz}\PYG{l+s+s2}{\PYGZdq{}}\PYG{p}{)}\PYG{p}{[}\PYG{l+s+s2}{\PYGZdq{}}\PYG{l+s+s2}{arr\PYGZus{}0}\PYG{l+s+s2}{\PYGZdq{}}\PYG{p}{]} 
\PYG{n}{sol\PYGZus{}dp}\PYG{o}{=}\PYG{n}{npfile}\PYG{o}{=}\PYG{n}{np}\PYG{o}{.}\PYG{n}{load}\PYG{p}{(}\PYG{l+s+s2}{\PYGZdq{}}\PYG{l+s+s2}{./temp/burgers\PYGZhy{}diffphys\PYGZhy{}solution.npz}\PYG{l+s+s2}{\PYGZdq{}}\PYG{p}{)}\PYG{p}{[}\PYG{l+s+s2}{\PYGZdq{}}\PYG{l+s+s2}{arr\PYGZus{}0}\PYG{l+s+s2}{\PYGZdq{}}\PYG{p}{]} 

\PYG{n}{divider} \PYG{o}{=} \PYG{n}{np}\PYG{o}{.}\PYG{n}{ones}\PYG{p}{(}\PYG{p}{[}\PYG{l+m+mi}{10}\PYG{p}{,}\PYG{l+m+mi}{33}\PYG{p}{]}\PYG{p}{)}\PYG{o}{*}\PYG{o}{\PYGZhy{}}\PYG{l+m+mf}{1.} \PYG{c+c1}{\PYGZsh{} we\PYGZsq{}ll sneak in a block of \PYGZhy{}1s to show a black divider in the image}
\PYG{n}{sbs} \PYG{o}{=} \PYG{n}{np}\PYG{o}{.}\PYG{n}{concatenate}\PYG{p}{(} \PYG{p}{[}\PYG{n}{sol\PYGZus{}gt}\PYG{p}{,} \PYG{n}{divider}\PYG{p}{,} \PYG{n}{sol\PYGZus{}pi}\PYG{p}{,} \PYG{n}{divider}\PYG{p}{,} \PYG{n}{sol\PYGZus{}dp}\PYG{p}{]}\PYG{p}{,} \PYG{n}{axis}\PYG{o}{=}\PYG{l+m+mi}{0}\PYG{p}{)}

\PYG{n+nb}{print}\PYG{p}{(}\PYG{l+s+s2}{\PYGZdq{}}\PYG{l+s+se}{\PYGZbs{}n}\PYG{l+s+s2}{Solutions Ground Truth (top), PINN (middle) , DiffPhys (bottom):}\PYG{l+s+s2}{\PYGZdq{}}\PYG{p}{)}
\PYG{n}{show\PYGZus{}state}\PYG{p}{(}\PYG{n}{np}\PYG{o}{.}\PYG{n}{reshape}\PYG{p}{(}\PYG{n}{sbs}\PYG{p}{,}\PYG{p}{[}\PYG{n}{N}\PYG{o}{*}\PYG{l+m+mi}{3}\PYG{o}{+}\PYG{l+m+mi}{20}\PYG{p}{,}\PYG{l+m+mi}{33}\PYG{p}{,}\PYG{l+m+mi}{1}\PYG{p}{]}\PYG{p}{)}\PYG{p}{)}
\end{sphinxVerbatim}

\begin{sphinxVerbatim}[commandchars=\\\{\}]
Solutions Ground Truth (top), PINN (middle) , DiffPhys (bottom):
\end{sphinxVerbatim}

\noindent\sphinxincludegraphics{{diffphys-code-burgers_23_1}.png}

It’s quite clearly visible here that the PINN solution (in the middle) recovers the overall shape of the solution, hence the temporal constraints are at least partially fulfilled. However, it doesn’t manage to capture the amplitudes of the GT solution very well.

The reconstruction from the optimization with a differentiable solver (at the bottom) is much closer to the ground truth thanks to an improved flow of gradients over the whole course of the sequence. In addition, it can leverage the grid\sphinxhyphen{}based discretization for both forward as well as backward passes, and in this way provide a more accurate signal to the unknown initial state. It is nonetheless visible that the reconstruction lacks certain “sharper” features of the GT version, e.g., visible in the bottom left corner of the solution image.

Let’s quantify these errors over the whole sequence:

\begin{sphinxVerbatim}[commandchars=\\\{\}]
\PYG{n}{err\PYGZus{}pi} \PYG{o}{=} \PYG{n}{np}\PYG{o}{.}\PYG{n}{sum}\PYG{p}{(} \PYG{n}{np}\PYG{o}{.}\PYG{n}{abs}\PYG{p}{(}\PYG{n}{sol\PYGZus{}pi}\PYG{o}{\PYGZhy{}}\PYG{n}{sol\PYGZus{}gt}\PYG{p}{)}\PYG{p}{)} \PYG{o}{/} \PYG{p}{(}\PYG{n}{STEPS}\PYG{o}{*}\PYG{n}{N}\PYG{p}{)}
\PYG{n}{err\PYGZus{}dp} \PYG{o}{=} \PYG{n}{np}\PYG{o}{.}\PYG{n}{sum}\PYG{p}{(} \PYG{n}{np}\PYG{o}{.}\PYG{n}{abs}\PYG{p}{(}\PYG{n}{sol\PYGZus{}dp}\PYG{o}{\PYGZhy{}}\PYG{n}{sol\PYGZus{}gt}\PYG{p}{)}\PYG{p}{)} \PYG{o}{/} \PYG{p}{(}\PYG{n}{STEPS}\PYG{o}{*}\PYG{n}{N}\PYG{p}{)}
\PYG{n+nb}{print}\PYG{p}{(}\PYG{l+s+s2}{\PYGZdq{}}\PYG{l+s+s2}{MAE PINN: }\PYG{l+s+si}{\PYGZob{}:7.5f\PYGZcb{}}\PYG{l+s+s2}{ }\PYG{l+s+se}{\PYGZbs{}n}\PYG{l+s+s2}{MAE DP:   }\PYG{l+s+si}{\PYGZob{}:7.5f\PYGZcb{}}\PYG{l+s+s2}{\PYGZdq{}}\PYG{o}{.}\PYG{n}{format}\PYG{p}{(}\PYG{n}{err\PYGZus{}pi}\PYG{p}{,}\PYG{n}{err\PYGZus{}dp}\PYG{p}{)}\PYG{p}{)}

\PYG{n+nb}{print}\PYG{p}{(}\PYG{l+s+s2}{\PYGZdq{}}\PYG{l+s+se}{\PYGZbs{}n}\PYG{l+s+s2}{Error GT to PINN (top) , DiffPhys (bottom):}\PYG{l+s+s2}{\PYGZdq{}}\PYG{p}{)}
\PYG{n}{show\PYGZus{}state}\PYG{p}{(}\PYG{n}{np}\PYG{o}{.}\PYG{n}{reshape}\PYG{p}{(} \PYG{n}{np}\PYG{o}{.}\PYG{n}{concatenate}\PYG{p}{(}\PYG{p}{[}\PYG{n}{sol\PYGZus{}pi}\PYG{o}{\PYGZhy{}}\PYG{n}{sol\PYGZus{}gt}\PYG{p}{,} \PYG{n}{divider}\PYG{p}{,} \PYG{n}{sol\PYGZus{}dp}\PYG{o}{\PYGZhy{}}\PYG{n}{sol\PYGZus{}gt}\PYG{p}{]}\PYG{p}{,}\PYG{n}{axis}\PYG{o}{=}\PYG{l+m+mi}{0}\PYG{p}{)} \PYG{p}{,}\PYG{p}{[}\PYG{n}{N}\PYG{o}{*}\PYG{l+m+mi}{2}\PYG{o}{+}\PYG{l+m+mi}{10}\PYG{p}{,}\PYG{l+m+mi}{33}\PYG{p}{,}\PYG{l+m+mi}{1}\PYG{p}{]}\PYG{p}{)}\PYG{p}{)}
\end{sphinxVerbatim}

\begin{sphinxVerbatim}[commandchars=\\\{\}]
MAE PINN: 0.22991 
MAE DP:   0.05811

Error GT to PINN (top) , DiffPhys (bottom):
\end{sphinxVerbatim}

\noindent\sphinxincludegraphics{{diffphys-code-burgers_25_1}.png}

That’s a pretty clear result: the PINN error is almost 4 times higher than the one from the Differentiable Physics (DP) reconstruction.

This difference also shows clearly in the jointly visualized image at the bottom: the magnitudes of the errors of the DP reconstruction are much closer to zero, as indicated by the purple color above.

A simple direct reconstruction problem like this one is always a good initial test for a DP solver. It can be tested independently before moving on to more complex setups, e.g., coupling it with an NN. If the direct optimization does not converge, there’s probably still something fundamentally wrong, and there’s no point involving an NN.

Now we have a first example to show similarities and differences of the two approaches. In the next section, we’ll present a discussion of the findings so far, before moving to more complex cases in the following chapter.

\section{Next steps}
\label{\detokenize{diffphys-code-burgers:next-steps}}
As with the PINN version, there’s variety of things that can be improved and experimented with using the code above:
\begin{itemize}
\item {} 
You can try to adjust the training parameters to further improve the reconstruction.

\item {} 
As for the PINN case, you can activate a different optimizer, and observe the changing (not necessarily improved) convergence behavior.

\item {} 
Vary the number of steps, or the resolution of the simulation and reconstruction.

\end{itemize}

\chapter{Discussion}
\label{\detokenize{diffphys-discuss:discussion}}\label{\detokenize{diffphys-discuss::doc}}
To summarize, the training via differentiable physics (DP) as described so far allows us
to integrate full numerical simulations into the training of deep neural networks.
As a consequence, this let’s the networks learn to \sphinxstyleemphasis{interact} with these simulations.
While we’ve only hinted at what could be
achieved via DP approaches it is nonetheless a good time to discuss some
additional properties, and summarize the pros and cons.

\sphinxincludegraphics{{divider4}.jpg}

\section{Time steps and iterations}
\label{\detokenize{diffphys-discuss:time-steps-and-iterations}}
When using DP approaches for learning applications,
there is a lot of flexibility w.r.t. the combination of DP and NN building blocks.
As some of the differences are subtle, the following section will go into more detail.
We’ll especially focus on solvers that repeat the PDE and NN evaluations multiple times,
e.g., to compute multiple states of the physical system over time.

To re\sphinxhyphen{}cap, here’s the previous figure about combining NNs and DP operators.
In the figure these operators look like a loss term: they typically don’t have weights,
and only provide a gradient that influences the optimization of the NN weights:

\begin{figure}[htbp]
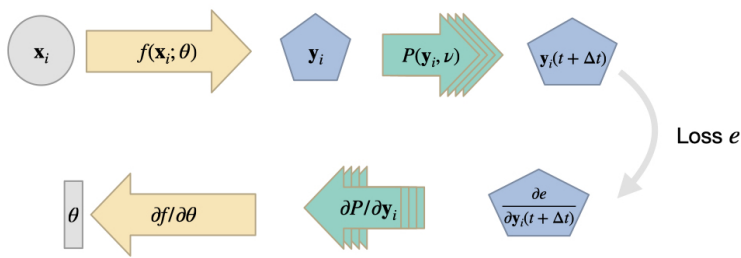

\centering
\capstart

\noindent\sphinxincludegraphics[height=220\sphinxpxdimen]{{diffphys-shortened}.jpg}
\caption{The DP approach as described in the previous chapters. A network produces an input to a PDE solver \(\mathcal P\), which provides a gradient for training during the backpropagation step.}\label{\detokenize{diffphys-discuss:diffphys-short}}\end{figure}

This setup can be seen as the network receiving information about how it’s output influences the outcome of the PDE solver. I.e., the gradient will provide information how to produce an NN output that minimizes the loss.
Similar to the previously described \sphinxstyleemphasis{physical losses} (from {\hyperref[\detokenize{physicalloss::doc}]{\sphinxcrossref{\DUrole{doc}{Physical Loss Terms}}}}), this can mean upholding a conservation law.

\sphinxstylestrong{Switching the Order}

However, with DP, there’s no real reason to be limited to this setup. E.g., we could imagine a swap of the NN and DP components, giving the following structure:

\begin{figure}[htbp]
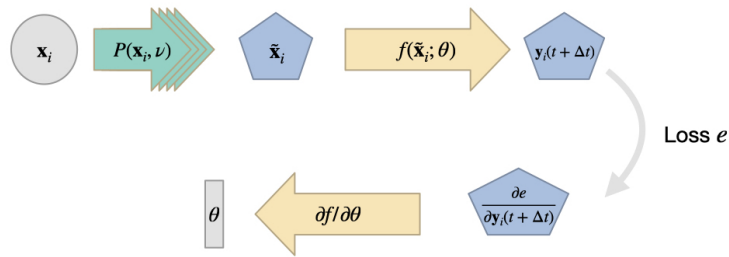

\centering
\capstart

\noindent\sphinxincludegraphics[height=220\sphinxpxdimen]{{diffphys-switched}.jpg}
\caption{A PDE solver produces an output which is processed by an NN.}\label{\detokenize{diffphys-discuss:diffphys-switch}}\end{figure}

In this case the PDE solver essentially represents an \sphinxstyleemphasis{on\sphinxhyphen{}the\sphinxhyphen{}fly} data generator. This is not necessarily always useful: this setup could be replaced by a pre\sphinxhyphen{}computation of the same inputs, as the PDE solver is not influenced by the NN. Hence, we could replace the \(\mathcal P\) invocations by a “loading” function. On the other hand, evaluating the PDE solver at training time with a randomized sampling of the parameter domain of interest can lead to an excellent sampling of the data distribution of the input, and hence yield accurate and stable NNs. If done correctly, the solver can alleviate the need to store and load large amounts of data, and instead produce them more quickly at training time, e.g., directly on a GPU.

\sphinxstylestrong{Time Stepping}

In general, there’s no combination of NN layers and DP operators that is \sphinxstyleemphasis{forbidden} (as long as their dimensions are compatible). One that makes particular sense is to “unroll” the iterations of a time stepping process of a simulator, and let the state of a system be influenced by an NN.

In this case we compute a (potentially very long) sequence of PDE solver steps in the forward pass. In\sphinxhyphen{}between these solver steps, an NN modifies the state of our system, which is then used to compute the next PDE solver step. During the backpropagation pass, we move backwards through all of these steps to evaluate contributions to the loss function (it can be evaluated in one or more places anywhere in the execution chain), and to backpropagte the gradient information through the DP and NN operators. This unrolling of solver iterations essentially gives feedback to the NN about how it’s “actions” influence the state of the physical system and resulting loss. Due to the iterative nature of this process, many errors increase exponentially over the course of iterations, and are extremely difficult to detect in a single evaluation. In these cases it is crucial to provide feedback to the NN at training time who the errors evolve over course of the iterations. Note that in this case, a pre\sphinxhyphen{}computation of the states is not possible, as the iterations depend on the state of the NN, which is unknown before training. Hence, a DP\sphinxhyphen{}based training is crucial to evaluate the correct gradient information at training time.

\begin{figure}[htbp]
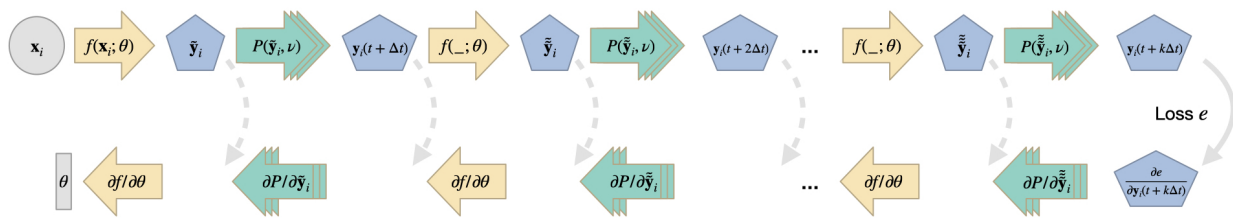

\centering
\capstart

\noindent\sphinxincludegraphics[height=180\sphinxpxdimen]{{diffphys-multistep}.jpg}
\caption{Time stepping with interleaved DP and NN operations for \(k\) solver iterations. The dashed gray arrows indicate optional intermediate evaluations of loss terms (similar to the solid gray arrow for the last step \(k\)).}\label{\detokenize{diffphys-discuss:diffphys-mulitstep}}\end{figure}

Note that in this picture (and the ones before) we have assumed an \sphinxstyleemphasis{additive} influence of the NN. Of course, any differentiable operator could be used here to integrate the NN result into the state of the PDE. E.g., multiplicative modifications can be more suitable in certain settings, or in others the NN could modify the parameters of the PDE in addition to or instead of the state space. Likewise, the loss function is problem dependent and can be computed in different ways.

DP setups with many time steps can be difficult to train: the gradients need to backpropagate through the full chain of PDE solver evaluations and NN evaluations. Typically, each of them represents a non\sphinxhyphen{}linear and complex function. Hence for larger numbers of steps, the vanishing and exploding gradient problem can make training difficult (see {\hyperref[\detokenize{diffphys-code-sol::doc}]{\sphinxcrossref{\DUrole{doc}{Reducing Numerical Errors with Deep Learning}}}} for some practical tips how to alleviate this).

\section{Alternatives: noise}
\label{\detokenize{diffphys-discuss:alternatives-noise}}
It is worth mentioning here that other works have proposed perturbing the inputs and
the iterations at training time with noise {[}\hyperlink{cite.references:id88}{SGGP+20}{]} (somewhat similar to
regularizers like dropout).
This can help to prevent overfitting to the training states, and hence shares similarities
with the goals of training with DP.

However, the noise is typically undirected, and hence not as accurate as training with
the actual evolutions of simulations. Hence, this noise can be a good starting point
for training that tends to overfit, but if possible, it is preferable to incorporate the
actual solver in the training loop via a DP approach.

\sphinxincludegraphics{{divider5}.jpg}

\section{Summary}
\label{\detokenize{diffphys-discuss:summary}}
To summarize the pros and cons of training NNs via DP:

✅ Pro:
\begin{itemize}
\item {} 
Uses physical model and numerical methods for discretization.

\item {} 
Efficiency and accuracy of selected methods carries over to training.

\item {} 
Very tight coupling of physical models and NNs possible.

\end{itemize}

❌ Con:
\begin{itemize}
\item {} 
Not compatible with all simulators (need to provide gradients).

\item {} 
Require more heavy machinery (in terms of framework support) than previously discussed methods.

\end{itemize}

Here, the last negative point (regarding heavy machinery) is bound to strongly improve in a fairly short amount of time. However, for now it’s important to keep in mind that not every simulator is suitable for DP training out of the box. Hence, in this book we’ll focus on examples using phiflow, which was designed for interfacing with deep learning frameworks.

Next we can target more some complex scenarios to showcase what can be achieved with differentiable physics.
This will also illustrate how the right selection of numerical methods for a DP operator yields improvements in terms of training accuracy.

\chapter{Differentiable Fluid Simulations}
\label{\detokenize{diffphys-code-ns:differentiable-fluid-simulations}}\label{\detokenize{diffphys-code-ns::doc}}
Next, we’ll target a more complex example with the Navier\sphinxhyphen{}Stokes equations as physical model. In line with {\hyperref[\detokenize{overview-ns-forw::doc}]{\sphinxcrossref{\DUrole{doc}{Navier\sphinxhyphen{}Stokes Forward Simulation}}}}, we’ll target a 2D case.

As optimization objective we’ll consider a more difficult variant of the previous Burgers example: the state of the observed density \(s\) should match a given target after \(n=20\) steps of simulation. In contrast to before, the observed quantity in the form of the marker field \(s\) cannot be changed in any way. Only the initial state of the velocity \(\mathbf{u}_0\) at \(t=0\) can be modified. This gives us a split between observable quantities for the loss formulation and quantities that we can interact with during the optimization (or later on via NNs).
\sphinxhref{https://colab.research.google.com/github/tum-pbs/pbdl-book/blob/main/diffphys-code-ns.ipynb}{{[}run in colab{]}}

\section{Physical Model}
\label{\detokenize{diffphys-code-ns:physical-model}}
We’ll use a Navier\sphinxhyphen{}Stokes model with velocity \(\mathbf{u}\), no explicit viscosity term, and a smoke marker density \(s\) that drives a simple Boussinesq buoyancy term \(\eta d\) adding a force along the y dimension. For the velocity this gives:
\begin{equation*}
\begin{split}\begin{aligned}
  \frac{\partial u_x}{\partial{t}} + \mathbf{u} \cdot \nabla u_x &= - \frac{1}{\rho} \nabla p 
  \\
  \frac{\partial u_y}{\partial{t}} + \mathbf{u} \cdot \nabla u_y &= - \frac{1}{\rho} \nabla p + \eta d
  \\
  \text{s.t.} \quad \nabla \cdot \mathbf{u} &= 0,
\end{aligned}\end{split}
\end{equation*}
With an additional transport equation for the passively advected marker density \(s\):
\begin{equation*}
\begin{split}\begin{aligned}
  \frac{\partial s}{\partial{t}} + \mathbf{u} \cdot \nabla s &= 0 
\end{aligned}\end{split}
\end{equation*}

\section{Formulation}
\label{\detokenize{diffphys-code-ns:formulation}}
With the notation from {\hyperref[\detokenize{overview-equations::doc}]{\sphinxcrossref{\DUrole{doc}{Models and Equations}}}} the inverse problem outlined above can be formulated as a  minimization problem
\begin{equation*}
\begin{split}
\text{arg min}_{\mathbf{u}_{0}} \sum_i \big( f(x_{t_e,i} ; \mathbf{u}_{0} )-y^*_{t_e,i} \big)^2 ,
\end{split}
\end{equation*}
where \(y^*_{t_e,i}\) are samples of the reference solution at a targeted time \(t_e\),
and \(x_{t_e,i}\) denotes the estimate of our simulator at the same
sampling locations and time.
The index \(i\) here runs over all discretized, spatial degrees of freedom in our fluid solver (we’ll have \(32\times40\) below).

In contrast to before, we are not dealing with pre\sphinxhyphen{}computed quantities anymore, but now \(x_{t_e,i}\) is a complex, non\sphinxhyphen{}linear function itself.
More specifically, the simulator starts with the initial velocity \(\mathbf{u}_0\) and density \(s_0\) to compute the \(x_{t_e,i}\), by \(n\) evaluations of the discretized PDE \(\mathcal{P}\).
This gives as simulated final state
\(y_{t_e,i} = s_{t_e} = \mathcal{P}^n(\mathbf{u}_0,s_0)\), where we will leave \(s_0\) fixed in the following, and focus on \(\mathbf{u}_0\) as our degrees of freedom.
Hence, the optimization can only change \(\mathbf{u}_0\) to align \(y_{t_e,i}\) with the references \(y^*_{t_e,i}\) as closely as possible.

\section{Starting the Implementation}
\label{\detokenize{diffphys-code-ns:starting-the-implementation}}
First, let’s get the loading of python modules out of the way. By importing \sphinxcode{\sphinxupquote{phi.torch.flow}}, we get fluid simulation functions that work within pytorch graphs and can provide gradients (\sphinxcode{\sphinxupquote{phi.tf.flow}} would be the alternative for tensorflow).

\begin{sphinxVerbatim}[commandchars=\\\{\}]
\PYG{c+ch}{\PYGZsh{}!pip install \PYGZhy{}\PYGZhy{}upgrade \PYGZhy{}\PYGZhy{}quiet phiflow}
\PYG{o}{!}pip install \PYGZhy{}\PYGZhy{}upgrade \PYGZhy{}\PYGZhy{}quiet git+https://github.com/tum\PYGZhy{}pbs/PhiFlow@develop

\PYG{k+kn}{from} \PYG{n+nn}{phi}\PYG{n+nn}{.}\PYG{n+nn}{torch}\PYG{n+nn}{.}\PYG{n+nn}{flow} \PYG{k+kn}{import} \PYG{o}{*}  
\PYG{k+kn}{import} \PYG{n+nn}{pylab} \PYG{c+c1}{\PYGZsh{} for visualizations later on}
\end{sphinxVerbatim}

\begin{sphinxVerbatim}[commandchars=\\\{\}]

\end{sphinxVerbatim}

\section{Batched simulations}
\label{\detokenize{diffphys-code-ns:batched-simulations}}
Now we can set up the simulation, which will work in line with the previous “regular” simulation example from the {\hyperref[\detokenize{overview-ns-forw::doc}]{\sphinxcrossref{\DUrole{doc}{Navier\sphinxhyphen{}Stokes Forward Simulation}}}}. However, now we’ll directly include an additional dimension, similar to a mini\sphinxhyphen{}batch used for NN training. For this, we’ll introduce a named dimension called \sphinxcode{\sphinxupquote{inflow\_loc}}. This dimension will exist “above” the previous spatial dimensions \sphinxcode{\sphinxupquote{y}}, \sphinxcode{\sphinxupquote{x}} and the channel dimensions \sphinxcode{\sphinxupquote{vector}}. As indicated by the name \sphinxcode{\sphinxupquote{inflow\_loc}}, the main differences for this dimension will lie in different locations of the inflow, in order to obtain different flow simulations. The named dimensions in phiflow make it very convenient to broadcast information across matching dimensions in different tensors.

The \sphinxcode{\sphinxupquote{Domain}} object is allocated just like before, but the \sphinxcode{\sphinxupquote{INFLOW\_LOCATION}} tensor now receives a string
\sphinxcode{\sphinxupquote{'inflow\_loc,vector'}} that indicates the names of the two dimensions. This leads to the creation of an \sphinxcode{\sphinxupquote{inflow\_loc}} dimensions in addition to the two spatial dimensions (the \sphinxcode{\sphinxupquote{vector}} part).

\begin{sphinxVerbatim}[commandchars=\\\{\}]
\PYG{c+c1}{\PYGZsh{} closed domain}
\PYG{n}{INFLOW\PYGZus{}LOCATION} \PYG{o}{=} \PYG{n}{tensor}\PYG{p}{(}\PYG{p}{[}\PYG{p}{(}\PYG{l+m+mi}{12}\PYG{p}{,} \PYG{l+m+mi}{4}\PYG{p}{)}\PYG{p}{,} \PYG{p}{(}\PYG{l+m+mi}{13}\PYG{p}{,} \PYG{l+m+mi}{6}\PYG{p}{)}\PYG{p}{,} \PYG{p}{(}\PYG{l+m+mi}{14}\PYG{p}{,} \PYG{l+m+mi}{5}\PYG{p}{)}\PYG{p}{,} \PYG{p}{(}\PYG{l+m+mi}{16}\PYG{p}{,} \PYG{l+m+mi}{5}\PYG{p}{)}\PYG{p}{]}\PYG{p}{,} \PYG{n}{batch}\PYG{p}{(}\PYG{l+s+s1}{\PYGZsq{}}\PYG{l+s+s1}{inflow\PYGZus{}loc}\PYG{l+s+s1}{\PYGZsq{}}\PYG{p}{)}\PYG{p}{,} \PYG{n}{channel}\PYG{p}{(}\PYG{l+s+s1}{\PYGZsq{}}\PYG{l+s+s1}{vector}\PYG{l+s+s1}{\PYGZsq{}}\PYG{p}{)}\PYG{p}{)}
\PYG{n}{INFLOW} \PYG{o}{=} \PYG{p}{(}\PYG{l+m+mf}{1.}\PYG{o}{/}\PYG{l+m+mf}{3.}\PYG{p}{)} \PYG{o}{*} \PYG{n}{CenteredGrid}\PYG{p}{(}\PYG{n}{Sphere}\PYG{p}{(}\PYG{n}{center}\PYG{o}{=}\PYG{n}{INFLOW\PYGZus{}LOCATION}\PYG{p}{,} \PYG{n}{radius}\PYG{o}{=}\PYG{l+m+mi}{3}\PYG{p}{)}\PYG{p}{,} \PYG{n}{extrapolation}\PYG{o}{.}\PYG{n}{BOUNDARY}\PYG{p}{,} \PYG{n}{x}\PYG{o}{=}\PYG{l+m+mi}{32}\PYG{p}{,} \PYG{n}{y}\PYG{o}{=}\PYG{l+m+mi}{40}\PYG{p}{,} \PYG{n}{bounds}\PYG{o}{=}\PYG{n}{Box}\PYG{p}{[}\PYG{l+m+mi}{0}\PYG{p}{:}\PYG{l+m+mi}{32}\PYG{p}{,} \PYG{l+m+mi}{0}\PYG{p}{:}\PYG{l+m+mi}{40}\PYG{p}{]}\PYG{p}{)}
\PYG{n}{BND} \PYG{o}{=} \PYG{n}{extrapolation}\PYG{o}{.}\PYG{n}{ZERO} \PYG{c+c1}{\PYGZsh{} closed, boundary conditions for velocity grid below}

\PYG{c+c1}{\PYGZsh{} uncomment this for a slightly different open domain case}
\PYG{c+c1}{\PYGZsh{}INFLOW\PYGZus{}LOCATION = tensor([(11, 6), (12, 4), (14, 5), (16, 5)], batch(\PYGZsq{}inflow\PYGZus{}loc\PYGZsq{}), channel(\PYGZsq{}vector\PYGZsq{}))}
\PYG{c+c1}{\PYGZsh{}INFLOW = (1./4.) * CenteredGrid(Sphere(center=INFLOW\PYGZus{}LOCATION, radius=3), extrapolation.BOUNDARY, x=32, y=40, bounds=Box[0:32, 0:40])}
\PYG{c+c1}{\PYGZsh{}BND = extrapolation.BOUNDARY \PYGZsh{} open boundaries }

\PYG{n}{INFLOW}\PYG{o}{.}\PYG{n}{shape}
\end{sphinxVerbatim}

\begin{sphinxVerbatim}[commandchars=\\\{\}]
(inflow\PYGZus{}locᵇ=4, xˢ=32, yˢ=40)
\end{sphinxVerbatim}

The last statement verifies that our \sphinxcode{\sphinxupquote{INFLOW}} grid likewise has an \sphinxcode{\sphinxupquote{inflow\_loc}} dimension in addition to the spatial dimensions \sphinxcode{\sphinxupquote{x}} and \sphinxcode{\sphinxupquote{y}}. You can test for the existence of a tensor dimension in phiflow with the \sphinxcode{\sphinxupquote{.exists}} boolean, which can be evaluated for any dimension name. E.g., above \sphinxcode{\sphinxupquote{INFLOW.inflow\_loc.exists}} will give \sphinxcode{\sphinxupquote{True}}, while \sphinxcode{\sphinxupquote{INFLOW.some\_unknown\_dim.exists}} will give \sphinxcode{\sphinxupquote{False}}. The \(^b\) superscript indicates that \sphinxcode{\sphinxupquote{inflow\_loc}} is a batch dimension.

Phiflow tensors are automatically broadcast to new dimensions via their names, and hence typically no\sphinxhyphen{}reshaping operations are required. E.g., you can easily add or multiply tensors with differing dimensions. Below we’ll multiply a staggered grid with a tensor of ones along the \sphinxcode{\sphinxupquote{inflow\_loc}} dimension to get a staggered velocity that has \sphinxcode{\sphinxupquote{x,y,inflow\_loc}} as dimensions via \sphinxcode{\sphinxupquote{StaggeredGrid(...) * math.ones(batch(inflow\_loc=4))}}.

We can easily simulate a few steps now starting with these different initial conditions. Thanks to the broadcasting, the exact same code we used for the single forward simulation in the overview chapter will produce four simulations with different smoke inflow positions.

\begin{sphinxVerbatim}[commandchars=\\\{\}]
\PYG{n}{smoke} \PYG{o}{=} \PYG{n}{CenteredGrid}\PYG{p}{(}\PYG{l+m+mi}{0}\PYG{p}{,} \PYG{n}{extrapolation}\PYG{o}{.}\PYG{n}{BOUNDARY}\PYG{p}{,} \PYG{n}{x}\PYG{o}{=}\PYG{l+m+mi}{32}\PYG{p}{,} \PYG{n}{y}\PYG{o}{=}\PYG{l+m+mi}{40}\PYG{p}{,} \PYG{n}{bounds}\PYG{o}{=}\PYG{n}{Box}\PYG{p}{[}\PYG{l+m+mi}{0}\PYG{p}{:}\PYG{l+m+mi}{32}\PYG{p}{,} \PYG{l+m+mi}{0}\PYG{p}{:}\PYG{l+m+mi}{40}\PYG{p}{]}\PYG{p}{)}  \PYG{c+c1}{\PYGZsh{} sampled at cell centers}
\PYG{n}{velocity} \PYG{o}{=} \PYG{n}{StaggeredGrid}\PYG{p}{(}\PYG{l+m+mi}{0}\PYG{p}{,} \PYG{n}{BND}\PYG{p}{,} \PYG{n}{x}\PYG{o}{=}\PYG{l+m+mi}{32}\PYG{p}{,} \PYG{n}{y}\PYG{o}{=}\PYG{l+m+mi}{40}\PYG{p}{,} \PYG{n}{bounds}\PYG{o}{=}\PYG{n}{Box}\PYG{p}{[}\PYG{l+m+mi}{0}\PYG{p}{:}\PYG{l+m+mi}{32}\PYG{p}{,} \PYG{l+m+mi}{0}\PYG{p}{:}\PYG{l+m+mi}{40}\PYG{p}{]}\PYG{p}{)}  \PYG{c+c1}{\PYGZsh{} sampled in staggered form at face centers }

\PYG{k}{def} \PYG{n+nf}{step}\PYG{p}{(}\PYG{n}{smoke}\PYG{p}{,} \PYG{n}{velocity}\PYG{p}{)}\PYG{p}{:}
  \PYG{n}{smoke} \PYG{o}{=} \PYG{n}{advect}\PYG{o}{.}\PYG{n}{mac\PYGZus{}cormack}\PYG{p}{(}\PYG{n}{smoke}\PYG{p}{,} \PYG{n}{velocity}\PYG{p}{,} \PYG{n}{dt}\PYG{o}{=}\PYG{l+m+mi}{1}\PYG{p}{)} \PYG{o}{+} \PYG{n}{INFLOW}
  \PYG{n}{buoyancy\PYGZus{}force} \PYG{o}{=} \PYG{n}{smoke} \PYG{o}{*} \PYG{p}{(}\PYG{l+m+mi}{0}\PYG{p}{,} \PYG{l+m+mi}{1}\PYG{p}{)} \PYG{o}{\PYGZgt{}\PYGZgt{}} \PYG{n}{velocity}
  \PYG{n}{velocity} \PYG{o}{=} \PYG{n}{advect}\PYG{o}{.}\PYG{n}{semi\PYGZus{}lagrangian}\PYG{p}{(}\PYG{n}{velocity}\PYG{p}{,} \PYG{n}{velocity}\PYG{p}{,} \PYG{n}{dt}\PYG{o}{=}\PYG{l+m+mi}{1}\PYG{p}{)} \PYG{o}{+} \PYG{n}{buoyancy\PYGZus{}force}
  \PYG{n}{velocity}\PYG{p}{,} \PYG{n}{\PYGZus{}} \PYG{o}{=} \PYG{n}{fluid}\PYG{o}{.}\PYG{n}{make\PYGZus{}incompressible}\PYG{p}{(}\PYG{n}{velocity}\PYG{p}{)}
  \PYG{k}{return} \PYG{n}{smoke}\PYG{p}{,} \PYG{n}{velocity}

\PYG{k}{for} \PYG{n}{\PYGZus{}} \PYG{o+ow}{in} \PYG{n+nb}{range}\PYG{p}{(}\PYG{l+m+mi}{20}\PYG{p}{)}\PYG{p}{:}
    \PYG{n}{smoke}\PYG{p}{,}\PYG{n}{velocity} \PYG{o}{=} \PYG{n}{step}\PYG{p}{(}\PYG{n}{smoke}\PYG{p}{,}\PYG{n}{velocity}\PYG{p}{)}

\PYG{c+c1}{\PYGZsh{} store and show final states (before optimization)}
\PYG{n}{smoke\PYGZus{}final} \PYG{o}{=} \PYG{n}{smoke} 
\PYG{n}{fig}\PYG{p}{,} \PYG{n}{axes} \PYG{o}{=} \PYG{n}{pylab}\PYG{o}{.}\PYG{n}{subplots}\PYG{p}{(}\PYG{l+m+mi}{1}\PYG{p}{,} \PYG{l+m+mi}{4}\PYG{p}{,} \PYG{n}{figsize}\PYG{o}{=}\PYG{p}{(}\PYG{l+m+mi}{10}\PYG{p}{,} \PYG{l+m+mi}{6}\PYG{p}{)}\PYG{p}{)}
\PYG{k}{for} \PYG{n}{i} \PYG{o+ow}{in} \PYG{n+nb}{range}\PYG{p}{(}\PYG{n}{INFLOW}\PYG{o}{.}\PYG{n}{shape}\PYG{o}{.}\PYG{n}{get\PYGZus{}size}\PYG{p}{(}\PYG{l+s+s1}{\PYGZsq{}}\PYG{l+s+s1}{inflow\PYGZus{}loc}\PYG{l+s+s1}{\PYGZsq{}}\PYG{p}{)}\PYG{p}{)}\PYG{p}{:}
  \PYG{n}{axes}\PYG{p}{[}\PYG{n}{i}\PYG{p}{]}\PYG{o}{.}\PYG{n}{imshow}\PYG{p}{(}\PYG{n}{smoke\PYGZus{}final}\PYG{o}{.}\PYG{n}{values}\PYG{o}{.}\PYG{n}{numpy}\PYG{p}{(}\PYG{l+s+s1}{\PYGZsq{}}\PYG{l+s+s1}{inflow\PYGZus{}loc,y,x}\PYG{l+s+s1}{\PYGZsq{}}\PYG{p}{)}\PYG{p}{[}\PYG{n}{i}\PYG{p}{,}\PYG{o}{.}\PYG{o}{.}\PYG{o}{.}\PYG{p}{]}\PYG{p}{,} \PYG{n}{origin}\PYG{o}{=}\PYG{l+s+s1}{\PYGZsq{}}\PYG{l+s+s1}{lower}\PYG{l+s+s1}{\PYGZsq{}}\PYG{p}{,} \PYG{n}{cmap}\PYG{o}{=}\PYG{l+s+s1}{\PYGZsq{}}\PYG{l+s+s1}{magma}\PYG{l+s+s1}{\PYGZsq{}}\PYG{p}{)}
  \PYG{n}{axes}\PYG{p}{[}\PYG{n}{i}\PYG{p}{]}\PYG{o}{.}\PYG{n}{set\PYGZus{}title}\PYG{p}{(}\PYG{l+s+sa}{f}\PYG{l+s+s2}{\PYGZdq{}}\PYG{l+s+s2}{Inflow }\PYG{l+s+si}{\PYGZob{}}\PYG{n}{INFLOW\PYGZus{}LOCATION}\PYG{o}{.}\PYG{n}{numpy}\PYG{p}{(}\PYG{l+s+s1}{\PYGZsq{}}\PYG{l+s+s1}{inflow\PYGZus{}loc,vector}\PYG{l+s+s1}{\PYGZsq{}}\PYG{p}{)}\PYG{p}{[}\PYG{n}{i}\PYG{p}{]}\PYG{l+s+si}{\PYGZcb{}}\PYG{l+s+s2}{\PYGZdq{}} \PYG{o}{+} \PYG{p}{(}\PYG{l+s+s2}{\PYGZdq{}}\PYG{l+s+s2}{, Reference}\PYG{l+s+s2}{\PYGZdq{}} \PYG{k}{if} \PYG{n}{i}\PYG{o}{==}\PYG{l+m+mi}{3} \PYG{k}{else} \PYG{l+s+s2}{\PYGZdq{}}\PYG{l+s+s2}{\PYGZdq{}}\PYG{p}{)}\PYG{p}{)}
\PYG{n}{pylab}\PYG{o}{.}\PYG{n}{tight\PYGZus{}layout}\PYG{p}{(}\PYG{p}{)}
\end{sphinxVerbatim}

\noindent\sphinxincludegraphics{{diffphys-code-ns_8_0}.png}

The last image shows the state of the advected smoke fields after 20 simulation steps. The final smoke shape of simulation \sphinxcode{\sphinxupquote{{[}3{]}}} with an inflow at \sphinxcode{\sphinxupquote{(16,5)}}, with the straight plume on the far right, will be our \sphinxstylestrong{reference state} below. The initial velocity of the other three will be modified in the optimization procedure below to match this reference.

(As a small side note: phiflow tensors will keep track of their chain of operations using the backend they were created for. E.g. a tensor created with NumPy will keep using NumPy/SciPy operations unless a PyTorch or TensorFlow tensor is also passed to the same operation. Thus, it is a good idea to verify that tensors are using the right backend once in a while, e.g., via \sphinxcode{\sphinxupquote{GRID.values.default\_backend}}.)

\section{Gradients}
\label{\detokenize{diffphys-code-ns:gradients}}
Let’s look at how to get gradients from our simulation. The first trivial step taken care of above was to include \sphinxcode{\sphinxupquote{phi.torch.flow}} to import differentiable operators from which to build our simulator.

Now we want to optimize the initial velocities so that all simulations arrive at a final state that is similar to the simulation on the right, where the inflow is located at \sphinxcode{\sphinxupquote{(16, 5)}}, i.e. centered along \sphinxcode{\sphinxupquote{x}}.
To achieve this, we record the gradients during the simulation and define a simple \(L^2\) based loss function. The loss function we’ll use is given by \(L = | s_{t_e} - s_{t_e}^* |^2\), where \(s_{t_e}\) denotes the smoke density, and \(s_{t_e}^*\)
denotes the reference state from the fourth simulation in our batch (both evaluated at the last time step \(t_e\)).
When evaluating the loss function we treat the reference state as an external constant via \sphinxcode{\sphinxupquote{field.stop\_gradient()}}.
As outlined at the top, \(s\) is a function of \(\mathbf{u}\) (via the advection equation), which in turn is given by the Navier\sphinxhyphen{}Stokes equations. Thus, via a chain of multiple time steps \(s\) depends in the initial velocity state \(\mathbf{u}_0\).

It is important that our initial velocity has the \sphinxcode{\sphinxupquote{inflow\_loc}} dimension before we record the gradients, such that we have the full “mini\sphinxhyphen{}batch” of four versions of our velocity (three of which will be updated via gradients in our optimization later on). To get the appropriate velocity tensor, we initialize a \sphinxcode{\sphinxupquote{StaggeredGrid}} with a tensor of zeros along the \sphinxcode{\sphinxupquote{inflow\_loc}} batch dimension. As the staggered grid already has \sphinxcode{\sphinxupquote{y,x}} and \sphinxcode{\sphinxupquote{vector}} dimensions, this gives the desired four dimensions, as verified by the print statement below.

Phiflow provides a unified API for gradients across different platforms by using functions that need to return a loss values, in addition to optional state values. It uses a loss function based interface, for which we define the \sphinxcode{\sphinxupquote{simulate}} function below. \sphinxcode{\sphinxupquote{simulate}} computes the \(L^2\) error outlined above and returns the evolved \sphinxcode{\sphinxupquote{smoke}} and \sphinxcode{\sphinxupquote{velocity}} states after 20 simulation steps.

\begin{sphinxVerbatim}[commandchars=\\\{\}]
\PYG{n}{initial\PYGZus{}smoke} \PYG{o}{=} \PYG{n}{CenteredGrid}\PYG{p}{(}\PYG{l+m+mi}{0}\PYG{p}{,} \PYG{n}{extrapolation}\PYG{o}{.}\PYG{n}{BOUNDARY}\PYG{p}{,} \PYG{n}{x}\PYG{o}{=}\PYG{l+m+mi}{32}\PYG{p}{,} \PYG{n}{y}\PYG{o}{=}\PYG{l+m+mi}{40}\PYG{p}{,} \PYG{n}{bounds}\PYG{o}{=}\PYG{n}{Box}\PYG{p}{[}\PYG{l+m+mi}{0}\PYG{p}{:}\PYG{l+m+mi}{32}\PYG{p}{,} \PYG{l+m+mi}{0}\PYG{p}{:}\PYG{l+m+mi}{40}\PYG{p}{]}\PYG{p}{)}
\PYG{n}{initial\PYGZus{}velocity} \PYG{o}{=} \PYG{n}{StaggeredGrid}\PYG{p}{(}\PYG{n}{math}\PYG{o}{.}\PYG{n}{zeros}\PYG{p}{(}\PYG{n}{batch}\PYG{p}{(}\PYG{n}{inflow\PYGZus{}loc}\PYG{o}{=}\PYG{l+m+mi}{4}\PYG{p}{)}\PYG{p}{)}\PYG{p}{,} \PYG{n}{BND}\PYG{p}{,} \PYG{n}{x}\PYG{o}{=}\PYG{l+m+mi}{32}\PYG{p}{,} \PYG{n}{y}\PYG{o}{=}\PYG{l+m+mi}{40}\PYG{p}{,} \PYG{n}{bounds}\PYG{o}{=}\PYG{n}{Box}\PYG{p}{[}\PYG{l+m+mi}{0}\PYG{p}{:}\PYG{l+m+mi}{32}\PYG{p}{,} \PYG{l+m+mi}{0}\PYG{p}{:}\PYG{l+m+mi}{40}\PYG{p}{]}\PYG{p}{)}
\PYG{n+nb}{print}\PYG{p}{(}\PYG{l+s+s2}{\PYGZdq{}}\PYG{l+s+s2}{Velocity dimensions: }\PYG{l+s+s2}{\PYGZdq{}}\PYG{o}{+}\PYG{n+nb}{format}\PYG{p}{(}\PYG{n}{initial\PYGZus{}velocity}\PYG{o}{.}\PYG{n}{shape}\PYG{p}{)}\PYG{p}{)}

\PYG{k}{def} \PYG{n+nf}{simulate}\PYG{p}{(}\PYG{n}{smoke}\PYG{p}{:} \PYG{n}{CenteredGrid}\PYG{p}{,} \PYG{n}{velocity}\PYG{p}{:} \PYG{n}{StaggeredGrid}\PYG{p}{)}\PYG{p}{:}
    \PYG{k}{for} \PYG{n}{\PYGZus{}} \PYG{o+ow}{in} \PYG{n+nb}{range}\PYG{p}{(}\PYG{l+m+mi}{20}\PYG{p}{)}\PYG{p}{:}
        \PYG{n}{smoke}\PYG{p}{,}\PYG{n}{velocity} \PYG{o}{=} \PYG{n}{step}\PYG{p}{(}\PYG{n}{smoke}\PYG{p}{,}\PYG{n}{velocity}\PYG{p}{)}
        
    \PYG{n}{loss} \PYG{o}{=} \PYG{n}{field}\PYG{o}{.}\PYG{n}{l2\PYGZus{}loss}\PYG{p}{(}\PYG{n}{smoke} \PYG{o}{\PYGZhy{}} \PYG{n}{field}\PYG{o}{.}\PYG{n}{stop\PYGZus{}gradient}\PYG{p}{(}\PYG{n}{smoke}\PYG{o}{.}\PYG{n}{inflow\PYGZus{}loc}\PYG{p}{[}\PYG{o}{\PYGZhy{}}\PYG{l+m+mi}{1}\PYG{p}{]}\PYG{p}{)} \PYG{p}{)}
    \PYG{c+c1}{\PYGZsh{} optionally, use smoother loss with diffusion steps \PYGZhy{} no difference here, but can be useful for more complex cases}
    \PYG{c+c1}{\PYGZsh{}loss = field.l2\PYGZus{}loss(diffuse.explicit(smoke \PYGZhy{} field.stop\PYGZus{}gradient(smoke.inflow\PYGZus{}loc[\PYGZhy{}1]), 1, 1, 10))}
    
    \PYG{k}{return} \PYG{n}{loss}\PYG{p}{,} \PYG{n}{smoke}\PYG{p}{,} \PYG{n}{velocity}
\end{sphinxVerbatim}

\begin{sphinxVerbatim}[commandchars=\\\{\}]
Velocity dimensions: (inflow\PYGZus{}locᵇ=4, xˢ=32, yˢ=40, vectorᵛ=2)
\end{sphinxVerbatim}

Phiflow’s \sphinxcode{\sphinxupquote{field.functional\_gradient()}} function is the central function to compute gradients. Next, we’ll use it to obtain the gradient with respect to the initial velocity. Since the velocity is the second argument of the \sphinxcode{\sphinxupquote{simulate()}} function, we pass \sphinxcode{\sphinxupquote{wrt={[}1{]}}}. (Phiflow also has a \sphinxcode{\sphinxupquote{field.spatial\_gradient}} function which instead computes derivatives of tensors along spatial dimensions, like \sphinxcode{\sphinxupquote{x,y}}.)

\sphinxcode{\sphinxupquote{functional\_gradient}} generates a gradient function. As a demonstration, the next cell evaluates the gradient once with the initial states for smoke and velocity. The last statement prints a summary of a part of the resulting gradient tensor.

\begin{sphinxVerbatim}[commandchars=\\\{\}]
\PYG{n}{sim\PYGZus{}grad} \PYG{o}{=} \PYG{n}{field}\PYG{o}{.}\PYG{n}{functional\PYGZus{}gradient}\PYG{p}{(}\PYG{n}{simulate}\PYG{p}{,} \PYG{n}{wrt}\PYG{o}{=}\PYG{p}{[}\PYG{l+m+mi}{1}\PYG{p}{]}\PYG{p}{,} \PYG{n}{get\PYGZus{}output}\PYG{o}{=}\PYG{k+kc}{False}\PYG{p}{)}
\PYG{p}{(}\PYG{n}{velocity\PYGZus{}grad}\PYG{p}{,}\PYG{p}{)} \PYG{o}{=} \PYG{n}{sim\PYGZus{}grad}\PYG{p}{(}\PYG{n}{initial\PYGZus{}smoke}\PYG{p}{,} \PYG{n}{initial\PYGZus{}velocity}\PYG{p}{)}

\PYG{n+nb}{print}\PYG{p}{(}\PYG{l+s+s2}{\PYGZdq{}}\PYG{l+s+s2}{Some gradient info: }\PYG{l+s+s2}{\PYGZdq{}} \PYG{o}{+} \PYG{n+nb}{format}\PYG{p}{(}\PYG{n}{velocity\PYGZus{}grad}\PYG{p}{)}\PYG{p}{)}
\PYG{n+nb}{print}\PYG{p}{(}\PYG{n+nb}{format}\PYG{p}{(}\PYG{n}{velocity\PYGZus{}grad}\PYG{o}{.}\PYG{n}{values}\PYG{o}{.}\PYG{n}{inflow\PYGZus{}loc}\PYG{p}{[}\PYG{l+m+mi}{0}\PYG{p}{]}\PYG{o}{.}\PYG{n}{vector}\PYG{p}{[}\PYG{l+m+mi}{0}\PYG{p}{]}\PYG{p}{)}\PYG{p}{)} \PYG{c+c1}{\PYGZsh{} one example, location 0, x component, automatically prints size \PYGZam{} content range}
\end{sphinxVerbatim}

\begin{sphinxVerbatim}[commandchars=\\\{\}]
Some gradient info: StaggeredGrid[(inflow\PYGZus{}locᵇ=4, xˢ=32, yˢ=40, vectorᵛ=2), size=(32, 40), extrapolation=0]
(xˢ=31, yˢ=40) float32  \PYGZhy{}17.366662979125977 \PYGZlt{} ... \PYGZlt{} 14.014090538024902
\end{sphinxVerbatim}

The last two lines just print some information about the resulting gradient field. Naturally, it has the same shape as the velocity itself: it’s a staggered grid with four inflow locations. The last line shows how to access the x\sphinxhyphen{}components of one of the gradients.

We could use this to take a look at the content of the computed gradient with regular plotting functions, e.g., by converting the x component of one of the simulations to a numpy array via \sphinxcode{\sphinxupquote{velocity\_grad.values.inflow\_loc{[}0{]}.vector{[}0{]}.numpy('y,x')}}. However, below we’ll use phiflow’s \sphinxcode{\sphinxupquote{view()}} function instead. It automatically analyzes the grid content and provides UI buttons to choose different viewing modes. You can use them to show arrows, single components of the 2\sphinxhyphen{}dimensional velocity vectors, or their magnitudes. (Because of its interactive nature, the corresponding image won’t show up outside of Jupyter, though.)

\begin{sphinxVerbatim}[commandchars=\\\{\}]
\PYG{c+c1}{\PYGZsh{} neat phiflow helper function:}
\PYG{n}{view}\PYG{p}{(}\PYG{n}{velocity\PYGZus{}grad}\PYG{p}{)}
\end{sphinxVerbatim}

Not surprisingly, the fourth gradient on the left is zero (it’s already matching the reference). The other three gradients have detected variations for the initial round inflow positions shown as positive and negative regions around the circular shape of the inflow. The ones for the larger distances on the left are also noticeably larger.

\section{Optimization}
\label{\detokenize{diffphys-code-ns:optimization}}
The gradient visualized above is just the linearized change that points in the direction of an increasing loss. Now we can proceed by updating the initial velocities in the opposite direction to minimize the loss, and iterate to find a minimizer.

This is a difficult task: the simulation is producing different dynamics due to the differing initial spatial density configuration. Our optimization should now find a single initial velocity state that gives the same state as the reference simulation at \(t=20\). Thus, after 20 non\sphinxhyphen{}linear update steps the simulation should reproduce the desired marker density state. It would be much easier to simply change the position of the marker inflow to arrive at this goal, but – to make things more difficult and interesting here – the inflow is \sphinxstyleemphasis{not} a degree of freedom. The optimizer can only change the initial velocity \(\mathbf{u}_0\).

The following cell implements a simple steepest gradient descent optimization: it re\sphinxhyphen{}evaluates the gradient function, and iterates several times to optimize \(\mathbf{u}_0\) with a learning rate (step size) \sphinxcode{\sphinxupquote{LR}}.

\sphinxcode{\sphinxupquote{field.functional\_gradient}} has a parameter \sphinxcode{\sphinxupquote{get\_output}} that determines whether the original results of the function (\sphinxcode{\sphinxupquote{simulate()}} in our case) are returned, or only the gradient. As it’s interesting to track how the loss evolves over the course of the iterations, let’s redefine the gradient function with \sphinxcode{\sphinxupquote{get\_output=True}}.

\begin{sphinxVerbatim}[commandchars=\\\{\}]
\PYG{n}{sim\PYGZus{}grad\PYGZus{}wloss} \PYG{o}{=} \PYG{n}{field}\PYG{o}{.}\PYG{n}{functional\PYGZus{}gradient}\PYG{p}{(}\PYG{n}{simulate}\PYG{p}{,} \PYG{n}{wrt}\PYG{o}{=}\PYG{p}{[}\PYG{l+m+mi}{1}\PYG{p}{]}\PYG{p}{,} \PYG{n}{get\PYGZus{}output}\PYG{o}{=}\PYG{k+kc}{True}\PYG{p}{)} \PYG{c+c1}{\PYGZsh{} if we need outputs...}

\PYG{n}{LR} \PYG{o}{=} \PYG{l+m+mf}{1e\PYGZhy{}03} 
\PYG{k}{for} \PYG{n}{optim\PYGZus{}step} \PYG{o+ow}{in} \PYG{n+nb}{range}\PYG{p}{(}\PYG{l+m+mi}{80}\PYG{p}{)}\PYG{p}{:}    
    \PYG{p}{(}\PYG{n}{loss}\PYG{p}{,} \PYG{n}{\PYGZus{}smoke}\PYG{p}{,} \PYG{n}{\PYGZus{}velocity}\PYG{p}{)}\PYG{p}{,} \PYG{p}{(}\PYG{n}{velocity\PYGZus{}grad}\PYG{p}{,}\PYG{p}{)} \PYG{o}{=} \PYG{n}{sim\PYGZus{}grad\PYGZus{}wloss}\PYG{p}{(}\PYG{n}{initial\PYGZus{}smoke}\PYG{p}{,} \PYG{n}{initial\PYGZus{}velocity}\PYG{p}{)}
    \PYG{n}{initial\PYGZus{}velocity} \PYG{o}{=} \PYG{n}{initial\PYGZus{}velocity} \PYG{o}{\PYGZhy{}} \PYG{n}{LR} \PYG{o}{*} \PYG{n}{velocity\PYGZus{}grad}
    \PYG{k}{if} \PYG{n}{optim\PYGZus{}step}\PYG{o}{\PYGZlt{}}\PYG{l+m+mi}{3} \PYG{o+ow}{or} \PYG{n}{optim\PYGZus{}step}\PYG{o}{\PYGZpc{}}\PYG{k}{10}==9: print(\PYGZsq{}Optimization step \PYGZpc{}d, loss: \PYGZpc{}f\PYGZsq{} \PYGZpc{} (optim\PYGZus{}step, np.sum(loss.numpy()) ))
\end{sphinxVerbatim}

\begin{sphinxVerbatim}[commandchars=\\\{\}]
Optimization step 0, loss: 1193.145020
Optimization step 1, loss: 1165.816650
Optimization step 2, loss: 1104.294556
Optimization step 10, loss: 861.661743
Optimization step 20, loss: 775.154846
Optimization step 30, loss: 747.199829
Optimization step 40, loss: 684.146729
Optimization step 50, loss: 703.087158
Optimization step 60, loss: 660.258423
Optimization step 70, loss: 649.957214
\end{sphinxVerbatim}

The loss should have gone down significantly, from above 1100 to below 700, and now we can also visualize the initial velocities that were obtained in the optimization.

The following images show the resulting three initial velocities in terms of their x (first set), and y components (second set of images). We’re skipping the fourth set with \sphinxcode{\sphinxupquote{inflow\_loc{[}0{]}}} because it only contains zeros.

\begin{sphinxVerbatim}[commandchars=\\\{\}]
\PYG{n}{fig}\PYG{p}{,} \PYG{n}{axes} \PYG{o}{=} \PYG{n}{pylab}\PYG{o}{.}\PYG{n}{subplots}\PYG{p}{(}\PYG{l+m+mi}{1}\PYG{p}{,} \PYG{l+m+mi}{3}\PYG{p}{,} \PYG{n}{figsize}\PYG{o}{=}\PYG{p}{(}\PYG{l+m+mi}{10}\PYG{p}{,} \PYG{l+m+mi}{4}\PYG{p}{)}\PYG{p}{)}
\PYG{k}{for} \PYG{n}{i} \PYG{o+ow}{in} \PYG{n+nb}{range}\PYG{p}{(}\PYG{n}{INFLOW}\PYG{o}{.}\PYG{n}{shape}\PYG{o}{.}\PYG{n}{get\PYGZus{}size}\PYG{p}{(}\PYG{l+s+s1}{\PYGZsq{}}\PYG{l+s+s1}{inflow\PYGZus{}loc}\PYG{l+s+s1}{\PYGZsq{}}\PYG{p}{)}\PYG{o}{\PYGZhy{}}\PYG{l+m+mi}{1}\PYG{p}{)}\PYG{p}{:}
  \PYG{n}{im} \PYG{o}{=} \PYG{n}{axes}\PYG{p}{[}\PYG{n}{i}\PYG{p}{]}\PYG{o}{.}\PYG{n}{imshow}\PYG{p}{(}\PYG{n}{initial\PYGZus{}velocity}\PYG{o}{.}\PYG{n}{staggered\PYGZus{}tensor}\PYG{p}{(}\PYG{p}{)}\PYG{o}{.}\PYG{n}{numpy}\PYG{p}{(}\PYG{l+s+s1}{\PYGZsq{}}\PYG{l+s+s1}{inflow\PYGZus{}loc,y,x,vector}\PYG{l+s+s1}{\PYGZsq{}}\PYG{p}{)}\PYG{p}{[}\PYG{n}{i}\PYG{p}{,}\PYG{o}{.}\PYG{o}{.}\PYG{o}{.}\PYG{p}{,}\PYG{l+m+mi}{0}\PYG{p}{]}\PYG{p}{,} \PYG{n}{origin}\PYG{o}{=}\PYG{l+s+s1}{\PYGZsq{}}\PYG{l+s+s1}{lower}\PYG{l+s+s1}{\PYGZsq{}}\PYG{p}{,} \PYG{n}{cmap}\PYG{o}{=}\PYG{l+s+s1}{\PYGZsq{}}\PYG{l+s+s1}{magma}\PYG{l+s+s1}{\PYGZsq{}}\PYG{p}{)}
  \PYG{n}{axes}\PYG{p}{[}\PYG{n}{i}\PYG{p}{]}\PYG{o}{.}\PYG{n}{set\PYGZus{}title}\PYG{p}{(}\PYG{l+s+sa}{f}\PYG{l+s+s2}{\PYGZdq{}}\PYG{l+s+s2}{Ini. vel. X }\PYG{l+s+si}{\PYGZob{}}\PYG{n}{INFLOW\PYGZus{}LOCATION}\PYG{o}{.}\PYG{n}{numpy}\PYG{p}{(}\PYG{l+s+s1}{\PYGZsq{}}\PYG{l+s+s1}{inflow\PYGZus{}loc,vector}\PYG{l+s+s1}{\PYGZsq{}}\PYG{p}{)}\PYG{p}{[}\PYG{n}{i}\PYG{p}{]}\PYG{l+s+si}{\PYGZcb{}}\PYG{l+s+s2}{\PYGZdq{}}\PYG{p}{)}
  \PYG{n}{pylab}\PYG{o}{.}\PYG{n}{colorbar}\PYG{p}{(}\PYG{n}{im}\PYG{p}{,}\PYG{n}{ax}\PYG{o}{=}\PYG{n}{axes}\PYG{p}{[}\PYG{n}{i}\PYG{p}{]}\PYG{p}{)}
\PYG{n}{pylab}\PYG{o}{.}\PYG{n}{tight\PYGZus{}layout}\PYG{p}{(}\PYG{p}{)}
\end{sphinxVerbatim}

\noindent\sphinxincludegraphics{{diffphys-code-ns_20_0}.png}

\begin{sphinxVerbatim}[commandchars=\\\{\}]
\PYG{n}{fig}\PYG{p}{,} \PYG{n}{axes} \PYG{o}{=} \PYG{n}{pylab}\PYG{o}{.}\PYG{n}{subplots}\PYG{p}{(}\PYG{l+m+mi}{1}\PYG{p}{,} \PYG{l+m+mi}{3}\PYG{p}{,} \PYG{n}{figsize}\PYG{o}{=}\PYG{p}{(}\PYG{l+m+mi}{10}\PYG{p}{,} \PYG{l+m+mi}{4}\PYG{p}{)}\PYG{p}{)}
\PYG{k}{for} \PYG{n}{i} \PYG{o+ow}{in} \PYG{n+nb}{range}\PYG{p}{(}\PYG{n}{INFLOW}\PYG{o}{.}\PYG{n}{shape}\PYG{o}{.}\PYG{n}{get\PYGZus{}size}\PYG{p}{(}\PYG{l+s+s1}{\PYGZsq{}}\PYG{l+s+s1}{inflow\PYGZus{}loc}\PYG{l+s+s1}{\PYGZsq{}}\PYG{p}{)}\PYG{o}{\PYGZhy{}}\PYG{l+m+mi}{1}\PYG{p}{)}\PYG{p}{:}
  \PYG{n}{im} \PYG{o}{=} \PYG{n}{axes}\PYG{p}{[}\PYG{n}{i}\PYG{p}{]}\PYG{o}{.}\PYG{n}{imshow}\PYG{p}{(}\PYG{n}{initial\PYGZus{}velocity}\PYG{o}{.}\PYG{n}{staggered\PYGZus{}tensor}\PYG{p}{(}\PYG{p}{)}\PYG{o}{.}\PYG{n}{numpy}\PYG{p}{(}\PYG{l+s+s1}{\PYGZsq{}}\PYG{l+s+s1}{inflow\PYGZus{}loc,y,x,vector}\PYG{l+s+s1}{\PYGZsq{}}\PYG{p}{)}\PYG{p}{[}\PYG{n}{i}\PYG{p}{,}\PYG{o}{.}\PYG{o}{.}\PYG{o}{.}\PYG{p}{,}\PYG{l+m+mi}{1}\PYG{p}{]}\PYG{p}{,} \PYG{n}{origin}\PYG{o}{=}\PYG{l+s+s1}{\PYGZsq{}}\PYG{l+s+s1}{lower}\PYG{l+s+s1}{\PYGZsq{}}\PYG{p}{,} \PYG{n}{cmap}\PYG{o}{=}\PYG{l+s+s1}{\PYGZsq{}}\PYG{l+s+s1}{magma}\PYG{l+s+s1}{\PYGZsq{}}\PYG{p}{)}
  \PYG{n}{axes}\PYG{p}{[}\PYG{n}{i}\PYG{p}{]}\PYG{o}{.}\PYG{n}{set\PYGZus{}title}\PYG{p}{(}\PYG{l+s+sa}{f}\PYG{l+s+s2}{\PYGZdq{}}\PYG{l+s+s2}{Ini. vel. Y }\PYG{l+s+si}{\PYGZob{}}\PYG{n}{INFLOW\PYGZus{}LOCATION}\PYG{o}{.}\PYG{n}{numpy}\PYG{p}{(}\PYG{l+s+s1}{\PYGZsq{}}\PYG{l+s+s1}{inflow\PYGZus{}loc,vector}\PYG{l+s+s1}{\PYGZsq{}}\PYG{p}{)}\PYG{p}{[}\PYG{n}{i}\PYG{p}{]}\PYG{l+s+si}{\PYGZcb{}}\PYG{l+s+s2}{\PYGZdq{}}\PYG{p}{)}
  \PYG{n}{pylab}\PYG{o}{.}\PYG{n}{colorbar}\PYG{p}{(}\PYG{n}{im}\PYG{p}{,}\PYG{n}{ax}\PYG{o}{=}\PYG{n}{axes}\PYG{p}{[}\PYG{n}{i}\PYG{p}{]}\PYG{p}{)}
\PYG{n}{pylab}\PYG{o}{.}\PYG{n}{tight\PYGZus{}layout}\PYG{p}{(}\PYG{p}{)}
\end{sphinxVerbatim}

\noindent\sphinxincludegraphics{{diffphys-code-ns_21_0}.png}

\section{Re\sphinxhyphen{}simulation}
\label{\detokenize{diffphys-code-ns:re-simulation}}
We can also visualize how the full simulation over the course of 20 steps turns out, given the new initial velocity conditions for each of the inflow locations. This is what happened internally at optimization time for every gradient calculation, and what was measured by our loss function. Hence, it’s good to get an intuition for which solutions the optimization has found.

Below, we re\sphinxhyphen{}run the forward simulation with the new initial conditions from \sphinxcode{\sphinxupquote{initial\_velocity}}:

\begin{sphinxVerbatim}[commandchars=\\\{\}]
\PYG{n}{smoke} \PYG{o}{=} \PYG{n}{initial\PYGZus{}smoke} 
\PYG{n}{velocity} \PYG{o}{=} \PYG{n}{initial\PYGZus{}velocity}

\PYG{k}{for} \PYG{n}{\PYGZus{}} \PYG{o+ow}{in} \PYG{n+nb}{range}\PYG{p}{(}\PYG{l+m+mi}{20}\PYG{p}{)}\PYG{p}{:}
    \PYG{n}{smoke}\PYG{p}{,}\PYG{n}{velocity} \PYG{o}{=} \PYG{n}{step}\PYG{p}{(}\PYG{n}{smoke}\PYG{p}{,}\PYG{n}{velocity}\PYG{p}{)}

\PYG{n}{fig}\PYG{p}{,} \PYG{n}{axes} \PYG{o}{=} \PYG{n}{pylab}\PYG{o}{.}\PYG{n}{subplots}\PYG{p}{(}\PYG{l+m+mi}{1}\PYG{p}{,} \PYG{l+m+mi}{4}\PYG{p}{,} \PYG{n}{figsize}\PYG{o}{=}\PYG{p}{(}\PYG{l+m+mi}{10}\PYG{p}{,} \PYG{l+m+mi}{6}\PYG{p}{)}\PYG{p}{)}
\PYG{k}{for} \PYG{n}{i} \PYG{o+ow}{in} \PYG{n+nb}{range}\PYG{p}{(}\PYG{n}{INFLOW}\PYG{o}{.}\PYG{n}{shape}\PYG{o}{.}\PYG{n}{get\PYGZus{}size}\PYG{p}{(}\PYG{l+s+s1}{\PYGZsq{}}\PYG{l+s+s1}{inflow\PYGZus{}loc}\PYG{l+s+s1}{\PYGZsq{}}\PYG{p}{)}\PYG{p}{)}\PYG{p}{:}
  \PYG{n}{axes}\PYG{p}{[}\PYG{n}{i}\PYG{p}{]}\PYG{o}{.}\PYG{n}{imshow}\PYG{p}{(}\PYG{n}{smoke\PYGZus{}final}\PYG{o}{.}\PYG{n}{values}\PYG{o}{.}\PYG{n}{numpy}\PYG{p}{(}\PYG{l+s+s1}{\PYGZsq{}}\PYG{l+s+s1}{inflow\PYGZus{}loc,y,x}\PYG{l+s+s1}{\PYGZsq{}}\PYG{p}{)}\PYG{p}{[}\PYG{n}{i}\PYG{p}{,}\PYG{o}{.}\PYG{o}{.}\PYG{o}{.}\PYG{p}{]}\PYG{p}{,} \PYG{n}{origin}\PYG{o}{=}\PYG{l+s+s1}{\PYGZsq{}}\PYG{l+s+s1}{lower}\PYG{l+s+s1}{\PYGZsq{}}\PYG{p}{,} \PYG{n}{cmap}\PYG{o}{=}\PYG{l+s+s1}{\PYGZsq{}}\PYG{l+s+s1}{magma}\PYG{l+s+s1}{\PYGZsq{}}\PYG{p}{)}
  \PYG{n}{axes}\PYG{p}{[}\PYG{n}{i}\PYG{p}{]}\PYG{o}{.}\PYG{n}{set\PYGZus{}title}\PYG{p}{(}\PYG{l+s+sa}{f}\PYG{l+s+s2}{\PYGZdq{}}\PYG{l+s+s2}{Inflow }\PYG{l+s+si}{\PYGZob{}}\PYG{n}{INFLOW\PYGZus{}LOCATION}\PYG{o}{.}\PYG{n}{numpy}\PYG{p}{(}\PYG{l+s+s1}{\PYGZsq{}}\PYG{l+s+s1}{inflow\PYGZus{}loc,vector}\PYG{l+s+s1}{\PYGZsq{}}\PYG{p}{)}\PYG{p}{[}\PYG{n}{i}\PYG{p}{]}\PYG{l+s+si}{\PYGZcb{}}\PYG{l+s+s2}{\PYGZdq{}} \PYG{o}{+} \PYG{p}{(}\PYG{l+s+s2}{\PYGZdq{}}\PYG{l+s+s2}{, Reference}\PYG{l+s+s2}{\PYGZdq{}} \PYG{k}{if} \PYG{n}{i}\PYG{o}{==}\PYG{l+m+mi}{3} \PYG{k}{else} \PYG{l+s+s2}{\PYGZdq{}}\PYG{l+s+s2}{\PYGZdq{}}\PYG{p}{)}\PYG{p}{)}
\PYG{n}{pylab}\PYG{o}{.}\PYG{n}{tight\PYGZus{}layout}\PYG{p}{(}\PYG{p}{)}
\end{sphinxVerbatim}

\noindent\sphinxincludegraphics{{diffphys-code-ns_23_0}.png}

Naturally, the image on the right is the same (this is the reference), and the other three simulations now exhibit a  shift towards the right. As the differences are a bit subtle, let’s visualize the difference between the target configuration and the different final states.

The following images contain the difference between the evolved simulated and target density. Hence, dark regions indicate where the target should be, but isn’t. The top row shows the original states with the initial velocity being zero, while the bottom row shows the versions after the optimization has tuned the initial velocities. Hence, in each column you can compare before (top) and after (bottom):

\begin{sphinxVerbatim}[commandchars=\\\{\}]
\PYG{n}{fig}\PYG{p}{,} \PYG{n}{axes} \PYG{o}{=} \PYG{n}{pylab}\PYG{o}{.}\PYG{n}{subplots}\PYG{p}{(}\PYG{l+m+mi}{2}\PYG{p}{,} \PYG{l+m+mi}{3}\PYG{p}{,} \PYG{n}{figsize}\PYG{o}{=}\PYG{p}{(}\PYG{l+m+mi}{10}\PYG{p}{,} \PYG{l+m+mi}{6}\PYG{p}{)}\PYG{p}{)}
\PYG{k}{for} \PYG{n}{i} \PYG{o+ow}{in} \PYG{n+nb}{range}\PYG{p}{(}\PYG{n}{INFLOW}\PYG{o}{.}\PYG{n}{shape}\PYG{o}{.}\PYG{n}{get\PYGZus{}size}\PYG{p}{(}\PYG{l+s+s1}{\PYGZsq{}}\PYG{l+s+s1}{inflow\PYGZus{}loc}\PYG{l+s+s1}{\PYGZsq{}}\PYG{p}{)}\PYG{o}{\PYGZhy{}}\PYG{l+m+mi}{1}\PYG{p}{)}\PYG{p}{:}
  \PYG{n}{axes}\PYG{p}{[}\PYG{l+m+mi}{0}\PYG{p}{,}\PYG{n}{i}\PYG{p}{]}\PYG{o}{.}\PYG{n}{imshow}\PYG{p}{(}\PYG{n}{smoke\PYGZus{}final}\PYG{o}{.}\PYG{n}{values}\PYG{o}{.}\PYG{n}{numpy}\PYG{p}{(}\PYG{l+s+s1}{\PYGZsq{}}\PYG{l+s+s1}{inflow\PYGZus{}loc,y,x}\PYG{l+s+s1}{\PYGZsq{}}\PYG{p}{)}\PYG{p}{[}\PYG{n}{i}\PYG{p}{,}\PYG{o}{.}\PYG{o}{.}\PYG{o}{.}\PYG{p}{]} \PYG{o}{\PYGZhy{}} \PYG{n}{smoke\PYGZus{}final}\PYG{o}{.}\PYG{n}{values}\PYG{o}{.}\PYG{n}{numpy}\PYG{p}{(}\PYG{l+s+s1}{\PYGZsq{}}\PYG{l+s+s1}{inflow\PYGZus{}loc,y,x}\PYG{l+s+s1}{\PYGZsq{}}\PYG{p}{)}\PYG{p}{[}\PYG{l+m+mi}{3}\PYG{p}{,}\PYG{o}{.}\PYG{o}{.}\PYG{o}{.}\PYG{p}{]}\PYG{p}{,} \PYG{n}{origin}\PYG{o}{=}\PYG{l+s+s1}{\PYGZsq{}}\PYG{l+s+s1}{lower}\PYG{l+s+s1}{\PYGZsq{}}\PYG{p}{,} \PYG{n}{cmap}\PYG{o}{=}\PYG{l+s+s1}{\PYGZsq{}}\PYG{l+s+s1}{magma}\PYG{l+s+s1}{\PYGZsq{}}\PYG{p}{)}
  \PYG{n}{axes}\PYG{p}{[}\PYG{l+m+mi}{0}\PYG{p}{,}\PYG{n}{i}\PYG{p}{]}\PYG{o}{.}\PYG{n}{set\PYGZus{}title}\PYG{p}{(}\PYG{l+s+sa}{f}\PYG{l+s+s2}{\PYGZdq{}}\PYG{l+s+s2}{Org. diff. }\PYG{l+s+si}{\PYGZob{}}\PYG{n}{INFLOW\PYGZus{}LOCATION}\PYG{o}{.}\PYG{n}{numpy}\PYG{p}{(}\PYG{l+s+s1}{\PYGZsq{}}\PYG{l+s+s1}{inflow\PYGZus{}loc,vector}\PYG{l+s+s1}{\PYGZsq{}}\PYG{p}{)}\PYG{p}{[}\PYG{n}{i}\PYG{p}{]}\PYG{l+s+si}{\PYGZcb{}}\PYG{l+s+s2}{\PYGZdq{}}\PYG{p}{)}
  \PYG{n}{pylab}\PYG{o}{.}\PYG{n}{colorbar}\PYG{p}{(}\PYG{n}{im}\PYG{p}{,}\PYG{n}{ax}\PYG{o}{=}\PYG{n}{axes}\PYG{p}{[}\PYG{l+m+mi}{0}\PYG{p}{,}\PYG{n}{i}\PYG{p}{]}\PYG{p}{)}
\PYG{k}{for} \PYG{n}{i} \PYG{o+ow}{in} \PYG{n+nb}{range}\PYG{p}{(}\PYG{n}{INFLOW}\PYG{o}{.}\PYG{n}{shape}\PYG{o}{.}\PYG{n}{get\PYGZus{}size}\PYG{p}{(}\PYG{l+s+s1}{\PYGZsq{}}\PYG{l+s+s1}{inflow\PYGZus{}loc}\PYG{l+s+s1}{\PYGZsq{}}\PYG{p}{)}\PYG{o}{\PYGZhy{}}\PYG{l+m+mi}{1}\PYG{p}{)}\PYG{p}{:}
  \PYG{n}{axes}\PYG{p}{[}\PYG{l+m+mi}{1}\PYG{p}{,}\PYG{n}{i}\PYG{p}{]}\PYG{o}{.}\PYG{n}{imshow}\PYG{p}{(}\PYG{n}{smoke}\PYG{o}{.}\PYG{n}{values}\PYG{o}{.}\PYG{n}{numpy}\PYG{p}{(}\PYG{l+s+s1}{\PYGZsq{}}\PYG{l+s+s1}{inflow\PYGZus{}loc,y,x}\PYG{l+s+s1}{\PYGZsq{}}\PYG{p}{)}\PYG{p}{[}\PYG{n}{i}\PYG{p}{,}\PYG{o}{.}\PYG{o}{.}\PYG{o}{.}\PYG{p}{]} \PYG{o}{\PYGZhy{}} \PYG{n}{smoke\PYGZus{}final}\PYG{o}{.}\PYG{n}{values}\PYG{o}{.}\PYG{n}{numpy}\PYG{p}{(}\PYG{l+s+s1}{\PYGZsq{}}\PYG{l+s+s1}{inflow\PYGZus{}loc,y,x}\PYG{l+s+s1}{\PYGZsq{}}\PYG{p}{)}\PYG{p}{[}\PYG{l+m+mi}{3}\PYG{p}{,}\PYG{o}{.}\PYG{o}{.}\PYG{o}{.}\PYG{p}{]}\PYG{p}{,} \PYG{n}{origin}\PYG{o}{=}\PYG{l+s+s1}{\PYGZsq{}}\PYG{l+s+s1}{lower}\PYG{l+s+s1}{\PYGZsq{}}\PYG{p}{,} \PYG{n}{cmap}\PYG{o}{=}\PYG{l+s+s1}{\PYGZsq{}}\PYG{l+s+s1}{magma}\PYG{l+s+s1}{\PYGZsq{}}\PYG{p}{)}
  \PYG{n}{axes}\PYG{p}{[}\PYG{l+m+mi}{1}\PYG{p}{,}\PYG{n}{i}\PYG{p}{]}\PYG{o}{.}\PYG{n}{set\PYGZus{}title}\PYG{p}{(}\PYG{l+s+sa}{f}\PYG{l+s+s2}{\PYGZdq{}}\PYG{l+s+s2}{Result }\PYG{l+s+si}{\PYGZob{}}\PYG{n}{INFLOW\PYGZus{}LOCATION}\PYG{o}{.}\PYG{n}{numpy}\PYG{p}{(}\PYG{l+s+s1}{\PYGZsq{}}\PYG{l+s+s1}{inflow\PYGZus{}loc,vector}\PYG{l+s+s1}{\PYGZsq{}}\PYG{p}{)}\PYG{p}{[}\PYG{n}{i}\PYG{p}{]}\PYG{l+s+si}{\PYGZcb{}}\PYG{l+s+s2}{\PYGZdq{}}\PYG{p}{)}
  \PYG{n}{pylab}\PYG{o}{.}\PYG{n}{colorbar}\PYG{p}{(}\PYG{n}{im}\PYG{p}{,}\PYG{n}{ax}\PYG{o}{=}\PYG{n}{axes}\PYG{p}{[}\PYG{l+m+mi}{1}\PYG{p}{,}\PYG{n}{i}\PYG{p}{]}\PYG{p}{)}
\end{sphinxVerbatim}

\noindent\sphinxincludegraphics{{diffphys-code-ns_25_0}.png}

These difference images clearly show that the optimization managed to align the upper region of the plumes very well. Each original image (at the top) shows a clear misalignment in terms of a black halo, while the states after optimization largely overlap the target smoke configuration of the reference, and exhibit differences closer to zero for the front of each smoke cloud.

Note that all three simulations need to “work” with a fixed inflow, hence they cannot simply “produce” marker density out of the blue to match the target. Also each simulation needs to take into account how the non\sphinxhyphen{}linear model equations change the state of the system over the course of 20 time steps. So the optimization goal is quite difficult, and it is not possible to exactly satisfy the constraints to match the reference simulation in this scenario. E.g., this is noticeable at the stems of the smoke plumes, which still show a black halo after the optimization. The optimization was not able to shift the inflow position, and hence needs to focus on aligning the upper regions of the plumes.

\section{Conclusions}
\label{\detokenize{diffphys-code-ns:conclusions}}
This example illustrated how the differentiable physics approach can easily be extended towards significantly more
complex PDEs. Above, we’ve optimized for a mini\sphinxhyphen{}batch of 20 steps of a full Navier\sphinxhyphen{}Stokes solver.

This is a powerful basis to bring NNs into the picture. As you might have noticed, our degrees of freedom were still a regular grid, and we’ve jointly solved a single inverse problem. There were three cases to solve as a mini\sphinxhyphen{}batch, of course, but nonetheless the setup still represents a direct optimization. Thus, in line with the PINN example of {\hyperref[\detokenize{physicalloss-code::doc}]{\sphinxcrossref{\DUrole{doc}{Burgers Optimization with a Physics\sphinxhyphen{}Informed NN}}}} we’ve not really dealt with a \sphinxstyleemphasis{machine learning} task here. However, that will be the topic of the next chapters.

\section{Next steps}
\label{\detokenize{diffphys-code-ns:next-steps}}
Based on the code example above, we can recommend experimenting with the following:
\begin{itemize}
\item {} 
Modify the setup of the simulation to differ more strongly across the four instances, run longer, or use a finer spatial discretization (i.e. larger grid size). Note that this will make the optimization problem tougher, and hence might not converge directly with this simple setup.

\item {} 
As a larger change, add a multi\sphinxhyphen{}resolution optimization to handle cases with larger differences. I.e., first solve with a coarse discretization, and then uses this solution as an initial guess for a finer discretization.

\end{itemize}

\chapter{Differentiable Physics versus Physics\sphinxhyphen{}informed Training}
\label{\detokenize{diffphys-dpvspinn:differentiable-physics-versus-physics-informed-training}}\label{\detokenize{diffphys-dpvspinn::doc}}
In the previous sections we’ve seen example reconstructions that used physical residuals as soft constraints, in the form of the PINNs, and reconstructions that used a differentiable physics (DP) solver. While both methods can find minimizers for similar inverse problems, the obtained solutions differ substantially, as does the behavior of the non\sphinxhyphen{}linear optimization problem that we get from each formulation. In the following we discuss these differences in more detail, and we will combine conclusions drawn from the behavior of the Burgers case of {\hyperref[\detokenize{physicalloss-code::doc}]{\sphinxcrossref{\DUrole{doc}{Burgers Optimization with a Physics\sphinxhyphen{}Informed NN}}}} and {\hyperref[\detokenize{diffphys-code-burgers::doc}]{\sphinxcrossref{\DUrole{doc}{Burgers Optimization with a Differentiable Physics Gradient}}}} with observations from research papers.

\sphinxincludegraphics{{divider3}.jpg}

\section{Compatibility with existing numerical methods}
\label{\detokenize{diffphys-dpvspinn:compatibility-with-existing-numerical-methods}}
It is very obvious that the PINN implementation is quite simple, which is a positive aspect, but at the same time it differs strongly from “typical” discretizations and solution approaches that are usually employed to solve PDEs like Burgers equation. The derivatives are computed via the neural network, and hence rely on a fairly accurate representation of the solution to provide a good direction for optimization problems.

The DP version on the other hand inherently relies on a numerical solver that is tied into the learning process. As such it requires a discretization of the problem at hand, and via this discretization can employ existing, and potentially powerful numerical techniques. This means solutions and derivatives can be evaluated with known and controllable accuracy, and can be evaluated efficiently.

\section{Discretization}
\label{\detokenize{diffphys-dpvspinn:discretization}}
The reliance on a suitable discretization requires some understanding and knowledge of the problem under consideration. A sub\sphinxhyphen{}optimal discretization can impede the learning process or, worst case, lead to diverging training runs. However, given the large body of theory and practical realizations of stable solvers for a wide variety of physical problems, this is typically not an unsurmountable obstacle.

The PINN approaches on the other hand do not require an a\sphinxhyphen{}priori choice of a discretization, and as such seems to be “discretization\sphinxhyphen{}less”. This, however, is only an advantage on first sight. As they yield solutions in a computer, they naturally \sphinxstyleemphasis{have} to discretize the problem, but they construct this discretization over the course of the training process, in a way that is not easily controllable from the outside. Thus, the resulting accuracy is determined by how well the training manages to estimate the complexity of the problem for realistic use cases, and how well the training data approximates the unknown regions of the solution.

As demonstrated with the Burgers example, the PINN solutions typically have significant difficulties propagating information \sphinxstyleemphasis{backward} in time. This is closely coupled to the efficiency of the method.

\section{Efficiency}
\label{\detokenize{diffphys-dpvspinn:efficiency}}
The PINN approaches typically perform a localized sampling and correction of the solutions, which means the corrections in the form of weight updates are likewise typically local. The fulfillment of boundary conditions in space and time can be correspondingly slow, leading to long training runs in practice.

A well\sphinxhyphen{}chosen discretization of a DP approach can remedy this behavior, and provide an improved flow of gradient information. At the same time, the reliance on a computational grid means that solutions can be obtained very quickly. Given an interpolation scheme or a set of basis functions, the solution can be sampled at any point in space or time given a very local neighborhood of the computational grid. Worst case, this can lead to slight memory overheads, e.g., by repeatedly storing mostly constant values of a solution.

For the PINN representation with fully\sphinxhyphen{}connected networks on the other hand, we need to make a full pass over the potentially large number of values in the whole network to obtain a sample of the solution at a single point. The network effectively needs to encode the full high\sphinxhyphen{}dimensional solution, and its size likewise determines the efficiency of derivative calculations.

\section{Efficiency continued}
\label{\detokenize{diffphys-dpvspinn:efficiency-continued}}
That being said, because the DP approaches can cover much larger solution manifolds, the structure of these manifolds is typically also difficult to learn. E.g., when training a network with a larger number of iterations (i.e. a long look\sphinxhyphen{}ahead into the future), this typically represents a signal that is more difficult to learn than a short look ahead.

As a consequence, these training runs not only take more computational resources per NN iteration, they also need longer to converge. Regarding resources, each computation of the look\sphinxhyphen{}ahead potentially requires a large number of simulation steps, and typically a similar amount of resources for the backpropagation step. Regarding convergence, the more complex signal that should be learned can take more training iterations and require larger NN structures.

\sphinxincludegraphics{{divider2}.jpg}

\section{Summary}
\label{\detokenize{diffphys-dpvspinn:summary}}
The following table summarizes these pros and cons of physics\sphinxhyphen{}informed (PI) and differentiable physics (DP) approaches:

\begin{savenotes}\sphinxattablestart
\centering
\begin{tabulary}{\linewidth}[t]{|T|T|T|}
\hline
\sphinxstyletheadfamily 
Method
&\sphinxstyletheadfamily 
✅ Pro
&\sphinxstyletheadfamily 
❌ Con
\\
\hline
\sphinxstylestrong{PI}
&
\sphinxhyphen{} Analytic derivatives via backpropagation.
&
\sphinxhyphen{} Expensive evaluation of NN, as well as derivative calculations.
\\
\hline

&
\sphinxhyphen{} Easy to implement.
&
\sphinxhyphen{} Incompatible with existing numerical methods.
\\
\hline

&

&
\sphinxhyphen{} No control of discretization.
\\
\hline

&

&

\\
\hline
\sphinxstylestrong{DP}
&
\sphinxhyphen{} Leverage existing numerical methods.
&
\sphinxhyphen{} More complicated implementation.
\\
\hline

&
\sphinxhyphen{} Efficient evaluation of simulation and derivatives.
&
\sphinxhyphen{} Require understanding of problem to choose suitable discretization.
\\
\hline

&

&

\\
\hline
\end{tabulary}
\par
\sphinxattableend\end{savenotes}

As a summary, both methods are definitely interesting, and have a lot of potential. There are numerous more complicated extensions and algorithmic modifications that change and improve on the various negative aspects we have discussed for both sides.

However, as of this writing, the physics\sphinxhyphen{}informed (PI) approach has clear limitations when it comes to performance and compatibility with existing numerical methods. Thus, when knowledge of the problem at hand is available, which typically is the case when we choose a suitable PDE model to constrain the learning process, employing a differentiable physics  solver can significantly improve the training process as well as the quality of the obtained solution. Next, we will target more complex settings, i.e., fluids with Navier\sphinxhyphen{}Stokes, to illustrate this in more detail.

\part{Complex Examples with DP}

\chapter{Complex Examples Overview}
\label{\detokenize{diffphys-examples:complex-examples-overview}}\label{\detokenize{diffphys-examples::doc}}
The following sections will give code examples of more complex cases to
show what can be achieved via differentiable physics training.

First, we’ll show a scenario that employs deep learning to represent the errors
of numerical simulations, following Um et al. {[}\hyperlink{cite.references:id6}{UBH+20}{]}.
This is a very fundamental task, and requires the learned model to closely
interact with a numerical solver. Hence, it’s a prime example of
situations where it’s crucial to bring the numerical solver into the
deep learning loop.

Next, we’ll show how to let NNs solve tough inverse problems, namely the long\sphinxhyphen{}term control
of a Navier\sphinxhyphen{}Stokes simulation, following Holl et al.  {[}\hyperlink{cite.references:id11}{HKT19}{]}.
This task requires long term planning,
and hence needs two networks, one to \sphinxstyleemphasis{predict} the evolution,
and another one to \sphinxstyleemphasis{act} to reach the desired goal. (Later on, in {\hyperref[\detokenize{reinflearn-code::doc}]{\sphinxcrossref{\DUrole{doc}{Controlling Burgers’ Equation with Reinforcement Learning}}}} we will compare
this approach to another DL variant using reinforcement learning.)

Both cases require quite a bit more resources than the previous examples, so you
can expect these notebooks to run longer (and it’s a good idea to use check\sphinxhyphen{}pointing
when working with these examples).

\chapter{Reducing Numerical Errors with Deep Learning}
\label{\detokenize{diffphys-code-sol:reducing-numerical-errors-with-deep-learning}}\label{\detokenize{diffphys-code-sol::doc}}
First, we’ll target numerical errors that arise in the discretization of a continuous PDE \(\mathcal P^*\), i.e. when we formulate \(\mathcal P\). This approach will demonstrate that, despite the lack of closed\sphinxhyphen{}form descriptions, discretization errors often are functions with regular and repeating structures and, thus, can be learned by a neural network. Once the network is trained, it can be evaluated locally to improve the solution of a PDE\sphinxhyphen{}solver, i.e., to reduce its numerical error. The resulting method is a hybrid one: it will always run (a coarse) PDE solver, and then improve it at runtime with corrections inferred by an NN.

Pretty much all numerical methods contain some form of iterative process: repeated updates over time for explicit solvers, or within a single update step for implicit solvers.
An example for the second case could be found \sphinxhref{https://github.com/tum-pbs/CG-Solver-in-the-Loop}{here},
but below we’ll target the first case, i.e. iterations over time.
\sphinxhref{https://colab.research.google.com/github/tum-pbs/pbdl-book/blob/main/diffphys-code-sol.ipynb}{{[}run in colab{]}}

\section{Problem formulation}
\label{\detokenize{diffphys-code-sol:problem-formulation}}
In the context of reducing errors, it’s crucial to have a \sphinxstyleemphasis{differentiable physics solver}, so that the learning process can take the reaction of the solver into account. This interaction is not possible with supervised learning or PINN training. Even small inference errors of a supervised NN can accumulate over time, and lead to a data distribution that differs from the distribution of the pre\sphinxhyphen{}computed data. This distribution shift can lead to sub\sphinxhyphen{}optimal results, or even cause blow\sphinxhyphen{}ups of the solver.

In order to learn the error function, we’ll consider two different discretizations of the same PDE \(\mathcal P^*\):
a \sphinxstyleemphasis{reference} version, which we assume to be accurate, with a discretized version
\(\mathcal P_r\), and solutions \(\mathbf r \in \mathscr R\), where \(\mathscr R\) denotes the manifold of solutions of \(\mathcal P_r\).
In parallel to this, we have a less accurate approximation of the same PDE, which we’ll refer to as the \sphinxstyleemphasis{source} version, as this will be the solver that our NN should later on interact with. Analogously,
we have \(\mathcal P_s\) with solutions \(\mathbf s \in \mathscr S\).
After training, we’ll obtain a \sphinxstyleemphasis{hybrid} solver that uses \(\mathcal P_s\) in conjunction with a trained network to obtain improved solutions, i.e., solutions that are closer to the ones produced by \(\mathcal P_r\).

\begin{figure}[htbp]
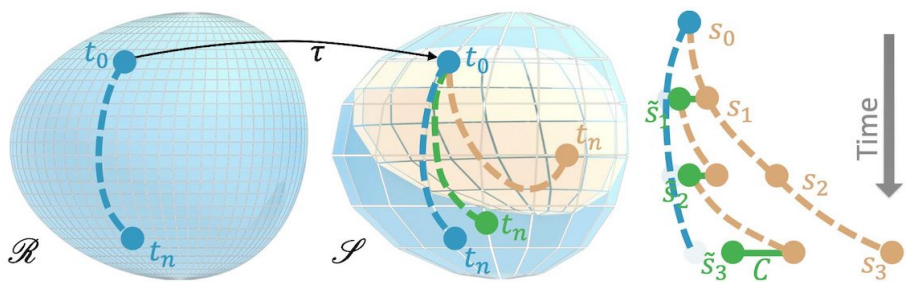

\centering
\capstart

\noindent\sphinxincludegraphics[height=150\sphinxpxdimen]{{diffphys-sol-manifolds}.jpeg}
\caption{Visual overview of coarse and reference manifolds}\label{\detokenize{diffphys-code-sol:diffphys-sol-manifolds}}\end{figure}

Let’s assume \(\mathcal{P}\) advances a solution by a time step \(\Delta t\), and let’s denote \(n\) consecutive steps by a superscript:
\(
\newcommand{\pde}{\mathcal{P}}
\newcommand{\pdec}{\pde_{s}}
\newcommand{\vc}[1]{\mathbf{s}_{#1}} 
\newcommand{\vr}[1]{\mathbf{r}_{#1}} 
\newcommand{\vcN}{\vs}          
\newcommand{\project}{\mathcal{T}}   
\pdec^n ( \mathcal{T} \vr{t} ) = \pdec(\pdec(\cdots \pdec( \mathcal{T} \vr{t}  )\cdots)) .
\)
The corresponding state of the simulation is
\(
\mathbf{s}_{t+n} = \mathcal{P}^n ( \mathcal{T} \mathbf{r}_{t} ) .
\)
Here we assume a mapping operator \(\mathcal{T}\) exists that transfers a reference solution to the source manifold. This could, e.g., be a simple downsampling operation.
Especially for longer sequences, i.e. larger \(n\), the source state
\(\newcommand{\vc}[1]{\mathbf{s}_{#1}} \vc{t+n}\)
will deviate from a corresponding reference state
\(\newcommand{\vr}[1]{\mathbf{r}_{#1}} \vr{t+n}\).
This is what we will address with an NN in the following.

As before, we’ll use an \(L^2\)\sphinxhyphen{}norm to quantify the deviations, i.e.,
an error function \(\newcommand{\loss}{e} 
\newcommand{\corr}{\mathcal{C}} 
\newcommand{\vc}[1]{\mathbf{s}_{#1}} 
\newcommand{\vr}[1]{\mathbf{r}_{#1}} 
\loss (\vc{t},\mathcal{T} \vr{t})=\Vert\vc{t}-\mathcal{T} \vr{t}\Vert_2\).
Our learning goal is to train at a correction operator
\(\mathcal{C} ( \mathbf{s} )\) such that
a solution to which the correction is applied has a lower error than the original unmodified (source)
solution: \(\newcommand{\loss}{e} 
\newcommand{\corr}{\mathcal{C}} 
\newcommand{\vr}[1]{\mathbf{r}_{#1}} 
\loss ( \mathcal{P}_{s}( \corr (\mathcal{T} \vr{t}) ) , \mathcal{T} \vr{t+1}) < \loss ( \mathcal{P}_{s}( \mathcal{T} \vr{t} ), \mathcal{T} \vr{t+1})\).

The correction function
\(\newcommand{\vcN}{\mathbf{s}} \newcommand{\corr}{\mathcal{C}} \corr (\vcN | \theta)\)
is represented as a deep neural network with weights \(\theta\)
and receives the state \(\mathbf{s}\) to infer an additive correction field with the same dimension.
To distinguish the original states \(\mathbf{s}\) from the corrected ones, we’ll denote the latter with an added tilde \(\tilde{\mathbf{s}}\).
The overall learning goal now becomes
\begin{equation*}
\begin{split}
\newcommand{\corr}{\mathcal{C}}  
\newcommand{\vr}[1]{\mathbf{r}_{#1}} 
\text{arg min}_\theta \big( ( \mathcal{P}_{s} \corr )^n ( \mathcal{T} \vr{t} ) - \mathcal{T} \vr{t+n} \big)^2
\end{split}
\end{equation*}
To simplify the notation, we’ve dropped the sum over different samples here (the \(i\) from previous versions).
A crucial bit that’s easy to overlook in the equation above, is that the correction depends on the modified states, i.e.
it is a function of
\(\tilde{\mathbf{s}}\), so we have
\(\newcommand{\vctN}{\tilde{\mathbf{s}}} \newcommand{\corr}{\mathcal{C}} \corr (\vctN | \theta)\).
These states actually evolve over time when training. They don’t exist beforehand.

\sphinxstylestrong{TL;DR}:
We’ll train a network \(\mathcal{C}\) to reduce the numerical errors of a simulator with a more accurate reference. It’s crucial to have the \sphinxstyleemphasis{source} solver realized as a differential physics operator, such that it can give gradients for an improved training of \(\mathcal{C}\).

\bigskip\hrule\bigskip

\section{Getting started with the implementation}
\label{\detokenize{diffphys-code-sol:getting-started-with-the-implementation}}
The following replicates an experiment) from \sphinxhref{https://ge.in.tum.de/publications/2020-um-solver-in-the-loop/}{Solver\sphinxhyphen{}in\sphinxhyphen{}the\sphinxhyphen{}loop: learning from differentiable physics to interact with iterative pde\sphinxhyphen{}solvers} {[}\hyperlink{cite.references:id11}{HKT19}{]}, further details can be found in section B.1 of the \sphinxhref{https://arxiv.org/pdf/2007.00016.pdf}{appendix} of the paper.

First, let’s download the prepared data set (for details on generation \& loading cf. \sphinxurl{https://github.com/tum-pbs/Solver-in-the-Loop}), and let’s get the data handling out of the way, so that we can focus on the \sphinxstyleemphasis{interesting} parts…

\begin{sphinxVerbatim}[commandchars=\\\{\}]
\PYG{k+kn}{import} \PYG{n+nn}{os}\PYG{o}{,} \PYG{n+nn}{sys}\PYG{o}{,} \PYG{n+nn}{logging}\PYG{o}{,} \PYG{n+nn}{argparse}\PYG{o}{,} \PYG{n+nn}{pickle}\PYG{o}{,} \PYG{n+nn}{glob}\PYG{o}{,} \PYG{n+nn}{random}\PYG{o}{,} \PYG{n+nn}{distutils}\PYG{n+nn}{.}\PYG{n+nn}{dir\PYGZus{}util}\PYG{o}{,} \PYG{n+nn}{urllib}\PYG{n+nn}{.}\PYG{n+nn}{request}

\PYG{n}{fname\PYGZus{}train} \PYG{o}{=} \PYG{l+s+s1}{\PYGZsq{}}\PYG{l+s+s1}{sol\PYGZhy{}karman\PYGZhy{}2d\PYGZhy{}train.pickle}\PYG{l+s+s1}{\PYGZsq{}}
\PYG{k}{if} \PYG{o+ow}{not} \PYG{n}{os}\PYG{o}{.}\PYG{n}{path}\PYG{o}{.}\PYG{n}{isfile}\PYG{p}{(}\PYG{n}{fname\PYGZus{}train}\PYG{p}{)}\PYG{p}{:}
  \PYG{n+nb}{print}\PYG{p}{(}\PYG{l+s+s2}{\PYGZdq{}}\PYG{l+s+s2}{Downloading training data (73MB), this can take a moment the first time...}\PYG{l+s+s2}{\PYGZdq{}}\PYG{p}{)}
  \PYG{n}{urllib}\PYG{o}{.}\PYG{n}{request}\PYG{o}{.}\PYG{n}{urlretrieve}\PYG{p}{(}\PYG{l+s+s2}{\PYGZdq{}}\PYG{l+s+s2}{https://physicsbaseddeeplearning.org/data/}\PYG{l+s+s2}{\PYGZdq{}}\PYG{o}{+}\PYG{n}{fname\PYGZus{}train}\PYG{p}{,} \PYG{n}{fname\PYGZus{}train}\PYG{p}{)}

\PYG{k}{with} \PYG{n+nb}{open}\PYG{p}{(}\PYG{n}{fname\PYGZus{}train}\PYG{p}{,} \PYG{l+s+s1}{\PYGZsq{}}\PYG{l+s+s1}{rb}\PYG{l+s+s1}{\PYGZsq{}}\PYG{p}{)} \PYG{k}{as} \PYG{n}{f}\PYG{p}{:} \PYG{n}{data\PYGZus{}preloaded} \PYG{o}{=} \PYG{n}{pickle}\PYG{o}{.}\PYG{n}{load}\PYG{p}{(}\PYG{n}{f}\PYG{p}{)}
\PYG{n+nb}{print}\PYG{p}{(}\PYG{l+s+s2}{\PYGZdq{}}\PYG{l+s+s2}{Loaded data, }\PYG{l+s+si}{\PYGZob{}\PYGZcb{}}\PYG{l+s+s2}{ training sims}\PYG{l+s+s2}{\PYGZdq{}}\PYG{o}{.}\PYG{n}{format}\PYG{p}{(}\PYG{n+nb}{len}\PYG{p}{(}\PYG{n}{data\PYGZus{}preloaded}\PYG{p}{)}\PYG{p}{)} \PYG{p}{)}
\end{sphinxVerbatim}

\begin{sphinxVerbatim}[commandchars=\\\{\}]
Downloading training data (73MB), this can take a moment the first time...
Loaded data, 6 training sims
\end{sphinxVerbatim}

Also let’s get installing / importing all the necessary libraries out of the way. And while we’re at it, we can set the random seed \sphinxhyphen{} obviously, 42 is the ultimate choice here 🙂

\begin{sphinxVerbatim}[commandchars=\\\{\}]
\PYG{c+ch}{\PYGZsh{}!pip install \PYGZhy{}\PYGZhy{}upgrade \PYGZhy{}\PYGZhy{}quiet phiflow}
\PYG{c+c1}{\PYGZsh{}!pip uninstall phiflow}
\PYG{o}{!}pip install \PYGZhy{}\PYGZhy{}upgrade \PYGZhy{}\PYGZhy{}quiet git+https://github.com/tum\PYGZhy{}pbs/PhiFlow@develop

\PYG{k+kn}{from} \PYG{n+nn}{phi}\PYG{n+nn}{.}\PYG{n+nn}{tf}\PYG{n+nn}{.}\PYG{n+nn}{flow} \PYG{k+kn}{import} \PYG{o}{*}
\PYG{k+kn}{import} \PYG{n+nn}{tensorflow} \PYG{k}{as} \PYG{n+nn}{tf}
\PYG{k+kn}{from} \PYG{n+nn}{tensorflow} \PYG{k+kn}{import} \PYG{n}{keras}

\PYG{n}{random}\PYG{o}{.}\PYG{n}{seed}\PYG{p}{(}\PYG{l+m+mi}{42}\PYG{p}{)} 
\PYG{n}{np}\PYG{o}{.}\PYG{n}{random}\PYG{o}{.}\PYG{n}{seed}\PYG{p}{(}\PYG{l+m+mi}{42}\PYG{p}{)}
\PYG{n}{tf}\PYG{o}{.}\PYG{n}{random}\PYG{o}{.}\PYG{n}{set\PYGZus{}seed}\PYG{p}{(}\PYG{l+m+mi}{42}\PYG{p}{)}
\end{sphinxVerbatim}

\begin{sphinxVerbatim}[commandchars=\\\{\}]

\end{sphinxVerbatim}

\section{Simulation setup}
\label{\detokenize{diffphys-code-sol:simulation-setup}}
Now we can set up the \sphinxstyleemphasis{source} simulation \(\mathcal{P}_{s}\).
Note that we won’t deal with
\(\mathcal{P}_{r}\)
below: the downsampled reference data is contained in the training data set. It was generated with a four times finer discretization. Below we’re focusing on the interaction of the source solver and the NN.

This code block and the next ones will define lots of functions, that will be used later on for training.

The \sphinxcode{\sphinxupquote{KarmanFlow}} solver below simulates a relatively standard wake flow case with a spherical obstacle in a rectangular domain, and an explicit viscosity solve to obtain different Reynolds numbers. This is the geometry of the setup:

\begin{figure}[htbp]
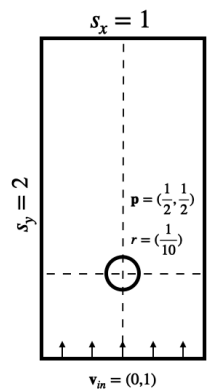

\centering
\capstart

\noindent\sphinxincludegraphics[height=200\sphinxpxdimen]{{diffphys-sol-domain}.png}
\caption{Domain setup for the wake flow case (sizes in the imlpementation are using an additional factor of 100).}\label{\detokenize{diffphys-code-sol:diffphys-sol-domain}}\end{figure}

The solver applies inflow boundary conditions for the y\sphinxhyphen{}velocity with a pre\sphinxhyphen{}multiplied mask (\sphinxcode{\sphinxupquote{vel\_BcMask}}), to set the y components at the bottom of the domain during the simulation step. This mask is created with the \sphinxcode{\sphinxupquote{HardGeometryMask}} from phiflow, which initializes the spatially shifted entries for the components of a staggered grid correctly. The simulation step is quite straight forward: it computes contributions for viscosity, inflow, advection and finally makes the resulting motion divergence free via an implicit pressure solve:

\begin{sphinxVerbatim}[commandchars=\\\{\}]
\PYG{k}{class} \PYG{n+nc}{KarmanFlow}\PYG{p}{(}\PYG{p}{)}\PYG{p}{:}
    \PYG{k}{def} \PYG{n+nf+fm}{\PYGZus{}\PYGZus{}init\PYGZus{}\PYGZus{}}\PYG{p}{(}\PYG{n+nb+bp}{self}\PYG{p}{,} \PYG{n}{domain}\PYG{p}{)}\PYG{p}{:}
        \PYG{n+nb+bp}{self}\PYG{o}{.}\PYG{n}{domain} \PYG{o}{=} \PYG{n}{domain}

        \PYG{n+nb+bp}{self}\PYG{o}{.}\PYG{n}{vel\PYGZus{}BcMask} \PYG{o}{=} \PYG{n+nb+bp}{self}\PYG{o}{.}\PYG{n}{domain}\PYG{o}{.}\PYG{n}{staggered\PYGZus{}grid}\PYG{p}{(}\PYG{n}{HardGeometryMask}\PYG{p}{(}\PYG{n}{Box}\PYG{p}{[}\PYG{p}{:}\PYG{l+m+mi}{5}\PYG{p}{,} \PYG{p}{:}\PYG{p}{]}\PYG{p}{)} \PYG{p}{)}
    
        \PYG{n+nb+bp}{self}\PYG{o}{.}\PYG{n}{inflow} \PYG{o}{=} \PYG{n+nb+bp}{self}\PYG{o}{.}\PYG{n}{domain}\PYG{o}{.}\PYG{n}{scalar\PYGZus{}grid}\PYG{p}{(}\PYG{n}{Box}\PYG{p}{[}\PYG{l+m+mi}{5}\PYG{p}{:}\PYG{l+m+mi}{10}\PYG{p}{,} \PYG{l+m+mi}{25}\PYG{p}{:}\PYG{l+m+mi}{75}\PYG{p}{]}\PYG{p}{)}         \PYG{c+c1}{\PYGZsh{} scale with domain if necessary!}
        \PYG{n+nb+bp}{self}\PYG{o}{.}\PYG{n}{obstacles} \PYG{o}{=} \PYG{p}{[}\PYG{n}{Obstacle}\PYG{p}{(}\PYG{n}{Sphere}\PYG{p}{(}\PYG{n}{center}\PYG{o}{=}\PYG{p}{[}\PYG{l+m+mi}{50}\PYG{p}{,} \PYG{l+m+mi}{50}\PYG{p}{]}\PYG{p}{,} \PYG{n}{radius}\PYG{o}{=}\PYG{l+m+mi}{10}\PYG{p}{)}\PYG{p}{)}\PYG{p}{]} 

    \PYG{k}{def} \PYG{n+nf}{step}\PYG{p}{(}\PYG{n+nb+bp}{self}\PYG{p}{,} \PYG{n}{density\PYGZus{}in}\PYG{p}{,} \PYG{n}{velocity\PYGZus{}in}\PYG{p}{,} \PYG{n}{re}\PYG{p}{,} \PYG{n}{res}\PYG{p}{,} \PYG{n}{buoyancy\PYGZus{}factor}\PYG{o}{=}\PYG{l+m+mi}{0}\PYG{p}{,} \PYG{n}{dt}\PYG{o}{=}\PYG{l+m+mf}{1.0}\PYG{p}{)}\PYG{p}{:}
        \PYG{n}{velocity} \PYG{o}{=} \PYG{n}{velocity\PYGZus{}in}
        \PYG{n}{density} \PYG{o}{=} \PYG{n}{density\PYGZus{}in}

        \PYG{c+c1}{\PYGZsh{} viscosity}
        \PYG{n}{velocity} \PYG{o}{=} \PYG{n}{phi}\PYG{o}{.}\PYG{n}{flow}\PYG{o}{.}\PYG{n}{diffuse}\PYG{o}{.}\PYG{n}{explicit}\PYG{p}{(}\PYG{n}{field}\PYG{o}{=}\PYG{n}{velocity}\PYG{p}{,} \PYG{n}{diffusivity}\PYG{o}{=}\PYG{l+m+mf}{1.0}\PYG{o}{/}\PYG{n}{re}\PYG{o}{*}\PYG{n}{dt}\PYG{o}{*}\PYG{n}{res}\PYG{o}{*}\PYG{n}{res}\PYG{p}{,} \PYG{n}{dt}\PYG{o}{=}\PYG{n}{dt}\PYG{p}{)}
        
        \PYG{c+c1}{\PYGZsh{} inflow boundary conditions}
        \PYG{n}{velocity} \PYG{o}{=} \PYG{n}{velocity}\PYG{o}{*}\PYG{p}{(}\PYG{l+m+mf}{1.0} \PYG{o}{\PYGZhy{}} \PYG{n+nb+bp}{self}\PYG{o}{.}\PYG{n}{vel\PYGZus{}BcMask}\PYG{p}{)} \PYG{o}{+} \PYG{n+nb+bp}{self}\PYG{o}{.}\PYG{n}{vel\PYGZus{}BcMask} \PYG{o}{*} \PYG{p}{(}\PYG{l+m+mi}{1}\PYG{p}{,}\PYG{l+m+mi}{0}\PYG{p}{)}

        \PYG{c+c1}{\PYGZsh{} advection }
        \PYG{n}{density} \PYG{o}{=} \PYG{n}{advect}\PYG{o}{.}\PYG{n}{semi\PYGZus{}lagrangian}\PYG{p}{(}\PYG{n}{density}\PYG{o}{+}\PYG{n+nb+bp}{self}\PYG{o}{.}\PYG{n}{inflow}\PYG{p}{,} \PYG{n}{velocity}\PYG{p}{,} \PYG{n}{dt}\PYG{o}{=}\PYG{n}{dt}\PYG{p}{)}
        \PYG{n}{velocity} \PYG{o}{=} \PYG{n}{advected\PYGZus{}velocity} \PYG{o}{=} \PYG{n}{advect}\PYG{o}{.}\PYG{n}{semi\PYGZus{}lagrangian}\PYG{p}{(}\PYG{n}{velocity}\PYG{p}{,} \PYG{n}{velocity}\PYG{p}{,} \PYG{n}{dt}\PYG{o}{=}\PYG{n}{dt}\PYG{p}{)}

        \PYG{c+c1}{\PYGZsh{} mass conservation (pressure solve)}
        \PYG{n}{pressure} \PYG{o}{=} \PYG{k+kc}{None}
        \PYG{n}{velocity}\PYG{p}{,} \PYG{n}{pressure} \PYG{o}{=} \PYG{n}{fluid}\PYG{o}{.}\PYG{n}{make\PYGZus{}incompressible}\PYG{p}{(}\PYG{n}{velocity}\PYG{p}{,} \PYG{n+nb+bp}{self}\PYG{o}{.}\PYG{n}{obstacles}\PYG{p}{)}
        \PYG{n+nb+bp}{self}\PYG{o}{.}\PYG{n}{solve\PYGZus{}info} \PYG{o}{=} \PYG{p}{\PYGZob{}} \PYG{l+s+s1}{\PYGZsq{}}\PYG{l+s+s1}{pressure}\PYG{l+s+s1}{\PYGZsq{}}\PYG{p}{:} \PYG{n}{pressure}\PYG{p}{,} \PYG{l+s+s1}{\PYGZsq{}}\PYG{l+s+s1}{advected\PYGZus{}velocity}\PYG{l+s+s1}{\PYGZsq{}}\PYG{p}{:} \PYG{n}{advected\PYGZus{}velocity} \PYG{p}{\PYGZcb{}}
        
        \PYG{k}{return} \PYG{p}{[}\PYG{n}{density}\PYG{p}{,} \PYG{n}{velocity}\PYG{p}{]}

\end{sphinxVerbatim}

\section{Network architecture}
\label{\detokenize{diffphys-code-sol:network-architecture}}
We’ll also define two alternative versions of a neural networks to represent
\(\newcommand{\vcN}{\mathbf{s}} \newcommand{\corr}{\mathcal{C}} \corr\). In both cases we’ll use fully convolutional networks, i.e. networks without any fully\sphinxhyphen{}connected layers. We’ll use Keras within tensorflow to define the layers of the network (mostly via \sphinxcode{\sphinxupquote{Conv2D}}), typically activated via ReLU and LeakyReLU functions, respectively.
The inputs to the network are:
\begin{itemize}
\item {} 
2 fields with x,y velocity

\item {} 
the Reynolds number as constant channel.

\end{itemize}

The output is:
\begin{itemize}
\item {} 
a 2 component field containing the x,y velocity.

\end{itemize}

First, let’s define a small network consisting only of four convolutional layers with ReLU activations (we’re also using keras here for simplicity). The input dimensions are determined from input tensor in the \sphinxcode{\sphinxupquote{inputs\_dict}} (it has three channels: u,v, and Re). Then we process the data via three conv layers with 32 features each, before reducing to 2 channels in the output.

\begin{sphinxVerbatim}[commandchars=\\\{\}]
\PYG{k}{def} \PYG{n+nf}{network\PYGZus{}small}\PYG{p}{(}\PYG{n}{inputs\PYGZus{}dict}\PYG{p}{)}\PYG{p}{:}
    \PYG{n}{l\PYGZus{}input} \PYG{o}{=} \PYG{n}{keras}\PYG{o}{.}\PYG{n}{layers}\PYG{o}{.}\PYG{n}{Input}\PYG{p}{(}\PYG{o}{*}\PYG{o}{*}\PYG{n}{inputs\PYGZus{}dict}\PYG{p}{)}
    \PYG{n}{block\PYGZus{}0} \PYG{o}{=} \PYG{n}{keras}\PYG{o}{.}\PYG{n}{layers}\PYG{o}{.}\PYG{n}{Conv2D}\PYG{p}{(}\PYG{n}{filters}\PYG{o}{=}\PYG{l+m+mi}{32}\PYG{p}{,} \PYG{n}{kernel\PYGZus{}size}\PYG{o}{=}\PYG{l+m+mi}{5}\PYG{p}{,} \PYG{n}{padding}\PYG{o}{=}\PYG{l+s+s1}{\PYGZsq{}}\PYG{l+s+s1}{same}\PYG{l+s+s1}{\PYGZsq{}}\PYG{p}{)}\PYG{p}{(}\PYG{n}{l\PYGZus{}input}\PYG{p}{)}
    \PYG{n}{block\PYGZus{}0} \PYG{o}{=} \PYG{n}{keras}\PYG{o}{.}\PYG{n}{layers}\PYG{o}{.}\PYG{n}{LeakyReLU}\PYG{p}{(}\PYG{p}{)}\PYG{p}{(}\PYG{n}{block\PYGZus{}0}\PYG{p}{)}

    \PYG{n}{l\PYGZus{}conv1} \PYG{o}{=} \PYG{n}{keras}\PYG{o}{.}\PYG{n}{layers}\PYG{o}{.}\PYG{n}{Conv2D}\PYG{p}{(}\PYG{n}{filters}\PYG{o}{=}\PYG{l+m+mi}{32}\PYG{p}{,} \PYG{n}{kernel\PYGZus{}size}\PYG{o}{=}\PYG{l+m+mi}{5}\PYG{p}{,} \PYG{n}{padding}\PYG{o}{=}\PYG{l+s+s1}{\PYGZsq{}}\PYG{l+s+s1}{same}\PYG{l+s+s1}{\PYGZsq{}}\PYG{p}{)}\PYG{p}{(}\PYG{n}{block\PYGZus{}0}\PYG{p}{)}
    \PYG{n}{l\PYGZus{}conv1} \PYG{o}{=} \PYG{n}{keras}\PYG{o}{.}\PYG{n}{layers}\PYG{o}{.}\PYG{n}{LeakyReLU}\PYG{p}{(}\PYG{p}{)}\PYG{p}{(}\PYG{n}{l\PYGZus{}conv1}\PYG{p}{)}
    \PYG{n}{l\PYGZus{}conv2} \PYG{o}{=} \PYG{n}{keras}\PYG{o}{.}\PYG{n}{layers}\PYG{o}{.}\PYG{n}{Conv2D}\PYG{p}{(}\PYG{n}{filters}\PYG{o}{=}\PYG{l+m+mi}{32}\PYG{p}{,} \PYG{n}{kernel\PYGZus{}size}\PYG{o}{=}\PYG{l+m+mi}{5}\PYG{p}{,} \PYG{n}{padding}\PYG{o}{=}\PYG{l+s+s1}{\PYGZsq{}}\PYG{l+s+s1}{same}\PYG{l+s+s1}{\PYGZsq{}}\PYG{p}{)}\PYG{p}{(}\PYG{n}{l\PYGZus{}conv1}\PYG{p}{)}
    \PYG{n}{block\PYGZus{}1} \PYG{o}{=} \PYG{n}{keras}\PYG{o}{.}\PYG{n}{layers}\PYG{o}{.}\PYG{n}{LeakyReLU}\PYG{p}{(}\PYG{p}{)}\PYG{p}{(}\PYG{n}{l\PYGZus{}conv2}\PYG{p}{)}

    \PYG{n}{l\PYGZus{}output} \PYG{o}{=} \PYG{n}{keras}\PYG{o}{.}\PYG{n}{layers}\PYG{o}{.}\PYG{n}{Conv2D}\PYG{p}{(}\PYG{n}{filters}\PYG{o}{=}\PYG{l+m+mi}{2}\PYG{p}{,}  \PYG{n}{kernel\PYGZus{}size}\PYG{o}{=}\PYG{l+m+mi}{5}\PYG{p}{,} \PYG{n}{padding}\PYG{o}{=}\PYG{l+s+s1}{\PYGZsq{}}\PYG{l+s+s1}{same}\PYG{l+s+s1}{\PYGZsq{}}\PYG{p}{)}\PYG{p}{(}\PYG{n}{block\PYGZus{}1}\PYG{p}{)} \PYG{c+c1}{\PYGZsh{} u, v}
    \PYG{k}{return} \PYG{n}{keras}\PYG{o}{.}\PYG{n}{models}\PYG{o}{.}\PYG{n}{Model}\PYG{p}{(}\PYG{n}{inputs}\PYG{o}{=}\PYG{n}{l\PYGZus{}input}\PYG{p}{,} \PYG{n}{outputs}\PYG{o}{=}\PYG{n}{l\PYGZus{}output}\PYG{p}{)}
\end{sphinxVerbatim}

For flexibility (and larger\sphinxhyphen{}scale tests later on), let’s also define a \sphinxstyleemphasis{proper} ResNet with a few more layers. This architecture is the one from the original paper, and will give a fairly good performance (\sphinxcode{\sphinxupquote{network\_small}} above will train faster, but give a sub\sphinxhyphen{}optimal performance at inference time).

\begin{sphinxVerbatim}[commandchars=\\\{\}]
\PYG{k}{def} \PYG{n+nf}{network\PYGZus{}medium}\PYG{p}{(}\PYG{n}{inputs\PYGZus{}dict}\PYG{p}{)}\PYG{p}{:}
    \PYG{n}{l\PYGZus{}input} \PYG{o}{=} \PYG{n}{keras}\PYG{o}{.}\PYG{n}{layers}\PYG{o}{.}\PYG{n}{Input}\PYG{p}{(}\PYG{o}{*}\PYG{o}{*}\PYG{n}{inputs\PYGZus{}dict}\PYG{p}{)}
    \PYG{n}{block\PYGZus{}0} \PYG{o}{=} \PYG{n}{keras}\PYG{o}{.}\PYG{n}{layers}\PYG{o}{.}\PYG{n}{Conv2D}\PYG{p}{(}\PYG{n}{filters}\PYG{o}{=}\PYG{l+m+mi}{32}\PYG{p}{,} \PYG{n}{kernel\PYGZus{}size}\PYG{o}{=}\PYG{l+m+mi}{5}\PYG{p}{,} \PYG{n}{padding}\PYG{o}{=}\PYG{l+s+s1}{\PYGZsq{}}\PYG{l+s+s1}{same}\PYG{l+s+s1}{\PYGZsq{}}\PYG{p}{)}\PYG{p}{(}\PYG{n}{l\PYGZus{}input}\PYG{p}{)}
    \PYG{n}{block\PYGZus{}0} \PYG{o}{=} \PYG{n}{keras}\PYG{o}{.}\PYG{n}{layers}\PYG{o}{.}\PYG{n}{LeakyReLU}\PYG{p}{(}\PYG{p}{)}\PYG{p}{(}\PYG{n}{block\PYGZus{}0}\PYG{p}{)}

    \PYG{n}{l\PYGZus{}conv1} \PYG{o}{=} \PYG{n}{keras}\PYG{o}{.}\PYG{n}{layers}\PYG{o}{.}\PYG{n}{Conv2D}\PYG{p}{(}\PYG{n}{filters}\PYG{o}{=}\PYG{l+m+mi}{32}\PYG{p}{,} \PYG{n}{kernel\PYGZus{}size}\PYG{o}{=}\PYG{l+m+mi}{5}\PYG{p}{,} \PYG{n}{padding}\PYG{o}{=}\PYG{l+s+s1}{\PYGZsq{}}\PYG{l+s+s1}{same}\PYG{l+s+s1}{\PYGZsq{}}\PYG{p}{)}\PYG{p}{(}\PYG{n}{block\PYGZus{}0}\PYG{p}{)}
    \PYG{n}{l\PYGZus{}conv1} \PYG{o}{=} \PYG{n}{keras}\PYG{o}{.}\PYG{n}{layers}\PYG{o}{.}\PYG{n}{LeakyReLU}\PYG{p}{(}\PYG{p}{)}\PYG{p}{(}\PYG{n}{l\PYGZus{}conv1}\PYG{p}{)}
    \PYG{n}{l\PYGZus{}conv2} \PYG{o}{=} \PYG{n}{keras}\PYG{o}{.}\PYG{n}{layers}\PYG{o}{.}\PYG{n}{Conv2D}\PYG{p}{(}\PYG{n}{filters}\PYG{o}{=}\PYG{l+m+mi}{32}\PYG{p}{,} \PYG{n}{kernel\PYGZus{}size}\PYG{o}{=}\PYG{l+m+mi}{5}\PYG{p}{,} \PYG{n}{padding}\PYG{o}{=}\PYG{l+s+s1}{\PYGZsq{}}\PYG{l+s+s1}{same}\PYG{l+s+s1}{\PYGZsq{}}\PYG{p}{)}\PYG{p}{(}\PYG{n}{l\PYGZus{}conv1}\PYG{p}{)}
    \PYG{n}{l\PYGZus{}skip1} \PYG{o}{=} \PYG{n}{keras}\PYG{o}{.}\PYG{n}{layers}\PYG{o}{.}\PYG{n}{add}\PYG{p}{(}\PYG{p}{[}\PYG{n}{block\PYGZus{}0}\PYG{p}{,} \PYG{n}{l\PYGZus{}conv2}\PYG{p}{]}\PYG{p}{)}
    \PYG{n}{block\PYGZus{}1} \PYG{o}{=} \PYG{n}{keras}\PYG{o}{.}\PYG{n}{layers}\PYG{o}{.}\PYG{n}{LeakyReLU}\PYG{p}{(}\PYG{p}{)}\PYG{p}{(}\PYG{n}{l\PYGZus{}skip1}\PYG{p}{)}

    \PYG{n}{l\PYGZus{}conv3} \PYG{o}{=} \PYG{n}{keras}\PYG{o}{.}\PYG{n}{layers}\PYG{o}{.}\PYG{n}{Conv2D}\PYG{p}{(}\PYG{n}{filters}\PYG{o}{=}\PYG{l+m+mi}{32}\PYG{p}{,} \PYG{n}{kernel\PYGZus{}size}\PYG{o}{=}\PYG{l+m+mi}{5}\PYG{p}{,} \PYG{n}{padding}\PYG{o}{=}\PYG{l+s+s1}{\PYGZsq{}}\PYG{l+s+s1}{same}\PYG{l+s+s1}{\PYGZsq{}}\PYG{p}{)}\PYG{p}{(}\PYG{n}{block\PYGZus{}1}\PYG{p}{)}
    \PYG{n}{l\PYGZus{}conv3} \PYG{o}{=} \PYG{n}{keras}\PYG{o}{.}\PYG{n}{layers}\PYG{o}{.}\PYG{n}{LeakyReLU}\PYG{p}{(}\PYG{p}{)}\PYG{p}{(}\PYG{n}{l\PYGZus{}conv3}\PYG{p}{)}
    \PYG{n}{l\PYGZus{}conv4} \PYG{o}{=} \PYG{n}{keras}\PYG{o}{.}\PYG{n}{layers}\PYG{o}{.}\PYG{n}{Conv2D}\PYG{p}{(}\PYG{n}{filters}\PYG{o}{=}\PYG{l+m+mi}{32}\PYG{p}{,} \PYG{n}{kernel\PYGZus{}size}\PYG{o}{=}\PYG{l+m+mi}{5}\PYG{p}{,} \PYG{n}{padding}\PYG{o}{=}\PYG{l+s+s1}{\PYGZsq{}}\PYG{l+s+s1}{same}\PYG{l+s+s1}{\PYGZsq{}}\PYG{p}{)}\PYG{p}{(}\PYG{n}{l\PYGZus{}conv3}\PYG{p}{)}
    \PYG{n}{l\PYGZus{}skip2} \PYG{o}{=} \PYG{n}{keras}\PYG{o}{.}\PYG{n}{layers}\PYG{o}{.}\PYG{n}{add}\PYG{p}{(}\PYG{p}{[}\PYG{n}{block\PYGZus{}1}\PYG{p}{,} \PYG{n}{l\PYGZus{}conv4}\PYG{p}{]}\PYG{p}{)}
    \PYG{n}{block\PYGZus{}2} \PYG{o}{=} \PYG{n}{keras}\PYG{o}{.}\PYG{n}{layers}\PYG{o}{.}\PYG{n}{LeakyReLU}\PYG{p}{(}\PYG{p}{)}\PYG{p}{(}\PYG{n}{l\PYGZus{}skip2}\PYG{p}{)}

    \PYG{n}{l\PYGZus{}conv5} \PYG{o}{=} \PYG{n}{keras}\PYG{o}{.}\PYG{n}{layers}\PYG{o}{.}\PYG{n}{Conv2D}\PYG{p}{(}\PYG{n}{filters}\PYG{o}{=}\PYG{l+m+mi}{32}\PYG{p}{,} \PYG{n}{kernel\PYGZus{}size}\PYG{o}{=}\PYG{l+m+mi}{5}\PYG{p}{,} \PYG{n}{padding}\PYG{o}{=}\PYG{l+s+s1}{\PYGZsq{}}\PYG{l+s+s1}{same}\PYG{l+s+s1}{\PYGZsq{}}\PYG{p}{)}\PYG{p}{(}\PYG{n}{block\PYGZus{}2}\PYG{p}{)}
    \PYG{n}{l\PYGZus{}conv5} \PYG{o}{=} \PYG{n}{keras}\PYG{o}{.}\PYG{n}{layers}\PYG{o}{.}\PYG{n}{LeakyReLU}\PYG{p}{(}\PYG{p}{)}\PYG{p}{(}\PYG{n}{l\PYGZus{}conv5}\PYG{p}{)}
    \PYG{n}{l\PYGZus{}conv6} \PYG{o}{=} \PYG{n}{keras}\PYG{o}{.}\PYG{n}{layers}\PYG{o}{.}\PYG{n}{Conv2D}\PYG{p}{(}\PYG{n}{filters}\PYG{o}{=}\PYG{l+m+mi}{32}\PYG{p}{,} \PYG{n}{kernel\PYGZus{}size}\PYG{o}{=}\PYG{l+m+mi}{5}\PYG{p}{,} \PYG{n}{padding}\PYG{o}{=}\PYG{l+s+s1}{\PYGZsq{}}\PYG{l+s+s1}{same}\PYG{l+s+s1}{\PYGZsq{}}\PYG{p}{)}\PYG{p}{(}\PYG{n}{l\PYGZus{}conv5}\PYG{p}{)}
    \PYG{n}{l\PYGZus{}skip3} \PYG{o}{=} \PYG{n}{keras}\PYG{o}{.}\PYG{n}{layers}\PYG{o}{.}\PYG{n}{add}\PYG{p}{(}\PYG{p}{[}\PYG{n}{block\PYGZus{}2}\PYG{p}{,} \PYG{n}{l\PYGZus{}conv6}\PYG{p}{]}\PYG{p}{)}
    \PYG{n}{block\PYGZus{}3} \PYG{o}{=} \PYG{n}{keras}\PYG{o}{.}\PYG{n}{layers}\PYG{o}{.}\PYG{n}{LeakyReLU}\PYG{p}{(}\PYG{p}{)}\PYG{p}{(}\PYG{n}{l\PYGZus{}skip3}\PYG{p}{)}

    \PYG{n}{l\PYGZus{}conv7} \PYG{o}{=} \PYG{n}{keras}\PYG{o}{.}\PYG{n}{layers}\PYG{o}{.}\PYG{n}{Conv2D}\PYG{p}{(}\PYG{n}{filters}\PYG{o}{=}\PYG{l+m+mi}{32}\PYG{p}{,} \PYG{n}{kernel\PYGZus{}size}\PYG{o}{=}\PYG{l+m+mi}{5}\PYG{p}{,} \PYG{n}{padding}\PYG{o}{=}\PYG{l+s+s1}{\PYGZsq{}}\PYG{l+s+s1}{same}\PYG{l+s+s1}{\PYGZsq{}}\PYG{p}{)}\PYG{p}{(}\PYG{n}{block\PYGZus{}3}\PYG{p}{)}
    \PYG{n}{l\PYGZus{}conv7} \PYG{o}{=} \PYG{n}{keras}\PYG{o}{.}\PYG{n}{layers}\PYG{o}{.}\PYG{n}{LeakyReLU}\PYG{p}{(}\PYG{p}{)}\PYG{p}{(}\PYG{n}{l\PYGZus{}conv7}\PYG{p}{)}
    \PYG{n}{l\PYGZus{}conv8} \PYG{o}{=} \PYG{n}{keras}\PYG{o}{.}\PYG{n}{layers}\PYG{o}{.}\PYG{n}{Conv2D}\PYG{p}{(}\PYG{n}{filters}\PYG{o}{=}\PYG{l+m+mi}{32}\PYG{p}{,} \PYG{n}{kernel\PYGZus{}size}\PYG{o}{=}\PYG{l+m+mi}{5}\PYG{p}{,} \PYG{n}{padding}\PYG{o}{=}\PYG{l+s+s1}{\PYGZsq{}}\PYG{l+s+s1}{same}\PYG{l+s+s1}{\PYGZsq{}}\PYG{p}{)}\PYG{p}{(}\PYG{n}{l\PYGZus{}conv7}\PYG{p}{)}
    \PYG{n}{l\PYGZus{}skip4} \PYG{o}{=} \PYG{n}{keras}\PYG{o}{.}\PYG{n}{layers}\PYG{o}{.}\PYG{n}{add}\PYG{p}{(}\PYG{p}{[}\PYG{n}{block\PYGZus{}3}\PYG{p}{,} \PYG{n}{l\PYGZus{}conv8}\PYG{p}{]}\PYG{p}{)}
    \PYG{n}{block\PYGZus{}4} \PYG{o}{=} \PYG{n}{keras}\PYG{o}{.}\PYG{n}{layers}\PYG{o}{.}\PYG{n}{LeakyReLU}\PYG{p}{(}\PYG{p}{)}\PYG{p}{(}\PYG{n}{l\PYGZus{}skip4}\PYG{p}{)}

    \PYG{n}{l\PYGZus{}conv9} \PYG{o}{=} \PYG{n}{keras}\PYG{o}{.}\PYG{n}{layers}\PYG{o}{.}\PYG{n}{Conv2D}\PYG{p}{(}\PYG{n}{filters}\PYG{o}{=}\PYG{l+m+mi}{32}\PYG{p}{,} \PYG{n}{kernel\PYGZus{}size}\PYG{o}{=}\PYG{l+m+mi}{5}\PYG{p}{,} \PYG{n}{padding}\PYG{o}{=}\PYG{l+s+s1}{\PYGZsq{}}\PYG{l+s+s1}{same}\PYG{l+s+s1}{\PYGZsq{}}\PYG{p}{)}\PYG{p}{(}\PYG{n}{block\PYGZus{}4}\PYG{p}{)}
    \PYG{n}{l\PYGZus{}conv9} \PYG{o}{=} \PYG{n}{keras}\PYG{o}{.}\PYG{n}{layers}\PYG{o}{.}\PYG{n}{LeakyReLU}\PYG{p}{(}\PYG{p}{)}\PYG{p}{(}\PYG{n}{l\PYGZus{}conv9}\PYG{p}{)}
    \PYG{n}{l\PYGZus{}convA} \PYG{o}{=} \PYG{n}{keras}\PYG{o}{.}\PYG{n}{layers}\PYG{o}{.}\PYG{n}{Conv2D}\PYG{p}{(}\PYG{n}{filters}\PYG{o}{=}\PYG{l+m+mi}{32}\PYG{p}{,} \PYG{n}{kernel\PYGZus{}size}\PYG{o}{=}\PYG{l+m+mi}{5}\PYG{p}{,} \PYG{n}{padding}\PYG{o}{=}\PYG{l+s+s1}{\PYGZsq{}}\PYG{l+s+s1}{same}\PYG{l+s+s1}{\PYGZsq{}}\PYG{p}{)}\PYG{p}{(}\PYG{n}{l\PYGZus{}conv9}\PYG{p}{)}
    \PYG{n}{l\PYGZus{}skip5} \PYG{o}{=} \PYG{n}{keras}\PYG{o}{.}\PYG{n}{layers}\PYG{o}{.}\PYG{n}{add}\PYG{p}{(}\PYG{p}{[}\PYG{n}{block\PYGZus{}4}\PYG{p}{,} \PYG{n}{l\PYGZus{}convA}\PYG{p}{]}\PYG{p}{)}
    \PYG{n}{block\PYGZus{}5} \PYG{o}{=} \PYG{n}{keras}\PYG{o}{.}\PYG{n}{layers}\PYG{o}{.}\PYG{n}{LeakyReLU}\PYG{p}{(}\PYG{p}{)}\PYG{p}{(}\PYG{n}{l\PYGZus{}skip5}\PYG{p}{)}

    \PYG{n}{l\PYGZus{}output} \PYG{o}{=} \PYG{n}{keras}\PYG{o}{.}\PYG{n}{layers}\PYG{o}{.}\PYG{n}{Conv2D}\PYG{p}{(}\PYG{n}{filters}\PYG{o}{=}\PYG{l+m+mi}{2}\PYG{p}{,}  \PYG{n}{kernel\PYGZus{}size}\PYG{o}{=}\PYG{l+m+mi}{5}\PYG{p}{,} \PYG{n}{padding}\PYG{o}{=}\PYG{l+s+s1}{\PYGZsq{}}\PYG{l+s+s1}{same}\PYG{l+s+s1}{\PYGZsq{}}\PYG{p}{)}\PYG{p}{(}\PYG{n}{block\PYGZus{}5}\PYG{p}{)}
    \PYG{k}{return} \PYG{n}{keras}\PYG{o}{.}\PYG{n}{models}\PYG{o}{.}\PYG{n}{Model}\PYG{p}{(}\PYG{n}{inputs}\PYG{o}{=}\PYG{n}{l\PYGZus{}input}\PYG{p}{,} \PYG{n}{outputs}\PYG{o}{=}\PYG{n}{l\PYGZus{}output}\PYG{p}{)}
\end{sphinxVerbatim}

Next, we’re coming to two functions which are pretty important: they transform the simulation state into an input tensor for the network, and vice versa. Hence, they’re the interface between \sphinxstyleemphasis{keras/tensorflow} and \sphinxstyleemphasis{phiflow}.

The \sphinxcode{\sphinxupquote{to\_keras}} function uses the two vector components via \sphinxcode{\sphinxupquote{vector{[}'x'{]}}} and \sphinxcode{\sphinxupquote{vector{[}'y'{]}}} to discard the outermost layer of the velocity field grids. This gives two tensors of equal size that can be combined.
It then adds a constant channel via \sphinxcode{\sphinxupquote{math.ones}} that is multiplied by the desired Reynolds number in \sphinxcode{\sphinxupquote{ext\_const\_channel}}. The resulting stack of grids is stacked along the \sphinxcode{\sphinxupquote{channels}} dimensions, and represents an  input to the neural network.

After network evaluation, we transform the output tensor back into a phiflow grid via the \sphinxcode{\sphinxupquote{to\_phiflow}} function.
It converts the 2\sphinxhyphen{}component tensor that is returned by the network into a phiflow staggered grid object, so that it is compatible with the velocity field of the fluid simulation.
(Note: these are two \sphinxstyleemphasis{centered} grids with different sizes, so we leave the work to the \sphinxcode{\sphinxupquote{domain.staggered\_grid}} function, which also sets physical size and boundary conditions as given by the domain object).

\begin{sphinxVerbatim}[commandchars=\\\{\}]
\PYG{k}{def} \PYG{n+nf}{to\PYGZus{}keras}\PYG{p}{(}\PYG{n}{dens\PYGZus{}vel\PYGZus{}grid\PYGZus{}array}\PYG{p}{,} \PYG{n}{ext\PYGZus{}const\PYGZus{}channel}\PYG{p}{)}\PYG{p}{:}
    \PYG{c+c1}{\PYGZsh{} align the sides the staggered velocity grid making its size the same as the centered grid}
    \PYG{k}{return} \PYG{n}{math}\PYG{o}{.}\PYG{n}{stack}\PYG{p}{(}
        \PYG{p}{[}
            \PYG{n}{math}\PYG{o}{.}\PYG{n}{pad}\PYG{p}{(} \PYG{n}{dens\PYGZus{}vel\PYGZus{}grid\PYGZus{}array}\PYG{p}{[}\PYG{l+m+mi}{1}\PYG{p}{]}\PYG{o}{.}\PYG{n}{vector}\PYG{p}{[}\PYG{l+s+s1}{\PYGZsq{}}\PYG{l+s+s1}{x}\PYG{l+s+s1}{\PYGZsq{}}\PYG{p}{]}\PYG{o}{.}\PYG{n}{values}\PYG{p}{,} \PYG{p}{\PYGZob{}}\PYG{l+s+s1}{\PYGZsq{}}\PYG{l+s+s1}{x}\PYG{l+s+s1}{\PYGZsq{}}\PYG{p}{:}\PYG{p}{(}\PYG{l+m+mi}{0}\PYG{p}{,}\PYG{l+m+mi}{1}\PYG{p}{)}\PYG{p}{\PYGZcb{}} \PYG{p}{,} \PYG{n}{math}\PYG{o}{.}\PYG{n}{extrapolation}\PYG{o}{.}\PYG{n}{ZERO}\PYG{p}{)}\PYG{p}{,}
            \PYG{n}{dens\PYGZus{}vel\PYGZus{}grid\PYGZus{}array}\PYG{p}{[}\PYG{l+m+mi}{1}\PYG{p}{]}\PYG{o}{.}\PYG{n}{vector}\PYG{p}{[}\PYG{l+s+s1}{\PYGZsq{}}\PYG{l+s+s1}{y}\PYG{l+s+s1}{\PYGZsq{}}\PYG{p}{]}\PYG{o}{.}\PYG{n}{y}\PYG{p}{[}\PYG{p}{:}\PYG{o}{\PYGZhy{}}\PYG{l+m+mi}{1}\PYG{p}{]}\PYG{o}{.}\PYG{n}{values}\PYG{p}{,}         \PYG{c+c1}{\PYGZsh{} v}
            \PYG{n}{math}\PYG{o}{.}\PYG{n}{ones}\PYG{p}{(}\PYG{n}{dens\PYGZus{}vel\PYGZus{}grid\PYGZus{}array}\PYG{p}{[}\PYG{l+m+mi}{0}\PYG{p}{]}\PYG{o}{.}\PYG{n}{shape}\PYG{p}{)}\PYG{o}{*}\PYG{n}{ext\PYGZus{}const\PYGZus{}channel} \PYG{c+c1}{\PYGZsh{} Re}
        \PYG{p}{]}\PYG{p}{,}
        \PYG{n}{math}\PYG{o}{.}\PYG{n}{channel}\PYG{p}{(}\PYG{l+s+s1}{\PYGZsq{}}\PYG{l+s+s1}{channels}\PYG{l+s+s1}{\PYGZsq{}}\PYG{p}{)}
    \PYG{p}{)}

\PYG{k}{def} \PYG{n+nf}{to\PYGZus{}phiflow}\PYG{p}{(}\PYG{n}{tf\PYGZus{}tensor}\PYG{p}{,} \PYG{n}{domain}\PYG{p}{)}\PYG{p}{:}
    \PYG{k}{return} \PYG{n}{domain}\PYG{o}{.}\PYG{n}{staggered\PYGZus{}grid}\PYG{p}{(}
        \PYG{n}{math}\PYG{o}{.}\PYG{n}{stack}\PYG{p}{(}
            \PYG{p}{[}
                \PYG{n}{math}\PYG{o}{.}\PYG{n}{tensor}\PYG{p}{(}\PYG{n}{tf}\PYG{o}{.}\PYG{n}{pad}\PYG{p}{(}\PYG{n}{tf\PYGZus{}tensor}\PYG{p}{[}\PYG{o}{.}\PYG{o}{.}\PYG{o}{.}\PYG{p}{,} \PYG{l+m+mi}{1}\PYG{p}{]}\PYG{p}{,} \PYG{p}{[}\PYG{p}{(}\PYG{l+m+mi}{0}\PYG{p}{,}\PYG{l+m+mi}{0}\PYG{p}{)}\PYG{p}{,} \PYG{p}{(}\PYG{l+m+mi}{0}\PYG{p}{,}\PYG{l+m+mi}{1}\PYG{p}{)}\PYG{p}{,} \PYG{p}{(}\PYG{l+m+mi}{0}\PYG{p}{,}\PYG{l+m+mi}{0}\PYG{p}{)}\PYG{p}{]}\PYG{p}{)}\PYG{p}{,} \PYG{n}{math}\PYG{o}{.}\PYG{n}{batch}\PYG{p}{(}\PYG{l+s+s1}{\PYGZsq{}}\PYG{l+s+s1}{batch}\PYG{l+s+s1}{\PYGZsq{}}\PYG{p}{)}\PYG{p}{,} \PYG{n}{math}\PYG{o}{.}\PYG{n}{spatial}\PYG{p}{(}\PYG{l+s+s1}{\PYGZsq{}}\PYG{l+s+s1}{y, x}\PYG{l+s+s1}{\PYGZsq{}}\PYG{p}{)}\PYG{p}{)}\PYG{p}{,} \PYG{c+c1}{\PYGZsh{} v}
                \PYG{n}{math}\PYG{o}{.}\PYG{n}{tensor}\PYG{p}{(} \PYG{n}{tf\PYGZus{}tensor}\PYG{p}{[}\PYG{o}{.}\PYG{o}{.}\PYG{o}{.}\PYG{p}{,}\PYG{p}{:}\PYG{o}{\PYGZhy{}}\PYG{l+m+mi}{1}\PYG{p}{,} \PYG{l+m+mi}{0}\PYG{p}{]}\PYG{p}{,} \PYG{n}{math}\PYG{o}{.}\PYG{n}{batch}\PYG{p}{(}\PYG{l+s+s1}{\PYGZsq{}}\PYG{l+s+s1}{batch}\PYG{l+s+s1}{\PYGZsq{}}\PYG{p}{)}\PYG{p}{,} \PYG{n}{math}\PYG{o}{.}\PYG{n}{spatial}\PYG{p}{(}\PYG{l+s+s1}{\PYGZsq{}}\PYG{l+s+s1}{y, x}\PYG{l+s+s1}{\PYGZsq{}}\PYG{p}{)}\PYG{p}{)}\PYG{p}{,} \PYG{c+c1}{\PYGZsh{} u }
            \PYG{p}{]}\PYG{p}{,} \PYG{n}{math}\PYG{o}{.}\PYG{n}{channel}\PYG{p}{(}\PYG{l+s+s1}{\PYGZsq{}}\PYG{l+s+s1}{vector}\PYG{l+s+s1}{\PYGZsq{}}\PYG{p}{)}
        \PYG{p}{)}
    \PYG{p}{)}
\end{sphinxVerbatim}

\bigskip\hrule\bigskip

\section{Data handling}
\label{\detokenize{diffphys-code-sol:data-handling}}
So far so good \sphinxhyphen{} we also need to take care of a few more mundane tasks, e.g., some data handling and randomization. Below we define a \sphinxcode{\sphinxupquote{Dataset}} class that stores all “ground truth” reference data (already downsampled).

We actually have a lot of data dimensions: multiple simulations, with many time steps, each with different fields. This makes the code below a bit more difficult to read.

The data format for the numpy array \sphinxcode{\sphinxupquote{dataPreloaded}}: is  \sphinxcode{\sphinxupquote{{[}'sim\_name', frame, field (dens \& vel){]}}}, where each field has dimension \sphinxcode{\sphinxupquote{{[}batch\sphinxhyphen{}size, y\sphinxhyphen{}size, x\sphinxhyphen{}size, channels{]}}} (this is the standard for a phiflow export).

\begin{sphinxVerbatim}[commandchars=\\\{\}]
\PYG{k}{class} \PYG{n+nc}{Dataset}\PYG{p}{(}\PYG{p}{)}\PYG{p}{:}
    \PYG{k}{def} \PYG{n+nf+fm}{\PYGZus{}\PYGZus{}init\PYGZus{}\PYGZus{}}\PYG{p}{(}\PYG{n+nb+bp}{self}\PYG{p}{,} \PYG{n}{data\PYGZus{}preloaded}\PYG{p}{,} \PYG{n}{num\PYGZus{}frames}\PYG{p}{,} \PYG{n}{num\PYGZus{}sims}\PYG{o}{=}\PYG{k+kc}{None}\PYG{p}{,} \PYG{n}{batch\PYGZus{}size}\PYG{o}{=}\PYG{l+m+mi}{1}\PYG{p}{,} \PYG{n}{is\PYGZus{}testset}\PYG{o}{=}\PYG{k+kc}{False}\PYG{p}{)}\PYG{p}{:}
        \PYG{n+nb+bp}{self}\PYG{o}{.}\PYG{n}{epoch}         \PYG{o}{=} \PYG{k+kc}{None}
        \PYG{n+nb+bp}{self}\PYG{o}{.}\PYG{n}{epochIdx}      \PYG{o}{=} \PYG{l+m+mi}{0}
        \PYG{n+nb+bp}{self}\PYG{o}{.}\PYG{n}{batch}         \PYG{o}{=} \PYG{k+kc}{None}
        \PYG{n+nb+bp}{self}\PYG{o}{.}\PYG{n}{batchIdx}      \PYG{o}{=} \PYG{l+m+mi}{0}
        \PYG{n+nb+bp}{self}\PYG{o}{.}\PYG{n}{step}          \PYG{o}{=} \PYG{k+kc}{None}
        \PYG{n+nb+bp}{self}\PYG{o}{.}\PYG{n}{stepIdx}       \PYG{o}{=} \PYG{l+m+mi}{0}

        \PYG{n+nb+bp}{self}\PYG{o}{.}\PYG{n}{dataPreloaded} \PYG{o}{=} \PYG{n}{data\PYGZus{}preloaded}
        \PYG{n+nb+bp}{self}\PYG{o}{.}\PYG{n}{batchSize}     \PYG{o}{=} \PYG{n}{batch\PYGZus{}size}

        \PYG{n+nb+bp}{self}\PYG{o}{.}\PYG{n}{numSims}       \PYG{o}{=} \PYG{n}{num\PYGZus{}sims}
        \PYG{n+nb+bp}{self}\PYG{o}{.}\PYG{n}{numBatches}    \PYG{o}{=} \PYG{n}{num\PYGZus{}sims}\PYG{o}{/}\PYG{o}{/}\PYG{n}{batch\PYGZus{}size}
        \PYG{n+nb+bp}{self}\PYG{o}{.}\PYG{n}{numFrames}     \PYG{o}{=} \PYG{n}{num\PYGZus{}frames}
        \PYG{n+nb+bp}{self}\PYG{o}{.}\PYG{n}{numSteps}      \PYG{o}{=} \PYG{n}{num\PYGZus{}frames}
        
        \PYG{c+c1}{\PYGZsh{} initialize directory keys (using naming scheme from SoL codebase)}
        \PYG{c+c1}{\PYGZsh{} constant additional per\PYGZhy{}sim channel: Reynolds numbers from data generation}
        \PYG{c+c1}{\PYGZsh{} hard coded for training and test data here}
        \PYG{k}{if} \PYG{o+ow}{not} \PYG{n}{is\PYGZus{}testset}\PYG{p}{:}
            \PYG{n+nb+bp}{self}\PYG{o}{.}\PYG{n}{dataSims} \PYG{o}{=} \PYG{p}{[}\PYG{l+s+s1}{\PYGZsq{}}\PYG{l+s+s1}{karman\PYGZhy{}fdt\PYGZhy{}hires\PYGZhy{}set/sim\PYGZus{}}\PYG{l+s+si}{\PYGZpc{}06d}\PYG{l+s+s1}{\PYGZsq{}}\PYG{o}{\PYGZpc{}}\PYG{k}{i} for i in range(num\PYGZus{}sims) ]
            \PYG{n}{ReNrs} \PYG{o}{=} \PYG{p}{[}\PYG{l+m+mf}{160000.0}\PYG{p}{,} \PYG{l+m+mf}{320000.0}\PYG{p}{,} \PYG{l+m+mf}{640000.0}\PYG{p}{,}  \PYG{l+m+mf}{1280000.0}\PYG{p}{,}  \PYG{l+m+mf}{2560000.0}\PYG{p}{,}  \PYG{l+m+mf}{5120000.0}\PYG{p}{]}
            \PYG{n+nb+bp}{self}\PYG{o}{.}\PYG{n}{extConstChannelPerSim} \PYG{o}{=} \PYG{p}{\PYGZob{}} \PYG{n+nb+bp}{self}\PYG{o}{.}\PYG{n}{dataSims}\PYG{p}{[}\PYG{n}{i}\PYG{p}{]}\PYG{p}{:}\PYG{p}{[}\PYG{n}{ReNrs}\PYG{p}{[}\PYG{n}{i}\PYG{p}{]}\PYG{p}{]} \PYG{k}{for} \PYG{n}{i} \PYG{o+ow}{in} \PYG{n+nb}{range}\PYG{p}{(}\PYG{n}{num\PYGZus{}sims}\PYG{p}{)} \PYG{p}{\PYGZcb{}}
        \PYG{k}{else}\PYG{p}{:}
            \PYG{n+nb+bp}{self}\PYG{o}{.}\PYG{n}{dataSims} \PYG{o}{=} \PYG{p}{[}\PYG{l+s+s1}{\PYGZsq{}}\PYG{l+s+s1}{karman\PYGZhy{}fdt\PYGZhy{}hires\PYGZhy{}testset/sim\PYGZus{}}\PYG{l+s+si}{\PYGZpc{}06d}\PYG{l+s+s1}{\PYGZsq{}}\PYG{o}{\PYGZpc{}}\PYG{k}{i} for i in range(num\PYGZus{}sims) ]
            \PYG{n}{ReNrs} \PYG{o}{=} \PYG{p}{[}\PYG{l+m+mf}{120000.0}\PYG{p}{,} \PYG{l+m+mf}{480000.0}\PYG{p}{,} \PYG{l+m+mf}{1920000.0}\PYG{p}{,} \PYG{l+m+mf}{7680000.0}\PYG{p}{]} 
            \PYG{n+nb+bp}{self}\PYG{o}{.}\PYG{n}{extConstChannelPerSim} \PYG{o}{=} \PYG{p}{\PYGZob{}} \PYG{n+nb+bp}{self}\PYG{o}{.}\PYG{n}{dataSims}\PYG{p}{[}\PYG{n}{i}\PYG{p}{]}\PYG{p}{:}\PYG{p}{[}\PYG{n}{ReNrs}\PYG{p}{[}\PYG{n}{i}\PYG{p}{]}\PYG{p}{]} \PYG{k}{for} \PYG{n}{i} \PYG{o+ow}{in} \PYG{n+nb}{range}\PYG{p}{(}\PYG{n}{num\PYGZus{}sims}\PYG{p}{)} \PYG{p}{\PYGZcb{}}

        \PYG{n+nb+bp}{self}\PYG{o}{.}\PYG{n}{dataFrames} \PYG{o}{=} \PYG{p}{[} \PYG{n}{np}\PYG{o}{.}\PYG{n}{arange}\PYG{p}{(}\PYG{n}{num\PYGZus{}frames}\PYG{p}{)} \PYG{k}{for} \PYG{n}{\PYGZus{}} \PYG{o+ow}{in} \PYG{n+nb+bp}{self}\PYG{o}{.}\PYG{n}{dataSims} \PYG{p}{]}  

        \PYG{c+c1}{\PYGZsh{} debugging example, check shape of a single marker density field:}
        \PYG{c+c1}{\PYGZsh{}print(format(self.dataPreloaded[self.dataSims[0]][0][0].shape )) }
        
        \PYG{c+c1}{\PYGZsh{} the data has the following shape [\PYGZsq{}sim\PYGZsq{}, frame, field (dens/vel)] where each field is [batch\PYGZhy{}size, y\PYGZhy{}size, x\PYGZhy{}size, channels]}
        \PYG{n+nb+bp}{self}\PYG{o}{.}\PYG{n}{resolution} \PYG{o}{=} \PYG{n+nb+bp}{self}\PYG{o}{.}\PYG{n}{dataPreloaded}\PYG{p}{[}\PYG{n+nb+bp}{self}\PYG{o}{.}\PYG{n}{dataSims}\PYG{p}{[}\PYG{l+m+mi}{0}\PYG{p}{]}\PYG{p}{]}\PYG{p}{[}\PYG{l+m+mi}{0}\PYG{p}{]}\PYG{p}{[}\PYG{l+m+mi}{0}\PYG{p}{]}\PYG{o}{.}\PYG{n}{shape}\PYG{p}{[}\PYG{l+m+mi}{1}\PYG{p}{:}\PYG{l+m+mi}{3}\PYG{p}{]}  

        \PYG{c+c1}{\PYGZsh{} compute data statistics for normalization}
        \PYG{n+nb+bp}{self}\PYG{o}{.}\PYG{n}{dataStats} \PYG{o}{=} \PYG{p}{\PYGZob{}}
            \PYG{l+s+s1}{\PYGZsq{}}\PYG{l+s+s1}{std}\PYG{l+s+s1}{\PYGZsq{}}\PYG{p}{:} \PYG{p}{(}
                \PYG{n}{np}\PYG{o}{.}\PYG{n}{std}\PYG{p}{(}\PYG{n}{np}\PYG{o}{.}\PYG{n}{concatenate}\PYG{p}{(}\PYG{p}{[}\PYG{n}{np}\PYG{o}{.}\PYG{n}{absolute}\PYG{p}{(}\PYG{n+nb+bp}{self}\PYG{o}{.}\PYG{n}{dataPreloaded}\PYG{p}{[}\PYG{n}{asim}\PYG{p}{]}\PYG{p}{[}\PYG{n}{i}\PYG{p}{]}\PYG{p}{[}\PYG{l+m+mi}{0}\PYG{p}{]}\PYG{o}{.}\PYG{n}{reshape}\PYG{p}{(}\PYG{o}{\PYGZhy{}}\PYG{l+m+mi}{1}\PYG{p}{)}\PYG{p}{)} \PYG{k}{for} \PYG{n}{asim} \PYG{o+ow}{in} \PYG{n+nb+bp}{self}\PYG{o}{.}\PYG{n}{dataSims} \PYG{k}{for} \PYG{n}{i} \PYG{o+ow}{in} \PYG{n+nb}{range}\PYG{p}{(}\PYG{n}{num\PYGZus{}frames}\PYG{p}{)}\PYG{p}{]}\PYG{p}{,} \PYG{n}{axis}\PYG{o}{=}\PYG{o}{\PYGZhy{}}\PYG{l+m+mi}{1}\PYG{p}{)}\PYG{p}{)}\PYG{p}{,} \PYG{c+c1}{\PYGZsh{} density}
                \PYG{n}{np}\PYG{o}{.}\PYG{n}{std}\PYG{p}{(}\PYG{n}{np}\PYG{o}{.}\PYG{n}{concatenate}\PYG{p}{(}\PYG{p}{[}\PYG{n}{np}\PYG{o}{.}\PYG{n}{absolute}\PYG{p}{(}\PYG{n+nb+bp}{self}\PYG{o}{.}\PYG{n}{dataPreloaded}\PYG{p}{[}\PYG{n}{asim}\PYG{p}{]}\PYG{p}{[}\PYG{n}{i}\PYG{p}{]}\PYG{p}{[}\PYG{l+m+mi}{1}\PYG{p}{]}\PYG{o}{.}\PYG{n}{reshape}\PYG{p}{(}\PYG{o}{\PYGZhy{}}\PYG{l+m+mi}{1}\PYG{p}{)}\PYG{p}{)} \PYG{k}{for} \PYG{n}{asim} \PYG{o+ow}{in} \PYG{n+nb+bp}{self}\PYG{o}{.}\PYG{n}{dataSims} \PYG{k}{for} \PYG{n}{i} \PYG{o+ow}{in} \PYG{n+nb}{range}\PYG{p}{(}\PYG{n}{num\PYGZus{}frames}\PYG{p}{)}\PYG{p}{]}\PYG{p}{,} \PYG{n}{axis}\PYG{o}{=}\PYG{o}{\PYGZhy{}}\PYG{l+m+mi}{1}\PYG{p}{)}\PYG{p}{)}\PYG{p}{,} \PYG{c+c1}{\PYGZsh{} x\PYGZhy{}velocity}
                \PYG{n}{np}\PYG{o}{.}\PYG{n}{std}\PYG{p}{(}\PYG{n}{np}\PYG{o}{.}\PYG{n}{concatenate}\PYG{p}{(}\PYG{p}{[}\PYG{n}{np}\PYG{o}{.}\PYG{n}{absolute}\PYG{p}{(}\PYG{n+nb+bp}{self}\PYG{o}{.}\PYG{n}{dataPreloaded}\PYG{p}{[}\PYG{n}{asim}\PYG{p}{]}\PYG{p}{[}\PYG{n}{i}\PYG{p}{]}\PYG{p}{[}\PYG{l+m+mi}{2}\PYG{p}{]}\PYG{o}{.}\PYG{n}{reshape}\PYG{p}{(}\PYG{o}{\PYGZhy{}}\PYG{l+m+mi}{1}\PYG{p}{)}\PYG{p}{)} \PYG{k}{for} \PYG{n}{asim} \PYG{o+ow}{in} \PYG{n+nb+bp}{self}\PYG{o}{.}\PYG{n}{dataSims} \PYG{k}{for} \PYG{n}{i} \PYG{o+ow}{in} \PYG{n+nb}{range}\PYG{p}{(}\PYG{n}{num\PYGZus{}frames}\PYG{p}{)}\PYG{p}{]}\PYG{p}{,} \PYG{n}{axis}\PYG{o}{=}\PYG{o}{\PYGZhy{}}\PYG{l+m+mi}{1}\PYG{p}{)}\PYG{p}{)}\PYG{p}{,} \PYG{c+c1}{\PYGZsh{} y\PYGZhy{}velocity}
            \PYG{p}{)}
        \PYG{p}{\PYGZcb{}}
        \PYG{n+nb+bp}{self}\PYG{o}{.}\PYG{n}{dataStats}\PYG{o}{.}\PYG{n}{update}\PYG{p}{(}\PYG{p}{\PYGZob{}}
            \PYG{l+s+s1}{\PYGZsq{}}\PYG{l+s+s1}{ext.std}\PYG{l+s+s1}{\PYGZsq{}}\PYG{p}{:} \PYG{p}{[} \PYG{n}{np}\PYG{o}{.}\PYG{n}{std}\PYG{p}{(}\PYG{p}{[}\PYG{n}{np}\PYG{o}{.}\PYG{n}{absolute}\PYG{p}{(}\PYG{n+nb+bp}{self}\PYG{o}{.}\PYG{n}{extConstChannelPerSim}\PYG{p}{[}\PYG{n}{asim}\PYG{p}{]}\PYG{p}{[}\PYG{l+m+mi}{0}\PYG{p}{]}\PYG{p}{)} \PYG{k}{for} \PYG{n}{asim} \PYG{o+ow}{in} \PYG{n+nb+bp}{self}\PYG{o}{.}\PYG{n}{dataSims}\PYG{p}{]}\PYG{p}{)} \PYG{p}{]} \PYG{c+c1}{\PYGZsh{} Reynolds Nr}
        \PYG{p}{\PYGZcb{}}\PYG{p}{)}

        \PYG{k}{if} \PYG{o+ow}{not} \PYG{n}{is\PYGZus{}testset}\PYG{p}{:}
            \PYG{n+nb}{print}\PYG{p}{(}\PYG{l+s+s2}{\PYGZdq{}}\PYG{l+s+s2}{Data stats: }\PYG{l+s+s2}{\PYGZdq{}}\PYG{o}{+}\PYG{n+nb}{format}\PYG{p}{(}\PYG{n+nb+bp}{self}\PYG{o}{.}\PYG{n}{dataStats}\PYG{p}{)}\PYG{p}{)}

    \PYG{c+c1}{\PYGZsh{} re\PYGZhy{}shuffle data for next epoch}
    \PYG{k}{def} \PYG{n+nf}{newEpoch}\PYG{p}{(}\PYG{n+nb+bp}{self}\PYG{p}{,} \PYG{n}{exclude\PYGZus{}tail}\PYG{o}{=}\PYG{l+m+mi}{0}\PYG{p}{,} \PYG{n}{shuffle\PYGZus{}data}\PYG{o}{=}\PYG{k+kc}{True}\PYG{p}{)}\PYG{p}{:}
        \PYG{n+nb+bp}{self}\PYG{o}{.}\PYG{n}{numSteps} \PYG{o}{=} \PYG{n+nb+bp}{self}\PYG{o}{.}\PYG{n}{numFrames} \PYG{o}{\PYGZhy{}} \PYG{n}{exclude\PYGZus{}tail}
        \PYG{n}{simSteps} \PYG{o}{=} \PYG{p}{[} \PYG{p}{(}\PYG{n}{asim}\PYG{p}{,} \PYG{n+nb+bp}{self}\PYG{o}{.}\PYG{n}{dataFrames}\PYG{p}{[}\PYG{n}{i}\PYG{p}{]}\PYG{p}{[}\PYG{l+m+mi}{0}\PYG{p}{:}\PYG{p}{(}\PYG{n+nb}{len}\PYG{p}{(}\PYG{n+nb+bp}{self}\PYG{o}{.}\PYG{n}{dataFrames}\PYG{p}{[}\PYG{n}{i}\PYG{p}{]}\PYG{p}{)}\PYG{o}{\PYGZhy{}}\PYG{n}{exclude\PYGZus{}tail}\PYG{p}{)}\PYG{p}{]}\PYG{p}{)} \PYG{k}{for} \PYG{n}{i}\PYG{p}{,}\PYG{n}{asim} \PYG{o+ow}{in} \PYG{n+nb}{enumerate}\PYG{p}{(}\PYG{n+nb+bp}{self}\PYG{o}{.}\PYG{n}{dataSims}\PYG{p}{)} \PYG{p}{]}
        \PYG{n}{sim\PYGZus{}step\PYGZus{}pair} \PYG{o}{=} \PYG{p}{[}\PYG{p}{]}
        \PYG{k}{for} \PYG{n}{i}\PYG{p}{,}\PYG{n}{\PYGZus{}} \PYG{o+ow}{in} \PYG{n+nb}{enumerate}\PYG{p}{(}\PYG{n}{simSteps}\PYG{p}{)}\PYG{p}{:}
            \PYG{n}{sim\PYGZus{}step\PYGZus{}pair} \PYG{o}{+}\PYG{o}{=} \PYG{p}{[} \PYG{p}{(}\PYG{n}{i}\PYG{p}{,} \PYG{n}{astep}\PYG{p}{)} \PYG{k}{for} \PYG{n}{astep} \PYG{o+ow}{in} \PYG{n}{simSteps}\PYG{p}{[}\PYG{n}{i}\PYG{p}{]}\PYG{p}{[}\PYG{l+m+mi}{1}\PYG{p}{]} \PYG{p}{]}  \PYG{c+c1}{\PYGZsh{} (sim\PYGZus{}idx, step) ...}

        \PYG{k}{if} \PYG{n}{shuffle\PYGZus{}data}\PYG{p}{:} \PYG{n}{random}\PYG{o}{.}\PYG{n}{shuffle}\PYG{p}{(}\PYG{n}{sim\PYGZus{}step\PYGZus{}pair}\PYG{p}{)}
        \PYG{n+nb+bp}{self}\PYG{o}{.}\PYG{n}{epoch} \PYG{o}{=} \PYG{p}{[} \PYG{n+nb}{list}\PYG{p}{(}\PYG{n}{sim\PYGZus{}step\PYGZus{}pair}\PYG{p}{[}\PYG{n}{i}\PYG{o}{*}\PYG{n+nb+bp}{self}\PYG{o}{.}\PYG{n}{numSteps}\PYG{p}{:}\PYG{p}{(}\PYG{n}{i}\PYG{o}{+}\PYG{l+m+mi}{1}\PYG{p}{)}\PYG{o}{*}\PYG{n+nb+bp}{self}\PYG{o}{.}\PYG{n}{numSteps}\PYG{p}{]}\PYG{p}{)} \PYG{k}{for} \PYG{n}{i} \PYG{o+ow}{in} \PYG{n+nb}{range}\PYG{p}{(}\PYG{n+nb+bp}{self}\PYG{o}{.}\PYG{n}{batchSize}\PYG{o}{*}\PYG{n+nb+bp}{self}\PYG{o}{.}\PYG{n}{numBatches}\PYG{p}{)} \PYG{p}{]}
        \PYG{n+nb+bp}{self}\PYG{o}{.}\PYG{n}{epochIdx} \PYG{o}{+}\PYG{o}{=} \PYG{l+m+mi}{1}
        \PYG{n+nb+bp}{self}\PYG{o}{.}\PYG{n}{batchIdx} \PYG{o}{=} \PYG{l+m+mi}{0}
        \PYG{n+nb+bp}{self}\PYG{o}{.}\PYG{n}{stepIdx} \PYG{o}{=} \PYG{l+m+mi}{0}

    \PYG{k}{def} \PYG{n+nf}{nextBatch}\PYG{p}{(}\PYG{n+nb+bp}{self}\PYG{p}{)}\PYG{p}{:}  
        \PYG{n+nb+bp}{self}\PYG{o}{.}\PYG{n}{batchIdx} \PYG{o}{+}\PYG{o}{=} \PYG{n+nb+bp}{self}\PYG{o}{.}\PYG{n}{batchSize}
        \PYG{n+nb+bp}{self}\PYG{o}{.}\PYG{n}{stepIdx} \PYG{o}{=} \PYG{l+m+mi}{0}

    \PYG{k}{def} \PYG{n+nf}{nextStep}\PYG{p}{(}\PYG{n+nb+bp}{self}\PYG{p}{)}\PYG{p}{:}
        \PYG{n+nb+bp}{self}\PYG{o}{.}\PYG{n}{stepIdx} \PYG{o}{+}\PYG{o}{=} \PYG{l+m+mi}{1}
\end{sphinxVerbatim}

The \sphinxcode{\sphinxupquote{nextEpoch}}, \sphinxcode{\sphinxupquote{nextBatch}}, and \sphinxcode{\sphinxupquote{nextStep}} functions will be called at training time to randomize the order of the training data.

Now we need one more function that compiles the data for a mini batch to train with, called \sphinxcode{\sphinxupquote{getData}} below. It returns batches of the desired size in terms of marker density, velocity, and Reynolds number.

\begin{sphinxVerbatim}[commandchars=\\\{\}]
\PYG{c+c1}{\PYGZsh{} for class Dataset():}
\PYG{k}{def} \PYG{n+nf}{getData}\PYG{p}{(}\PYG{n+nb+bp}{self}\PYG{p}{,} \PYG{n}{consecutive\PYGZus{}frames}\PYG{p}{)}\PYG{p}{:}
    \PYG{n}{d\PYGZus{}hi} \PYG{o}{=} \PYG{p}{[}
        \PYG{n}{np}\PYG{o}{.}\PYG{n}{concatenate}\PYG{p}{(}\PYG{p}{[}
            \PYG{n+nb+bp}{self}\PYG{o}{.}\PYG{n}{dataPreloaded}\PYG{p}{[}
                \PYG{n+nb+bp}{self}\PYG{o}{.}\PYG{n}{dataSims}\PYG{p}{[}\PYG{n+nb+bp}{self}\PYG{o}{.}\PYG{n}{epoch}\PYG{p}{[}\PYG{n+nb+bp}{self}\PYG{o}{.}\PYG{n}{batchIdx}\PYG{o}{+}\PYG{n}{i}\PYG{p}{]}\PYG{p}{[}\PYG{n+nb+bp}{self}\PYG{o}{.}\PYG{n}{stepIdx}\PYG{p}{]}\PYG{p}{[}\PYG{l+m+mi}{0}\PYG{p}{]}\PYG{p}{]} \PYG{c+c1}{\PYGZsh{} sim\PYGZus{}key}
            \PYG{p}{]}\PYG{p}{[}
                \PYG{n+nb+bp}{self}\PYG{o}{.}\PYG{n}{epoch}\PYG{p}{[}\PYG{n+nb+bp}{self}\PYG{o}{.}\PYG{n}{batchIdx}\PYG{o}{+}\PYG{n}{i}\PYG{p}{]}\PYG{p}{[}\PYG{n+nb+bp}{self}\PYG{o}{.}\PYG{n}{stepIdx}\PYG{p}{]}\PYG{p}{[}\PYG{l+m+mi}{1}\PYG{p}{]}\PYG{o}{+}\PYG{n}{j} \PYG{c+c1}{\PYGZsh{} frames}
            \PYG{p}{]}\PYG{p}{[}\PYG{l+m+mi}{0}\PYG{p}{]}
            \PYG{k}{for} \PYG{n}{i} \PYG{o+ow}{in} \PYG{n+nb}{range}\PYG{p}{(}\PYG{n+nb+bp}{self}\PYG{o}{.}\PYG{n}{batchSize}\PYG{p}{)}
        \PYG{p}{]}\PYG{p}{,} \PYG{n}{axis}\PYG{o}{=}\PYG{l+m+mi}{0}\PYG{p}{)} \PYG{k}{for} \PYG{n}{j} \PYG{o+ow}{in} \PYG{n+nb}{range}\PYG{p}{(}\PYG{n}{consecutive\PYGZus{}frames}\PYG{o}{+}\PYG{l+m+mi}{1}\PYG{p}{)}
    \PYG{p}{]}
    \PYG{n}{u\PYGZus{}hi} \PYG{o}{=} \PYG{p}{[}
        \PYG{n}{np}\PYG{o}{.}\PYG{n}{concatenate}\PYG{p}{(}\PYG{p}{[}
            \PYG{n+nb+bp}{self}\PYG{o}{.}\PYG{n}{dataPreloaded}\PYG{p}{[}
                \PYG{n+nb+bp}{self}\PYG{o}{.}\PYG{n}{dataSims}\PYG{p}{[}\PYG{n+nb+bp}{self}\PYG{o}{.}\PYG{n}{epoch}\PYG{p}{[}\PYG{n+nb+bp}{self}\PYG{o}{.}\PYG{n}{batchIdx}\PYG{o}{+}\PYG{n}{i}\PYG{p}{]}\PYG{p}{[}\PYG{n+nb+bp}{self}\PYG{o}{.}\PYG{n}{stepIdx}\PYG{p}{]}\PYG{p}{[}\PYG{l+m+mi}{0}\PYG{p}{]}\PYG{p}{]} \PYG{c+c1}{\PYGZsh{} sim\PYGZus{}key}
            \PYG{p}{]}\PYG{p}{[}
                \PYG{n+nb+bp}{self}\PYG{o}{.}\PYG{n}{epoch}\PYG{p}{[}\PYG{n+nb+bp}{self}\PYG{o}{.}\PYG{n}{batchIdx}\PYG{o}{+}\PYG{n}{i}\PYG{p}{]}\PYG{p}{[}\PYG{n+nb+bp}{self}\PYG{o}{.}\PYG{n}{stepIdx}\PYG{p}{]}\PYG{p}{[}\PYG{l+m+mi}{1}\PYG{p}{]}\PYG{o}{+}\PYG{n}{j} \PYG{c+c1}{\PYGZsh{} frames}
            \PYG{p}{]}\PYG{p}{[}\PYG{l+m+mi}{1}\PYG{p}{]}
            \PYG{k}{for} \PYG{n}{i} \PYG{o+ow}{in} \PYG{n+nb}{range}\PYG{p}{(}\PYG{n+nb+bp}{self}\PYG{o}{.}\PYG{n}{batchSize}\PYG{p}{)}
        \PYG{p}{]}\PYG{p}{,} \PYG{n}{axis}\PYG{o}{=}\PYG{l+m+mi}{0}\PYG{p}{)} \PYG{k}{for} \PYG{n}{j} \PYG{o+ow}{in} \PYG{n+nb}{range}\PYG{p}{(}\PYG{n}{consecutive\PYGZus{}frames}\PYG{o}{+}\PYG{l+m+mi}{1}\PYG{p}{)}
    \PYG{p}{]}
    \PYG{n}{v\PYGZus{}hi} \PYG{o}{=} \PYG{p}{[}
        \PYG{n}{np}\PYG{o}{.}\PYG{n}{concatenate}\PYG{p}{(}\PYG{p}{[}
            \PYG{n+nb+bp}{self}\PYG{o}{.}\PYG{n}{dataPreloaded}\PYG{p}{[}
                \PYG{n+nb+bp}{self}\PYG{o}{.}\PYG{n}{dataSims}\PYG{p}{[}\PYG{n+nb+bp}{self}\PYG{o}{.}\PYG{n}{epoch}\PYG{p}{[}\PYG{n+nb+bp}{self}\PYG{o}{.}\PYG{n}{batchIdx}\PYG{o}{+}\PYG{n}{i}\PYG{p}{]}\PYG{p}{[}\PYG{n+nb+bp}{self}\PYG{o}{.}\PYG{n}{stepIdx}\PYG{p}{]}\PYG{p}{[}\PYG{l+m+mi}{0}\PYG{p}{]}\PYG{p}{]} \PYG{c+c1}{\PYGZsh{} sim\PYGZus{}key}
            \PYG{p}{]}\PYG{p}{[}
                \PYG{n+nb+bp}{self}\PYG{o}{.}\PYG{n}{epoch}\PYG{p}{[}\PYG{n+nb+bp}{self}\PYG{o}{.}\PYG{n}{batchIdx}\PYG{o}{+}\PYG{n}{i}\PYG{p}{]}\PYG{p}{[}\PYG{n+nb+bp}{self}\PYG{o}{.}\PYG{n}{stepIdx}\PYG{p}{]}\PYG{p}{[}\PYG{l+m+mi}{1}\PYG{p}{]}\PYG{o}{+}\PYG{n}{j} \PYG{c+c1}{\PYGZsh{} frames}
            \PYG{p}{]}\PYG{p}{[}\PYG{l+m+mi}{2}\PYG{p}{]}
            \PYG{k}{for} \PYG{n}{i} \PYG{o+ow}{in} \PYG{n+nb}{range}\PYG{p}{(}\PYG{n+nb+bp}{self}\PYG{o}{.}\PYG{n}{batchSize}\PYG{p}{)}
        \PYG{p}{]}\PYG{p}{,} \PYG{n}{axis}\PYG{o}{=}\PYG{l+m+mi}{0}\PYG{p}{)} \PYG{k}{for} \PYG{n}{j} \PYG{o+ow}{in} \PYG{n+nb}{range}\PYG{p}{(}\PYG{n}{consecutive\PYGZus{}frames}\PYG{o}{+}\PYG{l+m+mi}{1}\PYG{p}{)}
    \PYG{p}{]}
    \PYG{n}{ext} \PYG{o}{=} \PYG{p}{[}
        \PYG{n+nb+bp}{self}\PYG{o}{.}\PYG{n}{extConstChannelPerSim}\PYG{p}{[}
            \PYG{n+nb+bp}{self}\PYG{o}{.}\PYG{n}{dataSims}\PYG{p}{[}\PYG{n+nb+bp}{self}\PYG{o}{.}\PYG{n}{epoch}\PYG{p}{[}\PYG{n+nb+bp}{self}\PYG{o}{.}\PYG{n}{batchIdx}\PYG{o}{+}\PYG{n}{i}\PYG{p}{]}\PYG{p}{[}\PYG{n+nb+bp}{self}\PYG{o}{.}\PYG{n}{stepIdx}\PYG{p}{]}\PYG{p}{[}\PYG{l+m+mi}{0}\PYG{p}{]}\PYG{p}{]}
        \PYG{p}{]}\PYG{p}{[}\PYG{l+m+mi}{0}\PYG{p}{]} \PYG{k}{for} \PYG{n}{i} \PYG{o+ow}{in} \PYG{n+nb}{range}\PYG{p}{(}\PYG{n+nb+bp}{self}\PYG{o}{.}\PYG{n}{batchSize}\PYG{p}{)}
    \PYG{p}{]}
    \PYG{k}{return} \PYG{p}{[}\PYG{n}{d\PYGZus{}hi}\PYG{p}{,} \PYG{n}{u\PYGZus{}hi}\PYG{p}{,} \PYG{n}{v\PYGZus{}hi}\PYG{p}{,} \PYG{n}{ext}\PYG{p}{]}
\end{sphinxVerbatim}

Note that the \sphinxcode{\sphinxupquote{density}} here denotes a passively advected marker field, and not the density of the fluid. Below we’ll be focusing on the velocity only, the marker density is tracked purely for visualization purposes.

After all the definitions we can finally run some code. We can define the dataset object with the downloaded data from the first cell.

\begin{sphinxVerbatim}[commandchars=\\\{\}]
\PYG{n}{nsims} \PYG{o}{=} \PYG{l+m+mi}{6}
\PYG{n}{batch\PYGZus{}size} \PYG{o}{=} \PYG{l+m+mi}{3}
\PYG{n}{simsteps} \PYG{o}{=} \PYG{l+m+mi}{500}

\PYG{n}{dataset} \PYG{o}{=} \PYG{n}{Dataset}\PYG{p}{(} \PYG{n}{data\PYGZus{}preloaded}\PYG{o}{=}\PYG{n}{data\PYGZus{}preloaded}\PYG{p}{,} \PYG{n}{num\PYGZus{}frames}\PYG{o}{=}\PYG{n}{simsteps}\PYG{p}{,} \PYG{n}{num\PYGZus{}sims}\PYG{o}{=}\PYG{n}{nsims}\PYG{p}{,} \PYG{n}{batch\PYGZus{}size}\PYG{o}{=}\PYG{n}{batch\PYGZus{}size} \PYG{p}{)}
\end{sphinxVerbatim}

\begin{sphinxVerbatim}[commandchars=\\\{\}]
Data stats: \PYGZob{}\PYGZsq{}std\PYGZsq{}: (2.6542656, 0.23155601, 0.3066732), \PYGZsq{}ext.std\PYGZsq{}: [1732512.6262166172]\PYGZcb{}
\end{sphinxVerbatim}

Additionally, we’ve defined several global variables to control the training and the simulation in the next code cells.

The most important and interesting one is \sphinxcode{\sphinxupquote{msteps}}. It defines the number of simulation steps that are unrolled at each training iteration. This directly influences the runtime of each training step, as we first have to simulate all steps forward, and then backpropagate the gradient through all \sphinxcode{\sphinxupquote{msteps}} simulation steps interleaved with the NN evaluations. However, this is where we’ll receive important feedback in terms of gradients how the inferred corrections actually influence a running simulation. Hence, larger \sphinxcode{\sphinxupquote{msteps}} are typically better.

In addition we define the resolution of the simulation in \sphinxcode{\sphinxupquote{source\_res}}, and allocate the fluid solver object called \sphinxcode{\sphinxupquote{simulator}}. In order to create grids, it requires access to a \sphinxcode{\sphinxupquote{Domain}} object, which mostly exists for convenience purposes: it stores resolution, physical size in \sphinxcode{\sphinxupquote{bounds}}, and boundary conditions of the domain. This information needs to be passed to every grid, and hence it’s convenient to have it in one place in the form of the \sphinxcode{\sphinxupquote{Domain}}. For the setup described above, we need different boundary conditions along x and y: closed walls, and free flow in and out of the domain, respecitvely.

We also instantiate the actual NN \sphinxcode{\sphinxupquote{network}} in the next cell.

\begin{sphinxVerbatim}[commandchars=\\\{\}]
\PYG{c+c1}{\PYGZsh{} one of the most crucial! how many simulation steps to look into the future while training}
\PYG{n}{msteps} \PYG{o}{=} \PYG{l+m+mi}{4}

\PYG{c+c1}{\PYGZsh{} \PYGZsh{} this is the actual resolution in terms of cells}
\PYG{n}{source\PYGZus{}res} \PYG{o}{=} \PYG{n+nb}{list}\PYG{p}{(}\PYG{n}{dataset}\PYG{o}{.}\PYG{n}{resolution}\PYG{p}{)}
\PYG{c+c1}{\PYGZsh{} \PYGZsh{} this is a virtual size, in terms of abstract units for the bounding box of the domain (it\PYGZsq{}s important for conversions or when rescaling to physical units)}
\PYG{n}{simulation\PYGZus{}length} \PYG{o}{=} \PYG{l+m+mf}{100.}

\PYG{c+c1}{\PYGZsh{} for readability}
\PYG{k+kn}{from} \PYG{n+nn}{phi}\PYG{n+nn}{.}\PYG{n+nn}{physics}\PYG{n+nn}{.}\PYG{n+nn}{\PYGZus{}boundaries} \PYG{k+kn}{import} \PYG{n}{Domain}\PYG{p}{,} \PYG{n}{OPEN}\PYG{p}{,} \PYG{n}{STICKY} \PYG{k}{as} \PYG{n}{CLOSED}

\PYG{n}{boundary\PYGZus{}conditions} \PYG{o}{=} \PYG{p}{\PYGZob{}}
    \PYG{l+s+s1}{\PYGZsq{}}\PYG{l+s+s1}{x}\PYG{l+s+s1}{\PYGZsq{}}\PYG{p}{:}\PYG{p}{(}\PYG{n}{phi}\PYG{o}{.}\PYG{n}{physics}\PYG{o}{.}\PYG{n}{\PYGZus{}boundaries}\PYG{o}{.}\PYG{n}{STICKY}\PYG{p}{,}\PYG{n}{phi}\PYG{o}{.}\PYG{n}{physics}\PYG{o}{.}\PYG{n}{\PYGZus{}boundaries}\PYG{o}{.}\PYG{n}{STICKY}\PYG{p}{)}\PYG{p}{,} 
    \PYG{l+s+s1}{\PYGZsq{}}\PYG{l+s+s1}{y}\PYG{l+s+s1}{\PYGZsq{}}\PYG{p}{:}\PYG{p}{(}\PYG{n}{phi}\PYG{o}{.}\PYG{n}{physics}\PYG{o}{.}\PYG{n}{\PYGZus{}boundaries}\PYG{o}{.}\PYG{n}{OPEN}\PYG{p}{,}  \PYG{n}{phi}\PYG{o}{.}\PYG{n}{physics}\PYG{o}{.}\PYG{n}{\PYGZus{}boundaries}\PYG{o}{.}\PYG{n}{OPEN}\PYG{p}{)} \PYG{p}{\PYGZcb{}}

\PYG{n}{domain} \PYG{o}{=} \PYG{n}{Domain}\PYG{p}{(}\PYG{n}{y}\PYG{o}{=}\PYG{n}{source\PYGZus{}res}\PYG{p}{[}\PYG{l+m+mi}{0}\PYG{p}{]}\PYG{p}{,} \PYG{n}{x}\PYG{o}{=}\PYG{n}{source\PYGZus{}res}\PYG{p}{[}\PYG{l+m+mi}{1}\PYG{p}{]}\PYG{p}{,} \PYG{n}{bounds}\PYG{o}{=}\PYG{n}{Box}\PYG{p}{[}\PYG{l+m+mi}{0}\PYG{p}{:}\PYG{l+m+mi}{2}\PYG{o}{*}\PYG{n}{simulation\PYGZus{}length}\PYG{p}{,} \PYG{l+m+mi}{0}\PYG{p}{:}\PYG{n}{simulation\PYGZus{}length}\PYG{p}{]}\PYG{p}{,} \PYG{n}{boundaries}\PYG{o}{=}\PYG{n}{boundary\PYGZus{}conditions}\PYG{p}{)}
\PYG{n}{simulator} \PYG{o}{=} \PYG{n}{KarmanFlow}\PYG{p}{(}\PYG{n}{domain}\PYG{o}{=}\PYG{n}{domain}\PYG{p}{)}

\PYG{n}{network} \PYG{o}{=} \PYG{n}{network\PYGZus{}small}\PYG{p}{(}\PYG{n+nb}{dict}\PYG{p}{(}\PYG{n}{shape}\PYG{o}{=}\PYG{p}{(}\PYG{n}{source\PYGZus{}res}\PYG{p}{[}\PYG{l+m+mi}{0}\PYG{p}{]}\PYG{p}{,}\PYG{n}{source\PYGZus{}res}\PYG{p}{[}\PYG{l+m+mi}{1}\PYG{p}{]}\PYG{p}{,} \PYG{l+m+mi}{3}\PYG{p}{)}\PYG{p}{)}\PYG{p}{)}
\PYG{n}{network}\PYG{o}{.}\PYG{n}{summary}\PYG{p}{(}\PYG{p}{)}
\end{sphinxVerbatim}

\begin{sphinxVerbatim}[commandchars=\\\{\}]
Model: \PYGZdq{}model\PYGZdq{}
\PYGZus{}\PYGZus{}\PYGZus{}\PYGZus{}\PYGZus{}\PYGZus{}\PYGZus{}\PYGZus{}\PYGZus{}\PYGZus{}\PYGZus{}\PYGZus{}\PYGZus{}\PYGZus{}\PYGZus{}\PYGZus{}\PYGZus{}\PYGZus{}\PYGZus{}\PYGZus{}\PYGZus{}\PYGZus{}\PYGZus{}\PYGZus{}\PYGZus{}\PYGZus{}\PYGZus{}\PYGZus{}\PYGZus{}\PYGZus{}\PYGZus{}\PYGZus{}\PYGZus{}\PYGZus{}\PYGZus{}\PYGZus{}\PYGZus{}\PYGZus{}\PYGZus{}\PYGZus{}\PYGZus{}\PYGZus{}\PYGZus{}\PYGZus{}\PYGZus{}\PYGZus{}\PYGZus{}\PYGZus{}\PYGZus{}\PYGZus{}\PYGZus{}\PYGZus{}\PYGZus{}\PYGZus{}\PYGZus{}\PYGZus{}\PYGZus{}\PYGZus{}\PYGZus{}\PYGZus{}\PYGZus{}\PYGZus{}\PYGZus{}\PYGZus{}\PYGZus{}
Layer (type)                 Output Shape              Param \PYGZsh{}   
=================================================================
input\PYGZus{}1 (InputLayer)         [(None, 64, 32, 3)]       0         
\PYGZus{}\PYGZus{}\PYGZus{}\PYGZus{}\PYGZus{}\PYGZus{}\PYGZus{}\PYGZus{}\PYGZus{}\PYGZus{}\PYGZus{}\PYGZus{}\PYGZus{}\PYGZus{}\PYGZus{}\PYGZus{}\PYGZus{}\PYGZus{}\PYGZus{}\PYGZus{}\PYGZus{}\PYGZus{}\PYGZus{}\PYGZus{}\PYGZus{}\PYGZus{}\PYGZus{}\PYGZus{}\PYGZus{}\PYGZus{}\PYGZus{}\PYGZus{}\PYGZus{}\PYGZus{}\PYGZus{}\PYGZus{}\PYGZus{}\PYGZus{}\PYGZus{}\PYGZus{}\PYGZus{}\PYGZus{}\PYGZus{}\PYGZus{}\PYGZus{}\PYGZus{}\PYGZus{}\PYGZus{}\PYGZus{}\PYGZus{}\PYGZus{}\PYGZus{}\PYGZus{}\PYGZus{}\PYGZus{}\PYGZus{}\PYGZus{}\PYGZus{}\PYGZus{}\PYGZus{}\PYGZus{}\PYGZus{}\PYGZus{}\PYGZus{}\PYGZus{}
conv2d (Conv2D)              (None, 64, 32, 32)        2432      
\PYGZus{}\PYGZus{}\PYGZus{}\PYGZus{}\PYGZus{}\PYGZus{}\PYGZus{}\PYGZus{}\PYGZus{}\PYGZus{}\PYGZus{}\PYGZus{}\PYGZus{}\PYGZus{}\PYGZus{}\PYGZus{}\PYGZus{}\PYGZus{}\PYGZus{}\PYGZus{}\PYGZus{}\PYGZus{}\PYGZus{}\PYGZus{}\PYGZus{}\PYGZus{}\PYGZus{}\PYGZus{}\PYGZus{}\PYGZus{}\PYGZus{}\PYGZus{}\PYGZus{}\PYGZus{}\PYGZus{}\PYGZus{}\PYGZus{}\PYGZus{}\PYGZus{}\PYGZus{}\PYGZus{}\PYGZus{}\PYGZus{}\PYGZus{}\PYGZus{}\PYGZus{}\PYGZus{}\PYGZus{}\PYGZus{}\PYGZus{}\PYGZus{}\PYGZus{}\PYGZus{}\PYGZus{}\PYGZus{}\PYGZus{}\PYGZus{}\PYGZus{}\PYGZus{}\PYGZus{}\PYGZus{}\PYGZus{}\PYGZus{}\PYGZus{}\PYGZus{}
leaky\PYGZus{}re\PYGZus{}lu (LeakyReLU)      (None, 64, 32, 32)        0         
\PYGZus{}\PYGZus{}\PYGZus{}\PYGZus{}\PYGZus{}\PYGZus{}\PYGZus{}\PYGZus{}\PYGZus{}\PYGZus{}\PYGZus{}\PYGZus{}\PYGZus{}\PYGZus{}\PYGZus{}\PYGZus{}\PYGZus{}\PYGZus{}\PYGZus{}\PYGZus{}\PYGZus{}\PYGZus{}\PYGZus{}\PYGZus{}\PYGZus{}\PYGZus{}\PYGZus{}\PYGZus{}\PYGZus{}\PYGZus{}\PYGZus{}\PYGZus{}\PYGZus{}\PYGZus{}\PYGZus{}\PYGZus{}\PYGZus{}\PYGZus{}\PYGZus{}\PYGZus{}\PYGZus{}\PYGZus{}\PYGZus{}\PYGZus{}\PYGZus{}\PYGZus{}\PYGZus{}\PYGZus{}\PYGZus{}\PYGZus{}\PYGZus{}\PYGZus{}\PYGZus{}\PYGZus{}\PYGZus{}\PYGZus{}\PYGZus{}\PYGZus{}\PYGZus{}\PYGZus{}\PYGZus{}\PYGZus{}\PYGZus{}\PYGZus{}\PYGZus{}
conv2d\PYGZus{}1 (Conv2D)            (None, 64, 32, 32)        25632     
\PYGZus{}\PYGZus{}\PYGZus{}\PYGZus{}\PYGZus{}\PYGZus{}\PYGZus{}\PYGZus{}\PYGZus{}\PYGZus{}\PYGZus{}\PYGZus{}\PYGZus{}\PYGZus{}\PYGZus{}\PYGZus{}\PYGZus{}\PYGZus{}\PYGZus{}\PYGZus{}\PYGZus{}\PYGZus{}\PYGZus{}\PYGZus{}\PYGZus{}\PYGZus{}\PYGZus{}\PYGZus{}\PYGZus{}\PYGZus{}\PYGZus{}\PYGZus{}\PYGZus{}\PYGZus{}\PYGZus{}\PYGZus{}\PYGZus{}\PYGZus{}\PYGZus{}\PYGZus{}\PYGZus{}\PYGZus{}\PYGZus{}\PYGZus{}\PYGZus{}\PYGZus{}\PYGZus{}\PYGZus{}\PYGZus{}\PYGZus{}\PYGZus{}\PYGZus{}\PYGZus{}\PYGZus{}\PYGZus{}\PYGZus{}\PYGZus{}\PYGZus{}\PYGZus{}\PYGZus{}\PYGZus{}\PYGZus{}\PYGZus{}\PYGZus{}\PYGZus{}
leaky\PYGZus{}re\PYGZus{}lu\PYGZus{}1 (LeakyReLU)    (None, 64, 32, 32)        0         
\PYGZus{}\PYGZus{}\PYGZus{}\PYGZus{}\PYGZus{}\PYGZus{}\PYGZus{}\PYGZus{}\PYGZus{}\PYGZus{}\PYGZus{}\PYGZus{}\PYGZus{}\PYGZus{}\PYGZus{}\PYGZus{}\PYGZus{}\PYGZus{}\PYGZus{}\PYGZus{}\PYGZus{}\PYGZus{}\PYGZus{}\PYGZus{}\PYGZus{}\PYGZus{}\PYGZus{}\PYGZus{}\PYGZus{}\PYGZus{}\PYGZus{}\PYGZus{}\PYGZus{}\PYGZus{}\PYGZus{}\PYGZus{}\PYGZus{}\PYGZus{}\PYGZus{}\PYGZus{}\PYGZus{}\PYGZus{}\PYGZus{}\PYGZus{}\PYGZus{}\PYGZus{}\PYGZus{}\PYGZus{}\PYGZus{}\PYGZus{}\PYGZus{}\PYGZus{}\PYGZus{}\PYGZus{}\PYGZus{}\PYGZus{}\PYGZus{}\PYGZus{}\PYGZus{}\PYGZus{}\PYGZus{}\PYGZus{}\PYGZus{}\PYGZus{}\PYGZus{}
conv2d\PYGZus{}2 (Conv2D)            (None, 64, 32, 32)        25632     
\PYGZus{}\PYGZus{}\PYGZus{}\PYGZus{}\PYGZus{}\PYGZus{}\PYGZus{}\PYGZus{}\PYGZus{}\PYGZus{}\PYGZus{}\PYGZus{}\PYGZus{}\PYGZus{}\PYGZus{}\PYGZus{}\PYGZus{}\PYGZus{}\PYGZus{}\PYGZus{}\PYGZus{}\PYGZus{}\PYGZus{}\PYGZus{}\PYGZus{}\PYGZus{}\PYGZus{}\PYGZus{}\PYGZus{}\PYGZus{}\PYGZus{}\PYGZus{}\PYGZus{}\PYGZus{}\PYGZus{}\PYGZus{}\PYGZus{}\PYGZus{}\PYGZus{}\PYGZus{}\PYGZus{}\PYGZus{}\PYGZus{}\PYGZus{}\PYGZus{}\PYGZus{}\PYGZus{}\PYGZus{}\PYGZus{}\PYGZus{}\PYGZus{}\PYGZus{}\PYGZus{}\PYGZus{}\PYGZus{}\PYGZus{}\PYGZus{}\PYGZus{}\PYGZus{}\PYGZus{}\PYGZus{}\PYGZus{}\PYGZus{}\PYGZus{}\PYGZus{}
leaky\PYGZus{}re\PYGZus{}lu\PYGZus{}2 (LeakyReLU)    (None, 64, 32, 32)        0         
\PYGZus{}\PYGZus{}\PYGZus{}\PYGZus{}\PYGZus{}\PYGZus{}\PYGZus{}\PYGZus{}\PYGZus{}\PYGZus{}\PYGZus{}\PYGZus{}\PYGZus{}\PYGZus{}\PYGZus{}\PYGZus{}\PYGZus{}\PYGZus{}\PYGZus{}\PYGZus{}\PYGZus{}\PYGZus{}\PYGZus{}\PYGZus{}\PYGZus{}\PYGZus{}\PYGZus{}\PYGZus{}\PYGZus{}\PYGZus{}\PYGZus{}\PYGZus{}\PYGZus{}\PYGZus{}\PYGZus{}\PYGZus{}\PYGZus{}\PYGZus{}\PYGZus{}\PYGZus{}\PYGZus{}\PYGZus{}\PYGZus{}\PYGZus{}\PYGZus{}\PYGZus{}\PYGZus{}\PYGZus{}\PYGZus{}\PYGZus{}\PYGZus{}\PYGZus{}\PYGZus{}\PYGZus{}\PYGZus{}\PYGZus{}\PYGZus{}\PYGZus{}\PYGZus{}\PYGZus{}\PYGZus{}\PYGZus{}\PYGZus{}\PYGZus{}\PYGZus{}
conv2d\PYGZus{}3 (Conv2D)            (None, 64, 32, 2)         1602      
=================================================================
Total params: 55,298
Trainable params: 55,298
Non\PYGZhy{}trainable params: 0
\PYGZus{}\PYGZus{}\PYGZus{}\PYGZus{}\PYGZus{}\PYGZus{}\PYGZus{}\PYGZus{}\PYGZus{}\PYGZus{}\PYGZus{}\PYGZus{}\PYGZus{}\PYGZus{}\PYGZus{}\PYGZus{}\PYGZus{}\PYGZus{}\PYGZus{}\PYGZus{}\PYGZus{}\PYGZus{}\PYGZus{}\PYGZus{}\PYGZus{}\PYGZus{}\PYGZus{}\PYGZus{}\PYGZus{}\PYGZus{}\PYGZus{}\PYGZus{}\PYGZus{}\PYGZus{}\PYGZus{}\PYGZus{}\PYGZus{}\PYGZus{}\PYGZus{}\PYGZus{}\PYGZus{}\PYGZus{}\PYGZus{}\PYGZus{}\PYGZus{}\PYGZus{}\PYGZus{}\PYGZus{}\PYGZus{}\PYGZus{}\PYGZus{}\PYGZus{}\PYGZus{}\PYGZus{}\PYGZus{}\PYGZus{}\PYGZus{}\PYGZus{}\PYGZus{}\PYGZus{}\PYGZus{}\PYGZus{}\PYGZus{}\PYGZus{}\PYGZus{}
\end{sphinxVerbatim}

\section{Interleaving simulation and NN}
\label{\detokenize{diffphys-code-sol:interleaving-simulation-and-nn}}
Now comes the \sphinxstylestrong{most crucial} step in the whole setup: we define a function that encapsulates the chain of simulation steps and network evaluations in each training step. After all the work defining helper functions, it’s actually pretty simple: we create a gradient tape via \sphinxcode{\sphinxupquote{tf.GradientTape()}} such that we can backpropagate later on. We then loop over \sphinxcode{\sphinxupquote{msteps}}, call the simulator via \sphinxcode{\sphinxupquote{simulator.step}} for an input state, and afterwards evaluate the correction via \sphinxcode{\sphinxupquote{network(to\_keras(...))}}. The NN correction is then added to the last simulation state in the \sphinxcode{\sphinxupquote{prediction}} list (we’re actually simply overwriting the last simulated velocity \sphinxcode{\sphinxupquote{prediction{[}\sphinxhyphen{}1{]}{[}1{]}}} with \sphinxcode{\sphinxupquote{prediction{[}\sphinxhyphen{}1{]}{[}1{]} + correction{[}\sphinxhyphen{}1{]}}}.

One other important thing that’s happening here is normalization: the inputs to the network are divided by the standard deviations in \sphinxcode{\sphinxupquote{dataset.dataStats}}. After evaluating the \sphinxcode{\sphinxupquote{network}}, we only have a velocity left, so we can simply multiply by the standard deviation of the velocity again (via \sphinxcode{\sphinxupquote{* dataset.dataStats{[}'std'{]}{[}1{]}}} and \sphinxcode{\sphinxupquote{{[}2{]}}}).

The \sphinxcode{\sphinxupquote{training\_step}} function also directly evaluates and returns the loss. Here, we can simply use an \(L^2\) loss over the whole sequence, i.e. the iteration over \sphinxcode{\sphinxupquote{msteps}}. This is requiring a few lines of code because we separately loop over ‘x’ and ‘y’ components, in order to normalize and compare to the ground truth values from the training data set.

The “learning” happens in the last two lines via \sphinxcode{\sphinxupquote{tape.gradient()}} and \sphinxcode{\sphinxupquote{opt.apply\_gradients()}}, which then contain the aggregated information about how to change the NN weights to nudge the simulation closer to the reference for the full chain of simulation steps.

\begin{sphinxVerbatim}[commandchars=\\\{\}]
\PYG{k}{def} \PYG{n+nf}{training\PYGZus{}step}\PYG{p}{(}\PYG{n}{dens\PYGZus{}gt}\PYG{p}{,} \PYG{n}{vel\PYGZus{}gt}\PYG{p}{,} \PYG{n}{Re}\PYG{p}{,} \PYG{n}{i\PYGZus{}step}\PYG{p}{)}\PYG{p}{:}
    \PYG{k}{with} \PYG{n}{tf}\PYG{o}{.}\PYG{n}{GradientTape}\PYG{p}{(}\PYG{p}{)} \PYG{k}{as} \PYG{n}{tape}\PYG{p}{:}
        \PYG{n}{prediction}\PYG{p}{,} \PYG{n}{correction} \PYG{o}{=} \PYG{p}{[} \PYG{p}{[}\PYG{n}{dens\PYGZus{}gt}\PYG{p}{[}\PYG{l+m+mi}{0}\PYG{p}{]}\PYG{p}{,}\PYG{n}{vel\PYGZus{}gt}\PYG{p}{[}\PYG{l+m+mi}{0}\PYG{p}{]}\PYG{p}{]} \PYG{p}{]}\PYG{p}{,} \PYG{p}{[}\PYG{l+m+mi}{0}\PYG{p}{]} \PYG{c+c1}{\PYGZsh{} predicted states with correction, inferred velocity corrections}

        \PYG{k}{for} \PYG{n}{i} \PYG{o+ow}{in} \PYG{n+nb}{range}\PYG{p}{(}\PYG{n}{msteps}\PYG{p}{)}\PYG{p}{:}
            \PYG{n}{prediction} \PYG{o}{+}\PYG{o}{=} \PYG{p}{[}
                \PYG{n}{simulator}\PYG{o}{.}\PYG{n}{step}\PYG{p}{(}
                    \PYG{n}{density\PYGZus{}in}\PYG{o}{=}\PYG{n}{prediction}\PYG{p}{[}\PYG{o}{\PYGZhy{}}\PYG{l+m+mi}{1}\PYG{p}{]}\PYG{p}{[}\PYG{l+m+mi}{0}\PYG{p}{]}\PYG{p}{,}
                    \PYG{n}{velocity\PYGZus{}in}\PYG{o}{=}\PYG{n}{prediction}\PYG{p}{[}\PYG{o}{\PYGZhy{}}\PYG{l+m+mi}{1}\PYG{p}{]}\PYG{p}{[}\PYG{l+m+mi}{1}\PYG{p}{]}\PYG{p}{,}
                    \PYG{n}{re}\PYG{o}{=}\PYG{n}{Re}\PYG{p}{,} \PYG{n}{res}\PYG{o}{=}\PYG{n}{source\PYGZus{}res}\PYG{p}{[}\PYG{l+m+mi}{1}\PYG{p}{]}\PYG{p}{,}
                \PYG{p}{)}
            \PYG{p}{]}       \PYG{c+c1}{\PYGZsh{} prediction: [[density1, velocity1], [density2, velocity2], ...]}

            \PYG{n}{model\PYGZus{}input} \PYG{o}{=} \PYG{n}{to\PYGZus{}keras}\PYG{p}{(}\PYG{n}{prediction}\PYG{p}{[}\PYG{o}{\PYGZhy{}}\PYG{l+m+mi}{1}\PYG{p}{]}\PYG{p}{,} \PYG{n}{Re}\PYG{p}{)}
            \PYG{n}{model\PYGZus{}input} \PYG{o}{/}\PYG{o}{=} \PYG{n}{math}\PYG{o}{.}\PYG{n}{tensor}\PYG{p}{(}\PYG{p}{[}\PYG{n}{dataset}\PYG{o}{.}\PYG{n}{dataStats}\PYG{p}{[}\PYG{l+s+s1}{\PYGZsq{}}\PYG{l+s+s1}{std}\PYG{l+s+s1}{\PYGZsq{}}\PYG{p}{]}\PYG{p}{[}\PYG{l+m+mi}{1}\PYG{p}{]}\PYG{p}{,} \PYG{n}{dataset}\PYG{o}{.}\PYG{n}{dataStats}\PYG{p}{[}\PYG{l+s+s1}{\PYGZsq{}}\PYG{l+s+s1}{std}\PYG{l+s+s1}{\PYGZsq{}}\PYG{p}{]}\PYG{p}{[}\PYG{l+m+mi}{2}\PYG{p}{]}\PYG{p}{,} \PYG{n}{dataset}\PYG{o}{.}\PYG{n}{dataStats}\PYG{p}{[}\PYG{l+s+s1}{\PYGZsq{}}\PYG{l+s+s1}{ext.std}\PYG{l+s+s1}{\PYGZsq{}}\PYG{p}{]}\PYG{p}{[}\PYG{l+m+mi}{0}\PYG{p}{]}\PYG{p}{]}\PYG{p}{,} \PYG{n}{channel}\PYG{p}{(}\PYG{l+s+s1}{\PYGZsq{}}\PYG{l+s+s1}{channels}\PYG{l+s+s1}{\PYGZsq{}}\PYG{p}{)}\PYG{p}{)} \PYG{c+c1}{\PYGZsh{} [u, v, Re]}
            \PYG{n}{model\PYGZus{}out} \PYG{o}{=} \PYG{n}{network}\PYG{p}{(}\PYG{n}{model\PYGZus{}input}\PYG{o}{.}\PYG{n}{native}\PYG{p}{(}\PYG{p}{[}\PYG{l+s+s1}{\PYGZsq{}}\PYG{l+s+s1}{batch}\PYG{l+s+s1}{\PYGZsq{}}\PYG{p}{,} \PYG{l+s+s1}{\PYGZsq{}}\PYG{l+s+s1}{y}\PYG{l+s+s1}{\PYGZsq{}}\PYG{p}{,} \PYG{l+s+s1}{\PYGZsq{}}\PYG{l+s+s1}{x}\PYG{l+s+s1}{\PYGZsq{}}\PYG{p}{,} \PYG{l+s+s1}{\PYGZsq{}}\PYG{l+s+s1}{channels}\PYG{l+s+s1}{\PYGZsq{}}\PYG{p}{]}\PYG{p}{)}\PYG{p}{,} \PYG{n}{training}\PYG{o}{=}\PYG{k+kc}{True}\PYG{p}{)}
            \PYG{n}{model\PYGZus{}out} \PYG{o}{*}\PYG{o}{=} \PYG{p}{[}\PYG{n}{dataset}\PYG{o}{.}\PYG{n}{dataStats}\PYG{p}{[}\PYG{l+s+s1}{\PYGZsq{}}\PYG{l+s+s1}{std}\PYG{l+s+s1}{\PYGZsq{}}\PYG{p}{]}\PYG{p}{[}\PYG{l+m+mi}{1}\PYG{p}{]}\PYG{p}{,} \PYG{n}{dataset}\PYG{o}{.}\PYG{n}{dataStats}\PYG{p}{[}\PYG{l+s+s1}{\PYGZsq{}}\PYG{l+s+s1}{std}\PYG{l+s+s1}{\PYGZsq{}}\PYG{p}{]}\PYG{p}{[}\PYG{l+m+mi}{2}\PYG{p}{]}\PYG{p}{]} \PYG{c+c1}{\PYGZsh{} [u, v]}
            \PYG{n}{correction} \PYG{o}{+}\PYG{o}{=} \PYG{p}{[} \PYG{n}{to\PYGZus{}phiflow}\PYG{p}{(}\PYG{n}{model\PYGZus{}out}\PYG{p}{,} \PYG{n}{domain}\PYG{p}{)} \PYG{p}{]}                         \PYG{c+c1}{\PYGZsh{} [velocity\PYGZus{}correction1, velocity\PYGZus{}correction2, ...]}

            \PYG{n}{prediction}\PYG{p}{[}\PYG{o}{\PYGZhy{}}\PYG{l+m+mi}{1}\PYG{p}{]}\PYG{p}{[}\PYG{l+m+mi}{1}\PYG{p}{]} \PYG{o}{=} \PYG{n}{prediction}\PYG{p}{[}\PYG{o}{\PYGZhy{}}\PYG{l+m+mi}{1}\PYG{p}{]}\PYG{p}{[}\PYG{l+m+mi}{1}\PYG{p}{]} \PYG{o}{+} \PYG{n}{correction}\PYG{p}{[}\PYG{o}{\PYGZhy{}}\PYG{l+m+mi}{1}\PYG{p}{]}
            \PYG{c+c1}{\PYGZsh{}prediction[\PYGZhy{}1][1] = correction[\PYGZhy{}1]}

        \PYG{c+c1}{\PYGZsh{} evaluate loss}
        \PYG{n}{loss\PYGZus{}steps\PYGZus{}x} \PYG{o}{=} \PYG{p}{[}
            \PYG{n}{tf}\PYG{o}{.}\PYG{n}{nn}\PYG{o}{.}\PYG{n}{l2\PYGZus{}loss}\PYG{p}{(}
                \PYG{p}{(}
                    \PYG{n}{vel\PYGZus{}gt}\PYG{p}{[}\PYG{n}{i}\PYG{p}{]}\PYG{o}{.}\PYG{n}{vector}\PYG{p}{[}\PYG{l+s+s1}{\PYGZsq{}}\PYG{l+s+s1}{x}\PYG{l+s+s1}{\PYGZsq{}}\PYG{p}{]}\PYG{o}{.}\PYG{n}{values}\PYG{o}{.}\PYG{n}{native}\PYG{p}{(}\PYG{p}{(}\PYG{l+s+s1}{\PYGZsq{}}\PYG{l+s+s1}{batch}\PYG{l+s+s1}{\PYGZsq{}}\PYG{p}{,} \PYG{l+s+s1}{\PYGZsq{}}\PYG{l+s+s1}{y}\PYG{l+s+s1}{\PYGZsq{}}\PYG{p}{,} \PYG{l+s+s1}{\PYGZsq{}}\PYG{l+s+s1}{x}\PYG{l+s+s1}{\PYGZsq{}}\PYG{p}{)}\PYG{p}{)}
                    \PYG{o}{\PYGZhy{}} \PYG{n}{prediction}\PYG{p}{[}\PYG{n}{i}\PYG{p}{]}\PYG{p}{[}\PYG{l+m+mi}{1}\PYG{p}{]}\PYG{o}{.}\PYG{n}{vector}\PYG{p}{[}\PYG{l+s+s1}{\PYGZsq{}}\PYG{l+s+s1}{x}\PYG{l+s+s1}{\PYGZsq{}}\PYG{p}{]}\PYG{o}{.}\PYG{n}{values}\PYG{o}{.}\PYG{n}{native}\PYG{p}{(}\PYG{p}{(}\PYG{l+s+s1}{\PYGZsq{}}\PYG{l+s+s1}{batch}\PYG{l+s+s1}{\PYGZsq{}}\PYG{p}{,} \PYG{l+s+s1}{\PYGZsq{}}\PYG{l+s+s1}{y}\PYG{l+s+s1}{\PYGZsq{}}\PYG{p}{,} \PYG{l+s+s1}{\PYGZsq{}}\PYG{l+s+s1}{x}\PYG{l+s+s1}{\PYGZsq{}}\PYG{p}{)}\PYG{p}{)}
                \PYG{p}{)}\PYG{o}{/}\PYG{n}{dataset}\PYG{o}{.}\PYG{n}{dataStats}\PYG{p}{[}\PYG{l+s+s1}{\PYGZsq{}}\PYG{l+s+s1}{std}\PYG{l+s+s1}{\PYGZsq{}}\PYG{p}{]}\PYG{p}{[}\PYG{l+m+mi}{1}\PYG{p}{]}
            \PYG{p}{)}
            \PYG{k}{for} \PYG{n}{i} \PYG{o+ow}{in} \PYG{n+nb}{range}\PYG{p}{(}\PYG{l+m+mi}{1}\PYG{p}{,}\PYG{n}{msteps}\PYG{o}{+}\PYG{l+m+mi}{1}\PYG{p}{)}
        \PYG{p}{]}
        \PYG{n}{loss\PYGZus{}steps\PYGZus{}x\PYGZus{}sum} \PYG{o}{=} \PYG{n}{tf}\PYG{o}{.}\PYG{n}{math}\PYG{o}{.}\PYG{n}{reduce\PYGZus{}sum}\PYG{p}{(}\PYG{n}{loss\PYGZus{}steps\PYGZus{}x}\PYG{p}{)}

        \PYG{n}{loss\PYGZus{}steps\PYGZus{}y} \PYG{o}{=} \PYG{p}{[}
            \PYG{n}{tf}\PYG{o}{.}\PYG{n}{nn}\PYG{o}{.}\PYG{n}{l2\PYGZus{}loss}\PYG{p}{(}
                \PYG{p}{(}
                    \PYG{n}{vel\PYGZus{}gt}\PYG{p}{[}\PYG{n}{i}\PYG{p}{]}\PYG{o}{.}\PYG{n}{vector}\PYG{p}{[}\PYG{l+s+s1}{\PYGZsq{}}\PYG{l+s+s1}{y}\PYG{l+s+s1}{\PYGZsq{}}\PYG{p}{]}\PYG{o}{.}\PYG{n}{values}\PYG{o}{.}\PYG{n}{native}\PYG{p}{(}\PYG{p}{(}\PYG{l+s+s1}{\PYGZsq{}}\PYG{l+s+s1}{batch}\PYG{l+s+s1}{\PYGZsq{}}\PYG{p}{,} \PYG{l+s+s1}{\PYGZsq{}}\PYG{l+s+s1}{y}\PYG{l+s+s1}{\PYGZsq{}}\PYG{p}{,} \PYG{l+s+s1}{\PYGZsq{}}\PYG{l+s+s1}{x}\PYG{l+s+s1}{\PYGZsq{}}\PYG{p}{)}\PYG{p}{)}
                    \PYG{o}{\PYGZhy{}} \PYG{n}{prediction}\PYG{p}{[}\PYG{n}{i}\PYG{p}{]}\PYG{p}{[}\PYG{l+m+mi}{1}\PYG{p}{]}\PYG{o}{.}\PYG{n}{vector}\PYG{p}{[}\PYG{l+s+s1}{\PYGZsq{}}\PYG{l+s+s1}{y}\PYG{l+s+s1}{\PYGZsq{}}\PYG{p}{]}\PYG{o}{.}\PYG{n}{values}\PYG{o}{.}\PYG{n}{native}\PYG{p}{(}\PYG{p}{(}\PYG{l+s+s1}{\PYGZsq{}}\PYG{l+s+s1}{batch}\PYG{l+s+s1}{\PYGZsq{}}\PYG{p}{,} \PYG{l+s+s1}{\PYGZsq{}}\PYG{l+s+s1}{y}\PYG{l+s+s1}{\PYGZsq{}}\PYG{p}{,} \PYG{l+s+s1}{\PYGZsq{}}\PYG{l+s+s1}{x}\PYG{l+s+s1}{\PYGZsq{}}\PYG{p}{)}\PYG{p}{)}
                \PYG{p}{)}\PYG{o}{/}\PYG{n}{dataset}\PYG{o}{.}\PYG{n}{dataStats}\PYG{p}{[}\PYG{l+s+s1}{\PYGZsq{}}\PYG{l+s+s1}{std}\PYG{l+s+s1}{\PYGZsq{}}\PYG{p}{]}\PYG{p}{[}\PYG{l+m+mi}{2}\PYG{p}{]}
            \PYG{p}{)}
            \PYG{k}{for} \PYG{n}{i} \PYG{o+ow}{in} \PYG{n+nb}{range}\PYG{p}{(}\PYG{l+m+mi}{1}\PYG{p}{,}\PYG{n}{msteps}\PYG{o}{+}\PYG{l+m+mi}{1}\PYG{p}{)}
        \PYG{p}{]}
        \PYG{n}{loss\PYGZus{}steps\PYGZus{}y\PYGZus{}sum} \PYG{o}{=} \PYG{n}{tf}\PYG{o}{.}\PYG{n}{math}\PYG{o}{.}\PYG{n}{reduce\PYGZus{}sum}\PYG{p}{(}\PYG{n}{loss\PYGZus{}steps\PYGZus{}y}\PYG{p}{)}

        \PYG{n}{loss} \PYG{o}{=} \PYG{p}{(}\PYG{n}{loss\PYGZus{}steps\PYGZus{}x\PYGZus{}sum} \PYG{o}{+} \PYG{n}{loss\PYGZus{}steps\PYGZus{}y\PYGZus{}sum}\PYG{p}{)}\PYG{o}{/}\PYG{n}{msteps}

        \PYG{n}{gradients} \PYG{o}{=} \PYG{n}{tape}\PYG{o}{.}\PYG{n}{gradient}\PYG{p}{(}\PYG{n}{loss}\PYG{p}{,} \PYG{n}{network}\PYG{o}{.}\PYG{n}{trainable\PYGZus{}variables}\PYG{p}{)}
        \PYG{n}{opt}\PYG{o}{.}\PYG{n}{apply\PYGZus{}gradients}\PYG{p}{(}\PYG{n+nb}{zip}\PYG{p}{(}\PYG{n}{gradients}\PYG{p}{,} \PYG{n}{network}\PYG{o}{.}\PYG{n}{trainable\PYGZus{}variables}\PYG{p}{)}\PYG{p}{)}

        \PYG{k}{return} \PYG{n}{math}\PYG{o}{.}\PYG{n}{tensor}\PYG{p}{(}\PYG{n}{loss}\PYG{p}{)}    
\end{sphinxVerbatim}

Once defined, we can prepare this function for executing the training step by calling phiflow’s \sphinxcode{\sphinxupquote{math.jit\_compile()}} function. It automatically maps to the correct pre\sphinxhyphen{}compilation step of the chosen backend. E.g., for TF this internally creates a computational graph, and optimizes the chain of operations. For JAX, it can even compile optimized GPU code (if JAX is set up correctly). Using the jit compilation can make a huge difference in terms of runtime.

\begin{sphinxVerbatim}[commandchars=\\\{\}]
\PYG{n}{training\PYGZus{}step\PYGZus{}jit} \PYG{o}{=} \PYG{n}{math}\PYG{o}{.}\PYG{n}{jit\PYGZus{}compile}\PYG{p}{(}\PYG{n}{training\PYGZus{}step}\PYG{p}{)}
\end{sphinxVerbatim}

\section{Training}
\label{\detokenize{diffphys-code-sol:training}}
For the training, we use a standard Adam optimizer, and run 15 epochs by default. This should be increased for the larger network or to obtain more accurate results. For longer training runs, it would also be beneficial to decrease the learning rate over the course of the epochs, but for simplicity, we’ll keep \sphinxcode{\sphinxupquote{LR}} constant here.

Optionally, this is also the right point to load a network state to resume training.

\begin{sphinxVerbatim}[commandchars=\\\{\}]
\PYG{n}{LR} \PYG{o}{=} \PYG{l+m+mf}{1e\PYGZhy{}4}
\PYG{n}{EPOCHS} \PYG{o}{=} \PYG{l+m+mi}{15}

\PYG{n}{opt} \PYG{o}{=} \PYG{n}{tf}\PYG{o}{.}\PYG{n}{keras}\PYG{o}{.}\PYG{n}{optimizers}\PYG{o}{.}\PYG{n}{Adam}\PYG{p}{(}\PYG{n}{learning\PYGZus{}rate}\PYG{o}{=}\PYG{n}{LR}\PYG{p}{)} 

\PYG{c+c1}{\PYGZsh{} optional, load existing network...}
\PYG{c+c1}{\PYGZsh{} set to epoch nr. to load existing network from there}
\PYG{n}{resume} \PYG{o}{=} \PYG{l+m+mi}{0}
\PYG{k}{if} \PYG{n}{resume}\PYG{o}{\PYGZgt{}}\PYG{l+m+mi}{0}\PYG{p}{:} 
    \PYG{n}{ld\PYGZus{}network} \PYG{o}{=} \PYG{n}{keras}\PYG{o}{.}\PYG{n}{models}\PYG{o}{.}\PYG{n}{load\PYGZus{}model}\PYG{p}{(}\PYG{l+s+s1}{\PYGZsq{}}\PYG{l+s+s1}{./nn\PYGZus{}epoch}\PYG{l+s+si}{\PYGZob{}:04d\PYGZcb{}}\PYG{l+s+s1}{.h5}\PYG{l+s+s1}{\PYGZsq{}}\PYG{o}{.}\PYG{n}{format}\PYG{p}{(}\PYG{n}{resume}\PYG{p}{)}\PYG{p}{)} 
    \PYG{c+c1}{\PYGZsh{}ld\PYGZus{}network = keras.models.load\PYGZus{}model(\PYGZsq{}./nn\PYGZus{}final.h5\PYGZsq{}) \PYGZsh{} or the last one}
    \PYG{n}{network}\PYG{o}{.}\PYG{n}{set\PYGZus{}weights}\PYG{p}{(}\PYG{n}{ld\PYGZus{}network}\PYG{o}{.}\PYG{n}{get\PYGZus{}weights}\PYG{p}{(}\PYG{p}{)}\PYG{p}{)}
    
\end{sphinxVerbatim}

Finally, we can start training the NN! This is very straight forward now, we simply loop over the desired number of iterations, get a batch each time via \sphinxcode{\sphinxupquote{getData}}, feed it into the source simulation input \sphinxcode{\sphinxupquote{source\_in}}, and compare it in the loss with the \sphinxcode{\sphinxupquote{reference}} data for the batch.

The setup above will automatically take care that the differentiable physics solver used here provides the right gradient information, and provides it to the tensorflow network. Be warned: due to the complexity of the setup, this training run can take a while… (If you have a saved \sphinxcode{\sphinxupquote{nn\_final.h5}} network from a previous run, you can potentially skip this block and load the previously trained model instead via the cell above.)

\begin{sphinxVerbatim}[commandchars=\\\{\}]
\PYG{n}{steps} \PYG{o}{=} \PYG{l+m+mi}{0}
\PYG{k}{for} \PYG{n}{j} \PYG{o+ow}{in} \PYG{n+nb}{range}\PYG{p}{(}\PYG{n}{EPOCHS}\PYG{p}{)}\PYG{p}{:}  \PYG{c+c1}{\PYGZsh{} training}
    \PYG{n}{dataset}\PYG{o}{.}\PYG{n}{newEpoch}\PYG{p}{(}\PYG{n}{exclude\PYGZus{}tail}\PYG{o}{=}\PYG{n}{msteps}\PYG{p}{)}
    \PYG{k}{if} \PYG{n}{j}\PYG{o}{\PYGZlt{}}\PYG{n}{resume}\PYG{p}{:}
        \PYG{n+nb}{print}\PYG{p}{(}\PYG{l+s+s1}{\PYGZsq{}}\PYG{l+s+s1}{resume: skipping }\PYG{l+s+si}{\PYGZob{}\PYGZcb{}}\PYG{l+s+s1}{ epoch}\PYG{l+s+s1}{\PYGZsq{}}\PYG{o}{.}\PYG{n}{format}\PYG{p}{(}\PYG{n}{j}\PYG{o}{+}\PYG{l+m+mi}{1}\PYG{p}{)}\PYG{p}{)}
        \PYG{n}{steps} \PYG{o}{+}\PYG{o}{=} \PYG{n}{dataset}\PYG{o}{.}\PYG{n}{numSteps}\PYG{o}{*}\PYG{n}{dataset}\PYG{o}{.}\PYG{n}{numBatches}
        \PYG{k}{continue}

    \PYG{k}{for} \PYG{n}{ib} \PYG{o+ow}{in} \PYG{n+nb}{range}\PYG{p}{(}\PYG{n}{dataset}\PYG{o}{.}\PYG{n}{numBatches}\PYG{p}{)}\PYG{p}{:}   
        \PYG{k}{for} \PYG{n}{i} \PYG{o+ow}{in} \PYG{n+nb}{range}\PYG{p}{(}\PYG{n}{dataset}\PYG{o}{.}\PYG{n}{numSteps}\PYG{p}{)}\PYG{p}{:} 

            \PYG{c+c1}{\PYGZsh{} batch: [[dens0, dens1, ...], [x\PYGZhy{}velo0, x\PYGZhy{}velo1, ...], [y\PYGZhy{}velo0, y\PYGZhy{}velo1, ...], [ReynoldsNr(s)]]            }
            \PYG{n}{batch} \PYG{o}{=} \PYG{n}{getData}\PYG{p}{(}\PYG{n}{dataset}\PYG{p}{,} \PYG{n}{consecutive\PYGZus{}frames}\PYG{o}{=}\PYG{n}{msteps}\PYG{p}{)}
            
            \PYG{n}{dens\PYGZus{}gt} \PYG{o}{=} \PYG{p}{[}   \PYG{c+c1}{\PYGZsh{} [density0:CenteredGrid, density1, ...]}
                \PYG{n}{domain}\PYG{o}{.}\PYG{n}{scalar\PYGZus{}grid}\PYG{p}{(}
                    \PYG{n}{math}\PYG{o}{.}\PYG{n}{tensor}\PYG{p}{(}\PYG{n}{batch}\PYG{p}{[}\PYG{l+m+mi}{0}\PYG{p}{]}\PYG{p}{[}\PYG{n}{k}\PYG{p}{]}\PYG{p}{,} \PYG{n}{math}\PYG{o}{.}\PYG{n}{batch}\PYG{p}{(}\PYG{l+s+s1}{\PYGZsq{}}\PYG{l+s+s1}{batch}\PYG{l+s+s1}{\PYGZsq{}}\PYG{p}{)}\PYG{p}{,} \PYG{n}{math}\PYG{o}{.}\PYG{n}{spatial}\PYG{p}{(}\PYG{l+s+s1}{\PYGZsq{}}\PYG{l+s+s1}{y, x}\PYG{l+s+s1}{\PYGZsq{}}\PYG{p}{)}\PYG{p}{)}
                \PYG{p}{)} \PYG{k}{for} \PYG{n}{k} \PYG{o+ow}{in} \PYG{n+nb}{range}\PYG{p}{(}\PYG{n}{msteps}\PYG{o}{+}\PYG{l+m+mi}{1}\PYG{p}{)}
            \PYG{p}{]}

            \PYG{n}{vel\PYGZus{}gt} \PYG{o}{=} \PYG{p}{[}   \PYG{c+c1}{\PYGZsh{} [velocity0:StaggeredGrid, velocity1, ...]}
                \PYG{n}{domain}\PYG{o}{.}\PYG{n}{staggered\PYGZus{}grid}\PYG{p}{(}
                    \PYG{n}{math}\PYG{o}{.}\PYG{n}{stack}\PYG{p}{(}
                        \PYG{p}{[}
                            \PYG{n}{math}\PYG{o}{.}\PYG{n}{tensor}\PYG{p}{(}\PYG{n}{batch}\PYG{p}{[}\PYG{l+m+mi}{2}\PYG{p}{]}\PYG{p}{[}\PYG{n}{k}\PYG{p}{]}\PYG{p}{,} \PYG{n}{math}\PYG{o}{.}\PYG{n}{batch}\PYG{p}{(}\PYG{l+s+s1}{\PYGZsq{}}\PYG{l+s+s1}{batch}\PYG{l+s+s1}{\PYGZsq{}}\PYG{p}{)}\PYG{p}{,} \PYG{n}{math}\PYG{o}{.}\PYG{n}{spatial}\PYG{p}{(}\PYG{l+s+s1}{\PYGZsq{}}\PYG{l+s+s1}{y, x}\PYG{l+s+s1}{\PYGZsq{}}\PYG{p}{)}\PYG{p}{)}\PYG{p}{,}
                            \PYG{n}{math}\PYG{o}{.}\PYG{n}{tensor}\PYG{p}{(}\PYG{n}{batch}\PYG{p}{[}\PYG{l+m+mi}{1}\PYG{p}{]}\PYG{p}{[}\PYG{n}{k}\PYG{p}{]}\PYG{p}{,} \PYG{n}{math}\PYG{o}{.}\PYG{n}{batch}\PYG{p}{(}\PYG{l+s+s1}{\PYGZsq{}}\PYG{l+s+s1}{batch}\PYG{l+s+s1}{\PYGZsq{}}\PYG{p}{)}\PYG{p}{,} \PYG{n}{math}\PYG{o}{.}\PYG{n}{spatial}\PYG{p}{(}\PYG{l+s+s1}{\PYGZsq{}}\PYG{l+s+s1}{y, x}\PYG{l+s+s1}{\PYGZsq{}}\PYG{p}{)}\PYG{p}{)}\PYG{p}{,}
                        \PYG{p}{]}\PYG{p}{,} \PYG{n}{math}\PYG{o}{.}\PYG{n}{channel}\PYG{p}{(}\PYG{l+s+s1}{\PYGZsq{}}\PYG{l+s+s1}{vector}\PYG{l+s+s1}{\PYGZsq{}}\PYG{p}{)}
                    \PYG{p}{)}
                \PYG{p}{)} \PYG{k}{for} \PYG{n}{k} \PYG{o+ow}{in} \PYG{n+nb}{range}\PYG{p}{(}\PYG{n}{msteps}\PYG{o}{+}\PYG{l+m+mi}{1}\PYG{p}{)}
            \PYG{p}{]}
            \PYG{n}{re\PYGZus{}nr} \PYG{o}{=} \PYG{n}{math}\PYG{o}{.}\PYG{n}{tensor}\PYG{p}{(}\PYG{n}{batch}\PYG{p}{[}\PYG{l+m+mi}{3}\PYG{p}{]}\PYG{p}{,} \PYG{n}{math}\PYG{o}{.}\PYG{n}{batch}\PYG{p}{(}\PYG{l+s+s1}{\PYGZsq{}}\PYG{l+s+s1}{batch}\PYG{l+s+s1}{\PYGZsq{}}\PYG{p}{)}\PYG{p}{)}

            \PYG{n}{loss} \PYG{o}{=} \PYG{n}{training\PYGZus{}step\PYGZus{}jit}\PYG{p}{(}\PYG{n}{dens\PYGZus{}gt}\PYG{p}{,} \PYG{n}{vel\PYGZus{}gt}\PYG{p}{,} \PYG{n}{re\PYGZus{}nr}\PYG{p}{,} \PYG{n}{math}\PYG{o}{.}\PYG{n}{tensor}\PYG{p}{(}\PYG{n}{steps}\PYG{p}{)}\PYG{p}{)} 
            
            \PYG{n}{steps} \PYG{o}{+}\PYG{o}{=} \PYG{l+m+mi}{1}
            \PYG{k}{if} \PYG{p}{(}\PYG{n}{j}\PYG{o}{==}\PYG{l+m+mi}{0} \PYG{o+ow}{and} \PYG{n}{ib}\PYG{o}{==}\PYG{l+m+mi}{0} \PYG{o+ow}{and} \PYG{n}{i}\PYG{o}{\PYGZlt{}}\PYG{l+m+mi}{3}\PYG{p}{)} \PYG{o+ow}{or} \PYG{p}{(}\PYG{n}{j}\PYG{o}{==}\PYG{l+m+mi}{0} \PYG{o+ow}{and} \PYG{n}{ib}\PYG{o}{==}\PYG{l+m+mi}{0} \PYG{o+ow}{and} \PYG{n}{i}\PYG{o}{\PYGZpc{}}\PYG{k}{128}==0) or (j\PYGZgt{}0 and ib==0 and i==400): \PYGZsh{} reduce output 
              \PYG{n+nb}{print}\PYG{p}{(}\PYG{l+s+s1}{\PYGZsq{}}\PYG{l+s+s1}{epoch }\PYG{l+s+si}{\PYGZob{}:03d\PYGZcb{}}\PYG{l+s+s1}{/}\PYG{l+s+si}{\PYGZob{}:03d\PYGZcb{}}\PYG{l+s+s1}{, batch }\PYG{l+s+si}{\PYGZob{}:03d\PYGZcb{}}\PYG{l+s+s1}{/}\PYG{l+s+si}{\PYGZob{}:03d\PYGZcb{}}\PYG{l+s+s1}{, step }\PYG{l+s+si}{\PYGZob{}:04d\PYGZcb{}}\PYG{l+s+s1}{/}\PYG{l+s+si}{\PYGZob{}:04d\PYGZcb{}}\PYG{l+s+s1}{: loss=}\PYG{l+s+si}{\PYGZob{}\PYGZcb{}}\PYG{l+s+s1}{\PYGZsq{}}\PYG{o}{.}\PYG{n}{format}\PYG{p}{(} \PYG{n}{j}\PYG{o}{+}\PYG{l+m+mi}{1}\PYG{p}{,} \PYG{n}{EPOCHS}\PYG{p}{,} \PYG{n}{ib}\PYG{o}{+}\PYG{l+m+mi}{1}\PYG{p}{,} \PYG{n}{dataset}\PYG{o}{.}\PYG{n}{numBatches}\PYG{p}{,} \PYG{n}{i}\PYG{o}{+}\PYG{l+m+mi}{1}\PYG{p}{,} \PYG{n}{dataset}\PYG{o}{.}\PYG{n}{numSteps}\PYG{p}{,} \PYG{n}{loss} \PYG{p}{)}\PYG{p}{)}
            
            \PYG{n}{dataset}\PYG{o}{.}\PYG{n}{nextStep}\PYG{p}{(}\PYG{p}{)}

        \PYG{n}{dataset}\PYG{o}{.}\PYG{n}{nextBatch}\PYG{p}{(}\PYG{p}{)}

    \PYG{k}{if} \PYG{n}{j}\PYG{o}{\PYGZpc{}}\PYG{k}{10}==9: network.save(\PYGZsq{}./nn\PYGZus{}epoch\PYGZob{}:04d\PYGZcb{}.h5\PYGZsq{}.format(j+1))

\PYG{c+c1}{\PYGZsh{} all done! save final version}
\PYG{n}{network}\PYG{o}{.}\PYG{n}{save}\PYG{p}{(}\PYG{l+s+s1}{\PYGZsq{}}\PYG{l+s+s1}{./nn\PYGZus{}final.h5}\PYG{l+s+s1}{\PYGZsq{}}\PYG{p}{)}\PYG{p}{;} \PYG{n+nb}{print}\PYG{p}{(}\PYG{l+s+s2}{\PYGZdq{}}\PYG{l+s+s2}{Training done, saved NN}\PYG{l+s+s2}{\PYGZdq{}}\PYG{p}{)}
\end{sphinxVerbatim}

\begin{sphinxVerbatim}[commandchars=\\\{\}]
epoch 001/015, batch 001/002, step 0001/0496: loss=2605.340576171875
epoch 001/015, batch 001/002, step 0002/0496: loss=1485.1646728515625
epoch 001/015, batch 001/002, step 0003/0496: loss=790.8267211914062
epoch 001/015, batch 001/002, step 0129/0496: loss=98.64994049072266
epoch 001/015, batch 001/002, step 0257/0496: loss=75.3546142578125
epoch 001/015, batch 001/002, step 0385/0496: loss=70.05519104003906
epoch 002/015, batch 001/002, step 0401/0496: loss=19.126527786254883
epoch 003/015, batch 001/002, step 0401/0496: loss=9.628664016723633
epoch 004/015, batch 001/002, step 0401/0496: loss=7.898053169250488
epoch 005/015, batch 001/002, step 0401/0496: loss=3.6936004161834717
epoch 006/015, batch 001/002, step 0401/0496: loss=3.172729730606079
epoch 007/015, batch 001/002, step 0401/0496: loss=2.8511123657226562
epoch 008/015, batch 001/002, step 0401/0496: loss=3.4968295097351074
epoch 009/015, batch 001/002, step 0401/0496: loss=1.6942076683044434
epoch 010/015, batch 001/002, step 0401/0496: loss=1.6551270484924316
epoch 011/015, batch 001/002, step 0401/0496: loss=1.9383186101913452
epoch 012/015, batch 001/002, step 0401/0496: loss=2.0140795707702637
epoch 013/015, batch 001/002, step 0401/0496: loss=1.4174892902374268
epoch 014/015, batch 001/002, step 0401/0496: loss=1.2593278884887695
epoch 015/015, batch 001/002, step 0401/0496: loss=1.250532627105713
\end{sphinxVerbatim}

\begin{sphinxVerbatim}[commandchars=\\\{\}]
Training done, saved NN
\end{sphinxVerbatim}

The loss should go down from above 1000 initially to below 10. This is a good sign, but of course it’s even more important to see how the NN\sphinxhyphen{}solver combination fares on new inputs. With this training approach we’ve realized a hybrid solver, consisting of a regular \sphinxstyleemphasis{source} simulator, and a network that was trained to specifically interact with this simulator for a chosen domain of simulation cases.

Let’s see how well this works by applying it to a set of test data inputs with new Reynolds numbers that were not part of the training data.

To keep things somewhat simple, we won’t aim for a high\sphinxhyphen{}performance version of our hybrid solver. For performance, please check out the external code base: the network trained here should be directly useable in \sphinxhref{https://github.com/tum-pbs/Solver-in-the-Loop/blob/master/karman-2d/karman\_apply.py}{this apply script}.

\bigskip\hrule\bigskip

\section{Evaluation}
\label{\detokenize{diffphys-code-sol:evaluation}}
In order to evaluate the performance of our DL\sphinxhyphen{}powered solver, we essentially only need to repeat the inner loop of each training iteration for more steps. While we were limited to \sphinxcode{\sphinxupquote{msteps}} evaluations at training time, we can now run our solver for arbitrary lengths. This is a good test for how well our solver has learned to keep the data within the desired distribution, and represents a generalization test for longer rollouts.

We can reuse the solver code from above, but in the following, we will consider two simulated versions: for comparison, we’ll run one reference simulation in the \sphinxstyleemphasis{source} space (i.e., without any modifications). This version receives the regular outputs of each evaluation of the simulator, and ignores the learned correction (stored in \sphinxcode{\sphinxupquote{steps\_source}} below). The second version, repeatedly computes the source solver plus the learned correction, and advances this state in the solver (\sphinxcode{\sphinxupquote{steps\_hybrid}}).

We also need a set of new data. Below, we’ll download a new set of Reynolds numbers (in between the ones used for training), so that we can later on run the unmodified simulator and the DL\sphinxhyphen{}powered one on these cases.

\begin{sphinxVerbatim}[commandchars=\\\{\}]
\PYG{n}{fname\PYGZus{}test} \PYG{o}{=} \PYG{l+s+s1}{\PYGZsq{}}\PYG{l+s+s1}{sol\PYGZhy{}karman\PYGZhy{}2d\PYGZhy{}test.pickle}\PYG{l+s+s1}{\PYGZsq{}}
\PYG{k}{if} \PYG{o+ow}{not} \PYG{n}{os}\PYG{o}{.}\PYG{n}{path}\PYG{o}{.}\PYG{n}{isfile}\PYG{p}{(}\PYG{n}{fname\PYGZus{}test}\PYG{p}{)}\PYG{p}{:}
  \PYG{n+nb}{print}\PYG{p}{(}\PYG{l+s+s2}{\PYGZdq{}}\PYG{l+s+s2}{Downloading test data (38MB), this can take a moment the first time...}\PYG{l+s+s2}{\PYGZdq{}}\PYG{p}{)}
  \PYG{n}{urllib}\PYG{o}{.}\PYG{n}{request}\PYG{o}{.}\PYG{n}{urlretrieve}\PYG{p}{(}\PYG{l+s+s2}{\PYGZdq{}}\PYG{l+s+s2}{https://physicsbaseddeeplearning.org/data/}\PYG{l+s+s2}{\PYGZdq{}}\PYG{o}{+}\PYG{n}{fname\PYGZus{}test}\PYG{p}{,} \PYG{n}{fname\PYGZus{}test}\PYG{p}{)}

\PYG{k}{with} \PYG{n+nb}{open}\PYG{p}{(}\PYG{n}{fname\PYGZus{}test}\PYG{p}{,} \PYG{l+s+s1}{\PYGZsq{}}\PYG{l+s+s1}{rb}\PYG{l+s+s1}{\PYGZsq{}}\PYG{p}{)} \PYG{k}{as} \PYG{n}{f}\PYG{p}{:} \PYG{n}{data\PYGZus{}test\PYGZus{}preloaded} \PYG{o}{=} \PYG{n}{pickle}\PYG{o}{.}\PYG{n}{load}\PYG{p}{(}\PYG{n}{f}\PYG{p}{)}
\PYG{n+nb}{print}\PYG{p}{(}\PYG{l+s+s2}{\PYGZdq{}}\PYG{l+s+s2}{Loaded test data, }\PYG{l+s+si}{\PYGZob{}\PYGZcb{}}\PYG{l+s+s2}{ training sims}\PYG{l+s+s2}{\PYGZdq{}}\PYG{o}{.}\PYG{n}{format}\PYG{p}{(}\PYG{n+nb}{len}\PYG{p}{(}\PYG{n}{data\PYGZus{}test\PYGZus{}preloaded}\PYG{p}{)}\PYG{p}{)} \PYG{p}{)}
\end{sphinxVerbatim}

\begin{sphinxVerbatim}[commandchars=\\\{\}]
Downloading test data (38MB), this can take a moment the first time...
Loaded test data, 4 training sims
\end{sphinxVerbatim}

Next we create a new dataset object \sphinxcode{\sphinxupquote{dataset\_test}} that organizes the data. We’re simply using the first batch of the unshuffled dataset, though.

A subtle but important point: we still have to use the normalization from the original training data set: \sphinxcode{\sphinxupquote{dataset.dataStats{[}'std'{]}}} values. The test data set has it’s own mean and standard deviation, and so the trained NN never saw this data before. The NN was trained with the data in \sphinxcode{\sphinxupquote{dataset}} above, and hence we have to use the constants from there for normalization to make sure the network receives values that it can relate to the data it was trained with.

\begin{sphinxVerbatim}[commandchars=\\\{\}]
\PYG{n}{dataset\PYGZus{}test} \PYG{o}{=} \PYG{n}{Dataset}\PYG{p}{(} \PYG{n}{data\PYGZus{}preloaded}\PYG{o}{=}\PYG{n}{data\PYGZus{}test\PYGZus{}preloaded}\PYG{p}{,} \PYG{n}{is\PYGZus{}testset}\PYG{o}{=}\PYG{k+kc}{True}\PYG{p}{,} \PYG{n}{num\PYGZus{}frames}\PYG{o}{=}\PYG{n}{simsteps}\PYG{p}{,} \PYG{n}{num\PYGZus{}sims}\PYG{o}{=}\PYG{l+m+mi}{4}\PYG{p}{,} \PYG{n}{batch\PYGZus{}size}\PYG{o}{=}\PYG{l+m+mi}{4} \PYG{p}{)}

\PYG{c+c1}{\PYGZsh{} we only need 1 batch with t=0 states to initialize the test simulations with}
\PYG{n}{dataset\PYGZus{}test}\PYG{o}{.}\PYG{n}{newEpoch}\PYG{p}{(}\PYG{n}{shuffle\PYGZus{}data}\PYG{o}{=}\PYG{k+kc}{False}\PYG{p}{)}
\PYG{n}{batch} \PYG{o}{=} \PYG{n}{getData}\PYG{p}{(}\PYG{n}{dataset\PYGZus{}test}\PYG{p}{,} \PYG{n}{consecutive\PYGZus{}frames}\PYG{o}{=}\PYG{l+m+mi}{0}\PYG{p}{)} 

\PYG{n}{re\PYGZus{}nr\PYGZus{}test} \PYG{o}{=} \PYG{n}{math}\PYG{o}{.}\PYG{n}{tensor}\PYG{p}{(}\PYG{n}{batch}\PYG{p}{[}\PYG{l+m+mi}{3}\PYG{p}{]}\PYG{p}{,} \PYG{n}{math}\PYG{o}{.}\PYG{n}{batch}\PYG{p}{(}\PYG{l+s+s1}{\PYGZsq{}}\PYG{l+s+s1}{batch}\PYG{l+s+s1}{\PYGZsq{}}\PYG{p}{)}\PYG{p}{)} \PYG{c+c1}{\PYGZsh{} Reynolds numbers}
\PYG{n+nb}{print}\PYG{p}{(}\PYG{l+s+s2}{\PYGZdq{}}\PYG{l+s+s2}{Reynolds numbers in test data set: }\PYG{l+s+s2}{\PYGZdq{}}\PYG{o}{+}\PYG{n+nb}{format}\PYG{p}{(}\PYG{n}{re\PYGZus{}nr\PYGZus{}test}\PYG{p}{)}\PYG{p}{)}
\end{sphinxVerbatim}

\begin{sphinxVerbatim}[commandchars=\\\{\}]
Reynolds numbers in test data set: (120000.0, 480000.0, 1920000.0, 7680000.0) along batchᵇ
\end{sphinxVerbatim}

Next we construct a \sphinxcode{\sphinxupquote{math.tensor}} as initial state for the centered marker fields, and a staggered grid from the next two indices of the test set batch. Similar to \sphinxcode{\sphinxupquote{to\_phiflow}} above, we can use \sphinxcode{\sphinxupquote{phi.math.stack()}} to combine two fields of appropriate size as a staggered grid.

\begin{sphinxVerbatim}[commandchars=\\\{\}]
\PYG{n}{source\PYGZus{}dens\PYGZus{}initial} \PYG{o}{=} \PYG{n}{math}\PYG{o}{.}\PYG{n}{tensor}\PYG{p}{(} \PYG{n}{batch}\PYG{p}{[}\PYG{l+m+mi}{0}\PYG{p}{]}\PYG{p}{[}\PYG{l+m+mi}{0}\PYG{p}{]}\PYG{p}{,} \PYG{n}{math}\PYG{o}{.}\PYG{n}{batch}\PYG{p}{(}\PYG{l+s+s1}{\PYGZsq{}}\PYG{l+s+s1}{batch}\PYG{l+s+s1}{\PYGZsq{}}\PYG{p}{)}\PYG{p}{,} \PYG{n}{math}\PYG{o}{.}\PYG{n}{spatial}\PYG{p}{(}\PYG{l+s+s1}{\PYGZsq{}}\PYG{l+s+s1}{y, x}\PYG{l+s+s1}{\PYGZsq{}}\PYG{p}{)}\PYG{p}{)}

\PYG{n}{source\PYGZus{}vel\PYGZus{}initial} \PYG{o}{=} \PYG{n}{domain}\PYG{o}{.}\PYG{n}{staggered\PYGZus{}grid}\PYG{p}{(}\PYG{n}{phi}\PYG{o}{.}\PYG{n}{math}\PYG{o}{.}\PYG{n}{stack}\PYG{p}{(}\PYG{p}{[}
    \PYG{n}{math}\PYG{o}{.}\PYG{n}{tensor}\PYG{p}{(}\PYG{n}{batch}\PYG{p}{[}\PYG{l+m+mi}{2}\PYG{p}{]}\PYG{p}{[}\PYG{l+m+mi}{0}\PYG{p}{]}\PYG{p}{,} \PYG{n}{math}\PYG{o}{.}\PYG{n}{batch}\PYG{p}{(}\PYG{l+s+s1}{\PYGZsq{}}\PYG{l+s+s1}{batch}\PYG{l+s+s1}{\PYGZsq{}}\PYG{p}{)}\PYG{p}{,}\PYG{n}{math}\PYG{o}{.}\PYG{n}{spatial}\PYG{p}{(}\PYG{l+s+s1}{\PYGZsq{}}\PYG{l+s+s1}{y, x}\PYG{l+s+s1}{\PYGZsq{}}\PYG{p}{)}\PYG{p}{)}\PYG{p}{,}
    \PYG{n}{math}\PYG{o}{.}\PYG{n}{tensor}\PYG{p}{(}\PYG{n}{batch}\PYG{p}{[}\PYG{l+m+mi}{1}\PYG{p}{]}\PYG{p}{[}\PYG{l+m+mi}{0}\PYG{p}{]}\PYG{p}{,} \PYG{n}{math}\PYG{o}{.}\PYG{n}{batch}\PYG{p}{(}\PYG{l+s+s1}{\PYGZsq{}}\PYG{l+s+s1}{batch}\PYG{l+s+s1}{\PYGZsq{}}\PYG{p}{)}\PYG{p}{,}\PYG{n}{math}\PYG{o}{.}\PYG{n}{spatial}\PYG{p}{(}\PYG{l+s+s1}{\PYGZsq{}}\PYG{l+s+s1}{y, x}\PYG{l+s+s1}{\PYGZsq{}}\PYG{p}{)}\PYG{p}{)}\PYG{p}{]}\PYG{p}{,} \PYG{n}{channel}\PYG{p}{(}\PYG{l+s+s1}{\PYGZsq{}}\PYG{l+s+s1}{vector}\PYG{l+s+s1}{\PYGZsq{}}\PYG{p}{)}\PYG{p}{)}\PYG{p}{)}
\end{sphinxVerbatim}

Now we can first run the \sphinxstyleemphasis{source} simulation for 120 steps as baseline:

\begin{sphinxVerbatim}[commandchars=\\\{\}]
\PYG{n}{source\PYGZus{}dens\PYGZus{}test}\PYG{p}{,} \PYG{n}{source\PYGZus{}vel\PYGZus{}test} \PYG{o}{=} \PYG{n}{source\PYGZus{}dens\PYGZus{}initial}\PYG{p}{,} \PYG{n}{source\PYGZus{}vel\PYGZus{}initial}
\PYG{n}{steps\PYGZus{}source} \PYG{o}{=} \PYG{p}{[}\PYG{p}{[}\PYG{n}{source\PYGZus{}dens\PYGZus{}test}\PYG{p}{,}\PYG{n}{source\PYGZus{}vel\PYGZus{}test}\PYG{p}{]}\PYG{p}{]}

\PYG{c+c1}{\PYGZsh{} note \PYGZhy{} math.jit\PYGZus{}compile() not useful for numpy solve... hence not necessary}
\PYG{k}{for} \PYG{n}{i} \PYG{o+ow}{in} \PYG{n+nb}{range}\PYG{p}{(}\PYG{l+m+mi}{120}\PYG{p}{)}\PYG{p}{:}
    \PYG{p}{[}\PYG{n}{source\PYGZus{}dens\PYGZus{}test}\PYG{p}{,}\PYG{n}{source\PYGZus{}vel\PYGZus{}test}\PYG{p}{]} \PYG{o}{=} \PYG{n}{simulator}\PYG{o}{.}\PYG{n}{step}\PYG{p}{(}
        \PYG{n}{density\PYGZus{}in}\PYG{o}{=}\PYG{n}{source\PYGZus{}dens\PYGZus{}test}\PYG{p}{,}
        \PYG{n}{velocity\PYGZus{}in}\PYG{o}{=}\PYG{n}{source\PYGZus{}vel\PYGZus{}test}\PYG{p}{,}
        \PYG{n}{re}\PYG{o}{=}\PYG{n}{re\PYGZus{}nr\PYGZus{}test}\PYG{p}{,}
        \PYG{n}{res}\PYG{o}{=}\PYG{n}{source\PYGZus{}res}\PYG{p}{[}\PYG{l+m+mi}{1}\PYG{p}{]}\PYG{p}{,}
    \PYG{p}{)}
    \PYG{n}{steps\PYGZus{}source}\PYG{o}{.}\PYG{n}{append}\PYG{p}{(} \PYG{p}{[}\PYG{n}{source\PYGZus{}dens\PYGZus{}test}\PYG{p}{,}\PYG{n}{source\PYGZus{}vel\PYGZus{}test}\PYG{p}{]} \PYG{p}{)}

\PYG{n+nb}{print}\PYG{p}{(}\PYG{l+s+s2}{\PYGZdq{}}\PYG{l+s+s2}{Source simulation steps }\PYG{l+s+s2}{\PYGZdq{}}\PYG{o}{+}\PYG{n+nb}{format}\PYG{p}{(}\PYG{n+nb}{len}\PYG{p}{(}\PYG{n}{steps\PYGZus{}source}\PYG{p}{)}\PYG{p}{)}\PYG{p}{)}
\end{sphinxVerbatim}

\begin{sphinxVerbatim}[commandchars=\\\{\}]
Source simulation steps 121
\end{sphinxVerbatim}

Next, we compute the corresponding states of our learned hybrid solver. Here, we closely follow the training code, however, now without any gradient tapes or loss computations. We only evaluate the NN in a forward pass for each simulated state to compute a correction field:

\begin{sphinxVerbatim}[commandchars=\\\{\}]
\PYG{n}{source\PYGZus{}dens\PYGZus{}test}\PYG{p}{,} \PYG{n}{source\PYGZus{}vel\PYGZus{}test} \PYG{o}{=} \PYG{n}{source\PYGZus{}dens\PYGZus{}initial}\PYG{p}{,} \PYG{n}{source\PYGZus{}vel\PYGZus{}initial}
\PYG{n}{steps\PYGZus{}hybrid} \PYG{o}{=} \PYG{p}{[}\PYG{p}{[}\PYG{n}{source\PYGZus{}dens\PYGZus{}test}\PYG{p}{,}\PYG{n}{source\PYGZus{}vel\PYGZus{}test}\PYG{p}{]}\PYG{p}{]}
        
\PYG{k}{for} \PYG{n}{i} \PYG{o+ow}{in} \PYG{n+nb}{range}\PYG{p}{(}\PYG{l+m+mi}{120}\PYG{p}{)}\PYG{p}{:}
    \PYG{p}{[}\PYG{n}{source\PYGZus{}dens\PYGZus{}test}\PYG{p}{,}\PYG{n}{source\PYGZus{}vel\PYGZus{}test}\PYG{p}{]} \PYG{o}{=} \PYG{n}{simulator}\PYG{o}{.}\PYG{n}{step}\PYG{p}{(}
        \PYG{n}{density\PYGZus{}in}\PYG{o}{=}\PYG{n}{source\PYGZus{}dens\PYGZus{}test}\PYG{p}{,}
        \PYG{n}{velocity\PYGZus{}in}\PYG{o}{=}\PYG{n}{source\PYGZus{}vel\PYGZus{}test}\PYG{p}{,}
        \PYG{n}{re}\PYG{o}{=}\PYG{n}{math}\PYG{o}{.}\PYG{n}{tensor}\PYG{p}{(}\PYG{n}{re\PYGZus{}nr\PYGZus{}test}\PYG{p}{)}\PYG{p}{,}
        \PYG{n}{res}\PYG{o}{=}\PYG{n}{source\PYGZus{}res}\PYG{p}{[}\PYG{l+m+mi}{1}\PYG{p}{]}\PYG{p}{,}
    \PYG{p}{)}
    \PYG{n}{model\PYGZus{}input} \PYG{o}{=} \PYG{n}{to\PYGZus{}keras}\PYG{p}{(}\PYG{p}{[}\PYG{n}{source\PYGZus{}dens\PYGZus{}test}\PYG{p}{,}\PYG{n}{source\PYGZus{}vel\PYGZus{}test}\PYG{p}{]}\PYG{p}{,} \PYG{n}{re\PYGZus{}nr\PYGZus{}test} \PYG{p}{)}
    \PYG{n}{model\PYGZus{}input} \PYG{o}{/}\PYG{o}{=} \PYG{n}{math}\PYG{o}{.}\PYG{n}{tensor}\PYG{p}{(}\PYG{p}{[}\PYG{n}{dataset}\PYG{o}{.}\PYG{n}{dataStats}\PYG{p}{[}\PYG{l+s+s1}{\PYGZsq{}}\PYG{l+s+s1}{std}\PYG{l+s+s1}{\PYGZsq{}}\PYG{p}{]}\PYG{p}{[}\PYG{l+m+mi}{1}\PYG{p}{]}\PYG{p}{,} \PYG{n}{dataset}\PYG{o}{.}\PYG{n}{dataStats}\PYG{p}{[}\PYG{l+s+s1}{\PYGZsq{}}\PYG{l+s+s1}{std}\PYG{l+s+s1}{\PYGZsq{}}\PYG{p}{]}\PYG{p}{[}\PYG{l+m+mi}{2}\PYG{p}{]}\PYG{p}{,} \PYG{n}{dataset}\PYG{o}{.}\PYG{n}{dataStats}\PYG{p}{[}\PYG{l+s+s1}{\PYGZsq{}}\PYG{l+s+s1}{ext.std}\PYG{l+s+s1}{\PYGZsq{}}\PYG{p}{]}\PYG{p}{[}\PYG{l+m+mi}{0}\PYG{p}{]}\PYG{p}{]}\PYG{p}{,} \PYG{n}{channel}\PYG{p}{(}\PYG{l+s+s1}{\PYGZsq{}}\PYG{l+s+s1}{channels}\PYG{l+s+s1}{\PYGZsq{}}\PYG{p}{)}\PYG{p}{)} \PYG{c+c1}{\PYGZsh{} [u, v, Re]}
    \PYG{n}{model\PYGZus{}out} \PYG{o}{=} \PYG{n}{network}\PYG{p}{(}\PYG{n}{model\PYGZus{}input}\PYG{o}{.}\PYG{n}{native}\PYG{p}{(}\PYG{p}{[}\PYG{l+s+s1}{\PYGZsq{}}\PYG{l+s+s1}{batch}\PYG{l+s+s1}{\PYGZsq{}}\PYG{p}{,} \PYG{l+s+s1}{\PYGZsq{}}\PYG{l+s+s1}{y}\PYG{l+s+s1}{\PYGZsq{}}\PYG{p}{,} \PYG{l+s+s1}{\PYGZsq{}}\PYG{l+s+s1}{x}\PYG{l+s+s1}{\PYGZsq{}}\PYG{p}{,} \PYG{l+s+s1}{\PYGZsq{}}\PYG{l+s+s1}{channels}\PYG{l+s+s1}{\PYGZsq{}}\PYG{p}{]}\PYG{p}{)}\PYG{p}{,} \PYG{n}{training}\PYG{o}{=}\PYG{k+kc}{False}\PYG{p}{)}
    \PYG{n}{model\PYGZus{}out} \PYG{o}{*}\PYG{o}{=} \PYG{p}{[}\PYG{n}{dataset}\PYG{o}{.}\PYG{n}{dataStats}\PYG{p}{[}\PYG{l+s+s1}{\PYGZsq{}}\PYG{l+s+s1}{std}\PYG{l+s+s1}{\PYGZsq{}}\PYG{p}{]}\PYG{p}{[}\PYG{l+m+mi}{1}\PYG{p}{]}\PYG{p}{,} \PYG{n}{dataset}\PYG{o}{.}\PYG{n}{dataStats}\PYG{p}{[}\PYG{l+s+s1}{\PYGZsq{}}\PYG{l+s+s1}{std}\PYG{l+s+s1}{\PYGZsq{}}\PYG{p}{]}\PYG{p}{[}\PYG{l+m+mi}{2}\PYG{p}{]}\PYG{p}{]} \PYG{c+c1}{\PYGZsh{} [u, v]}
    \PYG{n}{correction} \PYG{o}{=}  \PYG{n}{to\PYGZus{}phiflow}\PYG{p}{(}\PYG{n}{model\PYGZus{}out}\PYG{p}{,} \PYG{n}{domain}\PYG{p}{)} 
    \PYG{n}{source\PYGZus{}vel\PYGZus{}test} \PYG{o}{=} \PYG{n}{source\PYGZus{}vel\PYGZus{}test}\PYG{o}{+}\PYG{n}{correction}

    \PYG{n}{steps\PYGZus{}hybrid}\PYG{o}{.}\PYG{n}{append}\PYG{p}{(} \PYG{p}{[}\PYG{n}{source\PYGZus{}dens\PYGZus{}test}\PYG{p}{,}\PYG{n}{source\PYGZus{}vel\PYGZus{}test}\PYG{o}{+}\PYG{n}{correction}\PYG{p}{]} \PYG{p}{)}
    
\PYG{n+nb}{print}\PYG{p}{(}\PYG{l+s+s2}{\PYGZdq{}}\PYG{l+s+s2}{Steps with hybrid solver }\PYG{l+s+s2}{\PYGZdq{}}\PYG{o}{+}\PYG{n+nb}{format}\PYG{p}{(}\PYG{n+nb}{len}\PYG{p}{(}\PYG{n}{steps\PYGZus{}hybrid}\PYG{p}{)}\PYG{p}{)}\PYG{p}{)}
\end{sphinxVerbatim}

\begin{sphinxVerbatim}[commandchars=\\\{\}]
Steps with hybrid solver 121
\end{sphinxVerbatim}

Given the stored states, we quantify the improvements that the NN yields, and visualize the results.

In the following cells, the index \sphinxcode{\sphinxupquote{b}} chooses one of the four test simulations (by default index 0, the lowest Re outside the training data range), and computes the accumulated mean absolute error (MAE) over all time steps.

\begin{sphinxVerbatim}[commandchars=\\\{\}]
\PYG{k+kn}{import} \PYG{n+nn}{pylab}
\PYG{n}{b} \PYG{o}{=} \PYG{l+m+mi}{0} \PYG{c+c1}{\PYGZsh{} batch index for the following comparisons}

\PYG{n}{errors\PYGZus{}source}\PYG{p}{,} \PYG{n}{errors\PYGZus{}pred} \PYG{o}{=} \PYG{p}{[}\PYG{p}{]}\PYG{p}{,} \PYG{p}{[}\PYG{p}{]}
\PYG{k}{for} \PYG{n}{index} \PYG{o+ow}{in} \PYG{n+nb}{range}\PYG{p}{(}\PYG{l+m+mi}{100}\PYG{p}{)}\PYG{p}{:}
  \PYG{n}{vx\PYGZus{}ref} \PYG{o}{=} \PYG{n}{dataset\PYGZus{}test}\PYG{o}{.}\PYG{n}{dataPreloaded}\PYG{p}{[} \PYG{n}{dataset\PYGZus{}test}\PYG{o}{.}\PYG{n}{dataSims}\PYG{p}{[}\PYG{n}{b}\PYG{p}{]} \PYG{p}{]}\PYG{p}{[} \PYG{n}{index} \PYG{p}{]}\PYG{p}{[}\PYG{l+m+mi}{1}\PYG{p}{]}\PYG{p}{[}\PYG{l+m+mi}{0}\PYG{p}{,}\PYG{o}{.}\PYG{o}{.}\PYG{o}{.}\PYG{p}{]}
  \PYG{n}{vy\PYGZus{}ref} \PYG{o}{=} \PYG{n}{dataset\PYGZus{}test}\PYG{o}{.}\PYG{n}{dataPreloaded}\PYG{p}{[} \PYG{n}{dataset\PYGZus{}test}\PYG{o}{.}\PYG{n}{dataSims}\PYG{p}{[}\PYG{n}{b}\PYG{p}{]} \PYG{p}{]}\PYG{p}{[} \PYG{n}{index} \PYG{p}{]}\PYG{p}{[}\PYG{l+m+mi}{2}\PYG{p}{]}\PYG{p}{[}\PYG{l+m+mi}{0}\PYG{p}{,}\PYG{o}{.}\PYG{o}{.}\PYG{o}{.}\PYG{p}{]}
  \PYG{n}{vxs} \PYG{o}{=} \PYG{n}{vx\PYGZus{}ref} \PYG{o}{\PYGZhy{}} \PYG{n}{steps\PYGZus{}source}\PYG{p}{[}\PYG{n}{index}\PYG{p}{]}\PYG{p}{[}\PYG{l+m+mi}{1}\PYG{p}{]}\PYG{o}{.}\PYG{n}{values}\PYG{o}{.}\PYG{n}{vector}\PYG{p}{[}\PYG{l+m+mi}{1}\PYG{p}{]}\PYG{o}{.}\PYG{n}{numpy}\PYG{p}{(}\PYG{l+s+s1}{\PYGZsq{}}\PYG{l+s+s1}{batch,y,x}\PYG{l+s+s1}{\PYGZsq{}}\PYG{p}{)}\PYG{p}{[}\PYG{n}{b}\PYG{p}{,}\PYG{o}{.}\PYG{o}{.}\PYG{o}{.}\PYG{p}{]}
  \PYG{n}{vxh} \PYG{o}{=} \PYG{n}{vx\PYGZus{}ref} \PYG{o}{\PYGZhy{}} \PYG{n}{steps\PYGZus{}hybrid}\PYG{p}{[}\PYG{n}{index}\PYG{p}{]}\PYG{p}{[}\PYG{l+m+mi}{1}\PYG{p}{]}\PYG{o}{.}\PYG{n}{values}\PYG{o}{.}\PYG{n}{vector}\PYG{p}{[}\PYG{l+m+mi}{1}\PYG{p}{]}\PYG{o}{.}\PYG{n}{numpy}\PYG{p}{(}\PYG{l+s+s1}{\PYGZsq{}}\PYG{l+s+s1}{batch,y,x}\PYG{l+s+s1}{\PYGZsq{}}\PYG{p}{)}\PYG{p}{[}\PYG{n}{b}\PYG{p}{,}\PYG{o}{.}\PYG{o}{.}\PYG{o}{.}\PYG{p}{]}
  \PYG{n}{vys} \PYG{o}{=} \PYG{n}{vy\PYGZus{}ref} \PYG{o}{\PYGZhy{}} \PYG{n}{steps\PYGZus{}source}\PYG{p}{[}\PYG{n}{index}\PYG{p}{]}\PYG{p}{[}\PYG{l+m+mi}{1}\PYG{p}{]}\PYG{o}{.}\PYG{n}{values}\PYG{o}{.}\PYG{n}{vector}\PYG{p}{[}\PYG{l+m+mi}{0}\PYG{p}{]}\PYG{o}{.}\PYG{n}{numpy}\PYG{p}{(}\PYG{l+s+s1}{\PYGZsq{}}\PYG{l+s+s1}{batch,y,x}\PYG{l+s+s1}{\PYGZsq{}}\PYG{p}{)}\PYG{p}{[}\PYG{n}{b}\PYG{p}{,}\PYG{o}{.}\PYG{o}{.}\PYG{o}{.}\PYG{p}{]} 
  \PYG{n}{vyh} \PYG{o}{=} \PYG{n}{vy\PYGZus{}ref} \PYG{o}{\PYGZhy{}} \PYG{n}{steps\PYGZus{}hybrid}\PYG{p}{[}\PYG{n}{index}\PYG{p}{]}\PYG{p}{[}\PYG{l+m+mi}{1}\PYG{p}{]}\PYG{o}{.}\PYG{n}{values}\PYG{o}{.}\PYG{n}{vector}\PYG{p}{[}\PYG{l+m+mi}{0}\PYG{p}{]}\PYG{o}{.}\PYG{n}{numpy}\PYG{p}{(}\PYG{l+s+s1}{\PYGZsq{}}\PYG{l+s+s1}{batch,y,x}\PYG{l+s+s1}{\PYGZsq{}}\PYG{p}{)}\PYG{p}{[}\PYG{n}{b}\PYG{p}{,}\PYG{o}{.}\PYG{o}{.}\PYG{o}{.}\PYG{p}{]} 
  \PYG{n}{errors\PYGZus{}source}\PYG{o}{.}\PYG{n}{append}\PYG{p}{(}\PYG{n}{np}\PYG{o}{.}\PYG{n}{mean}\PYG{p}{(}\PYG{n}{np}\PYG{o}{.}\PYG{n}{abs}\PYG{p}{(}\PYG{n}{vxs}\PYG{p}{)}\PYG{p}{)} \PYG{o}{+} \PYG{n}{np}\PYG{o}{.}\PYG{n}{mean}\PYG{p}{(}\PYG{n}{np}\PYG{o}{.}\PYG{n}{abs}\PYG{p}{(}\PYG{n}{vys}\PYG{p}{)}\PYG{p}{)}\PYG{p}{)} 
  \PYG{n}{errors\PYGZus{}pred}\PYG{o}{.}\PYG{n}{append}\PYG{p}{(}\PYG{n}{np}\PYG{o}{.}\PYG{n}{mean}\PYG{p}{(}\PYG{n}{np}\PYG{o}{.}\PYG{n}{abs}\PYG{p}{(}\PYG{n}{vxh}\PYG{p}{)}\PYG{p}{)} \PYG{o}{+} \PYG{n}{np}\PYG{o}{.}\PYG{n}{mean}\PYG{p}{(}\PYG{n}{np}\PYG{o}{.}\PYG{n}{abs}\PYG{p}{(}\PYG{n}{vyh}\PYG{p}{)}\PYG{p}{)}\PYG{p}{)}

\PYG{n}{fig} \PYG{o}{=} \PYG{n}{pylab}\PYG{o}{.}\PYG{n}{figure}\PYG{p}{(}\PYG{p}{)}\PYG{o}{.}\PYG{n}{gca}\PYG{p}{(}\PYG{p}{)}
\PYG{n}{pltx} \PYG{o}{=} \PYG{n}{np}\PYG{o}{.}\PYG{n}{linspace}\PYG{p}{(}\PYG{l+m+mi}{0}\PYG{p}{,}\PYG{l+m+mi}{99}\PYG{p}{,}\PYG{l+m+mi}{100}\PYG{p}{)}
\PYG{n}{fig}\PYG{o}{.}\PYG{n}{plot}\PYG{p}{(}\PYG{n}{pltx}\PYG{p}{,} \PYG{n}{errors\PYGZus{}source}\PYG{p}{,} \PYG{n}{lw}\PYG{o}{=}\PYG{l+m+mi}{2}\PYG{p}{,} \PYG{n}{color}\PYG{o}{=}\PYG{l+s+s1}{\PYGZsq{}}\PYG{l+s+s1}{mediumblue}\PYG{l+s+s1}{\PYGZsq{}}\PYG{p}{,} \PYG{n}{label}\PYG{o}{=}\PYG{l+s+s1}{\PYGZsq{}}\PYG{l+s+s1}{Source}\PYG{l+s+s1}{\PYGZsq{}}\PYG{p}{)}  
\PYG{n}{fig}\PYG{o}{.}\PYG{n}{plot}\PYG{p}{(}\PYG{n}{pltx}\PYG{p}{,} \PYG{n}{errors\PYGZus{}pred}\PYG{p}{,}   \PYG{n}{lw}\PYG{o}{=}\PYG{l+m+mi}{2}\PYG{p}{,} \PYG{n}{color}\PYG{o}{=}\PYG{l+s+s1}{\PYGZsq{}}\PYG{l+s+s1}{green}\PYG{l+s+s1}{\PYGZsq{}}\PYG{p}{,} \PYG{n}{label}\PYG{o}{=}\PYG{l+s+s1}{\PYGZsq{}}\PYG{l+s+s1}{Hybrid}\PYG{l+s+s1}{\PYGZsq{}}\PYG{p}{)}
\PYG{n}{pylab}\PYG{o}{.}\PYG{n}{xlabel}\PYG{p}{(}\PYG{l+s+s1}{\PYGZsq{}}\PYG{l+s+s1}{Time step}\PYG{l+s+s1}{\PYGZsq{}}\PYG{p}{)}\PYG{p}{;} \PYG{n}{pylab}\PYG{o}{.}\PYG{n}{ylabel}\PYG{p}{(}\PYG{l+s+s1}{\PYGZsq{}}\PYG{l+s+s1}{Error}\PYG{l+s+s1}{\PYGZsq{}}\PYG{p}{)}\PYG{p}{;} \PYG{n}{fig}\PYG{o}{.}\PYG{n}{legend}\PYG{p}{(}\PYG{p}{)}

\PYG{n+nb}{print}\PYG{p}{(}\PYG{l+s+s2}{\PYGZdq{}}\PYG{l+s+s2}{MAE for source: }\PYG{l+s+s2}{\PYGZdq{}}\PYG{o}{+}\PYG{n+nb}{format}\PYG{p}{(}\PYG{n}{np}\PYG{o}{.}\PYG{n}{mean}\PYG{p}{(}\PYG{n}{errors\PYGZus{}source}\PYG{p}{)}\PYG{p}{)} \PYG{o}{+}\PYG{l+s+s2}{\PYGZdq{}}\PYG{l+s+s2}{ , and hybrid: }\PYG{l+s+s2}{\PYGZdq{}}\PYG{o}{+}\PYG{n+nb}{format}\PYG{p}{(}\PYG{n}{np}\PYG{o}{.}\PYG{n}{mean}\PYG{p}{(}\PYG{n}{errors\PYGZus{}pred}\PYG{p}{)}\PYG{p}{)} \PYG{p}{)}
\end{sphinxVerbatim}

\begin{sphinxVerbatim}[commandchars=\\\{\}]
MAE for source: 0.1363069713115692 , and hybrid: 0.05150971934199333
\end{sphinxVerbatim}

\noindent\sphinxincludegraphics{{diffphys-code-sol_42_1}.png}

Due to the complexity of the training, the performance can vary, but typically the overall MAE is ca. 160\% larger for the regular simulation compared to the hybrid simulator.
The gap is typically even bigger for other Reynolds numbers within the training data range.
The graph above also shows this behavior over time.

Let’s also visualize the differences of the two outputs by plotting the y component of the velocities over time. The two following code cells show six velocity snapshots for the batch index \sphinxcode{\sphinxupquote{b}} in intervals of 20 time steps.

\begin{sphinxVerbatim}[commandchars=\\\{\}]
\PYG{n}{c} \PYG{o}{=} \PYG{l+m+mi}{0}          \PYG{c+c1}{\PYGZsh{} channel selector, x=1 or y=0 }
\PYG{n}{interval} \PYG{o}{=} \PYG{l+m+mi}{20}  \PYG{c+c1}{\PYGZsh{} time interval}

\PYG{n}{fig}\PYG{p}{,} \PYG{n}{axes} \PYG{o}{=} \PYG{n}{pylab}\PYG{o}{.}\PYG{n}{subplots}\PYG{p}{(}\PYG{l+m+mi}{1}\PYG{p}{,} \PYG{l+m+mi}{6}\PYG{p}{,} \PYG{n}{figsize}\PYG{o}{=}\PYG{p}{(}\PYG{l+m+mi}{16}\PYG{p}{,} \PYG{l+m+mi}{5}\PYG{p}{)}\PYG{p}{)}    
\PYG{k}{for} \PYG{n}{i} \PYG{o+ow}{in} \PYG{n+nb}{range}\PYG{p}{(}\PYG{l+m+mi}{0}\PYG{p}{,}\PYG{l+m+mi}{6}\PYG{p}{)}\PYG{p}{:}
  \PYG{n}{v} \PYG{o}{=} \PYG{n}{steps\PYGZus{}source}\PYG{p}{[}\PYG{n}{i}\PYG{o}{*}\PYG{n}{interval}\PYG{p}{]}\PYG{p}{[}\PYG{l+m+mi}{1}\PYG{p}{]}\PYG{o}{.}\PYG{n}{values}\PYG{o}{.}\PYG{n}{vector}\PYG{p}{[}\PYG{n}{c}\PYG{p}{]}\PYG{o}{.}\PYG{n}{numpy}\PYG{p}{(}\PYG{l+s+s1}{\PYGZsq{}}\PYG{l+s+s1}{batch,y,x}\PYG{l+s+s1}{\PYGZsq{}}\PYG{p}{)}\PYG{p}{[}\PYG{n}{b}\PYG{p}{,}\PYG{o}{.}\PYG{o}{.}\PYG{o}{.}\PYG{p}{]}
  \PYG{n}{axes}\PYG{p}{[}\PYG{n}{i}\PYG{p}{]}\PYG{o}{.}\PYG{n}{imshow}\PYG{p}{(} \PYG{n}{v} \PYG{p}{,} \PYG{n}{origin}\PYG{o}{=}\PYG{l+s+s1}{\PYGZsq{}}\PYG{l+s+s1}{lower}\PYG{l+s+s1}{\PYGZsq{}}\PYG{p}{,} \PYG{n}{cmap}\PYG{o}{=}\PYG{l+s+s1}{\PYGZsq{}}\PYG{l+s+s1}{magma}\PYG{l+s+s1}{\PYGZsq{}}\PYG{p}{)}
  \PYG{n}{axes}\PYG{p}{[}\PYG{n}{i}\PYG{p}{]}\PYG{o}{.}\PYG{n}{set\PYGZus{}title}\PYG{p}{(}\PYG{l+s+sa}{f}\PYG{l+s+s2}{\PYGZdq{}}\PYG{l+s+s2}{ Source simulation t=}\PYG{l+s+si}{\PYGZob{}}\PYG{n}{i}\PYG{o}{*}\PYG{n}{interval}\PYG{l+s+si}{\PYGZcb{}}\PYG{l+s+s2}{ }\PYG{l+s+s2}{\PYGZdq{}}\PYG{p}{)}

\PYG{n}{pylab}\PYG{o}{.}\PYG{n}{tight\PYGZus{}layout}\PYG{p}{(}\PYG{p}{)}
\end{sphinxVerbatim}

\noindent\sphinxincludegraphics{{diffphys-code-sol_44_0}.png}

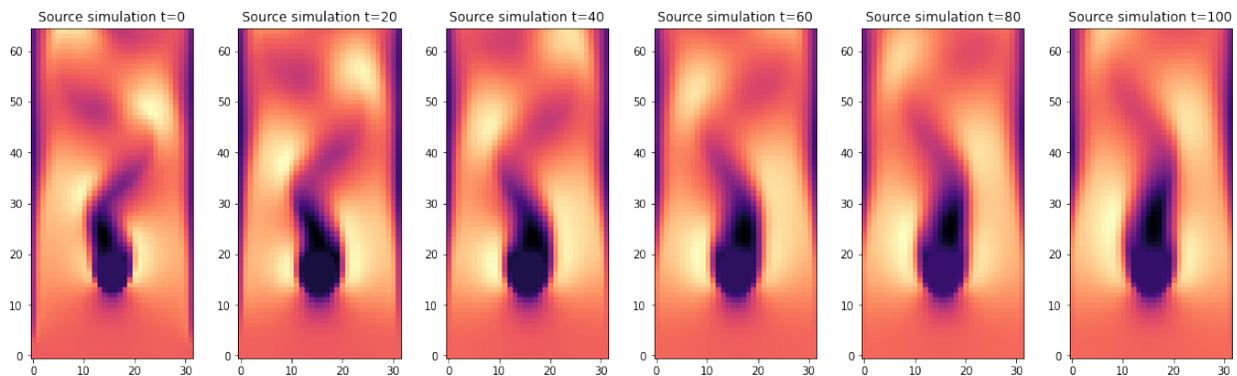
\begin{sphinxVerbatim}[commandchars=\\\{\}]
\PYG{n}{fig}\PYG{p}{,} \PYG{n}{axes} \PYG{o}{=} \PYG{n}{pylab}\PYG{o}{.}\PYG{n}{subplots}\PYG{p}{(}\PYG{l+m+mi}{1}\PYG{p}{,} \PYG{l+m+mi}{6}\PYG{p}{,} \PYG{n}{figsize}\PYG{o}{=}\PYG{p}{(}\PYG{l+m+mi}{16}\PYG{p}{,} \PYG{l+m+mi}{5}\PYG{p}{)}\PYG{p}{)}
\PYG{k}{for} \PYG{n}{i} \PYG{o+ow}{in} \PYG{n+nb}{range}\PYG{p}{(}\PYG{l+m+mi}{0}\PYG{p}{,}\PYG{l+m+mi}{6}\PYG{p}{)}\PYG{p}{:}
  \PYG{n}{v} \PYG{o}{=} \PYG{n}{steps\PYGZus{}hybrid}\PYG{p}{[}\PYG{n}{i}\PYG{o}{*}\PYG{n}{interval}\PYG{p}{]}\PYG{p}{[}\PYG{l+m+mi}{1}\PYG{p}{]}\PYG{o}{.}\PYG{n}{values}\PYG{o}{.}\PYG{n}{vector}\PYG{p}{[}\PYG{n}{c}\PYG{p}{]}\PYG{o}{.}\PYG{n}{numpy}\PYG{p}{(}\PYG{l+s+s1}{\PYGZsq{}}\PYG{l+s+s1}{batch,y,x}\PYG{l+s+s1}{\PYGZsq{}}\PYG{p}{)}\PYG{p}{[}\PYG{n}{b}\PYG{p}{,}\PYG{o}{.}\PYG{o}{.}\PYG{o}{.}\PYG{p}{]}
  \PYG{n}{axes}\PYG{p}{[}\PYG{n}{i}\PYG{p}{]}\PYG{o}{.}\PYG{n}{imshow}\PYG{p}{(} \PYG{n}{v} \PYG{p}{,} \PYG{n}{origin}\PYG{o}{=}\PYG{l+s+s1}{\PYGZsq{}}\PYG{l+s+s1}{lower}\PYG{l+s+s1}{\PYGZsq{}}\PYG{p}{,} \PYG{n}{cmap}\PYG{o}{=}\PYG{l+s+s1}{\PYGZsq{}}\PYG{l+s+s1}{magma}\PYG{l+s+s1}{\PYGZsq{}}\PYG{p}{)}
  \PYG{n}{axes}\PYG{p}{[}\PYG{n}{i}\PYG{p}{]}\PYG{o}{.}\PYG{n}{set\PYGZus{}title}\PYG{p}{(}\PYG{l+s+sa}{f}\PYG{l+s+s2}{\PYGZdq{}}\PYG{l+s+s2}{ Hybrid solver t=}\PYG{l+s+si}{\PYGZob{}}\PYG{n}{i}\PYG{o}{*}\PYG{n}{interval}\PYG{l+s+si}{\PYGZcb{}}\PYG{l+s+s2}{ }\PYG{l+s+s2}{\PYGZdq{}}\PYG{p}{)}
\PYG{n}{pylab}\PYG{o}{.}\PYG{n}{tight\PYGZus{}layout}\PYG{p}{(}\PYG{p}{)}
\end{sphinxVerbatim}

\noindent\sphinxincludegraphics{{diffphys-code-sol_45_0}.png}

They both start out with the same initial state at \(t=0\) (the downsampled solution from the reference solution manifold), and at \(t=20\) the solutions still share similarities. Over time, the source version strongly diffuses the structures in the flow and looses momentum. The flow behind the obstacles becomes straight, and lacks clear vortices.

The version produced by the hybrid solver does much better. It preserves the vortex shedding even after more than one hundred updates. Note that both outputs were produced by the same underlying solver. The second version just profits from the learned corrector which manages to revert the numerical errors of the source solver, including its overly strong dissipation.

We can also visually compare how the NN does w.r.t. reference data. The next cell plots one time step of the three versions: the reference data after 50 steps, and the re\sphinxhyphen{}simulated version of the source and our hybrid solver, together with a per\sphinxhyphen{}cell error of the two:

\begin{sphinxVerbatim}[commandchars=\\\{\}]
\PYG{n}{index} \PYG{o}{=} \PYG{l+m+mi}{50} \PYG{c+c1}{\PYGZsh{} time step index}
\PYG{n}{vx\PYGZus{}ref} \PYG{o}{=} \PYG{n}{dataset\PYGZus{}test}\PYG{o}{.}\PYG{n}{dataPreloaded}\PYG{p}{[} \PYG{n}{dataset\PYGZus{}test}\PYG{o}{.}\PYG{n}{dataSims}\PYG{p}{[}\PYG{n}{b}\PYG{p}{]} \PYG{p}{]}\PYG{p}{[} \PYG{n}{index} \PYG{p}{]}\PYG{p}{[}\PYG{l+m+mi}{1}\PYG{p}{]}\PYG{p}{[}\PYG{l+m+mi}{0}\PYG{p}{,}\PYG{o}{.}\PYG{o}{.}\PYG{o}{.}\PYG{p}{]}
\PYG{n}{vx\PYGZus{}src} \PYG{o}{=} \PYG{n}{steps\PYGZus{}source}\PYG{p}{[}\PYG{n}{index}\PYG{p}{]}\PYG{p}{[}\PYG{l+m+mi}{1}\PYG{p}{]}\PYG{o}{.}\PYG{n}{values}\PYG{o}{.}\PYG{n}{vector}\PYG{p}{[}\PYG{l+m+mi}{1}\PYG{p}{]}\PYG{o}{.}\PYG{n}{numpy}\PYG{p}{(}\PYG{l+s+s1}{\PYGZsq{}}\PYG{l+s+s1}{batch,y,x}\PYG{l+s+s1}{\PYGZsq{}}\PYG{p}{)}\PYG{p}{[}\PYG{n}{b}\PYG{p}{,}\PYG{o}{.}\PYG{o}{.}\PYG{o}{.}\PYG{p}{]}
\PYG{n}{vx\PYGZus{}hyb} \PYG{o}{=} \PYG{n}{steps\PYGZus{}hybrid}\PYG{p}{[}\PYG{n}{index}\PYG{p}{]}\PYG{p}{[}\PYG{l+m+mi}{1}\PYG{p}{]}\PYG{o}{.}\PYG{n}{values}\PYG{o}{.}\PYG{n}{vector}\PYG{p}{[}\PYG{l+m+mi}{1}\PYG{p}{]}\PYG{o}{.}\PYG{n}{numpy}\PYG{p}{(}\PYG{l+s+s1}{\PYGZsq{}}\PYG{l+s+s1}{batch,y,x}\PYG{l+s+s1}{\PYGZsq{}}\PYG{p}{)}\PYG{p}{[}\PYG{n}{b}\PYG{p}{,}\PYG{o}{.}\PYG{o}{.}\PYG{o}{.}\PYG{p}{]}

\PYG{n}{fig}\PYG{p}{,} \PYG{n}{axes} \PYG{o}{=} \PYG{n}{pylab}\PYG{o}{.}\PYG{n}{subplots}\PYG{p}{(}\PYG{l+m+mi}{1}\PYG{p}{,} \PYG{l+m+mi}{4}\PYG{p}{,} \PYG{n}{figsize}\PYG{o}{=}\PYG{p}{(}\PYG{l+m+mi}{14}\PYG{p}{,} \PYG{l+m+mi}{5}\PYG{p}{)}\PYG{p}{)}

\PYG{n}{axes}\PYG{p}{[}\PYG{l+m+mi}{0}\PYG{p}{]}\PYG{o}{.}\PYG{n}{imshow}\PYG{p}{(} \PYG{n}{vx\PYGZus{}ref} \PYG{p}{,} \PYG{n}{origin}\PYG{o}{=}\PYG{l+s+s1}{\PYGZsq{}}\PYG{l+s+s1}{lower}\PYG{l+s+s1}{\PYGZsq{}}\PYG{p}{,} \PYG{n}{cmap}\PYG{o}{=}\PYG{l+s+s1}{\PYGZsq{}}\PYG{l+s+s1}{magma}\PYG{l+s+s1}{\PYGZsq{}}\PYG{p}{)}
\PYG{n}{axes}\PYG{p}{[}\PYG{l+m+mi}{0}\PYG{p}{]}\PYG{o}{.}\PYG{n}{set\PYGZus{}title}\PYG{p}{(}\PYG{l+s+sa}{f}\PYG{l+s+s2}{\PYGZdq{}}\PYG{l+s+s2}{ Reference }\PYG{l+s+s2}{\PYGZdq{}}\PYG{p}{)}

\PYG{n}{axes}\PYG{p}{[}\PYG{l+m+mi}{1}\PYG{p}{]}\PYG{o}{.}\PYG{n}{imshow}\PYG{p}{(} \PYG{n}{vx\PYGZus{}src} \PYG{p}{,} \PYG{n}{origin}\PYG{o}{=}\PYG{l+s+s1}{\PYGZsq{}}\PYG{l+s+s1}{lower}\PYG{l+s+s1}{\PYGZsq{}}\PYG{p}{,} \PYG{n}{cmap}\PYG{o}{=}\PYG{l+s+s1}{\PYGZsq{}}\PYG{l+s+s1}{magma}\PYG{l+s+s1}{\PYGZsq{}}\PYG{p}{)}
\PYG{n}{axes}\PYG{p}{[}\PYG{l+m+mi}{1}\PYG{p}{]}\PYG{o}{.}\PYG{n}{set\PYGZus{}title}\PYG{p}{(}\PYG{l+s+sa}{f}\PYG{l+s+s2}{\PYGZdq{}}\PYG{l+s+s2}{ Source }\PYG{l+s+s2}{\PYGZdq{}}\PYG{p}{)}

\PYG{n}{axes}\PYG{p}{[}\PYG{l+m+mi}{2}\PYG{p}{]}\PYG{o}{.}\PYG{n}{imshow}\PYG{p}{(} \PYG{n}{vx\PYGZus{}hyb} \PYG{p}{,} \PYG{n}{origin}\PYG{o}{=}\PYG{l+s+s1}{\PYGZsq{}}\PYG{l+s+s1}{lower}\PYG{l+s+s1}{\PYGZsq{}}\PYG{p}{,} \PYG{n}{cmap}\PYG{o}{=}\PYG{l+s+s1}{\PYGZsq{}}\PYG{l+s+s1}{magma}\PYG{l+s+s1}{\PYGZsq{}}\PYG{p}{)}
\PYG{n}{axes}\PYG{p}{[}\PYG{l+m+mi}{2}\PYG{p}{]}\PYG{o}{.}\PYG{n}{set\PYGZus{}title}\PYG{p}{(}\PYG{l+s+sa}{f}\PYG{l+s+s2}{\PYGZdq{}}\PYG{l+s+s2}{ Learned }\PYG{l+s+s2}{\PYGZdq{}}\PYG{p}{)}

\PYG{c+c1}{\PYGZsh{} show error side by side}
\PYG{n}{err\PYGZus{}source} \PYG{o}{=} \PYG{n}{vx\PYGZus{}ref} \PYG{o}{\PYGZhy{}} \PYG{n}{vx\PYGZus{}src} 
\PYG{n}{err\PYGZus{}hybrid} \PYG{o}{=} \PYG{n}{vx\PYGZus{}ref} \PYG{o}{\PYGZhy{}} \PYG{n}{vx\PYGZus{}hyb} 
\PYG{n}{v} \PYG{o}{=} \PYG{n}{np}\PYG{o}{.}\PYG{n}{concatenate}\PYG{p}{(}\PYG{p}{[}\PYG{n}{err\PYGZus{}source}\PYG{p}{,}\PYG{n}{err\PYGZus{}hybrid}\PYG{p}{]}\PYG{p}{,} \PYG{n}{axis}\PYG{o}{=}\PYG{l+m+mi}{1}\PYG{p}{)}
\PYG{n}{axes}\PYG{p}{[}\PYG{l+m+mi}{3}\PYG{p}{]}\PYG{o}{.}\PYG{n}{imshow}\PYG{p}{(} \PYG{n}{v} \PYG{p}{,} \PYG{n}{origin}\PYG{o}{=}\PYG{l+s+s1}{\PYGZsq{}}\PYG{l+s+s1}{lower}\PYG{l+s+s1}{\PYGZsq{}}\PYG{p}{,} \PYG{n}{cmap}\PYG{o}{=}\PYG{l+s+s1}{\PYGZsq{}}\PYG{l+s+s1}{magma}\PYG{l+s+s1}{\PYGZsq{}}\PYG{p}{)}
\PYG{n}{axes}\PYG{p}{[}\PYG{l+m+mi}{3}\PYG{p}{]}\PYG{o}{.}\PYG{n}{set\PYGZus{}title}\PYG{p}{(}\PYG{l+s+sa}{f}\PYG{l+s+s2}{\PYGZdq{}}\PYG{l+s+s2}{ Errors: Source \PYGZam{} Learned}\PYG{l+s+s2}{\PYGZdq{}}\PYG{p}{)}

\PYG{n}{pylab}\PYG{o}{.}\PYG{n}{tight\PYGZus{}layout}\PYG{p}{(}\PYG{p}{)}
\end{sphinxVerbatim}

\noindent\sphinxincludegraphics{{diffphys-code-sol_47_0}.png}

This shows very clearly how the pure source simulation in the middle deviates from the reference on the left. The learned version stays much closer to the reference solution.

The two per\sphinxhyphen{}cell error images on the right also illustrate this: the source version has much larger errors (i.e. brighter colors) that show how it systematically underestimates the vortices that should form. The error for the learned version is much more evenly distributed and significantly smaller in magnitude.

This concludes our evaluation. Note that the improved behavior of the hybrid solver can be difficult to reliably measure with simple vector norms such as an MAE or \(L^2\) norm. To improve this, we’d need to employ other, domain\sphinxhyphen{}specific metrics. In this case, metrics for fluids based on vorticity and turbulence properties of the flow would be applicable. However, in this text, we instead want to focus on DL\sphinxhyphen{}related topics and target another inverse problem with differentiable physics solvers in the next chapter.

\section{Next steps}
\label{\detokenize{diffphys-code-sol:next-steps}}\begin{itemize}
\item {} 
Modify the training to further reduce the training error. With the \sphinxstyleemphasis{medium} network you should be able to get the loss down to around 1.

\item {} 
Turn off the differentiable physics training (by setting \sphinxcode{\sphinxupquote{msteps=1}}), and compare it with the DP version.

\item {} 
Likewise, train a network with a larger \sphinxcode{\sphinxupquote{msteps}} setting, e.g., 8 or 16. Note that due to the recurrent nature of the training, you’ll probably have to load a pre\sphinxhyphen{}trained state to stabilize the first iterations.

\item {} 
Use the external github code to generate new test data, and run your trained NN on these cases. You’ll see that a reduced training error not always directly correlates with an improved test performance.

\end{itemize}

\chapter{Solving Inverse Problems with NNs}
\label{\detokenize{diffphys-control:solving-inverse-problems-with-nns}}\label{\detokenize{diffphys-control::doc}}
Inverse problems encompass a large class of practical applications in science. In general, the goal here is not to directly compute a physical field like the velocity at a future time (this is the typical scenario for a \sphinxstyleemphasis{forward} solve), but instead more generically compute one or more parameters in the model equations such that certain constraints are fulfilled. A very common objective is to find the optimal setting for a single parameter given some constraints. E.g., this could be the global diffusion constant for an advection\sphinxhyphen{}diffusion model such that it fits measured data as accurately as possible. Inverse problems are encountered for any model parameter adjusted via observations, or the reconstruction of initial conditions, e.g., for particle imaging velocimetry (PIV). More complex cases aim for computing boundary geometries w.r.t. optimal conditions, e.g. to obtain a shape with minimal drag in a fluid flow.

A key aspect below will be that we’re not aiming for solving only a \sphinxstyleemphasis{single instance} of an inverse problem, but we’d like to use deep learning to solve a \sphinxstyleemphasis{larger collection} of inverse problems. Thus, unlike the physics\sphinxhyphen{}informed example of {\hyperref[\detokenize{physicalloss-code::doc}]{\sphinxcrossref{\DUrole{doc}{Burgers Optimization with a Physics\sphinxhyphen{}Informed NN}}}} or the differentiable physics (DP) optimization of {\hyperref[\detokenize{diffphys-code-ns::doc}]{\sphinxcrossref{\DUrole{doc}{Differentiable Fluid Simulations}}}}, where we’ve solved an optimization problem for specific instances of inverse problems, we now aim for training an NN that learns to solve a larger class of inverse problems, i.e., a whole solution manifold. Nonetheless, we of course need to rely on a certain degree of similarity for these problems, otherwise there’s nothing to learn (and the implied assumption of continuity in the solution manifold breaks down).

Below we will run a very challenging test case as a representative of these inverse problems: we will aim for computing a high dimensional control function that exerts forces over the full course of an incompressible fluid simulation in order to reach a desired goal state for a passively advected marker in the fluid. This means we only have very indirect constraints to be fulfilled (a single state at the end of a sequence), and a large number of degrees of freedom (the control force function is a space\sphinxhyphen{}time function with the same degrees of freedom as the flow field itself).

The \sphinxstyleemphasis{long\sphinxhyphen{}term} nature of the control is one of the aspects which makes this a tough inverse problem: any changes to the state of the physical system can lead to large change later on in time, and hence a controller needs to anticipate how the system will behave when it is influenced. This means an NN also needs to learn how the underlying physics evolve and change, and this is exactly where the gradients from the DP training come in to guide the learning task towards solution that can reach the goal.
\sphinxhref{https://colab.research.google.com/github/tum-pbs/pbdl-book/blob/main/diffphys-control.ipynb}{{[}run in colab{]}}

\section{Formulation}
\label{\detokenize{diffphys-control:formulation}}
With the notation from {\hyperref[\detokenize{overview-equations::doc}]{\sphinxcrossref{\DUrole{doc}{Models and Equations}}}} this gives the minimization problem
\begin{equation*}
\begin{split}
\text{arg min}_{\theta} \sum_m \sum_i (f(x_{m,i} ; \theta)-y^*_{m,i})^2 , 
\end{split}
\end{equation*}
where \(y^*_{m,i}\) denotes the samples of the target state of the marker field,
and \(x_{m,i}\) denotes the simulated state of the marker density.
As before, the index \(i\) samples our solution at different spatial locations (typically all grid cells), while the index \(m\) here indicates a large collection of different target states.

Our goal is to train two networks \(\mathrm{OP}\) and \(\mathrm{CFE}\) with weights
\(\theta_{\mathrm{OP}}\) and \(\theta_{\mathrm{CFE}}\) such that a sequence
\begin{equation*}
\begin{split}
\newcommand{\pde}{\mathcal{P}}
\newcommand{\net}{\mathrm{CFE}}
\mathbf{u}_{n},d_{n} = \pde(\net(~\pde(\net(\cdots \pde(\net( \mathbf{u}_0,d_0, d_{OP} ))\cdots)))) = (\pde~\net)^n ( \mathbf{u}_0,d_0, d_{OP} ) .
\end{split}
\end{equation*}
minimizes the loss above. The \(\mathrm{OP}\) network is a predictor that determines the state \(d_{OP}\) that the action of the \(\mathrm{CFE}\) should aim for, i.e., it does the longer term planning from which to determine the action. Given the target \(d^*\), it computes
\(d_{OP} = \mathrm{OP}(d,d^*)= f_{\mathrm{OP}}(d,d^{*};\theta_{\mathrm{OP}})\).
The \(\mathrm{CFE}\) acts additively on the velocity field by computing
\(\mathbf{u} + f_{\mathrm{CFE}}(\mathbf{u},d, f_{\mathrm{OP}}(d,d^{*};\theta_{\mathrm{OP}}) ;\theta_{\mathrm{CFE}}) \), where we’ve used \(f_{\mathrm{OP}}\) and \(f_{\mathrm{CFE}}\) to denote the NN representations of \(\mathrm{OP}\) and \(\mathrm{CFE}\), respectively, and \(d^{*}\) to denote the target density state. \(\theta_{\mathrm{OP}}\) and \(\theta_{\mathrm{CFE}}\) denote the corresponding network weights.

For this problem, the model PDE \(\mathcal{P}\) contains a discretized version of the incompressible Navier\sphinxhyphen{}Stokes equations in two dimensions for a velocity \(\mathbf{u}\):
\begin{equation*}
\begin{split}\begin{aligned}
  \frac{\partial u_x}{\partial{t}} + \mathbf{u} \cdot \nabla u_x &= - \frac{1}{\rho} \nabla p 
  \\
  \frac{\partial u_y}{\partial{t}} + \mathbf{u} \cdot \nabla u_y &= - \frac{1}{\rho} \nabla p 
  \\
  \text{s.t.} \quad \nabla \cdot \mathbf{u} &= 0,
\end{aligned}\end{split}
\end{equation*}
without explicit viscosity, and with an additional transport equation for the marker density \(d\) given by
\(\frac{\partial d}{\partial{t}} + \mathbf{u} \cdot \nabla d = 0\).

To summarize, we have a predictor \(\mathrm{OP}\) that gives us a direction, an actor \(\mathrm{CFE}\) that exerts a force on a physical model \(\mathcal{P}\). They all need to play hand in hand to reach a given target after \(n\) iterations of the simulation. As apparent from this formulation, it’s not a simple inverse problem, especially due to the fact that all three functions are non\sphinxhyphen{}linear. This is exactly why the gradients from the DP approach are so important. (The viewpoint above also indicates that \sphinxstyleemphasis{reinforcement learning} is a potential option. In {\hyperref[\detokenize{reinflearn-code::doc}]{\sphinxcrossref{\DUrole{doc}{Controlling Burgers’ Equation with Reinforcement Learning}}}} we’ll compare DP with these alternatives.)

\bigskip\hrule\bigskip

\section{Control of incompressible fluids}
\label{\detokenize{diffphys-control:control-of-incompressible-fluids}}
The next sections will walk you through all the necessary steps from data generation to network training using \sphinxhref{https://github.com/tum-pbs/PhiFlow}{ΦFlow}. Due to the complexity of the control problem,
we’ll start with a supervised initialization of the networks, before switching to a more accurate end\sphinxhyphen{}to\sphinxhyphen{}end training with DP.
(\sphinxstyleemphasis{Note: this example uses an older version 1.4.1 of ΦFlow.})

The code below replicates an inverse problem example (the shape transitions experiment) from \sphinxhref{https://ge.in.tum.de/publications/2020-iclr-holl/}{Learning to Control PDEs with Differentiable Physics} {[}\hyperlink{cite.references:id11}{HKT19}{]}, further details can be found in section D.2 of the paper’s \sphinxhref{https://openreview.net/pdf?id=HyeSin4FPB}{appendix}.

First we need to load phiflow and check out the \sphinxstyleemphasis{PDE\sphinxhyphen{}Control} git repository, which also contains some numpy arrays with initial shapes.

\begin{sphinxVerbatim}[commandchars=\\\{\}]
\PYG{o}{!}pip install \PYGZhy{}\PYGZhy{}upgrade \PYGZhy{}\PYGZhy{}quiet git+https://github.com/tum\PYGZhy{}pbs/PhiFlow@1.4.1

\PYG{k+kn}{import} \PYG{n+nn}{matplotlib}\PYG{n+nn}{.}\PYG{n+nn}{pyplot} \PYG{k}{as} \PYG{n+nn}{plt}
\PYG{k+kn}{from} \PYG{n+nn}{phi}\PYG{n+nn}{.}\PYG{n+nn}{flow} \PYG{k+kn}{import} \PYG{o}{*}

\PYG{k}{if} \PYG{o+ow}{not} \PYG{n}{os}\PYG{o}{.}\PYG{n}{path}\PYG{o}{.}\PYG{n}{isdir}\PYG{p}{(}\PYG{l+s+s1}{\PYGZsq{}}\PYG{l+s+s1}{PDE\PYGZhy{}Control}\PYG{l+s+s1}{\PYGZsq{}}\PYG{p}{)}\PYG{p}{:}
  \PYG{n+nb}{print}\PYG{p}{(}\PYG{l+s+s2}{\PYGZdq{}}\PYG{l+s+s2}{Cloning, PDE\PYGZhy{}Control repo, this can take a moment}\PYG{l+s+s2}{\PYGZdq{}}\PYG{p}{)}
  \PYG{n}{os}\PYG{o}{.}\PYG{n}{system}\PYG{p}{(}\PYG{l+s+s2}{\PYGZdq{}}\PYG{l+s+s2}{git clone \PYGZhy{}\PYGZhy{}recursive https://github.com/holl\PYGZhy{}/PDE\PYGZhy{}Control.git}\PYG{l+s+s2}{\PYGZdq{}}\PYG{p}{)}
    
\PYG{c+c1}{\PYGZsh{} now we can load the necessary phiflow libraries and helper functions}
\PYG{k+kn}{import} \PYG{n+nn}{sys}\PYG{p}{;} \PYG{n}{sys}\PYG{o}{.}\PYG{n}{path}\PYG{o}{.}\PYG{n}{append}\PYG{p}{(}\PYG{l+s+s1}{\PYGZsq{}}\PYG{l+s+s1}{PDE\PYGZhy{}Control/src}\PYG{l+s+s1}{\PYGZsq{}}\PYG{p}{)}
\PYG{k+kn}{from} \PYG{n+nn}{shape\PYGZus{}utils} \PYG{k+kn}{import} \PYG{n}{load\PYGZus{}shapes}\PYG{p}{,} \PYG{n}{distribute\PYGZus{}random\PYGZus{}shape}
\PYG{k+kn}{from} \PYG{n+nn}{control}\PYG{n+nn}{.}\PYG{n+nn}{pde}\PYG{n+nn}{.}\PYG{n+nn}{incompressible\PYGZus{}flow} \PYG{k+kn}{import} \PYG{n}{IncompressibleFluidPDE}
\PYG{k+kn}{from} \PYG{n+nn}{control}\PYG{n+nn}{.}\PYG{n+nn}{control\PYGZus{}training} \PYG{k+kn}{import} \PYG{n}{ControlTraining}
\PYG{k+kn}{from} \PYG{n+nn}{control}\PYG{n+nn}{.}\PYG{n+nn}{sequences} \PYG{k+kn}{import} \PYG{n}{StaggeredSequence}\PYG{p}{,} \PYG{n}{RefinedSequence}
\end{sphinxVerbatim}

\section{Data generation}
\label{\detokenize{diffphys-control:data-generation}}
Before starting the training, we have to generate a data set to train with, i.e., a set of ground truth time sequences \(u^*\). Due to the complexity of the training below, we’ll use a staged approach that pre\sphinxhyphen{}trains a supervised network as a rough initialization, and then refines it to learn control looking further and further ahead into the future. (This will be realized by training specialized NNs that deal with longer and longer sequences.)

First, let’s set up a domain and basic parameters of the data generation step.

\begin{sphinxVerbatim}[commandchars=\\\{\}]
\PYG{n}{domain} \PYG{o}{=} \PYG{n}{Domain}\PYG{p}{(}\PYG{p}{[}\PYG{l+m+mi}{64}\PYG{p}{,} \PYG{l+m+mi}{64}\PYG{p}{]}\PYG{p}{)}  \PYG{c+c1}{\PYGZsh{} 1D Grid resolution and physical size}
\PYG{n}{step\PYGZus{}count} \PYG{o}{=} \PYG{l+m+mi}{16}  \PYG{c+c1}{\PYGZsh{} how many solver steps to perform}
\PYG{n}{dt} \PYG{o}{=} \PYG{l+m+mf}{1.0}  \PYG{c+c1}{\PYGZsh{} Time increment per solver step}
\PYG{n}{example\PYGZus{}count} \PYG{o}{=} \PYG{l+m+mi}{1000}
\PYG{n}{batch\PYGZus{}size} \PYG{o}{=} \PYG{l+m+mi}{100}
\PYG{n}{data\PYGZus{}path} \PYG{o}{=} \PYG{l+s+s1}{\PYGZsq{}}\PYG{l+s+s1}{shape\PYGZhy{}transitions}\PYG{l+s+s1}{\PYGZsq{}}
\PYG{n}{pretrain\PYGZus{}data\PYGZus{}path} \PYG{o}{=} \PYG{l+s+s1}{\PYGZsq{}}\PYG{l+s+s1}{moving\PYGZhy{}squares}\PYG{l+s+s1}{\PYGZsq{}}
\PYG{n}{shape\PYGZus{}library} \PYG{o}{=} \PYG{n}{load\PYGZus{}shapes}\PYG{p}{(}\PYG{l+s+s1}{\PYGZsq{}}\PYG{l+s+s1}{PDE\PYGZhy{}Control/notebooks/shapes}\PYG{l+s+s1}{\PYGZsq{}}\PYG{p}{)}
\end{sphinxVerbatim}

The \sphinxcode{\sphinxupquote{shape\_library}} in the last line contains ten different shapes that we’ll use to intialize a marker density with at random positions.

This is what the shapes look like:

\begin{sphinxVerbatim}[commandchars=\\\{\}]
\PYG{k+kn}{import} \PYG{n+nn}{pylab}
\PYG{n}{pylab}\PYG{o}{.}\PYG{n}{subplots}\PYG{p}{(}\PYG{l+m+mi}{1}\PYG{p}{,} \PYG{n+nb}{len}\PYG{p}{(}\PYG{n}{shape\PYGZus{}library}\PYG{p}{)}\PYG{p}{,} \PYG{n}{figsize}\PYG{o}{=}\PYG{p}{(}\PYG{l+m+mi}{17}\PYG{p}{,} \PYG{l+m+mi}{5}\PYG{p}{)}\PYG{p}{)}
\PYG{k}{for} \PYG{n}{t} \PYG{o+ow}{in} \PYG{n+nb}{range}\PYG{p}{(}\PYG{n+nb}{len}\PYG{p}{(}\PYG{n}{shape\PYGZus{}library}\PYG{p}{)}\PYG{p}{)}\PYG{p}{:}
    \PYG{n}{pylab}\PYG{o}{.}\PYG{n}{subplot}\PYG{p}{(}\PYG{l+m+mi}{1}\PYG{p}{,} \PYG{n+nb}{len}\PYG{p}{(}\PYG{n}{shape\PYGZus{}library}\PYG{p}{)}\PYG{p}{,} \PYG{n}{t}\PYG{o}{+}\PYG{l+m+mi}{1}\PYG{p}{)}
    \PYG{n}{pylab}\PYG{o}{.}\PYG{n}{imshow}\PYG{p}{(}\PYG{n}{shape\PYGZus{}library}\PYG{p}{[}\PYG{n}{t}\PYG{p}{]}\PYG{p}{,} \PYG{n}{origin}\PYG{o}{=}\PYG{l+s+s1}{\PYGZsq{}}\PYG{l+s+s1}{lower}\PYG{l+s+s1}{\PYGZsq{}}\PYG{p}{)}
\end{sphinxVerbatim}

\noindent\sphinxincludegraphics{{diffphys-control_7_0}.png}

The following cell uses these shapes to create the dataset we want to train our network with.
Each example consists of a start and target (end) frame which are generated by placing a random shape from the \sphinxcode{\sphinxupquote{shape\_library}} somewhere within the domain.

\begin{sphinxVerbatim}[commandchars=\\\{\}]
\PYG{k}{for} \PYG{n}{scene} \PYG{o+ow}{in} \PYG{n}{Scene}\PYG{o}{.}\PYG{n}{list}\PYG{p}{(}\PYG{n}{data\PYGZus{}path}\PYG{p}{)}\PYG{p}{:} \PYG{n}{scene}\PYG{o}{.}\PYG{n}{remove}\PYG{p}{(}\PYG{p}{)}

\PYG{k}{for} \PYG{n}{\PYGZus{}} \PYG{o+ow}{in} \PYG{n+nb}{range}\PYG{p}{(}\PYG{n}{example\PYGZus{}count} \PYG{o}{/}\PYG{o}{/} \PYG{n}{batch\PYGZus{}size}\PYG{p}{)}\PYG{p}{:}
    \PYG{n}{scene} \PYG{o}{=} \PYG{n}{Scene}\PYG{o}{.}\PYG{n}{create}\PYG{p}{(}\PYG{n}{data\PYGZus{}path}\PYG{p}{,} \PYG{n}{count}\PYG{o}{=}\PYG{n}{batch\PYGZus{}size}\PYG{p}{,} \PYG{n}{copy\PYGZus{}calling\PYGZus{}script}\PYG{o}{=}\PYG{k+kc}{False}\PYG{p}{)}
    \PYG{n+nb}{print}\PYG{p}{(}\PYG{n}{scene}\PYG{p}{)}
    \PYG{n}{start} \PYG{o}{=} \PYG{n}{distribute\PYGZus{}random\PYGZus{}shape}\PYG{p}{(}\PYG{n}{domain}\PYG{o}{.}\PYG{n}{resolution}\PYG{p}{,} \PYG{n}{batch\PYGZus{}size}\PYG{p}{,} \PYG{n}{shape\PYGZus{}library}\PYG{p}{)}
    \PYG{n}{end\PYGZus{}\PYGZus{}} \PYG{o}{=} \PYG{n}{distribute\PYGZus{}random\PYGZus{}shape}\PYG{p}{(}\PYG{n}{domain}\PYG{o}{.}\PYG{n}{resolution}\PYG{p}{,} \PYG{n}{batch\PYGZus{}size}\PYG{p}{,} \PYG{n}{shape\PYGZus{}library}\PYG{p}{)}
    \PYG{p}{[}\PYG{n}{scene}\PYG{o}{.}\PYG{n}{write\PYGZus{}sim\PYGZus{}frame}\PYG{p}{(}\PYG{p}{[}\PYG{n}{start}\PYG{p}{]}\PYG{p}{,} \PYG{p}{[}\PYG{l+s+s1}{\PYGZsq{}}\PYG{l+s+s1}{density}\PYG{l+s+s1}{\PYGZsq{}}\PYG{p}{]}\PYG{p}{,} \PYG{n}{frame}\PYG{o}{=}\PYG{n}{f}\PYG{p}{)} \PYG{k}{for} \PYG{n}{f} \PYG{o+ow}{in} \PYG{n+nb}{range}\PYG{p}{(}\PYG{n}{step\PYGZus{}count}\PYG{p}{)}\PYG{p}{]}
    \PYG{n}{scene}\PYG{o}{.}\PYG{n}{write\PYGZus{}sim\PYGZus{}frame}\PYG{p}{(}\PYG{p}{[}\PYG{n}{end\PYGZus{}\PYGZus{}}\PYG{p}{]}\PYG{p}{,} \PYG{p}{[}\PYG{l+s+s1}{\PYGZsq{}}\PYG{l+s+s1}{density}\PYG{l+s+s1}{\PYGZsq{}}\PYG{p}{]}\PYG{p}{,} \PYG{n}{frame}\PYG{o}{=}\PYG{n}{step\PYGZus{}count}\PYG{p}{)}
\end{sphinxVerbatim}

\begin{sphinxVerbatim}[commandchars=\\\{\}]
shape\PYGZhy{}transitions/sim\PYGZus{}000000
shape\PYGZhy{}transitions/sim\PYGZus{}000100
shape\PYGZhy{}transitions/sim\PYGZus{}000200
shape\PYGZhy{}transitions/sim\PYGZus{}000300
shape\PYGZhy{}transitions/sim\PYGZus{}000400
shape\PYGZhy{}transitions/sim\PYGZus{}000500
shape\PYGZhy{}transitions/sim\PYGZus{}000600
shape\PYGZhy{}transitions/sim\PYGZus{}000700
shape\PYGZhy{}transitions/sim\PYGZus{}000800
shape\PYGZhy{}transitions/sim\PYGZus{}000900
\end{sphinxVerbatim}

Since this dataset does not contain any intermediate frames, it does not allow for supervised pretraining. This is because to pre\sphinxhyphen{}train a CFE network, two consecutive frames are required while to pretrain an \(\mathrm{OP}_n\) network, three frames with a distance of \(n/2\) are needed.

Instead, we create a second dataset which contains these intermediate frames. This does not need to be very close to the actual dataset since it’s only used for network initialization via pretraining. Here, we linearly move a rectangle around the domain.

\begin{sphinxVerbatim}[commandchars=\\\{\}]
\PYG{k}{for} \PYG{n}{scene} \PYG{o+ow}{in} \PYG{n}{Scene}\PYG{o}{.}\PYG{n}{list}\PYG{p}{(}\PYG{n}{pretrain\PYGZus{}data\PYGZus{}path}\PYG{p}{)}\PYG{p}{:} \PYG{n}{scene}\PYG{o}{.}\PYG{n}{remove}\PYG{p}{(}\PYG{p}{)}

\PYG{k}{for} \PYG{n}{scene\PYGZus{}index} \PYG{o+ow}{in} \PYG{n+nb}{range}\PYG{p}{(}\PYG{n}{example\PYGZus{}count} \PYG{o}{/}\PYG{o}{/} \PYG{n}{batch\PYGZus{}size}\PYG{p}{)}\PYG{p}{:}
    \PYG{n}{scene} \PYG{o}{=} \PYG{n}{Scene}\PYG{o}{.}\PYG{n}{create}\PYG{p}{(}\PYG{n}{pretrain\PYGZus{}data\PYGZus{}path}\PYG{p}{,} \PYG{n}{count}\PYG{o}{=}\PYG{n}{batch\PYGZus{}size}\PYG{p}{,} \PYG{n}{copy\PYGZus{}calling\PYGZus{}script}\PYG{o}{=}\PYG{k+kc}{False}\PYG{p}{)}
    \PYG{n+nb}{print}\PYG{p}{(}\PYG{n}{scene}\PYG{p}{)}
    \PYG{n}{pos0} \PYG{o}{=} \PYG{n}{np}\PYG{o}{.}\PYG{n}{random}\PYG{o}{.}\PYG{n}{randint}\PYG{p}{(}\PYG{l+m+mi}{10}\PYG{p}{,} \PYG{l+m+mi}{56}\PYG{p}{,} \PYG{p}{(}\PYG{n}{batch\PYGZus{}size}\PYG{p}{,} \PYG{l+m+mi}{2}\PYG{p}{)}\PYG{p}{)}  \PYG{c+c1}{\PYGZsh{} start position}
    \PYG{n}{pose} \PYG{o}{=} \PYG{n}{np}\PYG{o}{.}\PYG{n}{random}\PYG{o}{.}\PYG{n}{randint}\PYG{p}{(}\PYG{l+m+mi}{10}\PYG{p}{,} \PYG{l+m+mi}{56}\PYG{p}{,} \PYG{p}{(}\PYG{n}{batch\PYGZus{}size}\PYG{p}{,} \PYG{l+m+mi}{2}\PYG{p}{)}\PYG{p}{)}  \PYG{c+c1}{\PYGZsh{} end position}
    \PYG{n}{size} \PYG{o}{=} \PYG{n}{np}\PYG{o}{.}\PYG{n}{random}\PYG{o}{.}\PYG{n}{randint}\PYG{p}{(}\PYG{l+m+mi}{6}\PYG{p}{,}  \PYG{l+m+mi}{10}\PYG{p}{,}  \PYG{p}{(}\PYG{n}{batch\PYGZus{}size}\PYG{p}{,} \PYG{l+m+mi}{2}\PYG{p}{)}\PYG{p}{)}
    \PYG{k}{for} \PYG{n}{frame} \PYG{o+ow}{in} \PYG{n+nb}{range}\PYG{p}{(}\PYG{n}{step\PYGZus{}count}\PYG{o}{+}\PYG{l+m+mi}{1}\PYG{p}{)}\PYG{p}{:}
        \PYG{n}{time} \PYG{o}{=} \PYG{n}{frame} \PYG{o}{/} \PYG{n+nb}{float}\PYG{p}{(}\PYG{n}{step\PYGZus{}count} \PYG{o}{+} \PYG{l+m+mi}{1}\PYG{p}{)}
        \PYG{n}{pos} \PYG{o}{=} \PYG{n}{np}\PYG{o}{.}\PYG{n}{round}\PYG{p}{(}\PYG{n}{pos0} \PYG{o}{*} \PYG{p}{(}\PYG{l+m+mi}{1} \PYG{o}{\PYGZhy{}} \PYG{n}{time}\PYG{p}{)} \PYG{o}{+} \PYG{n}{pose} \PYG{o}{*} \PYG{n}{time}\PYG{p}{)}\PYG{o}{.}\PYG{n}{astype}\PYG{p}{(}\PYG{n}{np}\PYG{o}{.}\PYG{n}{int}\PYG{p}{)}
        \PYG{n}{density} \PYG{o}{=} \PYG{n}{AABox}\PYG{p}{(}\PYG{n}{lower}\PYG{o}{=}\PYG{n}{pos}\PYG{o}{\PYGZhy{}}\PYG{n}{size}\PYG{o}{/}\PYG{o}{/}\PYG{l+m+mi}{2}\PYG{p}{,} \PYG{n}{upper}\PYG{o}{=}\PYG{n}{pos}\PYG{o}{\PYGZhy{}}\PYG{n}{size}\PYG{o}{/}\PYG{o}{/}\PYG{l+m+mi}{2}\PYG{o}{+}\PYG{n}{size}\PYG{p}{)}\PYG{o}{.}\PYG{n}{value\PYGZus{}at}\PYG{p}{(}\PYG{n}{domain}\PYG{o}{.}\PYG{n}{center\PYGZus{}points}\PYG{p}{(}\PYG{p}{)}\PYG{p}{)}
        \PYG{n}{scene}\PYG{o}{.}\PYG{n}{write\PYGZus{}sim\PYGZus{}frame}\PYG{p}{(}\PYG{p}{[}\PYG{n}{density}\PYG{p}{]}\PYG{p}{,} \PYG{p}{[}\PYG{l+s+s1}{\PYGZsq{}}\PYG{l+s+s1}{density}\PYG{l+s+s1}{\PYGZsq{}}\PYG{p}{]}\PYG{p}{,} \PYG{n}{frame}\PYG{o}{=}\PYG{n}{frame}\PYG{p}{)}
\end{sphinxVerbatim}

\begin{sphinxVerbatim}[commandchars=\\\{\}]
moving\PYGZhy{}squares/sim\PYGZus{}000000
moving\PYGZhy{}squares/sim\PYGZus{}000100
moving\PYGZhy{}squares/sim\PYGZus{}000200
moving\PYGZhy{}squares/sim\PYGZus{}000300
moving\PYGZhy{}squares/sim\PYGZus{}000400
moving\PYGZhy{}squares/sim\PYGZus{}000500
moving\PYGZhy{}squares/sim\PYGZus{}000600
moving\PYGZhy{}squares/sim\PYGZus{}000700
moving\PYGZhy{}squares/sim\PYGZus{}000800
moving\PYGZhy{}squares/sim\PYGZus{}000900
\end{sphinxVerbatim}

\section{Supervised initialization}
\label{\detokenize{diffphys-control:supervised-initialization}}
First we define a split of the 1000 data samples into 100 test, 100 validation, and 800 training samples.

\begin{sphinxVerbatim}[commandchars=\\\{\}]
\PYG{n}{test\PYGZus{}range} \PYG{o}{=} \PYG{n+nb}{range}\PYG{p}{(}\PYG{l+m+mi}{100}\PYG{p}{)}
\PYG{n}{val\PYGZus{}range} \PYG{o}{=} \PYG{n+nb}{range}\PYG{p}{(}\PYG{l+m+mi}{100}\PYG{p}{,} \PYG{l+m+mi}{200}\PYG{p}{)}
\PYG{n}{train\PYGZus{}range} \PYG{o}{=} \PYG{n+nb}{range}\PYG{p}{(}\PYG{l+m+mi}{200}\PYG{p}{,} \PYG{l+m+mi}{1000}\PYG{p}{)}
\end{sphinxVerbatim}

The following cell trains all \(\mathrm{OP}_n \,\, \forall n\in\{2,4,8,16\}\). Here the \(n\) indicates the number of time steps for which the network predicts the target. In order to cover longer time horizons, we’re using factors of two here to hierarchically divide the time intervals during which the physical system should be controlled.

The \sphinxcode{\sphinxupquote{ControlTraining}} class is used to set up the corresponding optimization problem.
The loss for the supervised initialization is defined as the observation loss in terms of velocity at the center frame:
\begin{equation*}
\begin{split}
L_o^\textrm{sup} = \left| \mathrm{OP}\Big(d_{t_i},d_{t_j}\Big) - d^*_{(t_i+t_j)/2} \right|^2 .
\end{split}
\end{equation*}
Consequently, no sequence needs to be simulated (\sphinxcode{\sphinxupquote{sequence\_class=None}}) and an observation loss is required at frame \(\frac n 2\) (\sphinxcode{\sphinxupquote{obs\_loss\_frames={[}n // 2{]}}}).
The pretrained network checkpoints are stored in \sphinxcode{\sphinxupquote{supervised\_checkpoints}}.

\sphinxstyleemphasis{Note: The next cell will run for some time. The PDE\sphinxhyphen{}Control git repo comes with a set of pre\sphinxhyphen{}trained networks. So if you want to focus on the evaluation, you can skip the training and load the pretrained networks instead by commenting out the training cells, and uncommenting the cells for loading below.}

\begin{sphinxVerbatim}[commandchars=\\\{\}]
\PYG{n}{supervised\PYGZus{}checkpoints} \PYG{o}{=} \PYG{p}{\PYGZob{}}\PYG{p}{\PYGZcb{}}

\PYG{k}{for} \PYG{n}{n} \PYG{o+ow}{in} \PYG{p}{[}\PYG{l+m+mi}{2}\PYG{p}{,} \PYG{l+m+mi}{4}\PYG{p}{,} \PYG{l+m+mi}{8}\PYG{p}{,} \PYG{l+m+mi}{16}\PYG{p}{]}\PYG{p}{:}
    \PYG{n}{app} \PYG{o}{=} \PYG{n}{ControlTraining}\PYG{p}{(}\PYG{n}{n}\PYG{p}{,} \PYG{n}{IncompressibleFluidPDE}\PYG{p}{(}\PYG{n}{domain}\PYG{p}{,} \PYG{n}{dt}\PYG{p}{)}\PYG{p}{,}
                          \PYG{n}{datapath}\PYG{o}{=}\PYG{n}{pretrain\PYGZus{}data\PYGZus{}path}\PYG{p}{,} \PYG{n}{val\PYGZus{}range}\PYG{o}{=}\PYG{n}{val\PYGZus{}range}\PYG{p}{,} \PYG{n}{train\PYGZus{}range}\PYG{o}{=}\PYG{n}{train\PYGZus{}range}\PYG{p}{,} \PYG{n}{trace\PYGZus{}to\PYGZus{}channel}\PYG{o}{=}\PYG{k}{lambda} \PYG{n}{\PYGZus{}}\PYG{p}{:} \PYG{l+s+s1}{\PYGZsq{}}\PYG{l+s+s1}{density}\PYG{l+s+s1}{\PYGZsq{}}\PYG{p}{,}
                          \PYG{n}{obs\PYGZus{}loss\PYGZus{}frames}\PYG{o}{=}\PYG{p}{[}\PYG{n}{n}\PYG{o}{/}\PYG{o}{/}\PYG{l+m+mi}{2}\PYG{p}{]}\PYG{p}{,} \PYG{n}{trainable\PYGZus{}networks}\PYG{o}{=}\PYG{p}{[}\PYG{l+s+s1}{\PYGZsq{}}\PYG{l+s+s1}{OP}\PYG{l+s+si}{\PYGZpc{}d}\PYG{l+s+s1}{\PYGZsq{}} \PYG{o}{\PYGZpc{}} \PYG{n}{n}\PYG{p}{]}\PYG{p}{,}
                          \PYG{n}{sequence\PYGZus{}class}\PYG{o}{=}\PYG{k+kc}{None}\PYG{p}{)}\PYG{o}{.}\PYG{n}{prepare}\PYG{p}{(}\PYG{p}{)}
    \PYG{k}{for} \PYG{n}{i} \PYG{o+ow}{in} \PYG{n+nb}{range}\PYG{p}{(}\PYG{l+m+mi}{1000}\PYG{p}{)}\PYG{p}{:}
        \PYG{n}{app}\PYG{o}{.}\PYG{n}{progress}\PYG{p}{(}\PYG{p}{)}  \PYG{c+c1}{\PYGZsh{} Run Optimization for one batch}
    \PYG{n}{supervised\PYGZus{}checkpoints}\PYG{p}{[}\PYG{l+s+s1}{\PYGZsq{}}\PYG{l+s+s1}{OP}\PYG{l+s+si}{\PYGZpc{}d}\PYG{l+s+s1}{\PYGZsq{}} \PYG{o}{\PYGZpc{}} \PYG{n}{n}\PYG{p}{]} \PYG{o}{=} \PYG{n}{app}\PYG{o}{.}\PYG{n}{save\PYGZus{}model}\PYG{p}{(}\PYG{p}{)}
\end{sphinxVerbatim}

\begin{sphinxVerbatim}[commandchars=\\\{\}]
\PYG{n}{supervised\PYGZus{}checkpoints} \PYG{c+c1}{\PYGZsh{} this is where the checkpoints end up when re\PYGZhy{}training:}
\end{sphinxVerbatim}

\begin{sphinxVerbatim}[commandchars=\\\{\}]
\PYGZob{}\PYGZsq{}OP16\PYGZsq{}: \PYGZsq{}/root/phi/model/control\PYGZhy{}training/sim\PYGZus{}000003/checkpoint\PYGZus{}00001000\PYGZsq{},
 \PYGZsq{}OP2\PYGZsq{}: \PYGZsq{}/root/phi/model/control\PYGZhy{}training/sim\PYGZus{}000000/checkpoint\PYGZus{}00001000\PYGZsq{},
 \PYGZsq{}OP4\PYGZsq{}: \PYGZsq{}/root/phi/model/control\PYGZhy{}training/sim\PYGZus{}000001/checkpoint\PYGZus{}00001000\PYGZsq{},
 \PYGZsq{}OP8\PYGZsq{}: \PYGZsq{}/root/phi/model/control\PYGZhy{}training/sim\PYGZus{}000002/checkpoint\PYGZus{}00001000\PYGZsq{}\PYGZcb{}
\end{sphinxVerbatim}

\begin{sphinxVerbatim}[commandchars=\\\{\}]
\PYG{c+c1}{\PYGZsh{} supervised\PYGZus{}checkpoints = \PYGZob{}\PYGZsq{}OP\PYGZpc{}d\PYGZsq{} \PYGZpc{} n: \PYGZsq{}PDE\PYGZhy{}Control/networks/shapes/supervised/OP\PYGZpc{}d\PYGZus{}1000\PYGZsq{} \PYGZpc{} n for n in [2, 4, 8, 16]\PYGZcb{}}
\end{sphinxVerbatim}

This concludes the pretraining of the OP networks. These networks make it possible to at least perform a rough planning of the motions, which will be refined via end\sphinxhyphen{}to\sphinxhyphen{}end training below. However, beforehand we’ll initialize the \(\mathrm{CFE}\) networks such that we can perform \sphinxstyleemphasis{actions}, i.e., apply forces to the simulation. This is completely decoupled from the \(\mathrm{OP}\) networks.

\section{CFE pretraining with differentiable physics}
\label{\detokenize{diffphys-control:cfe-pretraining-with-differentiable-physics}}
To pretrain the \(\mathrm{CFE}\) networks, we set up a simulation with a single step of the differentiable solver.

The following cell trains the \(\mathrm{CFE}\) network from scratch. If you have a pretrained network at hand, you can skip the training and load the checkpoint by running the cell after.

\begin{sphinxVerbatim}[commandchars=\\\{\}]
\PYG{n}{app} \PYG{o}{=} \PYG{n}{ControlTraining}\PYG{p}{(}\PYG{l+m+mi}{1}\PYG{p}{,} \PYG{n}{IncompressibleFluidPDE}\PYG{p}{(}\PYG{n}{domain}\PYG{p}{,} \PYG{n}{dt}\PYG{p}{)}\PYG{p}{,}
                      \PYG{n}{datapath}\PYG{o}{=}\PYG{n}{pretrain\PYGZus{}data\PYGZus{}path}\PYG{p}{,} \PYG{n}{val\PYGZus{}range}\PYG{o}{=}\PYG{n}{val\PYGZus{}range}\PYG{p}{,} \PYG{n}{train\PYGZus{}range}\PYG{o}{=}\PYG{n}{train\PYGZus{}range}\PYG{p}{,} \PYG{n}{trace\PYGZus{}to\PYGZus{}channel}\PYG{o}{=}\PYG{k}{lambda} \PYG{n}{\PYGZus{}}\PYG{p}{:} \PYG{l+s+s1}{\PYGZsq{}}\PYG{l+s+s1}{density}\PYG{l+s+s1}{\PYGZsq{}}\PYG{p}{,}
                      \PYG{n}{obs\PYGZus{}loss\PYGZus{}frames}\PYG{o}{=}\PYG{p}{[}\PYG{l+m+mi}{1}\PYG{p}{]}\PYG{p}{,} \PYG{n}{trainable\PYGZus{}networks}\PYG{o}{=}\PYG{p}{[}\PYG{l+s+s1}{\PYGZsq{}}\PYG{l+s+s1}{CFE}\PYG{l+s+s1}{\PYGZsq{}}\PYG{p}{]}\PYG{p}{)}\PYG{o}{.}\PYG{n}{prepare}\PYG{p}{(}\PYG{p}{)}
\PYG{k}{for} \PYG{n}{i} \PYG{o+ow}{in} \PYG{n+nb}{range}\PYG{p}{(}\PYG{l+m+mi}{1000}\PYG{p}{)}\PYG{p}{:}
    \PYG{n}{app}\PYG{o}{.}\PYG{n}{progress}\PYG{p}{(}\PYG{p}{)}  \PYG{c+c1}{\PYGZsh{} Run Optimization for one batch}
\PYG{n}{supervised\PYGZus{}checkpoints}\PYG{p}{[}\PYG{l+s+s1}{\PYGZsq{}}\PYG{l+s+s1}{CFE}\PYG{l+s+s1}{\PYGZsq{}}\PYG{p}{]} \PYG{o}{=} \PYG{n}{app}\PYG{o}{.}\PYG{n}{save\PYGZus{}model}\PYG{p}{(}\PYG{p}{)}
\end{sphinxVerbatim}

\begin{sphinxVerbatim}[commandchars=\\\{\}]
\PYG{c+c1}{\PYGZsh{} supervised\PYGZus{}checkpoints[\PYGZsq{}CFE\PYGZsq{}] = \PYGZsq{}PDE\PYGZhy{}Control/networks/shapes/CFE/CFE\PYGZus{}2000\PYGZsq{}}
\end{sphinxVerbatim}

Note that we have not actually set up a simulation for the training, as the \(\mathrm{CFE}\) network only infers forces between pairs of states.

\begin{sphinxVerbatim}[commandchars=\\\{\}]
\PYG{c+c1}{\PYGZsh{} [TODO, show preview of CFE only?]}
\end{sphinxVerbatim}

\section{End\sphinxhyphen{}to\sphinxhyphen{}end training with differentiable physics}
\label{\detokenize{diffphys-control:end-to-end-training-with-differentiable-physics}}
Now that first versions of both network types exist, we can initiate the most important step of the setup at hand: the coupled end\sphinxhyphen{}to\sphinxhyphen{}end training of both networks via the differentiable fluid solver. While the pretraining stages relied on supervised training, the next step will yield a significantly improved quality for the control.

To initiate an end\sphinxhyphen{}to\sphinxhyphen{}end training of the \(\mathrm{CFE}\) and all \(\mathrm{OP}_n\) networks with the differentiable physics loss in phiflow, we create a new \sphinxcode{\sphinxupquote{ControlTraining}} instance with the staggered execution scheme.

The following cell builds the computational graph with \sphinxcode{\sphinxupquote{step\_count}} solver steps without initializing the network weights.

\begin{sphinxVerbatim}[commandchars=\\\{\}]
\PYG{n}{staggered\PYGZus{}app} \PYG{o}{=} \PYG{n}{ControlTraining}\PYG{p}{(}\PYG{n}{step\PYGZus{}count}\PYG{p}{,} \PYG{n}{IncompressibleFluidPDE}\PYG{p}{(}\PYG{n}{domain}\PYG{p}{,} \PYG{n}{dt}\PYG{p}{)}\PYG{p}{,}
                                \PYG{n}{datapath}\PYG{o}{=}\PYG{n}{data\PYGZus{}path}\PYG{p}{,} \PYG{n}{val\PYGZus{}range}\PYG{o}{=}\PYG{n}{val\PYGZus{}range}\PYG{p}{,} \PYG{n}{train\PYGZus{}range}\PYG{o}{=}\PYG{n}{train\PYGZus{}range}\PYG{p}{,} \PYG{n}{trace\PYGZus{}to\PYGZus{}channel}\PYG{o}{=}\PYG{k}{lambda} \PYG{n}{\PYGZus{}}\PYG{p}{:} \PYG{l+s+s1}{\PYGZsq{}}\PYG{l+s+s1}{density}\PYG{l+s+s1}{\PYGZsq{}}\PYG{p}{,}
                                \PYG{n}{obs\PYGZus{}loss\PYGZus{}frames}\PYG{o}{=}\PYG{p}{[}\PYG{n}{step\PYGZus{}count}\PYG{p}{]}\PYG{p}{,} \PYG{n}{trainable\PYGZus{}networks}\PYG{o}{=}\PYG{p}{[}\PYG{l+s+s1}{\PYGZsq{}}\PYG{l+s+s1}{CFE}\PYG{l+s+s1}{\PYGZsq{}}\PYG{p}{,} \PYG{l+s+s1}{\PYGZsq{}}\PYG{l+s+s1}{OP2}\PYG{l+s+s1}{\PYGZsq{}}\PYG{p}{,} \PYG{l+s+s1}{\PYGZsq{}}\PYG{l+s+s1}{OP4}\PYG{l+s+s1}{\PYGZsq{}}\PYG{p}{,} \PYG{l+s+s1}{\PYGZsq{}}\PYG{l+s+s1}{OP8}\PYG{l+s+s1}{\PYGZsq{}}\PYG{p}{,} \PYG{l+s+s1}{\PYGZsq{}}\PYG{l+s+s1}{OP16}\PYG{l+s+s1}{\PYGZsq{}}\PYG{p}{]}\PYG{p}{,}
                                \PYG{n}{sequence\PYGZus{}class}\PYG{o}{=}\PYG{n}{StaggeredSequence}\PYG{p}{,} \PYG{n}{learning\PYGZus{}rate}\PYG{o}{=}\PYG{l+m+mf}{5e\PYGZhy{}4}\PYG{p}{)}\PYG{o}{.}\PYG{n}{prepare}\PYG{p}{(}\PYG{p}{)}
\end{sphinxVerbatim}

\begin{sphinxVerbatim}[commandchars=\\\{\}]
App created. Scene directory is /root/phi/model/control\PYGZhy{}training/sim\PYGZus{}000005 (INFO), 2021\PYGZhy{}04\PYGZhy{}09 00:41:17,299n

Sequence class: \PYGZlt{}class \PYGZsq{}control.sequences.StaggeredSequence\PYGZsq{}\PYGZgt{} (INFO), 2021\PYGZhy{}04\PYGZhy{}09 00:41:17,305n

Partition length 16 sequence (from 0 to 16) at frame 8
Partition length 8 sequence (from 0 to 8) at frame 4
Partition length 4 sequence (from 0 to 4) at frame 2
Partition length 2 sequence (from 0 to 2) at frame 1
Execute \PYGZhy{}\PYGZgt{} 1
Execute \PYGZhy{}\PYGZgt{} 2
Partition length 2 sequence (from 2 to 4) at frame 3
Execute \PYGZhy{}\PYGZgt{} 3
Execute \PYGZhy{}\PYGZgt{} 4
Partition length 4 sequence (from 4 to 8) at frame 6
Partition length 2 sequence (from 4 to 6) at frame 5
Execute \PYGZhy{}\PYGZgt{} 5
Execute \PYGZhy{}\PYGZgt{} 6
Partition length 2 sequence (from 6 to 8) at frame 7
Execute \PYGZhy{}\PYGZgt{} 7
Execute \PYGZhy{}\PYGZgt{} 8
Partition length 8 sequence (from 8 to 16) at frame 12
Partition length 4 sequence (from 8 to 12) at frame 10
Partition length 2 sequence (from 8 to 10) at frame 9
Execute \PYGZhy{}\PYGZgt{} 9
Execute \PYGZhy{}\PYGZgt{} 10
Partition length 2 sequence (from 10 to 12) at frame 11
Execute \PYGZhy{}\PYGZgt{} 11
Execute \PYGZhy{}\PYGZgt{} 12
Partition length 4 sequence (from 12 to 16) at frame 14
Partition length 2 sequence (from 12 to 14) at frame 13
Execute \PYGZhy{}\PYGZgt{} 13
Execute \PYGZhy{}\PYGZgt{} 14
Partition length 2 sequence (from 14 to 16) at frame 15
Execute \PYGZhy{}\PYGZgt{} 15
Execute \PYGZhy{}\PYGZgt{} 16
Target loss: Tensor(\PYGZdq{}truediv\PYGZus{}16:0\PYGZdq{}, shape=(), dtype=float32) (INFO), 2021\PYGZhy{}04\PYGZhy{}09 00:41:44,654n

Force loss: Tensor(\PYGZdq{}truediv\PYGZus{}107:0\PYGZdq{}, shape=(), dtype=float32) (INFO), 2021\PYGZhy{}04\PYGZhy{}09 00:41:51,312n

Supervised loss at frame 16: Tensor(\PYGZdq{}truediv\PYGZus{}108:0\PYGZdq{}, shape=(), dtype=float32) (INFO), 2021\PYGZhy{}04\PYGZhy{}09 00:41:51,332n

Setting up loss (INFO), 2021\PYGZhy{}04\PYGZhy{}09 00:41:51,338n

Preparing data (INFO), 2021\PYGZhy{}04\PYGZhy{}09 00:42:32,417n

Initializing variables (INFO), 2021\PYGZhy{}04\PYGZhy{}09 00:42:32,443n

Model variables contain 0 total parameters. (INFO), 2021\PYGZhy{}04\PYGZhy{}09 00:42:36,418n

Validation (000000): Learning\PYGZus{}Rate: 0.0005, GT\PYGZus{}obs\PYGZus{}16: 399498.75, Loss\PYGZus{}reg\PYGZus{}unscaled: 1.2424506, Loss\PYGZus{}reg\PYGZus{}scale: 1.0, Loss: 798997.5 (INFO), 2021\PYGZhy{}04\PYGZhy{}09 00:42:59,618n
\end{sphinxVerbatim}

The next cell initializes the networks using the supervised checkpoints and then trains all networks jointly. You can increase the number of optimization steps or execute the next cell multiple times to further increase performance.

\sphinxstyleemphasis{Note: The next cell will run for some time. Optionally, you can skip this cell and load the pretrained networks instead with code in the cell below.}

\begin{sphinxVerbatim}[commandchars=\\\{\}]
\PYG{n}{staggered\PYGZus{}app}\PYG{o}{.}\PYG{n}{load\PYGZus{}checkpoints}\PYG{p}{(}\PYG{n}{supervised\PYGZus{}checkpoints}\PYG{p}{)}
\PYG{k}{for} \PYG{n}{i} \PYG{o+ow}{in} \PYG{n+nb}{range}\PYG{p}{(}\PYG{l+m+mi}{1000}\PYG{p}{)}\PYG{p}{:}
    \PYG{n}{staggered\PYGZus{}app}\PYG{o}{.}\PYG{n}{progress}\PYG{p}{(}\PYG{p}{)}  \PYG{c+c1}{\PYGZsh{} run staggered Optimization for one batch}
\PYG{n}{staggered\PYGZus{}checkpoint} \PYG{o}{=} \PYG{n}{staggered\PYGZus{}app}\PYG{o}{.}\PYG{n}{save\PYGZus{}model}\PYG{p}{(}\PYG{p}{)}
\end{sphinxVerbatim}

\begin{sphinxVerbatim}[commandchars=\\\{\}]
\PYG{c+c1}{\PYGZsh{} staggered\PYGZus{}checkpoint = \PYGZob{}net: \PYGZsq{}PDE\PYGZhy{}Control/networks/shapes/staggered/all\PYGZus{}53750\PYGZsq{} for net in [\PYGZsq{}CFE\PYGZsq{}, \PYGZsq{}OP2\PYGZsq{}, \PYGZsq{}OP4\PYGZsq{}, \PYGZsq{}OP8\PYGZsq{}, \PYGZsq{}OP16\PYGZsq{}]\PYGZcb{}}
\PYG{c+c1}{\PYGZsh{} staggered\PYGZus{}app.load\PYGZus{}checkpoints(staggered\PYGZus{}checkpoint)}
\end{sphinxVerbatim}

Now that the network is trained, we can infer some trajectories from the test set.
(This corresponds to Fig 5b and 18b of the \sphinxhref{https://openreview.net/pdf?id=HyeSin4FPB}{original paper}.)

The following cell takes the first one hundred configurations, i.e. our test set as defined by \sphinxcode{\sphinxupquote{test\_range}}, and let’s the network infer solutions for the corresponding inverse problems.

\begin{sphinxVerbatim}[commandchars=\\\{\}]
\PYG{n}{states} \PYG{o}{=} \PYG{n}{staggered\PYGZus{}app}\PYG{o}{.}\PYG{n}{infer\PYGZus{}all\PYGZus{}frames}\PYG{p}{(}\PYG{n}{test\PYGZus{}range}\PYG{p}{)}
\end{sphinxVerbatim}

Via the index list \sphinxcode{\sphinxupquote{batches}} below, you can choose to display some of the solutions. Each row shows a temporal sequence starting with the initial condition, and evolving the simulation with the NN control forces for 16 time steps. The last step, at \(t=16\) should match the target shown in the image on the far right.

\begin{sphinxVerbatim}[commandchars=\\\{\}]
\PYG{n}{batches} \PYG{o}{=} \PYG{p}{[}\PYG{l+m+mi}{0}\PYG{p}{,}\PYG{l+m+mi}{1}\PYG{p}{,}\PYG{l+m+mi}{2}\PYG{p}{]}

\PYG{n}{pylab}\PYG{o}{.}\PYG{n}{subplots}\PYG{p}{(}\PYG{n+nb}{len}\PYG{p}{(}\PYG{n}{batches}\PYG{p}{)}\PYG{p}{,} \PYG{l+m+mi}{10}\PYG{p}{,} \PYG{n}{sharey}\PYG{o}{=}\PYG{l+s+s1}{\PYGZsq{}}\PYG{l+s+s1}{row}\PYG{l+s+s1}{\PYGZsq{}}\PYG{p}{,} \PYG{n}{sharex}\PYG{o}{=}\PYG{l+s+s1}{\PYGZsq{}}\PYG{l+s+s1}{col}\PYG{l+s+s1}{\PYGZsq{}}\PYG{p}{,} \PYG{n}{figsize}\PYG{o}{=}\PYG{p}{(}\PYG{l+m+mi}{14}\PYG{p}{,} \PYG{l+m+mi}{6}\PYG{p}{)}\PYG{p}{)}
\PYG{n}{pylab}\PYG{o}{.}\PYG{n}{tight\PYGZus{}layout}\PYG{p}{(}\PYG{n}{w\PYGZus{}pad}\PYG{o}{=}\PYG{l+m+mi}{0}\PYG{p}{)}

\PYG{c+c1}{\PYGZsh{} solutions}
\PYG{k}{for} \PYG{n}{i}\PYG{p}{,} \PYG{n}{batch} \PYG{o+ow}{in} \PYG{n+nb}{enumerate}\PYG{p}{(}\PYG{n}{batches}\PYG{p}{)}\PYG{p}{:}
    \PYG{k}{for} \PYG{n}{t} \PYG{o+ow}{in} \PYG{n+nb}{range}\PYG{p}{(}\PYG{l+m+mi}{9}\PYG{p}{)}\PYG{p}{:}
        \PYG{n}{pylab}\PYG{o}{.}\PYG{n}{subplot}\PYG{p}{(}\PYG{n+nb}{len}\PYG{p}{(}\PYG{n}{batches}\PYG{p}{)}\PYG{p}{,} \PYG{l+m+mi}{10}\PYG{p}{,} \PYG{n}{t} \PYG{o}{+} \PYG{l+m+mi}{1} \PYG{o}{+} \PYG{n}{i} \PYG{o}{*} \PYG{l+m+mi}{10}\PYG{p}{)}
        \PYG{n}{pylab}\PYG{o}{.}\PYG{n}{title}\PYG{p}{(}\PYG{l+s+s1}{\PYGZsq{}}\PYG{l+s+s1}{t=}\PYG{l+s+si}{\PYGZpc{}d}\PYG{l+s+s1}{\PYGZsq{}} \PYG{o}{\PYGZpc{}} \PYG{p}{(}\PYG{n}{t} \PYG{o}{*} \PYG{l+m+mi}{2}\PYG{p}{)}\PYG{p}{)}
        \PYG{n}{pylab}\PYG{o}{.}\PYG{n}{imshow}\PYG{p}{(}\PYG{n}{states}\PYG{p}{[}\PYG{n}{t} \PYG{o}{*} \PYG{l+m+mi}{2}\PYG{p}{]}\PYG{o}{.}\PYG{n}{density}\PYG{o}{.}\PYG{n}{data}\PYG{p}{[}\PYG{n}{batch}\PYG{p}{,} \PYG{o}{.}\PYG{o}{.}\PYG{o}{.}\PYG{p}{,} \PYG{l+m+mi}{0}\PYG{p}{]}\PYG{p}{,} \PYG{n}{origin}\PYG{o}{=}\PYG{l+s+s1}{\PYGZsq{}}\PYG{l+s+s1}{lower}\PYG{l+s+s1}{\PYGZsq{}}\PYG{p}{)}

\PYG{c+c1}{\PYGZsh{} add targets}
\PYG{n}{testset} \PYG{o}{=} \PYG{n}{BatchReader}\PYG{p}{(}\PYG{n}{Dataset}\PYG{o}{.}\PYG{n}{load}\PYG{p}{(}\PYG{n}{staggered\PYGZus{}app}\PYG{o}{.}\PYG{n}{data\PYGZus{}path}\PYG{p}{,}\PYG{n}{test\PYGZus{}range}\PYG{p}{)}\PYG{p}{,} \PYG{n}{staggered\PYGZus{}app}\PYG{o}{.}\PYG{n}{\PYGZus{}channel\PYGZus{}struct}\PYG{p}{)}\PYG{p}{[}\PYG{n}{test\PYGZus{}range}\PYG{p}{]}
\PYG{k}{for} \PYG{n}{i}\PYG{p}{,} \PYG{n}{batch} \PYG{o+ow}{in} \PYG{n+nb}{enumerate}\PYG{p}{(}\PYG{n}{batches}\PYG{p}{)}\PYG{p}{:}
        \PYG{n}{pylab}\PYG{o}{.}\PYG{n}{subplot}\PYG{p}{(}\PYG{n+nb}{len}\PYG{p}{(}\PYG{n}{batches}\PYG{p}{)}\PYG{p}{,} \PYG{l+m+mi}{10}\PYG{p}{,} \PYG{n}{i} \PYG{o}{*} \PYG{l+m+mi}{10} \PYG{o}{+} \PYG{l+m+mi}{10}\PYG{p}{)}
        \PYG{n}{pylab}\PYG{o}{.}\PYG{n}{title}\PYG{p}{(}\PYG{l+s+s1}{\PYGZsq{}}\PYG{l+s+s1}{target}\PYG{l+s+s1}{\PYGZsq{}}\PYG{p}{)}
        \PYG{n}{pylab}\PYG{o}{.}\PYG{n}{imshow}\PYG{p}{(}\PYG{n}{testset}\PYG{p}{[}\PYG{l+m+mi}{1}\PYG{p}{]}\PYG{p}{[}\PYG{n}{i}\PYG{p}{,}\PYG{o}{.}\PYG{o}{.}\PYG{o}{.}\PYG{p}{,}\PYG{l+m+mi}{0}\PYG{p}{]}\PYG{p}{,} \PYG{n}{origin}\PYG{o}{=}\PYG{l+s+s1}{\PYGZsq{}}\PYG{l+s+s1}{lower}\PYG{l+s+s1}{\PYGZsq{}}\PYG{p}{)}
\end{sphinxVerbatim}

\noindent\sphinxincludegraphics{{diffphys-control_33_0}.png}

As you can see in the two right\sphinxhyphen{}most columns, the network does a very good job at solving these inverse problems: the fluid marker is pushed to the right spot and deformed in the right way to match the target.

What looks fairly simple here is actually a tricky task for a neural network:
it needs to guide a full 2D Navier\sphinxhyphen{}Stokes simulation over the course of 16 time integration steps. Hence, if the applied forces are slightly off or incoherent, the fluid can start swirling and moving chaotically. However, the network has learned to keep the motion together, and guide the  marker density to the target location.

Next, we quantify the achieved error rate by comparing the mean absolute error in terms of the final density configuration relative to the initial density. With the standard training setup above, the next cell should give a relative residual error of 5\sphinxhyphen{}6\%. Vice versa, this means that more than ca. 94\% of the marker density ends up in the right spot!

\begin{sphinxVerbatim}[commandchars=\\\{\}]
\PYG{n}{errors} \PYG{o}{=} \PYG{p}{[}\PYG{p}{]}
\PYG{k}{for} \PYG{n}{batch} \PYG{o+ow}{in} \PYG{n+nb}{enumerate}\PYG{p}{(}\PYG{n}{test\PYGZus{}range}\PYG{p}{)}\PYG{p}{:}
  \PYG{n}{initial} \PYG{o}{=} \PYG{n}{np}\PYG{o}{.}\PYG{n}{mean}\PYG{p}{(} \PYG{n}{np}\PYG{o}{.}\PYG{n}{abs}\PYG{p}{(} \PYG{n}{states}\PYG{p}{[}\PYG{l+m+mi}{0}\PYG{p}{]}\PYG{o}{.}\PYG{n}{density}\PYG{o}{.}\PYG{n}{data}\PYG{p}{[}\PYG{n}{batch}\PYG{p}{,} \PYG{o}{.}\PYG{o}{.}\PYG{o}{.}\PYG{p}{,} \PYG{l+m+mi}{0}\PYG{p}{]} \PYG{o}{\PYGZhy{}} \PYG{n}{testset}\PYG{p}{[}\PYG{l+m+mi}{1}\PYG{p}{]}\PYG{p}{[}\PYG{n}{batch}\PYG{p}{,}\PYG{o}{.}\PYG{o}{.}\PYG{o}{.}\PYG{p}{,}\PYG{l+m+mi}{0}\PYG{p}{]} \PYG{p}{)}\PYG{p}{)} 
  \PYG{n}{solution} \PYG{o}{=} \PYG{n}{np}\PYG{o}{.}\PYG{n}{mean}\PYG{p}{(} \PYG{n}{np}\PYG{o}{.}\PYG{n}{abs}\PYG{p}{(} \PYG{n}{states}\PYG{p}{[}\PYG{l+m+mi}{16}\PYG{p}{]}\PYG{o}{.}\PYG{n}{density}\PYG{o}{.}\PYG{n}{data}\PYG{p}{[}\PYG{n}{batch}\PYG{p}{,} \PYG{o}{.}\PYG{o}{.}\PYG{o}{.}\PYG{p}{,} \PYG{l+m+mi}{0}\PYG{p}{]} \PYG{o}{\PYGZhy{}} \PYG{n}{testset}\PYG{p}{[}\PYG{l+m+mi}{1}\PYG{p}{]}\PYG{p}{[}\PYG{n}{batch}\PYG{p}{,}\PYG{o}{.}\PYG{o}{.}\PYG{o}{.}\PYG{p}{,}\PYG{l+m+mi}{0}\PYG{p}{]} \PYG{p}{)}\PYG{p}{)} 
  \PYG{n}{errors}\PYG{o}{.}\PYG{n}{append}\PYG{p}{(} \PYG{n}{solution}\PYG{o}{/}\PYG{n}{initial} \PYG{p}{)}
\PYG{n+nb}{print}\PYG{p}{(}\PYG{l+s+s2}{\PYGZdq{}}\PYG{l+s+s2}{Relative MAE: }\PYG{l+s+s2}{\PYGZdq{}}\PYG{o}{+}\PYG{n+nb}{format}\PYG{p}{(}\PYG{n}{np}\PYG{o}{.}\PYG{n}{mean}\PYG{p}{(}\PYG{n}{errors}\PYG{p}{)}\PYG{p}{)}\PYG{p}{)}
\end{sphinxVerbatim}

\begin{sphinxVerbatim}[commandchars=\\\{\}]
Relative MAE: 0.05450168251991272
\end{sphinxVerbatim}

\section{Next steps}
\label{\detokenize{diffphys-control:next-steps}}
For further experiments with this source code, you can, e.g.:
\begin{itemize}
\item {} 
Change the \sphinxcode{\sphinxupquote{test\_range}} indices to look at different examples, or test the generalization of the trained controller networks by using new shapes as targets.

\item {} 
Try using a \sphinxcode{\sphinxupquote{RefinedSequence}} (instead of a \sphinxcode{\sphinxupquote{StaggeredSequence}}) to train with the prediction refinement scheme. This will yield a further improved control and reduced density error.

\end{itemize}

\chapter{Summary and Discussion}
\label{\detokenize{diffphys-outlook:summary-and-discussion}}\label{\detokenize{diffphys-outlook::doc}}
The previous sections have explained the \sphinxstyleemphasis{differentiable physics} approach for deep learning, and have given a range of examples: from a very basic gradient calculation, all the way to complex learning setups powered by advanced simulations. This is a good time to take a step back and evaluate: in the end, the differentiable physics components of these approaches are not too complicated. They are largely based on existing numerical methods, with a focus on efficiently using those methods not only to do a forward simulation, but also to compute gradient information. What is primarily exciting in this context are the implications that arise from the combination of these numerical methods with deep learning.

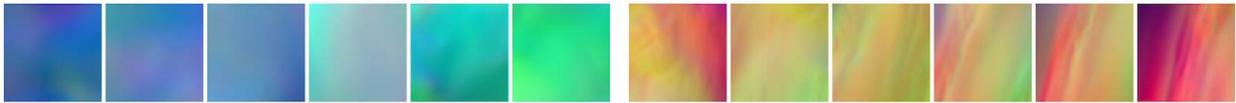
\sphinxincludegraphics{{divider6}.jpg}

\section{Integration}
\label{\detokenize{diffphys-outlook:integration}}
Most importantly, training via differentiable physics allows us to seamlessly bring the two fields together:
we can obtain \sphinxstyleemphasis{hybrid} methods, that use the best numerical methods that we have at our disposal for the simulation itself, as well as for the training process. We can then use the trained model to improve forward or backward solves. Thus, in the end, we have a solver that combines a \sphinxstyleemphasis{traditional} solver and a \sphinxstyleemphasis{learned} component that in combination can improve the capabilities of numerical methods.

\section{Interaction}
\label{\detokenize{diffphys-outlook:interaction}}
One key aspect that is important for these hybrids to work well is to let the NN \sphinxstyleemphasis{interact} with the PDE solver at training time. Differentiable simulations allow a trained model to “explore and experience” the physical environment, and receive directed feedback regarding its interactions throughout the solver iterations. This combination nicely fits into the broader context of machine learning as \sphinxstyleemphasis{differentiable programming}.

\section{Generalization}
\label{\detokenize{diffphys-outlook:generalization}}
The hybrid approach also bears particular promise for simulators: it improves generalizing capabilities of the trained models by letting the PDE\sphinxhyphen{}solver handle large\sphinxhyphen{}scale \sphinxstyleemphasis{changes to the data distribution} such that the learned model can focus on localized structures not captured by the discretization. While physical models generalize very well, learned models often specialize in data distributions seen at training time. This was, e.g., shown for the models reducing numerical errors of the previous chapter: the trained models can deal with solution manifolds with significant amounts of varying physical behavior, while simpler training variants quickly deteriorate over the course of recurrent time steps.

\sphinxincludegraphics{{divider7}.jpg}

Training NNs via differentiable physics solvers is a very generic approach that is applicable to a wide range of combinations of PDE\sphinxhyphen{}based models and deep learning. Nonetheless, the next chapters will discuss several variants that are orthogonal to the general DP version, or can yield benefits in more specialized settings.

\part{Reinforcement Learning}

\chapter{Introduction to Reinforcement Learning}
\label{\detokenize{reinflearn-intro:introduction-to-reinforcement-learning}}\label{\detokenize{reinflearn-intro::doc}}
Deep reinforcement learning, referred to as just \sphinxstyleemphasis{reinforcement learning} (RL) from now on, is a class of methods in the larger field of deep learning that lets an artificial intelligence agent explore the interactions with a surrounding environment. While doing this, the agent receives reward signals for its actions and tries to discern which actions contribute to higher rewards, to adapt its behavior accordingly. RL has been very successful at playing games such as Go {[}\hyperlink{cite.references:id90}{SSS+17}{]}, and it bears promise for engineering applications such as robotics.

The setup for RL generally consists of two parts: the environment and the agent. The environment receives actions \(a\) from the agent while supplying it with observations in the form of states \(s\), and rewards \(r\). The observations represent the fraction of the information from the respective environment state that the agent is able to perceive. The rewards are given by a predefined function, usually tailored to the environment and might contain, e.g., a game score, a penalty for wrong actions or a bounty for successfully finished tasks.

\begin{figure}[htbp]
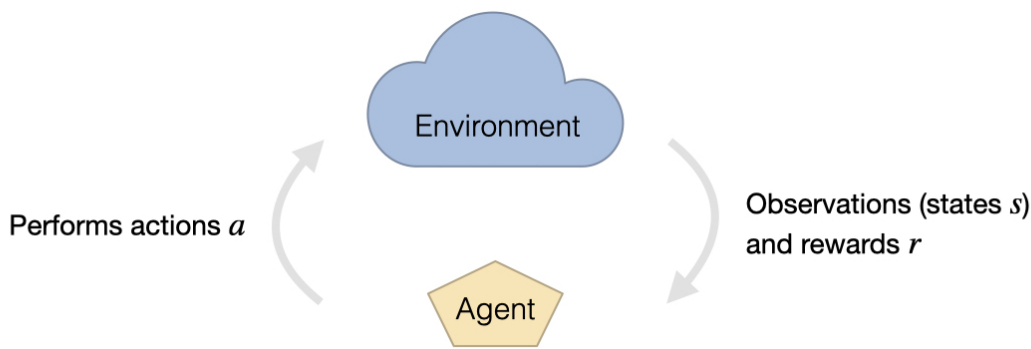

\centering
\capstart

\noindent\sphinxincludegraphics[height=200\sphinxpxdimen]{{rl-overview}.jpg}
\caption{Reinforcement learning is formulated in terms of an environment that gives observations in the form of states and rewards to an agent. The agent interacts with the environment by performing actions.}\label{\detokenize{reinflearn-intro:rl-overview}}\end{figure}

In its simplest form, the learning goal for reinforcement learning tasks can be formulated as
\begin{equation}\label{equation:reinflearn-intro:rl-learn-l2}
\begin{split}
\text{arg max}_{\theta} \mathbb{E}_{a \sim \pi(;s,\theta_p)} \big[ \sum_t r_t \big], 
\end{split}
\end{equation}
where the reward at time \(t\) (denoted by \(r_t\) above) is the result of an action \(a\) performed by an agent.
The agents choose their actions based on a neural network policy which decides via a set of given observations.
The policy \(\pi(a;s, \theta)\) returns the probability for the action, and is conditioned on the state \(s\) of the environment and the weights \(\theta\).

During the learning process the central aim of RL is to uses the combined information of state, action and corresponding rewards to increase the cumulative intensity of reward signals over each trajectory. To achieve this goal, multiple algorithms have been proposed, which can be roughly divided into two larger classes: \sphinxstyleemphasis{policy gradient} and \sphinxstyleemphasis{value\sphinxhyphen{}based} methods {[}\hyperlink{cite.references:id91}{SB18}{]}.

\section{Algorithms}
\label{\detokenize{reinflearn-intro:algorithms}}
In vanilla policy gradient methods, the trained neural networks directly select actions \(a\) from environment observations. In the learning process, an NN is trained to infer the probability of actions. Here, probabilities for actions leading to higher rewards in the rest of the respective trajectories are increased, while actions with smaller return are made less likely.

Value\sphinxhyphen{}based methods, such as \sphinxstyleemphasis{Q\sphinxhyphen{}Learning}, on the other hand work by optimizing a state\sphinxhyphen{}action value function, the so\sphinxhyphen{}called \sphinxstyleemphasis{Q\sphinxhyphen{}Function}. The network in this case receives state \(s\) and action \(a\) to predict the average cumulative reward resulting from this input for the remainder of the trajectory, i.e. \(Q(s,a)\). Actions are then chosen to maximize \(Q\) given the state.

In addition, \sphinxstyleemphasis{actor\sphinxhyphen{}critic} methods combine elements from both approaches. Here, the actions generated by a policy network are rated based on a corresponding change in state potential. These values are given by another neural network and approximate the expected cumulative reward from the given state. \sphinxstyleemphasis{Proximal policy optimization} (PPO) {[}\hyperlink{cite.references:id92}{SWD+17}{]} is one example from this class of algorithms and is our choice for the example task of this chapter, which is controlling Burgers’ equation as a physical environment.

\section{Proximal policy optimization}
\label{\detokenize{reinflearn-intro:proximal-policy-optimization}}
As PPO methods are an actor\sphinxhyphen{}critic approach, we need to train two interdependent networks: the actor, and the critic.
The objective of the actor inherently depends on the output of the critic network (it provides feedback which actions are worth performing), and likewise the critic depends on the actions generated by the actor network (this determines which states to explore).

This interdependence can promote instabilities, e.g., as strongly over\sphinxhyphen{} or underestimated state values can give wrong impulses during learning. Actions yielding higher rewards often also contribute to reaching states with higher informational value. As a consequence, when the \sphinxhyphen{} possibly incorrect \sphinxhyphen{} value estimate of individual samples are allowed to unrestrictedly affect the agent’s behavior, the learning progress can collapse.

DEBUG TEST t’s t’s agent’s vs’s TODO remove!!!

PPO was introduced as a method to specifically counteract this problem. The idea is to restrict the influence that individual state value estimates can have on the change of the actor’s behavior during learning. PPO is a popular choice especially when working on continuous action spaces. This can be attributed to the fact that it tends to achieve good results with a stable learning progress, while still being comparatively easy to implement.

\subsection{PPO\sphinxhyphen{}clip}
\label{\detokenize{reinflearn-intro:ppo-clip}}
More specifically, we will use the algorithm \sphinxstyleemphasis{PPO\sphinxhyphen{}clip} {[}\hyperlink{cite.references:id92}{SWD+17}{]}. This PPO variant sets a hard limit for the change in behavior caused by singular update steps. As such, the algorithm uses a previous network state (denoted by a subscript \(_p\) below) to limit the change per step of the learning process.
In the following, we will denote the network parameters of the actor network as \(\theta\) and those of the critic as \(\phi\).

\subsection{Actor}
\label{\detokenize{reinflearn-intro:actor}}
The actor computes a policy function returning the probability distribution for the actions conditioned by the current network parameters \(\theta\) and a state \(s\).
In the following we’ll denote the probability of choosing a specific action \(a\) from the distribution with \(\pi(a; s,\theta)\).
As mentioned above, the training procedure computes a certain number of weight updates using policy evaluations with a fixed previous network state \(\pi(a;s, \theta_p)\), and in intervals re\sphinxhyphen{}initializes the previous weights \(\theta_p\) from \(\theta\).
To limit the changes, the objective function makes use of a \(\text{clip}(a,b,c)\) function, which simply returns \(a\) clamped to the interval \([b,c]\).

\(\epsilon\) defines the bound for the deviation from the previous policy.
In combination, the objective for the actor is given by the following expression:
\begin{equation*}
\begin{split}\begin{aligned}
\text{arg max}_{\theta} \mathbb{E}_{a \sim \pi(;s,\theta_p)} \Big[ \text{min} \big(
    \frac{\pi(a;s,\theta)}{\pi(a;s,\theta_p)} 
        A(s, a; \phi), 
    \text{clip}(\frac{\pi(a;s,\theta)}{\pi(a;s,\theta_p)}, 1-\epsilon, 1+\epsilon)
        A(s, a; \phi) 
    \big) \Big]
\end{aligned}\end{split}
\end{equation*}
As the actor network is trained to provide the expected value, at training time an
additional standard deviation is used to sample values from a Gaussian distribution around this mean.
It is decreased over the course of the training, and at inference time we only evaluate the mean (i.e. a distribution with variance 0).

\subsection{Critic and advantage}
\label{\detokenize{reinflearn-intro:critic-and-advantage}}
The critic is represented by a value function \(V(s; \phi)\) that predicts the expected cumulative reward to be received from state \(s\).
Its objective is to minimize the squared advantage \(A\):
\begin{equation*}
\begin{split}\begin{aligned}
\text{arg min}_{\phi}\mathbb{E}_{a \sim \pi(;s,\theta_p)}[A(s, a; \phi)^2] \ , 
\end{aligned}\end{split}
\end{equation*}
where the advantage function \(A(s, a; \phi)\) builds upon \(V\): it’s goal is to evaluate the deviation from an average cumulative reward. I.e., we’re interested in estimating how much the decision made via \(\pi(;s,\theta_p)\) improves upon making random decisions (again, evaluated via the unchanging, previous network state \(\theta_p\)). We use the so\sphinxhyphen{}called Generalized Advantage Estimation (GAE) {[}\hyperlink{cite.references:id93}{SML+15}{]} to compute \(A\) as:
\begin{equation*}
\begin{split}\begin{aligned}
A(s_t, a_t; \phi) &= \sum\limits_{i=0}^{n-t-1}(\gamma\lambda)^i\delta_{t+i} \\
\delta_t &= r_t+\gamma \big( V(s_{t+1} ; \phi) - V(s_t ; \phi) \big)
\end{aligned}\end{split}
\end{equation*}
Here \(r_t\) describes the reward obtained in time step \(t\), while \(n\) denotes the total length of the trajectory. \(\gamma\) and \(\lambda\) are two hyperparameters which influence rewards and state value predictions from the distant futures have on the advantage calculation. They are typically set to values smaller than one.

The \(\delta_t\) in the formulation above represent a biased approximation of the true advantage. Hence
the GAE can be understood as a discounted cumulative sum of these estimates, from the current time step until the end of the trajectory.

\bigskip\hrule\bigskip

\section{Application to inverse problems}
\label{\detokenize{reinflearn-intro:application-to-inverse-problems}}
Reinforcement learning is widely used for trajectory optimization with multiple decision problems building upon one another. However, in the context of physical systems and PDEs, reinforcement learning algorithms are likewise attractive. In this setting, they can operate in a fashion that’s similar to supervised single shooting approaches by generating full trajectories and learning by comparing the final approximation to the target.

Still, the approaches differ in terms of how this optimization is performed. For example, reinforcement learning algorithms like PPO try to explore the action space during training by adding a random offset to the actions selected by the actor. This way, the algorithm can discover new behavioral patterns that are more refined than the previous ones.

The way how long term effects of generated forces are taken into account can also differ for physical systems. In a control force estimator setup with differentiable physics (DP) loss, as discussed e.g. in {\hyperref[\detokenize{diffphys-code-burgers::doc}]{\sphinxcrossref{\DUrole{doc}{Burgers Optimization with a Differentiable Physics Gradient}}}}, these dependencies are handled by passing the loss gradient through the simulation step back into previous time steps. Contrary to that, reinforcement learning usually treats the environment as a black box without gradient information. When using PPO, the value estimator network is instead used to track the long term dependencies by predicting the influence any action has for the future system evolution.

Working with Burgers’ equation as physical environment, the trajectory generation process can be summarized as follows. It shows how the simulation steps of the environment and the neural network evaluations of the agent are interleaved:
\begin{equation*}
\begin{split}
\mathbf{u}_{t+1}=\mathcal{P}(\mathbf{u}_t+\pi(\mathbf{u}_t; \mathbf{u}^*, t, \theta)\Delta t)
\end{split}
\end{equation*}
The \(*\) superscript (as usual) denotes a reference or target quantity, and hence here \(\mathbf{u}^*\) denotes a velocity target. For the continuous action space of the PDE, \(\pi\) directly computes an action in terms of a force, rather than probabilities for a discrete set of different actions.

The reward is calculated in a similar fashion as the Loss in the DP approach: it consists of two parts, one of which amounts to the negative square norm of the applied forces and is given at every time step. The other part adds a punishment proportional to the \(L^2\) distance between the final approximation and the target state at the end of each trajectory.
\begin{equation*}
\begin{split}\begin{aligned}
r_t   &=  r_t^f+r_t^o \\
r_t^f &= -\left\lVert{\mathbf{a}_t}\right\rVert_2^2 \\
r_t^o &=
\begin{cases}
    -\left\lVert{\mathbf{u}^*-\mathbf{u}_t}\right\rVert_2^2,&\text{if } t = n-1\\
    0,&\text{otherwise}
\end{cases}
\end{aligned}\end{split}
\end{equation*}

\section{Implementation}
\label{\detokenize{reinflearn-intro:implementation}}
In the following, we’ll describe a way to implement a PPO\sphinxhyphen{}based RL training for physical systems. This implementation is also the basis for the notebook of the next section, i.e., {\hyperref[\detokenize{reinflearn-code::doc}]{\sphinxcrossref{\DUrole{doc}{Controlling Burgers’ Equation with Reinforcement Learning}}}}. While this notebook provides a practical example, and an evaluation in comparison to DP training, we’ll first give a more generic overview below.

To train a reinforcement learning agent to control a PDE\sphinxhyphen{}governed system, the physical model has to be formalized as an RL environment. The \sphinxhref{https://github.com/DLR-RM/stable-baselines3}{stable\sphinxhyphen{}baselines3} framework, which we use in the following to implement a PPO training, uses a vectorized version of the \sphinxhref{https://github.com/openai/gym}{OpenAI gym environment}. This way, rollout collection can be performed on multiple trajectories in parallel for better resource utilization and wall time efficiency.

Vectorized environments require a definition of observation and action spaces, meaning the in\sphinxhyphen{} and output spaces of the agent policy. In our case, the former consists of the current physical states and the goal states, e.g., velocity fields, stacked along their channel dimension. Another channel is added for the elapsed time since the start of the simulation divided by the total trajectory length. The action space (the output) encompasses one force value for each cell of the velocity field.

The most relevant methods of vectorized environments are \sphinxcode{\sphinxupquote{reset}}, \sphinxcode{\sphinxupquote{step\_async}}, \sphinxcode{\sphinxupquote{step\_wait}} and \sphinxcode{\sphinxupquote{render}}. The first of these is used to start a new trajectory by computing initial and goal states and returning the first observation for each vectorized instance. As these instances in other applications are not bound to finish trajectories synchronously, \sphinxcode{\sphinxupquote{reset}} has to be called from within the environment itself when entering a terminal state. \sphinxcode{\sphinxupquote{step\_async}} and \sphinxcode{\sphinxupquote{step\_wait}} are the two main parts of the \sphinxcode{\sphinxupquote{step}} method, which takes actions, applies them to the velocity fields and performs one iteration of the physics models. The split into async and wait enables supporting vectorized environments that run each instance on separate threads. However, this is not required in our approach, as phiflow handles the simulation of  batches internally. The \sphinxcode{\sphinxupquote{render}} method is called to display training results, showing reconstructed trajectories in real time or rendering them to files.

Because of the strongly differing output spaces of the actor and critic networks, we use different architectures for each of them. The network yielding the actions uses a variant of the network architecture from Holl et al. {[}\hyperlink{cite.references:id11}{HKT19}{]}, in line with the \(CFE\) function performing actions there. The other network consists of a series of convolutions with kernel size 3, each followed by a max pooling layer with kernel size 2 and stride 2. After the feature maps have been downsampled to one value in this way, a final fully connected layer merges all channels, generating the predicted state value.

In the example implementation of the next chapter, the \sphinxcode{\sphinxupquote{BurgersTraining}} class manages all aspects of this training internally, including the setup of agent and environment and storing trained models and monitor logs to disk. It also includes a variant of the Burgers’ equation environment described above, which, instead of computing random trajectories, uses data from a predefined set. During training, the agent is evaluated in this environment in regular intervals to be able to compare the training progress to the DP method more accurately.

The next chapter will use this \sphinxcode{\sphinxupquote{BurgersTraining}} class to run a full PPO scenario, evaluate its performance, and compare it to an approach that uses more domain knowledge of the physical system, i.e., the gradient\sphinxhyphen{}based control training with a DP approach.

\chapter{Controlling Burgers’ Equation with Reinforcement Learning}
\label{\detokenize{reinflearn-code:controlling-burgers-equation-with-reinforcement-learning}}\label{\detokenize{reinflearn-code::doc}}
In the following, we will target inverse problems with Burgers equation as a testbed for reinforcement learning (RL). The setup is similar to the inverse problems previously targeted with differentiable physics (DP) training (cf. {\hyperref[\detokenize{diffphys-control::doc}]{\sphinxcrossref{\DUrole{doc}{Solving Inverse Problems with NNs}}}}), and hence we’ll also directly compare to these approaches below. Similar to before, Burgers equation is simple but non\sphinxhyphen{}linear with interesting dynamics, and hence a good starting point for RL experiments. In the following, the goal is to train a control force estimator network that should predict the forces needed to generate smooth transitions between two given states.
\sphinxhref{https://colab.research.google.com/github/tum-pbs/pbdl-book/blob/main/reinflearn-code.ipynb}{{[}run in colab{]}}

\section{Overview}
\label{\detokenize{reinflearn-code:overview}}
Reinforcement learning describes an agent perceiving an environment and taking actions inside it. It aims at maximizing an accumulated sum of rewards, which it receives for those actions by the environment. Thus, the agent learns empirically which actions to take in different situations. \sphinxstyleemphasis{Proximal policy optimization} \sphinxhref{https://arxiv.org/abs/1707.06347v2}{(PPO)} is a widely used reinforcement learning algorithm describing two neural networks: a policy NN selecting actions for given observations and a value estimator network rating the reward potential of those states. These value estimates form the loss of the policy network, given by the change in reward potential by the chosen action.

This notebook illustrates how PPO reinforcement learning can be applied to the described control problem of Burgers’ equation. In comparison to the DP approach, the RL method does not have access to a differentiable physics solver, it is \sphinxstyleemphasis{model\sphinxhyphen{}free}.

However, the goal of the value estimator NN is to compensate for this lack of a solver, as it tries to capture the long term effect of individual actions. Thus, an interesting question the following code example should answer is: can the model\sphinxhyphen{}free PPO reinforcement learning match the performance of the model\sphinxhyphen{}based DP training. We will compare this in terms of learning speed and the amount of required forces.

\section{Software installation}
\label{\detokenize{reinflearn-code:software-installation}}
This example uses the reinforcement learning framework \sphinxhref{https://github.com/DLR-RM/stable-baselines3}{stable\_baselines3} with \sphinxhref{https://arxiv.org/abs/1707.06347v2}{PPO} as reinforcement learning algorithm.
For the physical simulation, version 1.5.1 of the differentiable PDE solver \sphinxhref{https://github.com/tum-pbs/PhiFlow}{ΦFlow} is used.

After the RL training is completed, we’ll additionally compare it to a differentiable physics approach using a “control force estimator” (CFE) network from {\hyperref[\detokenize{diffphys-control::doc}]{\sphinxcrossref{\DUrole{doc}{Solving Inverse Problems with NNs}}}} (as introduced by {[}\hyperlink{cite.references:id11}{HKT19}{]}).

\begin{sphinxVerbatim}[commandchars=\\\{\}]
\PYG{o}{!}pip install stable\PYGZhy{}baselines3\PYG{o}{=}\PYG{o}{=}\PYG{l+m}{1}.1 \PYG{n+nv}{phiflow}\PYG{o}{=}\PYG{o}{=}\PYG{l+m}{1}.5.1
\PYG{o}{!}git clone https://github.com/Sh0cktr4p/PDE\PYGZhy{}Control\PYGZhy{}RL.git
\PYG{o}{!}git clone https://github.com/holl\PYGZhy{}/PDE\PYGZhy{}Control.git
\end{sphinxVerbatim}

Now we can import the necessary modules. Due to the scope of this example, there are quite a few modules to load.

\begin{sphinxVerbatim}[commandchars=\\\{\}]
\PYG{k+kn}{import} \PYG{n+nn}{sys}\PYG{p}{;} \PYG{n}{sys}\PYG{o}{.}\PYG{n}{path}\PYG{o}{.}\PYG{n}{append}\PYG{p}{(}\PYG{l+s+s1}{\PYGZsq{}}\PYG{l+s+s1}{PDE\PYGZhy{}Control/src}\PYG{l+s+s1}{\PYGZsq{}}\PYG{p}{)}\PYG{p}{;} \PYG{n}{sys}\PYG{o}{.}\PYG{n}{path}\PYG{o}{.}\PYG{n}{append}\PYG{p}{(}\PYG{l+s+s1}{\PYGZsq{}}\PYG{l+s+s1}{PDE\PYGZhy{}Control\PYGZhy{}RL/src}\PYG{l+s+s1}{\PYGZsq{}}\PYG{p}{)}
\PYG{k+kn}{import} \PYG{n+nn}{time}\PYG{o}{,} \PYG{n+nn}{csv}\PYG{o}{,} \PYG{n+nn}{os}\PYG{o}{,} \PYG{n+nn}{shutil}
\PYG{k+kn}{from} \PYG{n+nn}{tensorboard}\PYG{n+nn}{.}\PYG{n+nn}{backend}\PYG{n+nn}{.}\PYG{n+nn}{event\PYGZus{}processing}\PYG{n+nn}{.}\PYG{n+nn}{event\PYGZus{}accumulator} \PYG{k+kn}{import} \PYG{n}{EventAccumulator}
\PYG{k+kn}{from} \PYG{n+nn}{phi}\PYG{n+nn}{.}\PYG{n+nn}{flow} \PYG{k+kn}{import} \PYG{o}{*}
\PYG{k+kn}{import} \PYG{n+nn}{burgers\PYGZus{}plots} \PYG{k}{as} \PYG{n+nn}{bplt}
\PYG{k+kn}{import} \PYG{n+nn}{matplotlib}\PYG{n+nn}{.}\PYG{n+nn}{pyplot} \PYG{k}{as} \PYG{n+nn}{plt}
\PYG{k+kn}{from} \PYG{n+nn}{envs}\PYG{n+nn}{.}\PYG{n+nn}{burgers\PYGZus{}util} \PYG{k+kn}{import} \PYG{n}{GaussianClash}\PYG{p}{,} \PYG{n}{GaussianForce}
\end{sphinxVerbatim}

\begin{sphinxVerbatim}[commandchars=\\\{\}]

\end{sphinxVerbatim}

\section{Data generation}
\label{\detokenize{reinflearn-code:data-generation}}
First we generate a dataset which we will use to train the differentiable physics model on. We’ll also use it to evaluate the performance of both approaches during and after training. The code below simulates 1000 cases (i.e. phiflow “scenes”), and keeps 100 of them as validation and test cases, respectively. The remaining 800 are used for training.

\begin{sphinxVerbatim}[commandchars=\\\{\}]
\PYG{n}{DOMAIN} \PYG{o}{=} \PYG{n}{Domain}\PYG{p}{(}\PYG{p}{[}\PYG{l+m+mi}{32}\PYG{p}{]}\PYG{p}{,} \PYG{n}{box}\PYG{o}{=}\PYG{n}{box}\PYG{p}{[}\PYG{l+m+mi}{0}\PYG{p}{:}\PYG{l+m+mi}{1}\PYG{p}{]}\PYG{p}{)}     \PYG{c+c1}{\PYGZsh{} Size and shape of the fields}
\PYG{n}{VISCOSITY} \PYG{o}{=} \PYG{l+m+mf}{0.003}
\PYG{n}{STEP\PYGZus{}COUNT} \PYG{o}{=} \PYG{l+m+mi}{32}                         \PYG{c+c1}{\PYGZsh{} Trajectory length}
\PYG{n}{DT} \PYG{o}{=} \PYG{l+m+mf}{0.03}
\PYG{n}{DIFFUSION\PYGZus{}SUBSTEPS} \PYG{o}{=} \PYG{l+m+mi}{1}

\PYG{n}{DATA\PYGZus{}PATH} \PYG{o}{=} \PYG{l+s+s1}{\PYGZsq{}}\PYG{l+s+s1}{forced\PYGZhy{}burgers\PYGZhy{}clash}\PYG{l+s+s1}{\PYGZsq{}}
\PYG{n}{SCENE\PYGZus{}COUNT} \PYG{o}{=} \PYG{l+m+mi}{1000}
\PYG{n}{BATCH\PYGZus{}SIZE} \PYG{o}{=} \PYG{l+m+mi}{100}

\PYG{n}{TRAIN\PYGZus{}RANGE} \PYG{o}{=} \PYG{n+nb}{range}\PYG{p}{(}\PYG{l+m+mi}{200}\PYG{p}{,} \PYG{l+m+mi}{1000}\PYG{p}{)}
\PYG{n}{VAL\PYGZus{}RANGE} \PYG{o}{=} \PYG{n+nb}{range}\PYG{p}{(}\PYG{l+m+mi}{100}\PYG{p}{,} \PYG{l+m+mi}{200}\PYG{p}{)}
\PYG{n}{TEST\PYGZus{}RANGE} \PYG{o}{=} \PYG{n+nb}{range}\PYG{p}{(}\PYG{l+m+mi}{0}\PYG{p}{,} \PYG{l+m+mi}{100}\PYG{p}{)}
\end{sphinxVerbatim}

\begin{sphinxVerbatim}[commandchars=\\\{\}]
\PYG{k}{for} \PYG{n}{batch\PYGZus{}index} \PYG{o+ow}{in} \PYG{n+nb}{range}\PYG{p}{(}\PYG{n}{SCENE\PYGZus{}COUNT} \PYG{o}{/}\PYG{o}{/} \PYG{n}{BATCH\PYGZus{}SIZE}\PYG{p}{)}\PYG{p}{:}
    \PYG{n}{scene} \PYG{o}{=} \PYG{n}{Scene}\PYG{o}{.}\PYG{n}{create}\PYG{p}{(}\PYG{n}{DATA\PYGZus{}PATH}\PYG{p}{,} \PYG{n}{count}\PYG{o}{=}\PYG{n}{BATCH\PYGZus{}SIZE}\PYG{p}{)}
    \PYG{n+nb}{print}\PYG{p}{(}\PYG{n}{scene}\PYG{p}{)}
    \PYG{n}{world} \PYG{o}{=} \PYG{n}{World}\PYG{p}{(}\PYG{p}{)}
    \PYG{n}{u0} \PYG{o}{=} \PYG{n}{BurgersVelocity}\PYG{p}{(}
        \PYG{n}{DOMAIN}\PYG{p}{,} 
        \PYG{n}{velocity}\PYG{o}{=}\PYG{n}{GaussianClash}\PYG{p}{(}\PYG{n}{BATCH\PYGZus{}SIZE}\PYG{p}{)}\PYG{p}{,} 
        \PYG{n}{viscosity}\PYG{o}{=}\PYG{n}{VISCOSITY}\PYG{p}{,} 
        \PYG{n}{batch\PYGZus{}size}\PYG{o}{=}\PYG{n}{BATCH\PYGZus{}SIZE}\PYG{p}{,} 
        \PYG{n}{name}\PYG{o}{=}\PYG{l+s+s1}{\PYGZsq{}}\PYG{l+s+s1}{burgers}\PYG{l+s+s1}{\PYGZsq{}}
    \PYG{p}{)}
    \PYG{n}{u} \PYG{o}{=} \PYG{n}{world}\PYG{o}{.}\PYG{n}{add}\PYG{p}{(}\PYG{n}{u0}\PYG{p}{,} \PYG{n}{physics}\PYG{o}{=}\PYG{n}{Burgers}\PYG{p}{(}\PYG{n}{diffusion\PYGZus{}substeps}\PYG{o}{=}\PYG{n}{DIFFUSION\PYGZus{}SUBSTEPS}\PYG{p}{)}\PYG{p}{)}
    \PYG{n}{force} \PYG{o}{=} \PYG{n}{world}\PYG{o}{.}\PYG{n}{add}\PYG{p}{(}\PYG{n}{FieldEffect}\PYG{p}{(}\PYG{n}{GaussianForce}\PYG{p}{(}\PYG{n}{BATCH\PYGZus{}SIZE}\PYG{p}{)}\PYG{p}{,} \PYG{p}{[}\PYG{l+s+s1}{\PYGZsq{}}\PYG{l+s+s1}{velocity}\PYG{l+s+s1}{\PYGZsq{}}\PYG{p}{]}\PYG{p}{)}\PYG{p}{)}
    \PYG{n}{scene}\PYG{o}{.}\PYG{n}{write}\PYG{p}{(}\PYG{n}{world}\PYG{o}{.}\PYG{n}{state}\PYG{p}{,} \PYG{n}{frame}\PYG{o}{=}\PYG{l+m+mi}{0}\PYG{p}{)}
    \PYG{k}{for} \PYG{n}{frame} \PYG{o+ow}{in} \PYG{n+nb}{range}\PYG{p}{(}\PYG{l+m+mi}{1}\PYG{p}{,} \PYG{n}{STEP\PYGZus{}COUNT} \PYG{o}{+} \PYG{l+m+mi}{1}\PYG{p}{)}\PYG{p}{:}
        \PYG{n}{world}\PYG{o}{.}\PYG{n}{step}\PYG{p}{(}\PYG{n}{dt}\PYG{o}{=}\PYG{n}{DT}\PYG{p}{)}
        \PYG{n}{scene}\PYG{o}{.}\PYG{n}{write}\PYG{p}{(}\PYG{n}{world}\PYG{o}{.}\PYG{n}{state}\PYG{p}{,} \PYG{n}{frame}\PYG{o}{=}\PYG{n}{frame}\PYG{p}{)}
\end{sphinxVerbatim}

\begin{sphinxVerbatim}[commandchars=\\\{\}]
Cause: [Errno 2] No such file or directory: \PYGZsq{}\PYGZlt{}ipython\PYGZhy{}input\PYGZhy{}4\PYGZhy{}a61f572da032\PYGZgt{}\PYGZsq{}
\end{sphinxVerbatim}

\begin{sphinxVerbatim}[commandchars=\\\{\}]
forced\PYGZhy{}burgers\PYGZhy{}clash/sim\PYGZus{}000000
forced\PYGZhy{}burgers\PYGZhy{}clash/sim\PYGZus{}000100
forced\PYGZhy{}burgers\PYGZhy{}clash/sim\PYGZus{}000200
forced\PYGZhy{}burgers\PYGZhy{}clash/sim\PYGZus{}000300
forced\PYGZhy{}burgers\PYGZhy{}clash/sim\PYGZus{}000400
forced\PYGZhy{}burgers\PYGZhy{}clash/sim\PYGZus{}000500
forced\PYGZhy{}burgers\PYGZhy{}clash/sim\PYGZus{}000600
forced\PYGZhy{}burgers\PYGZhy{}clash/sim\PYGZus{}000700
forced\PYGZhy{}burgers\PYGZhy{}clash/sim\PYGZus{}000800
forced\PYGZhy{}burgers\PYGZhy{}clash/sim\PYGZus{}000900
\end{sphinxVerbatim}

\section{Training via reinforcement learning}
\label{\detokenize{reinflearn-code:training-via-reinforcement-learning}}
Next we set up the RL environment. The PPO approach uses a dedicated value estimator network (the “critic”) to predict the sum of rewards generated from a certain state. These predicted rewards are then used to update a policy network (the “actor”) which, analogously to the CFE network of {\hyperref[\detokenize{diffphys-control::doc}]{\sphinxcrossref{\DUrole{doc}{Solving Inverse Problems with NNs}}}}, predicts the forces to control the simulation.

\begin{sphinxVerbatim}[commandchars=\\\{\}]
\PYG{k+kn}{from} \PYG{n+nn}{experiment} \PYG{k+kn}{import} \PYG{n}{BurgersTraining}

\PYG{n}{N\PYGZus{}ENVS} \PYG{o}{=} \PYG{l+m+mi}{10}                         \PYG{c+c1}{\PYGZsh{} On how many environments to train in parallel, load balancing}
\PYG{n}{FINAL\PYGZus{}REWARD\PYGZus{}FACTOR} \PYG{o}{=} \PYG{n}{STEP\PYGZus{}COUNT}    \PYG{c+c1}{\PYGZsh{} Penalty for not reaching the goal state}
\PYG{n}{STEPS\PYGZus{}PER\PYGZus{}ROLLOUT} \PYG{o}{=} \PYG{n}{STEP\PYGZus{}COUNT} \PYG{o}{*} \PYG{l+m+mi}{10} \PYG{c+c1}{\PYGZsh{} How many steps to collect per environment between agent updates}
\PYG{n}{N\PYGZus{}EPOCHS} \PYG{o}{=} \PYG{l+m+mi}{10}                       \PYG{c+c1}{\PYGZsh{} How many epochs to perform during each agent update}
\PYG{n}{RL\PYGZus{}LEARNING\PYGZus{}RATE} \PYG{o}{=} \PYG{l+m+mf}{1e\PYGZhy{}4}             \PYG{c+c1}{\PYGZsh{} Learning rate for agent updates}
\PYG{n}{RL\PYGZus{}BATCH\PYGZus{}SIZE} \PYG{o}{=} \PYG{l+m+mi}{128}                 \PYG{c+c1}{\PYGZsh{} Batch size for agent updates}
\PYG{n}{RL\PYGZus{}ROLLOUTS} \PYG{o}{=} \PYG{l+m+mi}{500}                  \PYG{c+c1}{\PYGZsh{} Number of iterations for RL training}
\end{sphinxVerbatim}

To start training, we create a trainer object which manages the environment and the agent internally. Additionally, a directory for storing models, logs, and hyperparameters is created. This way, training can be continued at any later point using the same configuration. If the model folder specified in \sphinxcode{\sphinxupquote{exp\_name}} already exists, the agent within is loaded; otherwise, a new agent is created. For the PPO reinforcement learning algorithm, the implementation of \sphinxcode{\sphinxupquote{stable\_baselines3}} is used. The trainer class acts as a wrapper for this system. Under the hood, an instance of a \sphinxcode{\sphinxupquote{BurgersEnv}} gym environment is created, which is loaded into the PPO algorithm. It generates random initial states, precomputes corresponding ground truth simulations and handles the system evolution influenced by the agent’s actions. Furthermore, the trainer regularly evaluates the performance on the validation set by loading a different environment that uses the initial and target states of the validation set.

\subsection{Gym environment}
\label{\detokenize{reinflearn-code:gym-environment}}
The gym environment specification provides an interface leveraging the interaction with the agent. Environments implementing it must specify observation and action spaces, which represent the in\sphinxhyphen{} and output spaces of the agent. Further, they have to define a set of methods, the most important ones being \sphinxcode{\sphinxupquote{reset}}, \sphinxcode{\sphinxupquote{step}}, and \sphinxcode{\sphinxupquote{render}}.
\begin{itemize}
\item {} 
\sphinxcode{\sphinxupquote{reset}} is called after a trajectory has ended, to revert the environment to an initial state, and returns the corresponding observation.

\item {} 
\sphinxcode{\sphinxupquote{step}} takes an action given by the agent and iterates the environment to the next state. It returns the resulting observation, the received reward, a flag determining whether a terminal state has been reached and a dictionary for debugging and logging information.

\item {} 
\sphinxcode{\sphinxupquote{render}} is called to display the current environment state in a way the creator of the environment specifies. This function can be used to inspect the training results.

\end{itemize}

\sphinxcode{\sphinxupquote{stable\sphinxhyphen{}baselines3}} expands on the default gym environment by providing an interface for vectorized environments. This makes it possible to compute the forward pass for multiple trajectories simultaneously which can in turn increase time efficiency because of better resource utilization. In practice, this means that the methods now work on vectors of observations, actions, rewards, terminal state flags and info dictionaries. The step method is split into \sphinxcode{\sphinxupquote{step\_async}} and \sphinxcode{\sphinxupquote{step\_wait}}, making it possible to run individual instances of the environment on different threads.

\subsection{Physics simulation}
\label{\detokenize{reinflearn-code:physics-simulation}}
The environment for Burgers’ equation contains a \sphinxcode{\sphinxupquote{Burgers}} physics object provided by \sphinxcode{\sphinxupquote{phiflow}}. The states are internally stored as \sphinxcode{\sphinxupquote{BurgersVelocity}} objects. To create the initial states, the environment generates batches of random fields in the same fashion as in the data set generation process shown above. The observation space consists of the velocity fields of the current and target states stacked in the channel dimension with another channel specifying the current time step. Actions are taken in the form of a one dimensional array covering every velocity value. The \sphinxcode{\sphinxupquote{step}} method calls the physics object to advance the internal state by one time step, also applying the actions as a \sphinxcode{\sphinxupquote{FieldEffect}}.

The rewards encompass a penalty equal to the square norm of the generated forces at every time step. Additionally, the \(L^2\) distance to the target field, scaled by a predefined factor (\sphinxcode{\sphinxupquote{FINAL\_REWARD\_FACTOR}}) is subtracted at the end of each trajectory. The rewards are then normalized with a running estimate for the reward mean and standard deviation.

\subsection{Neural network setup}
\label{\detokenize{reinflearn-code:neural-network-setup}}
We use two different neural network architectures for the actor and critic respectively. The former uses the U\sphinxhyphen{}Net variant from {[}\hyperlink{cite.references:id11}{HKT19}{]}, while the latter consists of a series of 1D convolutional and pooling layers reducing the feature map size to one. The final operation is a convolution with kernel size one to combine the feature maps and retain one output value. The \sphinxcode{\sphinxupquote{CustomActorCriticPolicy}} class then makes it possible to use these two separate network architectures for the reinforcement learning agent.

By default, an agent is stored at \sphinxcode{\sphinxupquote{PDE\sphinxhyphen{}Control\sphinxhyphen{}RL/networks/rl\sphinxhyphen{}models/bench}}, and loaded if it exists. (If necessary, replace the \sphinxcode{\sphinxupquote{path}} below with a new one to start with a new model.) As the training takes quite long, we’re starting with a pre\sphinxhyphen{}trained agent here. It is already trained for 3500 iterations, and hence we’re only doing a “fine\sphinxhyphen{}tuning” below with another \sphinxcode{\sphinxupquote{RL\_ROLLOUTS=500}} iterations. These typically take around 2 hours, and hence the total training time of almost 18 hours would be too long for interactive tests. (However, the code provided here contains everything to train a model from scratch if you have the resources.)

\begin{sphinxVerbatim}[commandchars=\\\{\}]
\PYG{n}{rl\PYGZus{}trainer} \PYG{o}{=} \PYG{n}{BurgersTraining}\PYG{p}{(}
    \PYG{n}{path}\PYG{o}{=}\PYG{l+s+s1}{\PYGZsq{}}\PYG{l+s+s1}{PDE\PYGZhy{}Control\PYGZhy{}RL/networks/rl\PYGZhy{}models/bench}\PYG{l+s+s1}{\PYGZsq{}}\PYG{p}{,} \PYG{c+c1}{\PYGZsh{} Replace path to train a new model}
    \PYG{n}{domain}\PYG{o}{=}\PYG{n}{DOMAIN}\PYG{p}{,}
    \PYG{n}{viscosity}\PYG{o}{=}\PYG{n}{VISCOSITY}\PYG{p}{,}
    \PYG{n}{step\PYGZus{}count}\PYG{o}{=}\PYG{n}{STEP\PYGZus{}COUNT}\PYG{p}{,}
    \PYG{n}{dt}\PYG{o}{=}\PYG{n}{DT}\PYG{p}{,}
    \PYG{n}{diffusion\PYGZus{}substeps}\PYG{o}{=}\PYG{n}{DIFFUSION\PYGZus{}SUBSTEPS}\PYG{p}{,}
    \PYG{n}{n\PYGZus{}envs}\PYG{o}{=}\PYG{n}{N\PYGZus{}ENVS}\PYG{p}{,}
    \PYG{n}{final\PYGZus{}reward\PYGZus{}factor}\PYG{o}{=}\PYG{n}{FINAL\PYGZus{}REWARD\PYGZus{}FACTOR}\PYG{p}{,}
    \PYG{n}{steps\PYGZus{}per\PYGZus{}rollout}\PYG{o}{=}\PYG{n}{STEPS\PYGZus{}PER\PYGZus{}ROLLOUT}\PYG{p}{,}
    \PYG{n}{n\PYGZus{}epochs}\PYG{o}{=}\PYG{n}{N\PYGZus{}EPOCHS}\PYG{p}{,}
    \PYG{n}{learning\PYGZus{}rate}\PYG{o}{=}\PYG{n}{RL\PYGZus{}LEARNING\PYGZus{}RATE}\PYG{p}{,}
    \PYG{n}{batch\PYGZus{}size}\PYG{o}{=}\PYG{n}{RL\PYGZus{}BATCH\PYGZus{}SIZE}\PYG{p}{,}
    \PYG{n}{data\PYGZus{}path}\PYG{o}{=}\PYG{n}{DATA\PYGZus{}PATH}\PYG{p}{,}
    \PYG{n}{val\PYGZus{}range}\PYG{o}{=}\PYG{n}{VAL\PYGZus{}RANGE}\PYG{p}{,}
    \PYG{n}{test\PYGZus{}range}\PYG{o}{=}\PYG{n}{TEST\PYGZus{}RANGE}\PYG{p}{,}
\PYG{p}{)}
\end{sphinxVerbatim}

\begin{sphinxVerbatim}[commandchars=\\\{\}]
Tensorboard log path: PDE\PYGZhy{}Control\PYGZhy{}RL/networks/rl\PYGZhy{}models/bench/tensorboard\PYGZhy{}log
Loading existing agent from PDE\PYGZhy{}Control\PYGZhy{}RL/networks/rl\PYGZhy{}models/bench/agent.zip
\end{sphinxVerbatim}

The following cell is optional but very useful for debugging: it opens \sphinxstyleemphasis{tensorboard} inside the notebook to display the progress of the training. If a new model was created at a different location, please change the path accordingly. When resuming the learning process of a pre\sphinxhyphen{}trained agent, the new run is shown separately in tensorboard (enable reload via the cogwheel button).

The graph titled “forces” shows how the overall amount of forces generated by the network evolves during training. “rew\_unnormalized” shows the reward values without the normalization step described above. The corresponding values with normalization are shown under “rollout/ep\_rew\_mean”. “val\_set\_forces” outlines the performance of the agent on the validation set.

\begin{sphinxVerbatim}[commandchars=\\\{\}]
\PYG{o}{\PYGZpc{}}\PYG{k}{load\PYGZus{}ext} tensorboard
\PYG{o}{\PYGZpc{}}\PYG{k}{tensorboard} \PYGZhy{}\PYGZhy{}logdir PDE\PYGZhy{}Control\PYGZhy{}RL/networks/rl\PYGZhy{}models/bench/tensorboard\PYGZhy{}log
\end{sphinxVerbatim}

Now we are set up to start training the agent. The RL approach requires many iterations to explore the environment. Hence, the next cell typically takes multiple hours to execute (around 2h for 500 rollouts).

\begin{sphinxVerbatim}[commandchars=\\\{\}]
\PYG{n}{rl\PYGZus{}trainer}\PYG{o}{.}\PYG{n}{train}\PYG{p}{(}\PYG{n}{n\PYGZus{}rollouts}\PYG{o}{=}\PYG{n}{RL\PYGZus{}ROLLOUTS}\PYG{p}{,} \PYG{n}{save\PYGZus{}freq}\PYG{o}{=}\PYG{l+m+mi}{50}\PYG{p}{)}
\end{sphinxVerbatim}

\begin{sphinxVerbatim}[commandchars=\\\{\}]
Storing agent and hyperparameters to disk...
\end{sphinxVerbatim}

\begin{sphinxVerbatim}[commandchars=\\\{\}]
  \PYGZdq{}See the documentation of nn.Upsample for details.\PYGZdq{}.format(mode)
  return torch.max\PYGZus{}pool1d(input, kernel\PYGZus{}size, stride, padding, dilation, ceil\PYGZus{}mode)
\end{sphinxVerbatim}

\begin{sphinxVerbatim}[commandchars=\\\{\}]
Storing agent and hyperparameters to disk...
Storing agent and hyperparameters to disk...
Storing agent and hyperparameters to disk...
Storing agent and hyperparameters to disk...
Storing agent and hyperparameters to disk...
Storing agent and hyperparameters to disk...
Storing agent and hyperparameters to disk...
Storing agent and hyperparameters to disk...
Storing agent and hyperparameters to disk...
Storing agent and hyperparameters to disk...
Storing agent and hyperparameters to disk...
\end{sphinxVerbatim}

\section{RL evaluation}
\label{\detokenize{reinflearn-code:rl-evaluation}}
Now that we have a trained model, let’s take a look at the results. The leftmost plot shows the results of the reinforcement learning agent. As reference, next to it are shown the ground truth, i.e. the trajectory the agent should reconstruct, and the uncontrolled simulation where the system follows its natural evolution.

\begin{sphinxVerbatim}[commandchars=\\\{\}]
\PYG{n}{TEST\PYGZus{}SAMPLE} \PYG{o}{=} \PYG{l+m+mi}{0}    \PYG{c+c1}{\PYGZsh{} Change this to display a reconstruction of another scene}
\PYG{n}{rl\PYGZus{}frames}\PYG{p}{,} \PYG{n}{gt\PYGZus{}frames}\PYG{p}{,} \PYG{n}{unc\PYGZus{}frames} \PYG{o}{=} \PYG{n}{rl\PYGZus{}trainer}\PYG{o}{.}\PYG{n}{infer\PYGZus{}test\PYGZus{}set\PYGZus{}frames}\PYG{p}{(}\PYG{p}{)}

\PYG{n}{fig}\PYG{p}{,} \PYG{n}{axs} \PYG{o}{=} \PYG{n}{plt}\PYG{o}{.}\PYG{n}{subplots}\PYG{p}{(}\PYG{l+m+mi}{1}\PYG{p}{,} \PYG{l+m+mi}{3}\PYG{p}{,} \PYG{n}{figsize}\PYG{o}{=}\PYG{p}{(}\PYG{l+m+mf}{18.9}\PYG{p}{,} \PYG{l+m+mf}{9.6}\PYG{p}{)}\PYG{p}{)}
\PYG{n}{axs}\PYG{p}{[}\PYG{l+m+mi}{0}\PYG{p}{]}\PYG{o}{.}\PYG{n}{set\PYGZus{}title}\PYG{p}{(}\PYG{l+s+s2}{\PYGZdq{}}\PYG{l+s+s2}{Reinforcement Learning}\PYG{l+s+s2}{\PYGZdq{}}\PYG{p}{)}\PYG{p}{;} \PYG{n}{axs}\PYG{p}{[}\PYG{l+m+mi}{1}\PYG{p}{]}\PYG{o}{.}\PYG{n}{set\PYGZus{}title}\PYG{p}{(}\PYG{l+s+s2}{\PYGZdq{}}\PYG{l+s+s2}{Ground Truth}\PYG{l+s+s2}{\PYGZdq{}}\PYG{p}{)}\PYG{p}{;} \PYG{n}{axs}\PYG{p}{[}\PYG{l+m+mi}{2}\PYG{p}{]}\PYG{o}{.}\PYG{n}{set\PYGZus{}title}\PYG{p}{(}\PYG{l+s+s2}{\PYGZdq{}}\PYG{l+s+s2}{Uncontrolled}\PYG{l+s+s2}{\PYGZdq{}}\PYG{p}{)}
\PYG{k}{for} \PYG{n}{plot} \PYG{o+ow}{in} \PYG{n}{axs}\PYG{p}{:}
    \PYG{n}{plot}\PYG{o}{.}\PYG{n}{set\PYGZus{}ylim}\PYG{p}{(}\PYG{o}{\PYGZhy{}}\PYG{l+m+mi}{2}\PYG{p}{,} \PYG{l+m+mi}{2}\PYG{p}{)}\PYG{p}{;} \PYG{n}{plot}\PYG{o}{.}\PYG{n}{set\PYGZus{}xlabel}\PYG{p}{(}\PYG{l+s+s1}{\PYGZsq{}}\PYG{l+s+s1}{x}\PYG{l+s+s1}{\PYGZsq{}}\PYG{p}{)}\PYG{p}{;} \PYG{n}{plot}\PYG{o}{.}\PYG{n}{set\PYGZus{}ylabel}\PYG{p}{(}\PYG{l+s+s1}{\PYGZsq{}}\PYG{l+s+s1}{u(x)}\PYG{l+s+s1}{\PYGZsq{}}\PYG{p}{)}

\PYG{k}{for} \PYG{n}{frame} \PYG{o+ow}{in} \PYG{n+nb}{range}\PYG{p}{(}\PYG{l+m+mi}{0}\PYG{p}{,} \PYG{n}{STEP\PYGZus{}COUNT} \PYG{o}{+} \PYG{l+m+mi}{1}\PYG{p}{)}\PYG{p}{:}
    \PYG{n}{frame\PYGZus{}color} \PYG{o}{=} \PYG{n}{bplt}\PYG{o}{.}\PYG{n}{gradient\PYGZus{}color}\PYG{p}{(}\PYG{n}{frame}\PYG{p}{,} \PYG{n}{STEP\PYGZus{}COUNT}\PYG{o}{+}\PYG{l+m+mi}{1}\PYG{p}{)}\PYG{p}{;}
    \PYG{n}{axs}\PYG{p}{[}\PYG{l+m+mi}{0}\PYG{p}{]}\PYG{o}{.}\PYG{n}{plot}\PYG{p}{(}\PYG{n}{rl\PYGZus{}frames}\PYG{p}{[}\PYG{n}{frame}\PYG{p}{]}\PYG{p}{[}\PYG{n}{TEST\PYGZus{}SAMPLE}\PYG{p}{,}\PYG{p}{:}\PYG{p}{]}\PYG{p}{,} \PYG{n}{color}\PYG{o}{=}\PYG{n}{frame\PYGZus{}color}\PYG{p}{,} \PYG{n}{linewidth}\PYG{o}{=}\PYG{l+m+mf}{0.8}\PYG{p}{)}
    \PYG{n}{axs}\PYG{p}{[}\PYG{l+m+mi}{1}\PYG{p}{]}\PYG{o}{.}\PYG{n}{plot}\PYG{p}{(}\PYG{n}{gt\PYGZus{}frames}\PYG{p}{[}\PYG{n}{frame}\PYG{p}{]}\PYG{p}{[}\PYG{n}{TEST\PYGZus{}SAMPLE}\PYG{p}{,}\PYG{p}{:}\PYG{p}{]}\PYG{p}{,} \PYG{n}{color}\PYG{o}{=}\PYG{n}{frame\PYGZus{}color}\PYG{p}{,} \PYG{n}{linewidth}\PYG{o}{=}\PYG{l+m+mf}{0.8}\PYG{p}{)}
    \PYG{n}{axs}\PYG{p}{[}\PYG{l+m+mi}{2}\PYG{p}{]}\PYG{o}{.}\PYG{n}{plot}\PYG{p}{(}\PYG{n}{unc\PYGZus{}frames}\PYG{p}{[}\PYG{n}{frame}\PYG{p}{]}\PYG{p}{[}\PYG{n}{TEST\PYGZus{}SAMPLE}\PYG{p}{,}\PYG{p}{:}\PYG{p}{]}\PYG{p}{,} \PYG{n}{color}\PYG{o}{=}\PYG{n}{frame\PYGZus{}color}\PYG{p}{,} \PYG{n}{linewidth}\PYG{o}{=}\PYG{l+m+mf}{0.8}\PYG{p}{)}
\end{sphinxVerbatim}

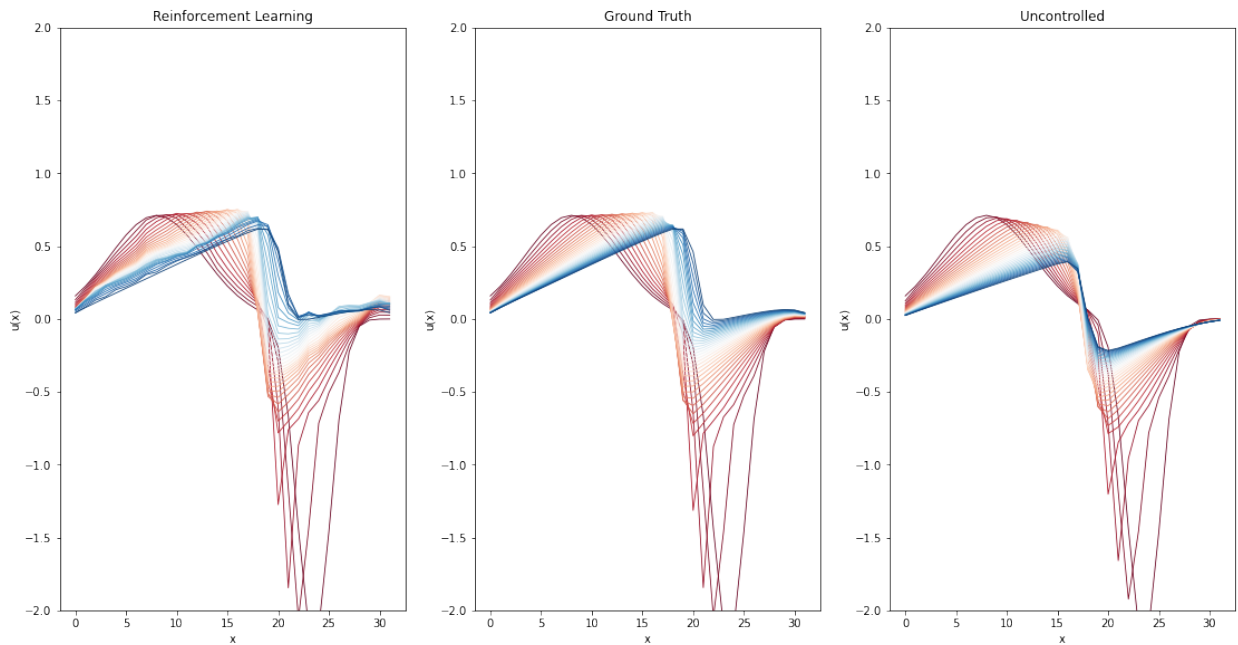
\begin{sphinxVerbatim}[commandchars=\\\{\}]
  \PYGZdq{}See the documentation of nn.Upsample for details.\PYGZdq{}.format(mode)
\end{sphinxVerbatim}

\noindent\sphinxincludegraphics{{reinflearn-code_21_1}.png}

As we can see, a trained reinforcement learning agent is able to reconstruct the trajectories fairly well. However, they still appear noticeably less smooth than the ground truth.

\section{Differentiable physics training}
\label{\detokenize{reinflearn-code:differentiable-physics-training}}
To classify the results of the reinforcement learning method, we now compare them to an approach using differentiable physics training. In contrast to the full approach from {\hyperref[\detokenize{diffphys-control::doc}]{\sphinxcrossref{\DUrole{doc}{Solving Inverse Problems with NNs}}}} which includes a second \sphinxstyleemphasis{OP} network, we aim for a direct control here. The OP network represents a separate “physics\sphinxhyphen{}predictor”, which is omitted here for fairness when comparing with the RL version.

The DP approach has access to the gradient data provided by the differentiable solver, making it possible to trace the loss over multiple time steps and enabling the model to comprehend long term effects of generated forces better. The reinforcement learning algorithm, on the other hand, is not limited by training set size like the DP algorithm, as new training samples are generated on policy. However, this also introduces additional simulation overhead during training, which can increase the time needed for convergence.

\begin{sphinxVerbatim}[commandchars=\\\{\}]
\PYG{k+kn}{from} \PYG{n+nn}{control}\PYG{n+nn}{.}\PYG{n+nn}{pde}\PYG{n+nn}{.}\PYG{n+nn}{burgers} \PYG{k+kn}{import} \PYG{n}{BurgersPDE}
\PYG{k+kn}{from} \PYG{n+nn}{control}\PYG{n+nn}{.}\PYG{n+nn}{control\PYGZus{}training} \PYG{k+kn}{import} \PYG{n}{ControlTraining}
\PYG{k+kn}{from} \PYG{n+nn}{control}\PYG{n+nn}{.}\PYG{n+nn}{sequences} \PYG{k+kn}{import} \PYG{n}{StaggeredSequence}
\end{sphinxVerbatim}

\begin{sphinxVerbatim}[commandchars=\\\{\}]
Could not load resample cuda libraries: CUDA binaries not found at /usr/local/lib/python3.7/dist\PYGZhy{}packages/phi/tf/cuda/build/resample.so. Run \PYGZdq{}python setup.py cuda\PYGZdq{} to compile them
\end{sphinxVerbatim}

\begin{sphinxVerbatim}[commandchars=\\\{\}]
\PYGZdl{} python setup.py tf\PYGZus{}cuda
before reinstalling phiflow.
\end{sphinxVerbatim}

The cell below sets up a model for training or to load an existing model checkpoint.

\begin{sphinxVerbatim}[commandchars=\\\{\}]
\PYG{n}{dp\PYGZus{}app} \PYG{o}{=} \PYG{n}{ControlTraining}\PYG{p}{(}
    \PYG{n}{STEP\PYGZus{}COUNT}\PYG{p}{,}
    \PYG{n}{BurgersPDE}\PYG{p}{(}\PYG{n}{DOMAIN}\PYG{p}{,} \PYG{n}{VISCOSITY}\PYG{p}{,} \PYG{n}{DT}\PYG{p}{)}\PYG{p}{,}
    \PYG{n}{datapath}\PYG{o}{=}\PYG{n}{DATA\PYGZus{}PATH}\PYG{p}{,}
    \PYG{n}{val\PYGZus{}range}\PYG{o}{=}\PYG{n}{VAL\PYGZus{}RANGE}\PYG{p}{,}
    \PYG{n}{train\PYGZus{}range}\PYG{o}{=}\PYG{n}{TRAIN\PYGZus{}RANGE}\PYG{p}{,}
    \PYG{n}{trace\PYGZus{}to\PYGZus{}channel}\PYG{o}{=}\PYG{k}{lambda} \PYG{n}{trace}\PYG{p}{:} \PYG{l+s+s1}{\PYGZsq{}}\PYG{l+s+s1}{burgers\PYGZus{}velocity}\PYG{l+s+s1}{\PYGZsq{}}\PYG{p}{,}
    \PYG{n}{obs\PYGZus{}loss\PYGZus{}frames}\PYG{o}{=}\PYG{p}{[}\PYG{p}{]}\PYG{p}{,}
    \PYG{n}{trainable\PYGZus{}networks}\PYG{o}{=}\PYG{p}{[}\PYG{l+s+s1}{\PYGZsq{}}\PYG{l+s+s1}{CFE}\PYG{l+s+s1}{\PYGZsq{}}\PYG{p}{]}\PYG{p}{,}
    \PYG{n}{sequence\PYGZus{}class}\PYG{o}{=}\PYG{n}{StaggeredSequence}\PYG{p}{,}
    \PYG{n}{batch\PYGZus{}size}\PYG{o}{=}\PYG{l+m+mi}{100}\PYG{p}{,}
    \PYG{n}{view\PYGZus{}size}\PYG{o}{=}\PYG{l+m+mi}{20}\PYG{p}{,}
    \PYG{n}{learning\PYGZus{}rate}\PYG{o}{=}\PYG{l+m+mf}{1e\PYGZhy{}3}\PYG{p}{,}
    \PYG{n}{learning\PYGZus{}rate\PYGZus{}half\PYGZus{}life}\PYG{o}{=}\PYG{l+m+mi}{1000}\PYG{p}{,}
    \PYG{n}{dt}\PYG{o}{=}\PYG{n}{DT}
\PYG{p}{)}\PYG{o}{.}\PYG{n}{prepare}\PYG{p}{(}\PYG{p}{)}
\end{sphinxVerbatim}

\begin{sphinxVerbatim}[commandchars=\\\{\}]
App created. Scene directory is /root/phi/model/control\PYGZhy{}training/sim\PYGZus{}000000 (INFO), 2021\PYGZhy{}08\PYGZhy{}04 10:11:58,466n

Sequence class: \PYGZlt{}class \PYGZsq{}control.sequences.StaggeredSequence\PYGZsq{}\PYGZgt{} (INFO), 2021\PYGZhy{}08\PYGZhy{}04 10:12:01,449n

Partition length 32 sequence (from 0 to 32) at frame 16
\end{sphinxVerbatim}

\begin{sphinxVerbatim}[commandchars=\\\{\}]

\end{sphinxVerbatim}

\begin{sphinxVerbatim}[commandchars=\\\{\}]
Partition length 16 sequence (from 0 to 16) at frame 8
Partition length 8 sequence (from 0 to 8) at frame 4
Partition length 4 sequence (from 0 to 4) at frame 2
Partition length 2 sequence (from 0 to 2) at frame 1
Execute \PYGZhy{}\PYGZgt{} 1
Execute \PYGZhy{}\PYGZgt{} 2
Partition length 2 sequence (from 2 to 4) at frame 3
Execute \PYGZhy{}\PYGZgt{} 3
Execute \PYGZhy{}\PYGZgt{} 4
Partition length 4 sequence (from 4 to 8) at frame 6
Partition length 2 sequence (from 4 to 6) at frame 5
Execute \PYGZhy{}\PYGZgt{} 5
Execute \PYGZhy{}\PYGZgt{} 6
Partition length 2 sequence (from 6 to 8) at frame 7
Execute \PYGZhy{}\PYGZgt{} 7
Execute \PYGZhy{}\PYGZgt{} 8
Partition length 8 sequence (from 8 to 16) at frame 12
Partition length 4 sequence (from 8 to 12) at frame 10
Partition length 2 sequence (from 8 to 10) at frame 9
Execute \PYGZhy{}\PYGZgt{} 9
Execute \PYGZhy{}\PYGZgt{} 10
Partition length 2 sequence (from 10 to 12) at frame 11
Execute \PYGZhy{}\PYGZgt{} 11
Execute \PYGZhy{}\PYGZgt{} 12
Partition length 4 sequence (from 12 to 16) at frame 14
Partition length 2 sequence (from 12 to 14) at frame 13
Execute \PYGZhy{}\PYGZgt{} 13
Execute \PYGZhy{}\PYGZgt{} 14
Partition length 2 sequence (from 14 to 16) at frame 15
Execute \PYGZhy{}\PYGZgt{} 15
Execute \PYGZhy{}\PYGZgt{} 16
Partition length 16 sequence (from 16 to 32) at frame 24
Partition length 8 sequence (from 16 to 24) at frame 20
Partition length 4 sequence (from 16 to 20) at frame 18
Partition length 2 sequence (from 16 to 18) at frame 17
Execute \PYGZhy{}\PYGZgt{} 17
Execute \PYGZhy{}\PYGZgt{} 18
Partition length 2 sequence (from 18 to 20) at frame 19
Execute \PYGZhy{}\PYGZgt{} 19
Execute \PYGZhy{}\PYGZgt{} 20
Partition length 4 sequence (from 20 to 24) at frame 22
Partition length 2 sequence (from 20 to 22) at frame 21
Execute \PYGZhy{}\PYGZgt{} 21
Execute \PYGZhy{}\PYGZgt{} 22
Partition length 2 sequence (from 22 to 24) at frame 23
Execute \PYGZhy{}\PYGZgt{} 23
Execute \PYGZhy{}\PYGZgt{} 24
Partition length 8 sequence (from 24 to 32) at frame 28
Partition length 4 sequence (from 24 to 28) at frame 26
Partition length 2 sequence (from 24 to 26) at frame 25
Execute \PYGZhy{}\PYGZgt{} 25
Execute \PYGZhy{}\PYGZgt{} 26
Partition length 2 sequence (from 26 to 28) at frame 27
Execute \PYGZhy{}\PYGZgt{} 27
Execute \PYGZhy{}\PYGZgt{} 28
Partition length 4 sequence (from 28 to 32) at frame 30
Partition length 2 sequence (from 28 to 30) at frame 29
Execute \PYGZhy{}\PYGZgt{} 29
Execute \PYGZhy{}\PYGZgt{} 30
Partition length 2 sequence (from 30 to 32) at frame 31
Execute \PYGZhy{}\PYGZgt{} 31
Execute \PYGZhy{}\PYGZgt{} 32
Target loss: Tensor(\PYGZdq{}truediv\PYGZus{}1:0\PYGZdq{}, shape=(), dtype=float32) (INFO), 2021\PYGZhy{}08\PYGZhy{}04 10:13:10,701n

Force loss: Tensor(\PYGZdq{}Sum\PYGZus{}97:0\PYGZdq{}, shape=(), dtype=float32) (INFO), 2021\PYGZhy{}08\PYGZhy{}04 10:13:14,221n

Setting up loss (INFO), 2021\PYGZhy{}08\PYGZhy{}04 10:13:14,223n

Preparing data (INFO), 2021\PYGZhy{}08\PYGZhy{}04 10:13:51,128n

INFO:tensorflow:Summary name Total Force is illegal; using Total\PYGZus{}Force instead.
Initializing variables (INFO), 2021\PYGZhy{}08\PYGZhy{}04 10:13:51,156n

Model variables contain 0 total parameters. (INFO), 2021\PYGZhy{}08\PYGZhy{}04 10:13:55,961n

Validation (000000): Learning\PYGZus{}Rate: 0.001, Loss\PYGZus{}reg\PYGZus{}unscaled: 205.98526, Loss\PYGZus{}reg\PYGZus{}scale: 1.0, Loss: 0.0, Total Force: 393.8109 (INFO), 2021\PYGZhy{}08\PYGZhy{}04 10:14:32,455n
\end{sphinxVerbatim}

Now we can execute the model training. This cell typically also takes a while to execute (ca. 2h for 1000 iterations).

\begin{sphinxVerbatim}[commandchars=\\\{\}]
\PYG{n}{DP\PYGZus{}TRAINING\PYGZus{}ITERATIONS} \PYG{o}{=} \PYG{l+m+mi}{10000}  \PYG{c+c1}{\PYGZsh{} Change this to change training duration}

\PYG{n}{dp\PYGZus{}training\PYGZus{}eval\PYGZus{}data} \PYG{o}{=} \PYG{p}{[}\PYG{p}{]}
\PYG{n}{start\PYGZus{}time} \PYG{o}{=} \PYG{n}{time}\PYG{o}{.}\PYG{n}{time}\PYG{p}{(}\PYG{p}{)}

\PYG{k}{for} \PYG{n}{epoch} \PYG{o+ow}{in} \PYG{n+nb}{range}\PYG{p}{(}\PYG{n}{DP\PYGZus{}TRAINING\PYGZus{}ITERATIONS}\PYG{p}{)}\PYG{p}{:}
    \PYG{n}{dp\PYGZus{}app}\PYG{o}{.}\PYG{n}{progress}\PYG{p}{(}\PYG{p}{)}
    \PYG{c+c1}{\PYGZsh{} Evaluate validation set at regular intervals to track learning progress}
    \PYG{c+c1}{\PYGZsh{} Size of intervals determined by RL epoch count per iteration for accurate comparison}
    \PYG{k}{if} \PYG{n}{epoch} \PYG{o}{\PYGZpc{}} \PYG{n}{N\PYGZus{}EPOCHS} \PYG{o}{==} \PYG{l+m+mi}{0}\PYG{p}{:}
        \PYG{n}{f} \PYG{o}{=} \PYG{n}{dp\PYGZus{}app}\PYG{o}{.}\PYG{n}{infer\PYGZus{}scalars}\PYG{p}{(}\PYG{n}{VAL\PYGZus{}RANGE}\PYG{p}{)}\PYG{p}{[}\PYG{l+s+s1}{\PYGZsq{}}\PYG{l+s+s1}{Total Force}\PYG{l+s+s1}{\PYGZsq{}}\PYG{p}{]} \PYG{o}{/} \PYG{n}{DT}
        \PYG{n}{dp\PYGZus{}training\PYGZus{}eval\PYGZus{}data}\PYG{o}{.}\PYG{n}{append}\PYG{p}{(}\PYG{p}{(}\PYG{n}{time}\PYG{o}{.}\PYG{n}{time}\PYG{p}{(}\PYG{p}{)} \PYG{o}{\PYGZhy{}} \PYG{n}{start\PYGZus{}time}\PYG{p}{,} \PYG{n}{epoch}\PYG{p}{,} \PYG{n}{f}\PYG{p}{)}\PYG{p}{)}
\end{sphinxVerbatim}

The trained model and the validation performance \sphinxcode{\sphinxupquote{val\_forces.csv}} with respect to iterations and wall time are saved on disk:

\begin{sphinxVerbatim}[commandchars=\\\{\}]
\PYG{n}{DP\PYGZus{}STORE\PYGZus{}PATH} \PYG{o}{=} \PYG{l+s+s1}{\PYGZsq{}}\PYG{l+s+s1}{networks/dp\PYGZhy{}models/bench}\PYG{l+s+s1}{\PYGZsq{}}
\PYG{k}{if} \PYG{o+ow}{not} \PYG{n}{os}\PYG{o}{.}\PYG{n}{path}\PYG{o}{.}\PYG{n}{exists}\PYG{p}{(}\PYG{n}{DP\PYGZus{}STORE\PYGZus{}PATH}\PYG{p}{)}\PYG{p}{:}
    \PYG{n}{os}\PYG{o}{.}\PYG{n}{makedirs}\PYG{p}{(}\PYG{n}{DP\PYGZus{}STORE\PYGZus{}PATH}\PYG{p}{)}

\PYG{c+c1}{\PYGZsh{} store training progress information}
\PYG{k}{with} \PYG{n+nb}{open}\PYG{p}{(}\PYG{n}{os}\PYG{o}{.}\PYG{n}{path}\PYG{o}{.}\PYG{n}{join}\PYG{p}{(}\PYG{n}{DP\PYGZus{}STORE\PYGZus{}PATH}\PYG{p}{,} \PYG{l+s+s1}{\PYGZsq{}}\PYG{l+s+s1}{val\PYGZus{}forces.csv}\PYG{l+s+s1}{\PYGZsq{}}\PYG{p}{)}\PYG{p}{,} \PYG{l+s+s1}{\PYGZsq{}}\PYG{l+s+s1}{at}\PYG{l+s+s1}{\PYGZsq{}}\PYG{p}{)} \PYG{k}{as} \PYG{n}{log\PYGZus{}file}\PYG{p}{:}
    \PYG{n}{logger} \PYG{o}{=} \PYG{n}{csv}\PYG{o}{.}\PYG{n}{DictWriter}\PYG{p}{(}\PYG{n}{log\PYGZus{}file}\PYG{p}{,} \PYG{p}{(}\PYG{l+s+s1}{\PYGZsq{}}\PYG{l+s+s1}{time}\PYG{l+s+s1}{\PYGZsq{}}\PYG{p}{,} \PYG{l+s+s1}{\PYGZsq{}}\PYG{l+s+s1}{epoch}\PYG{l+s+s1}{\PYGZsq{}}\PYG{p}{,} \PYG{l+s+s1}{\PYGZsq{}}\PYG{l+s+s1}{forces}\PYG{l+s+s1}{\PYGZsq{}}\PYG{p}{)}\PYG{p}{)}
    \PYG{n}{logger}\PYG{o}{.}\PYG{n}{writeheader}\PYG{p}{(}\PYG{p}{)}
    \PYG{k}{for} \PYG{p}{(}\PYG{n}{t}\PYG{p}{,} \PYG{n}{e}\PYG{p}{,} \PYG{n}{f}\PYG{p}{)} \PYG{o+ow}{in} \PYG{n}{dp\PYGZus{}training\PYGZus{}eval\PYGZus{}data}\PYG{p}{:}
        \PYG{n}{logger}\PYG{o}{.}\PYG{n}{writerow}\PYG{p}{(}\PYG{p}{\PYGZob{}}\PYG{l+s+s1}{\PYGZsq{}}\PYG{l+s+s1}{time}\PYG{l+s+s1}{\PYGZsq{}}\PYG{p}{:} \PYG{n}{t}\PYG{p}{,} \PYG{l+s+s1}{\PYGZsq{}}\PYG{l+s+s1}{epoch}\PYG{l+s+s1}{\PYGZsq{}}\PYG{p}{:} \PYG{n}{e}\PYG{p}{,} \PYG{l+s+s1}{\PYGZsq{}}\PYG{l+s+s1}{forces}\PYG{l+s+s1}{\PYGZsq{}}\PYG{p}{:} \PYG{n}{f}\PYG{p}{\PYGZcb{}}\PYG{p}{)}

\PYG{n}{dp\PYGZus{}checkpoint} \PYG{o}{=} \PYG{n}{dp\PYGZus{}app}\PYG{o}{.}\PYG{n}{save\PYGZus{}model}\PYG{p}{(}\PYG{p}{)}
\PYG{n}{shutil}\PYG{o}{.}\PYG{n}{move}\PYG{p}{(}\PYG{n}{dp\PYGZus{}checkpoint}\PYG{p}{,} \PYG{n}{DP\PYGZus{}STORE\PYGZus{}PATH}\PYG{p}{)}
\end{sphinxVerbatim}

\begin{sphinxVerbatim}[commandchars=\\\{\}]
\PYGZsq{}networks/dp\PYGZhy{}models/bench/checkpoint\PYGZus{}00010000\PYGZsq{}
\end{sphinxVerbatim}

Alternatively, uncomment the code in the cell below to load an existing network model.

\begin{sphinxVerbatim}[commandchars=\\\{\}]
\PYG{c+c1}{\PYGZsh{} dp\PYGZus{}path = \PYGZsq{}PDE\PYGZhy{}Control\PYGZhy{}RL/networks/dp\PYGZhy{}models/bench/checkpoint\PYGZus{}00020000/\PYGZsq{}}
\PYG{c+c1}{\PYGZsh{} networks\PYGZus{}to\PYGZus{}load = [\PYGZsq{}OP2\PYGZsq{}, \PYGZsq{}OP4\PYGZsq{}, \PYGZsq{}OP8\PYGZsq{}, \PYGZsq{}OP16\PYGZsq{}, \PYGZsq{}OP32\PYGZsq{}]}

\PYG{c+c1}{\PYGZsh{} dp\PYGZus{}app.load\PYGZus{}checkpoints(\PYGZob{}net: dp\PYGZus{}path for net in networks\PYGZus{}to\PYGZus{}load\PYGZcb{})}
\end{sphinxVerbatim}

Similar to the RL version, the next cell plots an example to visually show how well the DP\sphinxhyphen{}based model does. The leftmost plot again shows the learned results, this time of the DP\sphinxhyphen{}based model. Like above, the other two show the ground truth and the natural evolution.

\begin{sphinxVerbatim}[commandchars=\\\{\}]
\PYG{n}{dp\PYGZus{}frames} \PYG{o}{=} \PYG{n}{dp\PYGZus{}app}\PYG{o}{.}\PYG{n}{infer\PYGZus{}all\PYGZus{}frames}\PYG{p}{(}\PYG{n}{TEST\PYGZus{}RANGE}\PYG{p}{)}
\PYG{n}{dp\PYGZus{}frames} \PYG{o}{=} \PYG{p}{[}\PYG{n}{s}\PYG{o}{.}\PYG{n}{burgers}\PYG{o}{.}\PYG{n}{velocity}\PYG{o}{.}\PYG{n}{data} \PYG{k}{for} \PYG{n}{s} \PYG{o+ow}{in} \PYG{n}{dp\PYGZus{}frames}\PYG{p}{]}
\PYG{n}{\PYGZus{}}\PYG{p}{,} \PYG{n}{gt\PYGZus{}frames}\PYG{p}{,} \PYG{n}{unc\PYGZus{}frames} \PYG{o}{=} \PYG{n}{rl\PYGZus{}trainer}\PYG{o}{.}\PYG{n}{infer\PYGZus{}test\PYGZus{}set\PYGZus{}frames}\PYG{p}{(}\PYG{p}{)}

\PYG{n}{TEST\PYGZus{}SAMPLE} \PYG{o}{=} \PYG{l+m+mi}{0}    \PYG{c+c1}{\PYGZsh{} Change this to display a reconstruction of another scene}
\PYG{n}{fig}\PYG{p}{,} \PYG{n}{axs} \PYG{o}{=} \PYG{n}{plt}\PYG{o}{.}\PYG{n}{subplots}\PYG{p}{(}\PYG{l+m+mi}{1}\PYG{p}{,} \PYG{l+m+mi}{3}\PYG{p}{,} \PYG{n}{figsize}\PYG{o}{=}\PYG{p}{(}\PYG{l+m+mf}{18.9}\PYG{p}{,} \PYG{l+m+mf}{9.6}\PYG{p}{)}\PYG{p}{)}

\PYG{n}{axs}\PYG{p}{[}\PYG{l+m+mi}{0}\PYG{p}{]}\PYG{o}{.}\PYG{n}{set\PYGZus{}title}\PYG{p}{(}\PYG{l+s+s2}{\PYGZdq{}}\PYG{l+s+s2}{Differentiable Physics}\PYG{l+s+s2}{\PYGZdq{}}\PYG{p}{)}
\PYG{n}{axs}\PYG{p}{[}\PYG{l+m+mi}{1}\PYG{p}{]}\PYG{o}{.}\PYG{n}{set\PYGZus{}title}\PYG{p}{(}\PYG{l+s+s2}{\PYGZdq{}}\PYG{l+s+s2}{Ground Truth}\PYG{l+s+s2}{\PYGZdq{}}\PYG{p}{)}
\PYG{n}{axs}\PYG{p}{[}\PYG{l+m+mi}{2}\PYG{p}{]}\PYG{o}{.}\PYG{n}{set\PYGZus{}title}\PYG{p}{(}\PYG{l+s+s2}{\PYGZdq{}}\PYG{l+s+s2}{Uncontrolled}\PYG{l+s+s2}{\PYGZdq{}}\PYG{p}{)}

\PYG{k}{for} \PYG{n}{plot} \PYG{o+ow}{in} \PYG{n}{axs}\PYG{p}{:}
    \PYG{n}{plot}\PYG{o}{.}\PYG{n}{set\PYGZus{}ylim}\PYG{p}{(}\PYG{o}{\PYGZhy{}}\PYG{l+m+mi}{2}\PYG{p}{,} \PYG{l+m+mi}{2}\PYG{p}{)}
    \PYG{n}{plot}\PYG{o}{.}\PYG{n}{set\PYGZus{}xlabel}\PYG{p}{(}\PYG{l+s+s1}{\PYGZsq{}}\PYG{l+s+s1}{x}\PYG{l+s+s1}{\PYGZsq{}}\PYG{p}{)}
    \PYG{n}{plot}\PYG{o}{.}\PYG{n}{set\PYGZus{}ylabel}\PYG{p}{(}\PYG{l+s+s1}{\PYGZsq{}}\PYG{l+s+s1}{u(x)}\PYG{l+s+s1}{\PYGZsq{}}\PYG{p}{)}

\PYG{k}{for} \PYG{n}{frame} \PYG{o+ow}{in} \PYG{n+nb}{range}\PYG{p}{(}\PYG{l+m+mi}{0}\PYG{p}{,} \PYG{n}{STEP\PYGZus{}COUNT} \PYG{o}{+} \PYG{l+m+mi}{1}\PYG{p}{)}\PYG{p}{:}
    \PYG{n}{frame\PYGZus{}color} \PYG{o}{=} \PYG{n}{bplt}\PYG{o}{.}\PYG{n}{gradient\PYGZus{}color}\PYG{p}{(}\PYG{n}{frame}\PYG{p}{,} \PYG{n}{STEP\PYGZus{}COUNT}\PYG{o}{+}\PYG{l+m+mi}{1}\PYG{p}{)}
    \PYG{n}{axs}\PYG{p}{[}\PYG{l+m+mi}{0}\PYG{p}{]}\PYG{o}{.}\PYG{n}{plot}\PYG{p}{(}\PYG{n}{dp\PYGZus{}frames}\PYG{p}{[}\PYG{n}{frame}\PYG{p}{]}\PYG{p}{[}\PYG{n}{TEST\PYGZus{}SAMPLE}\PYG{p}{,}\PYG{p}{:}\PYG{p}{]}\PYG{p}{,} \PYG{n}{color}\PYG{o}{=}\PYG{n}{frame\PYGZus{}color}\PYG{p}{,} \PYG{n}{linewidth}\PYG{o}{=}\PYG{l+m+mf}{0.8}\PYG{p}{)}
    \PYG{n}{axs}\PYG{p}{[}\PYG{l+m+mi}{1}\PYG{p}{]}\PYG{o}{.}\PYG{n}{plot}\PYG{p}{(}\PYG{n}{gt\PYGZus{}frames}\PYG{p}{[}\PYG{n}{frame}\PYG{p}{]}\PYG{p}{[}\PYG{n}{TEST\PYGZus{}SAMPLE}\PYG{p}{,}\PYG{p}{:}\PYG{p}{]}\PYG{p}{,} \PYG{n}{color}\PYG{o}{=}\PYG{n}{frame\PYGZus{}color}\PYG{p}{,} \PYG{n}{linewidth}\PYG{o}{=}\PYG{l+m+mf}{0.8}\PYG{p}{)}
    \PYG{n}{axs}\PYG{p}{[}\PYG{l+m+mi}{2}\PYG{p}{]}\PYG{o}{.}\PYG{n}{plot}\PYG{p}{(}\PYG{n}{unc\PYGZus{}frames}\PYG{p}{[}\PYG{n}{frame}\PYG{p}{]}\PYG{p}{[}\PYG{n}{TEST\PYGZus{}SAMPLE}\PYG{p}{,}\PYG{p}{:}\PYG{p}{]}\PYG{p}{,} \PYG{n}{color}\PYG{o}{=}\PYG{n}{frame\PYGZus{}color}\PYG{p}{,} \PYG{n}{linewidth}\PYG{o}{=}\PYG{l+m+mf}{0.8}\PYG{p}{)}
\end{sphinxVerbatim}

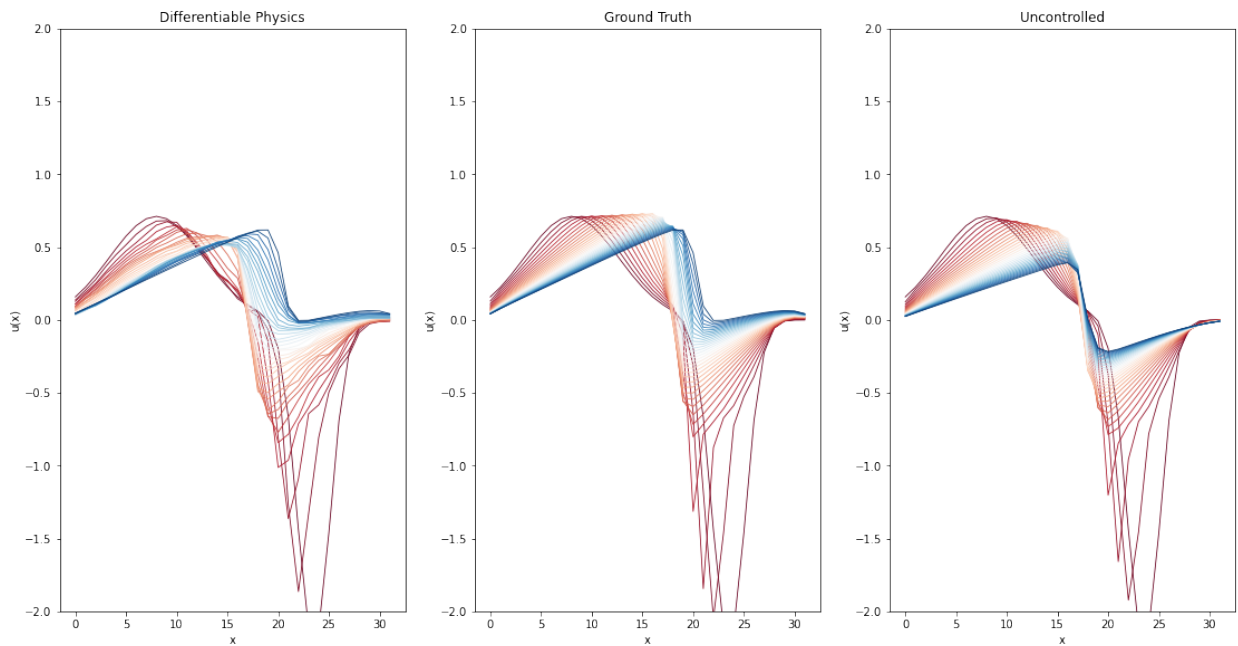
\begin{sphinxVerbatim}[commandchars=\\\{\}]
  \PYGZdq{}See the documentation of nn.Upsample for details.\PYGZdq{}.format(mode)
\end{sphinxVerbatim}

\noindent\sphinxincludegraphics{{reinflearn-code_34_1}.png}

The trained DP model also reconstructs the original trajectories closely. Furthermore, the generated results seem less noisy than using the RL agent.

With this, we have an RL and a DP version, which we can compare in more detail in the next section.

\bigskip\hrule\bigskip

\section{Comparison between RL and DP}
\label{\detokenize{reinflearn-code:comparison-between-rl-and-dp}}
Next, the results of both methods are compared in terms of visual quality of the resulting trajectories as well as quantitatively via the amount of generated forces. The latter provides insights about the performance of either approaches as both methods aspire to minimize this metric during training. This is also important as the task is trivially solved by applying a huge force at the last time step. Therefore, an ideal solution takes into account the dynamics of the PDE to apply as little forces as possible. Hence, this metric is a very good one to measure how well the network has learned about the underlying physical environment (Burgers equation in this example).

\begin{sphinxVerbatim}[commandchars=\\\{\}]
\PYG{k+kn}{import} \PYG{n+nn}{utils}
\PYG{k+kn}{import} \PYG{n+nn}{pandas} \PYG{k}{as} \PYG{n+nn}{pd}
\end{sphinxVerbatim}

\subsection{Trajectory comparison}
\label{\detokenize{reinflearn-code:trajectory-comparison}}
To compare the resulting trajectories, we generate trajectories from the test set with either method. Also, we collect the ground truth simulations and the natural evolution of the test set fields.

\begin{sphinxVerbatim}[commandchars=\\\{\}]
\PYG{n}{rl\PYGZus{}frames}\PYG{p}{,} \PYG{n}{gt\PYGZus{}frames}\PYG{p}{,} \PYG{n}{unc\PYGZus{}frames} \PYG{o}{=} \PYG{n}{rl\PYGZus{}trainer}\PYG{o}{.}\PYG{n}{infer\PYGZus{}test\PYGZus{}set\PYGZus{}frames}\PYG{p}{(}\PYG{p}{)}

\PYG{n}{dp\PYGZus{}frames} \PYG{o}{=} \PYG{n}{dp\PYGZus{}app}\PYG{o}{.}\PYG{n}{infer\PYGZus{}all\PYGZus{}frames}\PYG{p}{(}\PYG{n}{TEST\PYGZus{}RANGE}\PYG{p}{)}
\PYG{n}{dp\PYGZus{}frames} \PYG{o}{=} \PYG{p}{[}\PYG{n}{s}\PYG{o}{.}\PYG{n}{burgers}\PYG{o}{.}\PYG{n}{velocity}\PYG{o}{.}\PYG{n}{data} \PYG{k}{for} \PYG{n}{s} \PYG{o+ow}{in} \PYG{n}{dp\PYGZus{}frames}\PYG{p}{]}

\PYG{n}{frames} \PYG{o}{=} \PYG{p}{\PYGZob{}}
    \PYG{p}{(}\PYG{l+m+mi}{0}\PYG{p}{,} \PYG{l+m+mi}{0}\PYG{p}{)}\PYG{p}{:} \PYG{p}{(}\PYG{l+s+s1}{\PYGZsq{}}\PYG{l+s+s1}{Ground Truth}\PYG{l+s+s1}{\PYGZsq{}}\PYG{p}{,} \PYG{n}{gt\PYGZus{}frames}\PYG{p}{)}\PYG{p}{,}
    \PYG{p}{(}\PYG{l+m+mi}{0}\PYG{p}{,} \PYG{l+m+mi}{1}\PYG{p}{)}\PYG{p}{:} \PYG{p}{(}\PYG{l+s+s1}{\PYGZsq{}}\PYG{l+s+s1}{Uncontrolled}\PYG{l+s+s1}{\PYGZsq{}}\PYG{p}{,} \PYG{n}{unc\PYGZus{}frames}\PYG{p}{)}\PYG{p}{,}
    \PYG{p}{(}\PYG{l+m+mi}{1}\PYG{p}{,} \PYG{l+m+mi}{0}\PYG{p}{)}\PYG{p}{:} \PYG{p}{(}\PYG{l+s+s1}{\PYGZsq{}}\PYG{l+s+s1}{Reinforcement Learning}\PYG{l+s+s1}{\PYGZsq{}}\PYG{p}{,} \PYG{n}{rl\PYGZus{}frames}\PYG{p}{)}\PYG{p}{,}
    \PYG{p}{(}\PYG{l+m+mi}{1}\PYG{p}{,} \PYG{l+m+mi}{1}\PYG{p}{)}\PYG{p}{:} \PYG{p}{(}\PYG{l+s+s1}{\PYGZsq{}}\PYG{l+s+s1}{Differentiable Physics}\PYG{l+s+s1}{\PYGZsq{}}\PYG{p}{,} \PYG{n}{dp\PYGZus{}frames}\PYG{p}{)}\PYG{p}{,}
\PYG{p}{\PYGZcb{}}
\end{sphinxVerbatim}

\begin{sphinxVerbatim}[commandchars=\\\{\}]
  \PYGZdq{}See the documentation of nn.Upsample for details.\PYGZdq{}.format(mode)
\end{sphinxVerbatim}

\begin{sphinxVerbatim}[commandchars=\\\{\}]
\PYG{n}{TEST\PYGZus{}SAMPLE} \PYG{o}{=} \PYG{l+m+mi}{0}  \PYG{c+c1}{\PYGZsh{} Specifies which sample of the test set should be displayed}

\PYG{k}{def} \PYG{n+nf}{plot}\PYG{p}{(}\PYG{n}{axs}\PYG{p}{,} \PYG{n}{xy}\PYG{p}{,} \PYG{n}{title}\PYG{p}{,} \PYG{n}{field}\PYG{p}{)}\PYG{p}{:}
    \PYG{n}{axs}\PYG{p}{[}\PYG{n}{xy}\PYG{p}{]}\PYG{o}{.}\PYG{n}{set\PYGZus{}ylim}\PYG{p}{(}\PYG{o}{\PYGZhy{}}\PYG{l+m+mi}{2}\PYG{p}{,} \PYG{l+m+mi}{2}\PYG{p}{)}\PYG{p}{;} \PYG{n}{axs}\PYG{p}{[}\PYG{n}{xy}\PYG{p}{]}\PYG{o}{.}\PYG{n}{set\PYGZus{}title}\PYG{p}{(}\PYG{n}{title}\PYG{p}{)}
    \PYG{n}{axs}\PYG{p}{[}\PYG{n}{xy}\PYG{p}{]}\PYG{o}{.}\PYG{n}{set\PYGZus{}xlabel}\PYG{p}{(}\PYG{l+s+s1}{\PYGZsq{}}\PYG{l+s+s1}{x}\PYG{l+s+s1}{\PYGZsq{}}\PYG{p}{)}\PYG{p}{;} \PYG{n}{axs}\PYG{p}{[}\PYG{n}{xy}\PYG{p}{]}\PYG{o}{.}\PYG{n}{set\PYGZus{}ylabel}\PYG{p}{(}\PYG{l+s+s1}{\PYGZsq{}}\PYG{l+s+s1}{u(x)}\PYG{l+s+s1}{\PYGZsq{}}\PYG{p}{)}
    \PYG{n}{label} \PYG{o}{=} \PYG{l+s+s1}{\PYGZsq{}}\PYG{l+s+s1}{Initial state (red), final state (blue)}\PYG{l+s+s1}{\PYGZsq{}}
    \PYG{k}{for} \PYG{n}{step\PYGZus{}idx} \PYG{o+ow}{in} \PYG{n+nb}{range}\PYG{p}{(}\PYG{l+m+mi}{0}\PYG{p}{,} \PYG{n}{STEP\PYGZus{}COUNT} \PYG{o}{+} \PYG{l+m+mi}{1}\PYG{p}{)}\PYG{p}{:}
        \PYG{n}{color} \PYG{o}{=} \PYG{n}{bplt}\PYG{o}{.}\PYG{n}{gradient\PYGZus{}color}\PYG{p}{(}\PYG{n}{step\PYGZus{}idx}\PYG{p}{,} \PYG{n}{STEP\PYGZus{}COUNT}\PYG{o}{+}\PYG{l+m+mi}{1}\PYG{p}{)}
        \PYG{n}{axs}\PYG{p}{[}\PYG{n}{xy}\PYG{p}{]}\PYG{o}{.}\PYG{n}{plot}\PYG{p}{(}
            \PYG{n}{field}\PYG{p}{[}\PYG{n}{step\PYGZus{}idx}\PYG{p}{]}\PYG{p}{[}\PYG{n}{TEST\PYGZus{}SAMPLE}\PYG{p}{]}\PYG{o}{.}\PYG{n}{squeeze}\PYG{p}{(}\PYG{p}{)}\PYG{p}{,} \PYG{n}{color}\PYG{o}{=}\PYG{n}{color}\PYG{p}{,} \PYG{n}{linewidth}\PYG{o}{=}\PYG{l+m+mf}{0.8}\PYG{p}{,} \PYG{n}{label}\PYG{o}{=}\PYG{n}{label}\PYG{p}{)}
        \PYG{n}{label} \PYG{o}{=} \PYG{k+kc}{None}
    \PYG{n}{axs}\PYG{p}{[}\PYG{n}{xy}\PYG{p}{]}\PYG{o}{.}\PYG{n}{legend}\PYG{p}{(}\PYG{p}{)}

\PYG{n}{fig}\PYG{p}{,} \PYG{n}{axs} \PYG{o}{=} \PYG{n}{plt}\PYG{o}{.}\PYG{n}{subplots}\PYG{p}{(}\PYG{l+m+mi}{2}\PYG{p}{,} \PYG{l+m+mi}{2}\PYG{p}{,} \PYG{n}{figsize}\PYG{o}{=}\PYG{p}{(}\PYG{l+m+mf}{12.8}\PYG{p}{,} \PYG{l+m+mf}{9.6}\PYG{p}{)}\PYG{p}{)}
\PYG{k}{for} \PYG{n}{xy} \PYG{o+ow}{in} \PYG{n}{frames}\PYG{p}{:}
    \PYG{n}{plot}\PYG{p}{(}\PYG{n}{axs}\PYG{p}{,} \PYG{n}{xy}\PYG{p}{,} \PYG{o}{*}\PYG{n}{frames}\PYG{p}{[}\PYG{n}{xy}\PYG{p}{]}\PYG{p}{)}
\end{sphinxVerbatim}

\noindent\sphinxincludegraphics{{reinflearn-code_40_0}.png}

This diagram connects the two plots shown above after each training. Here we again see that the differentiable physics approach seems to generate less noisy trajectories than the RL agent, while both manage to approximate the ground truth.

\subsection{Comparison of exerted forces}
\label{\detokenize{reinflearn-code:comparison-of-exerted-forces}}
Next, we compute the forces the approaches have generated and applied for the test set trajectories.

\begin{sphinxVerbatim}[commandchars=\\\{\}]
\PYG{n}{gt\PYGZus{}forces} \PYG{o}{=} \PYG{n}{utils}\PYG{o}{.}\PYG{n}{infer\PYGZus{}forces\PYGZus{}sum\PYGZus{}from\PYGZus{}frames}\PYG{p}{(}
    \PYG{n}{gt\PYGZus{}frames}\PYG{p}{,} \PYG{n}{DOMAIN}\PYG{p}{,} \PYG{n}{DIFFUSION\PYGZus{}SUBSTEPS}\PYG{p}{,} \PYG{n}{VISCOSITY}\PYG{p}{,} \PYG{n}{DT}
\PYG{p}{)}
\PYG{n}{dp\PYGZus{}forces} \PYG{o}{=} \PYG{n}{utils}\PYG{o}{.}\PYG{n}{infer\PYGZus{}forces\PYGZus{}sum\PYGZus{}from\PYGZus{}frames}\PYG{p}{(}
    \PYG{n}{dp\PYGZus{}frames}\PYG{p}{,} \PYG{n}{DOMAIN}\PYG{p}{,} \PYG{n}{DIFFUSION\PYGZus{}SUBSTEPS}\PYG{p}{,} \PYG{n}{VISCOSITY}\PYG{p}{,} \PYG{n}{DT}
\PYG{p}{)}
\PYG{n}{rl\PYGZus{}forces} \PYG{o}{=} \PYG{n}{rl\PYGZus{}trainer}\PYG{o}{.}\PYG{n}{infer\PYGZus{}test\PYGZus{}set\PYGZus{}forces}\PYG{p}{(}\PYG{p}{)}
\end{sphinxVerbatim}

\begin{sphinxVerbatim}[commandchars=\\\{\}]
Sanity check \PYGZhy{} maximum deviation from target state: 0.000000
Sanity check \PYGZhy{} maximum deviation from target state: 0.000011
\end{sphinxVerbatim}

\begin{sphinxVerbatim}[commandchars=\\\{\}]
  \PYGZdq{}See the documentation of nn.Upsample for details.\PYGZdq{}.format(mode)
\end{sphinxVerbatim}

At first, we will compare the total sum of the forces that are generated by the RL and DP approaches and compare them to the ground truth.

\begin{sphinxVerbatim}[commandchars=\\\{\}]
\PYG{n}{plt}\PYG{o}{.}\PYG{n}{figure}\PYG{p}{(}\PYG{n}{figsize}\PYG{o}{=}\PYG{p}{(}\PYG{l+m+mi}{8}\PYG{p}{,} \PYG{l+m+mi}{6}\PYG{p}{)}\PYG{p}{)}
\PYG{n}{plt}\PYG{o}{.}\PYG{n}{bar}\PYG{p}{(}
    \PYG{p}{[}\PYG{l+s+s2}{\PYGZdq{}}\PYG{l+s+s2}{Reinforcement Learning}\PYG{l+s+s2}{\PYGZdq{}}\PYG{p}{,} \PYG{l+s+s2}{\PYGZdq{}}\PYG{l+s+s2}{Differentiable Physics}\PYG{l+s+s2}{\PYGZdq{}}\PYG{p}{,} \PYG{l+s+s2}{\PYGZdq{}}\PYG{l+s+s2}{Ground Truth}\PYG{l+s+s2}{\PYGZdq{}}\PYG{p}{]}\PYG{p}{,} 
    \PYG{p}{[}\PYG{n}{np}\PYG{o}{.}\PYG{n}{sum}\PYG{p}{(}\PYG{n}{rl\PYGZus{}forces}\PYG{p}{)}\PYG{p}{,} \PYG{n}{np}\PYG{o}{.}\PYG{n}{sum}\PYG{p}{(}\PYG{n}{dp\PYGZus{}forces}\PYG{p}{)}\PYG{p}{,} \PYG{n}{np}\PYG{o}{.}\PYG{n}{sum}\PYG{p}{(}\PYG{n}{gt\PYGZus{}forces}\PYG{p}{)}\PYG{p}{]}\PYG{p}{,} 
    \PYG{n}{color} \PYG{o}{=} \PYG{p}{[}\PYG{l+s+s2}{\PYGZdq{}}\PYG{l+s+s2}{\PYGZsh{}0065bd}\PYG{l+s+s2}{\PYGZdq{}}\PYG{p}{,} \PYG{l+s+s2}{\PYGZdq{}}\PYG{l+s+s2}{\PYGZsh{}e37222}\PYG{l+s+s2}{\PYGZdq{}}\PYG{p}{,} \PYG{l+s+s2}{\PYGZdq{}}\PYG{l+s+s2}{\PYGZsh{}a2ad00}\PYG{l+s+s2}{\PYGZdq{}}\PYG{p}{]}\PYG{p}{,}
    \PYG{n}{align}\PYG{o}{=}\PYG{l+s+s1}{\PYGZsq{}}\PYG{l+s+s1}{center}\PYG{l+s+s1}{\PYGZsq{}}\PYG{p}{,} \PYG{n}{label}\PYG{o}{=}\PYG{l+s+s1}{\PYGZsq{}}\PYG{l+s+s1}{Absolute forces comparison}\PYG{l+s+s1}{\PYGZsq{}} \PYG{p}{)}
\end{sphinxVerbatim}

\begin{sphinxVerbatim}[commandchars=\\\{\}]
\PYGZlt{}BarContainer object of 3 artists\PYGZgt{}
\end{sphinxVerbatim}

\noindent\sphinxincludegraphics{{reinflearn-code_45_1}.png}

As visualized in these bar plots, the DP approach learns to apply slightly lower forces than the RL model.
As both methods are on\sphinxhyphen{}par in terms of how well they reach the final target states, this is the main quantity we use to compare the performance of both methods.

In the following, the forces generated by the methods are also visually compared to the ground truth of the respective sample. Dots placed above the blue line denote stronger forces in the analyzed deep learning approach than in the ground truth and vice versa.

\begin{sphinxVerbatim}[commandchars=\\\{\}]
\PYG{n}{plt}\PYG{o}{.}\PYG{n}{figure}\PYG{p}{(}\PYG{n}{figsize}\PYG{o}{=}\PYG{p}{(}\PYG{l+m+mi}{12}\PYG{p}{,} \PYG{l+m+mi}{9}\PYG{p}{)}\PYG{p}{)}
\PYG{n}{plt}\PYG{o}{.}\PYG{n}{scatter}\PYG{p}{(}\PYG{n}{gt\PYGZus{}forces}\PYG{p}{,} \PYG{n}{rl\PYGZus{}forces}\PYG{p}{,} \PYG{n}{color}\PYG{o}{=}\PYG{l+s+s2}{\PYGZdq{}}\PYG{l+s+s2}{\PYGZsh{}0065bd}\PYG{l+s+s2}{\PYGZdq{}}\PYG{p}{,} \PYG{n}{label}\PYG{o}{=}\PYG{l+s+s1}{\PYGZsq{}}\PYG{l+s+s1}{RL}\PYG{l+s+s1}{\PYGZsq{}}\PYG{p}{)}
\PYG{n}{plt}\PYG{o}{.}\PYG{n}{scatter}\PYG{p}{(}\PYG{n}{gt\PYGZus{}forces}\PYG{p}{,} \PYG{n}{dp\PYGZus{}forces}\PYG{p}{,} \PYG{n}{color}\PYG{o}{=}\PYG{l+s+s2}{\PYGZdq{}}\PYG{l+s+s2}{\PYGZsh{}e37222}\PYG{l+s+s2}{\PYGZdq{}}\PYG{p}{,} \PYG{n}{label}\PYG{o}{=}\PYG{l+s+s1}{\PYGZsq{}}\PYG{l+s+s1}{DP}\PYG{l+s+s1}{\PYGZsq{}}\PYG{p}{)}
\PYG{n}{plt}\PYG{o}{.}\PYG{n}{plot}\PYG{p}{(}\PYG{p}{[}\PYG{n}{x} \PYG{o}{*} \PYG{l+m+mi}{100} \PYG{k}{for} \PYG{n}{x} \PYG{o+ow}{in} \PYG{n+nb}{range}\PYG{p}{(}\PYG{l+m+mi}{15}\PYG{p}{)}\PYG{p}{]}\PYG{p}{,} \PYG{p}{[}\PYG{n}{x} \PYG{o}{*} \PYG{l+m+mi}{100} \PYG{k}{for} \PYG{n}{x} \PYG{o+ow}{in} \PYG{n+nb}{range}\PYG{p}{(}\PYG{l+m+mi}{15}\PYG{p}{)}\PYG{p}{]}\PYG{p}{,} \PYG{n}{color}\PYG{o}{=}\PYG{l+s+s2}{\PYGZdq{}}\PYG{l+s+s2}{\PYGZsh{}a2ad00}\PYG{l+s+s2}{\PYGZdq{}}\PYG{p}{,} \PYG{n}{label}\PYG{o}{=}\PYG{l+s+s1}{\PYGZsq{}}\PYG{l+s+s1}{Same forces as original}\PYG{l+s+s1}{\PYGZsq{}}\PYG{p}{)}
\PYG{n}{plt}\PYG{o}{.}\PYG{n}{xlabel}\PYG{p}{(}\PYG{l+s+s1}{\PYGZsq{}}\PYG{l+s+s1}{ground truth}\PYG{l+s+s1}{\PYGZsq{}}\PYG{p}{)}\PYG{p}{;} \PYG{n}{plt}\PYG{o}{.}\PYG{n}{ylabel}\PYG{p}{(}\PYG{l+s+s1}{\PYGZsq{}}\PYG{l+s+s1}{reconstruction}\PYG{l+s+s1}{\PYGZsq{}}\PYG{p}{)}
\PYG{n}{plt}\PYG{o}{.}\PYG{n}{xlim}\PYG{p}{(}\PYG{l+m+mi}{0}\PYG{p}{,} \PYG{l+m+mi}{1200}\PYG{p}{)}\PYG{p}{;} \PYG{n}{plt}\PYG{o}{.}\PYG{n}{ylim}\PYG{p}{(}\PYG{l+m+mi}{0}\PYG{p}{,} \PYG{l+m+mi}{1200}\PYG{p}{)}\PYG{p}{;} \PYG{n}{plt}\PYG{o}{.}\PYG{n}{grid}\PYG{p}{(}\PYG{p}{)}\PYG{p}{;} \PYG{n}{plt}\PYG{o}{.}\PYG{n}{legend}\PYG{p}{(}\PYG{p}{)}
\end{sphinxVerbatim}

\begin{sphinxVerbatim}[commandchars=\\\{\}]
\PYGZlt{}matplotlib.legend.Legend at 0x7f4cbc6d5090\PYGZgt{}
\end{sphinxVerbatim}

\noindent\sphinxincludegraphics{{reinflearn-code_47_1}.png}

The graph shows that the orange dots of the DP training run are in general closer to the diagonal \sphinxhyphen{} i.e., this network learned to generate forces that are closer to the ground truth values.

The following plot displays the performance of all reinforcement learning, differentiable physics and ground truth with respect to individual samples.

\begin{sphinxVerbatim}[commandchars=\\\{\}]
\PYG{n}{w}\PYG{o}{=}\PYG{l+m+mf}{0.25}\PYG{p}{;} \PYG{n}{plot\PYGZus{}count}\PYG{o}{=}\PYG{l+m+mi}{20}   \PYG{c+c1}{\PYGZsh{} How many scenes to show}
\PYG{n}{plt}\PYG{o}{.}\PYG{n}{figure}\PYG{p}{(}\PYG{n}{figsize}\PYG{o}{=}\PYG{p}{(}\PYG{l+m+mf}{12.8}\PYG{p}{,} \PYG{l+m+mf}{9.6}\PYG{p}{)}\PYG{p}{)}
\PYG{n}{plt}\PYG{o}{.}\PYG{n}{bar}\PYG{p}{(} \PYG{p}{[}\PYG{n}{i} \PYG{o}{\PYGZhy{}} \PYG{n}{w} \PYG{k}{for} \PYG{n}{i} \PYG{o+ow}{in} \PYG{n+nb}{range}\PYG{p}{(}\PYG{n}{plot\PYGZus{}count}\PYG{p}{)}\PYG{p}{]}\PYG{p}{,} \PYG{n}{rl\PYGZus{}forces}\PYG{p}{[}\PYG{p}{:}\PYG{n}{plot\PYGZus{}count}\PYG{p}{]}\PYG{p}{,} \PYG{n}{color}\PYG{o}{=}\PYG{l+s+s2}{\PYGZdq{}}\PYG{l+s+s2}{\PYGZsh{}0065bd}\PYG{l+s+s2}{\PYGZdq{}}\PYG{p}{,} \PYG{n}{width}\PYG{o}{=}\PYG{n}{w}\PYG{p}{,} \PYG{n}{align}\PYG{o}{=}\PYG{l+s+s1}{\PYGZsq{}}\PYG{l+s+s1}{center}\PYG{l+s+s1}{\PYGZsq{}}\PYG{p}{,} \PYG{n}{label}\PYG{o}{=}\PYG{l+s+s1}{\PYGZsq{}}\PYG{l+s+s1}{RL}\PYG{l+s+s1}{\PYGZsq{}} \PYG{p}{)}
\PYG{n}{plt}\PYG{o}{.}\PYG{n}{bar}\PYG{p}{(} \PYG{p}{[}\PYG{n}{i}     \PYG{k}{for} \PYG{n}{i} \PYG{o+ow}{in} \PYG{n+nb}{range}\PYG{p}{(}\PYG{n}{plot\PYGZus{}count}\PYG{p}{)}\PYG{p}{]}\PYG{p}{,} \PYG{n}{dp\PYGZus{}forces}\PYG{p}{[}\PYG{p}{:}\PYG{n}{plot\PYGZus{}count}\PYG{p}{]}\PYG{p}{,} \PYG{n}{color}\PYG{o}{=}\PYG{l+s+s2}{\PYGZdq{}}\PYG{l+s+s2}{\PYGZsh{}e37222}\PYG{l+s+s2}{\PYGZdq{}}\PYG{p}{,} \PYG{n}{width}\PYG{o}{=}\PYG{n}{w}\PYG{p}{,} \PYG{n}{align}\PYG{o}{=}\PYG{l+s+s1}{\PYGZsq{}}\PYG{l+s+s1}{center}\PYG{l+s+s1}{\PYGZsq{}}\PYG{p}{,} \PYG{n}{label}\PYG{o}{=}\PYG{l+s+s1}{\PYGZsq{}}\PYG{l+s+s1}{DP}\PYG{l+s+s1}{\PYGZsq{}} \PYG{p}{)}
\PYG{n}{plt}\PYG{o}{.}\PYG{n}{bar}\PYG{p}{(} \PYG{p}{[}\PYG{n}{i} \PYG{o}{+} \PYG{n}{w} \PYG{k}{for} \PYG{n}{i} \PYG{o+ow}{in} \PYG{n+nb}{range}\PYG{p}{(}\PYG{n}{plot\PYGZus{}count}\PYG{p}{)}\PYG{p}{]}\PYG{p}{,} \PYG{n}{gt\PYGZus{}forces}\PYG{p}{[}\PYG{p}{:}\PYG{n}{plot\PYGZus{}count}\PYG{p}{]}\PYG{p}{,} \PYG{n}{color}\PYG{o}{=}\PYG{l+s+s2}{\PYGZdq{}}\PYG{l+s+s2}{\PYGZsh{}a2ad00}\PYG{l+s+s2}{\PYGZdq{}}\PYG{p}{,} \PYG{n}{width}\PYG{o}{=}\PYG{n}{w}\PYG{p}{,} \PYG{n}{align}\PYG{o}{=}\PYG{l+s+s1}{\PYGZsq{}}\PYG{l+s+s1}{center}\PYG{l+s+s1}{\PYGZsq{}}\PYG{p}{,} \PYG{n}{label}\PYG{o}{=}\PYG{l+s+s1}{\PYGZsq{}}\PYG{l+s+s1}{GT}\PYG{l+s+s1}{\PYGZsq{}} \PYG{p}{)}
\PYG{n}{plt}\PYG{o}{.}\PYG{n}{xlabel}\PYG{p}{(}\PYG{l+s+s1}{\PYGZsq{}}\PYG{l+s+s1}{Scenes}\PYG{l+s+s1}{\PYGZsq{}}\PYG{p}{)}\PYG{p}{;} \PYG{n}{plt}\PYG{o}{.}\PYG{n}{xticks}\PYG{p}{(}\PYG{n+nb}{range}\PYG{p}{(}\PYG{n}{plot\PYGZus{}count}\PYG{p}{)}\PYG{p}{)}
\PYG{n}{plt}\PYG{o}{.}\PYG{n}{ylabel}\PYG{p}{(}\PYG{l+s+s1}{\PYGZsq{}}\PYG{l+s+s1}{Forces}\PYG{l+s+s1}{\PYGZsq{}}\PYG{p}{)}\PYG{p}{;} \PYG{n}{plt}\PYG{o}{.}\PYG{n}{legend}\PYG{p}{(}\PYG{p}{)}\PYG{p}{;} \PYG{n}{plt}\PYG{o}{.}\PYG{n}{show}\PYG{p}{(}\PYG{p}{)}
\end{sphinxVerbatim}

\noindent\sphinxincludegraphics{{reinflearn-code_50_0}.png}

\section{Training progress comparison}
\label{\detokenize{reinflearn-code:training-progress-comparison}}
Although the quality of the control in terms of force magnitudes is the primary goal of the setup above, there are interesting differences in terms of how both methods behave at training time. The main difference of the physics\sphinxhyphen{}unaware RL training and the DP approach with its tightly coupled solver is that the latter results in a significantly faster convergence. I.e., the gradients provided by the numerical solver give a much better learning signal than the undirected exploration of the RL process. The behavior of the RL training, on the other hand, can in part be ascribed to the on\sphinxhyphen{}policy nature of training data collection and to the “brute\sphinxhyphen{}force” exploration of the reinforcement learning technique.

The next cell visualizes the training progress of both methods with respect to wall time.

\begin{sphinxVerbatim}[commandchars=\\\{\}]
\PYG{k}{def} \PYG{n+nf}{get\PYGZus{}dp\PYGZus{}val\PYGZus{}set\PYGZus{}forces}\PYG{p}{(}\PYG{n}{experiment\PYGZus{}path}\PYG{p}{)}\PYG{p}{:}
    \PYG{n}{path} \PYG{o}{=} \PYG{n}{os}\PYG{o}{.}\PYG{n}{path}\PYG{o}{.}\PYG{n}{join}\PYG{p}{(}\PYG{n}{experiment\PYGZus{}path}\PYG{p}{,} \PYG{l+s+s1}{\PYGZsq{}}\PYG{l+s+s1}{val\PYGZus{}forces.csv}\PYG{l+s+s1}{\PYGZsq{}}\PYG{p}{)}
    \PYG{n}{table} \PYG{o}{=} \PYG{n}{pd}\PYG{o}{.}\PYG{n}{read\PYGZus{}csv}\PYG{p}{(}\PYG{n}{path}\PYG{p}{)}
    \PYG{k}{return} \PYG{n+nb}{list}\PYG{p}{(}\PYG{n}{table}\PYG{p}{[}\PYG{l+s+s1}{\PYGZsq{}}\PYG{l+s+s1}{time}\PYG{l+s+s1}{\PYGZsq{}}\PYG{p}{]}\PYG{p}{)}\PYG{p}{,} \PYG{n+nb}{list}\PYG{p}{(}\PYG{n}{table}\PYG{p}{[}\PYG{l+s+s1}{\PYGZsq{}}\PYG{l+s+s1}{epoch}\PYG{l+s+s1}{\PYGZsq{}}\PYG{p}{]}\PYG{p}{)}\PYG{p}{,} \PYG{n+nb}{list}\PYG{p}{(}\PYG{n}{table}\PYG{p}{[}\PYG{l+s+s1}{\PYGZsq{}}\PYG{l+s+s1}{forces}\PYG{l+s+s1}{\PYGZsq{}}\PYG{p}{]}\PYG{p}{)}

\PYG{n}{rl\PYGZus{}w\PYGZus{}times}\PYG{p}{,} \PYG{n}{rl\PYGZus{}step\PYGZus{}nums}\PYG{p}{,} \PYG{n}{rl\PYGZus{}val\PYGZus{}forces} \PYG{o}{=} \PYG{n}{rl\PYGZus{}trainer}\PYG{o}{.}\PYG{n}{get\PYGZus{}val\PYGZus{}set\PYGZus{}forces\PYGZus{}data}\PYG{p}{(}\PYG{p}{)}
\PYG{n}{dp\PYGZus{}w\PYGZus{}times}\PYG{p}{,} \PYG{n}{dp\PYGZus{}epochs}\PYG{p}{,} \PYG{n}{dp\PYGZus{}val\PYGZus{}forces} \PYG{o}{=} \PYG{n}{get\PYGZus{}dp\PYGZus{}val\PYGZus{}set\PYGZus{}forces}\PYG{p}{(}\PYG{n}{DP\PYGZus{}STORE\PYGZus{}PATH}\PYG{p}{)} 

\PYG{n}{plt}\PYG{o}{.}\PYG{n}{figure}\PYG{p}{(}\PYG{n}{figsize}\PYG{o}{=}\PYG{p}{(}\PYG{l+m+mi}{12}\PYG{p}{,} \PYG{l+m+mi}{5}\PYG{p}{)}\PYG{p}{)}
\PYG{n}{plt}\PYG{o}{.}\PYG{n}{plot}\PYG{p}{(}\PYG{n}{np}\PYG{o}{.}\PYG{n}{array}\PYG{p}{(}\PYG{n}{rl\PYGZus{}w\PYGZus{}times}\PYG{p}{)} \PYG{o}{/} \PYG{l+m+mi}{3600}\PYG{p}{,} \PYG{n}{rl\PYGZus{}val\PYGZus{}forces}\PYG{p}{,} \PYG{n}{color}\PYG{o}{=}\PYG{l+s+s2}{\PYGZdq{}}\PYG{l+s+s2}{\PYGZsh{}0065bd}\PYG{l+s+s2}{\PYGZdq{}}\PYG{p}{,} \PYG{n}{label}\PYG{o}{=}\PYG{l+s+s1}{\PYGZsq{}}\PYG{l+s+s1}{RL}\PYG{l+s+s1}{\PYGZsq{}}\PYG{p}{)}
\PYG{n}{plt}\PYG{o}{.}\PYG{n}{plot}\PYG{p}{(}\PYG{n}{np}\PYG{o}{.}\PYG{n}{array}\PYG{p}{(}\PYG{n}{dp\PYGZus{}w\PYGZus{}times}\PYG{p}{)} \PYG{o}{/} \PYG{l+m+mi}{3600}\PYG{p}{,} \PYG{n}{dp\PYGZus{}val\PYGZus{}forces}\PYG{p}{,} \PYG{n}{color}\PYG{o}{=}\PYG{l+s+s2}{\PYGZdq{}}\PYG{l+s+s2}{\PYGZsh{}e37222}\PYG{l+s+s2}{\PYGZdq{}}\PYG{p}{,} \PYG{n}{label}\PYG{o}{=}\PYG{l+s+s1}{\PYGZsq{}}\PYG{l+s+s1}{DP}\PYG{l+s+s1}{\PYGZsq{}}\PYG{p}{)}
\PYG{n}{plt}\PYG{o}{.}\PYG{n}{xlabel}\PYG{p}{(}\PYG{l+s+s1}{\PYGZsq{}}\PYG{l+s+s1}{Wall time (hours)}\PYG{l+s+s1}{\PYGZsq{}}\PYG{p}{)}\PYG{p}{;} \PYG{n}{plt}\PYG{o}{.}\PYG{n}{ylabel}\PYG{p}{(}\PYG{l+s+s1}{\PYGZsq{}}\PYG{l+s+s1}{Forces}\PYG{l+s+s1}{\PYGZsq{}}\PYG{p}{)}
\PYG{n}{plt}\PYG{o}{.}\PYG{n}{ylim}\PYG{p}{(}\PYG{l+m+mi}{0}\PYG{p}{,} \PYG{l+m+mi}{1500}\PYG{p}{)}\PYG{p}{;} \PYG{n}{plt}\PYG{o}{.}\PYG{n}{grid}\PYG{p}{(}\PYG{p}{)}\PYG{p}{;} \PYG{n}{plt}\PYG{o}{.}\PYG{n}{legend}\PYG{p}{(}\PYG{p}{)}
\end{sphinxVerbatim}

\begin{sphinxVerbatim}[commandchars=\\\{\}]
\PYGZlt{}matplotlib.legend.Legend at 0x7f4cbc595c50\PYGZgt{}
\end{sphinxVerbatim}

\noindent\sphinxincludegraphics{{reinflearn-code_52_1}.png}

To conclude, the PPO reinforcement learning exerts higher forces in comparison to the differentiable physics approach. Hence, PPO yields a learned solution with slightly inferior quality. Additionally, the time needed for convergence is significantly higher in the RL case (both in terms of wall time and training iterations).

\section{Next steps}
\label{\detokenize{reinflearn-code:next-steps}}\begin{itemize}
\item {} 
See how different values for hyperparameters, such as learning rate, influence the training process

\item {} 
Work with fields of different resolution and see how the two approaches then compare to each other. Larger resolutions make the physical dynamics more complex, and hence harder to control

\item {} 
Use trained models in settings with different environment parameters (e.g. viscosity, dt) and test how well they generalize

\end{itemize}

\part{PBDL and Uncertainty}

\chapter{Introduction to Posterior Inference}
\label{\detokenize{bayesian-intro:introduction-to-posterior-inference}}\label{\detokenize{bayesian-intro::doc}}
We should keep in mind that for all measurements, models, and discretizations we have uncertainties. For measurements and observations, this typically appears in the form of measurement errors. Model equations equations, on the other hand, usually encompass only parts of a system we’re interested in (leaving the remainder as an uncertainty), while for numerical simulations we inherently introduce discretization errors. So a very important question to ask here is how we can be sure that an answer we obtain is the correct one. From a statisticians viewpoint, we’d like to know the posterior pobability distribution, a distribution that captures possible uncertainties we have about our model or data.

\section{Uncertainty}
\label{\detokenize{bayesian-intro:uncertainty}}
This admittedly becomes even more difficult in the context of machine learning:
we’re typically facing the task of approximating complex and unknown functions.
From a probabilistic perspective, the standard process of training an NN here
yields a \sphinxstyleemphasis{maximum likelihood estimation} (MLE) for the parameters of the network.
However, this MLE viewpoint does not take any of the uncertainties mentioned above into account:
for DL training, we likewise have a numerical optimization, and hence an inherent
approximation error and uncertainty regarding the learned representation.
Ideally, we should reformulate our the learning process such that it takes
its own uncertainties into account, and it should make
\sphinxstyleemphasis{posterior inference} possible,
i.e. learn to produce the full output distribution. However, this turns out to be an
extremely difficult task.

This is where so\sphinxhyphen{}called \sphinxstyleemphasis{Bayesian neural network} (BNN) approaches come into play. They
allow for a form of posterior inference by making assumptions about the probability
distributions of individual parameters of the network. This gives a distribution for the
parameters, with which we can evaluate the network multiple times to obtain different versions
of the output, and in this way sample the distribution of the output.

Nonetheless, the task
remains very challenging. Training a BNN is typically significantly more difficult
than training a regular NN. This should come as no surprise, as we’re trying to
learn something fundamentally different here: a full probability distribution
instead of a point estimate. (All previous chapters “just” dealt with
learning such point estimates, and the tasks were still far from trivial.)

\begin{sphinxadmonition}{note}{Aleatoric and Epistemic Uncertainty}

Although we won’t go into detail within the scope of this book, many works
distinguish two types of uncertainty which are important to mention here:
\begin{itemize}
\item {} 
\sphinxstyleemphasis{Aleatoric} uncertainty denotes uncertainty within the data, e.g., noise in measurements.

\item {} 
\sphinxstyleemphasis{Epistemic} uncertainty, on the other hand, describes uncertainties within a model such as a trained neural network.

\end{itemize}

In the following we’ll primarily target \sphinxstyleemphasis{epistemic} uncertainty via posterior inference.
However, as a word of caution: if they appear together, the different kinds of uncertainties (the two types above are not exhaustive) are very difficult to disentangle in practice.
\end{sphinxadmonition}

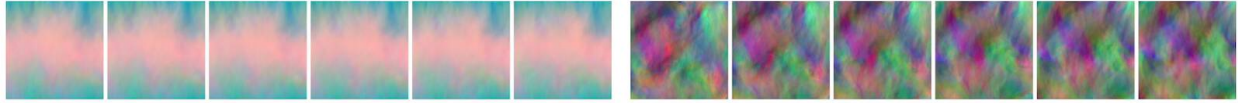
\sphinxincludegraphics{{divider5}.jpg}

\section{Introduction to Bayesian Neural Networks}
\label{\detokenize{bayesian-intro:introduction-to-bayesian-neural-networks}}
In order to combine posterior inference with Neural Networks, we can use standard techniques from Bayesian Modeling and combine them with the Deep Learning machinery. In Bayesian Modeling, we aim at learning \sphinxstyleemphasis{distributions} over the model parameters instead of these fixed point estimates. In the case of NNs, the model parameters are the weights and biases, summarized by \(\theta\), for a neural network \(f\). Our goal is therefore to learn a so\sphinxhyphen{}called \sphinxstyleemphasis{posterior distribution} \(p({\theta}|{D})\) from the data \(D\), which captures the uncertainty we have about the networks weights and biases \sphinxstyleemphasis{after} observing the data \(D\). This posterior distribution is the central quantity of interest here: if we can estimate it reasonably well, we can use it to make good predictions, but also assess uncertainties related to those predictions. For both objectives it is necessary to \sphinxstyleemphasis{marginalize} over the posterior distribution, i.e. integrate it out. A single prediction for input \(x_{i}\) can for example be obtained via
\begin{equation*}
\begin{split}
    \hat{y_{i}}=\int f(x_{i}; \theta) ~ p(\theta|D) ~ d\theta
\end{split}
\end{equation*}
Similarly, one can for instance compute the standard deviation in order to assess a measure of uncertainty over the prediction \(\hat{y_{i}}\).

\section{Prior distributions}
\label{\detokenize{bayesian-intro:prior-distributions}}
In order to obtain the required posterior distribution, in Bayesian modeling one first has to define a \sphinxstyleemphasis{prior distribution} \(p({\theta})\) over the network parameters. This prior distribution should encompass the knowledge we have about the network weights \sphinxstyleemphasis{before} training the model. We for instance know that the weights of neural networks are usually rather small and can be both positive and negative. Centered normal distributions with some small variance parameter are therefore a standard choice. For computational simplicity, they are typically also assumed to be independent from another. When observing data \({D}\), the prior distribution over the weights is updated to the  posterior according to Bayes rule:
\begin{equation*}
\begin{split}
    p({\theta}|{D}) = \frac{p({D}|{\theta})p({\theta})}{p({D})}
    \text{ . }
\end{split}
\end{equation*}
This is, we update our a\sphinxhyphen{}priori knowledge after observing data, i.e. we \sphinxstyleemphasis{learn} from data. The computation required for the Bayesian update is usually intractable, especially when dealing with non\sphinxhyphen{}trivial network architectures. Therefore, the posterior \(p({\theta}|{D})\) is approximated with an easy\sphinxhyphen{}to\sphinxhyphen{}evaluate variational distribution \(q_{\phi}(\theta)\), parametrized by \(\phi\). Again, independent Normal distributions are typically used for each weight. Hence, the parameters \(\phi\) contain all mean and variance parameters \(\mu, \sigma\) of those normal distributions.
The optimization goal is then to find a distribution \(q_{\phi}(\theta)\) that is close to the true posterior.
One way of assessing this closeness is the KL\sphinxhyphen{}divergence, a method used widely in practice for measuring the similarity of two distributions.

\section{Evidence lower bound}
\label{\detokenize{bayesian-intro:evidence-lower-bound}}
We cannot directly minimize the KL\sphinxhyphen{}divergence between approximate and true posterior \(KL(q_{\phi}({\theta})||p({\theta|D})\), because we do not have access to the true posterior distribution. It is however possible to show that one can equivalently maximize the so\sphinxhyphen{}called evidence lower bound (ELBO), a quantity well known from variational inference:
\begin{equation*}
\begin{split}
    \mathcal{L}(\phi)=  E_{q_{\phi}}[\log(p(D|{\theta}))] - KL(q_{\phi}({\theta})||p({\theta})) 
    \text{ , }
\end{split}
\end{equation*}
The ELBO (or negative ELBO, if one prefers to minimize instead of maximize) is the optimization objective for BNNs. The first term is an expected log\sphinxhyphen{}likelihood of the data. Maximizing it means explaining the data as well as possible. In practice, the log\sphinxhyphen{}likelihood is often a conventional loss functions like mean squared error (note that MSE can be seen as negative log\sphinxhyphen{}likelihood for normal noise with unit variance). The second term is the negative KL\sphinxhyphen{}divergence between the approximate posterior and the prior. For suitable prior and approximate posterior choices (like the ones above), this term can be computed analytically. Maximizing it means encouraging the approximate network weight distributions to stay close to the prior distribution. In that sense, the two terms of the ELBO  have opposite goals: The first term encourages the model to explain the data as well as possible, whereas the second term encourages the model to stay close to the (random) prior distributions, which implies randomness and regularization.

The expectation of the log\sphinxhyphen{}likelihood is typically not available in analytical form, but can be approximated in several ways. One can, for instance, use Monte\sphinxhyphen{}Carlo sampling and draw \(S\) samples from \(q_{\phi}({\theta})\). The expectation is then approximated via \(\frac{1}{S}\sum_{s=1}^{S}\log(p(D|{\theta_{s}})\). In practice, even a single sample, i.e. \(S=1\) can be enough. Furthermore, the expectation of the log\sphinxhyphen{}likelihood is typically not evaluated on the whole dataset, but approximated by a batch of data \(D_{batch}\), which enables the use of batch\sphinxhyphen{}wise stochastic gradient descent.

\section{The BNN training loop}
\label{\detokenize{bayesian-intro:the-bnn-training-loop}}
For \(S=1\), one iteration of the training loop then boils down to
\begin{enumerate}
\sphinxsetlistlabels{\arabic}{enumi}{enumii}{}{.}%
\item {} 
sampling a batch of data \(D_{batch}\)

\item {} 
sampling network weights and biases from \(q_{\phi}({\theta})\)

\item {} 
forwarding the batch of data through the network according to the sampled weights

\item {} 
evaluating the ELBO

\item {} 
backpropagating the ELBO through the network and updating \(\phi\)

\end{enumerate}

If \(S>1\), steps 2 and 3 have to be repeated \(S\) times in order to compute the ELBO. In that sense, training a variational BNN is fairly similar to training a conventional NN: We still use SGD and forward\sphinxhyphen{}backward passes to optimize our loss function. Only that now we are optimizing over distributions instead of single values. If you are curious on how one can backpropagate through distributions, you can, e.g., read about it \sphinxhref{https://arxiv.org/abs/1505.05424}{here}, and a more detailed introduction to Bayesian Neural Networks is available in Y. Gals \sphinxhref{https://mlg.eng.cam.ac.uk/yarin/thesis/thesis.pdf}{thesis}, in chapters 2 \& 3.

\section{Dropout as alternative}
\label{\detokenize{bayesian-intro:dropout-as-alternative}}
Previous work has shown that using dropout is mathematically equivalent to an approximation to the probabilistic deep Gaussian process. Furthermore, for a specific prior (satisfying the so\sphinxhyphen{}called KL\sphinxhyphen{}condition) and specific approximate posterior choice (a product of Bernoulli distributions), training a neural network with dropout and L2 regularization and training a variational BNN results in an equivalent optimization procedure. In other words, dropout neural networks are a form of Bayesian neural networks.

Obtaining uncertainty estimates is then as easy as training a conventional neural network with dropout and extending dropout, which traditionally has only been used during the training phase, to the prediction phase. This will lead to the predictions being random (because dropout will randomly drop some of the activations), which allows us to compute average and standard deviation statistics for single samples from the dataset, just like for the variational BNN case.
It is an ongoing discussion in the field whether variational or dropout\sphinxhyphen{}based methods are preferable.

\section{A practical example}
\label{\detokenize{bayesian-intro:a-practical-example}}
As a first real example for posterior inference with variational BNNs, let’s revisit the
case of turbulent flows around airfoils, from {\hyperref[\detokenize{supervised-airfoils::doc}]{\sphinxcrossref{\DUrole{doc}{Supervised training for RANS flows around airfoils}}}}.
However, in contrast to the point estimate learned in this section, we’ll now aim for
learning the full posterior.

\chapter{RANS Airfoil Flows with Bayesian Neural Nets}
\label{\detokenize{bayesian-code:rans-airfoil-flows-with-bayesian-neural-nets}}\label{\detokenize{bayesian-code::doc}}

\section{Overview}
\label{\detokenize{bayesian-code:overview}}
We are now considering the same setup as in the notebook {\hyperref[\detokenize{supervised-airfoils::doc}]{\sphinxcrossref{\DUrole{doc}{Supervised training for RANS flows around airfoils}}}}: A turbulent airflow around wing profiles, for which we’d like to know the average motion
and pressure distribution around this airfoil for different Reynolds numbers and angles of attack. In the earlier notebook, we tackled this by completely bypassing any physical solver and instead training a neural network that learns the quantities of interest. Now, we want to extend this approach to the variational Bayesian Neural Networks (BNNs) of the previous section. In contrast to traditional networks, that learn a single point estimate for each weight value, BNNs aim at learning a \sphinxstyleemphasis{distribution} over each weight parameter (e.g. a Gaussian with mean \(\mu\) and variance \(\sigma^{2}\)). During a forward\sphinxhyphen{}pass, each parameter in the network is then sampled from its corresponding approximate posterior distribution \(q_{\phi}(\theta)\). In that sense, the network parameters themselves are \sphinxstyleemphasis{random variables} and each forward pass becomes \sphinxstyleemphasis{stochastic}, because for a given input the predictions will vary with every forward\sphinxhyphen{}pass. This allows to assess how \sphinxstyleemphasis{uncertain} the network is: If the predictions vary a lot, we think that the network is uncertain about its output. \sphinxhref{https://colab.research.google.com/github/tum-pbs/pbdl-book/blob/main/bayesian-code.ipynb}{{[}run in colab{]}}

\subsection{Read in Data}
\label{\detokenize{bayesian-code:read-in-data}}
Like in the previous notebook we’ll skip the data generation process. This example is adapted from \sphinxhref{https://github.com/thunil/Deep-Flow-Prediction}{the Deep\sphinxhyphen{}Flow\sphinxhyphen{}Prediction codebase}, which you can check out for details. Here, we’ll simply download a small set of training data generated with a Spalart\sphinxhyphen{}Almaras RANS simulation in \sphinxhref{https://openfoam.org/}{OpenFOAM}.

\begin{sphinxVerbatim}[commandchars=\\\{\}]
\PYG{k+kn}{import} \PYG{n+nn}{numpy} \PYG{k}{as} \PYG{n+nn}{np}
\PYG{k+kn}{import} \PYG{n+nn}{os}\PYG{n+nn}{.}\PYG{n+nn}{path}\PYG{o}{,} \PYG{n+nn}{random}

\PYG{c+c1}{\PYGZsh{} get training data: as in the previous supervised example, either download or use gdrive}
\PYG{n+nb}{dir} \PYG{o}{=} \PYG{l+s+s2}{\PYGZdq{}}\PYG{l+s+s2}{./}\PYG{l+s+s2}{\PYGZdq{}}
\PYG{k}{if} \PYG{k+kc}{True}\PYG{p}{:}
    \PYG{k}{if} \PYG{o+ow}{not} \PYG{n}{os}\PYG{o}{.}\PYG{n}{path}\PYG{o}{.}\PYG{n}{isfile}\PYG{p}{(}\PYG{l+s+s1}{\PYGZsq{}}\PYG{l+s+s1}{data\PYGZhy{}airfoils.npz}\PYG{l+s+s1}{\PYGZsq{}}\PYG{p}{)}\PYG{p}{:}
        \PYG{k+kn}{import} \PYG{n+nn}{requests}
        \PYG{n+nb}{print}\PYG{p}{(}\PYG{l+s+s2}{\PYGZdq{}}\PYG{l+s+s2}{Downloading training data (300MB), this can take a few minutes the first time...}\PYG{l+s+s2}{\PYGZdq{}}\PYG{p}{)}
        \PYG{k}{with} \PYG{n+nb}{open}\PYG{p}{(}\PYG{l+s+s2}{\PYGZdq{}}\PYG{l+s+s2}{data\PYGZhy{}airfoils.npz}\PYG{l+s+s2}{\PYGZdq{}}\PYG{p}{,} \PYG{l+s+s1}{\PYGZsq{}}\PYG{l+s+s1}{wb}\PYG{l+s+s1}{\PYGZsq{}}\PYG{p}{)} \PYG{k}{as} \PYG{n}{datafile}\PYG{p}{:}
            \PYG{n}{resp} \PYG{o}{=} \PYG{n}{requests}\PYG{o}{.}\PYG{n}{get}\PYG{p}{(}\PYG{l+s+s1}{\PYGZsq{}}\PYG{l+s+s1}{https://dataserv.ub.tum.de/s/m1615239/download?path=}\PYG{l+s+si}{\PYGZpc{}2F}\PYG{l+s+s1}{\PYGZam{}files=dfp\PYGZhy{}data\PYGZhy{}400.npz}\PYG{l+s+s1}{\PYGZsq{}}\PYG{p}{,} \PYG{n}{verify}\PYG{o}{=}\PYG{k+kc}{False}\PYG{p}{)}
            \PYG{n}{datafile}\PYG{o}{.}\PYG{n}{write}\PYG{p}{(}\PYG{n}{resp}\PYG{o}{.}\PYG{n}{content}\PYG{p}{)}
\PYG{k}{else}\PYG{p}{:} \PYG{c+c1}{\PYGZsh{} cf supervised airfoil code:}
    \PYG{k+kn}{from} \PYG{n+nn}{google}\PYG{n+nn}{.}\PYG{n+nn}{colab} \PYG{k+kn}{import} \PYG{n}{drive}
    \PYG{n}{drive}\PYG{o}{.}\PYG{n}{mount}\PYG{p}{(}\PYG{l+s+s1}{\PYGZsq{}}\PYG{l+s+s1}{/content/gdrive}\PYG{l+s+s1}{\PYGZsq{}}\PYG{p}{)}
    \PYG{n+nb}{dir} \PYG{o}{=} \PYG{l+s+s2}{\PYGZdq{}}\PYG{l+s+s2}{./gdrive/My Drive/}\PYG{l+s+s2}{\PYGZdq{}}

\PYG{n}{npfile}\PYG{o}{=}\PYG{n}{np}\PYG{o}{.}\PYG{n}{load}\PYG{p}{(}\PYG{n+nb}{dir}\PYG{o}{+}\PYG{l+s+s1}{\PYGZsq{}}\PYG{l+s+s1}{data\PYGZhy{}airfoils.npz}\PYG{l+s+s1}{\PYGZsq{}}\PYG{p}{)}
\PYG{n+nb}{print}\PYG{p}{(}\PYG{l+s+s2}{\PYGZdq{}}\PYG{l+s+s2}{Loaded data, }\PYG{l+s+si}{\PYGZob{}\PYGZcb{}}\PYG{l+s+s2}{ training, }\PYG{l+s+si}{\PYGZob{}\PYGZcb{}}\PYG{l+s+s2}{ validation samples}\PYG{l+s+s2}{\PYGZdq{}}\PYG{o}{.}\PYG{n}{format}\PYG{p}{(}\PYG{n+nb}{len}\PYG{p}{(}\PYG{n}{npfile}\PYG{p}{[}\PYG{l+s+s2}{\PYGZdq{}}\PYG{l+s+s2}{inputs}\PYG{l+s+s2}{\PYGZdq{}}\PYG{p}{]}\PYG{p}{)}\PYG{p}{,}\PYG{n+nb}{len}\PYG{p}{(}\PYG{n}{npfile}\PYG{p}{[}\PYG{l+s+s2}{\PYGZdq{}}\PYG{l+s+s2}{vinputs}\PYG{l+s+s2}{\PYGZdq{}}\PYG{p}{]}\PYG{p}{)}\PYG{p}{)}\PYG{p}{)}
\PYG{n+nb}{print}\PYG{p}{(}\PYG{l+s+s2}{\PYGZdq{}}\PYG{l+s+s2}{Size of the inputs array: }\PYG{l+s+s2}{\PYGZdq{}}\PYG{o}{+}\PYG{n+nb}{format}\PYG{p}{(}\PYG{n}{npfile}\PYG{p}{[}\PYG{l+s+s2}{\PYGZdq{}}\PYG{l+s+s2}{inputs}\PYG{l+s+s2}{\PYGZdq{}}\PYG{p}{]}\PYG{o}{.}\PYG{n}{shape}\PYG{p}{)}\PYG{p}{)}

\PYG{c+c1}{\PYGZsh{} reshape to channels\PYGZus{}last for convencience}
\PYG{n}{X\PYGZus{}train} \PYG{o}{=} \PYG{n}{np}\PYG{o}{.}\PYG{n}{moveaxis}\PYG{p}{(}\PYG{n}{npfile}\PYG{p}{[}\PYG{l+s+s2}{\PYGZdq{}}\PYG{l+s+s2}{inputs}\PYG{l+s+s2}{\PYGZdq{}}\PYG{p}{]}\PYG{p}{,}\PYG{l+m+mi}{1}\PYG{p}{,}\PYG{o}{\PYGZhy{}}\PYG{l+m+mi}{1}\PYG{p}{)}
\PYG{n}{y\PYGZus{}train} \PYG{o}{=} \PYG{n}{np}\PYG{o}{.}\PYG{n}{moveaxis}\PYG{p}{(}\PYG{n}{npfile}\PYG{p}{[}\PYG{l+s+s2}{\PYGZdq{}}\PYG{l+s+s2}{targets}\PYG{l+s+s2}{\PYGZdq{}}\PYG{p}{]}\PYG{p}{,}\PYG{l+m+mi}{1}\PYG{p}{,}\PYG{o}{\PYGZhy{}}\PYG{l+m+mi}{1}\PYG{p}{)}

\PYG{n}{X\PYGZus{}val} \PYG{o}{=} \PYG{n}{np}\PYG{o}{.}\PYG{n}{moveaxis}\PYG{p}{(}\PYG{n}{npfile}\PYG{p}{[}\PYG{l+s+s2}{\PYGZdq{}}\PYG{l+s+s2}{vinputs}\PYG{l+s+s2}{\PYGZdq{}}\PYG{p}{]}\PYG{p}{,}\PYG{l+m+mi}{1}\PYG{p}{,}\PYG{o}{\PYGZhy{}}\PYG{l+m+mi}{1}\PYG{p}{)}
\PYG{n}{y\PYGZus{}val} \PYG{o}{=} \PYG{n}{np}\PYG{o}{.}\PYG{n}{moveaxis}\PYG{p}{(}\PYG{n}{npfile}\PYG{p}{[}\PYG{l+s+s2}{\PYGZdq{}}\PYG{l+s+s2}{vtargets}\PYG{l+s+s2}{\PYGZdq{}}\PYG{p}{]}\PYG{p}{,}\PYG{l+m+mi}{1}\PYG{p}{,}\PYG{o}{\PYGZhy{}}\PYG{l+m+mi}{1}\PYG{p}{)}
\end{sphinxVerbatim}

\begin{sphinxVerbatim}[commandchars=\\\{\}]
Downloading training data (300MB), this can take a few minutes the first time...
\end{sphinxVerbatim}

\begin{sphinxVerbatim}[commandchars=\\\{\}]

\end{sphinxVerbatim}

\begin{sphinxVerbatim}[commandchars=\\\{\}]
Loaded data, 320 training, 80 validation samples
Size of the inputs array: (320, 3, 128, 128)
\end{sphinxVerbatim}

\subsection{Look at Data}
\label{\detokenize{bayesian-code:look-at-data}}
Now we have some training data. We can look at it using the code we also used in the original notebook:

\begin{sphinxVerbatim}[commandchars=\\\{\}]
\PYG{k+kn}{import} \PYG{n+nn}{pylab}
\PYG{k+kn}{from} \PYG{n+nn}{matplotlib} \PYG{k+kn}{import} \PYG{n}{cm}

\PYG{c+c1}{\PYGZsh{} helper to show three target channels: normalized, with colormap, side by side}
\PYG{k}{def} \PYG{n+nf}{showSbs}\PYG{p}{(}\PYG{n}{a1}\PYG{p}{,}\PYG{n}{a2}\PYG{p}{,} \PYG{n}{bottom}\PYG{o}{=}\PYG{l+s+s2}{\PYGZdq{}}\PYG{l+s+s2}{NN Output}\PYG{l+s+s2}{\PYGZdq{}}\PYG{p}{,} \PYG{n}{top}\PYG{o}{=}\PYG{l+s+s2}{\PYGZdq{}}\PYG{l+s+s2}{Reference}\PYG{l+s+s2}{\PYGZdq{}}\PYG{p}{,} \PYG{n}{title}\PYG{o}{=}\PYG{k+kc}{None}\PYG{p}{)}\PYG{p}{:} 
  \PYG{n}{c}\PYG{o}{=}\PYG{p}{[}\PYG{p}{]}
  \PYG{k}{for} \PYG{n}{i} \PYG{o+ow}{in} \PYG{n+nb}{range}\PYG{p}{(}\PYG{l+m+mi}{3}\PYG{p}{)}\PYG{p}{:}
    \PYG{n}{b} \PYG{o}{=} \PYG{n}{np}\PYG{o}{.}\PYG{n}{flipud}\PYG{p}{(} \PYG{n}{np}\PYG{o}{.}\PYG{n}{concatenate}\PYG{p}{(}\PYG{p}{(}\PYG{n}{a2}\PYG{p}{[}\PYG{o}{.}\PYG{o}{.}\PYG{o}{.}\PYG{p}{,}\PYG{n}{i}\PYG{p}{]}\PYG{p}{,}\PYG{n}{a1}\PYG{p}{[}\PYG{o}{.}\PYG{o}{.}\PYG{o}{.}\PYG{p}{,}\PYG{n}{i}\PYG{p}{]}\PYG{p}{)}\PYG{p}{,}\PYG{n}{axis}\PYG{o}{=}\PYG{l+m+mi}{1}\PYG{p}{)}\PYG{o}{.}\PYG{n}{transpose}\PYG{p}{(}\PYG{p}{)}\PYG{p}{)}
    \PYG{n+nb}{min}\PYG{p}{,} \PYG{n}{mean}\PYG{p}{,} \PYG{n+nb}{max} \PYG{o}{=} \PYG{n}{np}\PYG{o}{.}\PYG{n}{min}\PYG{p}{(}\PYG{n}{b}\PYG{p}{)}\PYG{p}{,} \PYG{n}{np}\PYG{o}{.}\PYG{n}{mean}\PYG{p}{(}\PYG{n}{b}\PYG{p}{)}\PYG{p}{,} \PYG{n}{np}\PYG{o}{.}\PYG{n}{max}\PYG{p}{(}\PYG{n}{b}\PYG{p}{)}\PYG{p}{;} 
    \PYG{n}{b} \PYG{o}{\PYGZhy{}}\PYG{o}{=} \PYG{n+nb}{min}\PYG{p}{;} \PYG{n}{b} \PYG{o}{/}\PYG{o}{=} \PYG{p}{(}\PYG{n+nb}{max}\PYG{o}{\PYGZhy{}}\PYG{n+nb}{min}\PYG{p}{)}
    \PYG{n}{c}\PYG{o}{.}\PYG{n}{append}\PYG{p}{(}\PYG{n}{b}\PYG{p}{)}
  \PYG{n}{fig}\PYG{p}{,} \PYG{n}{axes} \PYG{o}{=} \PYG{n}{pylab}\PYG{o}{.}\PYG{n}{subplots}\PYG{p}{(}\PYG{l+m+mi}{1}\PYG{p}{,} \PYG{l+m+mi}{1}\PYG{p}{,} \PYG{n}{figsize}\PYG{o}{=}\PYG{p}{(}\PYG{l+m+mi}{16}\PYG{p}{,} \PYG{l+m+mi}{5}\PYG{p}{)}\PYG{p}{)}
  \PYG{n}{axes}\PYG{o}{.}\PYG{n}{set\PYGZus{}xticks}\PYG{p}{(}\PYG{p}{[}\PYG{p}{]}\PYG{p}{)}\PYG{p}{;} \PYG{n}{axes}\PYG{o}{.}\PYG{n}{set\PYGZus{}yticks}\PYG{p}{(}\PYG{p}{[}\PYG{p}{]}\PYG{p}{)}\PYG{p}{;} 
  \PYG{n}{im} \PYG{o}{=} \PYG{n}{axes}\PYG{o}{.}\PYG{n}{imshow}\PYG{p}{(}\PYG{n}{np}\PYG{o}{.}\PYG{n}{concatenate}\PYG{p}{(}\PYG{n}{c}\PYG{p}{,}\PYG{n}{axis}\PYG{o}{=}\PYG{l+m+mi}{1}\PYG{p}{)}\PYG{p}{,} \PYG{n}{origin}\PYG{o}{=}\PYG{l+s+s1}{\PYGZsq{}}\PYG{l+s+s1}{upper}\PYG{l+s+s1}{\PYGZsq{}}\PYG{p}{,} \PYG{n}{cmap}\PYG{o}{=}\PYG{l+s+s1}{\PYGZsq{}}\PYG{l+s+s1}{magma}\PYG{l+s+s1}{\PYGZsq{}}\PYG{p}{)}

  \PYG{n}{pylab}\PYG{o}{.}\PYG{n}{colorbar}\PYG{p}{(}\PYG{n}{im}\PYG{p}{)}\PYG{p}{;} \PYG{n}{pylab}\PYG{o}{.}\PYG{n}{xlabel}\PYG{p}{(}\PYG{l+s+s1}{\PYGZsq{}}\PYG{l+s+s1}{p, ux, uy}\PYG{l+s+s1}{\PYGZsq{}}\PYG{p}{)}\PYG{p}{;} \PYG{n}{pylab}\PYG{o}{.}\PYG{n}{ylabel}\PYG{p}{(}\PYG{l+s+s1}{\PYGZsq{}}\PYG{l+s+si}{\PYGZpc{}s}\PYG{l+s+s1}{           }\PYG{l+s+si}{\PYGZpc{}s}\PYG{l+s+s1}{\PYGZsq{}}\PYG{o}{\PYGZpc{}}\PYG{p}{(}\PYG{n}{bottom}\PYG{p}{,}\PYG{n}{top}\PYG{p}{)}\PYG{p}{)}
  \PYG{k}{if} \PYG{n}{title} \PYG{o+ow}{is} \PYG{o+ow}{not} \PYG{k+kc}{None}\PYG{p}{:} \PYG{n}{pylab}\PYG{o}{.}\PYG{n}{title}\PYG{p}{(}\PYG{n}{title}\PYG{p}{)}

\PYG{n}{NUM}\PYG{o}{=}\PYG{l+m+mi}{40}
\PYG{n+nb}{print}\PYG{p}{(}\PYG{l+s+s2}{\PYGZdq{}}\PYG{l+s+se}{\PYGZbs{}n}\PYG{l+s+s2}{Here are all 3 inputs are shown at the top (mask,in x, in y) }\PYG{l+s+se}{\PYGZbs{}n}\PYG{l+s+s2}{Side by side with the 3 output channels (p,vx,vy) at the bottom:}\PYG{l+s+s2}{\PYGZdq{}}\PYG{p}{)}
\PYG{n}{showSbs}\PYG{p}{(} \PYG{n}{X\PYGZus{}train}\PYG{p}{[}\PYG{n}{NUM}\PYG{p}{]}\PYG{p}{,}\PYG{n}{y\PYGZus{}train}\PYG{p}{[}\PYG{n}{NUM}\PYG{p}{]}\PYG{p}{,} \PYG{n}{bottom}\PYG{o}{=}\PYG{l+s+s2}{\PYGZdq{}}\PYG{l+s+s2}{Target Output}\PYG{l+s+s2}{\PYGZdq{}}\PYG{p}{,} \PYG{n}{top}\PYG{o}{=}\PYG{l+s+s2}{\PYGZdq{}}\PYG{l+s+s2}{Inputs}\PYG{l+s+s2}{\PYGZdq{}}\PYG{p}{,} \PYG{n}{title}\PYG{o}{=}\PYG{l+s+s2}{\PYGZdq{}}\PYG{l+s+s2}{Training data sample}\PYG{l+s+s2}{\PYGZdq{}}\PYG{p}{)}
\end{sphinxVerbatim}

\begin{sphinxVerbatim}[commandchars=\\\{\}]
Here are all 3 inputs are shown at the top (mask,in x, in y) 
Side by side with the 3 output channels (p,vx,vy) at the bottom:
\end{sphinxVerbatim}

\noindent\sphinxincludegraphics{{bayesian-code_4_1}.png}

Not surprisingly, the data still looks the same. For details, please check out the description in {\hyperref[\detokenize{supervised-airfoils::doc}]{\sphinxcrossref{\DUrole{doc}{Supervised training for RANS flows around airfoils}}}}.

\subsection{Neural Network Definition}
\label{\detokenize{bayesian-code:neural-network-definition}}
Now let’s look at how we can implement BNNs. Instead of PyTorch, we will use TensorFlow, in particular the extension TensorFlow Probability, which has easy\sphinxhyphen{}to\sphinxhyphen{}implement probabilistic layers. Like in the other notebook, we use a U\sphinxhyphen{}Net structure consisting of Convolutional blocks with skip\sphinxhyphen{}layer connections. For now, we only want to set up the decoder, i.e. second part of the U\sphinxhyphen{}Net as bayesian. For this, we will take advantage of TensorFlows \sphinxstyleemphasis{flipout} layers (in particular, the convolutional implementation).

In a forward pass, those layers automatically sample from the current posterior distribution and store the KL\sphinxhyphen{}divergence between prior and posterior in \sphinxstyleemphasis{model.losses}. One can specify the desired divergence measure (typically KL\sphinxhyphen{}divergence) and modify the prior and approximate posterior distributions, if other than normal distributions are desired. Other than that, the flipout layers can be used just like regular layers in sequential architectures. The code below implements a single convolutional block of the U\sphinxhyphen{}Net:

\begin{sphinxVerbatim}[commandchars=\\\{\}]
\PYG{k+kn}{import} \PYG{n+nn}{tensorflow} \PYG{k}{as} \PYG{n+nn}{tf}
\PYG{k+kn}{import} \PYG{n+nn}{tensorflow\PYGZus{}probability}\PYG{n+nn}{.}\PYG{n+nn}{python}\PYG{n+nn}{.}\PYG{n+nn}{distributions} \PYG{k}{as} \PYG{n+nn}{tfd}
\PYG{k+kn}{from} \PYG{n+nn}{tensorflow}\PYG{n+nn}{.}\PYG{n+nn}{keras} \PYG{k+kn}{import} \PYG{n}{Sequential}
\PYG{k+kn}{from} \PYG{n+nn}{tensorflow}\PYG{n+nn}{.}\PYG{n+nn}{keras}\PYG{n+nn}{.}\PYG{n+nn}{initializers} \PYG{k+kn}{import} \PYG{n}{RandomNormal}
\PYG{k+kn}{from} \PYG{n+nn}{tensorflow}\PYG{n+nn}{.}\PYG{n+nn}{keras}\PYG{n+nn}{.}\PYG{n+nn}{layers} \PYG{k+kn}{import} \PYG{n}{Input}\PYG{p}{,} \PYG{n}{Conv2D}\PYG{p}{,} \PYG{n}{Conv2DTranspose}\PYG{p}{,}\PYG{n}{UpSampling2D}\PYG{p}{,} \PYG{n}{BatchNormalization}\PYG{p}{,} \PYG{n}{ReLU}\PYG{p}{,} \PYG{n}{LeakyReLU}\PYG{p}{,} \PYG{n}{SpatialDropout2D}\PYG{p}{,} \PYG{n}{MaxPooling2D}
\PYG{k+kn}{from} \PYG{n+nn}{tensorflow\PYGZus{}probability}\PYG{n+nn}{.}\PYG{n+nn}{python}\PYG{n+nn}{.}\PYG{n+nn}{layers} \PYG{k+kn}{import} \PYG{n}{Convolution2DFlipout}
\PYG{k+kn}{from} \PYG{n+nn}{tensorflow}\PYG{n+nn}{.}\PYG{n+nn}{keras}\PYG{n+nn}{.}\PYG{n+nn}{models} \PYG{k+kn}{import} \PYG{n}{Model}

\PYG{k}{def} \PYG{n+nf}{tfBlockUnet}\PYG{p}{(}\PYG{n}{filters}\PYG{o}{=}\PYG{l+m+mi}{3}\PYG{p}{,} \PYG{n}{transposed}\PYG{o}{=}\PYG{k+kc}{False}\PYG{p}{,} \PYG{n}{kernel\PYGZus{}size}\PYG{o}{=}\PYG{l+m+mi}{4}\PYG{p}{,} \PYG{n}{bn}\PYG{o}{=}\PYG{k+kc}{True}\PYG{p}{,} \PYG{n}{relu}\PYG{o}{=}\PYG{k+kc}{True}\PYG{p}{,} \PYG{n}{pad}\PYG{o}{=}\PYG{l+s+s2}{\PYGZdq{}}\PYG{l+s+s2}{same}\PYG{l+s+s2}{\PYGZdq{}}\PYG{p}{,} \PYG{n}{dropout}\PYG{o}{=}\PYG{l+m+mf}{0.}\PYG{p}{,} \PYG{n}{flipout}\PYG{o}{=}\PYG{k+kc}{False}\PYG{p}{,}
              \PYG{n}{kdf}\PYG{o}{=}\PYG{k+kc}{None}\PYG{p}{,} \PYG{n}{name}\PYG{o}{=}\PYG{l+s+s1}{\PYGZsq{}}\PYG{l+s+s1}{\PYGZsq{}}\PYG{p}{)}\PYG{p}{:}
    \PYG{n}{block} \PYG{o}{=} \PYG{n}{Sequential}\PYG{p}{(}\PYG{n}{name}\PYG{o}{=}\PYG{n}{name}\PYG{p}{)}
    \PYG{k}{if} \PYG{n}{relu}\PYG{p}{:}
        \PYG{n}{block}\PYG{o}{.}\PYG{n}{add}\PYG{p}{(}\PYG{n}{ReLU}\PYG{p}{(}\PYG{p}{)}\PYG{p}{)}
    \PYG{k}{else}\PYG{p}{:}
        \PYG{n}{block}\PYG{o}{.}\PYG{n}{add}\PYG{p}{(}\PYG{n}{LeakyReLU}\PYG{p}{(}\PYG{l+m+mf}{0.2}\PYG{p}{)}\PYG{p}{)}
    \PYG{k}{if} \PYG{o+ow}{not} \PYG{n}{transposed}\PYG{p}{:}
        \PYG{n}{block}\PYG{o}{.}\PYG{n}{add}\PYG{p}{(}\PYG{n}{Conv2D}\PYG{p}{(}\PYG{n}{filters}\PYG{o}{=}\PYG{n}{filters}\PYG{p}{,} \PYG{n}{kernel\PYGZus{}size}\PYG{o}{=}\PYG{n}{kernel\PYGZus{}size}\PYG{p}{,} \PYG{n}{padding}\PYG{o}{=}\PYG{n}{pad}\PYG{p}{,}
                         \PYG{n}{kernel\PYGZus{}initializer}\PYG{o}{=}\PYG{n}{RandomNormal}\PYG{p}{(}\PYG{l+m+mf}{0.0}\PYG{p}{,} \PYG{l+m+mf}{0.02}\PYG{p}{)}\PYG{p}{,} \PYG{n}{activation}\PYG{o}{=}\PYG{k+kc}{None}\PYG{p}{,}\PYG{n}{strides}\PYG{o}{=}\PYG{p}{(}\PYG{l+m+mi}{2}\PYG{p}{,}\PYG{l+m+mi}{2}\PYG{p}{)}\PYG{p}{)}\PYG{p}{)}
    \PYG{k}{else}\PYG{p}{:}
        \PYG{n}{block}\PYG{o}{.}\PYG{n}{add}\PYG{p}{(}\PYG{n}{UpSampling2D}\PYG{p}{(}\PYG{n}{interpolation} \PYG{o}{=} \PYG{l+s+s1}{\PYGZsq{}}\PYG{l+s+s1}{bilinear}\PYG{l+s+s1}{\PYGZsq{}}\PYG{p}{)}\PYG{p}{)}
        \PYG{k}{if} \PYG{n}{flipout}\PYG{p}{:}
            \PYG{n}{block}\PYG{o}{.}\PYG{n}{add}\PYG{p}{(}\PYG{n}{Convolution2DFlipout}\PYG{p}{(}\PYG{n}{filters}\PYG{o}{=}\PYG{n}{filters}\PYG{p}{,} \PYG{n}{kernel\PYGZus{}size}\PYG{o}{=}\PYG{p}{(}\PYG{n}{kernel\PYGZus{}size}\PYG{o}{\PYGZhy{}}\PYG{l+m+mi}{1}\PYG{p}{)}\PYG{p}{,} \PYG{n}{strides}\PYG{o}{=}\PYG{p}{(}\PYG{l+m+mi}{1}\PYG{p}{,} \PYG{l+m+mi}{1}\PYG{p}{)}\PYG{p}{,} \PYG{n}{padding}\PYG{o}{=}\PYG{n}{pad}\PYG{p}{,}
                                           \PYG{n}{data\PYGZus{}format}\PYG{o}{=}\PYG{l+s+s2}{\PYGZdq{}}\PYG{l+s+s2}{channels\PYGZus{}last}\PYG{l+s+s2}{\PYGZdq{}}\PYG{p}{,} \PYG{n}{kernel\PYGZus{}divergence\PYGZus{}fn}\PYG{o}{=}\PYG{n}{kdf}\PYG{p}{,}
                                           \PYG{n}{activation}\PYG{o}{=}\PYG{k+kc}{None}\PYG{p}{)}\PYG{p}{)}
        \PYG{k}{else}\PYG{p}{:}
            \PYG{n}{block}\PYG{o}{.}\PYG{n}{add}\PYG{p}{(}\PYG{n}{Conv2D}\PYG{p}{(}\PYG{n}{filters}\PYG{o}{=}\PYG{n}{filters}\PYG{p}{,} \PYG{n}{kernel\PYGZus{}size}\PYG{o}{=}\PYG{p}{(}\PYG{n}{kernel\PYGZus{}size}\PYG{o}{\PYGZhy{}}\PYG{l+m+mi}{1}\PYG{p}{)}\PYG{p}{,} \PYG{n}{padding}\PYG{o}{=}\PYG{n}{pad}\PYG{p}{,}
                         \PYG{n}{kernel\PYGZus{}initializer}\PYG{o}{=}\PYG{n}{RandomNormal}\PYG{p}{(}\PYG{l+m+mf}{0.0}\PYG{p}{,} \PYG{l+m+mf}{0.02}\PYG{p}{)}\PYG{p}{,} \PYG{n}{strides}\PYG{o}{=}\PYG{p}{(}\PYG{l+m+mi}{1}\PYG{p}{,}\PYG{l+m+mi}{1}\PYG{p}{)}\PYG{p}{,} \PYG{n}{activation}\PYG{o}{=}\PYG{k+kc}{None}\PYG{p}{)}\PYG{p}{)}
    
    \PYG{n}{block}\PYG{o}{.}\PYG{n}{add}\PYG{p}{(}\PYG{n}{SpatialDropout2D}\PYG{p}{(}\PYG{n}{rate}\PYG{o}{=}\PYG{n}{dropout}\PYG{p}{)}\PYG{p}{)}
    
    \PYG{k}{if} \PYG{n}{bn}\PYG{p}{:}
        \PYG{n}{block}\PYG{o}{.}\PYG{n}{add}\PYG{p}{(}\PYG{n}{BatchNormalization}\PYG{p}{(}\PYG{n}{axis}\PYG{o}{=}\PYG{o}{\PYGZhy{}}\PYG{l+m+mi}{1}\PYG{p}{,} \PYG{n}{epsilon}\PYG{o}{=}\PYG{l+m+mf}{1e\PYGZhy{}05}\PYG{p}{,}\PYG{n}{momentum}\PYG{o}{=}\PYG{l+m+mf}{0.9}\PYG{p}{)}\PYG{p}{)}

    \PYG{k}{return} \PYG{n}{block}
\end{sphinxVerbatim}

Next we define the full network with these blocks \sphinxhyphen{} the structure is almost identical to the previous notebook. We manually define the kernel\sphinxhyphen{}divergence function as \sphinxcode{\sphinxupquote{kdf}} and rescale it with a factor called \sphinxcode{\sphinxupquote{kl\_scaling}}. There are two reasons for this:

First, we should only apply the kl\sphinxhyphen{}divergence once per epoch if we want to use the correct loss (like introduced in {\hyperref[\detokenize{bayesian-intro::doc}]{\sphinxcrossref{\DUrole{doc}{Introduction to Posterior Inference}}}}). Since we will use batch\sphinxhyphen{}wise training, we need to rescale the Kl\sphinxhyphen{}divergence by the number of batches, such that in every parameter update only \sphinxstyleemphasis{kdf / num\_batches} is added to the loss. During one epoch, \sphinxstyleemphasis{num\_batches} parameter updates are performed and the ‘full’ KL\sphinxhyphen{}divergence is used. This batch scaling is computed and passed to the network initialization via \sphinxcode{\sphinxupquote{kl\_scaling}} when instantiating the \sphinxcode{\sphinxupquote{Bayes\_DfpNet}} NN later on.

Second, by scaling the KL\sphinxhyphen{}divergence part of the loss up or down, we have a way of tuning how much randomness we want to allow in the network: If we neglect the KL\sphinxhyphen{}divergence completely, we would just minimize the regular loss (e.g. MSE or MAE), like in a conventional neural network. If we instead neglect the negative\sphinxhyphen{}log\sphinxhyphen{}likelihood, we would optimize the network such that we obtain random draws from the prior distribution. Balancing those extremes can be done by fine\sphinxhyphen{}tuning the scaling of the KL\sphinxhyphen{}divergence and is hard in practice.

\begin{sphinxVerbatim}[commandchars=\\\{\}]
\PYG{k}{def} \PYG{n+nf}{Bayes\PYGZus{}DfpNet}\PYG{p}{(}\PYG{n}{input\PYGZus{}shape}\PYG{o}{=}\PYG{p}{(}\PYG{l+m+mi}{128}\PYG{p}{,}\PYG{l+m+mi}{128}\PYG{p}{,}\PYG{l+m+mi}{3}\PYG{p}{)}\PYG{p}{,}\PYG{n}{expo}\PYG{o}{=}\PYG{l+m+mi}{5}\PYG{p}{,}\PYG{n}{dropout}\PYG{o}{=}\PYG{l+m+mf}{0.}\PYG{p}{,}\PYG{n}{flipout}\PYG{o}{=}\PYG{k+kc}{False}\PYG{p}{,}\PYG{n}{kl\PYGZus{}scaling}\PYG{o}{=}\PYG{l+m+mi}{10000}\PYG{p}{)}\PYG{p}{:}
    \PYG{n}{channels} \PYG{o}{=} \PYG{n+nb}{int}\PYG{p}{(}\PYG{l+m+mi}{2} \PYG{o}{*}\PYG{o}{*} \PYG{n}{expo} \PYG{o}{+} \PYG{l+m+mf}{0.5}\PYG{p}{)}
    \PYG{n}{kdf} \PYG{o}{=} \PYG{p}{(}\PYG{k}{lambda} \PYG{n}{q}\PYG{p}{,} \PYG{n}{p}\PYG{p}{,} \PYG{n}{\PYGZus{}}\PYG{p}{:} \PYG{n}{tfd}\PYG{o}{.}\PYG{n}{kl\PYGZus{}divergence}\PYG{p}{(}\PYG{n}{q}\PYG{p}{,} \PYG{n}{p}\PYG{p}{)} \PYG{o}{/} \PYG{n}{tf}\PYG{o}{.}\PYG{n}{cast}\PYG{p}{(}\PYG{n}{kl\PYGZus{}scaling}\PYG{p}{,} \PYG{n}{dtype}\PYG{o}{=}\PYG{n}{tf}\PYG{o}{.}\PYG{n}{float32}\PYG{p}{)}\PYG{p}{)}

    \PYG{n}{layer1}\PYG{o}{=}\PYG{n}{Sequential}\PYG{p}{(}\PYG{n}{name}\PYG{o}{=}\PYG{l+s+s1}{\PYGZsq{}}\PYG{l+s+s1}{layer1}\PYG{l+s+s1}{\PYGZsq{}}\PYG{p}{)}
    \PYG{n}{layer1}\PYG{o}{.}\PYG{n}{add}\PYG{p}{(}\PYG{n}{Conv2D}\PYG{p}{(}\PYG{n}{filters}\PYG{o}{=}\PYG{n}{channels}\PYG{p}{,}\PYG{n}{kernel\PYGZus{}size}\PYG{o}{=}\PYG{l+m+mi}{4}\PYG{p}{,}\PYG{n}{strides}\PYG{o}{=}\PYG{p}{(}\PYG{l+m+mi}{2}\PYG{p}{,}\PYG{l+m+mi}{2}\PYG{p}{)}\PYG{p}{,}\PYG{n}{padding}\PYG{o}{=}\PYG{l+s+s1}{\PYGZsq{}}\PYG{l+s+s1}{same}\PYG{l+s+s1}{\PYGZsq{}}\PYG{p}{,}\PYG{n}{activation}\PYG{o}{=}\PYG{k+kc}{None}\PYG{p}{,}\PYG{n}{data\PYGZus{}format}\PYG{o}{=}\PYG{l+s+s1}{\PYGZsq{}}\PYG{l+s+s1}{channels\PYGZus{}last}\PYG{l+s+s1}{\PYGZsq{}}\PYG{p}{)}\PYG{p}{)}
    \PYG{n}{layer2}\PYG{o}{=}\PYG{n}{tfBlockUnet}\PYG{p}{(}\PYG{n}{filters}\PYG{o}{=}\PYG{n}{channels}\PYG{o}{*}\PYG{l+m+mi}{2}\PYG{p}{,}\PYG{n}{transposed}\PYG{o}{=}\PYG{k+kc}{False}\PYG{p}{,}\PYG{n}{bn}\PYG{o}{=}\PYG{k+kc}{True}\PYG{p}{,} \PYG{n}{relu}\PYG{o}{=}\PYG{k+kc}{False}\PYG{p}{,}\PYG{n}{dropout}\PYG{o}{=}\PYG{n}{dropout}\PYG{p}{,}\PYG{n}{name}\PYG{o}{=}\PYG{l+s+s1}{\PYGZsq{}}\PYG{l+s+s1}{layer2}\PYG{l+s+s1}{\PYGZsq{}}\PYG{p}{)}
    \PYG{n}{layer3}\PYG{o}{=}\PYG{n}{tfBlockUnet}\PYG{p}{(}\PYG{n}{filters}\PYG{o}{=}\PYG{n}{channels}\PYG{o}{*}\PYG{l+m+mi}{2}\PYG{p}{,}\PYG{n}{transposed}\PYG{o}{=}\PYG{k+kc}{False}\PYG{p}{,}\PYG{n}{bn}\PYG{o}{=}\PYG{k+kc}{True}\PYG{p}{,} \PYG{n}{relu}\PYG{o}{=}\PYG{k+kc}{False}\PYG{p}{,}\PYG{n}{dropout}\PYG{o}{=}\PYG{n}{dropout}\PYG{p}{,}\PYG{n}{name}\PYG{o}{=}\PYG{l+s+s1}{\PYGZsq{}}\PYG{l+s+s1}{layer3}\PYG{l+s+s1}{\PYGZsq{}}\PYG{p}{)}        
    \PYG{n}{layer4}\PYG{o}{=}\PYG{n}{tfBlockUnet}\PYG{p}{(}\PYG{n}{filters}\PYG{o}{=}\PYG{n}{channels}\PYG{o}{*}\PYG{l+m+mi}{4}\PYG{p}{,}\PYG{n}{transposed}\PYG{o}{=}\PYG{k+kc}{False}\PYG{p}{,}\PYG{n}{bn}\PYG{o}{=}\PYG{k+kc}{True}\PYG{p}{,} \PYG{n}{relu}\PYG{o}{=}\PYG{k+kc}{False}\PYG{p}{,}\PYG{n}{dropout}\PYG{o}{=}\PYG{n}{dropout}\PYG{p}{,}\PYG{n}{name}\PYG{o}{=}\PYG{l+s+s1}{\PYGZsq{}}\PYG{l+s+s1}{layer4}\PYG{l+s+s1}{\PYGZsq{}}\PYG{p}{)}        
    \PYG{n}{layer5}\PYG{o}{=}\PYG{n}{tfBlockUnet}\PYG{p}{(}\PYG{n}{filters}\PYG{o}{=}\PYG{n}{channels}\PYG{o}{*}\PYG{l+m+mi}{8}\PYG{p}{,}\PYG{n}{transposed}\PYG{o}{=}\PYG{k+kc}{False}\PYG{p}{,}\PYG{n}{bn}\PYG{o}{=}\PYG{k+kc}{True}\PYG{p}{,} \PYG{n}{relu}\PYG{o}{=}\PYG{k+kc}{False}\PYG{p}{,}\PYG{n}{dropout}\PYG{o}{=}\PYG{n}{dropout}\PYG{p}{,}\PYG{n}{name}\PYG{o}{=}\PYG{l+s+s1}{\PYGZsq{}}\PYG{l+s+s1}{layer5}\PYG{l+s+s1}{\PYGZsq{}}\PYG{p}{)}        
    \PYG{n}{layer6}\PYG{o}{=}\PYG{n}{tfBlockUnet}\PYG{p}{(}\PYG{n}{filters}\PYG{o}{=}\PYG{n}{channels}\PYG{o}{*}\PYG{l+m+mi}{8}\PYG{p}{,}\PYG{n}{transposed}\PYG{o}{=}\PYG{k+kc}{False}\PYG{p}{,}\PYG{n}{bn}\PYG{o}{=}\PYG{k+kc}{True}\PYG{p}{,} \PYG{n}{relu}\PYG{o}{=}\PYG{k+kc}{False}\PYG{p}{,}\PYG{n}{dropout}\PYG{o}{=}\PYG{n}{dropout}\PYG{p}{,}\PYG{n}{kernel\PYGZus{}size}\PYG{o}{=}\PYG{l+m+mi}{2}\PYG{p}{,}\PYG{n}{pad}\PYG{o}{=}\PYG{l+s+s1}{\PYGZsq{}}\PYG{l+s+s1}{valid}\PYG{l+s+s1}{\PYGZsq{}}\PYG{p}{,}\PYG{n}{name}\PYG{o}{=}\PYG{l+s+s1}{\PYGZsq{}}\PYG{l+s+s1}{layer6}\PYG{l+s+s1}{\PYGZsq{}}\PYG{p}{)}        
    \PYG{n}{layer7}\PYG{o}{=}\PYG{n}{tfBlockUnet}\PYG{p}{(}\PYG{n}{filters}\PYG{o}{=}\PYG{n}{channels}\PYG{o}{*}\PYG{l+m+mi}{8}\PYG{p}{,}\PYG{n}{transposed}\PYG{o}{=}\PYG{k+kc}{False}\PYG{p}{,}\PYG{n}{bn}\PYG{o}{=}\PYG{k+kc}{True}\PYG{p}{,} \PYG{n}{relu}\PYG{o}{=}\PYG{k+kc}{False}\PYG{p}{,}\PYG{n}{dropout}\PYG{o}{=}\PYG{n}{dropout}\PYG{p}{,}\PYG{n}{kernel\PYGZus{}size}\PYG{o}{=}\PYG{l+m+mi}{2}\PYG{p}{,}\PYG{n}{pad}\PYG{o}{=}\PYG{l+s+s1}{\PYGZsq{}}\PYG{l+s+s1}{valid}\PYG{l+s+s1}{\PYGZsq{}}\PYG{p}{,}\PYG{n}{name}\PYG{o}{=}\PYG{l+s+s1}{\PYGZsq{}}\PYG{l+s+s1}{layer7}\PYG{l+s+s1}{\PYGZsq{}}\PYG{p}{)}        

    \PYG{c+c1}{\PYGZsh{} note, kernel size is internally reduced by one for the decoder part}
    \PYG{n}{dlayer7}\PYG{o}{=}\PYG{n}{tfBlockUnet}\PYG{p}{(}\PYG{n}{filters}\PYG{o}{=}\PYG{n}{channels}\PYG{o}{*}\PYG{l+m+mi}{8}\PYG{p}{,}\PYG{n}{transposed}\PYG{o}{=}\PYG{k+kc}{True}\PYG{p}{,}\PYG{n}{bn}\PYG{o}{=}\PYG{k+kc}{True}\PYG{p}{,} \PYG{n}{relu}\PYG{o}{=}\PYG{k+kc}{True}\PYG{p}{,}\PYG{n}{dropout}\PYG{o}{=}\PYG{n}{dropout}\PYG{p}{,} \PYG{n}{flipout}\PYG{o}{=}\PYG{n}{flipout}\PYG{p}{,}\PYG{n}{kdf}\PYG{o}{=}\PYG{n}{kdf}\PYG{p}{,} \PYG{n}{kernel\PYGZus{}size}\PYG{o}{=}\PYG{l+m+mi}{2}\PYG{p}{,}\PYG{n}{pad}\PYG{o}{=}\PYG{l+s+s1}{\PYGZsq{}}\PYG{l+s+s1}{valid}\PYG{l+s+s1}{\PYGZsq{}}\PYG{p}{,}\PYG{n}{name}\PYG{o}{=}\PYG{l+s+s1}{\PYGZsq{}}\PYG{l+s+s1}{dlayer7}\PYG{l+s+s1}{\PYGZsq{}}\PYG{p}{)}        
    \PYG{n}{dlayer6}\PYG{o}{=}\PYG{n}{tfBlockUnet}\PYG{p}{(}\PYG{n}{filters}\PYG{o}{=}\PYG{n}{channels}\PYG{o}{*}\PYG{l+m+mi}{8}\PYG{p}{,}\PYG{n}{transposed}\PYG{o}{=}\PYG{k+kc}{True}\PYG{p}{,}\PYG{n}{bn}\PYG{o}{=}\PYG{k+kc}{True}\PYG{p}{,} \PYG{n}{relu}\PYG{o}{=}\PYG{k+kc}{True}\PYG{p}{,}\PYG{n}{dropout}\PYG{o}{=}\PYG{n}{dropout}\PYG{p}{,} \PYG{n}{flipout}\PYG{o}{=}\PYG{n}{flipout}\PYG{p}{,}\PYG{n}{kdf}\PYG{o}{=}\PYG{n}{kdf}\PYG{p}{,} \PYG{n}{kernel\PYGZus{}size}\PYG{o}{=}\PYG{l+m+mi}{2}\PYG{p}{,}\PYG{n}{pad}\PYG{o}{=}\PYG{l+s+s1}{\PYGZsq{}}\PYG{l+s+s1}{valid}\PYG{l+s+s1}{\PYGZsq{}}\PYG{p}{,}\PYG{n}{name}\PYG{o}{=}\PYG{l+s+s1}{\PYGZsq{}}\PYG{l+s+s1}{dlayer6}\PYG{l+s+s1}{\PYGZsq{}}\PYG{p}{)}        
    \PYG{n}{dlayer5}\PYG{o}{=}\PYG{n}{tfBlockUnet}\PYG{p}{(}\PYG{n}{filters}\PYG{o}{=}\PYG{n}{channels}\PYG{o}{*}\PYG{l+m+mi}{4}\PYG{p}{,}\PYG{n}{transposed}\PYG{o}{=}\PYG{k+kc}{True}\PYG{p}{,}\PYG{n}{bn}\PYG{o}{=}\PYG{k+kc}{True}\PYG{p}{,} \PYG{n}{relu}\PYG{o}{=}\PYG{k+kc}{True}\PYG{p}{,}\PYG{n}{dropout}\PYG{o}{=}\PYG{n}{dropout}\PYG{p}{,} \PYG{n}{flipout}\PYG{o}{=}\PYG{n}{flipout}\PYG{p}{,}\PYG{n}{kdf}\PYG{o}{=}\PYG{n}{kdf}\PYG{p}{,}\PYG{n}{name}\PYG{o}{=}\PYG{l+s+s1}{\PYGZsq{}}\PYG{l+s+s1}{dlayer5}\PYG{l+s+s1}{\PYGZsq{}}\PYG{p}{)}        
    \PYG{n}{dlayer4}\PYG{o}{=}\PYG{n}{tfBlockUnet}\PYG{p}{(}\PYG{n}{filters}\PYG{o}{=}\PYG{n}{channels}\PYG{o}{*}\PYG{l+m+mi}{2}\PYG{p}{,}\PYG{n}{transposed}\PYG{o}{=}\PYG{k+kc}{True}\PYG{p}{,}\PYG{n}{bn}\PYG{o}{=}\PYG{k+kc}{True}\PYG{p}{,} \PYG{n}{relu}\PYG{o}{=}\PYG{k+kc}{True}\PYG{p}{,}\PYG{n}{dropout}\PYG{o}{=}\PYG{n}{dropout}\PYG{p}{,} \PYG{n}{flipout}\PYG{o}{=}\PYG{n}{flipout}\PYG{p}{,}\PYG{n}{kdf}\PYG{o}{=}\PYG{n}{kdf}\PYG{p}{,}\PYG{n}{name}\PYG{o}{=}\PYG{l+s+s1}{\PYGZsq{}}\PYG{l+s+s1}{dlayer4}\PYG{l+s+s1}{\PYGZsq{}}\PYG{p}{)}
    \PYG{n}{dlayer3}\PYG{o}{=}\PYG{n}{tfBlockUnet}\PYG{p}{(}\PYG{n}{filters}\PYG{o}{=}\PYG{n}{channels}\PYG{o}{*}\PYG{l+m+mi}{2}\PYG{p}{,}\PYG{n}{transposed}\PYG{o}{=}\PYG{k+kc}{True}\PYG{p}{,}\PYG{n}{bn}\PYG{o}{=}\PYG{k+kc}{True}\PYG{p}{,} \PYG{n}{relu}\PYG{o}{=}\PYG{k+kc}{True}\PYG{p}{,}\PYG{n}{dropout}\PYG{o}{=}\PYG{n}{dropout}\PYG{p}{,} \PYG{n}{flipout}\PYG{o}{=}\PYG{n}{flipout}\PYG{p}{,}\PYG{n}{kdf}\PYG{o}{=}\PYG{n}{kdf}\PYG{p}{,}\PYG{n}{name}\PYG{o}{=}\PYG{l+s+s1}{\PYGZsq{}}\PYG{l+s+s1}{dlayer3}\PYG{l+s+s1}{\PYGZsq{}}\PYG{p}{)}
    \PYG{n}{dlayer2}\PYG{o}{=}\PYG{n}{tfBlockUnet}\PYG{p}{(}\PYG{n}{filters}\PYG{o}{=}\PYG{n}{channels}  \PYG{p}{,}\PYG{n}{transposed}\PYG{o}{=}\PYG{k+kc}{True}\PYG{p}{,}\PYG{n}{bn}\PYG{o}{=}\PYG{k+kc}{True}\PYG{p}{,} \PYG{n}{relu}\PYG{o}{=}\PYG{k+kc}{True}\PYG{p}{,}\PYG{n}{dropout}\PYG{o}{=}\PYG{n}{dropout}\PYG{p}{,} \PYG{n}{flipout}\PYG{o}{=}\PYG{n}{flipout}\PYG{p}{,}\PYG{n}{kdf}\PYG{o}{=}\PYG{n}{kdf}\PYG{p}{,}\PYG{n}{name}\PYG{o}{=}\PYG{l+s+s1}{\PYGZsq{}}\PYG{l+s+s1}{dlayer2}\PYG{l+s+s1}{\PYGZsq{}}\PYG{p}{)}
    \PYG{n}{dlayer1}\PYG{o}{=}\PYG{n}{Sequential}\PYG{p}{(}\PYG{n}{name}\PYG{o}{=}\PYG{l+s+s1}{\PYGZsq{}}\PYG{l+s+s1}{outlayer}\PYG{l+s+s1}{\PYGZsq{}}\PYG{p}{)}
    \PYG{n}{dlayer1}\PYG{o}{.}\PYG{n}{add}\PYG{p}{(}\PYG{n}{ReLU}\PYG{p}{(}\PYG{p}{)}\PYG{p}{)}
    \PYG{n}{dlayer1}\PYG{o}{.}\PYG{n}{add}\PYG{p}{(}\PYG{n}{Conv2DTranspose}\PYG{p}{(}\PYG{l+m+mi}{3}\PYG{p}{,}\PYG{n}{kernel\PYGZus{}size}\PYG{o}{=}\PYG{l+m+mi}{4}\PYG{p}{,}\PYG{n}{strides}\PYG{o}{=}\PYG{p}{(}\PYG{l+m+mi}{2}\PYG{p}{,}\PYG{l+m+mi}{2}\PYG{p}{)}\PYG{p}{,}\PYG{n}{padding}\PYG{o}{=}\PYG{l+s+s1}{\PYGZsq{}}\PYG{l+s+s1}{same}\PYG{l+s+s1}{\PYGZsq{}}\PYG{p}{)}\PYG{p}{)}

    \PYG{c+c1}{\PYGZsh{} forward pass}
    \PYG{n}{inputs}\PYG{o}{=}\PYG{n}{Input}\PYG{p}{(}\PYG{n}{input\PYGZus{}shape}\PYG{p}{)}
    \PYG{n}{out1} \PYG{o}{=} \PYG{n}{layer1}\PYG{p}{(}\PYG{n}{inputs}\PYG{p}{)}
    \PYG{n}{out2} \PYG{o}{=} \PYG{n}{layer2}\PYG{p}{(}\PYG{n}{out1}\PYG{p}{)}
    \PYG{n}{out3} \PYG{o}{=} \PYG{n}{layer3}\PYG{p}{(}\PYG{n}{out2}\PYG{p}{)}
    \PYG{n}{out4} \PYG{o}{=} \PYG{n}{layer4}\PYG{p}{(}\PYG{n}{out3}\PYG{p}{)}
    \PYG{n}{out5} \PYG{o}{=} \PYG{n}{layer5}\PYG{p}{(}\PYG{n}{out4}\PYG{p}{)}
    \PYG{n}{out6} \PYG{o}{=} \PYG{n}{layer6}\PYG{p}{(}\PYG{n}{out5}\PYG{p}{)}
    \PYG{n}{out7} \PYG{o}{=} \PYG{n}{layer7}\PYG{p}{(}\PYG{n}{out6}\PYG{p}{)}
    \PYG{c+c1}{\PYGZsh{} ... bottleneck ...}
    \PYG{n}{dout6} \PYG{o}{=} \PYG{n}{dlayer7}\PYG{p}{(}\PYG{n}{out7}\PYG{p}{)}
    \PYG{n}{dout6\PYGZus{}out6} \PYG{o}{=} \PYG{n}{tf}\PYG{o}{.}\PYG{n}{concat}\PYG{p}{(}\PYG{p}{[}\PYG{n}{dout6}\PYG{p}{,}\PYG{n}{out6}\PYG{p}{]}\PYG{p}{,}\PYG{n}{axis}\PYG{o}{=}\PYG{l+m+mi}{3}\PYG{p}{)}
    \PYG{n}{dout6} \PYG{o}{=} \PYG{n}{dlayer6}\PYG{p}{(}\PYG{n}{dout6\PYGZus{}out6}\PYG{p}{)}
    \PYG{n}{dout6\PYGZus{}out5} \PYG{o}{=} \PYG{n}{tf}\PYG{o}{.}\PYG{n}{concat}\PYG{p}{(}\PYG{p}{[}\PYG{n}{dout6}\PYG{p}{,} \PYG{n}{out5}\PYG{p}{]}\PYG{p}{,} \PYG{n}{axis}\PYG{o}{=}\PYG{l+m+mi}{3}\PYG{p}{)}
    \PYG{n}{dout5} \PYG{o}{=} \PYG{n}{dlayer5}\PYG{p}{(}\PYG{n}{dout6\PYGZus{}out5}\PYG{p}{)}
    \PYG{n}{dout5\PYGZus{}out4} \PYG{o}{=} \PYG{n}{tf}\PYG{o}{.}\PYG{n}{concat}\PYG{p}{(}\PYG{p}{[}\PYG{n}{dout5}\PYG{p}{,} \PYG{n}{out4}\PYG{p}{]}\PYG{p}{,} \PYG{n}{axis}\PYG{o}{=}\PYG{l+m+mi}{3}\PYG{p}{)}
    \PYG{n}{dout4} \PYG{o}{=} \PYG{n}{dlayer4}\PYG{p}{(}\PYG{n}{dout5\PYGZus{}out4}\PYG{p}{)}
    \PYG{n}{dout4\PYGZus{}out3} \PYG{o}{=} \PYG{n}{tf}\PYG{o}{.}\PYG{n}{concat}\PYG{p}{(}\PYG{p}{[}\PYG{n}{dout4}\PYG{p}{,} \PYG{n}{out3}\PYG{p}{]}\PYG{p}{,} \PYG{n}{axis}\PYG{o}{=}\PYG{l+m+mi}{3}\PYG{p}{)}
    \PYG{n}{dout3} \PYG{o}{=} \PYG{n}{dlayer3}\PYG{p}{(}\PYG{n}{dout4\PYGZus{}out3}\PYG{p}{)}
    \PYG{n}{dout3\PYGZus{}out2} \PYG{o}{=} \PYG{n}{tf}\PYG{o}{.}\PYG{n}{concat}\PYG{p}{(}\PYG{p}{[}\PYG{n}{dout3}\PYG{p}{,} \PYG{n}{out2}\PYG{p}{]}\PYG{p}{,} \PYG{n}{axis}\PYG{o}{=}\PYG{l+m+mi}{3}\PYG{p}{)}
    \PYG{n}{dout2} \PYG{o}{=} \PYG{n}{dlayer2}\PYG{p}{(}\PYG{n}{dout3\PYGZus{}out2}\PYG{p}{)}
    \PYG{n}{dout2\PYGZus{}out1} \PYG{o}{=} \PYG{n}{tf}\PYG{o}{.}\PYG{n}{concat}\PYG{p}{(}\PYG{p}{[}\PYG{n}{dout2}\PYG{p}{,} \PYG{n}{out1}\PYG{p}{]}\PYG{p}{,} \PYG{n}{axis}\PYG{o}{=}\PYG{l+m+mi}{3}\PYG{p}{)}
    \PYG{n}{dout1} \PYG{o}{=} \PYG{n}{dlayer1}\PYG{p}{(}\PYG{n}{dout2\PYGZus{}out1}\PYG{p}{)}
    
    \PYG{k}{return} \PYG{n}{Model}\PYG{p}{(}\PYG{n}{inputs}\PYG{o}{=}\PYG{n}{inputs}\PYG{p}{,}\PYG{n}{outputs}\PYG{o}{=}\PYG{n}{dout1}\PYG{p}{)}
\end{sphinxVerbatim}

Let’s define the hyperparameters and create a tensorflow dataset to organize inputs and targets. Since we have 320 observations in the training set, for a batch\sphinxhyphen{}size of 10 we should rescale the KL\sphinxhyphen{}divergence with a factor of 320/10=32 in order apply the full KL\sphinxhyphen{}divergence just once per epoch. We will further scale the KL\sphinxhyphen{}divergence down by another factor of \sphinxcode{\sphinxupquote{KL\_PREF=5000}}, which has shown to work well in practice.

Furthermore, we will define a function that implements learning rate decay. Intuitively, this allows the optimization to be more precise (by making smaller steps) in later epochs, while still making fast progress (by making bigger steps) in the first epochs.

\begin{sphinxVerbatim}[commandchars=\\\{\}]
\PYG{k+kn}{import} \PYG{n+nn}{math} 
\PYG{k+kn}{import} \PYG{n+nn}{matplotlib}\PYG{n+nn}{.}\PYG{n+nn}{pyplot} \PYG{k}{as} \PYG{n+nn}{plt}

\PYG{n}{BATCH\PYGZus{}SIZE}\PYG{o}{=}\PYG{l+m+mi}{10}
\PYG{n}{LR}\PYG{o}{=}\PYG{l+m+mf}{0.001}
\PYG{n}{EPOCHS} \PYG{o}{=} \PYG{l+m+mi}{120}
\PYG{n}{KL\PYGZus{}PREF} \PYG{o}{=} \PYG{l+m+mi}{5000}

\PYG{n}{dataset} \PYG{o}{=} \PYG{n}{tf}\PYG{o}{.}\PYG{n}{data}\PYG{o}{.}\PYG{n}{Dataset}\PYG{o}{.}\PYG{n}{from\PYGZus{}tensor\PYGZus{}slices}\PYG{p}{(}\PYG{p}{(}\PYG{n}{X\PYGZus{}train}\PYG{p}{,} \PYG{n}{y\PYGZus{}train}\PYG{p}{)}\PYG{p}{)}\PYG{o}{.}\PYG{n}{shuffle}\PYG{p}{(}\PYG{n+nb}{len}\PYG{p}{(}\PYG{n}{X\PYGZus{}train}\PYG{p}{)}\PYG{p}{,}
    \PYG{n}{seed}\PYG{o}{=}\PYG{l+m+mi}{46168531}\PYG{p}{,} \PYG{n}{reshuffle\PYGZus{}each\PYGZus{}iteration}\PYG{o}{=}\PYG{k+kc}{False}\PYG{p}{)}\PYG{o}{.}\PYG{n}{batch}\PYG{p}{(}\PYG{n}{BATCH\PYGZus{}SIZE}\PYG{p}{,} \PYG{n}{drop\PYGZus{}remainder}\PYG{o}{=}\PYG{k+kc}{False}\PYG{p}{)}

\PYG{k}{def} \PYG{n+nf}{compute\PYGZus{}lr}\PYG{p}{(}\PYG{n}{i}\PYG{p}{,} \PYG{n}{epochs}\PYG{p}{,} \PYG{n}{minLR}\PYG{p}{,} \PYG{n}{maxLR}\PYG{p}{)}\PYG{p}{:}
  \PYG{k}{if} \PYG{n}{i} \PYG{o}{\PYGZlt{}} \PYG{n}{epochs} \PYG{o}{*} \PYG{l+m+mf}{0.5}\PYG{p}{:}
      \PYG{k}{return} \PYG{n}{maxLR}
  \PYG{n}{e} \PYG{o}{=} \PYG{p}{(}\PYG{n}{i} \PYG{o}{/} \PYG{n+nb}{float}\PYG{p}{(}\PYG{n}{epochs}\PYG{p}{)} \PYG{o}{\PYGZhy{}} \PYG{l+m+mf}{0.5}\PYG{p}{)} \PYG{o}{*} \PYG{l+m+mf}{2.}
  \PYG{c+c1}{\PYGZsh{} rescale second half to min/max range}
  \PYG{n}{fmin} \PYG{o}{=} \PYG{l+m+mf}{0.}
  \PYG{n}{fmax} \PYG{o}{=} \PYG{l+m+mf}{6.}
  \PYG{n}{e} \PYG{o}{=} \PYG{n}{fmin} \PYG{o}{+} \PYG{n}{e} \PYG{o}{*} \PYG{p}{(}\PYG{n}{fmax} \PYG{o}{\PYGZhy{}} \PYG{n}{fmin}\PYG{p}{)}
  \PYG{n}{f} \PYG{o}{=} \PYG{n}{math}\PYG{o}{.}\PYG{n}{pow}\PYG{p}{(}\PYG{l+m+mf}{0.5}\PYG{p}{,} \PYG{n}{e}\PYG{p}{)}
  \PYG{k}{return} \PYG{n}{minLR} \PYG{o}{+} \PYG{p}{(}\PYG{n}{maxLR} \PYG{o}{\PYGZhy{}} \PYG{n}{minLR}\PYG{p}{)} \PYG{o}{*} \PYG{n}{f}
\end{sphinxVerbatim}

We can visualize the learning rate decay: We start off with a constant rate and after half of the \sphinxcode{\sphinxupquote{EPOCHS}} we start to decay it exponentially, until arriving at half of the original learning rate.

\begin{sphinxVerbatim}[commandchars=\\\{\}]
\PYG{n}{lrs}\PYG{o}{=}\PYG{p}{[}\PYG{n}{compute\PYGZus{}lr}\PYG{p}{(}\PYG{n}{i}\PYG{p}{,} \PYG{n}{EPOCHS}\PYG{p}{,} \PYG{l+m+mf}{0.5}\PYG{o}{*}\PYG{n}{LR}\PYG{p}{,}\PYG{n}{LR}\PYG{p}{)} \PYG{k}{for} \PYG{n}{i} \PYG{o+ow}{in} \PYG{n+nb}{range}\PYG{p}{(}\PYG{n}{EPOCHS}\PYG{p}{)}\PYG{p}{]}
\PYG{n}{plt}\PYG{o}{.}\PYG{n}{plot}\PYG{p}{(}\PYG{n}{lrs}\PYG{p}{)}
\PYG{n}{plt}\PYG{o}{.}\PYG{n}{xlabel}\PYG{p}{(}\PYG{l+s+s1}{\PYGZsq{}}\PYG{l+s+s1}{Iteration}\PYG{l+s+s1}{\PYGZsq{}}\PYG{p}{)}
\PYG{n}{plt}\PYG{o}{.}\PYG{n}{ylabel}\PYG{p}{(}\PYG{l+s+s1}{\PYGZsq{}}\PYG{l+s+s1}{Learning Rate}\PYG{l+s+s1}{\PYGZsq{}}\PYG{p}{)}
\end{sphinxVerbatim}

\begin{sphinxVerbatim}[commandchars=\\\{\}]
Text(0, 0.5, \PYGZsq{}Learning Rate\PYGZsq{})
\end{sphinxVerbatim}

\noindent\sphinxincludegraphics{{bayesian-code_13_1}.png}

Let’s initialize the network. Here we’re finally computing the \sphinxcode{\sphinxupquote{kl\_scaling}} factor via \sphinxcode{\sphinxupquote{KL\_PREF}} and the batch size.

\begin{sphinxVerbatim}[commandchars=\\\{\}]
\PYG{k+kn}{from} \PYG{n+nn}{tensorflow}\PYG{n+nn}{.}\PYG{n+nn}{keras}\PYG{n+nn}{.}\PYG{n+nn}{optimizers} \PYG{k+kn}{import} \PYG{n}{RMSprop}\PYG{p}{,} \PYG{n}{Adam}

\PYG{n}{model}\PYG{o}{=}\PYG{n}{Bayes\PYGZus{}DfpNet}\PYG{p}{(}\PYG{n}{expo}\PYG{o}{=}\PYG{l+m+mi}{4}\PYG{p}{,}\PYG{n}{flipout}\PYG{o}{=}\PYG{k+kc}{True}\PYG{p}{,}\PYG{n}{kl\PYGZus{}scaling}\PYG{o}{=}\PYG{n}{KL\PYGZus{}PREF}\PYG{o}{*}\PYG{n+nb}{len}\PYG{p}{(}\PYG{n}{X\PYGZus{}train}\PYG{p}{)}\PYG{o}{/}\PYG{n}{BATCH\PYGZus{}SIZE}\PYG{p}{)}
\PYG{n}{optimizer} \PYG{o}{=} \PYG{n}{Adam}\PYG{p}{(}\PYG{n}{learning\PYGZus{}rate}\PYG{o}{=}\PYG{n}{LR}\PYG{p}{,} \PYG{n}{beta\PYGZus{}1}\PYG{o}{=}\PYG{l+m+mf}{0.5}\PYG{p}{,}\PYG{n}{beta\PYGZus{}2}\PYG{o}{=}\PYG{l+m+mf}{0.9999}\PYG{p}{)}

\PYG{n}{num\PYGZus{}params} \PYG{o}{=} \PYG{n}{np}\PYG{o}{.}\PYG{n}{sum}\PYG{p}{(}\PYG{p}{[}\PYG{n}{np}\PYG{o}{.}\PYG{n}{prod}\PYG{p}{(}\PYG{n}{v}\PYG{o}{.}\PYG{n}{get\PYGZus{}shape}\PYG{p}{(}\PYG{p}{)}\PYG{o}{.}\PYG{n}{as\PYGZus{}list}\PYG{p}{(}\PYG{p}{)}\PYG{p}{)} \PYG{k}{for} \PYG{n}{v} \PYG{o+ow}{in} \PYG{n}{model}\PYG{o}{.}\PYG{n}{trainable\PYGZus{}variables}\PYG{p}{]}\PYG{p}{)}
\PYG{n+nb}{print}\PYG{p}{(}\PYG{l+s+s1}{\PYGZsq{}}\PYG{l+s+s1}{The Bayesian U\PYGZhy{}Net has }\PYG{l+s+si}{\PYGZob{}\PYGZcb{}}\PYG{l+s+s1}{ parameters.}\PYG{l+s+s1}{\PYGZsq{}}\PYG{o}{.}\PYG{n}{format}\PYG{p}{(}\PYG{n}{num\PYGZus{}params}\PYG{p}{)}\PYG{p}{)}
\end{sphinxVerbatim}

\begin{sphinxVerbatim}[commandchars=\\\{\}]

\end{sphinxVerbatim}

\begin{sphinxVerbatim}[commandchars=\\\{\}]
The Bayesian U\PYGZhy{}Net has 846787 parameters.
\end{sphinxVerbatim}

In general, flipout layers come with twice as many parameters as their conventional counterparts, since instead of a single point estimate one has to learn both mean and variance parameters for the Gaussian posterior of the weights. As we only have flipout layers for the decoder part here, the resulting model has 846787 parameters, compared to the 585667 of the conventional NN.

\section{Training}
\label{\detokenize{bayesian-code:training}}
Now we are ready to run the training! Note that this might take a while (typically around 4 hours), as the flipout layers are significantly slower to train compared to regular layers.

\begin{sphinxVerbatim}[commandchars=\\\{\}]
\PYG{k+kn}{from} \PYG{n+nn}{tensorflow}\PYG{n+nn}{.}\PYG{n+nn}{keras}\PYG{n+nn}{.}\PYG{n+nn}{losses} \PYG{k+kn}{import} \PYG{n}{mae}
\PYG{k+kn}{import} \PYG{n+nn}{math}

\PYG{n}{kl\PYGZus{}losses}\PYG{o}{=}\PYG{p}{[}\PYG{p}{]}
\PYG{n}{mae\PYGZus{}losses}\PYG{o}{=}\PYG{p}{[}\PYG{p}{]}
\PYG{n}{total\PYGZus{}losses}\PYG{o}{=}\PYG{p}{[}\PYG{p}{]}
\PYG{n}{mae\PYGZus{}losses\PYGZus{}vali}\PYG{o}{=}\PYG{p}{[}\PYG{p}{]}

\PYG{k}{for} \PYG{n}{epoch} \PYG{o+ow}{in} \PYG{n+nb}{range}\PYG{p}{(}\PYG{n}{EPOCHS}\PYG{p}{)}\PYG{p}{:}
    \PYG{c+c1}{\PYGZsh{} compute learning rate \PYGZhy{} decay is implemented}
    \PYG{n}{currLr} \PYG{o}{=} \PYG{n}{compute\PYGZus{}lr}\PYG{p}{(}\PYG{n}{epoch}\PYG{p}{,}\PYG{n}{EPOCHS}\PYG{p}{,}\PYG{l+m+mf}{0.5}\PYG{o}{*}\PYG{n}{LR}\PYG{p}{,}\PYG{n}{LR}\PYG{p}{)}
    \PYG{k}{if} \PYG{n}{currLr} \PYG{o}{\PYGZlt{}} \PYG{n}{LR}\PYG{p}{:}
            \PYG{n}{tf}\PYG{o}{.}\PYG{n}{keras}\PYG{o}{.}\PYG{n}{backend}\PYG{o}{.}\PYG{n}{set\PYGZus{}value}\PYG{p}{(}\PYG{n}{optimizer}\PYG{o}{.}\PYG{n}{lr}\PYG{p}{,} \PYG{n}{currLr}\PYG{p}{)}

    \PYG{c+c1}{\PYGZsh{} iterate through training data }
    \PYG{n}{kl\PYGZus{}sum} \PYG{o}{=} \PYG{l+m+mi}{0}
    \PYG{n}{mae\PYGZus{}sum} \PYG{o}{=} \PYG{l+m+mi}{0}
    \PYG{n}{total\PYGZus{}sum}\PYG{o}{=}\PYG{l+m+mi}{0}
    \PYG{k}{for} \PYG{n}{i}\PYG{p}{,} \PYG{n}{traindata} \PYG{o+ow}{in} \PYG{n+nb}{enumerate}\PYG{p}{(}\PYG{n}{dataset}\PYG{p}{,} \PYG{l+m+mi}{0}\PYG{p}{)}\PYG{p}{:}
      \PYG{c+c1}{\PYGZsh{} forward pass and loss computation}
        \PYG{k}{with} \PYG{n}{tf}\PYG{o}{.}\PYG{n}{GradientTape}\PYG{p}{(}\PYG{p}{)} \PYG{k}{as} \PYG{n}{tape}\PYG{p}{:}
            \PYG{n}{inputs}\PYG{p}{,} \PYG{n}{targets} \PYG{o}{=} \PYG{n}{traindata}
            \PYG{n}{prediction} \PYG{o}{=} \PYG{n}{model}\PYG{p}{(}\PYG{n}{inputs}\PYG{p}{,} \PYG{n}{training}\PYG{o}{=}\PYG{k+kc}{True}\PYG{p}{)}
            \PYG{n}{loss\PYGZus{}mae} \PYG{o}{=} \PYG{n}{tf}\PYG{o}{.}\PYG{n}{reduce\PYGZus{}mean}\PYG{p}{(}\PYG{n}{mae}\PYG{p}{(}\PYG{n}{prediction}\PYG{p}{,} \PYG{n}{targets}\PYG{p}{)}\PYG{p}{)}
            \PYG{n}{kl}\PYG{o}{=}\PYG{n+nb}{sum}\PYG{p}{(}\PYG{n}{model}\PYG{o}{.}\PYG{n}{losses}\PYG{p}{)}
            \PYG{n}{loss\PYGZus{}value}\PYG{o}{=}\PYG{n}{kl}\PYG{o}{+}\PYG{n}{tf}\PYG{o}{.}\PYG{n}{cast}\PYG{p}{(}\PYG{n}{loss\PYGZus{}mae}\PYG{p}{,} \PYG{n}{dtype}\PYG{o}{=}\PYG{l+s+s1}{\PYGZsq{}}\PYG{l+s+s1}{float32}\PYG{l+s+s1}{\PYGZsq{}}\PYG{p}{)}
        \PYG{c+c1}{\PYGZsh{} backpropagate gradients and update parameters }
        \PYG{n}{gradients} \PYG{o}{=} \PYG{n}{tape}\PYG{o}{.}\PYG{n}{gradient}\PYG{p}{(}\PYG{n}{loss\PYGZus{}value}\PYG{p}{,} \PYG{n}{model}\PYG{o}{.}\PYG{n}{trainable\PYGZus{}variables}\PYG{p}{)}
        \PYG{n}{optimizer}\PYG{o}{.}\PYG{n}{apply\PYGZus{}gradients}\PYG{p}{(}\PYG{n+nb}{zip}\PYG{p}{(}\PYG{n}{gradients}\PYG{p}{,} \PYG{n}{model}\PYG{o}{.}\PYG{n}{trainable\PYGZus{}variables}\PYG{p}{)}\PYG{p}{)}
   
        \PYG{c+c1}{\PYGZsh{} store losses per batch}
        \PYG{n}{kl\PYGZus{}sum} \PYG{o}{+}\PYG{o}{=} \PYG{n}{kl}
        \PYG{n}{mae\PYGZus{}sum} \PYG{o}{+}\PYG{o}{=} \PYG{n}{tf}\PYG{o}{.}\PYG{n}{reduce\PYGZus{}mean}\PYG{p}{(}\PYG{n}{loss\PYGZus{}mae}\PYG{p}{)}
        \PYG{n}{total\PYGZus{}sum}\PYG{o}{+}\PYG{o}{=}\PYG{n}{tf}\PYG{o}{.}\PYG{n}{reduce\PYGZus{}mean}\PYG{p}{(}\PYG{n}{loss\PYGZus{}value}\PYG{p}{)}

    \PYG{c+c1}{\PYGZsh{} store losses per epoch}
    \PYG{n}{kl\PYGZus{}losses}\PYG{o}{+}\PYG{o}{=}\PYG{p}{[}\PYG{n}{kl\PYGZus{}sum}\PYG{o}{/}\PYG{n+nb}{len}\PYG{p}{(}\PYG{n}{dataset}\PYG{p}{)}\PYG{p}{]}
    \PYG{n}{mae\PYGZus{}losses}\PYG{o}{+}\PYG{o}{=}\PYG{p}{[}\PYG{n}{mae\PYGZus{}sum}\PYG{o}{/}\PYG{n+nb}{len}\PYG{p}{(}\PYG{n}{dataset}\PYG{p}{)}\PYG{p}{]}
    \PYG{n}{total\PYGZus{}losses}\PYG{o}{+}\PYG{o}{=}\PYG{p}{[}\PYG{n}{total\PYGZus{}sum}\PYG{o}{/}\PYG{n+nb}{len}\PYG{p}{(}\PYG{n}{dataset}\PYG{p}{)}\PYG{p}{]}

    \PYG{c+c1}{\PYGZsh{} validation}
    \PYG{n}{outputs} \PYG{o}{=} \PYG{n}{model}\PYG{o}{.}\PYG{n}{predict}\PYG{p}{(}\PYG{n}{X\PYGZus{}val}\PYG{p}{)}
    \PYG{n}{mae\PYGZus{}losses\PYGZus{}vali} \PYG{o}{+}\PYG{o}{=} \PYG{p}{[}\PYG{n}{tf}\PYG{o}{.}\PYG{n}{reduce\PYGZus{}mean}\PYG{p}{(}\PYG{n}{mae}\PYG{p}{(}\PYG{n}{y\PYGZus{}val}\PYG{p}{,} \PYG{n}{outputs}\PYG{p}{)}\PYG{p}{)}\PYG{p}{]}

    \PYG{k}{if} \PYG{n}{epoch}\PYG{o}{\PYGZlt{}}\PYG{l+m+mi}{3} \PYG{o+ow}{or} \PYG{n}{epoch}\PYG{o}{\PYGZpc{}}\PYG{k}{20}==0: 
        \PYG{n+nb}{print}\PYG{p}{(}\PYG{l+s+s1}{\PYGZsq{}}\PYG{l+s+s1}{Epoch }\PYG{l+s+si}{\PYGZob{}\PYGZcb{}}\PYG{l+s+s1}{/}\PYG{l+s+si}{\PYGZob{}\PYGZcb{}}\PYG{l+s+s1}{, total loss: }\PYG{l+s+si}{\PYGZob{}:.3f\PYGZcb{}}\PYG{l+s+s1}{, KL loss: }\PYG{l+s+si}{\PYGZob{}:.3f\PYGZcb{}}\PYG{l+s+s1}{, MAE loss: }\PYG{l+s+si}{\PYGZob{}:.4f\PYGZcb{}}\PYG{l+s+s1}{, MAE loss vali: }\PYG{l+s+si}{\PYGZob{}:.4f\PYGZcb{}}\PYG{l+s+s1}{\PYGZsq{}}\PYG{o}{.}\PYG{n}{format}\PYG{p}{(}\PYG{n}{epoch}\PYG{p}{,} \PYG{n}{EPOCHS}\PYG{p}{,} \PYG{n}{total\PYGZus{}losses}\PYG{p}{[}\PYG{o}{\PYGZhy{}}\PYG{l+m+mi}{1}\PYG{p}{]}\PYG{p}{,} \PYG{n}{kl\PYGZus{}losses}\PYG{p}{[}\PYG{o}{\PYGZhy{}}\PYG{l+m+mi}{1}\PYG{p}{]}\PYG{p}{,} \PYG{n}{mae\PYGZus{}losses}\PYG{p}{[}\PYG{o}{\PYGZhy{}}\PYG{l+m+mi}{1}\PYG{p}{]}\PYG{p}{,} \PYG{n}{mae\PYGZus{}losses\PYGZus{}vali}\PYG{p}{[}\PYG{o}{\PYGZhy{}}\PYG{l+m+mi}{1}\PYG{p}{]}\PYG{p}{)}\PYG{p}{)}

\end{sphinxVerbatim}

\begin{sphinxVerbatim}[commandchars=\\\{\}]
Epoch 0/120, total loss: 4.265, KL loss: 4.118, MAE loss: 0.1464, MAE loss vali: 0.0872
Epoch 1/120, total loss: 4.159, KL loss: 4.089, MAE loss: 0.0706, MAE loss vali: 0.0691
Epoch 2/120, total loss: 4.115, KL loss: 4.054, MAE loss: 0.0610, MAE loss vali: 0.0589
Epoch 20/120, total loss: 3.344, KL loss: 3.315, MAE loss: 0.0291, MAE loss vali: 0.0271
Epoch 40/120, total loss: 2.495, KL loss: 2.471, MAE loss: 0.0245, MAE loss vali: 0.0242
Epoch 60/120, total loss: 1.712, KL loss: 1.689, MAE loss: 0.0228, MAE loss vali: 0.0208
Epoch 80/120, total loss: 1.190, KL loss: 1.169, MAE loss: 0.0212, MAE loss vali: 0.0200
Epoch 100/120, total loss: 0.869, KL loss: 0.848, MAE loss: 0.0208, MAE loss vali: 0.0203
\end{sphinxVerbatim}

The BNN is trained! Let’s look at the loss. Since the loss consists of two separate parts, it is helpful to monitor both parts (MAE and KL).

\begin{sphinxVerbatim}[commandchars=\\\{\}]
\PYG{n}{fig}\PYG{p}{,}\PYG{n}{axs}\PYG{o}{=}\PYG{n}{plt}\PYG{o}{.}\PYG{n}{subplots}\PYG{p}{(}\PYG{n}{ncols}\PYG{o}{=}\PYG{l+m+mi}{3}\PYG{p}{,}\PYG{n}{nrows}\PYG{o}{=}\PYG{l+m+mi}{1}\PYG{p}{,}\PYG{n}{figsize}\PYG{o}{=}\PYG{p}{(}\PYG{l+m+mi}{20}\PYG{p}{,}\PYG{l+m+mi}{4}\PYG{p}{)}\PYG{p}{)}

\PYG{n}{axs}\PYG{p}{[}\PYG{l+m+mi}{0}\PYG{p}{]}\PYG{o}{.}\PYG{n}{plot}\PYG{p}{(}\PYG{n}{kl\PYGZus{}losses}\PYG{p}{,}\PYG{n}{color}\PYG{o}{=}\PYG{l+s+s1}{\PYGZsq{}}\PYG{l+s+s1}{red}\PYG{l+s+s1}{\PYGZsq{}}\PYG{p}{)}
\PYG{n}{axs}\PYG{p}{[}\PYG{l+m+mi}{0}\PYG{p}{]}\PYG{o}{.}\PYG{n}{set\PYGZus{}title}\PYG{p}{(}\PYG{l+s+s1}{\PYGZsq{}}\PYG{l+s+s1}{KL Loss (Train)}\PYG{l+s+s1}{\PYGZsq{}}\PYG{p}{)}
\PYG{n}{axs}\PYG{p}{[}\PYG{l+m+mi}{1}\PYG{p}{]}\PYG{o}{.}\PYG{n}{plot}\PYG{p}{(}\PYG{n}{mae\PYGZus{}losses}\PYG{p}{,}\PYG{n}{color}\PYG{o}{=}\PYG{l+s+s1}{\PYGZsq{}}\PYG{l+s+s1}{blue}\PYG{l+s+s1}{\PYGZsq{}}\PYG{p}{,}\PYG{n}{label}\PYG{o}{=}\PYG{l+s+s1}{\PYGZsq{}}\PYG{l+s+s1}{train}\PYG{l+s+s1}{\PYGZsq{}}\PYG{p}{)}
\PYG{n}{axs}\PYG{p}{[}\PYG{l+m+mi}{1}\PYG{p}{]}\PYG{o}{.}\PYG{n}{plot}\PYG{p}{(}\PYG{n}{mae\PYGZus{}losses\PYGZus{}vali}\PYG{p}{,}\PYG{n}{color}\PYG{o}{=}\PYG{l+s+s1}{\PYGZsq{}}\PYG{l+s+s1}{green}\PYG{l+s+s1}{\PYGZsq{}}\PYG{p}{,}\PYG{n}{label}\PYG{o}{=}\PYG{l+s+s1}{\PYGZsq{}}\PYG{l+s+s1}{val}\PYG{l+s+s1}{\PYGZsq{}}\PYG{p}{)}
\PYG{n}{axs}\PYG{p}{[}\PYG{l+m+mi}{1}\PYG{p}{]}\PYG{o}{.}\PYG{n}{set\PYGZus{}title}\PYG{p}{(}\PYG{l+s+s1}{\PYGZsq{}}\PYG{l+s+s1}{MAE Loss}\PYG{l+s+s1}{\PYGZsq{}}\PYG{p}{)}\PYG{p}{;} \PYG{n}{axs}\PYG{p}{[}\PYG{l+m+mi}{1}\PYG{p}{]}\PYG{o}{.}\PYG{n}{legend}\PYG{p}{(}\PYG{p}{)}

\PYG{n}{axs}\PYG{p}{[}\PYG{l+m+mi}{2}\PYG{p}{]}\PYG{o}{.}\PYG{n}{plot}\PYG{p}{(}\PYG{n}{total\PYGZus{}losses}\PYG{p}{,}\PYG{n}{label}\PYG{o}{=}\PYG{l+s+s1}{\PYGZsq{}}\PYG{l+s+s1}{Total}\PYG{l+s+s1}{\PYGZsq{}}\PYG{p}{,}\PYG{n}{color}\PYG{o}{=}\PYG{l+s+s1}{\PYGZsq{}}\PYG{l+s+s1}{black}\PYG{l+s+s1}{\PYGZsq{}}\PYG{p}{)}
\PYG{n}{axs}\PYG{p}{[}\PYG{l+m+mi}{2}\PYG{p}{]}\PYG{o}{.}\PYG{n}{plot}\PYG{p}{(}\PYG{n}{kl\PYGZus{}losses}\PYG{p}{,}\PYG{n}{label}\PYG{o}{=}\PYG{l+s+s1}{\PYGZsq{}}\PYG{l+s+s1}{KL}\PYG{l+s+s1}{\PYGZsq{}}\PYG{p}{,}\PYG{n}{color}\PYG{o}{=}\PYG{l+s+s1}{\PYGZsq{}}\PYG{l+s+s1}{red}\PYG{l+s+s1}{\PYGZsq{}}\PYG{p}{)}
\PYG{n}{axs}\PYG{p}{[}\PYG{l+m+mi}{2}\PYG{p}{]}\PYG{o}{.}\PYG{n}{plot}\PYG{p}{(}\PYG{n}{mae\PYGZus{}losses}\PYG{p}{,}\PYG{n}{label}\PYG{o}{=}\PYG{l+s+s1}{\PYGZsq{}}\PYG{l+s+s1}{MAE}\PYG{l+s+s1}{\PYGZsq{}}\PYG{p}{,}\PYG{n}{color}\PYG{o}{=}\PYG{l+s+s1}{\PYGZsq{}}\PYG{l+s+s1}{blue}\PYG{l+s+s1}{\PYGZsq{}}\PYG{p}{)}
\PYG{n}{axs}\PYG{p}{[}\PYG{l+m+mi}{2}\PYG{p}{]}\PYG{o}{.}\PYG{n}{set\PYGZus{}title}\PYG{p}{(}\PYG{l+s+s1}{\PYGZsq{}}\PYG{l+s+s1}{Total Train Loss}\PYG{l+s+s1}{\PYGZsq{}}\PYG{p}{)}\PYG{p}{;} \PYG{n}{axs}\PYG{p}{[}\PYG{l+m+mi}{2}\PYG{p}{]}\PYG{o}{.}\PYG{n}{legend}\PYG{p}{(}\PYG{p}{)}
\end{sphinxVerbatim}

\begin{sphinxVerbatim}[commandchars=\\\{\}]
\PYGZlt{}matplotlib.legend.Legend at 0x7f6285c60490\PYGZgt{}
\end{sphinxVerbatim}

\noindent\sphinxincludegraphics{{bayesian-code_19_1}.png}

This way, we can double\sphinxhyphen{}check if minimizing one part of the loss comes at the cost of increasing the other. For our case, we observe that both parts decrease smoothly. In particular, the MAE loss is not increasing for the validation set, indicating that we are not overfitting.

It is good practice to double\sphinxhyphen{}check how many layers added KL\sphinxhyphen{}losses. We can inspect \sphinxstyleemphasis{model.losses} for that. Since the decoder consists of 6 sequential blocks with flipout layers, we expect 6 entries in \sphinxstyleemphasis{model.losses}.

\begin{sphinxVerbatim}[commandchars=\\\{\}]
\PYG{c+c1}{\PYGZsh{} there should be 6 entries in model.losses since we have 6 blocks with flipout layers in our model}
\PYG{n+nb}{print}\PYG{p}{(}\PYG{l+s+s1}{\PYGZsq{}}\PYG{l+s+s1}{There are }\PYG{l+s+si}{\PYGZob{}\PYGZcb{}}\PYG{l+s+s1}{ entries in model.losses}\PYG{l+s+s1}{\PYGZsq{}}\PYG{o}{.}\PYG{n}{format}\PYG{p}{(}\PYG{n+nb}{len}\PYG{p}{(}\PYG{n}{model}\PYG{o}{.}\PYG{n}{losses}\PYG{p}{)}\PYG{p}{)}\PYG{p}{)}
\PYG{n+nb}{print}\PYG{p}{(}\PYG{n}{model}\PYG{o}{.}\PYG{n}{losses}\PYG{p}{)}
\end{sphinxVerbatim}

\begin{sphinxVerbatim}[commandchars=\\\{\}]
There are 6 entries in model.losses
[\PYGZlt{}tf.Tensor: shape=(), dtype=float32, numpy=0.03536303\PYGZgt{}, \PYGZlt{}tf.Tensor: shape=(), dtype=float32, numpy=0.06506929\PYGZgt{}, \PYGZlt{}tf.Tensor: shape=(), dtype=float32, numpy=0.2647468\PYGZgt{}, \PYGZlt{}tf.Tensor: shape=(), dtype=float32, numpy=0.09337218\PYGZgt{}, \PYGZlt{}tf.Tensor: shape=(), dtype=float32, numpy=0.0795429\PYGZgt{}, \PYGZlt{}tf.Tensor: shape=(), dtype=float32, numpy=0.075103864\PYGZgt{}]
\end{sphinxVerbatim}

Now let’s visualize how the BNN performs for unseen data from the validation set. Ideally, we would like to integrate out the parameters \(\theta\), i.e. marginalize in order to obtain a prediction. Since this is again hard to realize analytically, one usually approximates the integral via sampling from the posterior:
\begin{equation*}
\begin{split}
  \hat{y_{i}}=\int f(x_{i};\theta)q_{\phi}(\theta)d\theta\approx\frac{1}{R}\sum_{r=1}^{R}f(x_{i};\theta_{r})\end{split}
\end{equation*}
where each \(\theta_{r}\) is drawn from \(q_{\phi}(\theta)\). In practice, this just means performing \(R\) forward passes for each input \(x_{i}\) and computing the average. In the same spirit, one can obtain the standard deviation as a measure of uncertainty:
\(\sigma_{i}^{2} = \frac{1}{R-1}\sum_{r=1}^{R}(f(x_{i};\theta)-\hat{y_{i}})^{2}\).

Please note that both \(\hat{y_{i}}\) and \(\sigma_{i}^{2}\) still have shape \(128\times128\times3\), i.e. the mean and variance computations are performed \sphinxstyleemphasis{per\sphinxhyphen{}pixel} (but might be aggregated to a global measure afterwards).

\begin{sphinxVerbatim}[commandchars=\\\{\}]
\PYG{n}{REPS}\PYG{o}{=}\PYG{l+m+mi}{20}
\PYG{n}{preds}\PYG{o}{=}\PYG{n}{np}\PYG{o}{.}\PYG{n}{zeros}\PYG{p}{(}\PYG{n}{shape}\PYG{o}{=}\PYG{p}{(}\PYG{n}{REPS}\PYG{p}{,}\PYG{p}{)}\PYG{o}{+}\PYG{n}{X\PYGZus{}val}\PYG{o}{.}\PYG{n}{shape}\PYG{p}{)}
\PYG{k}{for} \PYG{n}{rep} \PYG{o+ow}{in} \PYG{n+nb}{range}\PYG{p}{(}\PYG{n}{REPS}\PYG{p}{)}\PYG{p}{:}
    \PYG{n}{preds}\PYG{p}{[}\PYG{n}{rep}\PYG{p}{,}\PYG{p}{:}\PYG{p}{,}\PYG{p}{:}\PYG{p}{,}\PYG{p}{:}\PYG{p}{,}\PYG{p}{:}\PYG{p}{]}\PYG{o}{=}\PYG{n}{model}\PYG{o}{.}\PYG{n}{predict}\PYG{p}{(}\PYG{n}{X\PYGZus{}val}\PYG{p}{)}
\PYG{n}{preds\PYGZus{}mean}\PYG{o}{=}\PYG{n}{np}\PYG{o}{.}\PYG{n}{mean}\PYG{p}{(}\PYG{n}{preds}\PYG{p}{,}\PYG{n}{axis}\PYG{o}{=}\PYG{l+m+mi}{0}\PYG{p}{)}
\PYG{n}{preds\PYGZus{}std}\PYG{o}{=}\PYG{n}{np}\PYG{o}{.}\PYG{n}{std}\PYG{p}{(}\PYG{n}{preds}\PYG{p}{,}\PYG{n}{axis}\PYG{o}{=}\PYG{l+m+mi}{0}\PYG{p}{)}
\end{sphinxVerbatim}

Before inspecting the mean and standard deviation computed in the previous cell, let’s visualize one of the outputs of the BNN. In the following plot, the input is shown in the first row, while the second row illustrates the result of a single forward pass.

\begin{sphinxVerbatim}[commandchars=\\\{\}]
\PYG{n}{NUM}\PYG{o}{=}\PYG{l+m+mi}{16}
\PYG{c+c1}{\PYGZsh{} show a single prediction}
\PYG{n}{showSbs}\PYG{p}{(}\PYG{n}{y\PYGZus{}val}\PYG{p}{[}\PYG{n}{NUM}\PYG{p}{]}\PYG{p}{,}\PYG{n}{preds}\PYG{p}{[}\PYG{l+m+mi}{0}\PYG{p}{]}\PYG{p}{[}\PYG{n}{NUM}\PYG{p}{]}\PYG{p}{,} \PYG{n}{top}\PYG{o}{=}\PYG{l+s+s2}{\PYGZdq{}}\PYG{l+s+s2}{Inputs}\PYG{l+s+s2}{\PYGZdq{}}\PYG{p}{,} \PYG{n}{bottom}\PYG{o}{=}\PYG{l+s+s2}{\PYGZdq{}}\PYG{l+s+s2}{Single forward pass}\PYG{l+s+s2}{\PYGZdq{}}\PYG{p}{)}
\end{sphinxVerbatim}

\noindent\sphinxincludegraphics{{bayesian-code_25_0}.png}

If you compare this image to one of the outputs from {\hyperref[\detokenize{supervised-airfoils::doc}]{\sphinxcrossref{\DUrole{doc}{Supervised training for RANS flows around airfoils}}}}, you’ll see that it doesn’t look to different on first sight. This is a good sign, it seems the network learned to produce the content of the pressure and velocity fields.

More importantly, though, we can now visualize the uncertainty over predictions more clearly by inspecting several samples from the posterior distribution  as well as the standard deviation for a given input. Below is code for a function that visualizes precisely that (uncertainty is shown with a different colormap in order to illustrate the differences to previous non\sphinxhyphen{}bayesian notebook).

\begin{sphinxVerbatim}[commandchars=\\\{\}]
\PYG{c+c1}{\PYGZsh{} plot repeated samples from posterior for some observations}
\PYG{k}{def} \PYG{n+nf}{plot\PYGZus{}BNN\PYGZus{}predictions}\PYG{p}{(}\PYG{n}{target}\PYG{p}{,} \PYG{n}{preds}\PYG{p}{,} \PYG{n}{pred\PYGZus{}mean}\PYG{p}{,} \PYG{n}{pred\PYGZus{}std}\PYG{p}{,} \PYG{n}{num\PYGZus{}preds}\PYG{o}{=}\PYG{l+m+mi}{5}\PYG{p}{,}\PYG{n}{channel}\PYG{o}{=}\PYG{l+m+mi}{0}\PYG{p}{)}\PYG{p}{:}
  \PYG{k}{if} \PYG{n}{num\PYGZus{}preds}\PYG{o}{\PYGZgt{}}\PYG{n+nb}{len}\PYG{p}{(}\PYG{n}{preds}\PYG{p}{)}\PYG{p}{:}
    \PYG{n+nb}{print}\PYG{p}{(}\PYG{l+s+s1}{\PYGZsq{}}\PYG{l+s+s1}{num\PYGZus{}preds was set to }\PYG{l+s+si}{\PYGZob{}\PYGZcb{}}\PYG{l+s+s1}{, but has to be smaller than the length of preds. Setting it to }\PYG{l+s+si}{\PYGZob{}\PYGZcb{}}\PYG{l+s+s1}{\PYGZsq{}}\PYG{o}{.}\PYG{n}{format}\PYG{p}{(}\PYG{n}{num\PYGZus{}preds}\PYG{p}{,}\PYG{n+nb}{len}\PYG{p}{(}\PYG{n}{preds}\PYG{p}{)}\PYG{p}{)}\PYG{p}{)}
    \PYG{n}{num\PYGZus{}preds} \PYG{o}{=} \PYG{n+nb}{len}\PYG{p}{(}\PYG{n}{preds}\PYG{p}{)}

  \PYG{c+c1}{\PYGZsh{} transpose and concatenate the frames that are to plot}
  \PYG{n}{to\PYGZus{}plot}\PYG{o}{=}\PYG{n}{np}\PYG{o}{.}\PYG{n}{concatenate}\PYG{p}{(}\PYG{p}{(}\PYG{n}{target}\PYG{p}{[}\PYG{p}{:}\PYG{p}{,}\PYG{p}{:}\PYG{p}{,}\PYG{n}{channel}\PYG{p}{]}\PYG{o}{.}\PYG{n}{transpose}\PYG{p}{(}\PYG{p}{)}\PYG{o}{.}\PYG{n}{reshape}\PYG{p}{(}\PYG{l+m+mi}{128}\PYG{p}{,}\PYG{l+m+mi}{128}\PYG{p}{,}\PYG{l+m+mi}{1}\PYG{p}{)}\PYG{p}{,}\PYG{n}{preds}\PYG{p}{[}\PYG{l+m+mi}{0}\PYG{p}{:}\PYG{n}{num\PYGZus{}preds}\PYG{p}{,}\PYG{p}{:}\PYG{p}{,}\PYG{p}{:}\PYG{p}{,}\PYG{n}{channel}\PYG{p}{]}\PYG{o}{.}\PYG{n}{transpose}\PYG{p}{(}\PYG{p}{)}\PYG{p}{,} 
                          \PYG{n}{pred\PYGZus{}mean}\PYG{p}{[}\PYG{p}{:}\PYG{p}{,}\PYG{p}{:}\PYG{p}{,}\PYG{n}{channel}\PYG{p}{]}\PYG{o}{.}\PYG{n}{transpose}\PYG{p}{(}\PYG{p}{)}\PYG{o}{.}\PYG{n}{reshape}\PYG{p}{(}\PYG{l+m+mi}{128}\PYG{p}{,}\PYG{l+m+mi}{128}\PYG{p}{,}\PYG{l+m+mi}{1}\PYG{p}{)}\PYG{p}{,}\PYG{n}{pred\PYGZus{}std}\PYG{p}{[}\PYG{p}{:}\PYG{p}{,}\PYG{p}{:}\PYG{p}{,}\PYG{n}{channel}\PYG{p}{]}\PYG{o}{.}\PYG{n}{transpose}\PYG{p}{(}\PYG{p}{)}\PYG{o}{.}\PYG{n}{reshape}\PYG{p}{(}\PYG{l+m+mi}{128}\PYG{p}{,}\PYG{l+m+mi}{128}\PYG{p}{,}\PYG{l+m+mi}{1}\PYG{p}{)}\PYG{p}{)}\PYG{p}{,}\PYG{n}{axis}\PYG{o}{=}\PYG{o}{\PYGZhy{}}\PYG{l+m+mi}{1}\PYG{p}{)}
  \PYG{n}{fig}\PYG{p}{,} \PYG{n}{axs} \PYG{o}{=} \PYG{n}{plt}\PYG{o}{.}\PYG{n}{subplots}\PYG{p}{(}\PYG{n}{nrows}\PYG{o}{=}\PYG{l+m+mi}{1}\PYG{p}{,}\PYG{n}{ncols}\PYG{o}{=}\PYG{n}{to\PYGZus{}plot}\PYG{o}{.}\PYG{n}{shape}\PYG{p}{[}\PYG{o}{\PYGZhy{}}\PYG{l+m+mi}{1}\PYG{p}{]}\PYG{p}{,}\PYG{n}{figsize}\PYG{o}{=}\PYG{p}{(}\PYG{l+m+mi}{20}\PYG{p}{,}\PYG{l+m+mi}{4}\PYG{p}{)}\PYG{p}{)}
  \PYG{k}{for} \PYG{n}{i} \PYG{o+ow}{in} \PYG{n+nb}{range}\PYG{p}{(}\PYG{n}{to\PYGZus{}plot}\PYG{o}{.}\PYG{n}{shape}\PYG{p}{[}\PYG{o}{\PYGZhy{}}\PYG{l+m+mi}{1}\PYG{p}{]}\PYG{p}{)}\PYG{p}{:}
    \PYG{n}{label}\PYG{o}{=}\PYG{l+s+s1}{\PYGZsq{}}\PYG{l+s+s1}{Target}\PYG{l+s+s1}{\PYGZsq{}} \PYG{k}{if} \PYG{n}{i}\PYG{o}{==}\PYG{l+m+mi}{0} \PYG{k}{else} \PYG{p}{(}\PYG{l+s+s1}{\PYGZsq{}}\PYG{l+s+s1}{Avg Pred}\PYG{l+s+s1}{\PYGZsq{}} \PYG{k}{if} \PYG{n}{i} \PYG{o}{==} \PYG{p}{(}\PYG{n}{num\PYGZus{}preds}\PYG{o}{+}\PYG{l+m+mi}{1}\PYG{p}{)} \PYG{k}{else} \PYG{p}{(}\PYG{l+s+s1}{\PYGZsq{}}\PYG{l+s+s1}{Std Dev (normalized)}\PYG{l+s+s1}{\PYGZsq{}} \PYG{k}{if} \PYG{n}{i} \PYG{o}{==} \PYG{p}{(}\PYG{n}{num\PYGZus{}preds}\PYG{o}{+}\PYG{l+m+mi}{2}\PYG{p}{)} \PYG{k}{else} \PYG{l+s+s1}{\PYGZsq{}}\PYG{l+s+s1}{Pred }\PYG{l+s+si}{\PYGZob{}\PYGZcb{}}\PYG{l+s+s1}{\PYGZsq{}}\PYG{o}{.}\PYG{n}{format}\PYG{p}{(}\PYG{n}{i}\PYG{p}{)}\PYG{p}{)}\PYG{p}{)}
    \PYG{n}{colmap} \PYG{o}{=} \PYG{n}{cm}\PYG{o}{.}\PYG{n}{viridis} \PYG{k}{if} \PYG{n}{i}\PYG{o}{==}\PYG{n}{to\PYGZus{}plot}\PYG{o}{.}\PYG{n}{shape}\PYG{p}{[}\PYG{o}{\PYGZhy{}}\PYG{l+m+mi}{1}\PYG{p}{]}\PYG{o}{\PYGZhy{}}\PYG{l+m+mi}{1} \PYG{k}{else} \PYG{n}{cm}\PYG{o}{.}\PYG{n}{magma}
    \PYG{n}{frame} \PYG{o}{=} \PYG{n}{np}\PYG{o}{.}\PYG{n}{flipud}\PYG{p}{(}\PYG{n}{to\PYGZus{}plot}\PYG{p}{[}\PYG{p}{:}\PYG{p}{,}\PYG{p}{:}\PYG{p}{,}\PYG{n}{i}\PYG{p}{]}\PYG{p}{)}
    \PYG{n+nb}{min}\PYG{o}{=}\PYG{n}{np}\PYG{o}{.}\PYG{n}{min}\PYG{p}{(}\PYG{n}{frame}\PYG{p}{)}\PYG{p}{;} \PYG{n+nb}{max} \PYG{o}{=} \PYG{n}{np}\PYG{o}{.}\PYG{n}{max}\PYG{p}{(}\PYG{n}{frame}\PYG{p}{)}
    \PYG{n}{frame} \PYG{o}{\PYGZhy{}}\PYG{o}{=} \PYG{n+nb}{min}\PYG{p}{;} \PYG{n}{frame} \PYG{o}{/}\PYG{o}{=}\PYG{p}{(}\PYG{n+nb}{max}\PYG{o}{\PYGZhy{}}\PYG{n+nb}{min}\PYG{p}{)}
    \PYG{n}{axs}\PYG{p}{[}\PYG{n}{i}\PYG{p}{]}\PYG{o}{.}\PYG{n}{imshow}\PYG{p}{(}\PYG{n}{frame}\PYG{p}{,}\PYG{n}{cmap}\PYG{o}{=}\PYG{n}{colmap}\PYG{p}{)}
    \PYG{n}{axs}\PYG{p}{[}\PYG{n}{i}\PYG{p}{]}\PYG{o}{.}\PYG{n}{axis}\PYG{p}{(}\PYG{l+s+s1}{\PYGZsq{}}\PYG{l+s+s1}{off}\PYG{l+s+s1}{\PYGZsq{}}\PYG{p}{)}
    \PYG{n}{axs}\PYG{p}{[}\PYG{n}{i}\PYG{p}{]}\PYG{o}{.}\PYG{n}{set\PYGZus{}title}\PYG{p}{(}\PYG{n}{label}\PYG{p}{)}

\PYG{n}{OBS\PYGZus{}IDX}\PYG{o}{=}\PYG{l+m+mi}{5}
\PYG{n}{plot\PYGZus{}BNN\PYGZus{}predictions}\PYG{p}{(}\PYG{n}{y\PYGZus{}val}\PYG{p}{[}\PYG{n}{OBS\PYGZus{}IDX}\PYG{p}{,}\PYG{o}{.}\PYG{o}{.}\PYG{o}{.}\PYG{p}{]}\PYG{p}{,}\PYG{n}{preds}\PYG{p}{[}\PYG{p}{:}\PYG{p}{,}\PYG{n}{OBS\PYGZus{}IDX}\PYG{p}{,}\PYG{p}{:}\PYG{p}{,}\PYG{p}{:}\PYG{p}{,}\PYG{p}{:}\PYG{p}{]}\PYG{p}{,}\PYG{n}{preds\PYGZus{}mean}\PYG{p}{[}\PYG{n}{OBS\PYGZus{}IDX}\PYG{p}{,}\PYG{o}{.}\PYG{o}{.}\PYG{o}{.}\PYG{p}{]}\PYG{p}{,}\PYG{n}{preds\PYGZus{}std}\PYG{p}{[}\PYG{n}{OBS\PYGZus{}IDX}\PYG{p}{,}\PYG{o}{.}\PYG{o}{.}\PYG{o}{.}\PYG{p}{]}\PYG{p}{)}
\end{sphinxVerbatim}

\noindent\sphinxincludegraphics{{bayesian-code_27_0}.png}

We are looking at channel 0, i.e. the pressure here. One can observe that the dark and bright regions vary quite a bit across predictions. It is reassuring to note that \sphinxhyphen{} at least from visual inspection \sphinxhyphen{} the average (i.e. marginal) prediction is closer to the target than most of the single forward passes.

It should also be noted that each frame was normalized for the visualization. Therefore, when looking at the uncertainty frame, we can infer where the network is uncertain, but now how uncertain it is in absolute values.

In order to assess a global measure of uncertainty we can however compute an average standard deviation over all samples in the validation set.

\begin{sphinxVerbatim}[commandchars=\\\{\}]
\PYG{c+c1}{\PYGZsh{} Average Prediction with total uncertainty}
\PYG{n}{uncertainty\PYGZus{}total} \PYG{o}{=} \PYG{n}{np}\PYG{o}{.}\PYG{n}{mean}\PYG{p}{(}\PYG{n}{np}\PYG{o}{.}\PYG{n}{abs}\PYG{p}{(}\PYG{n}{preds\PYGZus{}std}\PYG{p}{)}\PYG{p}{,}\PYG{n}{axis}\PYG{o}{=}\PYG{p}{(}\PYG{l+m+mi}{0}\PYG{p}{,}\PYG{l+m+mi}{1}\PYG{p}{,}\PYG{l+m+mi}{2}\PYG{p}{)}\PYG{p}{)}
\PYG{n}{preds\PYGZus{}mean\PYGZus{}global} \PYG{o}{=} \PYG{n}{np}\PYG{o}{.}\PYG{n}{mean}\PYG{p}{(}\PYG{n}{np}\PYG{o}{.}\PYG{n}{abs}\PYG{p}{(}\PYG{n}{preds}\PYG{p}{)}\PYG{p}{,}\PYG{n}{axis}\PYG{o}{=}\PYG{p}{(}\PYG{l+m+mi}{0}\PYG{p}{,}\PYG{l+m+mi}{1}\PYG{p}{,}\PYG{l+m+mi}{2}\PYG{p}{,}\PYG{l+m+mi}{3}\PYG{p}{)}\PYG{p}{)}
\PYG{n+nb}{print}\PYG{p}{(}\PYG{l+s+s2}{\PYGZdq{}}\PYG{l+s+se}{\PYGZbs{}n}\PYG{l+s+s2}{Average pixel prediction on validation set: }\PYG{l+s+se}{\PYGZbs{}n}\PYG{l+s+s2}{ pressure: }\PYG{l+s+si}{\PYGZob{}\PYGZcb{}}\PYG{l+s+s2}{ +\PYGZhy{} }\PYG{l+s+si}{\PYGZob{}\PYGZcb{}}\PYG{l+s+s2}{, }\PYG{l+s+se}{\PYGZbs{}n}\PYG{l+s+s2}{ ux: }\PYG{l+s+si}{\PYGZob{}\PYGZcb{}}\PYG{l+s+s2}{ +\PYGZhy{} }\PYG{l+s+si}{\PYGZob{}\PYGZcb{}}\PYG{l+s+s2}{,}\PYG{l+s+se}{\PYGZbs{}n}\PYG{l+s+s2}{ uy: }\PYG{l+s+si}{\PYGZob{}\PYGZcb{}}\PYG{l+s+s2}{ +\PYGZhy{} }\PYG{l+s+si}{\PYGZob{}\PYGZcb{}}\PYG{l+s+s2}{\PYGZdq{}}\PYG{o}{.}\PYG{n}{format}\PYG{p}{(}\PYG{n}{np}\PYG{o}{.}\PYG{n}{round}\PYG{p}{(}\PYG{n}{preds\PYGZus{}mean\PYGZus{}global}\PYG{p}{[}\PYG{l+m+mi}{0}\PYG{p}{]}\PYG{p}{,}\PYG{l+m+mi}{3}\PYG{p}{)}\PYG{p}{,}\PYG{n}{np}\PYG{o}{.}\PYG{n}{round}\PYG{p}{(}\PYG{n}{uncertainty\PYGZus{}total}\PYG{p}{[}\PYG{l+m+mi}{0}\PYG{p}{]}\PYG{p}{,}\PYG{l+m+mi}{3}\PYG{p}{)}\PYG{p}{,}\PYG{n}{np}\PYG{o}{.}\PYG{n}{round}\PYG{p}{(}\PYG{n}{preds\PYGZus{}mean\PYGZus{}global}\PYG{p}{[}\PYG{l+m+mi}{1}\PYG{p}{]}\PYG{p}{,}\PYG{l+m+mi}{3}\PYG{p}{)}\PYG{p}{,}\PYG{n}{np}\PYG{o}{.}\PYG{n}{round}\PYG{p}{(}\PYG{n}{uncertainty\PYGZus{}total}\PYG{p}{[}\PYG{l+m+mi}{1}\PYG{p}{]}\PYG{p}{,}\PYG{l+m+mi}{3}\PYG{p}{)}\PYG{p}{,}\PYG{n}{np}\PYG{o}{.}\PYG{n}{round}\PYG{p}{(}\PYG{n}{preds\PYGZus{}mean\PYGZus{}global}\PYG{p}{[}\PYG{l+m+mi}{2}\PYG{p}{]}\PYG{p}{,}\PYG{l+m+mi}{3}\PYG{p}{)}\PYG{p}{,}\PYG{n}{np}\PYG{o}{.}\PYG{n}{round}\PYG{p}{(}\PYG{n}{uncertainty\PYGZus{}total}\PYG{p}{[}\PYG{l+m+mi}{2}\PYG{p}{]}\PYG{p}{,}\PYG{l+m+mi}{3}\PYG{p}{)}\PYG{p}{)}\PYG{p}{)}
\end{sphinxVerbatim}

\begin{sphinxVerbatim}[commandchars=\\\{\}]
Average pixel prediction on validation set: 
 pressure: 0.025 +\PYGZhy{} 0.009, 
 ux: 0.471 +\PYGZhy{} 0.019,
 uy: 0.081 +\PYGZhy{} 0.016
\end{sphinxVerbatim}

For a run with standard settings, the uncertainties are on the order of 0.01 for all three fields. As the pressure field has a smaller mean, it’s uncertainty is larger in relative terms. This makes sense, as the pressure field is known to be more difficult to predict than the two velocity components.

\section{Test evaluation}
\label{\detokenize{bayesian-code:test-evaluation}}
Like in the case for a conventional neural network, let’s now look at \sphinxstylestrong{proper} test samples, i.e. OOD samples, for which in this case we’ll use new airfoil shapes. These are shapes that the network never saw in any training samples, and hence it tells us a bit about how well the network generalizes to new shapes.

As these samples are at least slightly OOD, we can draw conclusions about how well the network generalizes, which the validation data would not really tell us. In particular, we would like to investigate if the NN is more uncertain when handling OOD data. Like before, we first download the test samples …

\begin{sphinxVerbatim}[commandchars=\\\{\}]
\PYG{k}{if} \PYG{o+ow}{not} \PYG{n}{os}\PYG{o}{.}\PYG{n}{path}\PYG{o}{.}\PYG{n}{isfile}\PYG{p}{(}\PYG{l+s+s1}{\PYGZsq{}}\PYG{l+s+s1}{data\PYGZhy{}airfoils\PYGZhy{}test.npz}\PYG{l+s+s1}{\PYGZsq{}}\PYG{p}{)}\PYG{p}{:}
  \PYG{k+kn}{import} \PYG{n+nn}{urllib}\PYG{n+nn}{.}\PYG{n+nn}{request}
  \PYG{n}{url}\PYG{o}{=}\PYG{l+s+s2}{\PYGZdq{}}\PYG{l+s+s2}{https://physicsbaseddeeplearning.org/data/data\PYGZus{}test.npz}\PYG{l+s+s2}{\PYGZdq{}}
  \PYG{n+nb}{print}\PYG{p}{(}\PYG{l+s+s2}{\PYGZdq{}}\PYG{l+s+s2}{Downloading test data, this should be fast...}\PYG{l+s+s2}{\PYGZdq{}}\PYG{p}{)}
  \PYG{n}{urllib}\PYG{o}{.}\PYG{n}{request}\PYG{o}{.}\PYG{n}{urlretrieve}\PYG{p}{(}\PYG{n}{url}\PYG{p}{,} \PYG{l+s+s1}{\PYGZsq{}}\PYG{l+s+s1}{data\PYGZhy{}airfoils\PYGZhy{}test.npz}\PYG{l+s+s1}{\PYGZsq{}}\PYG{p}{)}

\PYG{n}{nptfile}\PYG{o}{=}\PYG{n}{np}\PYG{o}{.}\PYG{n}{load}\PYG{p}{(}\PYG{l+s+s1}{\PYGZsq{}}\PYG{l+s+s1}{data\PYGZhy{}airfoils\PYGZhy{}test.npz}\PYG{l+s+s1}{\PYGZsq{}}\PYG{p}{)}
\PYG{n+nb}{print}\PYG{p}{(}\PYG{l+s+s2}{\PYGZdq{}}\PYG{l+s+s2}{Loaded }\PYG{l+s+si}{\PYGZob{}\PYGZcb{}}\PYG{l+s+s2}{/}\PYG{l+s+si}{\PYGZob{}\PYGZcb{}}\PYG{l+s+s2}{ test samples}\PYG{l+s+s2}{\PYGZdq{}}\PYG{o}{.}\PYG{n}{format}\PYG{p}{(}\PYG{n+nb}{len}\PYG{p}{(}\PYG{n}{nptfile}\PYG{p}{[}\PYG{l+s+s2}{\PYGZdq{}}\PYG{l+s+s2}{test\PYGZus{}inputs}\PYG{l+s+s2}{\PYGZdq{}}\PYG{p}{]}\PYG{p}{)}\PYG{p}{,}\PYG{n+nb}{len}\PYG{p}{(}\PYG{n}{nptfile}\PYG{p}{[}\PYG{l+s+s2}{\PYGZdq{}}\PYG{l+s+s2}{test\PYGZus{}targets}\PYG{l+s+s2}{\PYGZdq{}}\PYG{p}{]}\PYG{p}{)}\PYG{p}{)}\PYG{p}{)}
\end{sphinxVerbatim}

\begin{sphinxVerbatim}[commandchars=\\\{\}]
Downloading test data, this should be fast...
Loaded 10/10 test samples
\end{sphinxVerbatim}

… and then repeat the procedure from above to evaluate the BNN on the test samples, and compute the marginalized prediction and uncertainty.

\begin{sphinxVerbatim}[commandchars=\\\{\}]
\PYG{n}{X\PYGZus{}test} \PYG{o}{=} \PYG{n}{np}\PYG{o}{.}\PYG{n}{moveaxis}\PYG{p}{(}\PYG{n}{nptfile}\PYG{p}{[}\PYG{l+s+s2}{\PYGZdq{}}\PYG{l+s+s2}{test\PYGZus{}inputs}\PYG{l+s+s2}{\PYGZdq{}}\PYG{p}{]}\PYG{p}{,}\PYG{l+m+mi}{1}\PYG{p}{,}\PYG{o}{\PYGZhy{}}\PYG{l+m+mi}{1}\PYG{p}{)}
\PYG{n}{y\PYGZus{}test} \PYG{o}{=} \PYG{n}{np}\PYG{o}{.}\PYG{n}{moveaxis}\PYG{p}{(}\PYG{n}{nptfile}\PYG{p}{[}\PYG{l+s+s2}{\PYGZdq{}}\PYG{l+s+s2}{test\PYGZus{}targets}\PYG{l+s+s2}{\PYGZdq{}}\PYG{p}{]}\PYG{p}{,}\PYG{l+m+mi}{1}\PYG{p}{,}\PYG{o}{\PYGZhy{}}\PYG{l+m+mi}{1}\PYG{p}{)}

\PYG{n}{REPS}\PYG{o}{=}\PYG{l+m+mi}{10}
\PYG{n}{preds\PYGZus{}test}\PYG{o}{=}\PYG{n}{np}\PYG{o}{.}\PYG{n}{zeros}\PYG{p}{(}\PYG{n}{shape}\PYG{o}{=}\PYG{p}{(}\PYG{n}{REPS}\PYG{p}{,}\PYG{p}{)}\PYG{o}{+}\PYG{n}{X\PYGZus{}test}\PYG{o}{.}\PYG{n}{shape}\PYG{p}{)}
\PYG{k}{for} \PYG{n}{rep} \PYG{o+ow}{in} \PYG{n+nb}{range}\PYG{p}{(}\PYG{n}{REPS}\PYG{p}{)}\PYG{p}{:}
    \PYG{n}{preds\PYGZus{}test}\PYG{p}{[}\PYG{n}{rep}\PYG{p}{,}\PYG{p}{:}\PYG{p}{,}\PYG{p}{:}\PYG{p}{,}\PYG{p}{:}\PYG{p}{,}\PYG{p}{:}\PYG{p}{]}\PYG{o}{=}\PYG{n}{model}\PYG{o}{.}\PYG{n}{predict}\PYG{p}{(}\PYG{n}{X\PYGZus{}test}\PYG{p}{)}
\PYG{n}{preds\PYGZus{}test\PYGZus{}mean}\PYG{o}{=}\PYG{n}{np}\PYG{o}{.}\PYG{n}{mean}\PYG{p}{(}\PYG{n}{preds\PYGZus{}test}\PYG{p}{,}\PYG{n}{axis}\PYG{o}{=}\PYG{l+m+mi}{0}\PYG{p}{)}
\PYG{n}{preds\PYGZus{}test\PYGZus{}std}\PYG{o}{=}\PYG{n}{np}\PYG{o}{.}\PYG{n}{std}\PYG{p}{(}\PYG{n}{preds\PYGZus{}test}\PYG{p}{,}\PYG{n}{axis}\PYG{o}{=}\PYG{l+m+mi}{0}\PYG{p}{)}
\PYG{n}{test\PYGZus{}loss} \PYG{o}{=} \PYG{n}{tf}\PYG{o}{.}\PYG{n}{reduce\PYGZus{}mean}\PYG{p}{(}\PYG{n}{mae}\PYG{p}{(}\PYG{n}{preds\PYGZus{}test\PYGZus{}mean}\PYG{p}{,} \PYG{n}{y\PYGZus{}test}\PYG{p}{)}\PYG{p}{)}

\PYG{n+nb}{print}\PYG{p}{(}\PYG{l+s+s2}{\PYGZdq{}}\PYG{l+s+se}{\PYGZbs{}n}\PYG{l+s+s2}{Average test error: }\PYG{l+s+si}{\PYGZob{}\PYGZcb{}}\PYG{l+s+s2}{\PYGZdq{}}\PYG{o}{.}\PYG{n}{format}\PYG{p}{(}\PYG{n}{test\PYGZus{}loss}\PYG{p}{)}\PYG{p}{)}
\end{sphinxVerbatim}

\begin{sphinxVerbatim}[commandchars=\\\{\}]
Average test error: 0.023046530292471824
\end{sphinxVerbatim}

\begin{sphinxVerbatim}[commandchars=\\\{\}]
\PYG{c+c1}{\PYGZsh{} Average Prediction with total uncertainty}
\PYG{n}{uncertainty\PYGZus{}test\PYGZus{}total} \PYG{o}{=} \PYG{n}{np}\PYG{o}{.}\PYG{n}{mean}\PYG{p}{(}\PYG{n}{np}\PYG{o}{.}\PYG{n}{abs}\PYG{p}{(}\PYG{n}{preds\PYGZus{}test\PYGZus{}std}\PYG{p}{)}\PYG{p}{,}\PYG{n}{axis}\PYG{o}{=}\PYG{p}{(}\PYG{l+m+mi}{0}\PYG{p}{,}\PYG{l+m+mi}{1}\PYG{p}{,}\PYG{l+m+mi}{2}\PYG{p}{)}\PYG{p}{)}
\PYG{n}{preds\PYGZus{}test\PYGZus{}mean\PYGZus{}global} \PYG{o}{=} \PYG{n}{np}\PYG{o}{.}\PYG{n}{mean}\PYG{p}{(}\PYG{n}{np}\PYG{o}{.}\PYG{n}{abs}\PYG{p}{(}\PYG{n}{preds\PYGZus{}test}\PYG{p}{)}\PYG{p}{,}\PYG{n}{axis}\PYG{o}{=}\PYG{p}{(}\PYG{l+m+mi}{0}\PYG{p}{,}\PYG{l+m+mi}{1}\PYG{p}{,}\PYG{l+m+mi}{2}\PYG{p}{,}\PYG{l+m+mi}{3}\PYG{p}{)}\PYG{p}{)}
\PYG{n+nb}{print}\PYG{p}{(}\PYG{l+s+s2}{\PYGZdq{}}\PYG{l+s+se}{\PYGZbs{}n}\PYG{l+s+s2}{Average pixel prediction on test set: }\PYG{l+s+se}{\PYGZbs{}n}\PYG{l+s+s2}{ pressure: }\PYG{l+s+si}{\PYGZob{}\PYGZcb{}}\PYG{l+s+s2}{ +\PYGZhy{} }\PYG{l+s+si}{\PYGZob{}\PYGZcb{}}\PYG{l+s+s2}{, }\PYG{l+s+se}{\PYGZbs{}n}\PYG{l+s+s2}{ ux: }\PYG{l+s+si}{\PYGZob{}\PYGZcb{}}\PYG{l+s+s2}{ +\PYGZhy{} }\PYG{l+s+si}{\PYGZob{}\PYGZcb{}}\PYG{l+s+s2}{,}\PYG{l+s+se}{\PYGZbs{}n}\PYG{l+s+s2}{ uy: }\PYG{l+s+si}{\PYGZob{}\PYGZcb{}}\PYG{l+s+s2}{ +\PYGZhy{} }\PYG{l+s+si}{\PYGZob{}\PYGZcb{}}\PYG{l+s+s2}{\PYGZdq{}}\PYG{o}{.}\PYG{n}{format}\PYG{p}{(}\PYG{n}{np}\PYG{o}{.}\PYG{n}{round}\PYG{p}{(}\PYG{n}{preds\PYGZus{}test\PYGZus{}mean\PYGZus{}global}\PYG{p}{[}\PYG{l+m+mi}{0}\PYG{p}{]}\PYG{p}{,}\PYG{l+m+mi}{3}\PYG{p}{)}\PYG{p}{,}\PYG{n}{np}\PYG{o}{.}\PYG{n}{round}\PYG{p}{(}\PYG{n}{uncertainty\PYGZus{}test\PYGZus{}total}\PYG{p}{[}\PYG{l+m+mi}{0}\PYG{p}{]}\PYG{p}{,}\PYG{l+m+mi}{3}\PYG{p}{)}\PYG{p}{,}\PYG{n}{np}\PYG{o}{.}\PYG{n}{round}\PYG{p}{(}\PYG{n}{preds\PYGZus{}test\PYGZus{}mean\PYGZus{}global}\PYG{p}{[}\PYG{l+m+mi}{1}\PYG{p}{]}\PYG{p}{,}\PYG{l+m+mi}{3}\PYG{p}{)}\PYG{p}{,}\PYG{n}{np}\PYG{o}{.}\PYG{n}{round}\PYG{p}{(}\PYG{n}{uncertainty\PYGZus{}test\PYGZus{}total}\PYG{p}{[}\PYG{l+m+mi}{1}\PYG{p}{]}\PYG{p}{,}\PYG{l+m+mi}{3}\PYG{p}{)}\PYG{p}{,}\PYG{n}{np}\PYG{o}{.}\PYG{n}{round}\PYG{p}{(}\PYG{n}{preds\PYGZus{}test\PYGZus{}mean\PYGZus{}global}\PYG{p}{[}\PYG{l+m+mi}{2}\PYG{p}{]}\PYG{p}{,}\PYG{l+m+mi}{3}\PYG{p}{)}\PYG{p}{,}\PYG{n}{np}\PYG{o}{.}\PYG{n}{round}\PYG{p}{(}\PYG{n}{uncertainty\PYGZus{}test\PYGZus{}total}\PYG{p}{[}\PYG{l+m+mi}{2}\PYG{p}{]}\PYG{p}{,}\PYG{l+m+mi}{3}\PYG{p}{)}\PYG{p}{)}\PYG{p}{)}
\end{sphinxVerbatim}

\begin{sphinxVerbatim}[commandchars=\\\{\}]
Average pixel prediction on test set: 
 pressure: 0.03 +\PYGZhy{} 0.012, 
 ux: 0.466 +\PYGZhy{} 0.024,
 uy: 0.091 +\PYGZhy{} 0.02
\end{sphinxVerbatim}

This is reassuring: The uncertainties on the OOD test set with new shapes are at least slightly higher than on the validation set.

\subsection{Visualizations}
\label{\detokenize{bayesian-code:visualizations}}
The following graph visualizes these measurements: it shows the mean absolute errors for validation and test sets side by side, together with the uncertainties of the predictions as error bars:

\begin{sphinxVerbatim}[commandchars=\\\{\}]
\PYG{c+c1}{\PYGZsh{} plot per channel MAE with uncertainty}
\PYG{n}{val\PYGZus{}loss\PYGZus{}c}\PYG{p}{,} \PYG{n}{test\PYGZus{}loss\PYGZus{}c} \PYG{o}{=} \PYG{p}{[}\PYG{p}{]}\PYG{p}{,} \PYG{p}{[}\PYG{p}{]}
\PYG{k}{for} \PYG{n}{channel} \PYG{o+ow}{in} \PYG{n+nb}{range}\PYG{p}{(}\PYG{l+m+mi}{3}\PYG{p}{)}\PYG{p}{:}
  \PYG{n}{val\PYGZus{}loss\PYGZus{}c}\PYG{o}{.}\PYG{n}{append}\PYG{p}{(} \PYG{n}{tf}\PYG{o}{.}\PYG{n}{reduce\PYGZus{}mean}\PYG{p}{(}\PYG{n}{mae}\PYG{p}{(}\PYG{n}{preds\PYGZus{}mean}\PYG{p}{[}\PYG{o}{.}\PYG{o}{.}\PYG{o}{.}\PYG{p}{,}\PYG{n}{channel}\PYG{p}{]}\PYG{p}{,} \PYG{n}{y\PYGZus{}val}\PYG{p}{[}\PYG{o}{.}\PYG{o}{.}\PYG{o}{.}\PYG{p}{,}\PYG{n}{channel}\PYG{p}{]}\PYG{p}{)}\PYG{p}{)} \PYG{p}{)}
  \PYG{n}{test\PYGZus{}loss\PYGZus{}c}\PYG{o}{.}\PYG{n}{append}\PYG{p}{(} \PYG{n}{tf}\PYG{o}{.}\PYG{n}{reduce\PYGZus{}mean}\PYG{p}{(}\PYG{n}{mae}\PYG{p}{(}\PYG{n}{preds\PYGZus{}test\PYGZus{}mean}\PYG{p}{[}\PYG{o}{.}\PYG{o}{.}\PYG{o}{.}\PYG{p}{,}\PYG{n}{channel}\PYG{p}{]}\PYG{p}{,} \PYG{n}{y\PYGZus{}test}\PYG{p}{[}\PYG{o}{.}\PYG{o}{.}\PYG{o}{.}\PYG{p}{,}\PYG{n}{channel}\PYG{p}{]}\PYG{p}{)}\PYG{p}{)} \PYG{p}{)}

\PYG{n}{fig}\PYG{p}{,} \PYG{n}{ax} \PYG{o}{=} \PYG{n}{plt}\PYG{o}{.}\PYG{n}{subplots}\PYG{p}{(}\PYG{p}{)}
\PYG{n}{ind} \PYG{o}{=} \PYG{n}{np}\PYG{o}{.}\PYG{n}{arange}\PYG{p}{(}\PYG{n+nb}{len}\PYG{p}{(}\PYG{n}{val\PYGZus{}loss\PYGZus{}c}\PYG{p}{)}\PYG{p}{)}\PYG{p}{;} \PYG{n}{width}\PYG{o}{=}\PYG{l+m+mf}{0.3}
\PYG{n}{bars1} \PYG{o}{=} \PYG{n}{ax}\PYG{o}{.}\PYG{n}{bar}\PYG{p}{(}\PYG{n}{ind} \PYG{o}{\PYGZhy{}} \PYG{n}{width}\PYG{o}{/}\PYG{l+m+mi}{2}\PYG{p}{,} \PYG{n}{val\PYGZus{}loss\PYGZus{}c}\PYG{p}{,} \PYG{n}{width}\PYG{p}{,} \PYG{n}{yerr}\PYG{o}{=}\PYG{n}{uncertainty\PYGZus{}total}\PYG{p}{,} \PYG{n}{capsize}\PYG{o}{=}\PYG{l+m+mi}{4}\PYG{p}{,} \PYG{n}{label}\PYG{o}{=}\PYG{l+s+s2}{\PYGZdq{}}\PYG{l+s+s2}{validation}\PYG{l+s+s2}{\PYGZdq{}}\PYG{p}{)}
\PYG{n}{bars2} \PYG{o}{=} \PYG{n}{ax}\PYG{o}{.}\PYG{n}{bar}\PYG{p}{(}\PYG{n}{ind} \PYG{o}{+} \PYG{n}{width}\PYG{o}{/}\PYG{l+m+mi}{2}\PYG{p}{,} \PYG{n}{test\PYGZus{}loss\PYGZus{}c}\PYG{p}{,} \PYG{n}{width}\PYG{p}{,} \PYG{n}{yerr}\PYG{o}{=}\PYG{n}{uncertainty\PYGZus{}test\PYGZus{}total}\PYG{p}{,} \PYG{n}{capsize}\PYG{o}{=}\PYG{l+m+mi}{4}\PYG{p}{,} \PYG{n}{label}\PYG{o}{=}\PYG{l+s+s2}{\PYGZdq{}}\PYG{l+s+s2}{test}\PYG{l+s+s2}{\PYGZdq{}}\PYG{p}{)}
\PYG{n}{ax}\PYG{o}{.}\PYG{n}{set\PYGZus{}ylabel}\PYG{p}{(}\PYG{l+s+s2}{\PYGZdq{}}\PYG{l+s+s2}{MAE \PYGZam{} Uncertainty}\PYG{l+s+s2}{\PYGZdq{}}\PYG{p}{)}
\PYG{n}{ax}\PYG{o}{.}\PYG{n}{set\PYGZus{}xticks}\PYG{p}{(}\PYG{n}{ind}\PYG{p}{)}\PYG{p}{;} \PYG{n}{ax}\PYG{o}{.}\PYG{n}{set\PYGZus{}xticklabels}\PYG{p}{(}\PYG{p}{(}\PYG{l+s+s1}{\PYGZsq{}}\PYG{l+s+s1}{P}\PYG{l+s+s1}{\PYGZsq{}}\PYG{p}{,} \PYG{l+s+s1}{\PYGZsq{}}\PYG{l+s+s1}{u\PYGZus{}x}\PYG{l+s+s1}{\PYGZsq{}}\PYG{p}{,} \PYG{l+s+s1}{\PYGZsq{}}\PYG{l+s+s1}{u\PYGZus{}y}\PYG{l+s+s1}{\PYGZsq{}}\PYG{p}{)}\PYG{p}{)}
\PYG{n}{ax}\PYG{o}{.}\PYG{n}{legend}\PYG{p}{(}\PYG{p}{)}\PYG{p}{;} \PYG{n}{plt}\PYG{o}{.}\PYG{n}{tight\PYGZus{}layout}\PYG{p}{(}\PYG{p}{)}
\end{sphinxVerbatim}

\noindent\sphinxincludegraphics{{bayesian-code_37_0}.png}

The mean error is clearly larger, and the slightly larger uncertainties of the predictions are likewise visible via the error bars.

In general it is hard to obtain a calibrated uncertainty estimate, but since we are dealing with a fairly simple problem here, the BNN is able to estimate the uncertainty reasonably well.

The next graph shows the differences of the BNN predictions for a single case of the test set (using the same style as for the validation sample above):

\begin{sphinxVerbatim}[commandchars=\\\{\}]
\PYG{n}{OBS\PYGZus{}IDX}\PYG{o}{=}\PYG{l+m+mi}{5}
\PYG{n}{plot\PYGZus{}BNN\PYGZus{}predictions}\PYG{p}{(}\PYG{n}{y\PYGZus{}test}\PYG{p}{[}\PYG{n}{OBS\PYGZus{}IDX}\PYG{p}{,}\PYG{o}{.}\PYG{o}{.}\PYG{o}{.}\PYG{p}{]}\PYG{p}{,}\PYG{n}{preds\PYGZus{}test}\PYG{p}{[}\PYG{p}{:}\PYG{p}{,}\PYG{n}{OBS\PYGZus{}IDX}\PYG{p}{,}\PYG{p}{:}\PYG{p}{,}\PYG{p}{:}\PYG{p}{,}\PYG{p}{:}\PYG{p}{]}\PYG{p}{,}\PYG{n}{preds\PYGZus{}test\PYGZus{}mean}\PYG{p}{[}\PYG{n}{OBS\PYGZus{}IDX}\PYG{p}{,}\PYG{o}{.}\PYG{o}{.}\PYG{o}{.}\PYG{p}{]}\PYG{p}{,}\PYG{n}{preds\PYGZus{}test\PYGZus{}std}\PYG{p}{[}\PYG{n}{OBS\PYGZus{}IDX}\PYG{p}{,}\PYG{o}{.}\PYG{o}{.}\PYG{o}{.}\PYG{p}{]}\PYG{p}{)}
\end{sphinxVerbatim}

\noindent\sphinxincludegraphics{{bayesian-code_39_0}.png}

We can also visualize several shapes from the test set together with the corresponding marginalized prediction and uncertainty.

\begin{sphinxVerbatim}[commandchars=\\\{\}]
\PYG{n}{IDXS} \PYG{o}{=} \PYG{p}{[}\PYG{l+m+mi}{1}\PYG{p}{,}\PYG{l+m+mi}{3}\PYG{p}{,}\PYG{l+m+mi}{8}\PYG{p}{]}
\PYG{n}{CHANNEL} \PYG{o}{=} \PYG{l+m+mi}{0}
\PYG{n}{fig}\PYG{p}{,} \PYG{n}{axs} \PYG{o}{=} \PYG{n}{plt}\PYG{o}{.}\PYG{n}{subplots}\PYG{p}{(}\PYG{n}{nrows}\PYG{o}{=}\PYG{n+nb}{len}\PYG{p}{(}\PYG{n}{IDXS}\PYG{p}{)}\PYG{p}{,}\PYG{n}{ncols}\PYG{o}{=}\PYG{l+m+mi}{3}\PYG{p}{,}\PYG{n}{sharex}\PYG{o}{=}\PYG{k+kc}{True}\PYG{p}{,} \PYG{n}{sharey} \PYG{o}{=} \PYG{k+kc}{True}\PYG{p}{,} \PYG{n}{figsize} \PYG{o}{=} \PYG{p}{(}\PYG{l+m+mi}{9}\PYG{p}{,}\PYG{n+nb}{len}\PYG{p}{(}\PYG{n}{IDXS}\PYG{p}{)}\PYG{o}{*}\PYG{l+m+mi}{3}\PYG{p}{)}\PYG{p}{)}
\PYG{k}{for} \PYG{n}{i}\PYG{p}{,} \PYG{n}{idx} \PYG{o+ow}{in} \PYG{n+nb}{enumerate}\PYG{p}{(}\PYG{n}{IDXS}\PYG{p}{)}\PYG{p}{:}
  \PYG{n}{axs}\PYG{p}{[}\PYG{n}{i}\PYG{p}{]}\PYG{p}{[}\PYG{l+m+mi}{0}\PYG{p}{]}\PYG{o}{.}\PYG{n}{imshow}\PYG{p}{(}\PYG{n}{np}\PYG{o}{.}\PYG{n}{flipud}\PYG{p}{(}\PYG{n}{X\PYGZus{}test}\PYG{p}{[}\PYG{n}{idx}\PYG{p}{,}\PYG{p}{:}\PYG{p}{,}\PYG{p}{:}\PYG{p}{,}\PYG{n}{CHANNEL}\PYG{p}{]}\PYG{o}{.}\PYG{n}{transpose}\PYG{p}{(}\PYG{p}{)}\PYG{p}{)}\PYG{p}{,} \PYG{n}{cmap}\PYG{o}{=}\PYG{n}{cm}\PYG{o}{.}\PYG{n}{magma}\PYG{p}{)}
  \PYG{n}{axs}\PYG{p}{[}\PYG{n}{i}\PYG{p}{]}\PYG{p}{[}\PYG{l+m+mi}{1}\PYG{p}{]}\PYG{o}{.}\PYG{n}{imshow}\PYG{p}{(}\PYG{n}{np}\PYG{o}{.}\PYG{n}{flipud}\PYG{p}{(}\PYG{n}{preds\PYGZus{}test\PYGZus{}mean}\PYG{p}{[}\PYG{n}{idx}\PYG{p}{,}\PYG{p}{:}\PYG{p}{,}\PYG{p}{:}\PYG{p}{,}\PYG{n}{CHANNEL}\PYG{p}{]}\PYG{o}{.}\PYG{n}{transpose}\PYG{p}{(}\PYG{p}{)}\PYG{p}{)}\PYG{p}{,} \PYG{n}{cmap}\PYG{o}{=}\PYG{n}{cm}\PYG{o}{.}\PYG{n}{magma}\PYG{p}{)}
  \PYG{n}{axs}\PYG{p}{[}\PYG{n}{i}\PYG{p}{]}\PYG{p}{[}\PYG{l+m+mi}{2}\PYG{p}{]}\PYG{o}{.}\PYG{n}{imshow}\PYG{p}{(}\PYG{n}{np}\PYG{o}{.}\PYG{n}{flipud}\PYG{p}{(}\PYG{n}{preds\PYGZus{}test\PYGZus{}std}\PYG{p}{[}\PYG{n}{idx}\PYG{p}{,}\PYG{p}{:}\PYG{p}{,}\PYG{p}{:}\PYG{p}{,}\PYG{n}{CHANNEL}\PYG{p}{]}\PYG{o}{.}\PYG{n}{transpose}\PYG{p}{(}\PYG{p}{)}\PYG{p}{)}\PYG{p}{,} \PYG{n}{cmap}\PYG{o}{=}\PYG{n}{cm}\PYG{o}{.}\PYG{n}{viridis}\PYG{p}{)}
\PYG{n}{axs}\PYG{p}{[}\PYG{l+m+mi}{0}\PYG{p}{]}\PYG{p}{[}\PYG{l+m+mi}{0}\PYG{p}{]}\PYG{o}{.}\PYG{n}{set\PYGZus{}title}\PYG{p}{(}\PYG{l+s+s1}{\PYGZsq{}}\PYG{l+s+s1}{Shape}\PYG{l+s+s1}{\PYGZsq{}}\PYG{p}{)}
\PYG{n}{axs}\PYG{p}{[}\PYG{l+m+mi}{0}\PYG{p}{]}\PYG{p}{[}\PYG{l+m+mi}{1}\PYG{p}{]}\PYG{o}{.}\PYG{n}{set\PYGZus{}title}\PYG{p}{(}\PYG{l+s+s1}{\PYGZsq{}}\PYG{l+s+s1}{Avg Pred}\PYG{l+s+s1}{\PYGZsq{}}\PYG{p}{)}
\PYG{n}{axs}\PYG{p}{[}\PYG{l+m+mi}{0}\PYG{p}{]}\PYG{p}{[}\PYG{l+m+mi}{2}\PYG{p}{]}\PYG{o}{.}\PYG{n}{set\PYGZus{}title}\PYG{p}{(}\PYG{l+s+s1}{\PYGZsq{}}\PYG{l+s+s1}{Std. Dev}\PYG{l+s+s1}{\PYGZsq{}}\PYG{p}{)}
\end{sphinxVerbatim}

\begin{sphinxVerbatim}[commandchars=\\\{\}]
Text(0.5, 1.0, \PYGZsq{}Std. Dev\PYGZsq{})
\end{sphinxVerbatim}

\noindent\sphinxincludegraphics{{bayesian-code_41_1}.png}

As we can see, the shapes from the test set differ quite a bit from another. Nevertheless, the uncertainty estimate is reasonably distributed. It is especially high in the boundary layer around the airfoil, and in the regions of the low pressure pocket.

\subsection{Discussion}
\label{\detokenize{bayesian-code:discussion}}
Despite these promising results, there are still several issues with Bayesian Neural Nets, limiting their use in many practical applications. One serious drawback is the need for additional scaling of the KL\sphinxhyphen{}loss and the fact that there is no convincing argument on why it is necessary yet (read e.g. \sphinxhref{http://proceedings.mlr.press/v119/wenzel20a/wenzel20a.pdf}{here} or \sphinxhref{https://arxiv.org/abs/2008.05912}{here}).
Furthermore, some people think that assuming independent normal distributions as variational approximations to the posterior is an oversimplification since in practice the weights are actually highly correlated (\sphinxhref{https://arxiv.org/abs/1909.00719}{paper}). Other people instead argue that this might not be an issue, as long as the networks in use are deep enough (\sphinxhref{https://arxiv.org/abs/2002.03704}{paper}). On top of that, there is research on different (e.g. heavy\sphinxhyphen{}tailed) priors other than normals and many other aspects of BNNs.

\section{Next steps}
\label{\detokenize{bayesian-code:next-steps}}
But now it’s time to experiment with BNNs yourself.
\begin{itemize}
\item {} 
One interesting thing to look at is how the behavior of our BNN changes, if we adjust the KL\sphinxhyphen{}prefactor. In the training loop above we set it to 5000 without further justification. You can check out what happens, if you use a value of 1, as it is suggested by the theory, instead of 5000. According to our implementation, this should make the network ‘more bayesian’, since we assign larger importance to the KL\sphinxhyphen{}divergence than before.

\item {} 
So far, we have only worked with variational BNNs, implemented via TensorFlows probabilistic layers. Recall that there is a simpler way of getting uncertainty estimates: Using dropout not only at training, but also at inference time. You can check out how the outputs change for that case. In order to do so, you can, for instance, just pass a non\sphinxhyphen{}zero dropout rate to the network specification and change the prediction phase in the above implementation from \sphinxstyleemphasis{model.predict(…)} to \sphinxstyleemphasis{model(…, training=True)}. Setting the \sphinxstyleemphasis{training=True} flag will tell TensorFlow to forward the input as if it were training data and hence, it will apply dropout. Please note that the \sphinxstyleemphasis{training=True} flag can also affect other features of the network. Batch normalization, for instance, works differently in training and prediction mode. As long as we don’t deal with overly different data and use sufficiently large batch\sphinxhyphen{}sizes, this should not introduce large errors, though. Sensible dropout rates to start experimenting with are e.g. around 0.1.

\end{itemize}

\part{Fast Forward Topics}

\chapter{Additional Topics}
\label{\detokenize{others-intro:additional-topics}}\label{\detokenize{others-intro::doc}}
The next sections will give a shorter introduction to other topics that are highly
interesting in the context of physics\sphinxhyphen{}based deep learning. These topics (for now) do
not come with executable notebooks, but we will still point to existing open source
implementations for each of them.

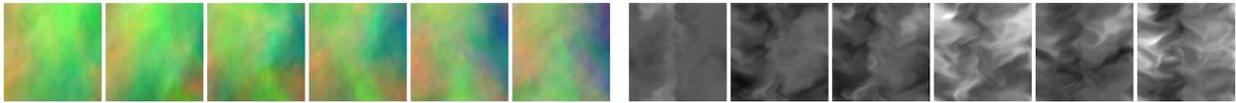
\sphinxincludegraphics{{divider4}.jpg}

More specifically, we will look at:
\begin{itemize}
\item {} 
Model reduction and time series predictions, i.e., using to DL predict the evolution of a physical system in a latent space.	
This typically replaces a numerical solver, and we can make use of special techniques from the DL area that target time series.

\item {} 
Generative models are likewise an own topic in DL, and here especially generative adversarial networks were shown to be powerful tools. They also represent a highly interesting training approach involving to separate NNs.

\item {} 
Meshless methods and unstructured meshes are an important topic for classical simulations. Here, we’ll look at a specific Lagrangian method that employs learning in the context of dynamic, particle\sphinxhyphen{}based representations.

\end{itemize}

\chapter{Model Reduction and Time Series}
\label{\detokenize{others-timeseries:model-reduction-and-time-series}}\label{\detokenize{others-timeseries::doc}}
An inherent challenge for many practical PDE solvers is the large dimensionality of the resulting problems.
Our model \(\mathcal{P}\) is typically discretized with \(\mathcal{O}(n^3)\) samples for a 3 dimensional
problem (with \(n\) denoting the number of samples along one axis),
and for time\sphinxhyphen{}dependent phenomena we additionally have a discretization along
time. The latter typically scales in accordance to the spatial dimensions. This gives an
overall samples count on the order of \(\mathcal{O}(n^4)\). Not surprisingly,
the workload in these situations quickly explodes for larger \(n\) (and for all practical high\sphinxhyphen{}fidelity applications we want \(n\) to be as large as possible).

One popular way to reduce the complexity is to map a spatial state of our system \(\mathbf{s_t} \in \mathbb{R}^{n^3}\)
into a much lower dimensional state \(\mathbf{c_t} \in \mathbb{R}^{m}\), with \(m \ll n^3\). Within this latent space,
we estimate the evolution of our system by inferring a new state \(\mathbf{c_{t+1}}\), which we then decode to obtain \(\mathbf{s_{t+1}}\). In order for this to work, it’s crucial that we can choose \(m\) large enough that it captures all important structures in our solution manifold, and that the time prediction of \(\mathbf{c_{t+1}}\) can be computed efficiently, such that we obtain a gain in performance despite the additional encoding and decoding steps. In practice, the explosion in terms of unknowns for regular simulations (the \(\mathcal{O}(n^3)\) above) coupled with a super\sphinxhyphen{}linear complexity for computing a new state \(\mathbf{s_t}\) directly makes this approach very expensive, while working with the latent space points \(\mathbf{c}\) very quickly pays off for small \(m\).

However, it’s crucial that encoder and decoder do a good job at reducing the dimensionality of the problem. This is a very good task for DL approaches. Furthermore, we then need a time evolution of the latent space states \(\mathbf{c}\), and for most practical model equations, we cannot find closed form solutions to evolve \(\mathbf{c}\). Hence, this likewise poses a very good problem for DL. To summarize, we’re facing two challenges: learning a good spatial encoding and decoding, together with learning an accurate time evolution.
Below, we will describe an approach to solve this problem following Wiewel et al.
{[}\hyperlink{cite.references:id18}{WBT19}{]} \& {[}\hyperlink{cite.references:id8}{WKA+20}{]}, which in turn employs
the encoder/decoder of Kim et al. {[}\hyperlink{cite.references:id19}{KAT+19}{]}.

\begin{figure}[htbp]
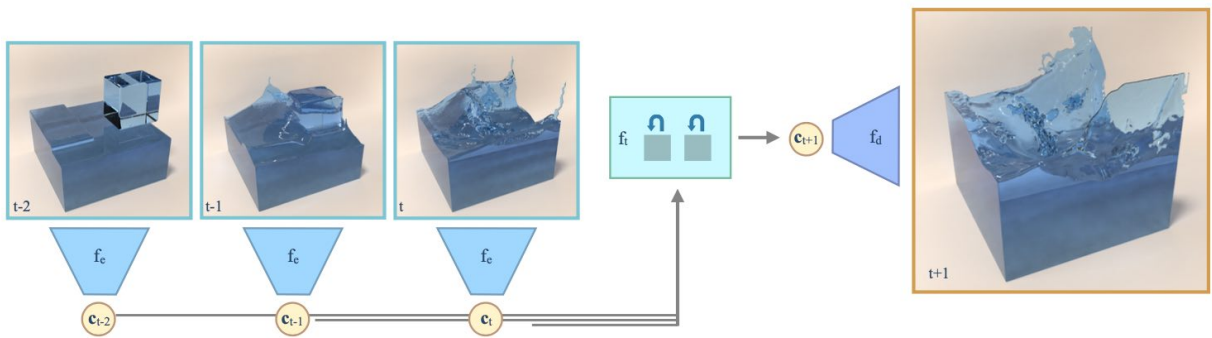

\centering
\capstart

\noindent\sphinxincludegraphics{{others-timeseries-lsp-overview}.jpg}
\caption{For time series predictions with ROMs, we encode the state of our system with an encoder \(f_e\), predict
the time evolution with \(f_t\), and then decode the full spatial information with a decoder \(f_d\).}\label{\detokenize{others-timeseries:timeseries-lsp-overview}}\end{figure}

\section{Reduced order models}
\label{\detokenize{others-timeseries:reduced-order-models}}
Reducing the dimension and complexity of computational models, often called \sphinxstyleemphasis{reduced order modeling} (ROM) or \sphinxstyleemphasis{model reduction}, is a classic topic in the computational field. Traditional techniques often employ techniques such as principal component analysis to arrive at a basis for a chosen space of solution. However, being linear by construction, these approaches have inherent limitations when representing complex, non\sphinxhyphen{}linear solution manifolds. In practice, all “interesting” solutions are highly non\sphinxhyphen{}linear, and hence DL has received a substantial amount of interest as a way to learn non\sphinxhyphen{}linear representations. Due to the non\sphinxhyphen{}linearity, DL representations can potentially yield a high accuracy with fewer degrees of freedom in the reduced model compared to classic approaches.

The canonical NN for reduced models is an \sphinxstyleemphasis{autoencoder}. This denotes a network whose sole task is to reconstruct a given input \(x\) while passing it through a bottleneck that is typically located in or near the middle of the stack of layers of the NN. The data in the bottleneck then represents the compressed, latent space representation \(\mathbf{c}\). The part of the network leading up to the bottleneck  \(\mathbf{c}\) is the encoder \(f_e\), and the part after it the decoder \(f_d\). In combination, the learning task can be written as
\begin{equation*}
\begin{split}
\text{arg min}_{\theta_e,\theta_d} | f_d( f_e(\mathbf{s};\theta_e) ;\theta_d) - \mathbf{s} |_2^2
\end{split}
\end{equation*}
with the encoder
\(f_e: \mathbb{R}^{n^3} \rightarrow \mathbb{R}^{m}\) with weights \(\theta_e\),
and the decoder
\(f_d: \mathbb{R}^{m} \rightarrow \mathbb{R}^{n^3}\) with weights \(\theta_d\). For this
learning objective we do not require any other data than the \(\mathbf{s}\), as these represent
inputs as well as the reference outputs.

Autoencoder networks are typically realized as stacks of convolutional layers.
While the details of these layers can be chosen flexibly, a key property of all
autoencoder architectures is that no connection between encoder and decoder part may
exist. Hence, the network has to be separable for encoder and decoder.
This is natural, as any connections (or information) shared between encoder and decoder
would prevent using the encoder or decoder in a standalone manner. E.g., the decoder has to be able to decode a full state \(\mathbf{s}\) purely from a latent space point \(\mathbf{c}\).

\subsection{Autoencoder variants}
\label{\detokenize{others-timeseries:autoencoder-variants}}
One popular variant of autoencoders is worth a mention here: the so\sphinxhyphen{}called \sphinxstyleemphasis{variational autoencoders}, or VAEs. These autoencoders follow the structure above, but additionally employ a loss term to shape the latent space of \(\mathbf{c}\). Its goal is to let the latent space follow a known distribution. This makes it possible to draw samples in latent space without workarounds such as having to project samples into the latent space.

Typically we use a normal distribution as target, which makes the latent space an \(m\) dimensional unit cube: each dimension should have a zero mean and unit standard deviation.
This approach is especially useful if the decoder should be used as a generative model. E.g., we can then produce \(\mathbf{c}\) samples directly, and decode them to obtain full states.
While this is very useful for applications such as constructing generative models for faces or other types of natural images, it is less crucial in a simulation setting. Here we want to obtain a latent space that facilitates the temporal prediction, rather than being able to easily produce samples from it.

\section{Time series}
\label{\detokenize{others-timeseries:time-series}}
The goal of the temporal prediction is to compute a latent space state at time \(t+1\) given one or more previous
latent space states.
The most straight\sphinxhyphen{}forward way to formulate the corresponding minimization problem is
\begin{equation*}
\begin{split}
\text{arg min}_{\theta_p} | f_p( \mathbf{c}_{t};\theta_p) - \mathbf{c}_{t+1} |_2^2
\end{split}
\end{equation*}
where the prediction network is denoted by \(f_p\) to distinguish it from encoder and decoder, above.
This already implies that we’re facing a recurrent task: any \(ith\) step is
the result of \(i\) evaluations of \(f_p\), i.e. \(\mathbf{c}_{t+i} = f_p^{(i)}( \mathbf{c}_{t};\theta_p)\).
As there is an inherent per\sphinxhyphen{}evaluation error, it is typically important to train this process
for more than a single step, such that the \(f_p\) network “sees” the drift it produces in terms
of the latent space states over time.

\begin{sphinxadmonition}{note}{Koopman operators}

In classical dynamical systems literature, a data\sphinxhyphen{}driven prediction of future states
is typically formulated in terms of the so\sphinxhyphen{}called \sphinxstyleemphasis{Koopman operator}, which usually takes
the form of a matrix, i.e. uses a linear approach.

Traditional works have focused on obtaining good \sphinxstyleemphasis{Koopman operators} that yield
a high accuracy in combination with a basis to span the space of solutions. In the approach
outlined above the \(f_p\) network can be seen as a non\sphinxhyphen{}linear Koopman operator.
\end{sphinxadmonition}

In order for this approach to work, we either need an appropriate history of previous
states to uniquely identify the right next state, or our network has to internally
store the previous history of states it has seen.

For the former variant, the prediction network \(f_p\) receives more than
a single \(\mathbf{c}_{t}\). For the latter variant, we can turn to algorithms
from the subfield of \sphinxstyleemphasis{recurrent neural networks} (RNNs). A variety of architectures
have been proposed to encode and store temporal states of a system, the most
popular ones being
\sphinxstyleemphasis{long short\sphinxhyphen{}term memory} (LSTM) networks,
\sphinxstyleemphasis{gated recurrent units} (GRUs), or
lately attention\sphinxhyphen{}based \sphinxstyleemphasis{transformer} networks.
No matter which variant is used, these approaches always work with fully\sphinxhyphen{}connected layers
as the latent space vectors do not exhibit any spatial structure, but typically represent
a seemingly random collection of values.
Due to the fully\sphinxhyphen{}connected layers, the prediction networks quickly grow in terms
of their parameter count, and thus require a relatively small latent\sphinxhyphen{}space dimension \(m\).
Luckily, this is in line with our main goals, as outlined at the top.

\section{End\sphinxhyphen{}to\sphinxhyphen{}end training}
\label{\detokenize{others-timeseries:end-to-end-training}}
In the formulation above we have clearly split the en\sphinxhyphen{} / decoding and the time prediction parts.
However, in practice an \sphinxstyleemphasis{end\sphinxhyphen{}to\sphinxhyphen{}end} training of all networks involved in a certain task
is usually preferable, as the networks can adjust their behavior in accordance with the other
components involved in the task.

For the time prediction, we can formulate the objective in terms of \(\mathbf{s}\), and use en\sphinxhyphen{} and decoder in the
time prediction to compute the loss:
\begin{equation*}
\begin{split}
\text{arg min}_{\theta_e,\theta_p,\theta_d} | f_d( f_p( f_e( \mathbf{s}_{t} ;\theta_e)  ;\theta_p) ;\theta_d) - \mathbf{s}_{t+1} |_2^2
\end{split}
\end{equation*}
Ideally, this step is furthermore unrolled over time to stabilize the evolution over time.
The resulting training will be significantly more expensive, as more weights need to be trained at once,
and a much larger number of intermediate states needs to be processed. However, the increased
cost typically pays off with a reduced overall inference error.

\begin{figure}[htbp]
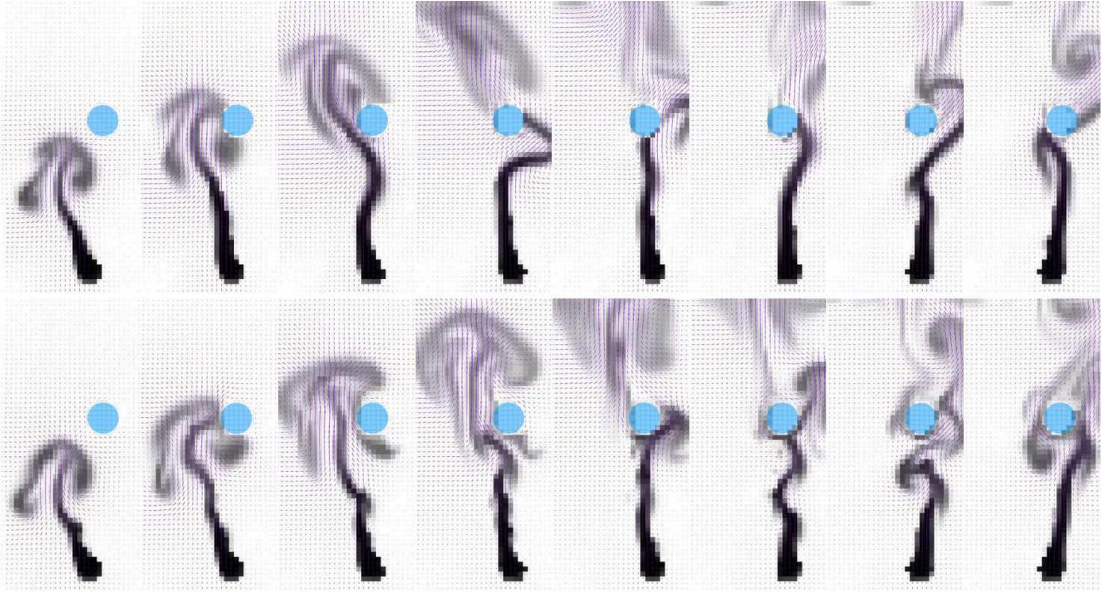

\centering
\capstart

\noindent\sphinxincludegraphics[height=300\sphinxpxdimen]{{others-timeseries-lss-subdiv-prediction}.jpg}
\caption{Several time frames of an example prediction from {[}\hyperlink{cite.references:id8}{WKA+20}{]}, which additionally couples the
learned time evolution with a numerically solved advection step.
The learned prediction is shown at the top, the reference simulation at the bottom.}\label{\detokenize{others-timeseries:timeseries-lss-subdiv-prediction}}\end{figure}

To summarize, DL allows us to move from linear subspaces to non\sphinxhyphen{}linear manifolds, and provides a basis for performing
complex steps (such as time evolutions) in the resulting latent space.

\section{Source code}
\label{\detokenize{others-timeseries:source-code}}
In order to make practical experiments in this area of deep learning, we can
recommend this
\sphinxhref{https://github.com/wiewel/LatentSpaceSubdivision}{latent space simulation code},
which realizes an end\sphinxhyphen{}to\sphinxhyphen{}end training for encoding and prediction.
Alternatively, this
\sphinxhref{https://github.com/byungsook/deep-fluids}{learned model reduction code} focuses on the
encoding and decoding aspects.

Both are available as open source and use a combination of TensorFlow and mantaflow
as DL and fluid simulation frameworks.

\chapter{Generative Adversarial Networks}
\label{\detokenize{others-GANs:generative-adversarial-networks}}\label{\detokenize{others-GANs::doc}}
A fundamental problem in machine learning is to fully represent
all possible states of a variable \(\mathbf{x}\) under consideration,
i.e. to capture its full distribution.
For this task, \sphinxstyleemphasis{generative adversarial networks} (GANs) were
shown to be powerful tools in DL. They are important when the data has ambiguous solutions,
and no differentiable physics model is available to disambiguate the data. In such a case
a supervised learning would yield an undesirable averaging that can be prevented with
a GAN approach.

\begin{figure}[htbp]
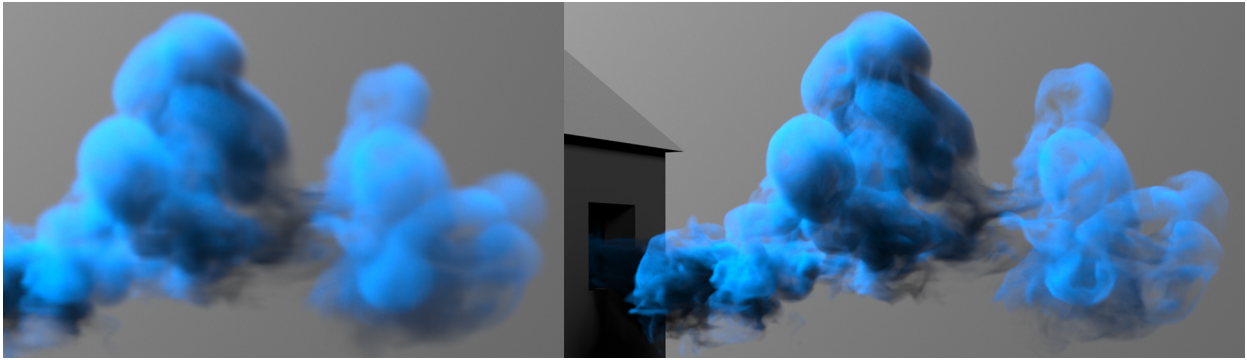

\centering
\capstart

\noindent\sphinxincludegraphics{{others-GANs-tempoGAN}.jpg}
\caption{GANs were shown to work well for tasks such
as the inference of super\sphinxhyphen{}resolution solutions where the range of possible
results can be highly ambiguous.}\label{\detokenize{others-GANs:others-gans-tempogan}}\end{figure}

\section{Maximum likelihood estimation}
\label{\detokenize{others-GANs:maximum-likelihood-estimation}}
To train a GAN we have to briefly turn to classification problems.
For these, the learning objective takes a slightly different form than the
regression objective in equation \eqref{equation:overview-equations:learn-l2} of {\hyperref[\detokenize{overview-equations::doc}]{\sphinxcrossref{\DUrole{doc}{Models and Equations}}}}:
We now want to maximize the likelihood of a learned representation
\(f\) that assigns a probability to an input \(\mathbf{x}_i\) given a set of weights \(\theta\).
This yields a maximization problem of the form
\begin{equation}\label{equation:others-GANs:mle-prob}
\begin{split}
\text{arg max}_{\theta} \Pi_i f(\mathbf{x}_i;\theta) ,
\end{split}
\end{equation}
the classic \sphinxstyleemphasis{maximum likelihood estimation} (MLE). In practice, it is typically turned into
a sum of negative log likelihoods to give the learning objective
\begin{equation*}
\begin{split}
\text{arg min}_{\theta} - \sum_i \text{ log} f(\mathbf{x}_i;\theta) .
\end{split}
\end{equation*}
There are quite a few equivalent viewpoints for this fundamental expression:
e.g., it can be seen as minimizing the KL\sphinxhyphen{}divergence between the empirical distribution
as given by our training data set and the learned one. It likewise represents a maximization
of the expectation as defined by the training data, i.e.
\(\mathbb E \text{ log} f(\mathbf{x}_i;\theta)\).
This in turn is the same as the classical cross\sphinxhyphen{}entropy loss for classification problems,
i.e., a classifier with a sigmoid as activation function..
The takeaway message here is that the wide\sphinxhyphen{}spread training via cross entropy
is effectively a maximum likelihood estimation for probabilities over the inputs,
as defined in equation \eqref{equation:others-GANs:mle-prob}.

\section{Adversarial training}
\label{\detokenize{others-GANs:adversarial-training}}
MLE is a crucial component for GANs: here we have a \sphinxstyleemphasis{generator} that is typically
similar to a decoder network, e.g., the second half of an autoencoder from {\hyperref[\detokenize{others-timeseries::doc}]{\sphinxcrossref{\DUrole{doc}{Model Reduction and Time Series}}}}.
For regular GANs, the generator receives a random input vector, denoted with \(\mathbf{z}\),
from which it should produce the desired output.

However, instead of directly training the generator, we employ a second network
that serves as loss for the generator. This second network is called \sphinxstyleemphasis{discriminator},
and it has a classification task: to distinguish the generated samples from “real” ones.
The real ones are typically provided in the form of a training data set, samples of
which will be denoted as \(\mathbf{x}\) below.

For regular GANs training the classification task of the discriminator is typically formulated as
\begin{equation*}
\begin{split}
\text{arg min}_{\theta_d} 
    - \frac{1}{2}\mathbb E \text{ log} D(\mathbf{y}) 
    - \frac{1}{2}\mathbb E \text{ log} (1 - D(G(\mathbf{z}))
\end{split}
\end{equation*}
which, as outlined above, is a standard binary cross\sphinxhyphen{}entropy training for the class of real samples
\(\mathbf{y}\), and the generated ones \(G(\mathbf{z})\). With the formulation above, the discriminator
is trained to maximize the loss via producing an output of 1 for the real samples, and 0 for the generated ones.

The key for the generator loss is to employ the discriminator and produce samples that are classified as
real by the discriminator:
\begin{equation*}
\begin{split}
\text{arg min}_{\theta_g} 
    - \frac{1}{2}\mathbb E \text{ log} D(G(\mathbf{z}))
\end{split}
\end{equation*}
Typically, this training is alternated, performing one step for \(D\) and then one for \(G\).
Thus the \(D\) network is kept constant, and provides a gradient to “steer” \(G\) in the right direction
to produce samples that are indistinguishable from the real ones. As \(D\) is likewise an NN, it is
differentiable by construction, and can provide the necessary gradients.

\section{Regularization}
\label{\detokenize{others-GANs:regularization}}
Due to the coupled, alternating training, GAN training has a reputation of being finicky in practice.
Instead of a single, non\sphinxhyphen{}linear optimization problem, we now have two coupled ones, for which we need
to find a fragile balance. (Otherwise we’ll get the dreaded \sphinxstyleemphasis{mode\sphinxhyphen{}collapse} problem: once one of the two network “collapses” to a trivial solution, the coupled training breaks down.)

To alleviate this problem, regularization is often crucial to achieve a stable training. In the simplest case,
we can add an \(L^1\) regularizer w.r.t. reference data with a small coefficient for the generator \(G\). Along those lines, pre\sphinxhyphen{}training the generator in a supervised fashion can help to start with a stable state for \(G\). (However, then \(D\) usually also needs a certain amount of pre\sphinxhyphen{}training to keep the balance.)

\section{Conditional GANs}
\label{\detokenize{others-GANs:conditional-gans}}
For physical problems the regular GANs which generate solutions from the randomized latent\sphinxhyphen{}space
\(\mathbf{z}\) above are not overly useful. Rather, we often have inputs such as parameters, boundary conditions or partial solutions which should be used to infer an output. Such cases represent \sphinxstyleemphasis{conditional} GANs,
which means that instead of \(G(\mathbf{z})\), we now have \(G(\mathbf{x})\), where \(\mathbf{x}\) denotes the input data.

A good scenario for conditional GANs are super\sphinxhyphen{}resolution networks: These have the task to compute a high\sphinxhyphen{}resolution output given a sparse or low\sphinxhyphen{}resolution input solution.

\bigskip\hrule\bigskip

\section{Ambiguous solutions}
\label{\detokenize{others-GANs:ambiguous-solutions}}
One of the main advantages of GANs is that they can prevent an undesirable
averaging for ambiguous data. E.g., consider the case of super\sphinxhyphen{}resolution: a
low\sphinxhyphen{}resolution observation that serves as input typically has an infinite number
of possible high\sphinxhyphen{}resolution solutions that would fit the low\sphinxhyphen{}res input.

If a data set contains multiple such cases, and we employ supervised training,
the network will reliably learn the mean. This averaged solution usually is one
that is clearly undesirable, and unlike any of the individual solutions from which it was
computed. This is the \sphinxstyleemphasis{multi\sphinxhyphen{}modality} problem, i.e. different modes existing as valid
equally valid solutions to a problem. For fluids, this can, e.g., happen when
we’re facing bifurcations, as discussed in {\hyperref[\detokenize{intro-teaser::doc}]{\sphinxcrossref{\DUrole{doc}{A Teaser Example}}}}.

The following image shows a clear example of how well GANs can circumvent
this problem:

\begin{figure}[htbp]
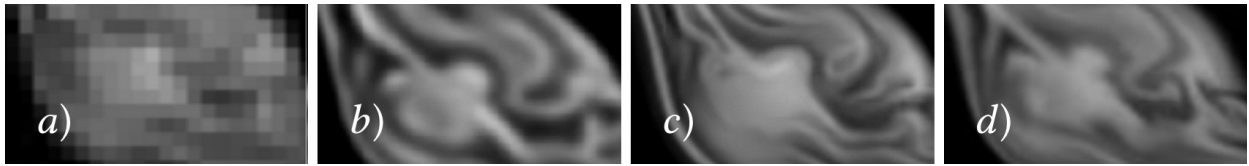

\centering
\capstart

\noindent\sphinxincludegraphics{{others-GANs-tempoGAN-fig3}.jpg}
\caption{A super\sphinxhyphen{}resolution example: a) input, b) supervised result, c) GAN result, d) high\sphinxhyphen{}resolution reference.}\label{\detokenize{others-GANs:gans-tempogan-fig3}}\end{figure}

\section{Spatio\sphinxhyphen{}temporal super\sphinxhyphen{}resolution}
\label{\detokenize{others-GANs:spatio-temporal-super-resolution}}
Naturally, the GAN approach is not limited to spatial resolutions. Previous work
has demonstrated that the concept of learned self\sphinxhyphen{}supervision extends to space\sphinxhyphen{}time
solutions, e.g., in the context of super\sphinxhyphen{}resolution for fluid simulations {[}\hyperlink{cite.references:id20}{XFCT18}{]}.

The following example compares the time derivatives of different solutions:

\begin{figure}[htbp]
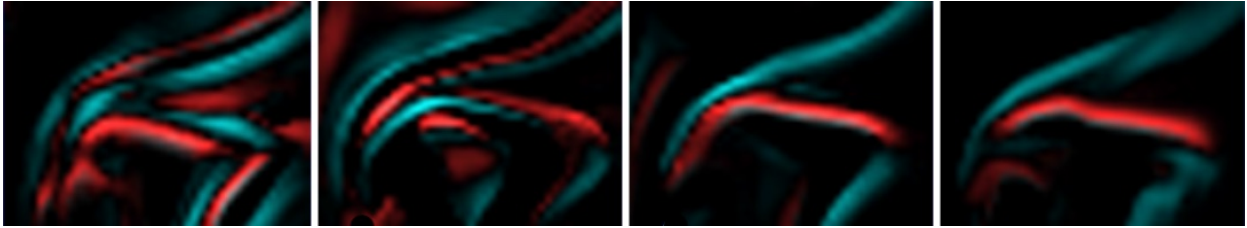

\centering
\capstart

\noindent\sphinxincludegraphics{{others-GANs-tempoGAN-fig4}.jpg}
\caption{From left to right, time derivatives for: a spatial GAN (i.e. not time aware), a temporally supervised learning, a spatio\sphinxhyphen{}temporal GAN, and a reference solution.}\label{\detokenize{others-GANs:gans-tempogan-fig4}}\end{figure}

As can be seen, the GAN trained with spatio\sphinxhyphen{}temporal self\sphinxhyphen{}supervision (second from right) closely matches the reference solution on the far right. In this case the discriminator receives reference solutions over time (in the form of triplets), such that it can learn to judge whether the temporal evolution of a generated solution matches that of the reference.

\section{Physical generative models}
\label{\detokenize{others-GANs:physical-generative-models}}
As a last example, GANs were also shown to be able to
accurately capture solution manifolds of PDEs parametrized by physical parameters {[}\hyperlink{cite.references:id4}{CTS+21}{]}.
In this work, Navier\sphinxhyphen{}Stokes solutions parametrized by varying buoyancies, vorticity content, boundary conditions,
and obstacle geometries were learned by an NN.

This is a highly challenging solution manifold, and requires an extended “cyclic” GAN approach
that pushes the discriminator to take all the physical parameters under consideration into account.
Interestingly, the generator learns to produce realistic and accurate solutions despite
being trained purely on data, i.e. without explicit help in the form of a differentiable physics solver setup.

\begin{figure}[htbp]
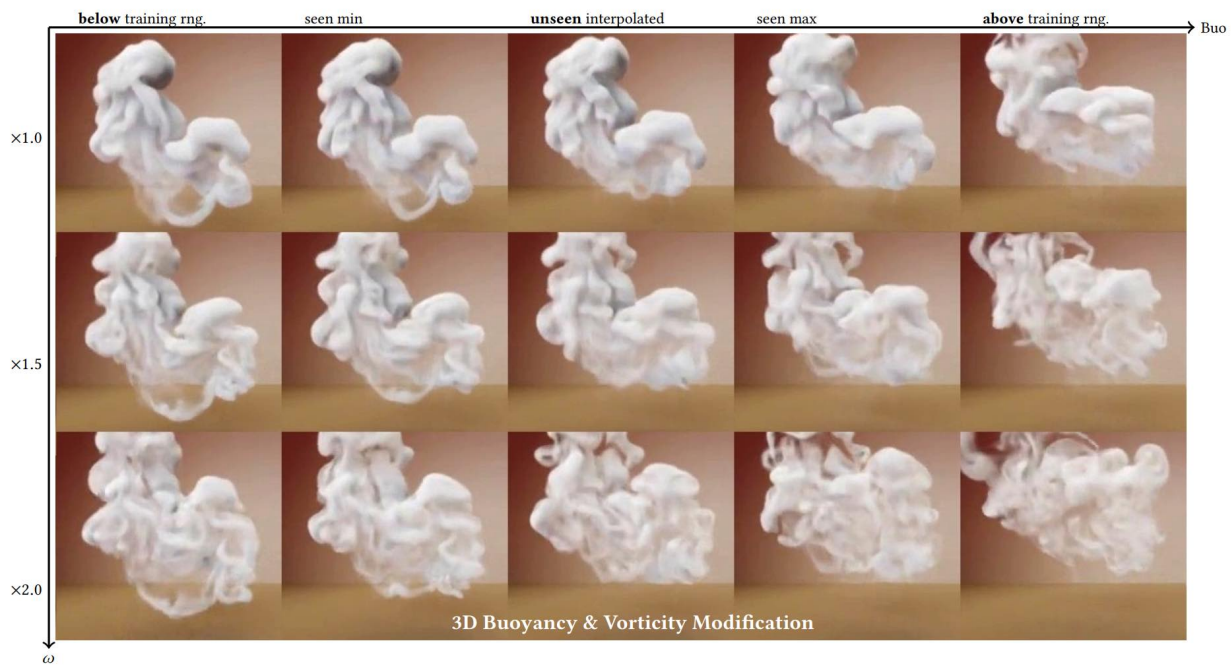

\centering
\capstart

\noindent\sphinxincludegraphics{{others-GANs-meaningful-fig11}.jpg}
\caption{A range of example outputs of a physically\sphinxhyphen{}parametrized GAN {[}\hyperlink{cite.references:id4}{CTS+21}{]}.
The network can successfully extrapolate to buoyancy settings beyond the
range of values seen at training time.}\label{\detokenize{others-GANs:others-gans-meaningful-fig11}}\end{figure}

\bigskip\hrule\bigskip

\section{Discussion}
\label{\detokenize{others-GANs:discussion}}
GANs are a powerful learning tool. Note that the discriminator \(D\) is really “just” a learned
loss function: we can completely discard it at inference time, once the generator is fully trained.
Hence it’s also not overly crucial how much resources it needs.

However, despite being very powerful tools, it is (given the current state\sphinxhyphen{}of\sphinxhyphen{}the\sphinxhyphen{}art) questionable
whether GANs make sense when we have access to a reasonable PDE model. If we can discretize the model
equations and include them with a differentiable physics (DP) training (cf. {\hyperref[\detokenize{diffphys::doc}]{\sphinxcrossref{\DUrole{doc}{Introduction to Differentiable Physics}}}}),
this will most likely give
better results than trying to approximate the PDE model with a discriminator.
The DP training can yield similar benefits to GAN training: it yields a local gradient via the
discretized simulator, and in this way prevents undesirable averaging across samples.
Hence, combinations of DP training and GANs are also bound to not perform better than either of
them in isolation.

That being said, GANs can nonetheless be attractive in situations where DP training is
infeasible due to black\sphinxhyphen{}box solvers without gradients

\section{Source code}
\label{\detokenize{others-GANs:source-code}}
Due to the complexity of the training setup, we only refer to external open source
implementations for practical experiments with physical GANs. E.g.,
the spatio\sphinxhyphen{}temporal GAN from {[}\hyperlink{cite.references:id20}{XFCT18}{]} is available at
\sphinxurl{https://github.com/thunil/tempoGAN}.

It also includes several extensions for stabilization, such as
\(L^1\) regularization, and generator\sphinxhyphen{}discriminator balancing.

\chapter{Unstructured Meshes and Meshless Methods}
\label{\detokenize{others-lagrangian:unstructured-meshes-and-meshless-methods}}\label{\detokenize{others-lagrangian::doc}}
For all computer\sphinxhyphen{}based methods we need to find a suitable \sphinxstyleemphasis{discrete} representation.
While this is straight\sphinxhyphen{}forward for cases such as data consisting only of integers, it is more challenging
for continuously changing quantities such as the temperature in a room.
While the previous examples have focused on aspects beyond discretization
(and used Cartesian grids as a placeholder), the following chapter will target
scenarios where learning with dynamically changing and adaptive discretization has a benefit.

\section{Types of computational meshes}
\label{\detokenize{others-lagrangian:types-of-computational-meshes}}
Generally speaking, we can distinguish three types of computational meshes (or “grids”)
with which discretizations are typically performed:
\begin{itemize}
\item {} 
\sphinxstylestrong{structured} meshes: Structured meshes have a regular
arrangement of the sample points, and an implicitly defined connectivity.
In the simplest case it’s a dense Cartesian grid.

\item {} 
\sphinxstylestrong{unstructured} meshes: On the other hand can have an arbitrary connectivity and arrangement. The flexibility gained from this typically also leads to an increased computational cost.

\item {} 
\sphinxstylestrong{meshless} or particle\sphinxhyphen{}based finally share arbitrary arrangements of the sample points with unstructured meshes, but in contrast implicitly define connectivity via neighborhoods, i.e. a suitable distance metric.

\end{itemize}

Structured meshes are currently very well supported within DL algorithms due to their
similarity to image data, and hence they typically simplify implementations and allow
for using stable, established DL components (especially regular convolutional layers).
However, for target functions that exhibit an uneven mix of smooth and complex
regions, the other two mesh types can have advantages.

\begin{figure}[htbp]
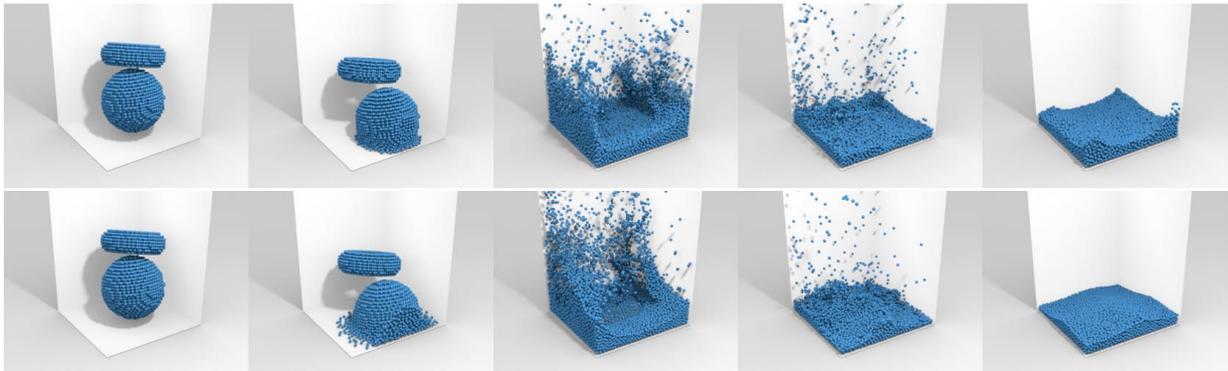

\centering
\capstart

\noindent\sphinxincludegraphics{{others-lagrangian-cconv-dfsph}.jpg}
\caption{Lagrangian simulations of liquids: the sampling points move with the material, and undergo large changes. In the
top row timesteps of a learned simulation, in the bottom row the traditional SPH solver.}\label{\detokenize{others-lagrangian:others-lagrangian-cconv-dfsph}}\end{figure}

\section{Unstructured meshes and graph neural networks}
\label{\detokenize{others-lagrangian:unstructured-meshes-and-graph-neural-networks}}
Within computational sciences the generation of improved mesh structures
is a challenging and ongoing effort. The numerous H\sphinxhyphen{},C\sphinxhyphen{} and O\sphinxhyphen{}type meshes
which were proposed with numerous variations over the years for flows around
airfoils are a good example.

Unstructured meshes offer the largest flexibility here on the meshing side,
but of course need to be supported by the simulator. Interestingly,
unstructured meshes share many properties with \sphinxstyleemphasis{graph} neural networks (GNNs),
which extend the classic ideas of DL on Cartesian grids to graph structures.
Despite growing support, working with GNNs typically causes a fair
amount of additional complexity in an implementation, and the arbitrary
connectivities call for \sphinxstyleemphasis{message\sphinxhyphen{}passing} approaches between the nodes of a graph.
This message passing is usually realized using fully\sphinxhyphen{}connected layers, instead of convolutions.

Thus, in the following, we will focus on a particle\sphinxhyphen{}based method {[}\hyperlink{cite.references:id14}{UPTK19}{]}, which offers
the same flexibility in terms of spatial adaptivity as GNNs. These were previously employed for
a very similar goal {[}\hyperlink{cite.references:id88}{SGGP+20}{]}, however, the method below
enables a real convolution operator for learning the physical relationships.

\section{Meshless and particle\sphinxhyphen{}based methods}
\label{\detokenize{others-lagrangian:meshless-and-particle-based-methods}}
Organizing connectivity explicitly is particularly challenging in dynamic cases,
e.g., for Lagrangian representations of moving materials where the
connectivity quickly becomes obsolete over time.
In such situations, methods that rely on flexible, re\sphinxhyphen{}computed connectivities
are a good choice. Operations are then defined in terms of a spatial
neighborhood around the sampling locations (also called “particles” or just “points”),
and due to the lack of an explicit mesh\sphinxhyphen{}structure these methods are also known as “meshless” methods.
Arguably, different unstructured, graph and meshless variants can typically be translated
from one to the other, but nonetheless the rough distinction outlined above
gives an indicator for how a method works.

In the following, we will discuss an example targeting splashing liquids as a particularly challenging case.
For these simulations, the fluid material moves significantly and is often distributed very non\sphinxhyphen{}uniformly.

The general outline of a learned, particle\sphinxhyphen{}based simulation is similar to a
DL method working on a Cartesian grid: we store data such as the velocity
at certain locations, and then repeatedly perform convolutions to create
a latent space at each location. Each convolution reads in the latent space content
within its support and produces a result, which is activated with a suitable
non\sphinxhyphen{}linear function such as ReLU. This is done multiple times in parallel to produce a latent space
vector, and the resulting latent space vectors at each location serve as inputs
for the next stage of convolutions. After expanding
the size of the latent space over the course of a few layers, it is contracted again
to produce the desired result, e.g., an acceleration.

\section{Continuous convolutions}
\label{\detokenize{others-lagrangian:continuous-convolutions}}
A generic, discrete convolution operator to compute the convolution \((f*g)\) between
functions \(f\) and \(g\) has the form
\begin{equation*}
\begin{split}
(f*g)(\mathbf{x}) = \sum_{\mathbf{\tau} \in \Omega} f(\mathbf{x} + \mathbf{\tau}) g(\mathbf{\tau}),
\end{split}
\end{equation*}
where \(\tau\) denotes the offset vector, and \(\Omega\) defines the support of the filter function (typically \(g\)).

We transfer this idea to particles and point clouds by evaluating a convolution on a set of \(i\) locations \(\mathbf{x}_i\) in a radial neighborhood \(\mathcal N(\mathbf{x}, R)\) around \(\mathbf{x}\). Here, \(R\) denotes the radius within which the convolution should have support.
We define a continuous version of the convolution following {[}\hyperlink{cite.references:id14}{UPTK19}{]}:
\begin{equation*}
\begin{split}
(f*g)(\mathbf{x}) = \sum_{i} f(\mathbf{x}_i) \; g(\Lambda(\mathbf{x}_i - \mathbf{x})).
\end{split}
\end{equation*}
Here, the mapping \(\Lambda\) plays a central role: it represents
a mapping from the unit ball to the unit cube, which allows us to use a simple grid
to represent the unknowns in the convolution kernel. This greatly simplifies
the construction and handling of the convolution kernel, and is illustrated in the following figure:

\begin{figure}[htbp]
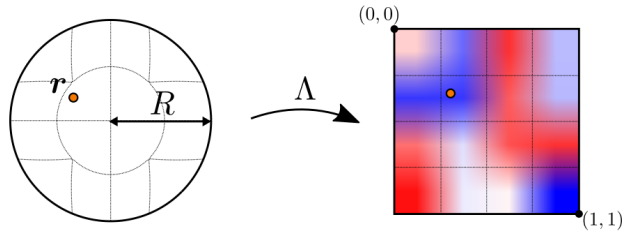

\centering
\capstart

\noindent\sphinxincludegraphics[height=120\sphinxpxdimen]{{others-lagrangian-kernel}.png}
\caption{The unit ball to unit cube mapping employed for the kernel function of the continuous convolution.}\label{\detokenize{others-lagrangian:others-lagrangian-kernel}}\end{figure}

In a physical setting, e.g., the simulation of fluid dynamics, we can additionally introduce a radial
weighting function, denoted as \(a\) below to make sure the kernel has a smooth falloff. This yields
\begin{equation*}
\begin{split}
(f*g)(\mathbf{x}) = \frac{1}{ a_{\mathcal N} } \sum_{i} a(\mathbf{x}_i, \mathbf{x})\; f(\mathbf{x}_i) \; g(\Lambda(\mathbf{x}_i - \mathbf{x})), 
\end{split}
\end{equation*}
where \(a_{\mathcal N}\) denotes a normalization factor
\(a_{\mathcal N} = \sum_{i \in \mathcal N(\mathbf{x}, R)} a(\mathbf{x}_i, \mathbf{x})\).
There’s is quite some flexibility for \(a\), but below we’ll use the following weighting function
\begin{equation*}
\begin{split}
        a(\mathbf{x}_i,\mathbf{x}) = 
        \begin{cases} 
                \left(1 - \frac{\Vert \mathbf{x}_i-\mathbf{x} \Vert_2^2}{R^2}\right)^3  & \text{for } \Vert \mathbf{x}_i-\mathbf{x} \Vert_2 < R\\
                0 & \text{else}.
        \end{cases}
\end{split}
\end{equation*}
This ensures that the learned influence smoothly drops to zero for each of the individual convolutions.

For a lean architecture, a small fully\sphinxhyphen{}connected layer can be added for each convolution to process
the content of the destination particle itself. This makes it possible to use relatively small
kernels with even sizes, e.g., sizes of \(4^3\) {[}\hyperlink{cite.references:id14}{UPTK19}{]}.

\section{Learning the dynamics of liquids}
\label{\detokenize{others-lagrangian:learning-the-dynamics-of-liquids}}
The architecture outlined above can then be trained with a
collection of randomized reference data from a particle\sphinxhyphen{}based Navier\sphinxhyphen{}Stokes solver.
The resulting network yields a good accuracy with a very small and efficient model. E.g.,
compared to GNN\sphinxhyphen{}based approaches the continuous convolution requires significantly fewer
weights and is faster to evaluate.

Interestingly, a particularly tough case for such a learned
solver is a container of liquid that should come to rest. If the training data is not specifically
engineered to contain many such cases, the network receives only a relatively small
of such cases at training time. Moreover, a simulation typically takes many steps to come
to rest (many more than are unrolled for training). Hence the network is not explicitly trained
to reproduce such behavior.

Nonetheless, an interesting side\sphinxhyphen{}effect of having a trained NN for such a liquid simulation
by construction provides a differentiable solver. Based on a pre\sphinxhyphen{}trained network, the learned solver
then supports optimization via gradient descent, e.g., w.r.t. input parameters such as viscosity.

\begin{figure}[htbp]
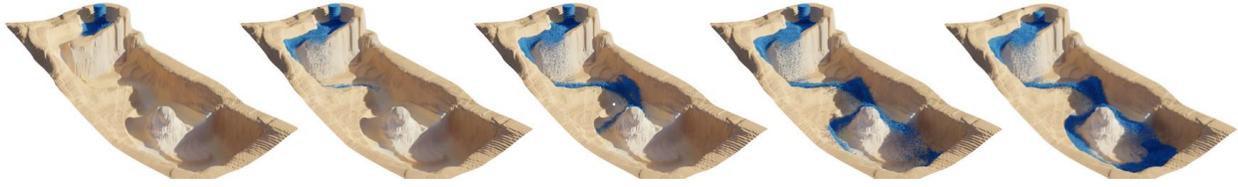

\centering
\capstart

\noindent\sphinxincludegraphics{{others-lagrangian-canyon}.jpg}
\caption{An example of a particle\sphinxhyphen{}based liquid spreading in a landscape scenario, simulated with
learned approach using continuous convolutions {[}\hyperlink{cite.references:id14}{UPTK19}{]}.}\label{\detokenize{others-lagrangian:others-lagrangian-canyon}}\end{figure}

\section{Source code}
\label{\detokenize{others-lagrangian:source-code}}
For a practical implementation of the continuous convolutions, another important step
is a fast collection of neighboring particles for \(\mathcal N\). An efficient example implementation
can be found at
\sphinxurl{https://github.com/intel-isl/DeepLagrangianFluids},
together with training code for learning the dynamics of liquids, an example of which is
shown in the figure above.

\part{End Matter}

\chapter{Outlook}
\label{\detokenize{outlook:outlook}}\label{\detokenize{outlook::doc}}
Despite the lengthy discussions and numerous examples, we’ve really just barely scratched the surface regarding the possibilities that arise in the context of physics\sphinxhyphen{}based deep learning.

Most importantly, the techniques that were explained in the previous chapter have an enormous potential to influence all computational methods of the next decades. As demonstrated many times in the code examples, there’s no magic involved, but deep learning gives us very powerful tools to represent and approximate non\sphinxhyphen{}linear functions. And deep learning by no means makes existing numerical methods deprecated. Rather, the two are an ideal combination.

A topic that we have not touched at all so far is, that – of course – in the end our goal is to improve human understanding of our world. And here the view of neural networks as “black boxes” is clearly outdated. It is simply another numerical method that humans can employ, and the physical fields predicted by a network are as interpretable as the outcome of a traditional simulation. Nonetheless, it is important to further improve the tools for analyzing learned networks, and to extract condensed formulations of the patterns and regularities the networks have found in the solution manifolds.

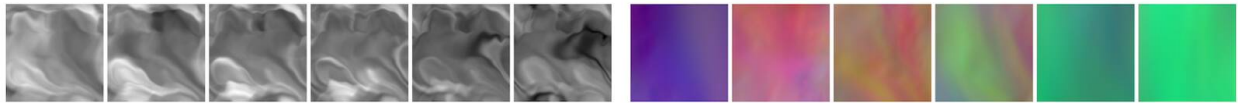
\sphinxincludegraphics{{divider2}.jpg}

\section{Some specific directions}
\label{\detokenize{outlook:some-specific-directions}}
Beyond this long term outlook, there are many interesting and immediate steps.
And while the examples with Burgers equation and Navier\sphinxhyphen{}Stokes solvers are clearly non\sphinxhyphen{}trivial, there’s a wide variety of other potential PDE models that the techniques of this book can be applied to. To name just a few promising examples from other fields:
\begin{itemize}
\item {} 
PDEs for chemical reactions often show complex behavior due to the interactions of multiple species. Here, and especially interesting direction is to train models that quickly learn to predict the evolution of an experiment or machine, and adjust control knobs to stabilize it, i.e., an online \sphinxstyleemphasis{control} setting.

\item {} 
Plasma simulations share a lot with vorticity\sphinxhyphen{}based formulations for fluids, but additionally introduce terms to handle electric and magnetic interactions within the material. Likewise, controllers for plasma fusion experiments and generators are an excellent topic with plenty of potential for DL with differentiable physics.

\item {} 
Finally, weather and climate are crucial topics for humanity, and highly complex systems of fluid flows interacting with a multitude of phenomena on the surface of our planet. Accurately modeling all these interacting systems and predicting their long\sphinxhyphen{}term behavior shows a lot of promise to benefit from DL approaches that can interface with numerical simulations.

\end{itemize}

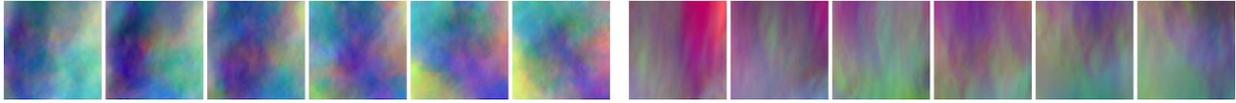
\sphinxincludegraphics{{divider3}.jpg}

\section{Closing remarks}
\label{\detokenize{outlook:closing-remarks}}
So overall, there’s lots of exciting research work left to do \sphinxhyphen{} the next years and decades definitely won’t be boring. 👍

\begin{figure}[htbp]
\centering

\noindent\sphinxincludegraphics[height=200\sphinxpxdimen]{{logo}.jpg}
\end{figure}

\chapter{References}
\label{\detokenize{references:references}}\label{\detokenize{references::doc}}

\chapter{Notation and Abbreviations}
\label{\detokenize{notation:notation-and-abbreviations}}\label{\detokenize{notation::doc}}

\section{Math notation:}
\label{\detokenize{notation:math-notation}}

\begin{savenotes}\sphinxattablestart
\centering
\begin{tabulary}{\linewidth}[t]{|T|T|}
\hline
\sphinxstyletheadfamily 
Symbol
&\sphinxstyletheadfamily 
Meaning
\\
\hline
\(A\)
&
matrix
\\
\hline
\(\eta\)
&
learning rate or step size
\\
\hline
\(\Gamma\)
&
boundary of computational domain \(\Omega\)
\\
\hline
\(f^{*}\)
&
generic function to be approximated, typically  unknown
\\
\hline
\(f\)
&
approximate version of \(f^{*}\)
\\
\hline
\(\Omega\)
&
computational domain
\\
\hline
\(\mathcal P^*\)
&
continuous/ideal physical model
\\
\hline
\(\mathcal P\)
&
discretized physical model, PDE
\\
\hline
\(\theta\)
&
neural network params
\\
\hline
\(t\)
&
time dimension
\\
\hline
\(\mathbf{u}\)
&
vector\sphinxhyphen{}valued velocity
\\
\hline
\(x\)
&
neural network input or spatial coordinate
\\
\hline
\(y\)
&
neural network output
\\
\hline
\end{tabulary}
\par
\sphinxattableend\end{savenotes}

\section{Summary of the most important abbreviations:}
\label{\detokenize{notation:summary-of-the-most-important-abbreviations}}

\begin{savenotes}\sphinxattablestart
\centering
\begin{tabulary}{\linewidth}[t]{|T|T|}
\hline
\sphinxstyletheadfamily 
ABbreviation
&\sphinxstyletheadfamily 
Meaning
\\
\hline
BNN
&
Bayesian neural network
\\
\hline
CNN
&
Convolutional neural network
\\
\hline
DL
&
Deep Learning
\\
\hline
GD
&
(steepest) Gradient Descent
\\
\hline
MLP
&
Multi\sphinxhyphen{}Layer Perceptron, a neural network with fully connected layers
\\
\hline
NN
&
Neural network (a generic one, in contrast to, e.g., a CNN or MLP)
\\
\hline
PDE
&
Partial Differential Equation
\\
\hline
PBDL
&
Physics\sphinxhyphen{}Based Deep Learning
\\
\hline
SGD
&
Stochastic Gradient Descent
\\
\hline
\end{tabulary}
\par
\sphinxattableend\end{savenotes}

\begin{sphinxthebibliography}{OMalleyB}
\bibitem[AAC+19]{references:id95}
Ilge Akkaya, Marcin Andrychowicz, Maciek Chociej, Mateusz Litwin, Bob McGrew, and others. Solving rubik's cube with a robot hand. \sphinxstyleemphasis{arXiv:1910.07113}, 2019.
\bibitem[CT21]{references:id2}
Li\sphinxhyphen{}Wei Chen and Nils Thuerey. Towards high\sphinxhyphen{}accuracy deep learning inference of compressible turbulent flows over aerofoils. In \sphinxstyleemphasis{arXiv}. 2021. URL: \sphinxurl{https://ge.in.tum.de/publications/}.
\bibitem[CTS+21]{references:id4}
Mengyu Chu, Nils Thuerey, Hans\sphinxhyphen{}Peter Seidel, Christian Theobalt, and Rhaleb Zayer. Learning Meaningful Controls for Fluids. \sphinxstyleemphasis{ACM Trans. Graph.}, 2021. URL: \sphinxurl{https://people.mpi-inf.mpg.de/~mchu/gvv-den2vel/den2vel.html}.
\bibitem[Gol90]{references:id79}
H Goldstine. \sphinxstyleemphasis{A history of scientific computing}. ACM, 1990.
\bibitem[HKT19]{references:id11}
Philipp Holl, Vladlen Koltun, and Nils Thuerey. Learning to control pdes with differentiable physics. In \sphinxstyleemphasis{International Conference on Learning Representations}. 2019. URL: \sphinxurl{https://ge.in.tum.de/publications/2020-iclr-holl/}.
\bibitem[KAT+19]{references:id19}
Byungsoo Kim, Vinicius C Azevedo, Nils Thuerey, Theodore Kim, Markus Gross, and Barbara Solenthaler. Deep Fluids: A Generative Network for Parameterized Fluid Simulations. \sphinxstyleemphasis{Comp. Grap. Forum}, 38(2):12, 2019. URL: \sphinxurl{http://www.byungsoo.me/project/deep-fluids/}.
\bibitem[KB14]{references:id89}
Diederik P Kingma and Jimmy Ba. Adam: a method for stochastic optimization. \sphinxstyleemphasis{arXiv:1412.6980}, 2014.
\bibitem[KSA+21]{references:id94}
Dmitrii Kochkov, Jamie A Smith, Ayya Alieva, Qing Wang, Michael P Brenner, and Stephan Hoyer. Machine learning–accelerated computational fluid dynamics. \sphinxstyleemphasis{Proceedings of the National Academy of Sciences}, 2021.
\bibitem[KUT20]{references:id7}
Georg Kohl, Kiwon Um, and Nils Thuerey. Learning similarity metrics for numerical simulations. \sphinxstyleemphasis{International Conference on Machine Learning}, 2020. URL: \sphinxurl{https://ge.in.tum.de/publications/2020-lsim-kohl/}.
\bibitem[KSH12]{references:id85}
Alex Krizhevsky, Ilya Sutskever, and Geoffrey E Hinton. Imagenet classification with deep convolutional neural networks. In \sphinxstyleemphasis{Advances in Neural Information Processing Systems}. 2012.
\bibitem[MLA+19]{references:id83}
Rajesh Maingi, Arnold Lumsdaine, Jean Paul Allain, Luis Chacon, SA Gourlay, and others. Summary of the fesac transformative enabling capabilities panel report. \sphinxstyleemphasis{Fusion Science and Technology}, 75(3):167–177, 2019.
\bibitem[OMalleyBK+16]{references:id84}
Peter JJ O’Malley, Ryan Babbush, Ian D Kivlichan, Jonathan Romero, Jarrod R McClean, Rami Barends, Julian Kelly, Pedram Roushan, Andrew Tranter, Nan Ding, and others. Scalable quantum simulation of molecular energies. \sphinxstyleemphasis{Physical Review X}, 6(3):031007, 2016.
\bibitem[Qur19]{references:id87}
Mohammed Al Quraishi. Alphafold at casp13. \sphinxstyleemphasis{Bioinformatics}, 35(22):4862–4865, 2019.
\bibitem[RWC+19]{references:id86}
Alec Radford, Jeffrey Wu, Rewon Child, David Luan, Dario Amodei, and Ilya Sutskever. Language models are unsupervised multitask learners. \sphinxstyleemphasis{OpenAI blog}, 1(8):9, 2019.
\bibitem[RK18]{references:id81}
Maziar Raissi and George Em Karniadakis. Hidden physics models: machine learning of nonlinear partial differential equations. \sphinxstyleemphasis{Journal of Computational Physics}, 357:125–141, 2018.
\bibitem[SGGP+20]{references:id88}
Alvaro Sanchez\sphinxhyphen{}Gonzalez, Jonathan Godwin, Tobias Pfaff, Rex Ying, Jure Leskovec, and Peter Battaglia. Learning to simulate complex physics with graph networks. In \sphinxstyleemphasis{International Conference on Machine Learning}, 8459–8468. 2020.
\bibitem[SML+15]{references:id93}
John Schulman, Philipp Moritz, Sergey Levine, Michael Jordan, and Pieter Abbeel. High\sphinxhyphen{}dimensional continuous control using generalized advantage estimation. \sphinxstyleemphasis{arXiv:1506.02438}, 2015.
\bibitem[SWD+17]{references:id92}
John Schulman, Filip Wolski, Prafulla Dhariwal, Alec Radford, and Oleg Klimov. Proximal policy optimization algorithms. \sphinxstyleemphasis{arXiv:1707.06347}, 2017.
\bibitem[SSS+17]{references:id90}
David Silver, Julian Schrittwieser, Karen Simonyan, Ioannis Antonoglou, Aja Huang, and others. Mastering the game of Go without human knowledge. \sphinxstyleemphasis{Nature}, 2017.
\bibitem[Sto14]{references:id82}
Thomas Stocker. \sphinxstyleemphasis{Climate change 2013: the physical science basis: Working Group I contribution to the Fifth assessment report of the Intergovernmental Panel on Climate Change}. Cambridge university press, 2014.
\bibitem[SB18]{references:id91}
Richard S Sutton and Andrew G Barto. \sphinxstyleemphasis{Reinforcement learning: An introduction}. MIT press, 2018.
\bibitem[TWPH20]{references:id16}
Nils Thuerey, Konstantin Weissenow, Lukas Prantl, and Xiangyu Hu. Deep learning methods for reynolds\sphinxhyphen{}averaged navier–stokes simulations of airfoil flows. \sphinxstyleemphasis{AIAA Journal}, 58(1):25–36, 2020. URL: \sphinxurl{https://ge.in.tum.de/publications/2018-deep-flow-pred/}.
\bibitem[TSSP17]{references:id80}
Jonathan Tompson, Kristofer Schlachter, Pablo Sprechmann, and Ken Perlin. Accelerating eulerian fluid simulation with convolutional networks. In \sphinxstyleemphasis{Proceedings of Machine Learning Research}, 3424–3433. 2017.
\bibitem[UBH+20]{references:id6}
Kiwon Um, Robert Brand, Philipp Holl, Raymond Fei, and Nils Thuerey. Solver\sphinxhyphen{}in\sphinxhyphen{}the\sphinxhyphen{}loop: learning from differentiable physics to interact with iterative pde\sphinxhyphen{}solvers. \sphinxstyleemphasis{Advances in Neural Information Processing Systems}, 2020. URL: \sphinxurl{https://ge.in.tum.de/publications/2020-um-solver-in-the-loop/}.
\bibitem[UPTK19]{references:id14}
Benjamin Ummenhofer, Lukas Prantl, Nils Thuerey, and Vladlen Koltun. Lagrangian fluid simulation with continuous convolutions. In \sphinxstyleemphasis{International Conference on Learning Representations}. 2019. URL: \sphinxurl{https://ge.in.tum.de/publications/2020-ummenhofer-iclr/}.
\bibitem[WBT19]{references:id18}
Steffen Wiewel, Moritz Becher, and Nils Thuerey. Latent\sphinxhyphen{}space Physics: Towards Learning the Temporal Evolution of Fluid Flow. \sphinxstyleemphasis{Comp. Grap. Forum}, 38(2):12, 2019. URL: \sphinxurl{https://ge.in.tum.de/publications/latent-space-physics/}.
\bibitem[WKA+20]{references:id8}
Steffen Wiewel, Byungsoo Kim, Vinicius C Azevedo, Barbara Solenthaler, and Nils Thuerey. Latent space subdivision: stable and controllable time predictions for fluid flow. \sphinxstyleemphasis{Symposium on Computer Animation}, 2020. URL: \sphinxurl{https://ge.in.tum.de/publications/2020-lssubdiv-wiewel/}.
\bibitem[XFCT18]{references:id20}
You Xie, Erik Franz, Mengyu Chu, and Nils Thuerey. tempoGAN: A Temporally Coherent, Volumetric GAN for Super\sphinxhyphen{}resolution Fluid Flow. \sphinxstyleemphasis{ACM Trans. Graph.}, 2018. URL: \sphinxurl{https://ge.in.tum.de/publications/tempogan/}.
\end{sphinxthebibliography}

\renewcommand{\indexname}{Index}
\printindex
\end{document}